\documentclass[
  a4paper,
  12pt,
  headsepline,
  numbers=noenddot,
  captions=oneline,
  captions=tableheading,
  BCOR12mm,
  headinclude,
  chapterprefix,
  appendixprefix,
  index=totoc,
  bibliography=totoc
]{scrbook}

\usepackage{dissertation}
\usepackage{scrhack}
\usepackage{color}

\usepackage{amssymb}

\usepackage{xspace}
\usepackage{xcolor}

\usepackage{amsmath,amssymb}
\usepackage{bm}

\newcommand{\argmin}{\operatornamewithlimits{argmin}}
\newcommand{\ci}{\mathrel{\text{\scalebox{1}{$\perp\mkern-10mu\perp$}}}}

\usepackage{algorithm}
\usepackage{algorithmic}

\usepackage{subfigure}
\usepackage{tikz}
\usetikzlibrary{bayesnet,matrix,calc}
\usepackage{graphicx}
\newcommand{\mycaption}[2]{\caption{\textbf{#1.}\xspace#2}}
\usepackage{pbox}
\usepackage{wrapfig}

\usepackage{booktabs}
\usepackage{array}
\usepackage{multirow}
\usepackage{multicol}
\usepackage{longtable}

\usepackage{natbib}
\usepackage{bibentry}

\setcitestyle{square,numbers,comma}

\usepackage{hyperref}

\degree{}
\name{Varun Jampani}
\thesis{Learning Inference Models for Computer Vision}
\keywordlist{Computer Vision, Machine Learning, Inference, Generative Models,
Discriminative Models, Deep Learning}
\hometown{Tenali, Indien}

\dean{Prof.~Dr.~Wolfgang Rosenstiel}

\reviewerone{Prof.~Dr.~Hendrik P.~A. Lensch}
\reviewertwo{Prof.~Dr.~Peter V. Gehler}
\reviewerthree{Dr.~Iasonas Kokkinos}

\hypersetup{
              bookmarksopen={true},
              bookmarksopenlevel={0},
              bookmarksnumbered={true},
              breaklinks={true},
              colorlinks={false},
              pdfpagemode={UseOutlines},
              pdftitle={\thethesis},
              pdfauthor={\thename},
              pdfsubject={Dissertation},
              pdfkeywords={\thekeywordlist},
              pdfview={FitV},
              pdffitwindow={true},
              pdfstartview={FitV},
              pdfnewwindow={false},
              pdfdisplaydoctitle={true},
              pdfhighlight={/P},
              plainpages={false},
              unicode={true},
              urlcolor={blue}
}
\extratitle{\begin{center}\normalfont\sffamily\thethesis\end{center}}
\title{\thethesis}
\author{\Large Dissertation\\
\normalsize der Mathematisch-Naturwissenschaftlichen Fakult\"at\\
\normalsize der Eberhard Karls Universit\"at T\"ubingen\\
\normalsize zur Erlangung des Grades eines\\
\normalsize Doktors der Naturwissenschaften\\
\normalsize (Dr.~rer.~nat.)}
\date{\ \\[2ex]\normalsize  vorgelegt von\\
\Large \thename\\
\normalsize aus \thehometown}
\publishers{\normalsize T\"ubingen\\\normalsize 2016}
\lowertitleback{
Gedruckt mit Genehmigung der Mathematisch-Naturwissenschaftlichen Fakult\"at der
Eberhard Karls Universit\"at T\"ubingen. \\
\\
\begin{tabular}{lp{.55\textwidth}}
Tag der m\"undlichen Qualifikation:& 02.05.2017\\
Dekan:                             & \thedean\\
1.~Berichterstatter:               & \thereviewerone\\
2.~Berichterstatter:               & \thereviewertwo\\
3.~Berichterstatter:               & \thereviewerthree\\
\end{tabular}}
\dedication{\normalsize To my grandfather \emph{K. Anjaneyulu}} 

\DeclareOldFontCommand{\bf}{\normalfont\bfseries}{\mathbf}

\begin{document}
\frontmatter
\selectlanguage{ngerman}
\maketitle
\selectlanguage{english}

%
%
\chapter{Abstract}
\label{chap:abstract}

Computer vision can be understood as the ability to perform \emph{inference} on
image data. Breakthroughs in computer vision technology are often marked by advances
in inference techniques, as even the model design is often dictated by the
complexity of inference in them. This thesis proposes learning based inference schemes
and demonstrates applications in computer vision. We propose techniques for inference
in both generative and discriminative computer vision models.

Despite their intuitive appeal, the use of generative models in vision is hampered
by the difficulty of posterior inference, which is often too complex or too slow to
be practical. We propose techniques for improving inference in two widely used
techniques: Markov Chain Monte Carlo (MCMC) sampling and message-passing inference.
Our inference strategy is to learn separate discriminative models that assist Bayesian
inference in a generative model. Experiments on a range of generative vision models show
that the proposed techniques accelerate the inference process and/or converge to better solutions.

A main complication in the design of discriminative models is the inclusion of prior
knowledge in a principled way. For better inference in discriminative models, we propose
techniques that modify the original model itself, as inference is simple evaluation of
the model. We concentrate on convolutional neural network (CNN) models and propose a
generalization of standard spatial convolutions, which are the basic building blocks of CNN
architectures, to bilateral convolutions. First, we generalize the existing use of
bilateral filters and then propose new neural network architectures with learnable
bilateral filters, which we call `Bilateral Neural Networks'. We show how the bilateral
filtering modules can be used for modifying existing CNN architectures for better image
segmentation and propose a neural network approach for temporal information propagation
in videos. Experiments demonstrate the potential of the proposed bilateral networks on
a wide range of vision tasks and datasets.

In summary, we propose learning based techniques for better inference in several
computer vision models ranging from inverse graphics to freely
parameterized neural networks. In generative vision models, our inference techniques
alleviate some of the crucial hurdles in Bayesian posterior inference, paving new ways
for the use of model based machine learning in vision. In discriminative CNN models,
the proposed filter generalizations aid in the design of new neural network architectures
that can handle sparse high-dimensional data as well as provide a way for incorporating
prior knowledge into CNNs.
\selectlanguage{ngerman}
\chapter{Zusammenfassung}

Maschinelles Sehen kann als die F\"ahigkeit verstanden werden Bilddaten zu interpretieren. Durchbr\"uche in diesem Feld gehen oft einher mit Fortschritten in Inferenztechniken, da die Komplexit\"at der Inferenz die Komplexit\"at der verwendeten Modelle bestimmt. Diese Arbeit beschreibt lernbasierte Inferenzmechanismen und zeigt Anwendungen im maschinellen Sehen auf, wobei auf Techniken f\"ur Inferenz in sowohl generativen als auch diskriminativen Modellen eingegangen wird.

Obwohl naheliegend und intuitiv verst\"andlich, sind generative Modelle im maschinellen Sehen h\"aufig nur eingeschr\"ankt nutzbar, da die Berechnung der A-Posteriori-Wahrscheinlichkeiten oft zu komplex oder zu langsam ist, um praktikabel zu sein. Wir beschreiben Techniken zur Verbesserung der Inferenz in zwei weit verbreiteten Inferenzverfahren: `Markov Chain Monte Carlo Sampling' (MCMC) und `Message-Passing'. Die vorgeschlagene Verbesserung besteht darin, mehrere diskriminative Modelle zu lernen, die die Grundlage f\"ur Bayes'sche Inferenz \"uber einem generativen Modell bilden. Wir demonstrieren anhand einer Reihe von generativen Modellen, dass die beschriebenen Techniken den Inferenzprozess beschleunigen und/oder zu besseren L\"osungen konvergieren.

Eine der gr\"o{\ss}ten Schwierigkeiten bei der Verwendung von diskriminativen Modellen ist die systematische Ber\"ucksichtigung von Vorkenntnissen. Zur Verbesserung der Inferenz in diskriminativen Modellen schlagen wir Techniken vor die das ursprüngliche Modell selbst verändern, da Inferenz in diesen die schlichte Auswertung des Modells ist. Wir konzentrieren uns auf `Convolutional Neural Networks' (CNN) und schlagen eine Generalisierung der Faltungsoperation vor, die den Kern jeder CNN-Architektur bildet. Dazu verallgemeinern wir bilaterale Filter und pr\"asentieren eine neue Netzarchitektur mit trainierbaren bilateralen Filtern, die wir `Bilaterale Neuronale Netze' nennen. Wir zeigen, wie die bilateralen Filtermodule verwendet werden können, um existierende Netzwerkarchitekturen für Bildsegmentierung zu verbessern und entwickeln ein auf Bilateralen Netzen basierendes Modell zur zeitlichen Integration von Information für Videoanalyse. Experimente mit einer breiten Palette von Anwendungen und Datens\"atzen zeigen das Potenzial der vorgeschlagenen bilateralen Netzwerke.

Zusammenfassend schlagen wir Lernmethoden f\"ur bessere Inferenz in einer Reihe von Modellen des maschinellen Sehens vor, von inversen Renderern bis zu trainierbaren neuronalen Netzwerken. Unsere Inferenz-Techniken helfen bei der Berechnung der A-Posteriori-Wahrscheinlichkeiten in generativen Modellen und erm\"oglichen so neue Ans\"atze des modellbasierten machinellen Lernens im Bereich des maschinellen Sehens. In diskriminativen Modellen wie CNNs helfen die vorgeschlagenen verallgemeinerten Filter beim Entwurf neuer Netzarchitekturen, die sowohl hochdimensionale Daten verarbeiten k\"onnen als auch Vorkenntnisse in die Inferenz einbeziehen.
\selectlanguage{english}
\chapter{Acknowledgements}
\label{chap:ack}

First, my heartfelt thanks to my PhD advisor Prof. Peter Gehler without whom
this thesis work would not have materialized. His suggestions helped me work
on very interesting research problems in computer vision and machine learning
whereas stimulating discussions with him helped me propose compelling techniques
for those problems. He also taught me how to be critical about various aspects of
a research project. I thank Dr. Sebastian Nowozin for being a co-advisor in my
first PhD research project and teaching me how to be rigorous in research.

Thanks to Prof.~Michael Black for allowing me to be a part of his vibrant research
group in T\"ubingen, where I conducted my PhD research. Special thanks to
Prof. Bipin Indurkhya for teaching me how to be mature in our reasoning and thinking
while doing research.~Thanks to my masters advisor Prof.~Jayanthi Sivaswamy for her
guidance during the start of my research career. Thanks to Dr. Pushmeet Kohli,
Dr. Ali Eslami, Dr. John Winn and Dr. Daniel Tarlow for giving me a great internship
opportunity at Microsoft Research that helped me getting a broader perspective over
the fields of computer vision and machine learning.
Thanks to Prof. Jan Kautz for giving me a wonderful opportunity at Nvidia
to further pursue my research interests after my PhD.

I am thankful to Prof. Hendrik Lensch, Prof. Peter Gehler, Prof. Felix Wichmann and
Prof. Matthias Bethge for being a part of my thesis committee and
thanks to Prof. Hendrik Lensch and Prof. Peter Gehler for providing reviews to
this thesis work. Special thanks to Dr. Iasonas Kokkinos for providing an additional
review for this thesis.
Thanks to Thomas Nestmeyer, Dr. Laura Sevilla, Angjoo Kanazawa,
Fatma G\"uney, Jonas Wulff and Christoph Lassner for their helpful feedback
and comments on this thesis manuscript.

I am fortunate to have collaborated with several great researchers during my
PhD: Dr. Martin Kiefel, Raghudeep Gadde, Dr. Sebastian Nowozin, Dr. Ali Eslami,
Dr. John Winn, Dr. Pushmeet Kohli, Dr. Matthew Loper, Prof. Michael Black,
Dr. Laura Sevilla, Dr. Deqing Sun, Daniel Kappler, Dr. Renaud Marlet and Dr. Daniel Tarlow.
I am very grateful to them.
Special thanks to Raghudeep Gadde and Dr. Laura Sevilla for trusting me and
letting me be a part of their research projects. I am highly thankful to Raghudeep Gadde
for the countless hours we spent together on brainstorming various research questions
and techniques.

I want to express my gratitude to my wonderful colleagues at Max-Planck Institute
Thomas Nestmeyer, Raghudeep Gadde, Abhilash Srikantha, Dr. Laura Sevilla, Dr. Silvia Zuffi, Jonas Wulff,
Fatma G\"uney, Dr. Martin Kiefel,
Dr. Andreas Lehrmann, Christoph Lassner, Sergey Prokudin, Dr.~Osman Ulusoy,
Dr. Federica Bogo, Daniel Kappler, Prof. Michael Black,
Dr. Andreas Geiger, Prof. J\"urgen Gall, Dr. Chaohui Wang, Dr. Javier Romero,
Dr. Hueihan Jhuang, Dr. Dimitris Tzionas, Dr. Gerard Pons-Moll,
Dr. Cristina Cifuentes, Dr. Naejin Kong, Naureen Mohamed, Angjoo Kanazawa,
Dr. Ijaz Akhter, Dr. S{\o}ren Hauberg, Dr. Matthew Loper, Dr. Aggeliki Tsoli,
Dr. Sergi Pujades, Siyu Tang, Dr. Si Yong Yeo, Joel Janai, Yiyi Liao and others
for their insightful discussions, technical and non-technical conversations and making my
road to obtaining my PhD less bumpy. Thanks to Melanie Feldhofer, Jon Anning
and Nicole Overbaugh for their great support in administrative matters.

I am highly indebted to my friends Srikar Ravulapalli, Srujan Sunkoju, Srinivas Sunkara,
Vamshi Velagapuri, Chandrasekhar Varkala, Sunil Soni, Rohit Gernapudi, Raghudeep Gadde,
Seetharamakrishna Madamsetti, Tenzin Bety, Tsering Bhuti, Phurbu Thackchoe, Varsha Eluri
and Dharani Kodali for always supporting, helping and encouraging me in the times of need.
Thanks to Dr. Sowmya Vajjala and Dr. Lorenzo Ducci
for the fun times we had in and around T\"ubingen and also patiently listening to my
ravings about both research and life in general.

Finally, I would like to thank my family members who have given me great freedom and support
to pursue my interests. I am very thankful to my parents
Dr. Chandrasekhar Jampani and Sudharani Kodali, my brother Prasanth Jampani, my sister-in-law Neelima Kamani,
my grandmother Krishnakumari Kodali for their great support during all these years.
Special thanks to my late grandfather and my favorite person Anjaneyulu Kodali for
understanding me so well and encouraging me in different aspects of life. He will always
be in my heart. Lastly, but importantly, I am very grateful to my wife
Dr. Yuyu Liu for her love and continued support over the last few years and for sharing
the ups and downs of my life.

\tableofcontents

\chapter{Symbols and Notation}
\label{chap:symbols}

\newcommand{\obs}{\mathbf{x}}
\newcommand{\target}{\mathbf{y}}
\newcommand{\targetProp}{\bar{\mathbf{y}}}
\newcommand{\params}{\mathbf{\theta}}
\newcommand{\f}{\mathbf{f}}
\newcommand{\func}{\mathcal{F}}

\newcommand{\eg}{\textit{e.g.}\@\xspace}
\newcommand{\ie}{\textit{i.e.}\@\xspace}
\newcommand{\ala}{\textit{\'{a}~la}\@\xspace}
\newcommand{\etal}{\textit{et~al.}\@\xspace}

Unless otherwise mentioned, we use the following notation and symbols in this
thesis. Here, we only list those symbols which are used across multiple chapters.
Those symbols that are specific to particular sections or chapters are not
listed here.

\begin{longtable}[l]{p{50pt} p{300pt}}
\toprule
\textbf{Symbol}	& \textbf{Description} \\
\midrule
$\obs$	 	& Observation variables (in vectorized form)\\
$\target$	 	& Target variables (in vectorized form)\\
$\targetProp$	 	& Intermediate/Proposed target variables\\
$K_t,m_t,\cdots$	 	& Random variables at time step $t$\\
$\params$	 	& Set of all model parameters \\
$\alpha, \beta, \mu, \gamma$	 & Model or training parameters \\
$\f$	 	& Pixel or superpixel features such as $(x,y,r,g,b)$ \\
$P(\cdot|\cdot)$	&  Probability distribution or density \\
$\mathcal{N}(\cdot|\cdot)$	& Gaussian distribution \\
$\psi_u$ & Unary potential at each pixel/superpixel \\
$\psi_p$ & Pairwise potential between two pixels/superpixels \\
$L(\cdot), E(\cdot)$ & Loss/Objective/Energy function \\
$\func(\cdot)$ & Generic function relating input to output variables \\
$\Lambda$ & Diagonal matrix for scaling image features say $(x,y,r,g,b)$ \\
\bottomrule
\end{longtable}

\chapter{Acronyms}
\label{chap:abbrv}
\vspace{-0.5cm}
\begin{longtable}[l]{p{80pt} p{300pt}}
\toprule
\textbf{Acronym}	& \textbf{Full Form} \\
\midrule
AR & Acceptance Rate \\
BCL	 	& Bilateral Convolutional Layer\\
BI	 	& Bilateral Inception\\
BNN	 	& Bilateral Neural Network\\
BP & Belief Propagation \\
CMP & Consensus Message Passing \\
CNN	 	& Convolutional Neural Network\\
CPU & Central Processing Unit \\
CRF & Conditional Random Field \\
DDMCMC & Data Driven Markov Chain Monte Carlo\\
EP & Expectation Propagation \\
FC & Fully Connected \\
GPU & Graphics Processing Unit \\
IoU & Intersection over Union \\
INF-INDMH	 	& Informed Independent Metropolis Hastings\\
INF-MIXMH	 	& Informed Mixture Metropolis Hastings\\
INF-MH	 	& Informed Metropolis Hastings\\
KDE & Kernel Density Estimation \\
MAP & Maximum a Posteriori \\
MCMC 	& Markov Chain Monte Carlo\\
MF & Mean-Field \\
MH	 	& Metropolis Hastings\\
MHWG & Metropolis Hastings Within Gibbs \\
MP & Message Passing \\
PGM  & Probabilistic Graphical Model \\
PSNR & Peak Signal to Noise Ratio \\
PSRF & Potential Scale Reduction Factor \\
PT & Parallel Tempering \\
REG-MH & Regenerative Metropolis Hastings \\
RF & Random Forest \\
RMSE & Root Mean Square Error \\
VMP & Variational Message Passing \\
VPN & Video Propagation Network \\
\bottomrule
\end{longtable}

\mainmatter

\chapter{Introduction}
\label{chap:intro}

Computer vision is the task of inferring properties of the world from the
observed visual data. The observed visual data can originate from a variety of
sensors such as color or depth cameras, laser scans etc.
The properties of the world range from low-level material properties
such as reflectance to high-level object properties such as 3D shape and
pose. The field of computer vision encompasses a broad range of problems
involving a variety of sensor data and world properties. Some example
problems include: `Inferring 3D pose and shape of an object from a depth image';
`Inferring the actions of the persons from video' etc.
Computer vision is hard because of variability in lighting, shape and texture in
the scene. Moreover, there is sensor noise and the image signal is non-additive
due to occlusion. Vision problems are inherently ambiguous as the sensor data
is an incomplete representation of the richer 3D world. As a result, probabilistic
frameworks are typically employed to deal with such ambiguity.

Following~\cite{prince2012computer}, there are three main components in any vision
system: \textit{model}, \textit{learning} algorithm and \textit{inference} algorithm.
The \textit{Model} forms the core of any vision system describing the mathematical
relationship between the observed data and the desirable properties of the world.
The set of parameters and/or structure of this mathematical model
is learned using a \textit{learning} algorithm. Once the model
is learned, an \textit{inference} algorithm is used to predict the world properties
from a given observation.

Since models form the core of any vision system, let us briefly discuss the
two broad categories in computer vision models:
\textit{Generative} and \textit{Discriminative}
models, which can be viewed as complementary and inverse to each other.
Let us denote the observed data as a vector of random variables $\obs \in \mathbb{R}^k$ and
the target world properties as another vector of random variables $\target \in \mathbb{R}^l$.
For example, $\obs$ can be a vectorized representation of image pixels and
$\target$ can be a vector representing the parameterized shape of an object
in the image. Generative models characterize the probability of observed data
given the world properties $P(\obs|\target,\params)$ (called `likelihood')
as well as a prior on target variables $P(\target)$,
where $\params$ denotes the parameters of the model.
Some example generative models include graphics systems and probabilistic graphical models.
In this thesis, we use the
term `generative' more loosely in the sense that any model which characterizes
the likelihood $P(\obs|\target,\params)$ and/or prior over target
variables $P(\target)$ is considered a `generative' model.
Discriminative models characterize the probability of world properties
given the observed data $P(\target|\obs,\params)$ (called `posterior distribution').
In other words, generative models model the image formation process as a function
of world parameters, whereas discriminative approaches model the target world
parameters as a function of the given image.
Once the model is defined, a learning algorithm is used to learn the model
parameters $\theta$ and then an inference algorithm is used to predict the
posterior distribution $P(\target|\obs,\params)$. In Chapter~\ref{chap:models},
we will discuss more about these models along with the inference and learning in them.

Depending on the type of model, a specialized learning or inference algorithm
may not be required. For example, in the case of manually specified generative models
(e.g.\ graphics system or fully specified Bayesian network),
there is no need for a specialized learning
algorithm since all the model parameters are already hand specified,
but specialized inference techniques are required to invert such
models. In the case of discriminative models, where the
posterior distribution $P(\target|\obs,\params)$ is directly modeled, (e.g.\ neural
networks or random forests), the inference mechanism just reduces to a simple
evaluation of the model.

This dissertation focuses on improving the \textit{inference} in
prominent computer vision models. Inference plays a crucial role in any vision
system as this would produce the desired end result for a given observation.
The breakthroughs in computer vision technology are often marked by the advances
in the inference techniques as even the model design is often dictated by the
complexity of the inference in them.
The inference result is what matters at the end and
having a model with high fidelity but with no feasible or practical inference scheme
(for instance recent photo-realistic graphics systems) is of little use for
addressing vision problems.
Thus, better inference techniques not only improve the existing computer
vision systems but also help to develop better models. This thesis work
proposes techniques for better inference in existing and widely used computer
vision models.

\section{Thesis Overview}

In this section, we will discuss the objective of this thesis followed by the
organization and contributions of various chapters.

\subsection{Objective}

The main aim of the work presented in this thesis is to improve the performance
of inference algorithms used in different computer vision models. Inference is
highly inter-linked with model design and depending on the type of model, we
propose different techniques to improve inference.

Generative models characterize the image formation process and the inference is
typically performed via Bayesian inference techniques. Despite their
intuitive appeal, the use of generative models in vision is hampered by the
difficulty of posterior inference~(estimating $P(\target|\obs,\params)$).
Existing inference techniques are either often too
complex or too slow to be practical. In this thesis, we aim to alleviate some
of these inference challenges in generative models. Specifically, we concentrate
on improving two different inference schemes for generative vision models: First one is
Markov chain Monte Carlo (MCMC) inference for inverting graphics
engines and the second one is message passing inference in layered graphical models.
A common strategy that we follow to improce inference in generative models is
to learn a new discriminative model that is separate from the given generative
model and propose modified inference schemes that make use of this new discriminative
model for better inference.

Discriminative models directly model the posterior distribution $P(\target|\obs,\params)$
of the desired world parameters given the observed data. Thus the inference amounts
to simple evaluation of the model. One of the main limitations of inference
in discriminative models in the lack of principled techniques to incorporate
our prior knowledge about the task. This is especially the case with the prominent
convolutional neural network (CNN) models. In this thesis, we concentrate on CNN
models and propose techniques to improve inference in them. We modify the original
CNN models and make them more amenable for the incorporation of prior knowledge.

In summary, the aim of this thesis is to improve inference in general computer
vision models. We do this by leveraging machine learning techniques to learn
a new model for inference that is either separate from the original model (in case
of generative models) or modifying the original model itself (in case of
discriminative models). The work in this thesis deals
with the construction and learning of such inference models and how such models
can be integrated into the original vision models for better inference. We propose
techniques for inference in diverse computer vision models ranging from hand-specified
graphics systems to freely-parameterized neural network models. We concentrate on three
types of models which are prevalent in modern computer vision systems:
1.~Graphics systems; 2.~Layered graphical models and 3.~Convolutional neural networks.

\subsection{Organization and Contributions}

Since models form the core part of any vision system and this thesis involves the construction of
new models for inference, in Chapter~\ref{chap:models}, we give an overview of different
computer vision models along with the learning and inference mechanisms that are
usually employed in them. In addition, we review some existing techniques that aim to
improve inference in vision models by combining generative and discriminative approaches.

\vspace{-0.3cm}
\paragraph{Part I: Inference in Generative Vision Models} In Part I
(Chapters~\ref{chap:infsampler} and~\ref{chap:cmp})
of the thesis, we propose techniques for inference in generative computer vision models.

\vspace{-0.3cm}
\paragraph{Chapter~\ref{chap:infsampler} - The Informed Sampler}
In this Chapter, we propose a new sampling technique for inference
in complex generative models like graphics systems. Markov chain Monte Carlo (MCMC)
sampling is one of the most widely used and most generic inference scheme
in such complex generative models. Although generic, in practice,
MCMC sampling suffers from slow convergence unless the posterior is a unimodal low-dimensional
distribution. By leveraging discriminative learning techniques with ancillary clustering
and random forest models, we devise a mixture sampling technique that helps in faster mixing
without losing the acceptance rate. We call this `Informed Sampler' and demonstrate
it using challenging generative graphics models and a popular model of
human bodies~\cite{hirshberg2012coregistration}. Our method is similar in spirit
to ‘Data Driven Markov Chain Monte Carlo’ methods~\cite{zhu2000integrating}.

\vspace{-0.3cm}
\paragraph{Chapter~\ref{chap:cmp} - Consensus Message Passing}
In this Chapter, we concentrate on layered and loopy graphical models that are prevalent
in computer vision applications. When factors (relationship between the variables)
in such graphical models are from a pre-defined
family of distributions, inference is generally performed using
standard message passing techniques such as `expectation propagation'~\cite{Minka2001} and
`variational message passing'~\cite{Winn2005}. We observe that these inference techniques
fail to converge or even diverge when the graphical model is loopy with a large number of
variables. The failure of these inference techniques can be attributed to the algorithm's inability
to determine the values of a relatively small number of influential variables which we call
global variables (e.g.\ light in a scene). Without accurate estimation of these global
variables, it can be very difficult for message passing to make meaningful progress on the
other variables in the model. As a remedy, we exploit the layered structure of the model
and learn ancillary random forest models that learn to predict these influential variables
and use them for better message passing inference. We call this method `Consensus Message Passing' (CMP)
and demonstrate it on a variety of layered vision models. Experiments show that CMP
leads to significantly more accurate inference results whilst preserving the computational
efficiency of standard message passing.

\vspace{-0.3cm}
\paragraph{Part II: Inference in Discriminative Vision Models}
In Part II (Chapters~\ref{chap:bnn},~\ref{chap:vpn} and~\ref{chap:binception}) of the
thesis, we focus on inference in discriminative CNN models.

\vspace{-0.3cm}
\paragraph{Chapter~\ref{chap:bnn} - Learning Sparse High Dimensional Filters}
2D spatial convolutions form the basic unit of CNN models.
Spatial convolutions are perhaps the simplest, fastest and
most used way of propagating information across pixels.
Despite their staggering success in a
wide range of vision tasks, spatial convolutions have several drawbacks: There
are no well-established ways of incorporating prior knowledge into spatial filters;
Spatial convolutions quickly get intractable when filtering data of
increasing dimensionality; and the receptive fields of the filters are image-agnostic.
Spatial convolutions are usually confined to a local neighborhood
of pixels and thus many deep layers of spatial convolutions or post-processing conditional
random field (CRF) formulations are required for long-range propagation of
information across pixels. Bilateral filtering~\cite{aurich1995non, tomasi1998bilateral},
on the other hand, provides a simple yet powerful
framework for long range information propagation across pixels. But the traditional use of
bilateral filtering is confined to a manually chosen parametric from, usually a Gaussian filter.
In this chapter, we generalize the bilateral filter parameterization using a sparse high-dimensional
linear approximation and derive a gradient descent algorithm, so the filter parameters can be learned from data.
We demonstrate the use of learned bilateral filters in several diverse applications where Gaussian bilateral
filters are traditionally employed: color up-sampling, depth up-sampling~\cite{kopf2007joint}
and 3D mesh denoising~\cite{fleishman2003bilateral}. The ability to learn
generic high-dimensional sparse filters allows us to stack several parallel and sequential filters like in
convolutional neural networks (CNN) resulting in a generalization of 2D CNNs which we call `Bilateral Neural
Networks' (BNN). We demonstrate the use of BNNs using an illustrative segmentation problem and sparse character
recognition. Gaussian bilateral filters are also employed for mean-field inference in fully connected
conditional random fields (DenseCRF)~\cite{krahenbuhl2012efficient}. Existing works on DenseCRFs
are confined to using Gaussian pairwise potentials due to the traditional use of Gaussian kernels
in bilateral filtering. By learning bilateral filters, we remove the need of confining
to Gaussian pairwise potentials which has the added advantage of directly learning the pairwise potentials
for a given task. We showcase the use of learning edge potentials in DenseCRF with experiments
on semantic segmentation and material segmentation. In summary, we propose a general technique
for learning sparse high-dimensional filters that help in improving the model and inference in DenseCRF
models and also generalizes 2D CNNs.

\vspace{-0.3cm}
\paragraph{Chapter~\ref{chap:vpn} - Video Propagation Networks}
Videos carry redundant information across frames and the information propagation across video
frames is valuable for many computer vision applications such as video segmentation, color
propagation etc. In this chapter, we propose a novel neural network architecture for video
information propagation. We leverage learnable bilateral filters, developed in the previous
chapter, and propose a `Video Propagation Network' (VPN) that processes video frames in an
adaptive manner. The model is applied online: it propagates information forward without
the need to access future frames other than the current ones. In particular we combine
two components, a temporal bilateral network for dense and video adaptive filtering,
followed by a spatial network to refine features and increased flexibility.
We present experiments on video object segmentation and semantic video segmentation
and show increased performance comparing to the best previous task-specific methods,
while having favorable runtime. Additionally we demonstrate our approach on an example
regression task of propagating color in a grayscale video.

\vspace{-0.3cm}
\paragraph{Chapter~\ref{chap:binception} - Bilateral Inception Networks}
In this Chapter, we propose a new CNN module which we call the `Bilateral Inception' (BI)
module that can be inserted into \emph{existing} segmentation CNN models. BI modules help in image adaptive
long-range information propagation across intermediate CNN units at multiple scales.
We show empirically that this alleviates some of the need for standard post-processing inference
techniques such as DenseCRF.~In addition, our module helps in recovering the full resolution of segmentation result,
which is generally lost due to max-pooling and striding. Experiments on different base
segmentation networks and datasets showed that our BI modules result in
reliable performance gains in terms of both speed and accuracy in comparison to traditionally
employed DenseCRF/Deconvolution techniques and also recently introduced dense pixel prediction techniques.

\section{List of Publications}

\nobibliography*

The contributions in this thesis mainly comprise of work from the following publications
~\cite{jampani2014,jampani15aistats,kiefel15bnn,arxivpaper,gadde16bilateralinception,
jampani16vpn}:

\begin{itemize}

\item \bibentry{jampani2014}.~\cite{jampani2014}
\item \bibentry{jampani15aistats}.~\cite{jampani15aistats}
\item \bibentry{kiefel15bnn}.~\cite{kiefel15bnn}
\item \bibentry{arxivpaper}.~\cite{arxivpaper}
\item \bibentry{gadde16bilateralinception}.~\cite{gadde16bilateralinception}
\item \bibentry{jampani16vpn}.~\cite{jampani16vpn}

\end{itemize}

The following publications~\cite{jampani15wacv,jampani15wacv,gadde2016efficient}
are a part of my PhD research but are outside the scope of this thesis:

\begin{itemize}

\item \bibentry{jampani15wacv}.~\cite{jampani15wacv}
\item \bibentry{sevilla:CVPR:2016}.~\cite{sevilla:CVPR:2016}
\item \bibentry{gadde2016efficient}.~\cite{gadde2016efficient}

\end{itemize}

\chapter{Models and Inference in Computer Vision}
\label{chap:models}

Models play a central role in the study of both biological and artificial
vision. Helmholtz, in the 19th century, popularized the human vision as a result
of psychological inference in learned models~\cite{frey2003advances,cahan1993hermann}
as opposed to native processing in lower visual system or eyes.
With the advent of computers in the 20th century, researchers are able to
formulate, learn and evaluate several computational models of vision.
More recently, with powerful parallel computing hardware like graphics processing units
(GPU), researchers are able to learn and do inference in highly complex models with even millions
of parameters. In this chapter, we present an overview of different computer vision models and
discuss the inference and learning techniques therein. Since this thesis work
mainly constitutes the development of new inference techniques, we emphasize the
difficulty of inference in different models and discuss several remedies proposed
in the literature.

\section{Models in Computer Vision}

Models describe the mathematical relationship between the observed data and the
desired properties of the world. Computer vision models are often
probabilistic in nature due to the inherent ambiguity in vision problems.
Due to the broad range of problems in computer vision, there is no single model
that can work well for various vision tasks. Depending on the nature of problem and
the availability of data, different models work well for different scenarios.
Visual data is complex with variability arising due
world properties such as occlusion, lighting, texture, geometry, depth ordering
etc. It is very difficult to model the relationship between all the aspects of the world
and the visual data, and do inference therein. Vision models usually are highly
specific to one or few aspects of the world.

As briefly mentioned in Chapter~\ref{chap:intro}, computer vision models can
be broadly classified into two types: \textit{Generative} and \textit{Discriminative}
models, which can be viewed as complementary and inverse to each other.
Generative models characterize the probability of observed data given the world
properties $P(\obs|\target,\params)$ and discriminative models characterize
the probability of world properties given the observed data $P(\target|\obs,\params)$,
where $\params$ denotes the parameters of the model.
In other words, generative models model the image formation process as a function
of world parameters, whereas discriminative approaches model the desired world
parameters as a function of the given image. As mentioned in Chapter~\ref{chap:intro},
we use the term `generative' more loosely in the sense that any model which characterizes
the likelihood $P(\obs|\target,\params)$ and/or prior over target
variables $P(\target)$ is considered a `generative' model.
Next, we give an overview of these two complementary models discussing the
advantages and disadvantages of both.

\begin{figure}[th!]
	\centering
  \subfigure[Graphics Renderings]{
		\setlength\fboxsep{-0.3mm}
		\setlength\fboxrule{0pt}
		\parbox[b]{5cm}{
			\centering
			\small
      \fbox{\includegraphics[width=5cm]{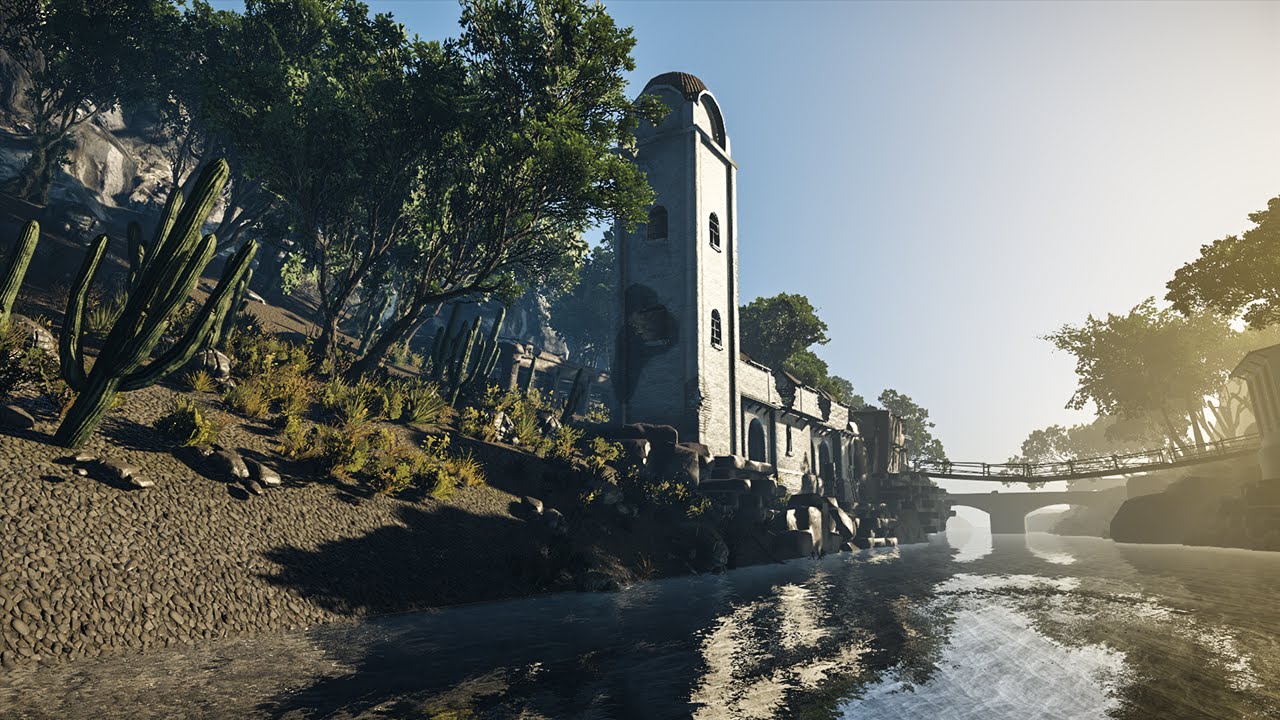}}\\
      \vspace{4.5mm}
			\fbox{\includegraphics[width=5cm]{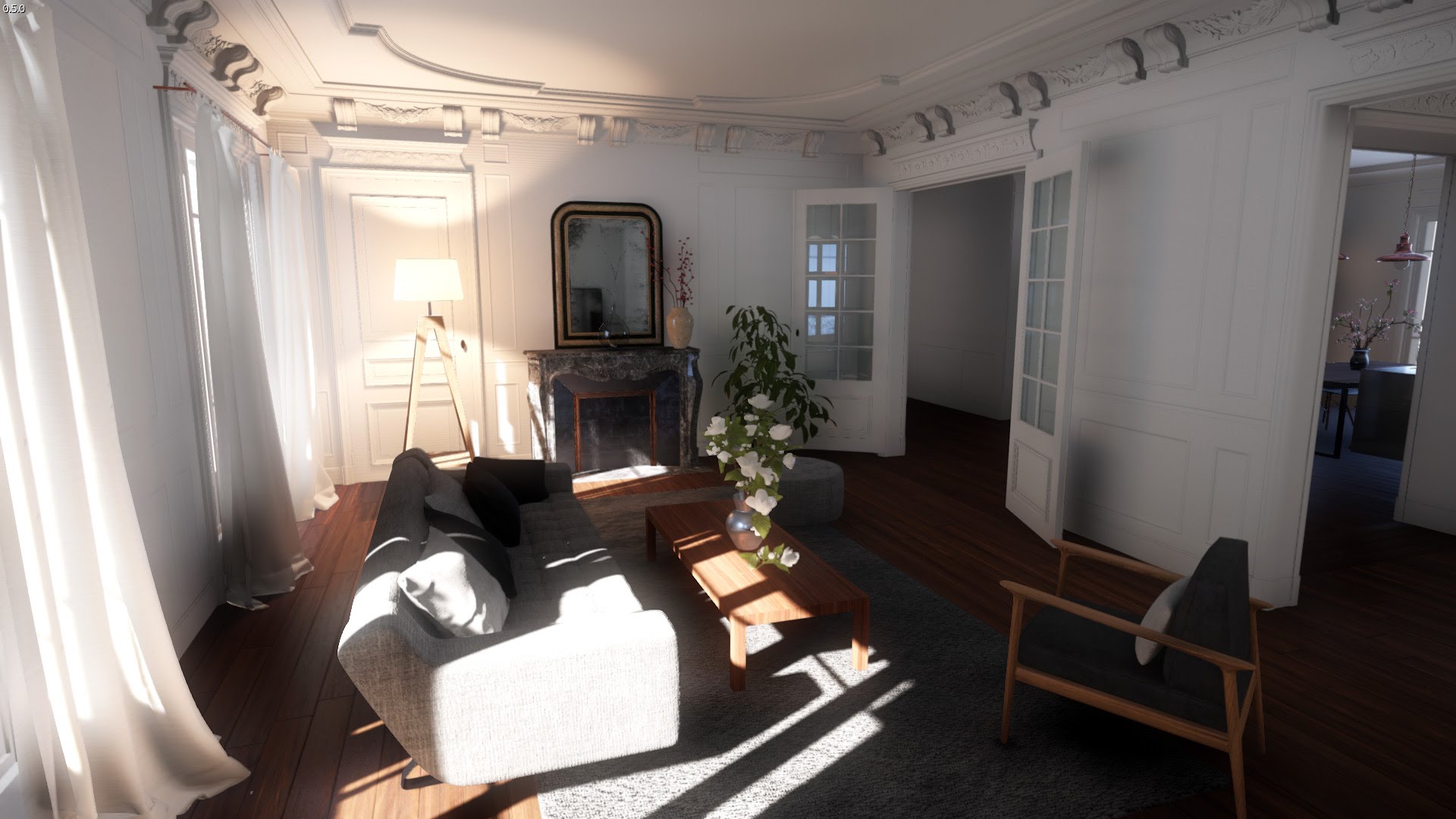}}\\
      \vspace{4.5mm}
			\fbox{\includegraphics[width=5cm]{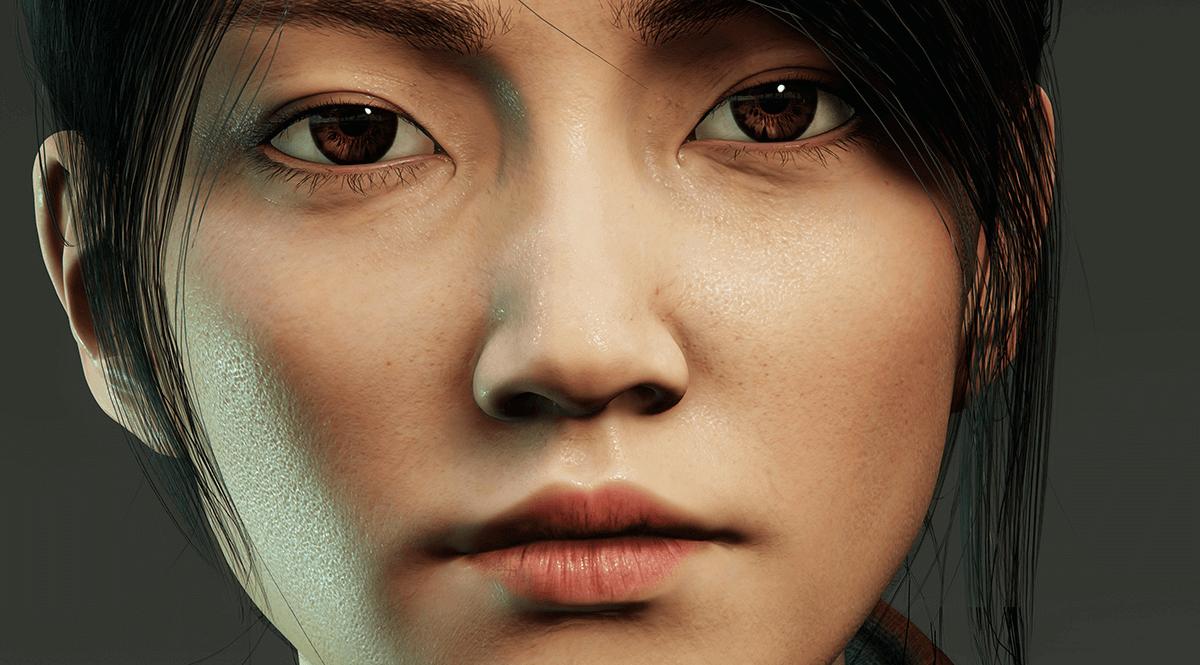}} \\
		}
		\label{fig:sample-graphics}
	}
  \hspace{1cm}
	\subfigure[A face generative model]{
		\begin{tikzpicture}

			%

			\node[obs] 											(x)			{$x_i$}; %
			\node[latent, above=of x]                			(z)			{$z_i$}; %

			\factor[above=of z, yshift=10mm] 					{times}		{below left: $\times$} {} {}; %

			\node[latent, above=of times, yshift=-5mm]			(s)			{$s_i$}; %

			\factor[above=of x, yshift=1mm] 					{noise1}	{left:Gaussian} {} {}; %

			\node[latent, left=of times]                		(r)			{$r_i$}; %

			\factor[above=of r] 								{pr}		{} {} {}; %

			\factor[above=of s, yshift=10mm] 					{inner}		{left:Product} {} {}; %

			\node[latent, above=of inner, yshift=-5mm] 			(n)			{$\mathbf{n}_i$}; %
			\node[latent, right=of inner]                		(l)			{$\mathbf{l}$}; %

			\factor[above=of l] 								{pl}		{} {} {}; %

			\factor[above=of n] 								{pn}		{} {} {}; %

			\factoredge {z} 		{noise1} 		{x}; %
			\factoredge {s} 		{times} 		{z}; %
			\factoredge {r} 		{times} 		{}; %
			\factoredge {n} 		{inner} 		{s}; %
			\factoredge {l} 		{inner} 		{}; %
			\factoredge {} 			{pn} 			{n}; %
			\factoredge {} 			{pl} 			{l}; %
			\factoredge {} 			{pr} 			{r}; %

			\plate {} {(pn) (x) (r)} {}; %

			%
			%
			%



			\node[yshift=0.35cm, xshift=-0.83cm] at (inner) { Inner };

		\end{tikzpicture}
		\label{fig:face-model}
	}
  \subfigure[Example face data]{
		\setlength\fboxsep{-0.3mm}
		\setlength\fboxrule{0pt}
		\parbox[b]{3.4cm}{
			\centering
			\small
			\fbox{\includegraphics[width=1.3cm]{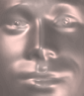}}\\
			Normal $\{\mathbf{n}_i \}$ \vspace{4.5mm} \\
			\fbox{\includegraphics[width=1.3cm]{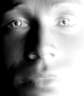}} \\
			Shading $\{ s_i \}$ \vspace{4.5mm} \\
			\fbox{\includegraphics[width=1.3cm]{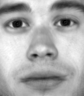}} \\
			Reflectance $\{ r_i \}$ \vspace{4.5mm} \\
			\fbox{\includegraphics[width=1.3cm]{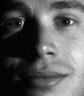}} \\
			Observed image $\{ x_i \}$\\
		}
		\label{fig:face-data}
	}
	\mycaption{Sample generative models in computer vision}
  {(a)~Sample renderings from modern graphics engines~\cite{cryengine,lumberyard}.
  Modern graphics provide renderings with stunning level of realism and vision
  problems can be approached as inverting such graphics systems. Images courtesy
  from the official websites of CryEngine~\cite{cryengine} and Lumberyard~\cite{lumberyard}.
  (b)~A layered graphical model (factor graph) for faces (explained in Sec.~\ref{sec:pgm}).
	Here the vision problem
  could be inferring the reflectance map, normal map and light direction from a given
  face image. (c) Sample face data from Yale B face dataset~\cite{Georghiades2001,Lee2005}.}
	\label{fig:sample-gen-models}
\end{figure}

\section{Generative Vision Models}
\label{sec:gen-models}

A conceptually elegant view on computer vision is to consider a generative
model of the physical image formation process. The observed image becomes a
function of unobserved variables of interest (for instance the presence and positions
of objects) and nuisance variables (for instance light sources and shadows).
When building such a generative model, we can think of a scene description
$\target$ that produces an image $\obs=G(\target,\theta)$ using a deterministic
rendering engine $G$ with parameters $\params$, or more generally,
results in a distribution over images, $P(\obs|\target,\params)$.
Generative models provide a powerful framework for probabilistic reasoning and
are applicable across a wide variety of domains, including computational biology,
natural language processing, and computer vision. For example, in computer vision,
one can use graphical models to express the process by which a face is lit and
rendered into an image, incorporating knowledge of surface normals, lighting and
even the approximate symmetry of human faces. Models that make effective use of this
information will generalize well, and they will require less labelled training data
than their discriminative counterparts (e.g.\ random forests or neural networks) in order
to make accurate predictions.

Given an image observation $\obs$ and a prior over scenes $P(\target)$ we can
then perform \textit{Bayesian inference} to obtain the posterior distribution over the
target variables $P(\target | \obs,\params)$
(also called `posterior inference'):

\begin{equation}
P(\target | \obs,\params) =
\frac{P(\obs|\target,\params)P(\target)}{P(\obs)} =
\frac{P(\obs|\target,\params)P(\target)}{\sum_{\target'} P(\obs|\target',\params)P(\target')}.
\label{eqn:bayes}
\end{equation}

The summation in the denominator runs over all the possible values of $\target$ variable and
would become integral in the case of continuous
variables. $P(\obs|\target,\params)$ is the \emph{likelihood} of the observed data. Inspecting
the above Bayesian formula shows that it is straight forward to compute the numerator
as that involves simply evaluating the generative model. The key difficulty with
Bayesian inference is computing the denominator in Eq.~\ref{eqn:bayes}.
Often, it is not feasible to evaluate the summation
over all the possible target variables even for slightly non-trivial models. This
makes it difficult to obtain closed-form solutions for posterior inference resulting
in the development and use of several approximate inference techniques such
as Markov Chain Monte Carlo (MCMC) sampling and variational inference, which we
will briefly discuss later.

There are different types of generative models with varying model fidelity and
complexity of inference therein. In general, generative models which accurately
model the image formation process (e.g.\ graphics engines) have complex
non-linear functions resulting in a challenging inference task. Building
a good generative model for a given task involves finding a good trade-off
between model fidelity and inference complexity.

Figure~\ref{fig:sample-gen-models} shows some sample generative models in computer vision.
Advances in graphics and physics of light transport resulted in generative models with
high fidelity as is evident in modern computer games and animation movies.
But it is difficult to invert such complex models.
Probabilistic graphical models provide the widely adopted framework
for generative computer vision models. Although graphical models have
less fidelity and only model one or two aspects of the world properties,
they are generally preferred over graphic systems as inference
in them is faster and more reliable. Next, we will discuss these two prominent
generative vision models: \textit{Inverse graphics} and
\textit{Probabilistic graphical models}.

\subsection{Inverse Graphics}
\label{sec:inv-graphics}

\begin{figure}[t]
\begin{center}
\centerline{\includegraphics[width=0.7\columnwidth]{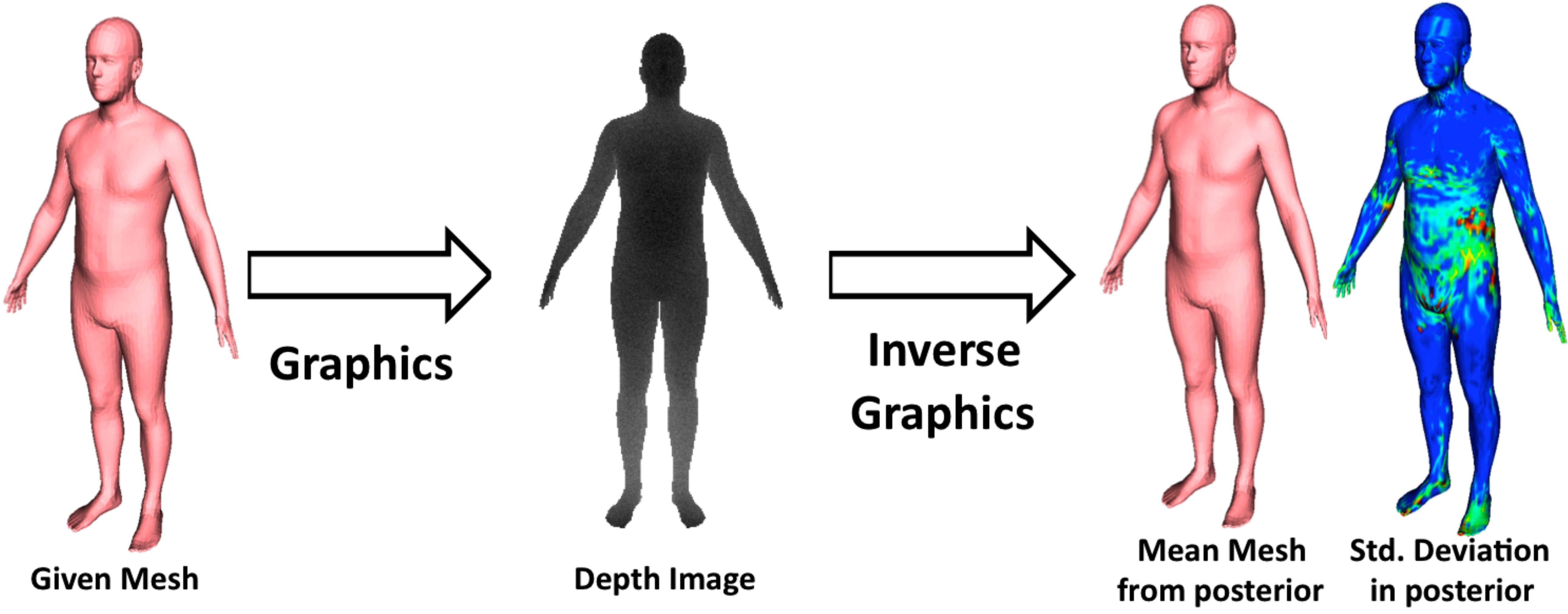}}
\mycaption{An example `inverse graphics' problem}{A graphics engine renders a 3D
  body mesh and a depth image using an artificial camera. By Inverse
  Graphics we refer to the process of estimating the posterior
  probability over possible bodies given the depth image.}
\label{fig:invgraphicsteaser}
\end{center}
\end{figure}

Modern graphics engines
(e.g., game engines like CryEngine~\cite{cryengine} and Lumberyard~\cite{lumberyard})
leverage dedicated hardware setups and
provide real-time renderings with stunning level of realism.
Some sample renderings from modern game engines~\cite{cryengine,lumberyard}
are shown in Fig.~\ref{fig:sample-graphics}.
Vision problems can be tackled with posterior
inference (Eq.~\ref{eqn:bayes}) in such accurate computer graphics systems.
This approach for solving vision problems can be understood as
\textit{Inverse Graphics}~\cite{baumgart1974inversegraphics}. The
target variables $\target$ correspond to the input to the graphics system and
the observation variables are the output $\obs=G(\target,\theta)$. The
deterministic graphics system $G$ can be converted into a probabilistic generative
model by defining a prior over the target variables (input to the graphics system) $P(\target)$
and also defining an approximate likelihood function $P(\obs|G(\target,\theta))$
characterizing the model imperfections. If the model imperfections are neglected,
the likelihood can be given using the \emph{delta} function
$P(\obs|\target) = \delta(\obs-G(\target,\theta))$.
An example `inverse graphics' problem, which we tackle in the next chapter,
is depicted in Fig.~\ref{fig:invgraphicsteaser},
where the graphics engine renders a depth image given 3D body mesh
and camera parameters, and a vision problem would be the inverse estimation
of body shape given a depth image.

Modern graphic engines are based on physics principles and thus most of the
rendering parameters $\params$ are set to mimic the real world physics. Learning in graphics
generative models mainly involves learning the prior $P(\target)$
over the target world properties which are input to the graphics system.
Depending on the type of model, several learned priors are proposed in the
literature. An example is the SCAPE model~\cite{anguelov2005scape,hirshberg2012coregistration}
for modeling the prior over human body shape and pose for the example
shown in Fig.~\ref{fig:invgraphicsteaser}.

Since modern rendering engines involve complex non-linear functions, it is
usually not feasible to obtain a closed-form solution for the posterior distribution
(Eq.~\ref{eqn:bayes}). Even several approximate inference techniques like
variational optimization techniques~\cite{koller2009probabilistic} cannot
be easily employed for posterior inference in complex graphics systems.
`Monte Carlo' sampling provides a generic inference technique for
such complex models and can be used even when the internals
of the graphics systems are not known. The aim of sampling methods is to
characterize the posterior distribution with \emph{independent and identically
distributed} samples. There exists many Monte Carlo sampling
strategies such as uniform sampling, rejection sampling,
importance sampling~\cite{hammersley1954poor,rosenbluth1955monte} etc.
Here, we limit our discussion to the widely used Markov Chain Monte Carlo (MCMC)
sampling techniques.

\subsubsection*{MCMC Sampling}
MCMC sampling~\cite{metropolis1953} is a
particular instance of sampling methods that generates
a sequence of target variables (samples) by simulating a \textit{reversible}
Markov chain.

\begin{equation}
\target_1 \to \target_2 \to \target_3 \to \dots  \to \target_n
\end{equation}•

This Markov chain of samples approximates the target distribution (posterior
distribution in the case of `inverse graphics'). The Markov property states that
at every sequence step $t$, given the present sample $\target_t$ in the sequence,
the next sample $\target_{t+1}$ is independent of all the previous samples
$P(\target_{t+1}|\target_t,\dots,\target_1) = P(\target_{t+1}|\target_t)$.

Let us denote the target distribution with $\pi(\cdot)$. In the inverse
graphics setting, the target distribution is the posterior $\pi(\target)=P(\target | \obs,\params)$.
And, let us denote the transition probability between the two states (samples) in the
Markov chain be $T(\target_{t+1}|\target_{t})$ or $T(\target_t\to\target_{t+1}) \in [0,1]$. One way to ensure that the
Markov chain is \emph{reversible} and converging to the target distribution is to check whether the
following \emph{detailed balance condition} holds for any two states
$\target_t$ and $\targetProp$~\cite{koller2009probabilistic}:

\begin{equation}
\pi(\target_t) T(\target_t \to \targetProp) = \pi(\targetProp) T(\targetProp \to \target_t)
\end{equation}

Note that the above detailed balance condition is satisfied when the transition probability distribution is
close to the target distribution. Since we do not know the target distribution, designing
transition probability distributions that satisfies the detailed balance condition is difficult.
The Metropolis-Hastings (MH) algorithm~\cite{metropolis1953}, instead of devising special transition probabilities,
introduces the \emph{acceptance probability} to each Markov chain transition
$A(\target_t \to \targetProp) \in [0,1]$. The detailed balance condition then becomes:

\begin{equation}
\pi(\target_t) T(\target_t \to \targetProp) A(\target_t\to\targetProp) =
\pi(\targetProp) T(\targetProp \to \target_t) A(\targetProp\to\target_t)
\end{equation}

It can be verified~\cite{koller2009probabilistic} that the following acceptance
probability satisfies the above detailed balance condition:

\begin{equation}
A(\target_t\to\targetProp) = \min\left(1,\frac{\pi(\targetProp)
T(\targetProp \to \target_t)}{\pi(\target_t) T(\target_t \to \targetProp)} \right)
\end{equation}

With the use of the above acceptance rule, instead of designing task-specific
transition probability distributions,
any transition probability distribution $T$ with non-zero probability over
the range of all target variables can be used for MH sampling. $T$ is also
called `proposal distribution' since it is used to propose the next sample in the
Markov chain, which is then either accepted or rejected based on the acceptance probability.
Below, we summarize the Metropolis-Hastings (MH) MCMC algorithm.

\paragraph{Metropolis-Hastings (MH) MCMC:}
Sampling from $\pi(\cdot)$ consists of repeating the
following two steps~\cite{liu2001montecarlo}:

\begin{enumerate}
\item Propose a transition using a \textit{proposal distribution $T$}
  and the current state $\target_t$
\begin{equation*}
\targetProp \sim T(\cdot|\target_t)
\end{equation*}
\item Accept or reject the transition based on Metropolis Hastings (MH) acceptance rule:
\begin{equation*}
\target_{t+1} = \left\{
  \begin{array}{cl}
      \targetProp,  & \textrm{rand}(0,1) < \min\left(1,\frac{\pi(\targetProp)
T(\targetProp \to \target_t)}{\pi(\target_t) T(\target_t \to \targetProp)} \right), \\
     \target_t, & \textrm{otherwise.}
  \end{array}
\right.
\end{equation*}
\end{enumerate}

Different MCMC techniques mainly differ in the type of the
proposal distribution $T$. Note that we do not need to compute the target (posterior)
probabilities, but only the \textit{ratio} of
posterior probabilities $\frac{\pi(\targetProp)}{\pi(\target_t)}$.
This makes MCMC sampling suitable for the inverse graphics setting
where it is not feasible to get a closed-form solution for the normalization
constant in the posterior distribution (denominator in Eq.~\ref{eqn:bayes}).

The key aspect of the MH sampling is the number of steps it requires until
it converges to the target distribution. If the proposal distribution is very
different from the target distribution, the samples tend to be frequently rejected
resulting in a long wait for convergence. In practice, it is difficult to
measure the convergence of any sampler since we do not know the target
distribution. In Chapter~\ref{chap:infsampler}, we discuss several diagnostic
measures that indicate the convergence of MCMC sampling. In the case of
`inverse graphics', each forward rendering step takes a considerable amount of time and
we would like the sampler to accept as many samples as possible. A key for improving
the MH sampling efficiency is to design the proposal distributions that match
the target posterior distribution. In Chapter~\ref{chap:infsampler}, we devise
such technique by leveraging discriminative learning techniques for learning
the proposal distribution. Refer to~\cite{liu2001montecarlo,koller2009probabilistic}
for more details on MCMC sampling. In Chapter~\ref{chap:infsampler}, we study
the behavior of MCMC sampling and its variants for inverting graphics
engines and propose techniques for improving the sampling efficiency.

\subsection{Probabilistic Graphical Models}
\label{sec:pgm}

Probabilistic graphical models (PGM) provide a rigorous mathematical framework,
based on probability and graph theory, for modeling the relationship between
the world and image properties.
PGMs have been popular not only in computer vision but also in
related fields such as natural language processing,
speech processing, etc. Several model representations, learning and inference
schemes haven been developed in the PGM literature and even a concise description
of them would be an inundating task and is outside the scope of this thesis. Refer
to~\cite{koller2009probabilistic} for a comprehensive overview of PGMs.
PGMs generally represent input-output relationships with
factorized functions and are typically confined to a restricted domain so that
efficient inference techniques can be applied.

PGMs are popular models of choice when the joint distribution of all the target
and observation variables can be factorized into independent distributions
each involving a subset of variables. This factorization of the joint distribution
is represented with the structure of the graph, where each node represents a subset
of variables and edges between the nodes represent the joint or conditional
distributions between the corresponding node variables.

\begin{figure}[t]
	\centering
		\begin{tikzpicture}

			\node[latent] (var_a) {a};
			\node[latent, right=of var_a, xshift=4mm] (var_f) {f};
			\node[latent, right=of var_f, xshift=4mm] (var_ce) {c,e};
			\node[latent, below=of var_f, yshift=-4mm] (var_b) {b};
			\node[latent, below=of var_ce, yshift=-4mm] (var_d) {d};

			\factor[left=of var_a, xshift=-2mm] {factor_a} {$i$} {} {};
			\factor[left=of var_f, xshift=-2mm] {factor_f} {$m$} {} {};
			\factor[below right=of var_a, xshift=5mm, yshift=-5mm] {factor_ab} {$j$} {} {};
			\factor[above right=of var_b, xshift=4mm, yshift=4mm] {factor_fceb} {$l$} {} {};
			\factor[right =of var_b, xshift=2mm] {factor_bd} {$k$} {} {};
			\factor[right =of var_f, xshift=2mm] {factor_fce} {$n$} {} {};

			\edge[-] {factor_a} {var_a};
			\edge[-] {factor_f} {var_f};
			\edge[-] {var_a} {var_b};
			\edge[-] {var_b} {var_ce};
			\edge[-] {var_f} {factor_fceb};
			\edge[-] {var_b} {var_d};
			\edge[-] {var_f} {var_ce};

			\edge[->, transform canvas={xshift=8.4mm,yshift=-2.4mm,scale=0.8},black!60!green] {var_ce} {factor_fceb};
			\edge[->, transform canvas={xshift=4mm,yshift=-2mm,scale=0.8},black!60!green] {var_f} {factor_fceb};
			\edge[->, transform canvas={xshift=6.5mm,yshift=-4mm,scale=0.8},red] {factor_fceb} {var_b};
			\edge[->, transform canvas={xshift=0.56cm,yshift=-5.7mm,scale=0.8},black!60!green] {var_b} {factor_bd};
			\edge[->, transform canvas={xshift=0.2cm,yshift=-4mm,scale=0.8},red] {factor_ab} {var_b};

		\end{tikzpicture}
		\label{fig:example-graph}
	\mycaption{An example factor graph}{Variable nodes (circles) and factor nodes (squares) representing
	the factorized function
	$P(a,b,c,d,e,f)=\psi_i(a)\psi_j(a,b)\psi_k(b,d)\psi_l(b,c,e,f)\psi_m(f)\psi_n(c,e,f)$. Also shown are sample
	variable-to-factor messages (green arrows) and, factor-to-variable messages (red arrows).}
\end{figure}

\subsubsection{Factor Graph Representation}

Factor graphs provide a useful visualization and mathematical formalization for representing probabilistic
graphical models. Factor graphs are \emph{bipartite} graphs where nodes in the graph
are divided into two types: `variable' nodes represented as circles and `factor' nodes
represented as squares. Variable nodes represent the random variables in the graphical model and
factor nodes represent the statistical relationship between the variable nodes that they are connected to.
Every edge in the factor graph connects a variable node to factor node. That is,
there are no \emph{direct} edges among the variable nodes or factor nodes. An example
factor graph is shown in Fig.~\ref{fig:example-graph}, where we represent variable nodes
as $a,b,c,\cdots$ and factor nodes as $i,j,k,\cdots$. The factor function associated with
a factor node $a$ is represented as $f_a$ and the states of variables associated with
a variable node $i$ is represented as $i$.
The joint function in Fig.~\ref{fig:example-graph},
$P(a,b,c,d,e)$ is factorized into five independent factors
$\psi_i(a)\psi_j(a,b)\psi_k(b,d)\psi_l(b,c,e,f)\psi_m(f)f_n(c,e,f)$.
Each factor in the graph represent a function and the product of all factor functions makes
up the joint function. In the context of probabilistic generative models, these
functions are probability distributions. The edges in the factor can be either directed
or un-directed representing the joint and conditional distributions respectively.
In the case of discrete variables, the factor functions are probability tables
assigning the probabilities for all the states of the factor variables. And, in the
case of continuous variables, the factor functions are probability density functions.
Another example of factor graph, representing a generative model of faces, is
shown in Fig.~\ref{fig:face-model}. We will discuss more about this face model later
in this section.

Graphical representations like factor graphs have several advantages. They provide
an intuitive way of representing the statistical dependencies between the variables.
Given the factor graph, it is easy to visualize \emph{conditional independence}
between variables.
For a factor graph with undirected edges, variables $s$ and $t$ are independent
given a set of variables $\nu$ ($s \ci t | \nu$)
if every path between $s$ and $t$ have some node $v\in\nu$.
For instance, in the factor graph shown in Fig.~\ref{fig:example-graph},
variable $d$ is independent of $a,c,e$ and $f$ given $b$ is observed: $d \ci a,c,e,f | b$.
Perhaps the most important advantage of graphical representations like factor graphs
is that we can perform Bayesian inference by using the graph structure. For example, the widely used
\emph{message-passing} inference is performed by passing messages between factor and variable nodes.

\subsubsection{Message Passing} Message Passing (MP) inference forms a general class
of algorithms that are used to estimate the factor graph distributions,
i.e. maximizing or minimizing the joint distribution (P(a,b,c,d,e,f) in Fig.~\ref{fig:example-graph}).
MP inference proceeds by passing messages (distributions or density functions)
between variable and factor nodes. Here, we describe the message passing in the
famous `Sum-Product' belief-propagation (BP) algorithm.
Messages are probability distributions that represent the beliefs
over the variables to/from which the messages are being sent.
MP inference in factor graphs has two types of messages: Messages
from variable to factor nodes $\mu_{a\rightarrow i}$ (green arrows in Fig.~\ref{fig:example-graph})
and messages from factor to variable nodes $\mu_{i\rightarrow a}$
(red arrows in Fig.~\ref{fig:example-graph}).
In the case of Sum-Product BP, these messages are defined as follows.

\textit{Variable-to-Factor Message:} A message from variable to factor node
is the product of all the messages that the variable node receives from its
neighboring factor nodes except the recipient factor node.
In Fig.~\ref{fig:example-graph}, the message $\mu_{b \rightarrow m}$
from variable $b$ to factor $m$ is the product of the messages that $b$ receives,
i.e. $\mu_{k \rightarrow b} \mu_{l \rightarrow b}$. In general, a message from
a variable node $g$ to factor node $r$ is defined as:

\begin{equation}
\mu_{g \rightarrow r}(x_g) = \prod_{\hat{r}\in N(g)\setminus \{r\}} \mu_{\hat{r} \rightarrow g}(x_g),
\end{equation}

where $N(g)$ is the set of neighboring nodes to $g$ and $x_g$ represent the states
of $g$.

\textit{Factor-to-Variable Message:} A message from factor to variable node
is first computed as the product of factor function with all the incoming messages
from variables except the target variable. The resulting product is then marginalized
over all the variables except the one associated with the target node variables.
For example, in Fig.~\ref{fig:example-graph}, the message from $l$ to $b$ is computed
by marginalizing the product $\psi_l(b,c,e,f)\mu_{c,e\rightarrow l}\mu_{f\rightarrow l}$
over non-target variables $c,e,f$. In general, a message from factor node $r$
to variable node $g$ is defined as:

\begin{equation}
\mu _{r\to g}(x_g)=\sum _{x':V_{r}\setminus g} \psi_{r}(x')\prod_{\hat{g}\in
N(r)\setminus \{g\}}\mu _{\hat{g}\to r}(x_{\hat{g}}),
\end{equation}

where $N(r)$ represents the neighboring nodes of $r$ and $V_r$ represent all the variables
attached to $r$. The summation in the above equation becomes an integral when
dealing with continuous variables.

In a typical message passing inference run, the above messages are repeatedly computed
and sent between factor and variable nodes. Depending on the structure of the factor graph,
different message priorities are used. Upon convergence or passing messages for a
pre-defined number of iterations, the marginal distribution at each variable node is
computed as the product of all its incoming messages from the neighboring factor nodes:

\begin{equation}
P(x_g) = \prod_{\hat{r}\in N(g)} \mu_{\hat{r} \rightarrow g}(x_g).
\end{equation}

Messages represent a probability of each possible state of a variable.
In the case of discrete variables with small number of states, it is easy to represent
messages as probability distribution tables. However, in the case of continuous variables,
the messages must be full functions of the variables. Some approximations are typically
required to represent messages with continuous distributions.
Different message passing algorithms differ in the way the messages are approximated.
Some MP algorithms assume Gaussian form for the messages with either restricting
the type of factor functions~\cite{weiss2001correctness} that always result in Gaussian messages or
projecting the computed messages into Gaussian form~\cite{Minka2001}.
Some other MP algorithms use mixture of Gaussians~\cite{sudderth2010nonparametric} or set of
particles/samples~\cite{ihler2009particle} for representing messages.

\subsubsection{Variational Inference}
Except for small graphical models, exact Bayesian inference (say, with
the above mentioned Sum-Product BP) in PGMs is
usually not possible. This is especially true for computer vision models which
typically involve several hundreds or thousands of variables.
Variational Bayesian inference is one of the widely used technique for performing
inference in PGMs. These methods try to find an
approximation $Q(\target)$ to the true posterior distribution $P(\target|\obs)$
via optimization techniques.
This approximate distribution is usually taken from a known simpler family
of distributions (such as exponential family) and the optimization is
performed over the space of that family of distributions. For example, for the
above joint distribution function $P(a,b,c,d,e,f)$,
an approximation could be a factorized distribution
$Q = Q_1(a) Q_2(b) Q_3(c) Q_4(d) Q_5(e) Q_6(f)$, each from an exponential family of distributions.
The most commonly used optimization function is minimizing the
Kullback-Leibler (KL) divergence of $P$ from $Q$ in order to find a close approximation
to $P$:

\begin{equation}
D_{KL}(Q||P) = \underbrace{\sum_z Q(\target) \frac{Q(\target)}{P(\target,\obs)}}_{-\mathcal{L}(Q)} +
\log P(\obs)
\label{eqn:kl}
\vspace{-0.2cm}
\end{equation}

Minimizing the above KL-divergence translates to maximizing the variational
lower bound $\mathcal{L}(Q)$ as $P(\obs)$ is constant.
$\mathcal{L}(Q)$ is also called \textit{energy functional} or
\textit{variational free energy} as it can be written as the sum of energy
$\mathbb{E}_Q[\log P(\target,\obs)]$ and the entropy of $Q$. $Q$ is chosen
in such a way that $\mathcal{L}(Q)$ becomes tractable and maximizable.

In general, the energy functional is maximized by passing messages between different
variable nodes in the factor graph (e.g. factor graphs in
Figs.~\ref{fig:example-graph},~\ref{fig:face-model}).
Following~\cite{koller2009probabilistic}, depending on the type of approximations
to $Q$ and the energy functional, methods for maximizing $\mathcal{L}(Q)$ can
be categorized into three types. The first category of methods optimizes approximate
versions of the energy functions by passing messages in the simplified versions of the
given factor graph. This includes the
\textit{loopy belief propagation}~\cite{frey1998revolution,weiss2000correctness} algorithm.
The second category of methods try to maximize the exact energy functional but
uses approximate message propagation steps, for example approximating complex
messages with distributions from the exponential family. This is equivalent to using
relaxed consistency constraints on $Q$. These class of methods are also known as
\textit{expectation propagation} (EP)~\cite{Minka2001} algorithms.
The third commonly used category of methods maximize the exact energy functional
but restrict $Q$ to simple factorized distributions, which is called \textit{mean-field}
approximation. One of the most commonly used message passing technique for
optimizing the energy functional is \textit{variational message passing} (VMP)~\cite{Winn2005}.
In Chapter~\ref{chap:cmp}, we show how EP and VMP fail to converge for inference
in model shown in Fig.~\ref{fig:face-model} and propose a remedy for that.

Several other inference techniques such as MCMC sampling can be used for inference
in PGMs. Refer to~\cite{fox2012tutorial} for a tutorial on variational Bayesian
inference and to~\cite{koller2009probabilistic,wainwright2008graphical}
for a comprehensive review of various inference techniques in PGMs.

\subsubsection{Two Example Models}
\label{sec:example_modesl_crf}
Here, we give a brief overview of two popular types of
PGMs in vision which are later used in this thesis:
Layered graphical models and fully connected CRFs.

\paragraph{Layered Graphical Models:} Several vision models are hierarchical
in nature and can be naturally expressed
with layered graphical models. Figure~\ref{fig:face-model} shows an example
layered \textit{factor graph} model for faces. Here, a vision task could be:
Given an observation of pixels $\mathbf{x} = \{x_i\}$,
we wish to infer the reflectance value $r_i$ and normal vector $\mathbf{n_i}$
for each pixel $i$ (see Fig.~\ref{fig:face-data}).
The model shown in Fig.~\ref{fig:face-model} represents the following approximate image
formation process: $x_i = (\mathbf{n_i} \cdot \mathbf{l}) \times r_i + \epsilon$,
thereby assuming Lambertian reflection and an infinitely distant directional
light source with variable intensity. Each factor in the graph is a
conditional probability distribution providing the factorization for the
joint distribution:

\begin{equation}
P(\obs, \mathbf{z}, \mathbf{r}, \mathbf{s}, \mathbf{n}, \mathbf{l})
= P(\mathbf{n}) P(\mathbf{l}) P(\mathbf{r}) \prod_i P(x_i) P(x_i | z_i)
P(z_i | r_i, s_i) P(s_i | \mathbf{n}_i, \mathbf{l}),
\end{equation}

where $s_i$ and $z_i$ represent the intermediate shading and non-noisy
image observation variables. We omitted the model parameters $\params$
in the above equation for the sake of simplicity.
A vision task could to estimate the posterior distribution
$P(\mathbf{r}, \mathbf{s}, \mathbf{n}, \mathbf{l} | \obs)$.
Note that this generative model is only a crude
approximation of the true image formation process
(e.g. each pixel is modeled independently and it does not account for
shadows or specularities). Such approximations are customary to PGMs as
several PGM inference techniques cannot be applied for models with complex
non-linear factors. Note that even for a relatively small image of
size $96 \times 84$, the face model contains over 48,000 latent variables
and 56,000 factors, and as we will show in Chapter~\ref{chap:cmp},
standard message passing routinely fails to converge to
accurate solutions.

\definecolor{city_1}{RGB}{128, 64, 128}
\definecolor{city_2}{RGB}{244, 35, 232}
\definecolor{city_3}{RGB}{70, 70, 70}
\definecolor{city_4}{RGB}{102, 102, 156}
\definecolor{city_5}{RGB}{190, 153, 153}
\definecolor{city_6}{RGB}{153, 153, 153}
\definecolor{city_7}{RGB}{250, 170, 30}
\definecolor{city_8}{RGB}{220, 220, 0}
\definecolor{city_9}{RGB}{107, 142, 35}
\definecolor{city_10}{RGB}{152, 251, 152}
\definecolor{city_11}{RGB}{70, 130, 180}
\definecolor{city_12}{RGB}{220, 20, 60}
\definecolor{city_13}{RGB}{255, 0, 0}
\definecolor{city_14}{RGB}{0, 0, 142}
\definecolor{city_15}{RGB}{0, 0, 70}
\definecolor{city_16}{RGB}{0, 60, 100}
\definecolor{city_17}{RGB}{0, 80, 100}
\definecolor{city_18}{RGB}{0, 0, 230}
\definecolor{city_19}{RGB}{119, 11, 32}
\begin{figure}[t]
  \scriptsize 
  \centering
 \fcolorbox{white}{city_1}{\rule{0pt}{1pt}\rule{1pt}{0pt}} Road~~
 \fcolorbox{white}{city_2}{\rule{0pt}{1pt}\rule{1pt}{0pt}} Sidewalk~~
 \fcolorbox{white}{city_3}{\rule{0pt}{1pt}\rule{1pt}{0pt}} Building~~
 \fcolorbox{white}{city_4}{\rule{0pt}{1pt}\rule{1pt}{0pt}} Wall~~
 \fcolorbox{white}{city_5}{\rule{0pt}{1pt}\rule{1pt}{0pt}} Fence~~
 \fcolorbox{white}{city_6}{\rule{0pt}{1pt}\rule{1pt}{0pt}} Pole~~
 \fcolorbox{white}{city_7}{\rule{0pt}{1pt}\rule{1pt}{0pt}} Traffic Light~~
 \fcolorbox{white}{city_8}{\rule{0pt}{1pt}\rule{1pt}{0pt}} Traffic Sign~~\\
 \fcolorbox{white}{city_9}{\rule{0pt}{1pt}\rule{1pt}{0pt}} Vegetation~~
 \fcolorbox{white}{city_10}{\rule{0pt}{1pt}\rule{1pt}{0pt}} Terrain~~
 \fcolorbox{white}{city_11}{\rule{0pt}{1pt}\rule{1pt}{0pt}} Sky~~
 \fcolorbox{white}{city_12}{\rule{0pt}{1pt}\rule{1pt}{0pt}} Person~~
 \fcolorbox{white}{city_13}{\rule{0pt}{1pt}\rule{1pt}{0pt}} Rider~~
 \fcolorbox{white}{city_14}{\rule{0pt}{1pt}\rule{1pt}{0pt}} Car~~
 \fcolorbox{white}{city_15}{\rule{0pt}{1pt}\rule{1pt}{0pt}} Truck~~\\
 \fcolorbox{white}{city_16}{\rule{0pt}{1pt}\rule{1pt}{0pt}} Bus~~
 \fcolorbox{white}{city_17}{\rule{0pt}{1pt}\rule{1pt}{0pt}} Train~~
 \fcolorbox{white}{city_18}{\rule{0pt}{1pt}\rule{1pt}{0pt}} Motorcycle~~
 \fcolorbox{white}{city_19}{\rule{0pt}{1pt}\rule{1pt}{0pt}} Bicycle~~\\
  \subfigure[Sample Image]{%
    \includegraphics[width=.45\columnwidth]{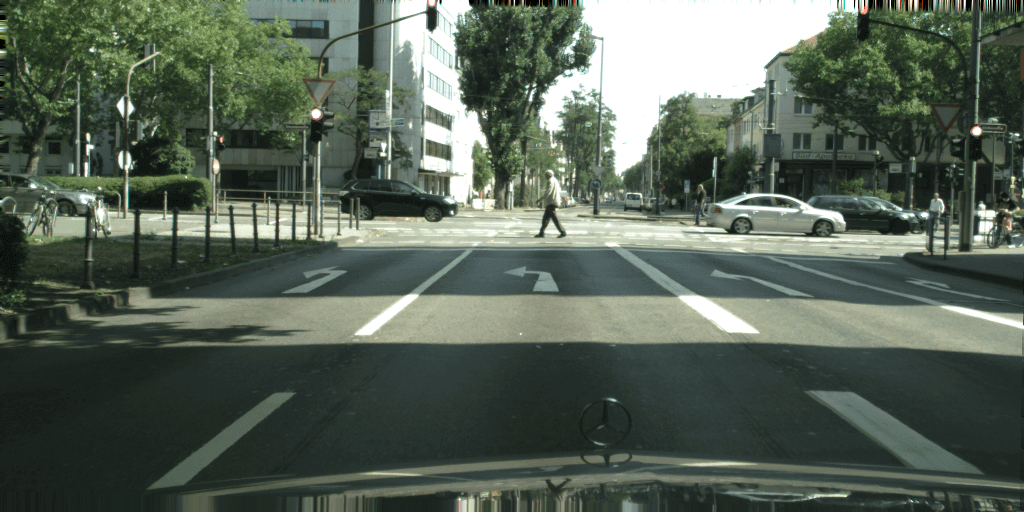}
  }\hspace{4.5mm}
  \subfigure[Ground Truth Semantics]{%
    \includegraphics[width=.45\columnwidth]{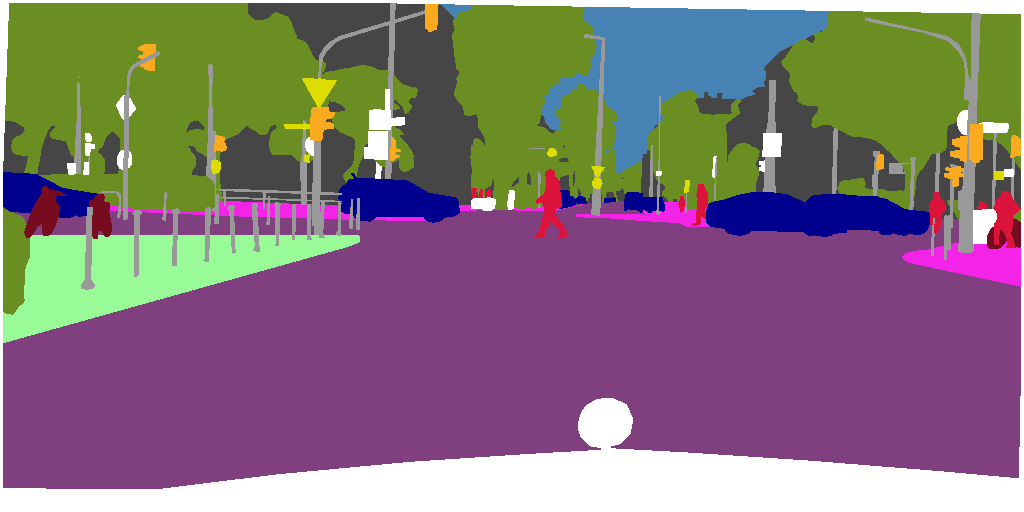}
  }
  \mycaption{Illustration of semantic segmentation task}{A sample image from
  Cityscapes street scene dataset~\cite{Cordts2015Cvprw} and the corresponding
  ground truth semantic labels.}
\label{fig:example-seg}
\end{figure}

\paragraph{Fully Connected CRFs:} Fully connected conditional random fields,
also known as DenseCRFs, are CRF models where every variable in the image is connected to
every other variable via pairwise edge potentials. For illustration purposes,
let us consider the task of semantic segmentation which is labelling each pixel
in a given image with a semantic class. See Fig.~\ref{fig:example-seg} for an
illustration. For the segmentation problem, DenseCRFs
are generally used to encode the prior knowledge about the problem:
`Pixels that are spatially and photometrically similar are more likely to have the same label'.

For an image $\obs$ with $n$ pixels, the semantic segmentation task is to produce a
labelling $\target$ with discrete values
$\{y_1,\dots,y_n\}$ in the label space $y_i\in\{1,\ldots,\mathcal{L}\}$. The DenseCRF model has
unary potentials $\psi_u(y)\in\mathbb{R}$, e.g., these can be the output of CNNs.
The pairwise potentials, as introduced in~\cite{krahenbuhl2012efficient},
are of the form $\psi^{ij}_p(y_i,y_j) = \mu(y_i,y_j) k(\f_i,\f_j)$
where $\mu$ is a label compatibility matrix, $k$ is a Gaussian kernel
$k(\f_i,\f_j)=\exp(-(\f_i-\f_j)^{\top}\Sigma^{-1}(\f_i-\f_j))$ and
the vectors $\f_i$ are feature vectors at each point. Commonly used features are
position and color values at the pixels (e.g., $\f=(x,y,r,g,b)^\top$).
In the DenseCRF model, the energy functional for an image $\obs$ thus reads:

\begin{equation}
P(\target|\obs)\propto\exp(-\sum_i\psi_u(y_i)-\sum_{i>j}\psi^{ij}_p(y_i,y_j)).
\end{equation}

Because of the dense connectivity, exact MAP or marginal inference is intractable. The main result
of~\cite{krahenbuhl2012efficient} is to derive the mean-field approximation for this model and to relate it
to bilateral filtering which enables tractable approximate inference.
As described above, mean-field approximation is a type of variational inference
where the approximate distribution $Q$ is considered to be fully-factorized
across pixels:
$Q=\prod_{i \in n} Q_i(x_i)$. Variational inference then solves for $Q$ by minimizing
the KL divergence of $P$ from $Q$(see Eq.~\ref{eqn:kl}).The work of~\cite{krahenbuhl2012efficient}
showed that the inference can be performed with efficient bilateral filtering
~\cite{aurich1995non, smith97ijcv, tomasi1998bilateral, adams2010fast} operations.
Specifially, mean-field inference results in a fixed point equation which
can be solved iteratively $t=0,1,\ldots$ to update the marginal distributions $Q_i$:

\begin{equation}
Q^{t+1}_i(x_i) = \frac{1}{Z_i} \exp(-\psi_u(x_i) - \sum_{l \in \mathcal{L}}\underbrace{\sum_{j \ne i}
\psi^{ij}_p(x_i,l) Q^{t}_j(l)}_\text{bilateral filtering}),
\label{eq:mfupdate}
\end{equation}

where $Z_i$ denotes a normalization constant and can be easily computed as $Q_i$ is
a single dimensional distribution. Although we used semantic segmentation task
for illustration purposes, DenseCRFs are shown to be useful for tackling
other tasks such as material segmentation~\cite{bell2015minc}, optical flow
estimation~\cite{sun2013fully} and intrinsic image decomposition~\cite{bell2014intrinsic}.

One of the fundamental limitations of the existing use of DenseCRFs is the
confinement of pairwise potentials $\psi^{ij}_p(y_i,y_j)$ to be Gaussian as
bilateral filtering is traditionally implemented with a Gaussian kernel. In Chapter~\ref{chap:bnn},
we show how we can learn a more general form of bilateral filters and
apply that technique for learning pairwise edge potentials in DenseCRF.

\subsection{Advantages and Limitations}
\label{sec:gen-adv-limits}

This generative modeling view is appealing as it is relatively easy to incorporate our knowledge
of physics and light transport into models and was advocated since the late
1970~\cite{horn1977imageintensities,
grenander1976patterntheory,zhu1997learning,mumford2010patterntheory,
mansinghka2013approximate,yuille2006vision}.
For example, the knowledge of how light reflects on objects with different
material properties or the knowledge of how roads and buildings are structured
are relatively easy to incorporate into generative models.
Due to incorporation of strong prior knowledge into the systems, generative
models usually work better when there is little or no data available for a
particular problem.
Since a single generative model can model different world and image
characteristics, it can be used for many different applications.
In addition, it is easier to diagnose the flaws in generative model as most
of the model is manually designed.

Despite its intuitive appeal and advantages,
in practice, generative models are used only for a few vision problems.
The few successes of the idea have been in limited settings.
In the successful examples, either
the generative model was restricted to few high-level latent
variables, e.g.,~\cite{oliver2000humaninteractions},
or restricted to a set of image transformations in a fixed reference
frame, e.g.,~\cite{black2000imageappearance},
or it modeled only a limited aspect such as
object shape masks~\cite{eslami2012shapeboltzmann},
or the generative model was merely used to generate
training data for a discriminative model~\cite{shotton2011kinect,gaidon2016virtual,
ros2016synthia,richter2016playing,shafaei2016play}.
With all its intuitive appeal, its beauty and simplicity, it is fair to say
that the track record of generative models in computer vision is poor.
As a result, the field of computer vision is now dominated by
efficient but data-hungry discriminative models,
the use of empirical risk minimization for learning,
and energy minimization on heuristic objective functions for inference.

Why did generative models not succeed? There are two key problems that
need to be addressed, the design of an accurate generative model, and
the inference therein.
The first key problem which is the design of accurate generative model is partly
addressed by recent advances in graphics. Although modern graphics provide
rendering with stunning level of realism, priors of world parameters are difficult
to characterize. This results in complex priors together with more complex
forward models for accurate generative models which in turn results in
difficult inference.

This brings us to the second key problem in the generative world view which is
the difficulty of posterior inference at test time.
This difficulty stems from a number of reasons:
\emph{first}, the target variable $\target$ is typically high-dimensional and so is
the posterior.
\emph{Second}, given $\target$, the image formation process realizes complex and
\emph{dynamic} dependency structures, for example when objects occlude or
self-occlude each other. These intrinsic ambiguities result in multi-modal
posterior distributions.
\emph{Third}, while most renderers are real-time, each simulation of the
forward process is expensive and prevents exhaustive enumeration. Overall,
the limitations of generative approaches out-weigh their advantages making
them not succeed in building practical computer vision systems.

Despite these limitations, we still believe in the usefulness of generative
models in computer vision, but argue that we need to leverage existing discriminative
or even heuristic computer vision methods for alleviating some of the difficulties in the
posterior inference. Inference techniques proposed in this thesis are steps
in this direction.

\section{Discriminative Vision Models}

With the advances in internet and image capturing technology, there is an explosive
growth of visual data during the last few years. Moreover, presence of crowd-sourcing
platforms like `Amazon Mechanical Turk'~\cite{mturk} make it easier to annotate large
amounts of data by millions of people. Discriminative models directly model the
contingency of world properties on the observed data $P(\target|\obs)$.
Unlike generative models, discriminative
models are task-specific and learning takes the central role in defining the model.
Discriminative models comprise of functions directly approximating
the posterior distribution $P(\target|\obs, \params)$, where $\params$ denote
the parameters of the model. Supervised learning with annotated training data
is usually employed to fit the model parameters to the given task.
Since discriminative models directly characterize
the posterior distribution, inference is reduced to simple evaluation of
the model.

Due to the availability of large amounts of training data and the computing power
that can handle rich high-capacity models, discriminative models have been very
successful in many vision problems. In addition, inference is fast
since this involves a simple evaluation of the model. This makes discriminative
models particularly attractive for many practical applications. Many
mathematical functions that are rich enough to capture the relationship between
the observation and target variables can be used as discriminative models.
Hence, many types of discriminative models have been used in the computer vision
literature.

Discriminative models are traditionally partitioned into two modules:
\textit{feature extraction} and \textit{prediction} modules.
Before the advent of modern convolutional neural networks (CNNs),
these two components are studied separately in the literature.
We briefly discuss these two components in the discriminative models.

\paragraph{Feature Extraction:}
Depending on the type of vision task, features are extracted either at all
pixels (points) in the observed data or only at some key points. For example,
registering two images taken from different view-points requires finding
corresponding points (key points) in each image and then matching.
For such tasks, an additional step of key point detection is required before
feature computation. For image classification, a single feature vector
is extracted for the entire image.

An ideal feature representation should be compact, efficient to compute
and invariant to specific transformations. As an example,
for semantic segmentation, features should be invariant to intra-class
variations such as illumination, scale, rotation, object articulations, etc.,
while being sensitive to changes across different semantic categories.
Several feature extraction schemes have
been proposed in the vision literature, most of them are hand crafted.
Some popular choices include SIFT~\cite{lowe1999object}, HoG~\cite{bay2006surf},
SURF~\cite{dalal2005histograms}, DAISY~\cite{tola2008fast}, etc.
Models for feature extraction and prediction are plentiful and discussing
all of them is outside the scope of this thesis.
With the recent advances in CNNs, feature extraction is coupled with
prediction which are learned together end-to-end.

\paragraph{Prediction:}
Once the image features are extracted, the task is to estimate the
posterior distribution $P(\target|\f(\obs))$, where $\f(\obs)$ denotes the features.
A common strategy is to learn a rich parametric or non-parametric model with
supervised learning techniques. This makes the availability of training data
crucial for discriminative approaches. Several learning based prediction
models have become popular in tackling vision tasks including support vector
machines (SVM)~\cite{cortes1995support},
boosting~\cite{schapire1990strength, freund1995desicion},
random forests~\cite{breiman2001random,ho1995random},
deep convolutional neural networks~\cite{lecun1998gradient},
etc. Refer to~\cite{friedman2001elements}
for a review of different prediction techniques.
Next, we briefly review random forests and CNN models as we either make use of
or propose extensions to these models in this thesis.

\subsection{Random Forests}
\label{sec:forests}

Random forests~\cite{breiman2001random,ho1995random}
are an ensemble of $K$ randomly trained prediction (classification or regression)
trees, where each tree $T(\target|\f(\obs),\params^k)$ represents a non-linear
function approximating the posterior $P(\target|\f(\obs))$.
The trees are typically binary trees and can be viewed as performing a
discriminative hierarchical clustering of the feature space.
And a simple model fit (e.g.,\ linear model) is used in each cluster.

Trees are grown incrementally from the root node to the leaves and
each node represents a partition of the feature space. These partitions can be any linear
or non-linear functions, but the simple axis-aligned partitions are the most used ones
due to their simplicity and efficiency. For simplicity, let us assume the partition functions
are axis-aligned. At each node, a feature $\kappa$ and
its split value $\tau$ are chosen to split the feature space,
so as to minimize an energy function $E$. Let us
consider training the $j^{th}$ node in a $k^{th}$ tree. Let all the data points falling
in that node be $\mathcal{S}_j$ (due to splitting of its ancestor nodes)
and $\mathcal{T}_j$ denotes the discrete set of
\textit{randomly} selected feature axes $\{(\kappa,\tau)_i\}$
(feature indices and their corresponding values)
for the node $j$. Training the $j^{th}$ node corresponds to choosing the optimal
split $\theta^k_j \in \mathcal{T}_j$ among the randomly chosen splits that
minimizes an energy function $E$:

\begin{equation}
\theta^k_j = \argmin_{\gamma \in \mathcal{T}_j} E(\mathcal{S}_j,\gamma)
\end{equation}

Depending on the type of task and data, different energy functions $E$ are
used. Each split $\gamma$ partitions the training data $\mathcal{S}_j$ in
the node $j$ into two parts $\mathcal{S}^L_j$ and $\mathcal{S}^R_j$ which are
assigned to left and right child nodes respectively.
A common energy function measures how well a regression/classification model
fit the data in each of the left and right child nodes created by a split $\gamma$:

\begin{equation}
E(\mathcal{S}_j,\gamma) = - (M(\mathcal{S}^L_j, \beta) + M(\mathcal{S}^R_j, \beta)).
\label{eqn:forest_energy}
\end{equation}

Where $M(\mathcal{S}_j,\beta)$ denotes the model likelihood i.e., how well the model with parameters
$\beta$ can explain the data $\mathcal{S}_j$.
For example, in the case of regression tasks, $M$ can be a linear regression fit
and in the case of classification
(like in semantic segmentation), $M$ can be the classification accuracy.
Like this, the trees are recursively grown by splitting the leaf nodes into left and right
child nodes. The set of all node splits $\theta^k=\{\theta^k_j\}_{j=1,\cdots,J}$ represents
the parameters of the $k^{th}$ tree.
Once a tree is trained, a simple prediction
model is fitted to the data in the leaf nodes. A deep tree might overfit the data
and a shallow tree would under-fit the data and miss important structure. Restricting
the tree size corresponds to regularizing the model and size should be adaptively
chosen based on the training data. Some training stopping criteria include setting the
maximum depth of the trees; minimum number of data points in each node;
a threshold for energy function $E$, etc.

Random forests are distinguished from other tree-based
supervised learning techniques such as boosted decision trees, by the way different
trees are trained in a forest. Each tree in a random forest is trained independently
and randomness is added either in terms of choosing a random subset of training data for each
tree (called \textit{bagging}~\cite{breiman1996bagging}) and/or randomly
choosing the split candidates (feature indices and their values) at each node.
Typically, the estimates across the trees $T(\target|\f(\obs),\params^k)$
are averaged to get the final model $P(\target|\f(\obs))$:

\begin{equation}
P(\target|\f(\obs)) = \frac{1}{K}\sum_{k} T(\target|\f(\obs),\params^k).
\end{equation}

Due to the randomness, different trees are identically distributed
resulting in a low-variance estimate when the final estimate is taken as the average
across the trees.
Random forests are highly flexible and several different types of models are
conceivable with using a combination of different splitting criteria.
Due to their simplicity and flexibility, random forests have become a popular
choice for supervised learning in vision.
Random forests are easy to implement and train. Also, they can be easily adapted
to a wide range of classification and regression tasks with relatively simple
changes to the model. Moreover, they are non-parametric in
nature with the ability to consume large amounts of training data.
Random forests are successfully applied for vision tasks such as
human pose estimation~\cite{shotton2011kinect}, semantic segmentation~\cite{shotton2008semantic}, etc.
In the case of semantic segmentation, a popular model is to extract TextonBoost
features~\cite{shotton2006textonboost}
at each pixel and then train a random forest classifier to predict the class
label at each pixel.
One of the crucial advantages of random forests with respect
to neural networks is that the loss function $E$ need not be differentiable.
In Chapter~\ref{chap:infsampler}, we use random forest models
to improve inference in inverse graphics via our informed
sampler approach. In Chapter~\ref{chap:cmp}, we use random forests
for predicting messages resulting in improved variational inference in
layered graphical models. Refer to~\cite{Criminisi2013} for a comprehensive
overview of random forests and their applications in computer vision and medical
image analysis.

\subsection{Convolutional Neural Networks}
\label{sec:cnn}

Neural networks are a class of models with complex parametric non-linear
functions relating the input $\obs$ to the target $\target$. The complex non-linear
function is usually realized by stacking a series of simple and differentiable
linear and non-linear functions:

\begin{equation}
P(\target|\obs,\params) = \func_1(\func_2(\cdots \func_k(\obs,\theta_k)\cdots, \params_2),\params_1).
\end{equation}

Learning involves finding the parameters $\{\params_1, \params_2, \cdots, \params_k\}$
that best approximates the desired relationship between the input and target variables.
The component functions are usually simple linear functions such as convolutions
$\func(\mathbf{s},\theta)=\mathbf{W(\params)}\mathbf{s}+b$ (where $\mathbf{s} \in \mathbb{R}^q$,
$\mathbf{W} \in \mathbb{R}^{p \times q}$) interleaved
with simple non-linear functions such as rectified linear units (ReLU) $\func(\mathbf{s})=max(0,\mathbf{s})$.
A linear function together with a non-linearity is usually called a single layer
in the network. Intermediate layers in a neural network are also called \textit{hidden}
layers and the number of units in intermediate layers determine the \textit{width}
of the network.
A theoretical result~\cite{csaji2001approximation,hornik1991approximation}
is that any complex continuous function can be approximated by a simple two
layered neural network, given sufficient number of intermediate units (width of
the network).
From a practical point of view, neural networks are attractive because of their
fast inference (simple forward pass through the network) and an end-to-end prediction
(going from input to output variables without the intermediate handcrafted feature
extraction) capabilities.

Convolutional neural networks (CNN) are a special class of neural networks
tailored for processing 2D or higher dimensional visual data on a grid.
The main characteristic of CNNs is the use of spatial convolutions instead of
\textit{fully-connected} matrix-vector multiplications for building linear
functions. This greatly reduces the amount of parameters due to parameter sharing
across different spatial locations and speeds up the
network computation and training. One of the main hurdles for the success of
CNNs was the lack of computational resources required to train models
with millions of parameters. Recent availability of large datasets together
with efficient model and training implementations in GPUs made it possible
to successfully apply CNNs to real-world vision tasks. Since CNNs typically have
millions of parameters, they are highly prone to overfit the training data.
Advances in simple yet powerful regularization techniques (such as
DropOut~\cite{srivastava2014dropout}) are another reason for the successful deployment of
CNN models. Currently, CNNs are state-of-the-art
in many traditional vision problems such as image classification~\cite{he2015deep,krizhevsky2012imagenet},
object detection~\cite{girshick2014rich,redmon2015you,ren2015faster},
semantic segmentation~\cite{long2014fully,chen2016deeplab}, etc.

\begin{figure}[t!]
 \centering
 \includegraphics[width=0.9\columnwidth]{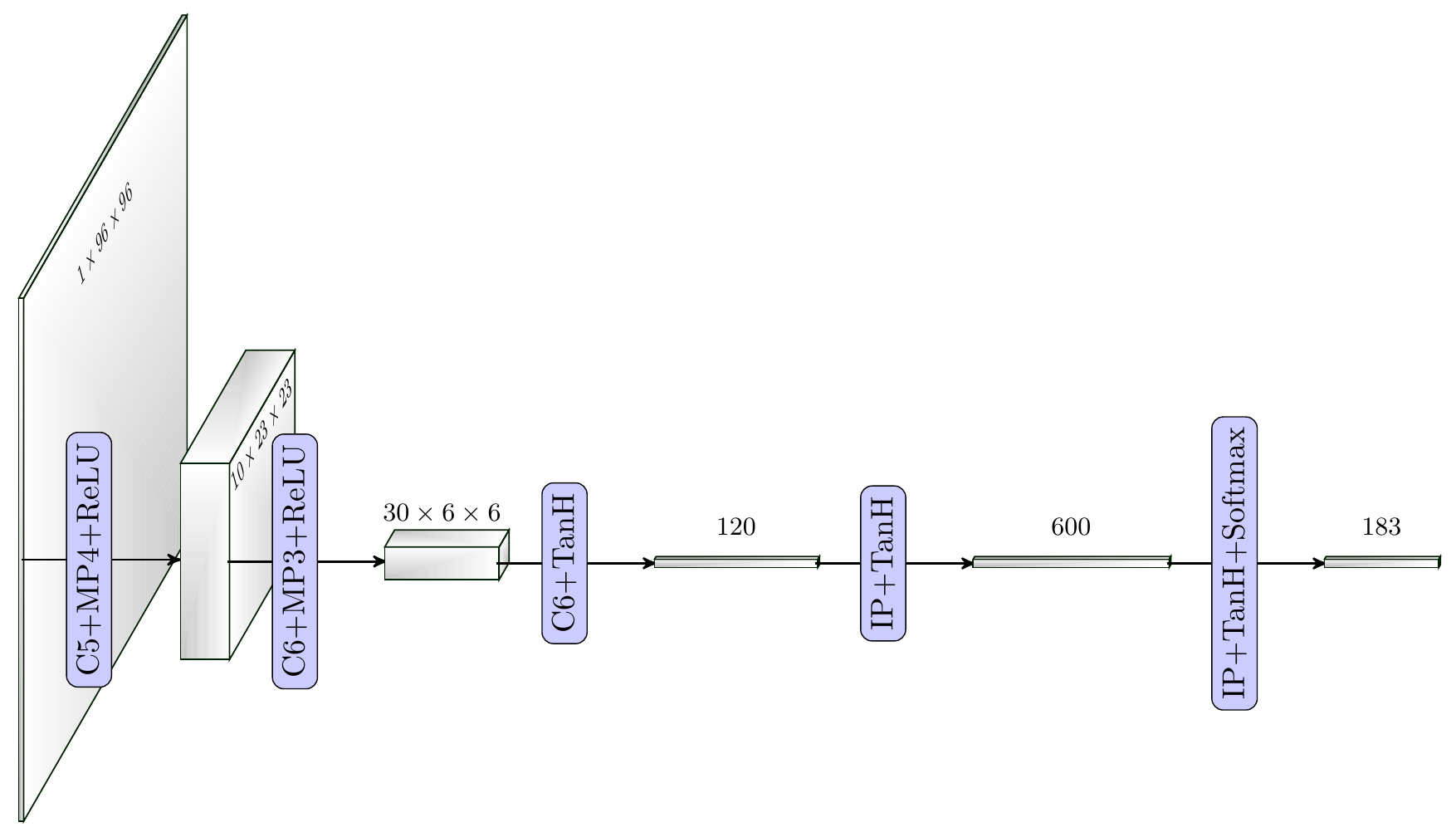}
 \mycaption{Sample CNN architecture for character recognition}
 {LeNet-7~\cite{lecun1998mnist} architecture generally used for character recognition
 (used for Assamese character recognition in Chapter~\ref{chap:bnn}).
 `C$n$' corresponds to convolution layer with $n\times n$ filters; `MP$n$' corresponds
 to max-pooling layer with window size $n$; `IP' corresponds to fully-connected
 inner-product layer; `ReLU' and `TanH' corresponds to rectified linear units
 and tanh non-linear layers and `Softmax' layer produces probabilities for 183
 output classes.}
\label{fig:lenet7}
\end{figure}

CNN architectures are typically composed of the following layers: Convolution,
pooling, non-linearity, fully-connected (FC) and loss layers.
Convolution layers are simple spatial convolutions,
pooling layers do spatial downsampling
and FC layers connect each output unit to all input units. Non-linear layers
are simple yet important functions that model non-linearities in the CNN
model. Some popular non-linear functions include ReLU, TanH, sigmoid function,
etc. Loss layers are problem specific layers that are used at the end of the
network and implement the differentiable empirical loss $\sigma(\hat{\target}, \target^*)$
between the predicted target $\hat{\target}$ and the ground truth target $\target^*$.
Figure~\ref{fig:lenet7} shows a sample CNN architecture generally used for character
recognition. The input is a grayscale image $\obs$ with the size $96\times96$ and the
output is a vector of probabilities for each class $P(\target|\obs)$. Also shown
are the sizes of intermediate CNN representations.

Training the parameters $\params_k$ of each layer involves back-propagating
the empirical loss from the loss layers backwards to the early CNN layers.
To avoid over-fitting to the training data, the loss is usually augmented
with a regularization over the network parameters. The optimization objective
for a given dataset with $m$ training instances is given as an average loss $L$:

\begin{equation}
L(\params) = \frac{1}{m} \sum_{i=1}^{m} \sigma(\hat{\target_i}, \target_i^{*}) + \lambda r(\params),
\end{equation}

where $r$ denotes the regularization over the parameters $\params$ with weight $\lambda$.
Then the parameters $\theta$ are updated using gradient descent methods
such as stochastic gradient descent (SGD), AdaDelta~\cite{zeiler2012adadelta},
Adam~\cite{kingma2014adam}, etc.
The parameter update steps to update a single parameter $\params^i \in \params$ in SGD are given as:

\begin{align}
\begin{split}
v_{t+1} &= \mu v_t - \gamma \nabla L(\params^i_t)
\\
\params^i_{t+1} &= \params_t + v_{t+1}
\end{split}
\end{align}

where $v, \gamma, \mu \in \mathbb{R}$,
$v_t$ denotes the parameter update in the previous step and
$\nabla L(\params^i_t)$ denotes the gradient of loss $L$ with respect
to the parameter $\params^i$. Thus the parameter update $v_{t+1}$ is a weighted
combination of the previous update and the negative gradient of loss $L$.
The weights $\mu$ and $\gamma$ are called \textit{momentum} and
\textit{learning rate} respectively which are generally chosen to obtain
good performance on a given validation data.
In practice, since the size of the dataset $m$ is large, only a small subset (batch)
of dataset is used for computing the loss and updating parameters in each step.

Instead of computing the gradients of loss $L$ with respect to all the network parameters
in one step, the gradients are back-propagated across the layers.
Thus, one of the fundamental requirements for a component function (except the first layer)
in CNNs is that it should be differentiable with respect to both its inputs
and its parameters. Once the network is trained,
inference is a simple forward pass through the network. Like in many discriminative
models, inference in CNNs amounts to evaluation of the model.

In Chapter~\ref{chap:bnn}, we generalize the standard spatial convolutions found
in CNNs to sparse high-dimensional filters and in Chapter~\ref{chap:binception},
we propose an efficient CNN module for long-range spatial propagation of information
across intermediate CNN representations. There are several other types of
neural networks, such as recurrent neural networks,
that are currently being used in the computer vision community.
Refer to~\cite{Goodfellow-et-al-2016-Book,888} for more details regarding CNNs or
neural networks in general.

\subsection{Advantages and Limitations}

Discriminative models are mainly data-driven methods where inference amounts to
simple evaluation of the model. In general, discriminative models are
fast, robust to model mismatch and are also high performing when
trained with enough amounts of data. Discriminative models are attractive
because of their practical utility and also their flexibility in terms of
being able to use same model architecture and training for different vision tasks.

\begin{figure}[t!]
 \centering
 \includegraphics[width=0.9\columnwidth]{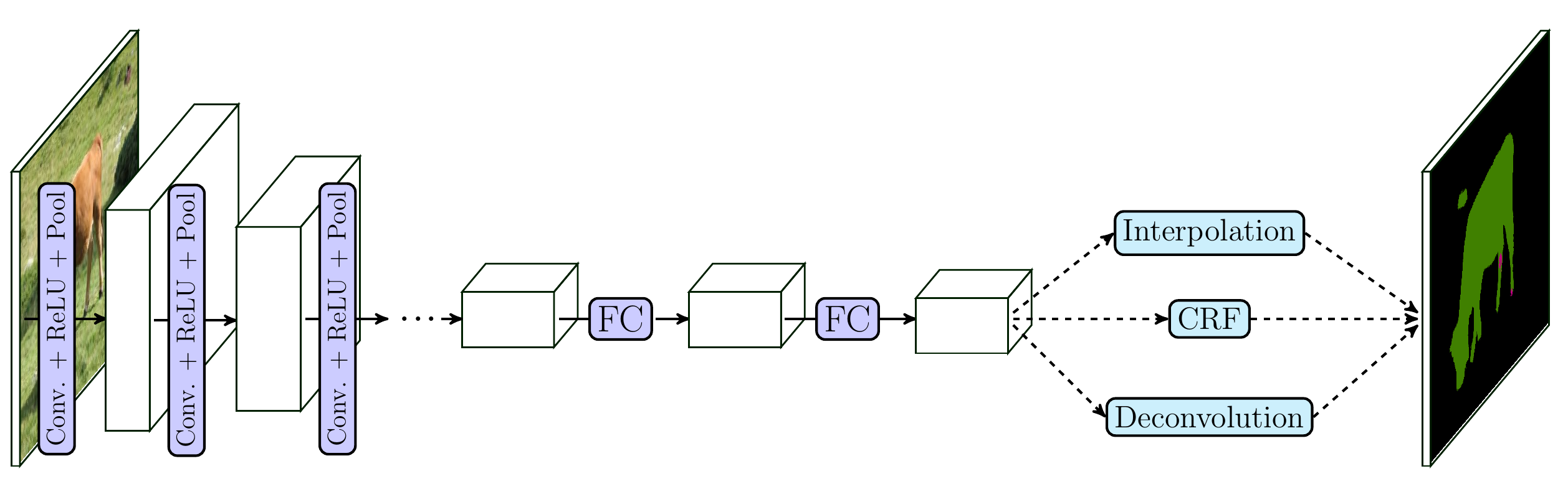}
 \mycaption{Sample CNN architecture for semantic segmentation}
 {Semantic segmentation CNN architectures typically consists of convolution (Conv.),
 pooling (Pool) and $1\times1$ convolution layers (FC) interleaved with non-linearities (ReLU).
 The use of pooling results in lower resolution CNN output which is generally up-sampled
 with either interpolation, deconvolution and/or CRF techniques. CRF techniques also help
 in incorporating prior knowledge about semantic segmentation.}
\label{fig:seg-net}
\end{figure}

On the other hand, discriminative models also have several limitations.

\begin{itemize}
\item Discriminative models are data hungry and typically fail to work where
there is little data available. Availability of large datasets for several vision
tasks and online annotation tools like Amazon Mechanical Turk~\cite{mturk} help
in mitigating this limitation.
\item Since discriminative models tend to have a large number of parameters
which are directly learned from data, when the performance is not as expected,
it is difficult to find the cause of the problem and accordingly modify the models.
As a result, there are no guaranteed ways
to find the right model architecture for a given task.
These problems are generally handled with either regularizations on model complexity
or with trial-and-error strategy on various model architectures.
\item One of the fundamental limitations of discriminative models is the lack of key
approaches to inject prior knowledge into the models. This is especially true
in the case of end-to-end trained models like CNNs. In the case of hand crafted
features, we can inject some prior context in the form of image or pixel features.
It is not easy to inject the knowledge of generative models into discriminative
approaches such as CNNs. For example, in the case of semantic segmentation,
post-processing steps such as DenseCRFs are generally employed to model the
relationship (prior knowledge) between the pixels and output labels.
Figure~\ref{fig:seg-net} shows a prominent CNN architecture for semantic segmentation.
\item Discriminative models are task-specific and a single trained model can not
be easily transferred to other vision problems.
Although several transfer learning techniques~\cite{pan2010survey} exists that
can help transfer the knowledge across different discriminative models,
they are generally task specific and have limited success.
One of the main advantages of CNNs, in comparison to other discriminative models,
is that a CNN trained on image classification task is shown to perform reasonably well on other related
tasks such as semantic segmentation, object detection etc. with only minor adaptions
to the new task.

\end{itemize}

Recently, hybrid models combining generative and discriminative models are proposed
to alleviate some of the limitations in both and make use of their complementary
advantages. We will discuss these models in the next section.

\section{Combining Generative and Discriminative Models}
\label{sec:gen-disc-comb}

As we have argued in the previous sections, generative and discriminative
models have complementary advantages and limitations. Generative models have the
advantage of incorporating prior knowledge while being slow; whereas discriminative
models are fast and robust but it is difficult to incorporate prior knowledge.
Typically, generative models suffer from high bias due to model mismatch, whereas
discriminative models suffer from higher variance.
In general, discriminative models work well and are robust to model
mismatch when the available annotated training data is large. If the
available data is small in comparison to the required model complexity, we need
ways to constrain model parameters with the use of prior knowledge. Generative
models provide principled ways to incorporate such prior knowledge and can even
make use of unlabelled data which is generally abundant.
The work of~\cite{jordan2002discriminative} is one of the first comparative studies on generative and
discriminative models resulting in the common knowledge of using discriminative
models when the data is abundant, otherwise use generative approach for
a given problem.

We hypothesize that combining generative and discriminative
models can leverage the advantages in both. At the same time, combining these complementary
models can also bring forward the limitations in both. The generative and discriminative
models are often studied in isolation, but during the past decade,
several synergistic combinations of generative and discriminative models have been proposed.

There are 3 ways in which generative and discriminative models can be combined.
1. Use a generative model to improve the model or the inference in the discriminative model
(indicated as `Generative $\rightarrow $ Discriminative'); 2. Use a discriminative model for
improving the model and/or inference in the generative model (indicated as
`Discriminative $\rightarrow $ Generative'); and 3. Hybrid generative and discriminative
models (indicated as `Generative $\leftrightarrow $ Discriminative').

\subsubsection{Generative $\rightarrow $ Discriminative}

One way to use generative models for improving discriminative models is by
feature extraction using generative models. The work of~\cite{jaakkola1999exploiting} showed that
the gradients of the generative models
can be used as features in discriminative models.
The gradients of a generative model are called `Fisher vectors' and are
particularly useful for building kernel functions (Fisher kernels)
that can be used in kernel based techniques such as SVMs.

Another popular way to incorporate generative prior knowledge
is to provide prior constraints while training discriminative models.
For instance, CNN models can be trained
with extra loss layers encoding the prior relationship between the output
variables. The overall training loss is a combination
of the discriminative prediction loss and also a generative prior loss.
A related strategy for training CNNs is to first train a discriminative CNN
using generative prior loss with large amounts of unlabelled data,
and then fine-tune the network using discriminative prediction loss with
limited labelled data. Instead of training a single discriminative
model with prior constraints,~\cite{tu2010auto} proposed to train a sequence of
discriminative predictors, each taking as input not only the input features
but also features from previous stage predictions. This way, it is easy to
incorporate the prior constraints on output target variables using features
extracted on predictions. This technique is called `Auto-Context'~\cite{tu2010auto} and
the sequence of predictors is usually trained with stacked generalization
method~\cite{wolpert1992stacked}. Despite being simple, auto-context method
is shown to be powerful and useful in many vision problems
(for e.g.,~\cite{tu2010auto,jiang2009efficient,jampani15wacv}).

More recently, structured prediction layers~\cite{ionescu2015matrix,zheng2015conditional,schwing2015fully,chandra2016fast}
are introduced into
discriminative CNN frameworks. These layers are mainly adapted from the models and inference
techniques in generative models. For example, in the case of semantic segmentation
CNNs, mean-field inference in fully-connected CRFs can be formulated as
recurrent neural network modules~\cite{domke2013learning,zheng2015conditional}
and is used to augment the existing
CNN architectures resulting in better performance. The work of~\cite{chandra2016fast}
proposed a way to incorporate Gaussian CRFs into end-to-end trained semantic
segmentation CNNs. In Chapter~\ref{chap:bnn}, we generalize the standard spatial convolutions
in CNNs to sparse high-dimensional filters and show it can be used to incorporate
structured prior knowledge into CNNs resulting in better model and inference.
In Chapter~\ref{chap:binception}, we propose a specialized structured prediction
module to be used in CNNs for dense pixel prediction tasks such as semantic
segmentation.

\subsubsection{Discriminative $\rightarrow $ Generative}

Although generative models provide an elegant formalism to encode prior
knowledge about the problem, their use is mainly hampered by the difficulty in
the posterior inference. Since discriminative models directly characterize the posterior
distribution, they have the potential to be useful for inference in a corresponding
generative model.

Since many of the generative models require approximate
Bayesian inference for estimating the posterior distribution, some components of
the Bayesian inference can be completely replaced with discriminative models.
\textit{Inference machines}~\cite{Ross2011} are a successful example of such technique.
Inference machines pose the message passing inference in a given generative model
as a sequence of computations that can be performed efficiently by training
discriminative models like random forests. Instead of learning complex potential
functions and computing messages between the variables, discriminative predictors
that directly learn to pass the messages are proposed. This technique
is shown to perform well on real world tasks~\cite{Ross2011,ramakrishna2014pose,shapovalov2013spatial}
such as human pose estimation, 3D surface layout estimation and 3D point cloud estimation.
Inference machines help bridging the gap between the message
passing and random forest techniques. Similar technique~\cite{Heess2013} is also shown to be
useful to predict messages for
expectation propagation~\cite{Minka2001} inference in generative models. More recently,
~\cite{lin2015deeply} proposed to use discriminative deep learning models for predicting
messages in message passing inference.

By completely replacing the components of Bayesian inference with discriminative predictors,
we lose the theoretical guarantees from the original inference
techniques. However,
discriminative models can still be used to improve the inference process.
Data driven Markov chain Monte Carlo (DDMCMC)~\cite{tu2002image} methods
leverage discriminative models to speed up the MCMC inference.
DDMCMC methods have been used in image segmentation~\cite{tu2002image}, object
recognition~\cite{zhu2000integrating}, and human pose estimation~\cite{lee2004proposal}.
In Chapters~\ref{chap:infsampler} and~\ref{chap:cmp}, we propose
principal techniques for leveraging discriminative models for Bayesian
inference in inverse graphics and layered graphical models respectively.

Another way of using discriminative models to improve
generative approaches is to use discriminative prediction loss for training
the generative model parameters. This is called `discriminatively training
generative models'~\cite{bouchard2004tradeoff,holub2005discriminative,yakhnenko2005discriminatively}
and is akin to using a generative prior loss for training discriminative models.
Such models are also called hybrid models~\cite{lasserre2006principled} (discussed
more below) if different parameters are used for defining discriminative and
generative models.

\subsubsection{Generative $\leftrightarrow $ Discriminative}

It is possible to define both discriminative and generative models for the same
task and train them together. This synergistic training
can help in a better model fit in both. With
such hybrid models, it is possible to  train with both unlabelled and labelled
data together~\cite{lasserre2006principled,minka2005discriminative}.

Recent advances in deep learning showed that neural network models can also be
used as good approximators for generative models of images
(for e.g.,~\cite{dosovitskiy2015learning,gregor2015draw,theis2015generative}). Thus, it is
possible to define a hybrid model with different neural networks approximating
the corresponding generative and discriminative models for a task, and then train them together.
One popular model in this category is `Auto-encoding variational Bayes'
~\cite{kingma2013auto,rezende2014stochastic}. Here,
a generative model with exponential family distributions is approximated with
a neural network. At the same time, variational Bayesian posterior inference in that model is
approximated with a different neural network. Both generative and discriminative
(inference) networks are trained by minimizing the variational lower bound (Eq.~\ref{eqn:kl}).
The work of ~\cite{eslami2016attend} uses such hybrid models with recurrent neural networks
and an attention mechanism to tackle vision problems involving multiple unknown number of objects in an image.
These models are shown to perform well on small scale vision problems such as
character recognition and are not scaled for tackling mainstream vision problems.
The formulation of such hybrid models is elegant and has potential to be useful
for many vision tasks.

Very recently, in a similar spirit to auto-context,~\cite{saining16} proposes to learn
a top-down CNN for capturing the contextual relationships between the
target variables. The top-down generative CNN learns to predict the target variables
from the surrounding target variables (context).
This top-down CNN is then coupled with original discriminative
CNN to serve as top-down constraints for intermediate CNN representations.
As an advantage over the auto-context framework where different models are learned
at different stages, a single discriminative model is learned and shown to be sufficient.

Hybrid generative and discriminative CNN models are a very active area of
research with different architectures being proposed frequently.
We only discussed a few model architectures here.
It is plausible that
in the near future, hybrid generative and discriminative models dominate
the field of computer vision.

\section{Discussion and Conclusions}

In this chapter, we have discussed various generative and discriminative computer vision models.
Models form the core of any computer vision system,
we choose to discuss some prominent models in relation
to this thesis while highlighting the advantages and limitations in
popular generative and discriminative models.

Generative models are conceptually elegant to incorporate prior knowledge about
the task while their use is mainly hampered by the difficulty of posterior inference.
Discriminative models, on the other hand, are robust to model mismatch while
being fast but require large amount of labelled data and there is a lack of standard
approaches for incorporating prior knowledge into them. Hybrid generative and
discriminative models, discussed in the previous section, try to bridge the gap
between these two complementary approaches.

The main aim of this thesis is to improve inference with in various computer
vision models. In Part I of the thesis, we concentrate on improving inference
in generative vision models. We do this by learning separate discriminative models
and propose algorithms for better inference in prominent generative models in
vision namely inverse graphics models (Chapter~\ref{chap:infsampler})
and layered graphical models (Chapter~\ref{chap:cmp}).

In Part II of the thesis, we concentrate on improving inference in discriminative
vision models. Since inference is simple evaluation of the model in discriminative models, we
propose techniques for modifying the model itself enabling the introduction of
prior knowledge into CNN models. Specifically, we generalize the standard spatial convolutions
in prominent CNN models to sparse high-dimensional filtering (Chapter~\ref{chap:bnn}) and then
propose a neural network approach for propagating information across video frames (Chapter~\ref{chap:vpn}).
In Chapter~\ref{chap:binception}, we propose a new CNN module that can be added to existing
segmentation CNN architectures that helps in image adaptive filtering of intermediate CNN units.

\part{Inference in Generative Vision Models}
\chapter{The Informed Sampler}
\label{chap:infsampler}

\newcommand{\INDLMH}{INF-INDMH\xspace}
\newcommand{\MIXLMH}{INF-MH\xspace}
\newcommand{\INFBMHWG}{INF-BMHWG\xspace}
\newcommand{\REGMH}{REG-MH\xspace}
\newcommand{\REGAMH}{REG-MH\xspace}
\newcommand{\MIXLMHRF}{INF-RFMH\xspace}
\newcommand{\MH}{MH\xspace}
\newcommand{\MHWG}{MHWG\xspace}
\newcommand{\PT}{PT\xspace}
\newcommand{\BMHWG}{BMHWG\xspace}
\newcommand{\obsI}{\hat{I}}

\newcommand{\thetaProp}{\bar{\theta}}

In the previous chapter, we briefly discussed the generative models (Section~\ref{sec:gen-models})
with their advantages and limitations. With all its intuitive appeal, beauty
and simplicity, it is fair to say that the track record of generative models
in computer vision is poor, which is mainly due to the difficulty of posterior inference.
As a result the computer vision community has favored efficient discriminative
approaches.
We still believe in the usefulness of generative models in computer vision,
but argue that we need to leverage existing discriminative or even heuristic
computer vision methods.
In this chapter, we implement this idea in a principled way with an
\emph{informed sampler}, which is a mixture sampling technique,
 and in careful experiments demonstrate its use on
challenging generative models which contain renderer programs as their components.
We concentrate on posterior inference in `inverse graphics' models which is briefly
described in Section~\ref{sec:inv-graphics}.
With experiments on diverse `inverse graphics' models,
we show that the informed sampler, using simple discriminative proposals based on existing
computer vision technology, achieves significant improvements of inference.

\section{Introduction}
\label{sec:introduction}

\begin{figure}[t]
\begin{center}
\centerline{\includegraphics[width=0.7\columnwidth]{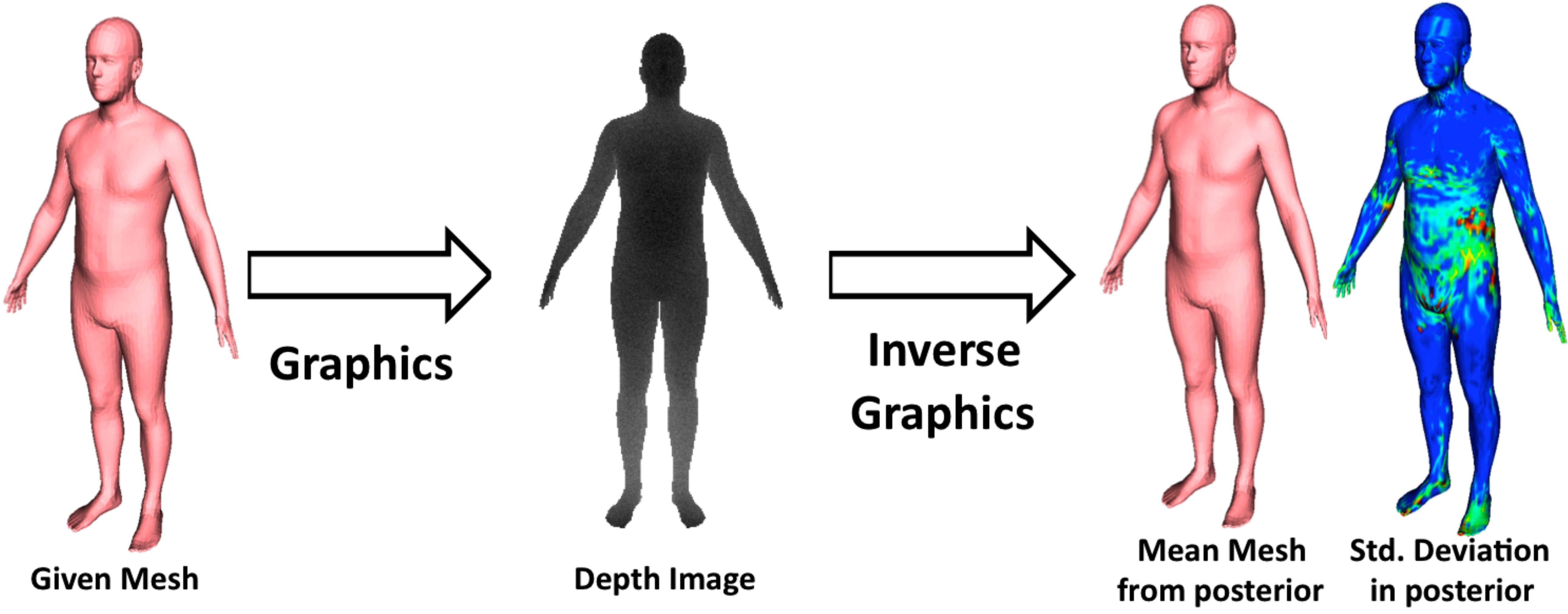}}
\mycaption{An example `inverse graphics' problem}{A graphics engine renders a 3D
  body mesh and a depth image using an artificial camera. By `Inverse
  Graphics', we refer to the process of estimating the posterior
  probability over possible bodies given the depth image.}
\label{fig:teaser}
\end{center}
\end{figure}

As discussed in Section~\ref{sec:inv-graphics}, modern computer graphic systems
that leverage dedicated hardware setups produce a stunning level of realism with
high frame rates. We believe that these systems will find its way in
the design of generative models and will open up exciting modeling
opportunities.
This observation motivates the research question of this chapter, the
design of a general inference technique for efficient posterior
inference in accurate computer graphics systems. As such it can be
understood as an instance of \emph{Inverse Graphics}~\cite{baumgart1974inversegraphics},
which is briefly discussed in Section~\ref{sec:inv-graphics} and
illustrated in Fig.~\ref{fig:teaser} with one of our applications.

The key problem in the `inverse graphics' is the difficulty of
posterior inference at test-time.
This difficulty stems from a number of reasons as outlined in
Section~\ref{sec:gen-adv-limits}.
Our aim in this work is to devise an inference technique that is general
and allow reuse in several different models and novel scenarios.
On the other hand we want to maintain correctness in terms of the probabilistic
estimates that they produce. One way to improve on inference
efficiency in generative models is to leverage existing computer vision
features and discriminative models.
In this chapter, we propose the \emph{informed sampler}, a Markov chain
Monte Carlo (MCMC) method with discriminative proposal distributions.
It can be understood as an instance of a data-driven MCMC
method~\cite{zhu2000integrating}, and our aim is to design a method that
is general enough such that it can be applied across different
problems and is not tailored to a particular application.

During sampling, the informed sampler leverages computer vision
features and discriminative models to make informed proposals for the state
of latent variables and
these proposals are accepted or rejected based on the generative model.
The informed sampler is simple and easy to implement, but it enables inference
in generative models that were out of reach for current \emph{uninformed}
samplers.
We demonstrate this claim on challenging models that incorporate
rendering engines, object occlusion, ill-posedness, and
multi-modality. We carefully assess convergence statistics for the
samplers to investigate their correctness about the probabilistic
estimates.
Our informed sampler uses existing computer vision technology such as
histogram-of-gradients features (HoG)~\cite{dalal2005histograms}, and the OpenCV
library,~\cite{bradski2008opencv}, to produce informed proposals.
Likewise one of our models is an existing computer vision model,
the \emph{BlendSCAPE} model, a parametric model of human
bodies~\cite{hirshberg2012coregistration}.

In Section~\ref{sec:related-chap3},
we discuss related work and explain our informed
sampler approach in Section~\ref{sec:model-chap3}.
Section~\ref{sec:experiments-chap3} presents baseline methods
and experimental setup. Then we present experimental analysis of
informed sampler with three diverse
problems of estimating camera extrinsics (Section~\ref{sec:room}),
occlusion reasoning (Section~\ref{sec:tiles}) and estimating body shape (Section~\ref{sec:bodyshape}).
We conclude with a discussion of future work in Section~\ref{sec:discussion-chap3}.

\section{Related Work}
\label{sec:related-chap3}

This work stands at the intersection of computer vision, computer
graphics, and machine learning; it builds on previous approaches we
will discuss below.

There is a vast literature on approaches to solve computer vision
problems by means of generative models. We mention some works that
also use an accurate graphics process as a generative model. This
includes applications such as indoor scene
understanding~\cite{del2012bayesian}, human pose
estimation~\cite{lee2004proposal}, hand pose
estimation~\cite{de2008model} and many more. Most of these works are
however interested in inferring MAP solutions, rather than the full
posterior distribution.

Our method is similar in spirit to a \emph{Data Driven
Markov Chain Monte Carlo} (DDMCMC) methods that use a bottom-up
approach to help convergence of MCMC sampling. DDMCMC methods have
been used in image segmentation~\cite{tu2002image}, object
recognition~\cite{zhu2000integrating}, and human pose
estimation~\cite{lee2004proposal}. The idea of making Markov samplers
data dependent is very general, but in the works mentioned above,
lead to highly problem specific implementations, mostly using
approximate likelihood functions. Since these methods provide specialized
solutions for a particular problem, they are not easily transferable to
new problems.
In contrast, we aim to provide a simple,
yet efficient and general inference technique for problems where an accurate
generative model exists.
Because our method is general we believe that it is easy to adapt to a variety
of new models and tasks.

The idea to invert graphics~\cite{baumgart1974inversegraphics} in order to
understand scenes also has roots in the computer graphics community under the term
`inverse rendering'. The goal of inverse rendering however is to
derive a direct mathematical model for the forward light transport
process and then to analytically invert it. The work
of~\cite{ramamoorthi2001signal} falls in this category. The authors
formulate the light reflection problem as a convolution, to then
cast the inverse light transport problem as a deconvolution.
While this is a very elegant way to pose the problem, it requires
a specification of the inverse process, a requirement generative
modeling approaches try to circumvent.

Our approach can also be viewed as an instance of a probabilistic
programming approach. In the recent work
of~\cite{mansinghka2013approximate}, the authors combine graphics
modules in a probabilistic programming language to formulate an
approximate Bayesian computation. Inference is then implemented using
Metropolis-Hastings (\MH) sampling. This approach is appealing in its
generality and elegance, however we show that for our graphics
problems, a plain \MH sampling approach is not sufficient to achieve reliable
inference and that our proposed informed sampler can achieve robust
convergence in these challenging models.
Another piece of work from~\cite{stuhlmueller2013nips} is similar to our
proposed inference method in that the knowledge about the forward process is
learned as ``stochastic inverses'', then applied for MCMC sampling in a
Bayesian network.
In the present work, we devise an MCMC sampler that works in both a
multi-modal problem as well as for inverting an existing piece of image
rendering code. In summary, our method can be understood in a similar
context as the above-mentioned papers,
including~\cite{mansinghka2013approximate}.


\section{The Informed Sampler}
\label{sec:model-chap3}

In general, inference about the posterior distribution is challenging
because for a complex model $p(\obs|\target)$ no closed-form
simplifications can be made. This is especially true in the case that
we consider, where $p(\obs|\target)$ corresponds to a graphics
engine rendering images.
Despite this apparent complexity, we observe the following: for many
computer vision applications there exist well performing
discriminative approaches, which, given the image, predict some
target variables $\target$ or distributions thereof. These do not correspond
to the posterior distribution that we are interested in, but,
\emph{intuitively} the availability of discriminative inference
methods should make the task of inferring $p(\target|\obs)$ easier.
\emph{Furthermore} a physically accurate generative model can be used in an
offline stage prior to inference to generate as many samples as we would like
or can afford computationally.  Again, \emph{intuitively} this should allow us
to prepare and summarize useful information about the distribution in order to
accelerate the test-time inference.

Concretely, in our case we will use a discriminative method to provide a
global density $T_G(\target|\obs)$,
which we then use in a valid MCMC inference method.
The standard Metropolis-Hasting Markov Chain Monte Carlo (MCMC) is already
described in Section~\ref{sec:inv-graphics} of the previous chapter, where
in each time-step, a proposal is made with a proposal distribution which
is then either accepted or rejected based on the acceptance probability:

\begin{enumerate}
\item Propose a transition using a \textit{proposal distribution $T$}
  and the current state $\target_t$
\begin{equation*}
\targetProp \sim T(\cdot|\target_t)
\end{equation*}
\item Accept or reject the transition based on Metropolis Hastings (MH) acceptance rule:
\begin{equation*}
\target_{t+1} = \left\{
  \begin{array}{cl}
      \targetProp,  & \textrm{rand}(0,1) < \min\left(1,\frac{\pi(\targetProp)
T(\targetProp \to \target_t)}{\pi(\target_t) T(\target_t \to \targetProp)} \right), \\
     \target_t, & \textrm{otherwise.}
  \end{array}
\right.
\end{equation*}
\end{enumerate}

Refer to Section~\ref{sec:inv-graphics} for more details about MCMC sampling.
Different MCMC techniques mainly differ in the type of the
proposal distribution $T$. Next, we describe our informed proposal distribution
which we use in standard Metropolis-Hastings sampling resulting in
our proposed `Informed Sampler' technique.

\subsection{Informed Proposal Distribution}
We use a common mixture kernel for Metropolis-Hastings (MH) sampling. Given
the present target sample $\target_t$, the informed proposal distribution
for MH sampling is given as:
\begin{equation}
T_{\alpha}(\cdot | \obs, \target_t) =
	\alpha \: T_L(\cdot | \target_t) + (1-\alpha) \: T_G(\cdot |
        \obs).
\label{eq:proposal}
\end{equation}
Here $T_L$ is an ordinary \emph{local} proposal distribution, for example a
multivariate Normal distribution centered around the current sample $\target_t$,
and $T_G$ is a \emph{global} proposal distribution independent
of the current state.
We inject knowledge by conditioning the global proposal
distribution $T_G$ on the image observation $\obs$.
We learn the informed proposal $T_G(\cdot | \obs)$ discriminatively in an
offline training stage using a non-parametric density estimator described
below.

The mixture parameter $\alpha \in [0,1]$ controls the contribution of each
proposal, for $\alpha=1$ we recover \MH.
For $\alpha=0$ the proposal $T_{\alpha}$ would be identical to $T_G(\cdot |
\obs)$ and the resulting Metropolis sampler would be a valid Metropolized
independence sampler~\cite{liu2001montecarlo}.
With $\alpha=0$, we call this baseline method `Informed Independent \MH' (\INDLMH).
For intermediate values, $\alpha\in(0,1)$, we combine local with global moves
in a valid Markov chain.
We call this method `Informed Metropolis Hastings' (\MIXLMH).

\subsection{Discriminatively Learning $T_G$}
The key step in the construction of $T_G$ is to include some discriminative
information about the sample $\obs$.
Ideally we would hope to have $T_G$ propose global moves which improve mixing
and even allow mixing between multiple modes, whereas the local proposal $T_L$ is
responsible for exploring the density locally.
To see that this is possible in principle, consider the case of a perfect
global proposal where $T_G$ matches the true posterior distribution, that is,
$T_G(\target | \obs)=P(\target | \obs)$.
In this case, we would get independent samples with $\alpha=0$ because every
proposal is accepted.
In practice $T_G$ is only an approximation to the true posterior
$P(\target|\obs)$.
If the approximation is good enough then the mixture of local and global
proposals will have a high acceptance rate and explore the density rapidly.

In principle, we can use any conditional density estimation technique
for learning a proposal $T_G$ from samples.
Typically high-dimensional density estimation is difficult and even
more so in the conditional case; however, in our case we do have the
true generating process available to provide example pairs $(\target,\obs)$.
Therefore we use a simple but scalable non-parametric density
estimation method based on clustering a feature representation of the
observed image, $v(\obs) \in \mathbb{R}^d$. For each cluster we
then estimate an unconditional density over $\target$ using kernel
density estimation (KDE). We chose this simple setup since it can
easily be reused in many different scenarios, in the experiments we
solve diverse problems using the same method. This method yields a
valid transition kernel for which detailed balance holds.
In addition to the KDE estimate for the global transition kernel we
also experimented with a random forest approach that maps the
observations to transition kernels $T_G$. More details will be given
in Section~\ref{sec:bodyshape}.

    \begin{algorithm}[t]
      \caption{Learning a global proposal $T_G(\target|\obs)$}
      \label{alg:training}
      \begin{algorithmic}
        \STATE 1. Simulate $\{(\target^{(i)},\obs^{(i)})\}_{i=1,\dots,n}$
        from $p(\obs|\target) \: p(\target)$ \STATE 2. Compute a feature
        representation $v(\obs^{(i)})$ \STATE 3. Perform k-means
        clustering of $\{v(\obs^{(i)})\}_i$
        \STATE 4. For each cluster $C_j \subset \{1,\dots,n\}$, fit a kernel
        density estimate $\textrm{KDE}(C_j)$ to the vectors
        $\target^{\{C_j\}}$
      \end{algorithmic}
    \end{algorithm}

    \begin{algorithm}[t]
      \caption{\MIXLMH (Informed Metropolis-Hastings)}
      \label{alg:sampling}
      \begin{algorithmic}
        \STATE \textbf{Input:} observed image $\obs$ \STATE $T_L$
        $\leftarrow$ Local proposal distribution (Gaussian) \STATE $C$
        $\leftarrow$ cluster for $v(\obs)$ \STATE $T_G$ $\leftarrow
        KDE(C)$ (as obtained by Alg.~\ref{alg:training}) \STATE $T =
        \alpha T_L + (1-\alpha) T_G$
        \STATE $\pi(\target|\obs)$ $\leftarrow$ Posterior distribution $P(\target|\obs)$
        \STATE Initialize $\target_1$
        \FOR{$t=1$ {\bfseries to} $N-1$}
        \STATE 1. Sample $\targetProp \sim T(\cdot)$
        \STATE 2. $\gamma =
        \min\left(1,\frac{\pi(\targetProp|\obs) T(\targetProp \to
          \target_t)}{\pi(\target_t|\obs) T(\target_t \to
          \targetProp)} \right)$
        \IF{rand$(0,1) < \gamma$}
        \STATE $\target_{t+1} = \targetProp$
        \ELSE
        \STATE $\target_{t+1} = \target_t$
        \ENDIF
        \ENDFOR
      \end{algorithmic}
    \end{algorithm}
For the feature representation, we leverage successful discriminative
features and heuristics developed in the computer vision community. Different
task specific feature representations can be used in order to provide
invariance to small changes in $\target$ and to nuisance parameters.
The main inference method remains the same across all problems.

We construct the KDE for each cluster and we use a relatively small kernel bandwidth in
order to accurately represent the high probability regions in the posterior.
This is similar in spirit to using only high probability regions as ``darts''
in the \emph{Darting Monte Carlo} sampling technique
of~\cite{sminchisescu2011generalized}.
We summarize the offline training in Algorithm~\ref{alg:training}.

At test time, this method has the advantage that given an image $\obs$
we only need to identify the corresponding cluster once using
$v(\obs)$ in order to sample efficiently from the kernel density $T_G$.
We show the full procedure in Algorithm~\ref{alg:sampling}.

This method yields a transition kernel that is a mixture kernel of a
reversible symmetric Metropolis-Hastings kernel and a metropolized
independence sampler. The combined transition kernel $T$ is hence also
reversible. Because the measure of each kernel dominates the support
of the posterior, the kernel is ergodic and has the correct stationary
distribution~\cite{brooks2011mcmchandbook}.
This ensures correctness of the inference and in the experiments we
investigate the efficiency of the different methods in terms of convergence
statistics.

\section{Setup and Baseline Methods}
\label{sec:experiments-chap3}

In the remainder of the chapter we demonstrate the proposed method in three different
experimental setups.
For all experiments, we use four parallel chains initialized at different
random locations sampled from the prior.
The reported numbers are median statistics over
multiple test images except when noted otherwise.

\subsection{Baseline Methods}
\paragraph{Metropolis Hastings (\MH)}
Described in Section~\ref{sec:inv-graphics},
corresponds to $\alpha=1$, we use a symmetric diagonal
Gaussian distribution, centered at $\target_t$.
\paragraph{Metropolis Hastings within Gibbs (\MHWG)}
We use a Metropolis Hastings scheme in a Gibbs sampler, that is, we draw from
one-dimensional conditional distributions for proposing moves and the Markov
chain is updated along one dimension at a time.  We further use a blocked
variant of this \MHWG~sampler,
where we update blocks of dimensions at a time, and denote it by \BMHWG.
\paragraph{Parallel Tempering (\PT)}
We use Parallel Tempering to address the problem of sampling from
multi-modal
distributions~\cite{swendsen1986replicamontecarlo,geyer1991paralleltempering}.
This technique is also known as ``replica exchange MCMC
sampling''~\cite{hukushima1996exchange}.
We run different parallel chains at different temperatures $T$, sampling
$\pi(\cdot)^{\frac{1}{T}}$ and at each sampling step propose to exchange two
randomly chosen chains. In our experiments we run three chains at
temperature levels $T\in\{1,3,27\}$ that were found to be best working
out of all combinations in $\{1,3,9,27\}$ for all experiments
individually. The highest temperature levels corresponds to an almost
flat distribution.
\paragraph{Regeneration Sampler (\REGAMH)}
We implemented a regenerative MCMC method
\cite{mykland1995regeneration} that performs
adaption~\cite{gilks1998adaptive} of the proposal distribution during
sampling.
Adapting the proposal distribution with existing MCMC samples is not
straight-forward as this would potentially violate the Markov property
of the chain~\cite{atchade2005adaptivemcmc}.
One approach is to identify \textit{times of
  regeneration} at which the chain can be restarted and the proposal
distribution can be adapted using samples drawn previously.
Several approaches to identify good regeneration times in a
general Markov chain have been proposed~\cite{athreya1978new,
  nummelin1978splitting}. We build on~\cite{mykland1995regeneration}
that proposed two \textit{splitting} methods for finding the
regeneration times. Here, we briefly describe the method that we
implemented in this study.

Let the present state of the sampler be $\target_t$ and let the independent
global proposal distribution be $T_G$. When $\targetProp \sim T_G$ is accepted
according to the MH acceptance rule, the probability of a regeneration is
given by:

\begin{equation}
r(\target_t,\targetProp) = \left\{
  \begin{array}{ll}
    \max\{ \frac{c}{w(\target_t)}, \frac{c}{w(\targetProp)}  \},& \textrm{if $w(\target_t)>c$ and $w(\targetProp)>c$},\\
    \max\{ \frac{w(\target_t)}{c}, \frac{w(\targetProp)}{c}  \},& \textrm{if $w(\target_t)<c$ and $w(\targetProp)<c$},\\
    1,    & \textrm{otherwise},
  \end{array}
\right.
\label{reg_eq}
\end{equation}
where $c > 0$ is an arbitrary constant and $w(\target_t) =
\frac{\pi(\target_t)}{T_G(\target_t)}$. The value of $c$ can be set to maximize the
regeneration probability. At every sampling step, if a sample from the
independent proposal distribution is accepted, we compute regeneration
probability using equation~\ref{reg_eq}. If a regeneration occurs,
the present sample is discarded and replaced with one from the
independent proposal distribution $T_G$. We use the same mixture
proposal distribution as in our informed sampling approach where we
initialize the global proposal $T_G$ with a prior distribution and at
times of regeneration fit a KDE to the existing samples. This becomes
the new adapted distribution $T_G$. Refer to
\cite{mykland1995regeneration} for more details of this regeneration
technique. In the work of~\cite{ahn2013distributed}, this regeneration
technique is used with success in a Darting Monte Carlo sampler.

We use the mixture kernel (Eq.~\ref{eq:proposal}) as proposal distribution and
adapt only the global part $T_G(\cdot|\obs)$. This is initialized as
the prior over target variables $P(\target)$ and at times of regeneration we fit a KDE to
the already drawn samples. For comparison we used the same mixture coefficient
$\alpha$ as for \MIXLMH.

\subsection{MCMC Diagnostics}\label{sec:mcmcdiagnostics}
We use established methods for monitoring the convergence of our MCMC
method~\cite{kass1998roundtable,flegal2008mcmcdiagnostics}. In
particular, we report different diagnostics. We compare the different
samplers with respect to the number of iterations instead of time. The
forward graphics process significantly dominates the runtime and
therefore the iterations in our experiments correspond linearly to the
runtime.

\paragraph{Acceptance Rate (AR)}
The ratio of accepted samples to the total Markov chain length.  The higher
the acceptance rate, the fewer samples we need to approximate the
posterior.  Acceptance rate indicates how well the proposal distribution
approximates the true distribution \emph{locally}.
\paragraph{Potential Scale Reduction Factor (PSRF)}
The PSRF diagnostics~\cite{gelman1992psrf,brooks1998convergence} is
derived by comparing within-chain variances with between-chain
variances of sample statistics.  For this, it requires independent
runs of multiple chains (4 in our case) in parallel. Because our
sample $\target$ is multi-dimensional, we estimate the PSRF for each
parameter dimension separately and take the maximum PSRF value as
final PSRF value. A value close to one indicates that all chains
characterize the same distribution. This does not imply convergence,
as the chains may all collectively miss a mode. However, a PSRF value
much larger than one is a certain sign of the lack of convergence.
PSRF also indicates how well the sampler visits different modes of
a multi-modal distribution.

\paragraph{Root Mean Square Error (RMSE)}
During our experiments we have access to the input parameters
$\target^*$ that generated the image. To assess whether the posterior
distribution recovers the \emph{correct} value, we report the RMSE between
the posterior expectation $\mathbb{E}_{P(\target |\obs)}[G(\target)]$
and the value $G(\target^*)$ of the generating input. Since there is
noise being added to the observation, we do not have access to the
ground truth posterior expectation and therefore this measure is only
an indicator. Under convergence all samplers would agree on the same
correct value.

\subsection{Parameter Selection}

For each sampler we individually selected hyper-parameters that gave
the best PSRF value after $10k$ iterations. In case the PSRF does not
differ for multiple values, we chose the one with the highest acceptance
rate.  We include a detailed analysis of the baseline samplers and parameter
selection in Appendix~\ref{appendix:chap3}.

\section{Experiments}

We studied the use of informed sampler with three different problem
scenarios: Estimating camera extrinsics (Section~\ref{sec:room}),
occlusion reasoning (Section~\ref{sec:tiles}) and
estimating body shape (Section~\ref{sec:bodyshape}).

\subsection{Estimating Camera Extrinsics}
\label{sec:room}

\begin{figure}[!h]
\begin{center}
\centerline{\includegraphics[width=.7\columnwidth]{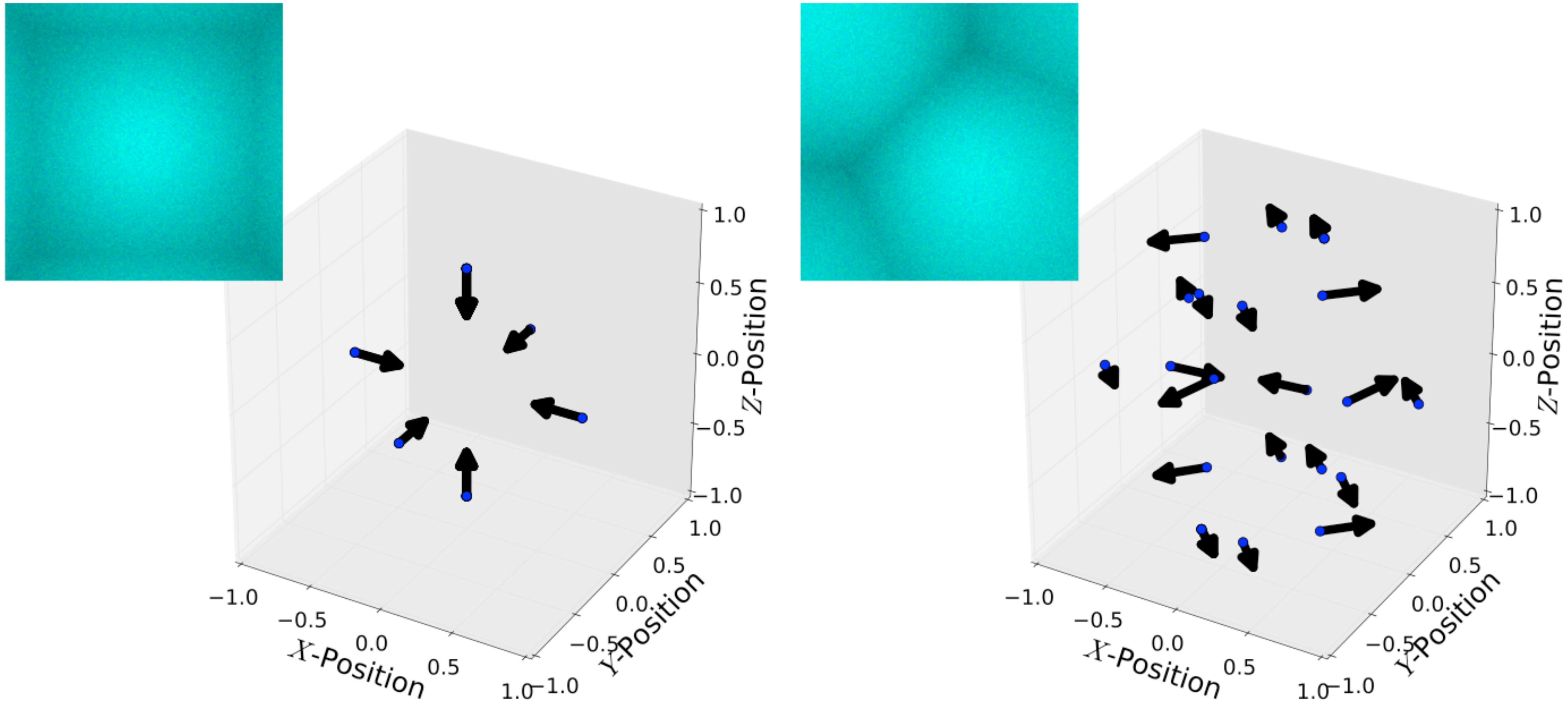}}
\mycaption{Sample images for the experiment on `Estimating Camera Extrinsics'}
  {Two rendered room images with possible camera
  positions and headings that produce the same image. Not shown are
  the orientations; in the left example all six headings can
  be rolled by 90,180, and 270 degrees for the same image.}
\label{fig:room}
\end{center}
\end{figure}

We implement the following simple graphics scenario to create a
challenging multi-modal problem. We render a cubical
room of edge length 2 with a point light source in the center of the
room, $(0,0,0)$, from a camera somewhere inside the
room. The camera parameters are described by its $(x,y,z)$-position
and the orientation, specified by yaw, pitch, and roll angles. The
inference process consists of estimating the posterior over these 6D
camera parameters $\target$. See Fig.~\ref{fig:room} for
two example renderings. Posterior inference is a highly multi-modal problem
because the room is a cubical and thus symmetric. There are 24
different camera parameters that will result in the same
image. This is also shown in Fig.~\ref{fig:room} where we plot the
position and orientation (but not camera roll) of all camera
parameters that create the same image. A rendering of a
$200\times 200$ image with a resolution of $32bit$ using a single core
on an Intel Xeon 2.66GHz machine takes about $11\textrm{ms}$ on
average.

A small amount of isotropic Gaussian noise is added to the rendered
image $G(\target)$, using a standard deviation of $\sigma=0.02$. The
posterior distribution we try to infer is then:
$p(\target|\obs) \propto p(\obs|\target)p(\target) =
\mathcal{N}(\obs|G(\target),\sigma^2) \: \textrm{Uniform}(\target)$.
The uniform prior over location parameters ranges between $-1.0$ and $1.0$ and
the prior over angle parameters is modeled with wrapped uniform distribution over $[-\pi,\pi]$.

To learn the informed part of the proposal distribution from data, we
computed a histogram of oriented gradients (HOG)
descriptor~\cite{dalal2005histograms} from the image, using 9
gradient orientations and cells of size $20\times20$ yielding a feature vector
$v(\obs)\in\mathbb{R}^{900}$. We generated $300k$ training
images using a uniform prior over the camera extrinsic parameters, and
performed k-means clustering using $5k$ cluster centers based on the HOG feature
vectors. For each cluster cell, we then computed and stored a KDE for
the 6 dimensional camera parameters, following the steps in
Algorithm~\ref{alg:training}. As test data, we create 30 images using
extrinsic parameters sampled uniform at random over their range.

\subsubsection{Results}
\label{sec:roomresults}

\begin{figure*}[!t]
\begin{center}
\centerline{\includegraphics[width=\textwidth]{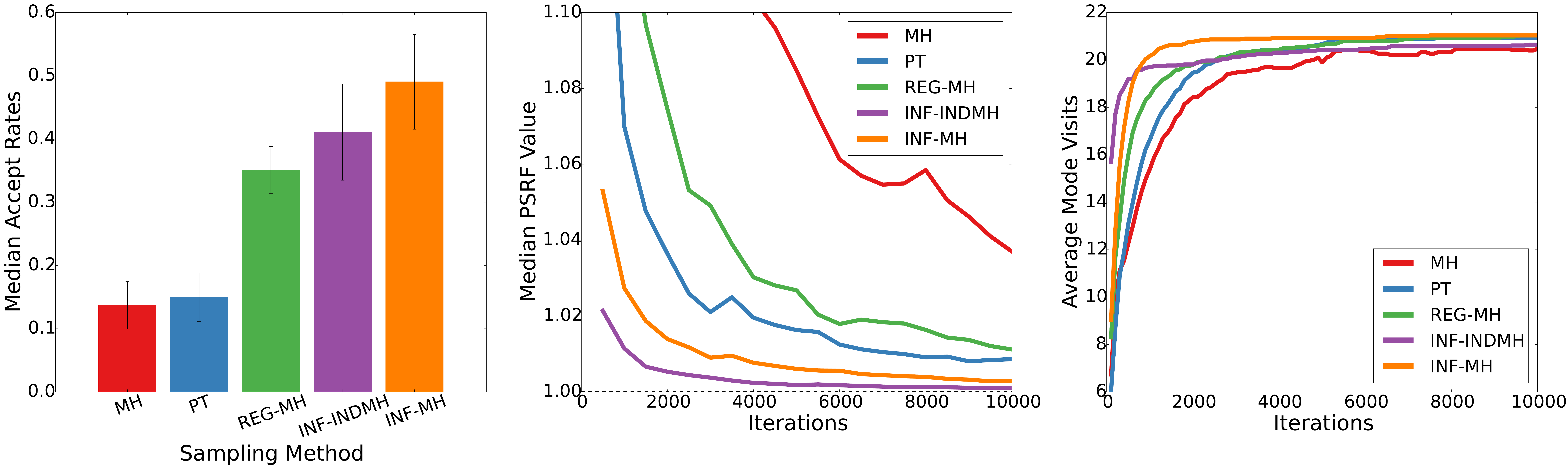}}
\mycaption{Results of the `Estimating Camera Extrinsics' experiment}
  {Acceptance Rates (left), PSRFs (middle), and average number of modes
  visited (right) for different sampling methods. We plot the
  median/average statistics over 30 test examples.}
\label{fig:camPose_ALL}
\vspace{-0.3cm}
\end{center}
\end{figure*}

We show results in Fig.~\ref{fig:camPose_ALL}. We observe that both
\MH and \PT yield low acceptance rates (AR) compared to other methods.
However parallel tempering appears to overcome the multi-modality
better and improves over \MH in terms of convergence. The same holds
for the regeneration technique, we observe many regenerations, good
convergence and AR. Both \INDLMH and \MIXLMH converge quickly.

In this experimental setup, we have access to the 24 different exact modes.
We analyze how quickly the samplers visit
the modes and whether or not they capture all of them. For every
different instance, the pairwise distances between the modes changes,
therefore we chose to define `visiting a mode' in the following way.
We compute a Voronoi tessellation with the modes as centers. A mode is
visited if a sample falls into its corresponding Voronoi cell, that
is, it is closer than to any other mode. Sampling uniform at random
would quickly find the modes (depending on the cell sizes) but is not
a valid sampler that characterizes the desired posterior distribution.
We also experimented with balls of different radii
around the modes and found a similar behavior to the one we report
here. Figure~\ref{fig:camPose_ALL} (right) shows results for various
samplers. We find that \MIXLMH discovers different modes quicker when
compared to other baseline samplers. Just sampling from the global
proposal distribution \INDLMH is initially visiting more modes (it is
not being held back by local steps) but is dominated by \MIXLMH over
some range. This indicates that the mixture kernel takes advantage of
both local and global moves, either one of them is exploring slower.
Also in most examples, all samplers miss some modes under our
definition, the average number of discovered modes is 21 for \MIXLMH
and even lower for \MH.

\begin{figure*}[h]
\centerline{\includegraphics[width=.9\textwidth]{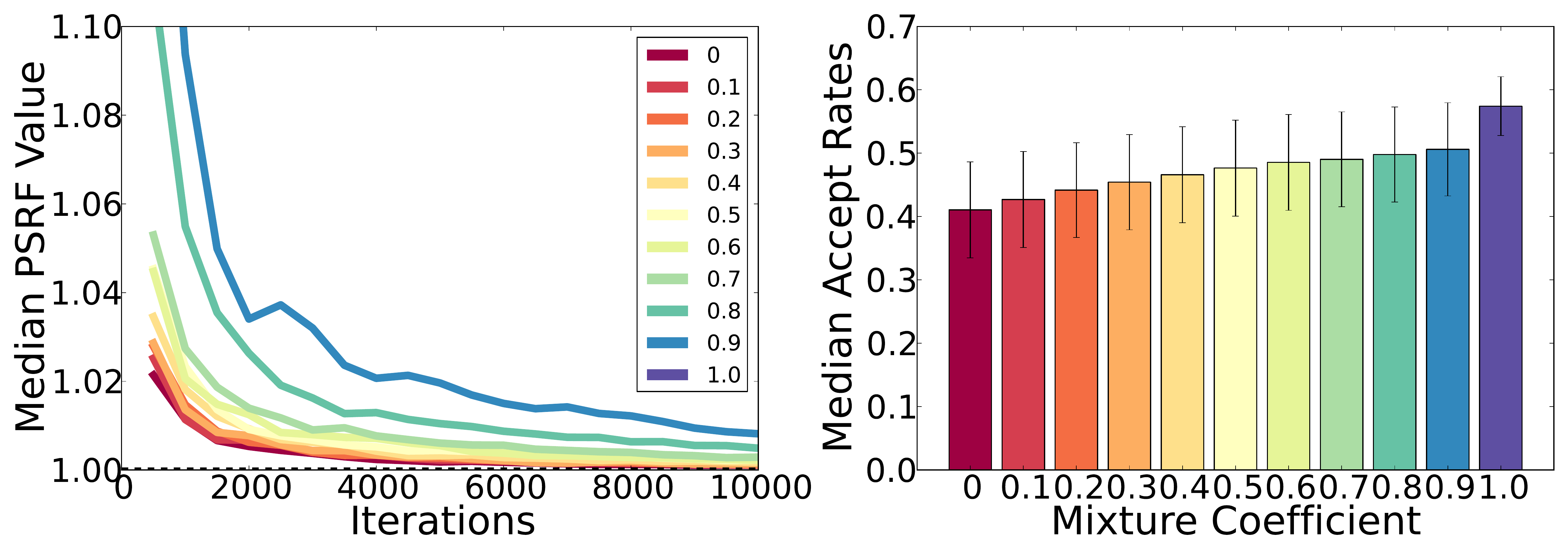}}
\mycaption{Role of mixture coefficient}
{PRSFs and Acceptance rates corresponding to various mixture coefficients ($\alpha$) of \MIXLMH sampling in `Estimating Camera Extrinsics' experiment.}
\label{fig:exp1_alpha:chap3}
\end{figure*}

Figure~\ref{fig:exp1_alpha:chap3} shows the effect of mixture
coefficient ($\alpha$) on the informed sampling
\MIXLMH. Since there is no significant difference in PSRF values for
$0 \le \alpha \le 0.7$, we chose $0.7$ due to its high acceptance
rate.
Likewise, the parameters of the baseline samplers are chosen based on the PSRF and acceptance rate metrics.
See Appendix~\ref{appendix:chap3:room} for the analysis of the baseline samplers and the parameter selection.

\begin{figure*}[!t]
\begin{center}
\centerline{\includegraphics[width=0.9\textwidth]{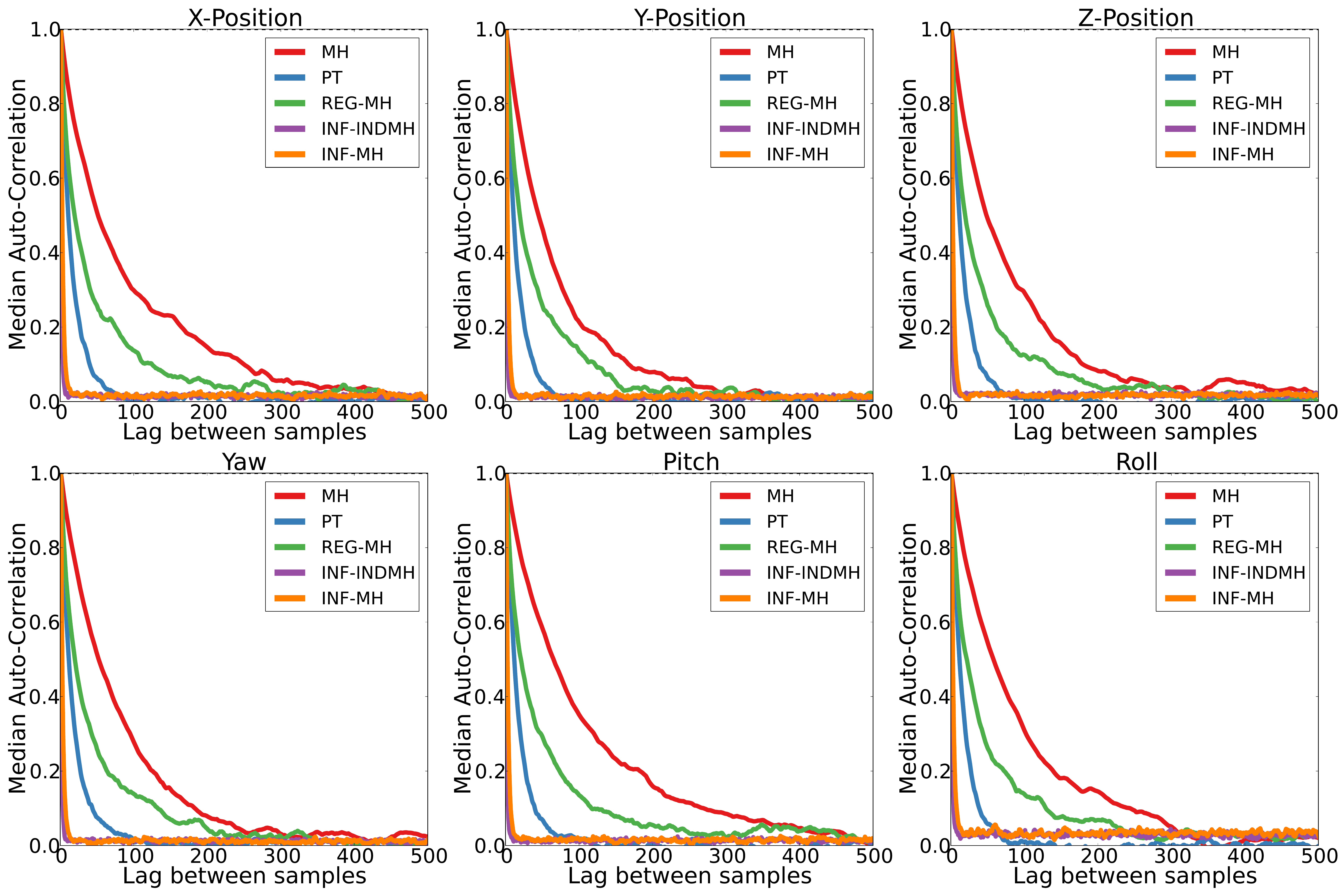}}
\mycaption{Independence of obtained samples}{Auto-Correlation of samples obtained by different
      sampling techniques in camera extrinsics experiment, for each of
      the six extrinsic camera parameters.}
\label{fig:camPose_AC}
\vspace{-0.3cm}
\end{center}
\end{figure*}

We also tested the \MHWG sampler and found that it did not converge
even after $100k$ iterations, with a PSRF value around 3. This is to be
expected since single variable updates will not traverse the
multi-modal posterior distributions fast enough due to the high
correlation of the camera parameters. In Fig.~\ref{fig:camPose_AC}, we plot the median
auto-correlation of samples obtained by different sampling techniques,
separately for each of the six extrinsic camera parameters. The
informed sampling approach (\MIXLMH~and \INDLMH) appears to produce samples which
are more independent compared to other baseline samplers.

As expected, some knowledge of the multi-modal structure of the
posterior needs to be available for the sampler to perform
well. The methods \INDLMH and \MIXLMH have this
information and perform better than baseline methods and \REGMH.

\subsection{Occluding Tiles}
\label{sec:tiles}

\begin{figure*}[th]
\begin{center}
\centerline{\includegraphics[width=\textwidth]{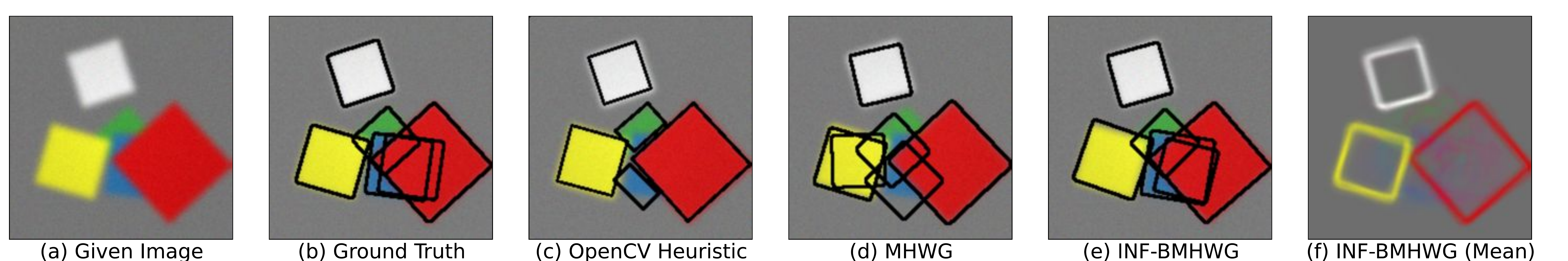}}
\mycaption{A visual result in `Occluding Tiles' experiment}
  {(a) A sample rendered image, (b) Ground truth squares, (c) Rectangle fitting
  with OpenCV MinAreaRect algorithm and
  most probable estimates from 5000 samples obtained by (d) \MHWG
  sampler (best baseline) and (e) the \INFBMHWG sampler. (f)
  Posterior expectation of the square boundaries obtained by \INFBMHWG
  sampling (The first 2000 samples are discarded as burn-in).}
\label{fig:occlusion_main}
\end{center}
\end{figure*}

In a second experiment, we render images depicting a fixed number
of six quadratic tiles placed at a random location $(x, y)$ in the image at a
random depth $z$ and orientation $\theta$. We blur the image and add
a bit of Gaussian random noise ($\sigma = 0.02$). An example is
depicted in Fig.~\ref{fig:occlusion_main}(a), note that all the
tiles are of the same size, but farther away tiles look smaller. A rendering
of one $200 \times 200$ image takes about $25\textrm{ms}$ on average. Here, as prior,
we again use the uniform distribution over the 3D cube for tile location parameters, and
wrapped uniform distribution over $[-\frac{\pi}{4},\frac{\pi}{4}]$ for tile orientation angle.
To avoid label switching issues, each tile is given a fixed color and is not changed during the inference.

We chose this experiment such that it resembles the `dead leaves model'
of~\cite{lee2001occlusion}, because it has properties that are
commonplace in computer vision. It is a scene composed of several
objects that are independent, except for occlusion, which complicates
the problem. If occlusion did not exist, the task is readily solved
using a standard OpenCV~\cite{bradski2008opencv} rectangle finding
algorithm ($\textrm{minAreaRect}$). The output of such an algorithm can be seen
in Fig.~\ref{fig:occlusion_main}(c), and we use this algorithm as a discriminative
source of information. This problem is higher dimensional than the
previous one (24, due to 6 tiles of 4 parameters). Inference becomes
more challenging in higher dimensions and our approach without
modification does not scale well with increasing dimensionality. One
way to approach this problem, is to factorize the joint distribution
into blocks and learn informed proposals separately. In the present
experiment, we observed that both baseline samplers and the plain
informed sampling fail when proposing all parameters jointly. Since
the tiles are independent except for the occlusion, we can approximate
the full joint distribution as product of block distributions where
each block corresponds to the parameters of a single tile. To estimate the full
posterior distribution, we learn global proposal distributions for
each block separately and use a block-Gibbs like scheme in our sampler
where we propose changes to one tile at a time, alternating between tiles.

The experimental protocol is the same as before, we render
500$k$ images, apply the OpenCV MinAreaRect algorithm to fit rectangles and take their
found four parameters as features for clustering (10$k$ clusters).
Again KDE distributions are fit to each cluster and at test time, we assign
the observed image to its corresponding
cluster.  The KDE in that chosen cluster determines the global sampler $T_G$
for that tile.  We then use $T_G$ to propose an update to all 4 parameters of
the tile. We refer to this procedure as
\INFBMHWG. Empirically we find $\alpha = 0.8$ to be optimal for \INFBMHWG sampling.
Analysis of various samplers is presented in Appendix~\ref{appendix:chap3:tiles}.

\subsubsection{Results}

An example visual result is shown in Fig.~\ref{fig:occlusion_main}.
We found that the the \MH and \MIXLMH samplers fail entirely on this
problem. Both use a proposal distribution for the entire state and due to the high
dimensionality, there is almost no acceptance ($< 1\%$) and thus they do
not reach convergence. The \MHWG sampler, updating one dimension at a time, is
found to be the best among the baseline samplers with acceptance rate of around $42\%$,
followed by a block sampler that samples each tile separately.
The OpenCV algorithm produces a reasonable initial guess but fails in
occlusion cases.

\begin{figure}[t]
\begin{center}
\centerline{\includegraphics[width=\textwidth]{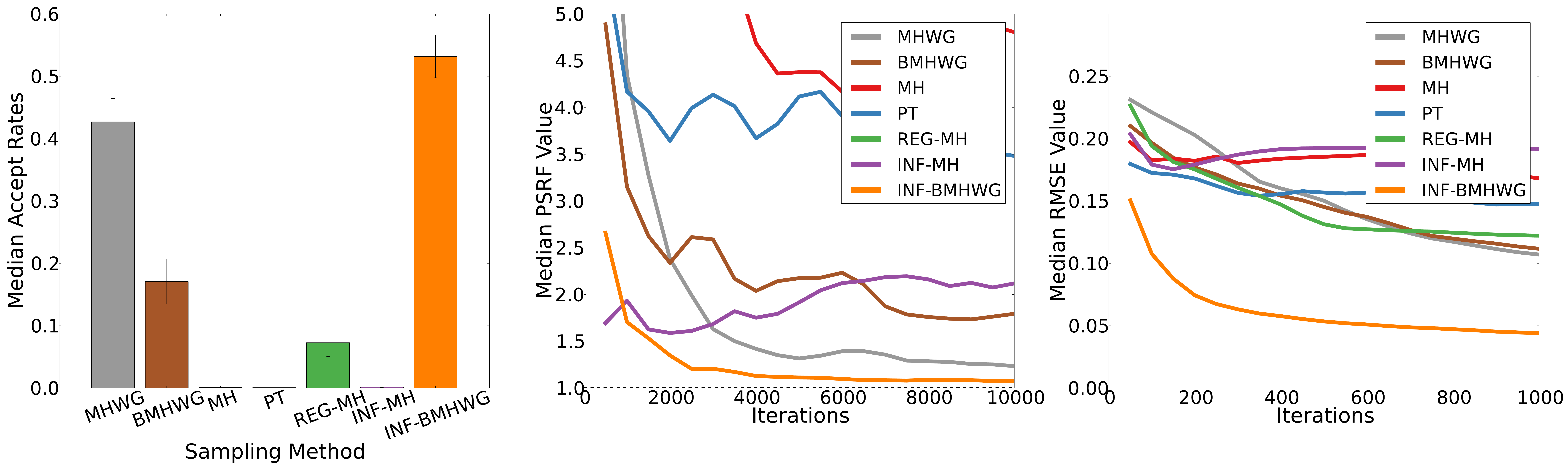}}
\mycaption{Results of the `Occluding Tiles' experiment}
  {Acceptance Rates (left), PSRFs (middle), and RMSEs (right)
  for different sampling methods. Median results for 10 test examples.}
\label{fig:occlusion_ALL}
\end{center}
\end{figure}

The median diagnostic statistical curves for 10 test examples are shown in
Fig.~\ref{fig:occlusion_ALL}, \INFBMHWG by far produces lower
reconstruction errors.
The block wise informed sampler \INFBMHWG converges quicker, with
higher acceptance rates ($\approx 53\%$), and lower reconstruction
error. Also in Fig~\ref{fig:occlusion_main}(f) the
posterior distribution is visualized, fully visible tiles are more
localized, position and orientation of occluded tiles more uncertain.
Figure~\ref{fig:exp2_visual_more}, in the Appendix~\ref{appendix:chap3},
shows some more visual results.
Although the model is relatively simple, all the baseline samplers
perform poorly and discriminative information is crucial to enable
accurate inference. Here the discriminative information is provided by a
readily available heuristic in the OpenCV library.

This experiment illustrates a variation of the informed sampling
strategy that can be applied to sampling from high-dimensional
distributions. Inference methods for general high-dimensional
distributions is an active area of research and intrinsically difficult.
The occluding tiles experiment is simple but illustrates this point,
namely that all non-block baseline samplers fail. Block sampling is a common
strategy in such scenarios and many computer vision problems have such
block-structure. Again the informed sampler improves in convergence
speed over the baseline method. Other techniques that produce better
fits to the conditional (block-) marginals should give faster
convergence.

\subsection{Estimating Body Shape}
\label{sec:bodyshape}

The last experiment is motivated by a real world problem: estimating
the 3D body shape of a person from a single static depth image. With
the recent availability of cheap active depth sensors, the use of RGBD
data has become ubiquitous in computer
vision~\cite{shao2013rgbd,han2013computervisionkinect}.

\begin{figure*}[t]
\begin{center}
\centerline{\includegraphics[width=\columnwidth]{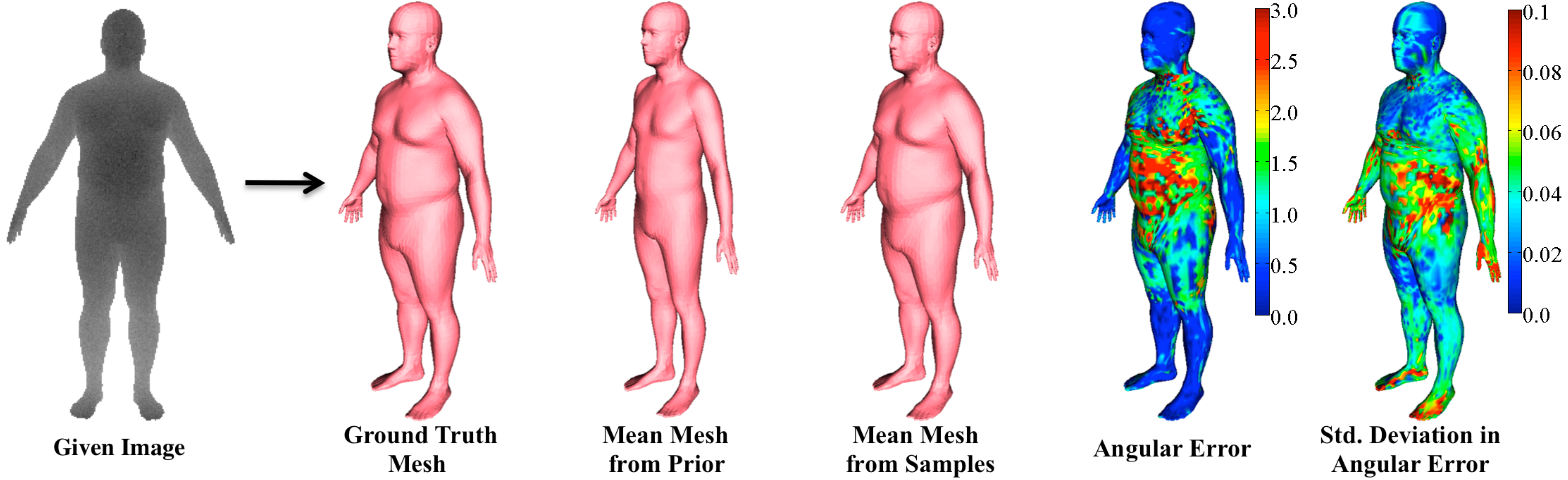}}
\mycaption{Inference of body shape from a depth image}
  {A sample test result showing the result of 3D mesh
  reconstruction with the first 1000 samples obtained using our
  \MIXLMH sampling method. We visualize the angular error (in degrees) between the
  estimated and ground truth edge and project onto the mesh.}
\label{fig:meshVariances}
\end{center}
\end{figure*}

To represent a human body, we use
the \emph{BlendSCAPE} model~\cite{hirshberg2012coregistration}, which updates the
originally proposed SCAPE model~\cite{anguelov2005scape} with better
training and blend weights. This model produces a 3D mesh of a human
body as shown in Fig.~\ref{fig:meshVariances} as a function of
shape and pose parameters. The shape parameters allow us to represent
bodies of many builds and sizes, and includes a statistical
characterization (being roughly Gaussian). These parameters control
directions in deformation space, which were learned from a corpus of
roughly 2000 3D mesh models registered to scans of human bodies via
PCA. The pose parameters are joint angles which indirectly control
local orientations of predefined parts of the model.

Our model uses 57 pose parameters and any number of shape parameters
to produce a 3D mesh with 10,777 vertices. We use the first 7 SCAPE
components to represent the shape of a person and keep the pose fixed.
The camera viewpoint, orientation, and pose of the person is held fixed. Thus a
rendering process takes $\target\in\mathbb{R}^7$, generates a 3D mesh
representation of it and projects it through a virtual depth camera to
create a depth image of the person $\obs$. This can be done in various
resolutions, we chose $430\times 260$ with depth values represented as
$32\textrm{bit}$ numbers in the interval $[0,4]$. On average, a full render path takes about
$28\textrm{ms}$. We add Gaussian noise with standard
deviation of $0.02$ to the created depth image. See
Fig.\ref{fig:meshVariances}(left) for an example.

We used very simple low level features for feature representation. In
order to learn the global proposal distribution we compute depth
histogram features on a $15\times 10$ grid on the image. For each cell
we record the mean and variance of the depth values. Additionally we
add the height and the width of the body silhouette as features
resulting in a feature vector $v(\obs)\in\mathbb{R}^{302}$. As
normalization, each feature dimension is divided by the maximum value
in the training set. We used $400k$ training images sampled from the
standard normal prior distribution and 10$k$ clusters to learn the KDE
proposal distributions in each cluster cell.

For this experiment, we also experimented with a different conditional
density estimation approach using a forest of random regression
trees~\cite{breiman1984cart,breiman2001random}.
See Section~\ref{sec:forests} for a brief overview of random forests.
In the previous experiments, utilizing the KDE estimates, the
discriminative information entered through the feature representation.
Then, suppose if there was no relation between some observed features and the
variables that we are trying to infer, we would require a large number
of samples to reliably estimate the densities in the different
clusters. The regression forest can adaptively partition the parameter
space based on observed features and is able to ignore uninformative features, thus may lead to better
fits of the conditional densities. It can thus be understood as the
adaptive version of the k-means clustering technique that solely
relies on the used metric (Euclidean in our case).

In particular, we use the same features as for k-means clustering but
grow the regression trees using a mean square error criterion for
scoring the split functions. A forest of 10 binary trees with a depth
of 15 is grown, with the constraint of having a minimum of 40 training
points per leaf node. Then for each of the leaf nodes, a KDE is
trained as before. At test time the regression forest yields a mixture
of KDEs as the global proposal distribution. We denote this method as
\MIXLMHRF in the experiments.

Instead of using one KDE model for each cluster, we could also
explore a regression approach, for example using a
discriminative linear regression model to map observations into
proposal distributions. By using informative covariates in the
regression model, one should be able to overcome the curse of
dimensionality. Such a semi-parametric approach would allow to capture
explicit parametric dependencies of the variables (for example linear
dependencies) and combine them with non-parametric estimates of the
residuals. We are exploring this technique as future work.

Again, we chose parameters for all samplers individually, based on
empirical mixing rates. For informed samplers, we chose $\alpha = 0.8$,
and a local proposal standard deviation of 0.05. The full analysis for all
samplers is included in Appendix~\ref{appendix:chap3:body}.

\subsubsection{Results}
\label{sec:bodyresults}

\begin{figure}[h]
\begin{center}
\centerline{\includegraphics[width=\columnwidth]{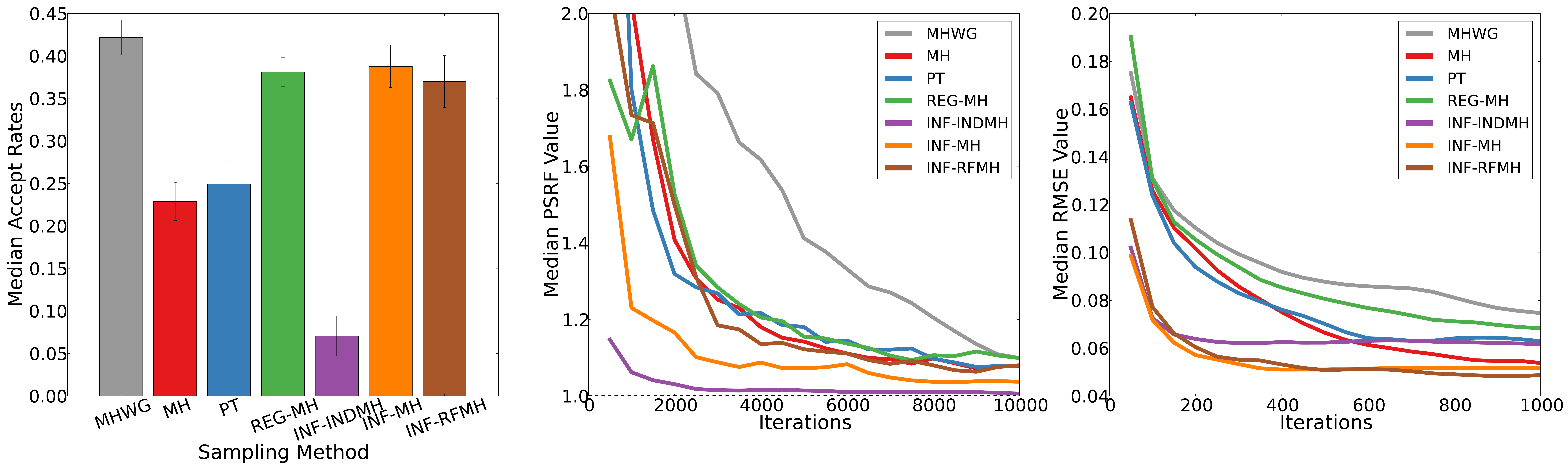}}
\mycaption{Results of the `Body Shape' experiment}{Acceptance Rates (left), PSRFs (middle), and RMSEs (right)
  for different sampling methods in the body shape experiment. Median results over 10 test examples.}
\label{fig:bodyShape_ALL}
\end{center}
\end{figure}

We tested the different approaches on 10 test images that are
generated by the shape parameters drawn from the standard normal prior
distribution. Figure~\ref{fig:bodyShape_ALL} summarizes the results of the
sampling methods. We make the following observations. The baselines
methods \MH, \MHWG, and \PT show inferior convergence results and, \MH and
\PT also suffer from lower acceptance rates. Just sampling from
the distribution of the discriminative step (\INDLMH) is not enough,
because the low acceptance rate indicates that the global proposals do not represent the
correct posterior distribution. However, combined with a local
proposal in a mixture kernel, we achieve a higher acceptance rate, faster convergence
and a decrease in RMSE. The regression forest approach has slower
convergence than \MIXLMH. In this example, the regeneration sampler
\REGMH does not improve over simpler baseline methods. We attribute
this to rare regenerations which may be improved with more specialized
methods.

We believe that our simple choice of depth image representation can
also be significantly improved. For example, features can be
computed from identified body parts, something that the simple
histogram features have not taken into account. In the computer vision
literature, some discriminative approaches for pose estimation do
exist, most prominent being the influential work on pose recovery in parts for
the Kinect Xbox system~\cite{shotton2011kinect}.
In future work, we plan to use similar methods to deal with pose variation and
complicated dependencies between parameters and observations.

\subsubsection{3D Mesh Reconstruction}

In Fig.~\ref{fig:meshVariances}, we show a sample 3D body mesh
reconstruction result using the \MIXLMH sampler after only 1000
iterations. We visualized the difference of the mean posterior and the
ground truth 3D mesh in terms of mesh edge directions. One can observe
that most differences are in the belly region and the feet of the
person. The retrieved posterior distribution allows us to assess the
model uncertainty.
To visualize the posterior variance we record standard deviation over
the edge directions for all mesh edges. This is back-projected to
achieve the visualization in Fig.~\ref{fig:meshVariances}(right). We see
that posterior variance is higher in regions of higher error, that is, our
model predicts its own uncertainty correctly~\cite{dawid1982calibration}. In a
real-world body scanning scenario, this information will be
beneficial; for example, when scanning from multiple viewpoints or in
an experimental design scenario, it helps in selecting the next best
pose and viewpoint to record.
Fig.~\ref{fig:exp3_bodyMeshes}, in appendix, shows more 3D mesh reconstruction
results using our sampling approach.

\begin{figure}[!t]
  \begin{center}
    \centerline{\includegraphics[width=0.9\columnwidth]{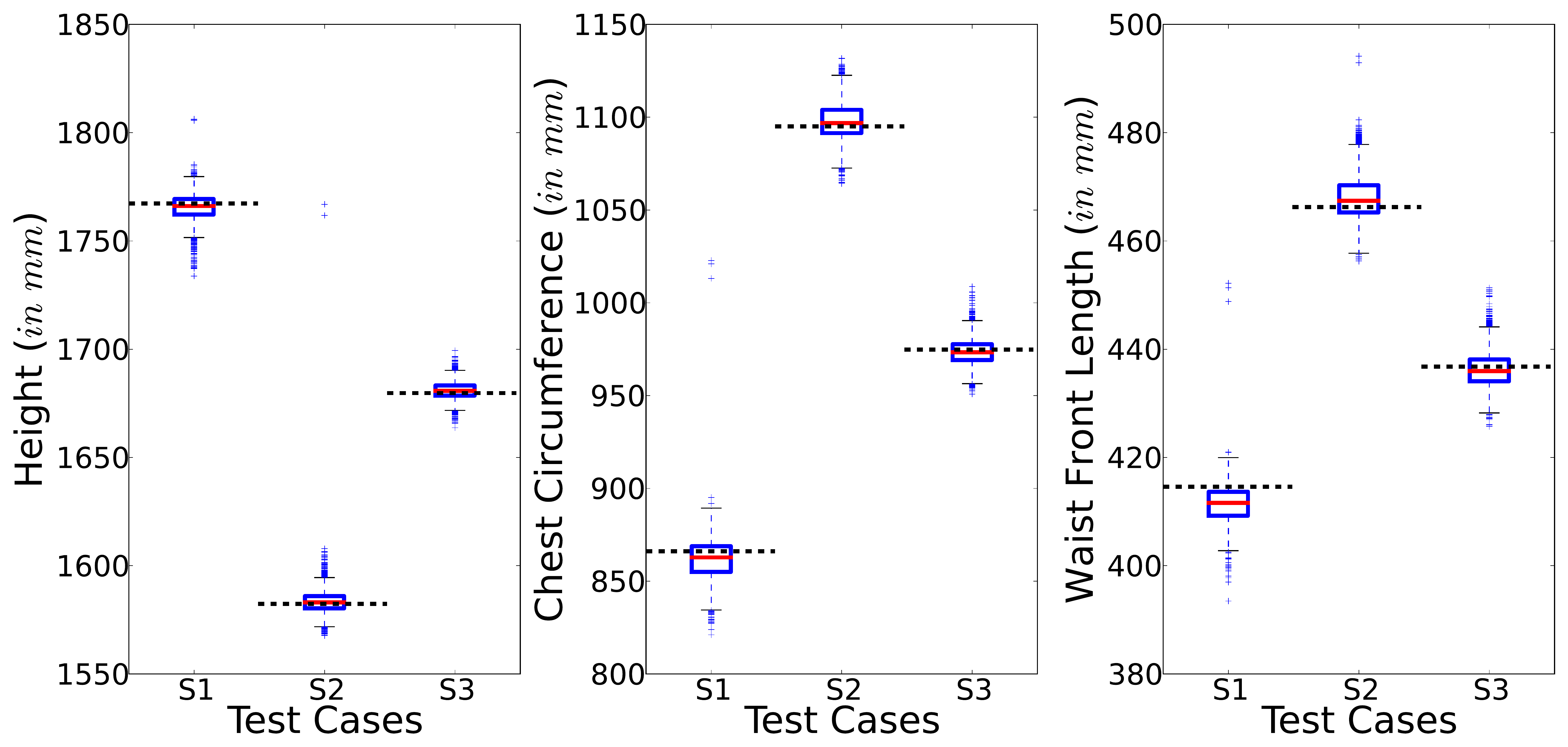}}
    \mycaption{Body measurements with quantified uncertainty}
      {Box plots of three body measurements for three test
      subjects, computed from the first $10k$ samples obtained by the
      \MIXLMH sampler. Dotted lines indicate measurements
      corresponding to ground truth SCAPE parameters.}
    \label{bodyMeasurements}
  \end{center}
\end{figure}

\subsubsection{Body Measurements}
Predicting body measurements has many applications including clothing,
sizing and ergonomic design. Given pixel observations, one may wish to
infer a distribution over measurements (such as height and chest
circumference). Fortunately, our original shape training corpus
includes a host of 47 different per-subject measurements, obtained by
professional anthropometrists; this allows us to relate shape
parameters to measurements. Among many possible forms of regression,
regularized linear regression~\cite{zou2005regularization} was found
to best predict measurements from shape parameters. This linear
relationship allows us to transform any posterior distribution over
SCAPE parameters into a posterior over measurements, as shown in
Fig.~\ref{bodyMeasurements}. We report for three randomly chosen
subjects' (S1, S2, and S3) results on three out of the 47
measurements. The dashed lines corresponds to ground truth
values. Our estimate not only faithfully recovers the true value but also
yields a characterization of the full conditional posterior.

\subsubsection{Incomplete Evidence}

\begin{figure}[!t]
\begin{center}
\centerline{\includegraphics[width=1.0\columnwidth]{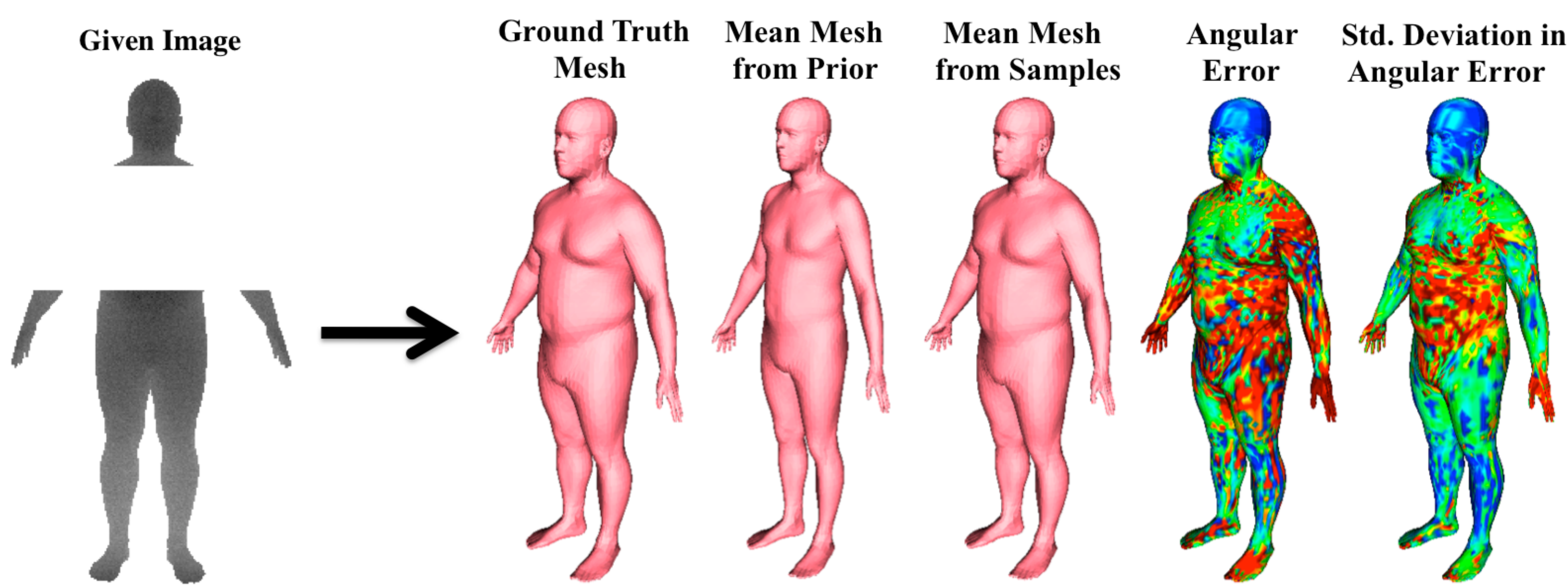}}
\mycaption{Inference with incomplete evidence}
  {Mean 3D mesh and corresponding errors and uncertainties
  (std. deviations) in mesh edge directions, for the same test case as in Fig.~\ref{fig:meshVariances},
  computed from first $10k$ samples of our \MIXLMH sampling method
  with (bottom row) occlusion mask in image evidence.
(blue indicates small values and red indicates high values)}
\label{fig:occlusionMeshes}
\end{center}
\end{figure}

Another advantage of using a generative model is the ability
to reason with missing observations. We perform a simple
experiment by occluding a portion of the observed depth image.
We use the same inference and learning codes, with the same parametrization
and features as in the non-occlusion case, but retrain the model to account for
the changes in the forward graphics process. The result of \MIXLMH,
computed on the first $10k$ samples is shown in
Fig.~\ref{fig:occlusionMeshes}. The 3D reconstruction is reasonable even
under large occlusion; the error and the edge direction variance did
increase as expected.

\section{Discussion and Conclusions}
\label{sec:discussion-chap3}
This work proposes a method to incorporate discriminative methods into
Bayesian inference in a principled way. We augment a sampling
technique with discriminative information to enable inference with
global accurate generative models. Empirical results on three
challenging and diverse computer vision experiments are discussed. We
carefully analyze the convergence behavior of several different
baselines and find that the informed sampler performs well across all
different scenarios. This sampler is applicable to general scenarios
and in this work we leverage the accurate forward process for offline
training, a setting frequently found in computer vision applications.
The main focus is the generality of the approach, this inference
technique should be applicable to many different problems and not be
tailored to a particular problem.

We show that even for very simple scenarios, most baseline samplers
perform poorly or fail completely. By including a global
image-conditioned proposal distribution that is informed through
discriminative inference we can improve sampling performance. We
deliberately use a simple learning technique (KDEs on k-means cluster
cells and a forest of regression trees) to enable easy reuse in other
applications. Using stronger and more tailored discriminative models
should lead to better performance. We see this as a way where top-down
inference is combined with bottom-up proposals in a probabilistic
setting.

There are some avenues for future work; we understand this method as
an initial step into the direction of general inference techniques for
accurate generative computer vision models. Identifying conditional
dependence structure should improve results, e.g.
recently~\cite{stuhlmueller2013nips} used structure in Bayesian
networks to identify such dependencies.
One assumption in our work is that we use an accurate generative
model. Relaxing this assumption to allow for more general scenarios
where the generative model is known only approximately is important
future work. In particular for high-level computer vision problems
such as scene or object understanding there are no accurate generative
models available yet but there is a clear trend towards physically
more accurate 3D representations of the world.
This more general setting is different to the one we consider in this
chapter, but we believe that some ideas can be carried over. For
example, we could create the informed proposal distributions from
manually annotated data that is readily available in many computer
vision data sets. Another problem domain are trans-dimensional models,
that require different sampling techniques like reversible jump MCMC
methods~\cite{green1995reversible,brooks2011mcmchandbook}.
We are investigating general techniques to \emph{inform} this
sampler in similar ways as described in this chapter.

We believe that generative models are useful in many computer vision
scenarios and that the interplay between computer graphics and
computer vision is a prime candidate for studying probabilistic
inference and probabilistic
programming~\cite{mansinghka2013approximate}.
However, current inference techniques need to be improved on many
fronts: efficiency, ease of usability, and generality. Our method is a
step towards this direction: the informed sampler leverages the power
of existing discriminative and heuristic techniques to enable a
principled Bayesian treatment in rich graphics generative models.
Our emphasis is on generality; we aimed to create a method that can be easily
reused in other scenarios with existing code bases.
The presented results are a successful example of the inversion of an
involved rendering pass.
In the future, we plan to investigate ways to combine existing computer
vision techniques with principled generative models, with the aim of
being general rather than problem specific.

\chapter{Consensus Message Passing}
\label{chap:cmp}

\tikzset{
  cluster/.style={rectangle, minimum height=0.4cm, minimum width=0.4cm, inner sep=0.05cm, draw},
  mylatent/.style={circle, minimum height=0.5cm, minimum width=0.5cm, inner sep=0.05cm, draw},
  myfactor/.style={rectangle, minimum height=0.05cm, minimum width=0.05cm, fill=black, scale=0.75},
  myinitfactor/.style={rectangle, minimum height=0.1cm, minimum width=0.1cm, fill=red, scale=0.75},
}

\newcommand{\METHOD}{Consensus Message Passing\@\xspace}
\newcommand{\Method}{Consensus message passing\@\xspace}
\newcommand{\method}{consensus message passing\@\xspace}
\newcommand{\MTD}{CMP\@\xspace}

\makeatletter
\renewcommand{\thesubfigure}{\alph{subfigure}}
\renewcommand{\@thesubfigure}{(\thesubfigure)\hskip\subfiglabelskip}
\makeatother

In the last chapter, we have proposed a technique to speed up the sampling process
for inverse graphics. Despite having faster convergence with techniques
like informed sampler, sampling based inference is often too slow for practical applications.
An alternative inference approach in vision, which is often faster, is
message-passing in factor graph models (see Section~\ref{sec:pgm}).
Generative models in computer vision tend to be large, loopy and layered as discussed in Section~\ref{sec:pgm}.
We find that widely-used, general-purpose message passing inference algorithms such as Expectation Propagation (EP) and Variational Message Passing (VMP) fail on the simplest of vision models. With these models in mind, we introduce a modification to message passing that learns to exploit their layered structure by passing \textit{consensus} messages that guide inference towards good solutions. Experiments on a variety of problems show that the proposed technique leads to significantly more accurate inference results, not only when compared to standard EP and VMP, but also when compared to competitive bottom-up discriminative models. Refer to Section~\ref{sec:pgm} for an overview
of factor graphs and message-passing inference.

\section{Introduction}
\label{sec:introduction}

As discussed in Section~\ref{sec:gen-adv-limits}, perhaps, the most significant challenge of the generative modeling framework is that inference can be very hard. Sampling-based methods run the risk of slow convergence, while message passing-based methods (which are the focus of this chapter) can converge slowly, converge to bad solutions, or fail to converge at all. Whilst significant efforts have been made to improve the accuracy of message passing algorithms (\eg by using structured variational approximations), many challenges remain, including difficulty of implementation, the problem of computational cost and the question of how the structured approximation should be chosen. The work
in this chapter aims to alleviate these problems for general-purpose message-passing algorithms.

Our initial observation is that general purpose message passing inference algorithms (\eg EP and VMP; \cite{Minka2001,Winn2005}) fail on even the simplest of computer vision models. We claim that in these models the failure can be attributed to the algorithms' inability to determine the values of a relatively small number of influential variables which we call `global' variables. Without accurate estimation of these global variables, it can be very difficult for message passing to make meaningful progress on the other variables in the model.

Latent variables in vision models are often organized in a layered structure (also discussed in
Section~\ref{sec:pgm}), where the observed image pixels are at the bottom and high-level scene parameters are at
the top. Additionally, knowledge about the values of the variables at level $l$ is sufficient to reason about
any global variable at layer $l+1$. With these properties in mind, we develop a method called \emph{\METHOD}
(\MTD) that learns to exploit such layered structures and estimate global variables during the early stages of
inference.

Experimental results on a variety of problems show that \MTD leads to significantly more accurate inference results while preserving the computational efficiency of standard message passing. The implication of this work is twofold. First, it adds a useful tool to the toolbox of techniques for improving general-purpose inference, and second, in doing so it overcomes a bottleneck that has restricted the use of model-based machine learning in computer vision.

This chapter is organized as follows.~In Section~\ref{sec:related-work-chap4},
we discuss related work and explain our CMP approach in Section~\ref{sec:method-chap4}.
Then we present experimental analysis of
CMP with three diverse generative vision models in Section~\ref{sec:experiments-chap4}.
We conclude with a discussion in Section~\ref{sec:discussion-chap4}.

\section{Related Work}
\label{sec:related-work-chap4}

Inspiration for \MTD stems from the kinds of distinctions that have been made for decades between so-called `intuitive', bottom-up, fast inference techniques, and iterative `rational' inference techniques~\citep{Hinton1990}. \MTD can be seen as an implementation of such ideas in the context of message passing, where the consensus messages form the `intuitive' part of inference and the following standard message passing forms the `rational' part. Analogues to intuitive and rational inference also exist for sampling, where bottom-up techniques are used to compute proposals for MCMC, leading to significant speedup in inference~\citep{tu2002image, stuhlmueller2013nips} (Chapter~\ref{chap:infsampler}).
The works of~\cite{rezende2014stochastic} and~\cite{kingma2013auto} proposed techniques for learning the parameters of both the generative model and the corresponding recognition model.

The idea of `learning to infer' also has a long history. Early examples include~\cite{Hinton1995}, where a dedicated set of `recognition' parameters are learned to drive inference. In more modern instances of such ideas \citep{Munoz2010, Ross2011, Domke2011, shapovalov2013spatial, Munoz2013}, message passing is performed  by a sequence of predictions defined by a graphical model, and the predictors are jointly trained to ensure that the system produces coherent labelings. However, in these techniques, the resulting inference procedure no longer corresponds to the original (or perhaps to any) graphical model. An important distinction of \MTD is that the predictors fit completely within the framework of message passing and final inference results correspond to valid fixed points in the original model of interest.

Finally, we note recent works of~\cite{Heess2013} and~\cite{Eslami2014} that make use of regressors (neural networks and random forests, respectively) to learn to pass EP messages. These works are concerned with reducing the computational cost of computing individual messages and do not make any attempt to change the accuracy or rate of convergence in message passing inference as a whole. In contrast, \MTD learns to pass messages specifically with the aim of reducing the total number of iterations required for accurate inference in a given generative model.

\section{\METHOD}
\label{sec:method-chap4}

\Method exploits the layered characteristic of vision models in order to overcome the aforementioned inference challenges. For illustration, two layers of latent variables of such a model are shown in Fig.~\ref{fig:types-a} using factor graph notation (black). Here the latent variables below ($\mathbf{h}^b = \{ h^b_k\}$) are a function of the latent variables above ($\mathbf{h}^a = \{ h^a_k\}$) and the global variables $p$ and $q$, where $k$ ranges over pixels (in this case $|k|=3$). As we will see in the experiments that follow, this is a recurring pattern that appears in many models of interest in vision. For example, in the case of face modeling, the $\mathbf{h}^a$ variables correspond to the normals $\mathbf{n}_i$, the global variable $p$ to the light vector $\mathbf{l}$, and $\mathbf{h}^b$ to the shading intensities $s_i$ (see Fig.~\ref{fig:shading-model}).

Our reasoning follows a recursive structure. Assume for a moment that in Fig.~\ref{fig:types-a}, the messages from the layer below to the inter-layer factors (blue) are both informative and accurate (\eg\ due to being close to the observed pixels). We will refer to these messages collectively as \textit{contextual messages}. It would be desirable, for purposes of both speed and accuracy, that we could ensure that the messages sent to the layer above ($\mathbf{h}^a$) are also accurate and informative. If we had access to an oracle that could give us the correct belief for the global variables ($p$ and $q$) for the image, we could send accurate initial messages from $p$ and $q$ and then compute informative and accurate messages from the inter-layer factors to the layer above.

In practice, however, we do not have access to such an oracle. In this work we train regressors to \textit{predict} the values of the global variables given all the messages from the layer below. Should this prediction be good enough, the messages to the layer above will be informative and accurate, and the inductive argument will hold for further layers (if any) above in the factor graph. We describe how these regressors are trained in Section~\ref{sec:training-chap4}. The approach consists of the following two components:

\begin{enumerate}

\item Before inference, for each global variable in different layers of the model, we train a regressor to predict some oracle's value for the target variable given the values of all the messages from the layer below (\ie the \textit{contextual messages}, Fig.~\ref{fig:types-a}, blue),

\item During inference, each regressor sends this belief in the form of a \textit{consensus message} (Fig.~\ref{fig:types-a}, red) to its target variable.

\end{enumerate}

In some models, it will be useful to employ a second type of \MTD, illustrated graphically in Fig.~\ref{fig:types-b}, where global layer variables are absent and loops in the graphical model are due to global variables in other layers. In this case, a consensus message is sent to each variable in the latent layer above, given all the contextual messages.

\begin{figure}[t]
	\centering
	\subfigure[Type A]{
		\begin{tikzpicture}
			\matrix at (0, 0.2) [matrix, column sep=0.1cm, row sep=0.21cm,ampersand replacement=\&]
			{
				\node { }; \&
				\node (y1e) { $\vdots$ }; \&
				\node (y2e) { $\vdots$ }; \&
				\node (y3e) { $\vdots$ }; \&
				\node { }; \\

				\node (a) [mylatent] { $p$ }; \&
				\node (y1) [mylatent] { $h^a_1$ }; \&
				\node (y2) [mylatent] { $h^a_2$ }; \&
				\node (y3) [mylatent] { $h^a_3$ }; \&
				\node (b) [mylatent] { $q$ }; \\

				\& \& \& \& \\

				\node { }; \&
				\node (cl1) [myfactor] {  }; \&
				\node (cl2) [myfactor] {  }; \&
				\node (cl3) [myfactor] {  }; \&
				\node { }; \& \\

				\& \& \& \& \\
				\& \& \& \& \\
				\& \& \& \& \\

				\node { }; \&
				\node (x1) [mylatent] { $h^b_1$ }; \&
				\node (x2) [mylatent] { $h^b_2$ }; \&
				\node (x3) [mylatent] { $h^b_3$ }; \&
				\node { }; \\

				\node { }; \&
				\node (x1e) { $\vdots$ }; \&
				\node (x2e) { $\vdots$ }; \&
				\node (x3e) { $\vdots$ }; \&
				\node { }; \\
			};

			\draw[red!20] (-1.9, 0) -- (1.8, 0);
			\draw[red, dotted] (-1.9, 0) -- (1.8, 0);

			\fill (a.south) ++ (0, -1.1) circle (3pt) [fill=red] { };
			\fill (b.south) ++ (0, -1.1) circle (3pt) [fill=red] { };

			\node[yshift=-1.45cm] at (a.south) { $\textcolor{red}{\Delta^p}$ };
			\node[yshift=-1.45cm] at (b.south) { $\textcolor{red}{\Delta^q}$ };

			\draw[-stealth, cyan] (0.1, -0.2) -- (0.1, 0.2);
			\draw[-stealth, cyan] (0.99, -0.2) -- (0.99, 0.2);
			\draw[-stealth, cyan] (-0.8, -0.2) -- (-0.8, 0.2);

			\fill (0.1, 0) circle (1pt) [fill=cyan] { };
			\fill (0.99, 0) circle (1pt) [fill=cyan] { };
			\fill (-0.8, 0) circle (1pt) [fill=cyan] { };

			\draw[-stealth, red] (1.6, 0.25) -- (1.6, 0.9);
			\draw[-stealth, red] (-1.68, 0.25) -- (-1.68, 0.9);

			\draw (y1.north) -- (y1e);
			\draw (y2.north) -- (y2e);
			\draw (y3.north) -- (y3e);

			\draw (a.south) -- (cl1.north);
			\draw (a.south) -- (cl2.north);
			\draw (a.south) -- (cl3.north);
			\draw (b.south) -- (cl1.north);
			\draw (b.south) -- (cl2.north);
			\draw (b.south) -- (cl3.north);
			\draw (y1.south) -- (cl1);
			\draw (y2.south) -- (cl2);
			\draw (y3.south) -- (cl3);
			\draw [->] (cl1) -- (x1.north);
			\draw [->] (cl2) -- (x2.north);
			\draw [->] (cl3) -- (x3.north);

			\draw (x1.south) -- (x1e);
			\draw (x2.south) -- (x2e);
			\draw (x3.south) -- (x3e);

		\end{tikzpicture}
		\label{fig:types-a}
	}
	\subfigure[Type B]{
		\begin{tikzpicture}

			\matrix at (0, 0.2) [matrix, column sep=0.4cm, row sep=0.217cm,ampersand replacement=\&]
			{
				\node (y1e) { $\vdots$ }; \&
				\node (y2e) { $\vdots$ }; \&
				\node (y3e) { $\vdots$ }; \\

				\node (y1) [mylatent] { $h^a_1$ }; \&
				\node (y2) [mylatent] { $h^a_2$ }; \&
				\node (y3) [mylatent] { $h^a_3$ }; \\

				\& \& \\

				\node (cl1) [myfactor] {  }; \&
				\node (cl2) [myfactor] {  }; \&
				\node (cl3) [myfactor] {  }; \\

				\& \& \\
				\& \& \\
				\& \& \\

				\node (x1) [mylatent] { $h^b_1$ }; \&
				\node (x2) [mylatent] { $h^b_2$ }; \&
				\node (x3) [mylatent] { $h^b_3$ }; \\

				\node (x1e) { $\vdots$ }; \&
				\node (x2e) { $\vdots$ }; \&
				\node (x3e) { $\vdots$ }; \\
			};

			\draw[red!20] (-1.9, 0) -- (1.6, 0);
			\draw[red, dotted] (-1.9, 0) -- (1.6, 0);

			\fill (y1.south) ++ (-0.5, -0.98) circle (3pt) [fill=red] { };
			\fill (y2.south) ++ (-0.5, -0.98) circle (3pt) [fill=red] { };
			\fill (y3.south) ++ (-0.5, -0.98) circle (3pt) [fill=red] { };

			\node[xshift=-0.5cm, yshift=-1.3cm] at (y1.south) { $\textcolor{red}{\Delta^1}$ };
			\node[xshift=-0.5cm, yshift=-1.3cm] at (y2.south) { $\textcolor{red}{\Delta^2}$ };
			\node[xshift=-0.5cm, yshift=-1.3cm] at (y3.south) { $\textcolor{red}{\Delta^3}$ };

			\draw[-stealth, cyan] (0.17, -0.2) -- (0.17, 0.2);
			\draw[-stealth, cyan] (1.35, -0.2) -- (1.35, 0.2);
			\draw[-stealth, cyan] (-1.03, -0.2) -- (-1.03, 0.2);

			\fill (0.17, 0) circle (1pt) [fill=cyan] { };
			\fill (1.35, 0) circle (1pt) [fill=cyan] { };
			\fill (-1.03, 0) circle (1pt) [fill=cyan] { };

			\draw[-stealth, red] (-0.46, 0.23) -- (-0.13, 0.87);
			\draw[-stealth, red] (-1.65, 0.23) -- (-1.32, 0.87);
			\draw[-stealth, red] (0.71, 0.23) -- (1.04, 0.87);

			\draw (y1.north) -- (y1e);
			\draw (y2.north) -- (y2e);
			\draw (y3.north) -- (y3e);

			\draw (y1.south) -- (cl1);
			\draw (y2.south) -- (cl2);
			\draw (y3.south) -- (cl3);
			\draw [->] (cl1) -- (x1.north);
			\draw [->] (cl2) -- (x2.north);
			\draw [->] (cl3) -- (x3.north);

			\draw (x1.south) -- (x1e);
			\draw (x2.south) -- (x2e);
			\draw (x3.south) -- (x3e);

		\end{tikzpicture}
		\label{fig:types-b}
	}
	\mycaption{\Method}{Vision models tend to be large, layered and loopy. (a)~Two adjacent layers of the latent variables of a model of this kind (black). In \MTD, consensus messages (red) are computed from contextual messages (blue) and sent to global variables ($p$ and $q$), guiding inference in the layer. (b)~\Method of a different kind for situations where loops in the graphical model are due to global variables in other layers.}
	\label{fig:types}
\end{figure}

Any message passing schedule can be used subject to the constraint that the consensus messages are given maximum priority within a layer and that they are sent bottom up. Naturally, a consensus message can only be sent after its contextual messages have been computed. It is desirable to be able to ensure that the fixed point (result at convergence) reached under this scheme is also a fixed point of standard message passing in the model. One approach for this is to reduce the certainty of the consensus messages over the course of inference, or to only pass them in the first few iterations. In our experiments we found that even passing consensus messages only in the first iteration led to accurate inference, and therefore we follow this strategy for the remainder of the chapter. It is worth emphasizing that message-passing equations remain unchanged and we used the same scheduling scheme in all our experiments (\ie no need for manual tuning).

It is important to highlight a crucial difference between \method and heuristic \textit{initialization}. In the latter, predictions are made from the \textit{observations} no matter how high up in the hierarchy the target variable is, whereas in \MTD predictions are made using \textit{messages} that are sent from variables immediately below the target variables of interest. The CMP prediction task is much simpler, since the relationship between the target variables and the variables in the layer immediately below is much less complex than the relationship between the target variables and the observations. Furthermore, we know from the layered structure of the model that all relevant information from the observations is contained in the variables in the layer below.  This is because target variables at layer $l+1$ are conditionally independent of all layers $l-1$ and below, given the values of layer $l$.

One final note on the capacity of the regressors. Of course it is true that an infinite capacity regressor can make perfect predictions given enough data (whether using \MTD or heuristic initialization). However, we are interested in practical ways of obtaining accurate results for models of increasing complexity, where lack of capable regressors and unlimited data is inevitable. One important feature of \MTD is that it makes use of predictors in a scalable way, since regressions are only made between adjacent latent layers.

\subsection{Predicting Messages for CMP}
\label{sec:training-chap4}

To recap, the goal is to perform inference in a layered model of observed variables $\obs$ with latent variables $\mathbf{h}$. Each predictor $\Delta^t$ (with target $t$) is a function of a collection of its contextual messages $\mathbf{c} = \{ c_k \}$ (incoming from the latent layer below $\mathbf{h}^b$), that produces the consensus message $m$, \ie $m = \Delta^t(\mathbf{c}).$

We adopt an approach in which we \textit{learn} a function for this task that is parameterized by $\bm{\theta}$, \ie $\overline{m} \equiv \func(\mathbf{c}|\bm{\theta}).$ This can be seen as an instance of the canonical regression task. For a given family of regressors $\func$, the goal of training is to find parameters $\bm{\theta}$ that capture the relationship between context and consensus message pairs $\{ (\mathbf{c}_d, m_d) \}_{d=1...D}$ in some set of training examples.

\subsubsection{Choice of Predictor Training Data}

First we discuss how this training data is obtained. There can be at least three different sources:

\textbf{1.\,\,Beliefs at Convergence.} Standard message passing inference is run in the model for a large number of iterations until convergence and for a collection of different observations $\{ \obs_d \}$. Message passing is scheduled in precisely the same way as it would be if \MTD were present, however no consensus messages are sent. For each observation $\obs_d$, the collection of the marginals of the latent variables in the layer below the predictor ($\mathbf{h}^b_d = \{ h^b_{dk} \}$, (see \eg Fig.~\ref{fig:types-a}) at the \textit{first} iteration of message passing is considered to be the context $\mathbf{c}_d$, and the marginal of the target variable $t$ at the \textit{last} iteration of message passing is considered to be the oracle message $m_d$. The intuition is that during inference on new problems, a predictor trained in this way would send messages that \textit{accelerate}
convergence to the fixed-point that message passing would have reached by itself anyway. This technique is only useful if standard message passing works but is slow.

\textbf{2.\,\,Samples from the Model.} First a collection of samples from the model is generated, giving us both the observation $\obs_d$ and its corresponding latent variables $\mathbf{h}_d$ for each sample. Standard message passing inference is then run on the observations $\{ \obs_d \}$ only for a single iteration. Message passing is scheduled as before. For each observation $\obs_d$, the marginals of the latent variables in the layer below $\mathbf{h}^b_d$ at the \textit{first} iteration of message passing is the context $\mathbf{c}_d$, and the oracle message $m_d$ is considered to be a point-mass centered at the sampled value of the target variable $t$. The intuition is that during inference on new problems, a predictor trained in this way would send messages that guide inference to a fixed-point in which the marginal of the target variable $t$ is close to its sampled value. This technique is useful if standard message passing fails to reach good fixed points no matter how long it is run for.

\textbf{3.\,\,Labelled Data.} As above, except the latent variables of interest $\mathbf{h}_d$ are set from real data instead of being sampled from the model. The oracle message $m_d$ is therefore a point-mass centered at the label provided for the target variable $t$ for observation $\mathbf{X}_d$. The aim is that during inference on new problems, a predictor trained in this way would send messages that guide inference to a fixed-point in which the marginal of the target variable $t$ is close to its labelled value, even in the presence of a degree of model mismatch. We demonstrate each of the strategies in the experiments in Section~\ref{sec:experiments-chap4}.

\subsubsection{Random Regression Forests for CMP}

Our goal is to learn a mapping $\func$ from contextual messages $\mathbf{c}$ to the consensus message $m$ from training data $\{ (\mathbf{c}_d, m_d) \}_{d=1...D}$. This is challenging since the inputs and outputs of the regression problem are both messages (\ie distributions), and special care needs to be taken to account for this fact. We follow closely the methodology of~\cite{Eslami2014}, which use random forests to predict outgoing messages from a factor given the incoming messages to it. Please refer to Section~\ref{sec:forests} for a brief review of random forests.

In approximate message passing (\eg EP~\cite{Minka2001} and VMP~\cite{Winn2005}), messages can be represented using only a few numbers, \eg a Gaussian message can be represented by its natural parameters. We represent the contextual messages $\mathbf{c}$ collectively, in two different ways: first as a concatenation of the parameters of its constituent messages which we refer to as `regression parameterization' and denote by $\mathbf{r}_\textrm{c}$; and second as a vector of features computed on the set which we refer to as `tree parameterization' and denote by $\mathbf{t}_\textrm{c}$. This parametrization typically contains features of the set as a whole (\eg moments of their means). We represent the outgoing message $m$ by a vector of real valued numbers $\mathbf{r}_\textrm{m}$.

\textbf{Prediction Model.} Each leaf node is associated with a subset of the labelled training data. During testing, a previously unseen set of contextual messages represented by $\mathbf{t}_\textrm{c}$ traverses the tree until it reaches a leaf which by construction is likely to contain similar training examples. Therefore, we use the statistics of the data gathered in that leaf to predict the consensus message with a multivariate regression model of the form: $\mathbf{r}_\textrm{m} = \mathbf{W} \cdot \mathbf{r}_\textrm{c} + \epsilon$ where $\epsilon$ is a vector of normal error terms. We use the learned matrix of coefficients $\mathbf{W}$ at test time to make predictions $\overline{\mathbf{r}}_\textrm{m}$ for a given $\mathbf{r}_\textrm{c}$. To recap, $\mathbf{t}_\textrm{c}$ is used to traverse the contextual messages down to leaves, and $\mathbf{r}_\textrm{c}$ is used by a linear regressor to predict the parameters $\mathbf{r}_\textrm{m}$ of the consensus message.

\textbf{Training Objective Function.} Recall the training procedure of random forests from Section~\ref{sec:forests}. Each node is in a tree represents a partition of feature space and
split function in each node is chosen in a greedy manner minimizing a splitting criterion $E$.
A common split criterion, which we also use here, is the sum of data likelihood in the node's
left and right child clusters (see Eq.~\ref{eqn:forest_energy}). We use the `fit residual' as
defined in~\citep{Eslami2014} as the likelihood (model fit) function for optimizing splits at
each node. In other words, this objective function splits the training data at each node in a way that the relationship between the incoming and outgoing messages is well captured by the regression model in each child.

\textbf{Ensemble Model.} During testing, a set of contextual messages simultaneously traverses every tree in the forest from their roots until it reaches their leaves. Combining the predictions into a single forest prediction might be done by averaging the parameters $\overline{\mathbf{r}}_\textrm{m}^t$ of the predicted messages $\overline{m}^t$ by each tree $t$, however this would be sensitive to the chosen parameterization for the messages. Instead we compute the moment average $\overline{m}$ of the distributions $\{ \overline{m}^t \}$ by averaging the first few moments of the predictions across trees, and solving for the distribution parameters which match the averaged moments (see \eg~\cite{Grosse2013}).

\section{Experiments}
\label{sec:experiments-chap4}

We first illustrate the application of \MTD to two diagnostic models: one of circles and a second of squares. We then use the approach to improve inference in a more challenging vision model: intrinsic images of faces. In the first experiment, the predictors are trained on beliefs at convergence, in the second on samples from the model, and in the third on annotated labels, showcasing various use-cases of \MTD. We show that in all cases, the proposed technique leads to significantly more accurate inference results whilst preserving the computational efficiency of message passing. The experiments were performed using Infer.NET~\cite{InferNET2012} with default settings, unless stated otherwise. For random forest predictors, we set the number of trees in each forest to 8.

\subsection{A Generative Model of Circles}
\label{sec:circle}

\begin{figure}[t]
	\centering
	\subfigure[]{
		\setlength\fboxsep{-0.3mm}
		\setlength\fboxrule{0.5pt}
		\parbox[b]{3.1cm}{
			\fbox{\includegraphics[width=\linewidth]{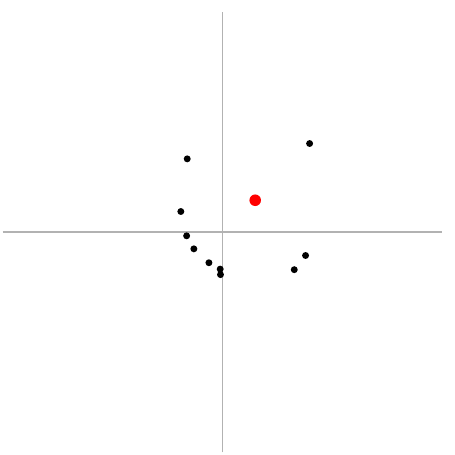}} \\
			\vspace{1mm}
			\fbox{\includegraphics[width=\linewidth]{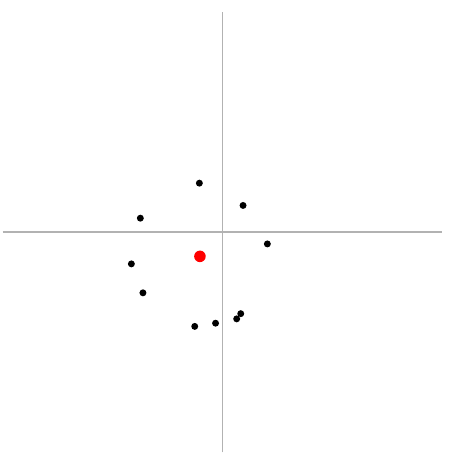}}
		}
		\hspace{0.1cm}
		\label{fig:circle-data}
	}
	\subfigure[]{
		\begin{tikzpicture}

			\draw[red!20] (-2.3, 2.5) -- (2, 2.5);
			\draw[red, dotted] (-2.3, 2.5) -- (2, 2.5);

			\node[obs] 											(x)			{$\mathbf{x}_i$}; %
			\node[latent, above=of x, yshift=-2mm]                			(z)			{$\mathbf{z}_i$}; %

			\factor[above=of x]				 					{noise}		{left:Gaussian} {} {}; %
			\factor[above=of z, yshift=10mm] 					{sum}		{left:Sum} {} {}; %

			\node[latent, right=of sum]                			(c)			{$\mathbf{c}$}; %
			\node[latent, above=of sum, yshift=-7mm]   			(p)			{$\mathbf{p}_i$}; %

			\factor[above=of p] 								{circle}	{above:Circle\,\,\,\,\,} {} {}; %
			\factor[above=of c] 								{pc}		{} {} {}; %

			\node[latent, left=of circle]                		(a)			{$a_i$}; %
			\node[latent, right=of circle]                		(r)			{$r$}; %

			\factor[above=of a] 								{pa}		{} {} {}; %
			\factor[above=of r] 								{pr}		{} {} {}; %

			\factoredge {z} 		{noise} 		{x}; %
			\factoredge {p} 		{sum} 			{}; %
			\factoredge {a} 		{circle} 		{p}; %
			\factoredge {r} 		{circle} 		{}; %
			\factoredge {c} 		{sum} 			{z}; %
			\factoredge {} 			{pc} 			{c}; %
			\factoredge {} 			{pa} 			{a}; %
			\factoredge {} 			{pr} 			{r}; %

			\plate {} {(pa) (a) (p) (z) (x)} {}; %

			\fill (c.south) ++ (0, -0.52) circle (3pt) [fill=red] { };

			\draw [-stealth, red] (1.45, 2.65) -- (1.45, 2.95);

			\node[yshift=-0.9cm] at (c.south) { $\textcolor{red}{\Delta^c}$ };)

		\end{tikzpicture}
		\label{fig:circle-model}
	}
	\mycaption{The circle problem}{(a)~Given a sample of points on a circle (black), we wish to infer the circle's center (red) and its radius. Two sets of samples are shown. (b)~The graphical model for this problem.}
	\label{fig:circle}
\end{figure}

\begin{figure}[t]
	\centering
	\subfigure[Center]{
		\includegraphics[width=0.46\linewidth]{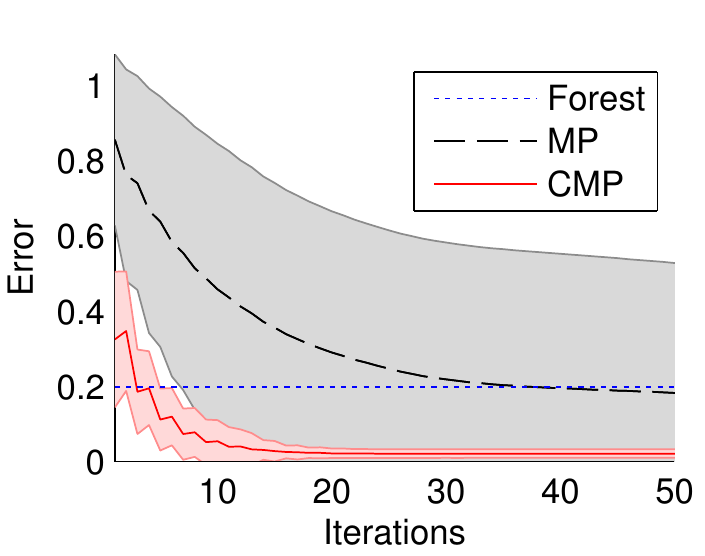}
	}
	\subfigure[Radius]{
		\includegraphics[width=0.46\linewidth]{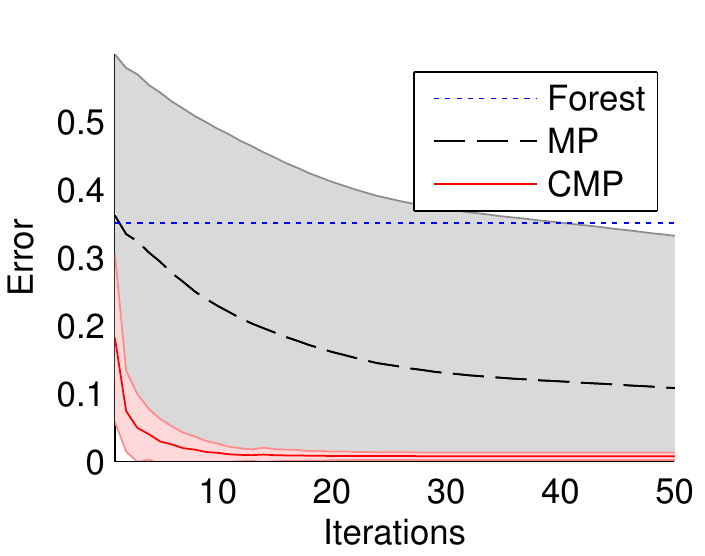}
	}
	\mycaption{Accelerated inference using \MTD for the circle problem}{(a)~Distance of the mean of the marginal posterior of center $c$ from its true value as a function of number of inference iterations (Forest: direct prediction, MP: standard VMP, \MTD: VMP with consensus). \Method significantly accelerates convergence. (b)~Similar plot for radius $r$. }
	\label{fig:circle-results}
\end{figure}

We begin by studying the behavior of standard message passing on a simplified Gauss and Ceres problem~\citep{Teets1999}.
Given a noisy sample of points $\obs = \{ \mathbf{x}_i \}_{i=1...N}$ on a circle in the 2D plane (Fig.~\ref{fig:circle-data}, black, $\mathcal{N}(0,0.01)$ noise on each axis), the aim is to infer the coordinates of the circle's center $\mathbf{c}$ (Fig.~\ref{fig:circle-data}, red) and its radius $r$. We can express the data generation process using a graphical model (Fig.~\ref{fig:circle-model}). The Cartesian point $(0, r)$ is rotated $a_i$ radians to generate $\mathbf{p}_i$, then translated by $\mathbf{c}$ to generate the latent $\mathbf{z}_i$, which finally produces the noisy observation $\mathbf{x}_i$.  This model can be expressed in a few lines of code in Infer.NET. The circle model is interesting for our purposes ,since it is both layered (the $\mathbf{z}_i$s, $\mathbf{p}_i$s and $a_i$s each form a layer) and loopy (due to the presence of two variables outside the plate).

We use this example to highlight the fact that although inference may require many iterations of message passing, message initialization can have a significant effect on the speed of convergence, and to demonstrate how this can be done automatically using \MTD.

Vanilla message passing inference in this model can take a surprisingly large number of iterations to converge. We draw 10 points $\{ \mathbf{x}_i \}$ from circles with random centers and radii, run VMP and record the accuracy of the marginals of the latent variables at each iteration. We repeat the experiment 50 times and plot results in Fig.~\ref{fig:circle-results} (dashed black). As can be seen from the figure, the marginals contain significant errors even after 50 iterations of message passing.

We then experiment with \method. A predictor $\Delta^c$ is trained to send a consensus message to $\mathbf{c}$ in the initial stages of inference, given the messages coming up from all of the $\mathbf{z}_i$ (indicated graphically in Fig.~\ref{fig:circle-model}, red). The predictor is trained on final beliefs at 100 iterations of standard message passing on $D=500$ sample problems.

As can be seen in Fig.~\ref{fig:circle-results} (red), this single consensus message has the effect of significantly increasing the rate of convergence (as indicated by slope) and also inference robustness (as indicated by error bars). For comparison, we also plot how well a regressor of the same capacity as the one used by \MTD can directly estimate the latent variables without using the graphical model in Fig.~\ref{fig:circle-results} (blue). \Method gives us the best of both worlds in this example: speed that is more comparable to one-shot bottom-up prediction and the accuracy of message passing inference in a good model for the problem.

\begin{figure}[t]
	\centering
	\subfigure[]{
		\setlength\fboxsep{-0.2mm}
		\setlength\fboxrule{0.7pt}
		\parbox[b]{2.3cm}{
			\fbox{\includegraphics[width=0.45\linewidth]{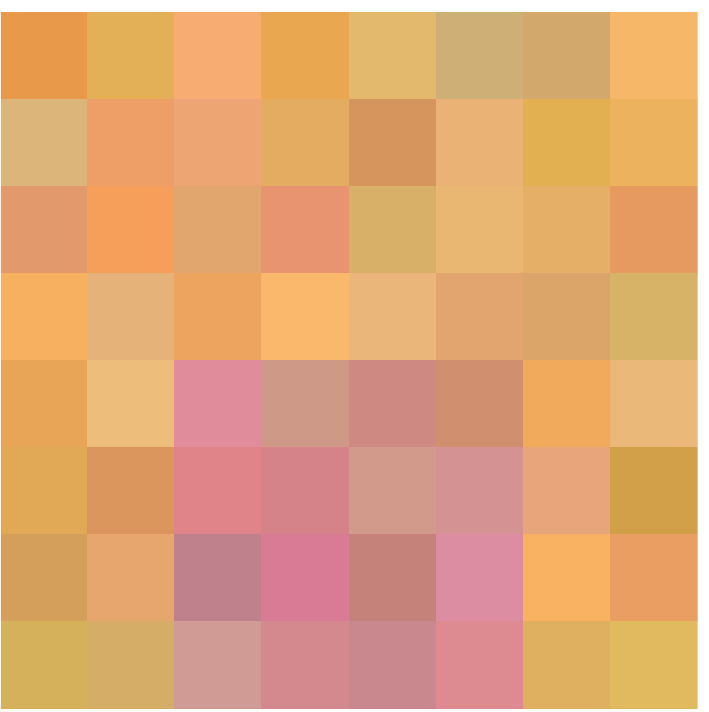}} \hspace*{0mm}
			\fbox{\includegraphics[width=0.45\linewidth]{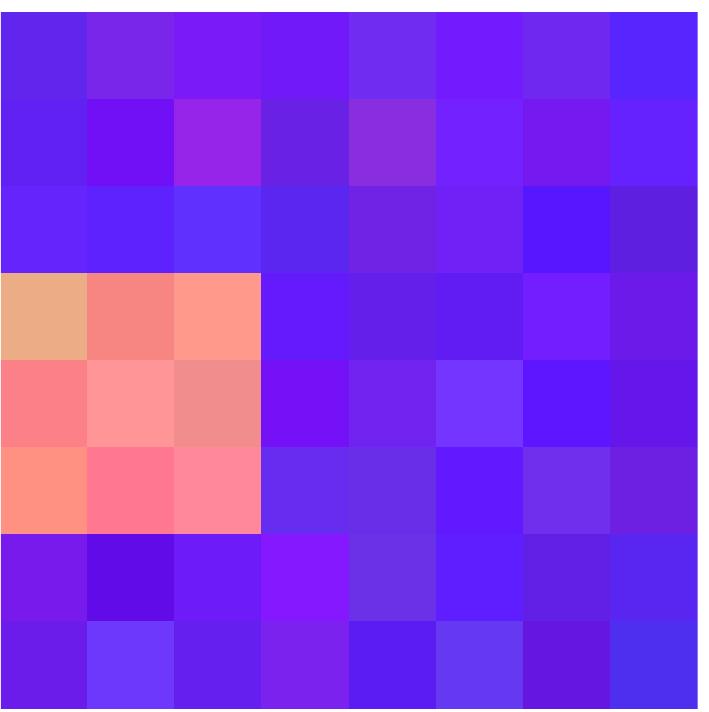}} \vspace*{-2.4mm} \\
			\fbox{\includegraphics[width=0.45\linewidth]{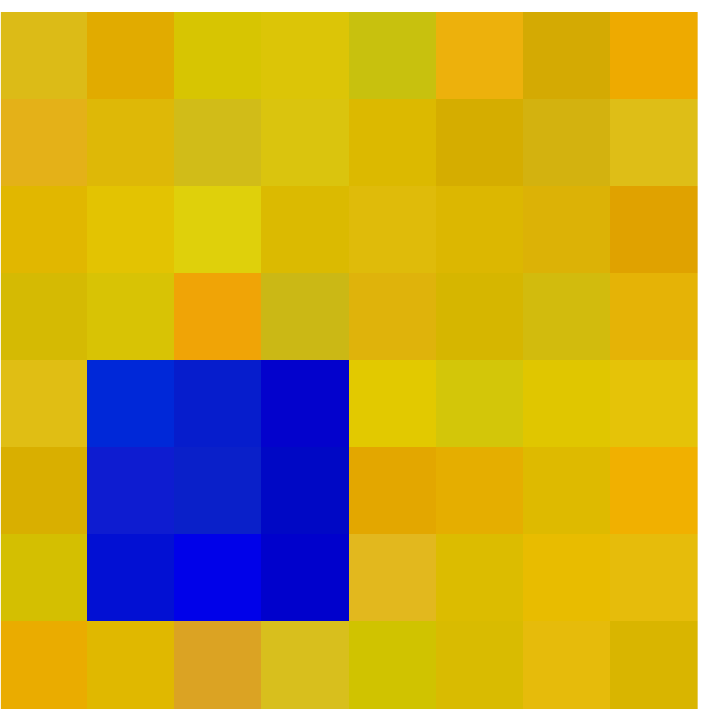}} \hspace*{0mm}
			\fbox{\includegraphics[width=0.45\linewidth]{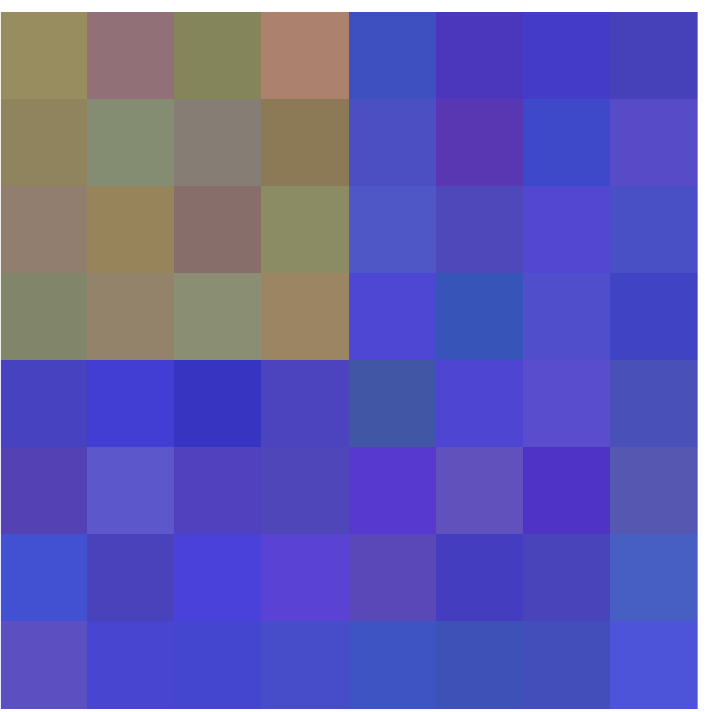}} \vspace*{-2.4mm} \\
			\fbox{\includegraphics[width=0.45\linewidth]{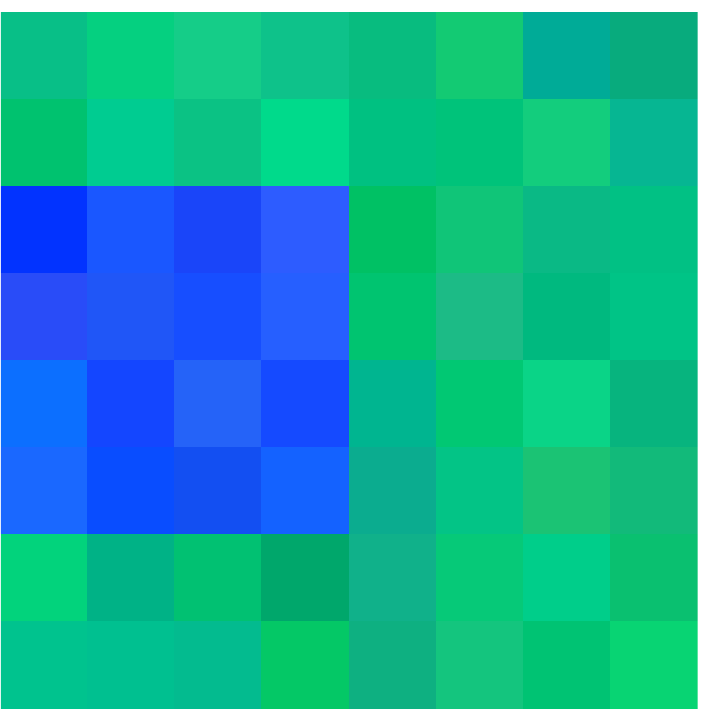}} \hspace*{0mm}
			\fbox{\includegraphics[width=0.45\linewidth]{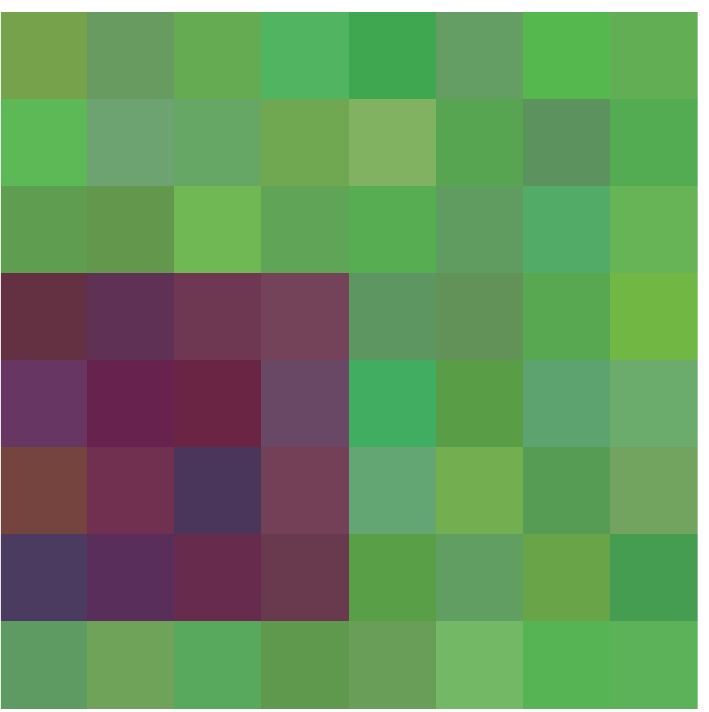}} \vspace*{-2.4mm} \\
			\fbox{\includegraphics[width=0.45\linewidth]{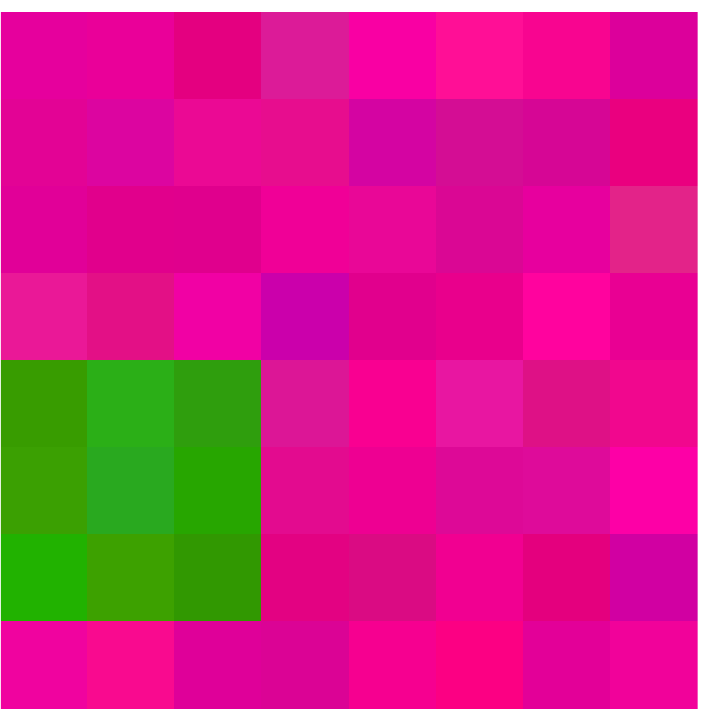}} \hspace*{0mm}
			\fbox{\includegraphics[width=0.45\linewidth]{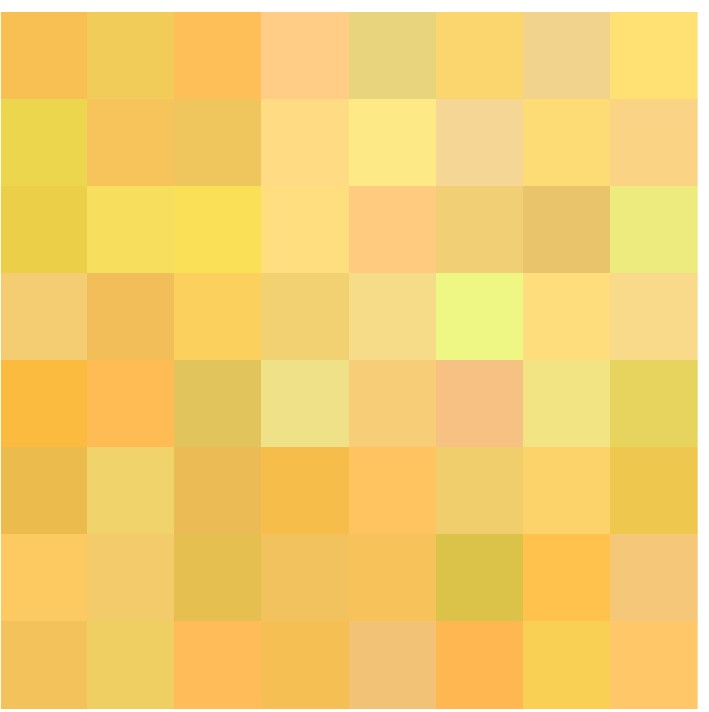}} \vspace*{-2.4mm} \\
			\fbox{\includegraphics[width=0.45\linewidth]{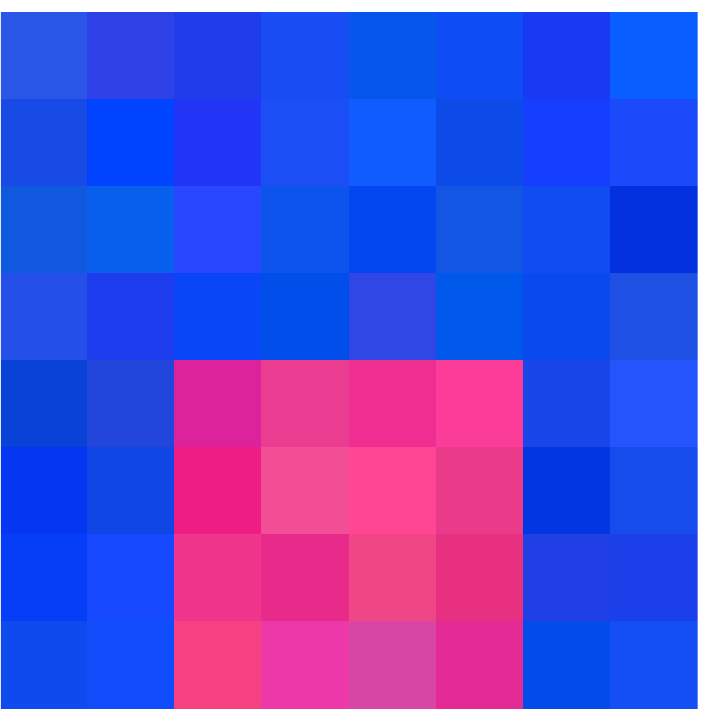}} \hspace*{0mm}
			\fbox{\includegraphics[width=0.45\linewidth]{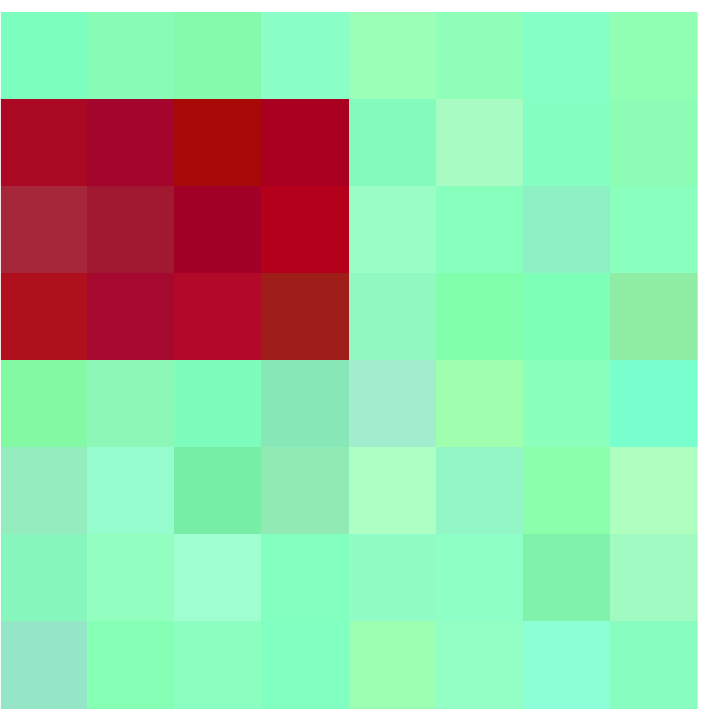}} \vspace*{-2.4mm} \\
			\fbox{\includegraphics[width=0.45\linewidth]{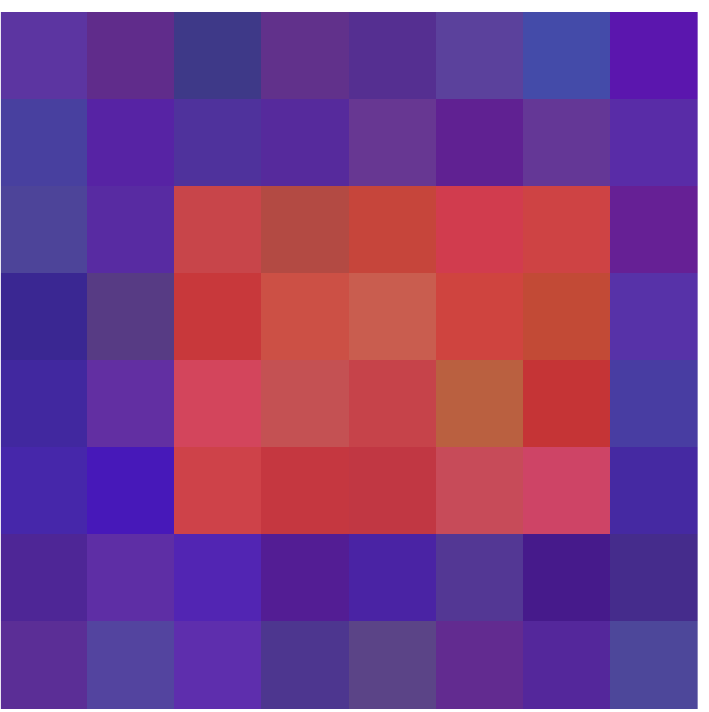}} \hspace*{0mm}
			\fbox{\includegraphics[width=0.45\linewidth]{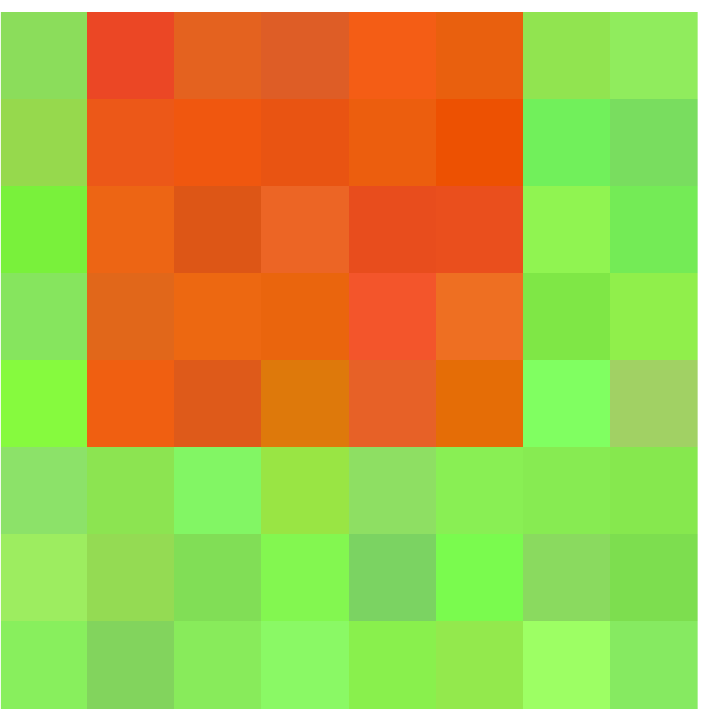}} \vspace*{-2.4mm} \\
			\fbox{\includegraphics[width=0.45\linewidth]{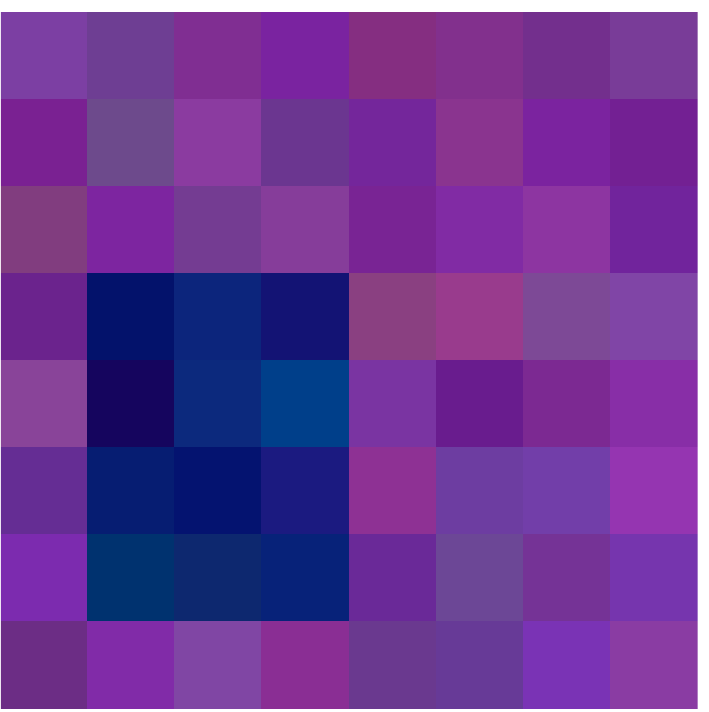}} \hspace*{0mm}
			\fbox{\includegraphics[width=0.45\linewidth]{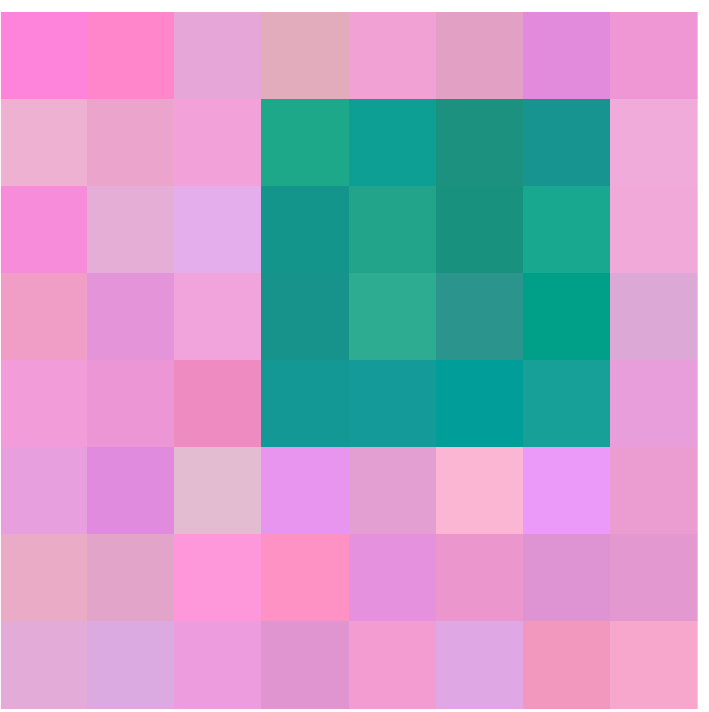}}
		}
		\hspace{0.2cm}
		\label{fig:square-data}
	}
	\subfigure[]{
		\begin{tikzpicture}

			\draw[red!20] (-2, 2.7) -- (2, 2.7);
			\draw[red, dotted] (-2, 2.7) -- (2, 2.7);

			\draw[red!20] (-2, 5.5) -- (2, 5.5);
			\draw[red, dotted] (-2, 5.5) -- (2, 5.5);

			\node[obs] 											(x)			{$\mathbf{x}_i$}; %
			\node[latent, above=of x]                			(z)			{$\mathbf{z}_i$}; %

			\factor[above=of x, yshift=1mm] 					{noise}		{left:Gaussian} {} {}; %
			\factor[above=of z, yshift=10mm] 					{gate}		{below left:Gate} {} {}; %

			\node[latent, right=of gate]                		(fg)		{$\mathrm{\mathbf{fg}}$}; %
			\node[latent, left=of gate]                			(bg)		{$\mathrm{\mathbf{bg}}$}; %
			\node[latent, above=of gate, yshift=-5mm]			(s)			{$s_i$}; %

			\factor[above=of s, yshift=10mm] 					{insq}		{below left, yshift=-2mm:Square} {} {}; %
			\factor[above=of fg] 								{pfg}		{} {} {}; %
			\factor[above=of bg] 								{pbg}		{} {} {}; %

			\node[latent, left=of insq]                			(c)			{$\mathbf{c}$}; %
			\node[latent, right=of insq]                		(l)			{$l$}; %
			\node[latent, above=of insq, yshift=-5mm, draw=none](p)			{$p_i$}; %

			\factor[above=of c] 								{pc}		{} {} {}; %
			\factor[above=of l] 								{pr}		{} {} {}; %

			\factoredge {z} 		{noise} 		{x}; %
			\factoredge {s} 		{gate} 			{z}; %
			\factoredge {c} 		{insq} 			{}; %
			\factoredge {l} 		{insq} 			{}; %
			\factoredge {p} 		{insq} 			{s}; %
			\factoredge {fg} 		{gate} 			{}; %
			\factoredge {bg} 		{gate} 			{}; %
			\factoredge {} 			{pfg} 			{fg}; %
			\factoredge {} 			{pbg} 			{bg}; %
			\factoredge {} 			{pc} 			{c}; %
			\factoredge {} 			{pr} 			{l}; %

			\plate {} {(p) (x)} {}; %

			\fill (bg.south) ++ (0, -0.52) circle (3pt) [fill=red] { };
			\fill (fg.south) ++ (0, -0.52) circle (3pt) [fill=red] { };

			\fill (l.south) ++ (0, -0.52) circle (3pt) [fill=red] { };

			\draw [-stealth, red] (-1.45, 2.85) -- (-1.45, 3.15);
			\draw [-stealth, red] (1.45, 2.85) -- (1.45, 3.15);

			\draw [-stealth, red] (1.45, 5.65) -- (1.45, 5.95);

			\node[yshift=-0.9cm] at (fg.south) { $\textcolor{red}{\Delta^{\mathrm{\mathbf{fg}}}}$ };
			\node[yshift=-0.9cm] at (bg.south) { $\textcolor{red}{\Delta^{\mathrm{\mathbf{bg}}}}$ };

			\node[yshift=-0.9cm] at (l.south) { $\textcolor{red}{\Delta^l}$ };)

		\end{tikzpicture}
		\hspace{0.1cm}
		\label{fig:square-model}
	}
	\mycaption{The square problem}{(a)~We wish to infer the square's center and its side length. (b)~A graphical model for this problem. $s_i$ is a boolean variable indicating the square's presence at position $p_i$. Depending on the value of $s_i$, the gate copies the appropriate color ($\mathrm{\mathbf{fg}}$ or $\mathrm{\mathbf{bg}}$) to $\mathbf{z}_i$.}
	\label{fig:square}
\end{figure}

\begin{figure}[t]
	\centering
	\subfigure[Center]{
		\includegraphics[width=0.4\linewidth]{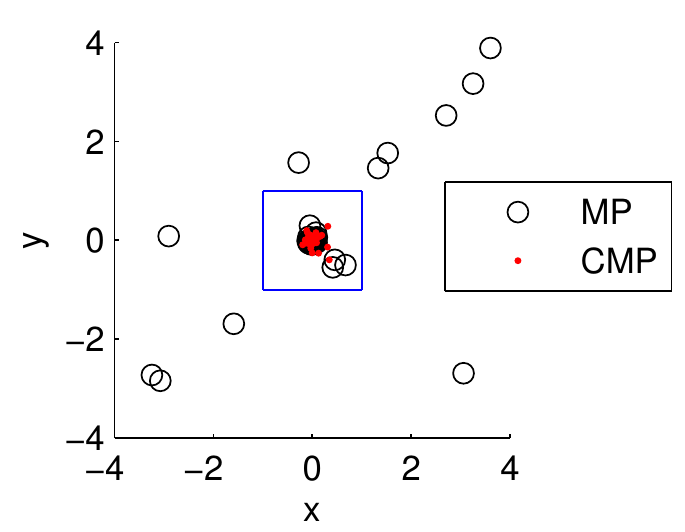}
		\label{fig:square-results-bullseye}
	}
	\subfigure[Center]{
		\includegraphics[width=0.4\linewidth]{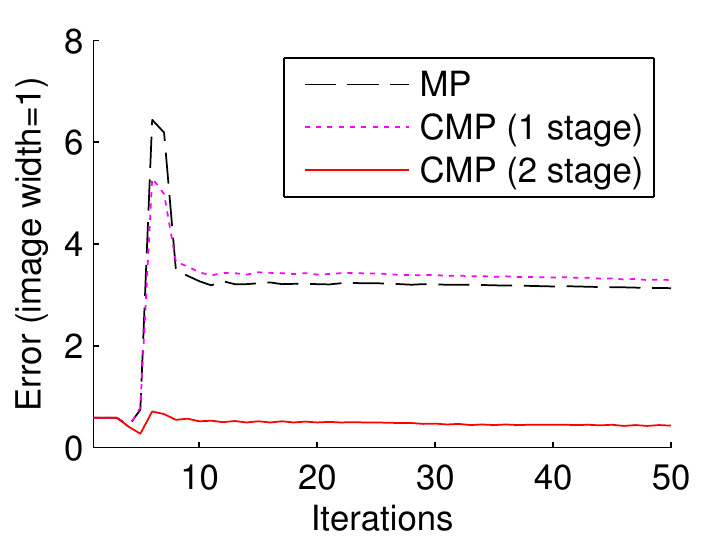}
		\label{fig:square-results-center}
	}
	\subfigure[Side length]{
		\includegraphics[width=0.4\linewidth]{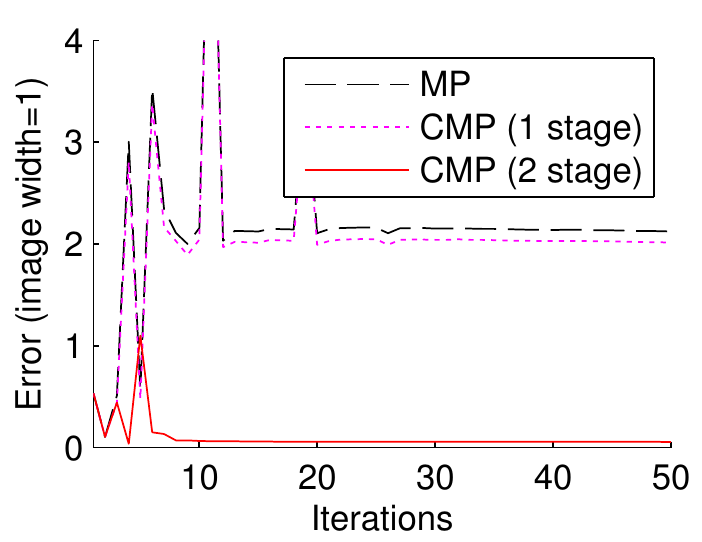}
		\label{fig:square-results-radius}
	}
	\subfigure[BG color]{
		\includegraphics[width=0.4\linewidth]{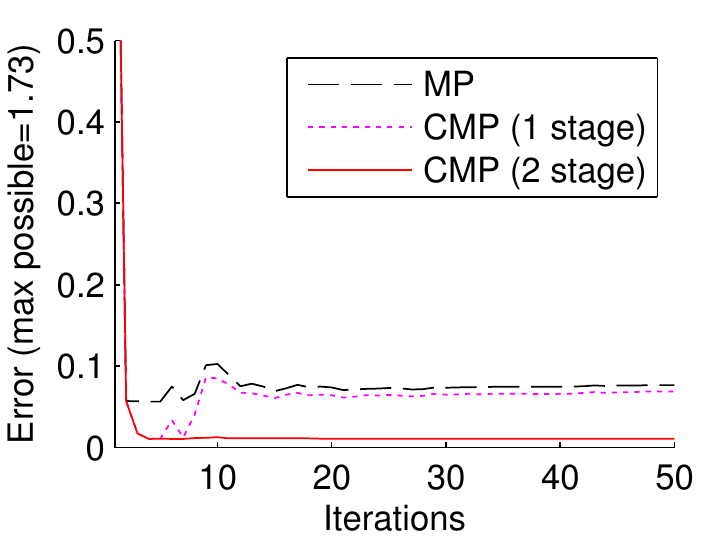}
		\label{fig:square-results-bgcolor}
	}
	\mycaption{Robustified inference using \MTD for the square problem}{(a)~Position of inferred centers relative to ground-truth. Image boundaries shown in blue for scale. (b,c,d)~Distance of the mean of the posterior of $\mathbf{c}$, $l$ and $\mathrm{\mathbf{bg}}$ from their true values. \MTD consistently increases inference accuracy. Results have been averaged over 50 different problems. 1 stage \MTD only makes use of the lower predictors $\Delta^\mathrm{\mathbf{fg}}$ and $\Delta^\mathrm{\mathbf{bg}}$.}
	\label{fig:square-results}
\end{figure}

\subsection{A Generative Model of Squares}
\label{sec:square}

Next, we turn our attention to a more challenging problem for which even the best message passing scheme that we could devise frequently finds completely inaccurate solutions. The task is to infer the center $\mathbf{c}$ and side length $r$ of a square in an image (Fig.~\ref{fig:square-data}). Unlike the previous problem where we knew that all points belonged to the circle, here we must first determine which pixels belong to the square and which do not. To do so we might also wish to reason about the color of the foreground $\mathrm{\mathbf{fg}}$ and background $\mathrm{\mathbf{bg}}$, making the task of inference significantly harder. The graphical model for this problem is shown in Fig.~\ref{fig:square-model}. Let $\mathbf{c}$ and $l$ denote square center and side length respectively. At each pixel position $p_i$, $s_i$ is a boolean variable indicating the square's presence. Depending on the value of $s_i$, the gate copies the appropriate color ($\mathrm{\mathbf{fg}}$ or $\mathrm{\mathbf{bg}}$) to $\mathbf{z}_i$.

We experiment with 50 test images (themselves samples from the model), perform inference using EP and with a sequential schedule, recording the accuracy of the marginals of the latent variables at each iteration. We additionally place damping with step size 0.95 on messages from the square factor to the center $\mathbf{c}$. We found these choices led to the best performing standard message passing algorithm. Despite this, we observed inference accuracy to be disappointingly poor (see Fig.~\ref{fig:square-results}). In Fig.~\ref{fig:square-results-bullseye} we see that, for many images, message passing converges to highly inaccurate marginals for the center. The low quality of inference can also be seen in quantitative results of Figs.~\ref{fig:square-results}(b-d).

We implement \MTD predictors at two different layers of the model (see Fig.~\ref{fig:square-model}, red). In the first layer, $\Delta^\mathrm{\mathbf{fg}}$ and $\Delta^\mathrm{\mathbf{bg}}$ send consensus messages to $\mathrm{\mathbf{fg}}$ and $\mathrm{\mathbf{bg}}$ respectively, given the messages coming up from all of the $\mathbf{z}_i$ which take the form of independent Gaussians centered at the appearances of the observed pixels (we use a Gaussian noise model). Therefore $\Delta^\mathrm{\mathbf{fg}}$ and $\Delta^\mathrm{\mathbf{bg}}$ effectively make initial guesses of the values of the foreground and background colors in the image given the observed image. Split features in the internal nodes of the regression forest are designed to test for equality of two randomly chosen pixel positions, and sparse regressors are used at the leaves to prevent overfitting.

In the second layer, $\Delta^l$ sends a consensus message to $l$ given the messages coming up from all of the $s_i$. The messages from $s_i$ take the form of independent Bernoullis indicating the algorithm's current beliefs about the presence of the square at each pixel. Therefore, the predictor's job is to predict the square's side length from this probabilistic segmentation map. Note that it is much easier to implement a regressor to perform this task (effectively one only needs to count) than it is to do so using the original observed image pixels $x_i$. We find these predictors to be sufficient for stable inference and so we do not implement a fourth predictor for $\mathbf{c}$. We experiment with single stage \MTD, where only the lower predictors $\Delta^\mathrm{\mathbf{fg}}$ and $\Delta^\mathrm{\mathbf{bg}}$ are active, and with two stage \MTD, where all three predictors are active. The predictors are trained on $D=500$ samples from the model.

The results of these experiments are shown in Fig.~\ref{fig:square-results}. We observe that \MTD significantly improves the accuracy of inference for the center $\mathbf{c}$ (Figs.~\ref{fig:square-results-bullseye}, \ref{fig:square-results-center}) but also for the other latent variables (Figs.~\ref{fig:square-results-radius}, \ref{fig:square-results-bgcolor}). Note that single stage \MTD appears to be insufficient for guiding message passing to good solutions. Whereas in circle example \MTD accelerated convergence, this example demonstrates how it can make inference possible in models that were outside the capabilities of standard message passing.

\subsection{A Generative Model of Faces}
\label{sec:shading}

\begin{figure}[t]
	\centering
	\subfigure[]{
		\setlength\fboxsep{-0.3mm}
		\setlength\fboxrule{0pt}
		\parbox[b]{3cm}{
			\centering
			\small
			\fbox{\includegraphics[width=1.3cm]{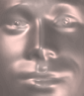}}\\
			Normal $\{\mathbf{n}_i \}$ \vspace{4.5mm} \\
			\fbox{\includegraphics[width=1.3cm]{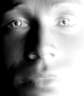}} \\
			Shading $\{ s_i \}$ \vspace{4.5mm} \\
			\fbox{\includegraphics[width=1.3cm]{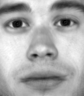}} \\
			Reflectance $\{ r_i \}$ \vspace{4.5mm} \\
			\fbox{\includegraphics[width=1.3cm]{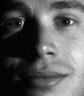}} \\
			Observed image $\{ x_i \}$\\
		}
		\hspace{0.1cm}
		\label{fig:shading-data}
	}
	\subfigure[]{
		\begin{tikzpicture}

			\draw[red!20] (-2.4, 2.7) -- (2, 2.7);
			\draw[red, dotted] (-2.4, 2.7) -- (2, 2.7);

			\draw[red!20] (-2.4, 5.5) -- (2, 5.5);
			\draw[red, dotted] (-2.4, 5.5) -- (2, 5.5);

			\node[obs] 											(x)			{$x_i$}; %
			\node[latent, above=of x]                			(z)			{$z_i$}; %

			\factor[above=of z, yshift=10mm] 					{times}		{below left: $\times$} {} {}; %

			\node[latent, above=of times, yshift=-5mm]			(s)			{$s_i$}; %

			\factor[above=of x, yshift=1mm] 					{noise1}	{left:Gaussian} {} {}; %

			\node[latent, left=of times]                		(r)			{$r_i$}; %

			\factor[above=of r] 								{pr}		{} {} {}; %

			\factor[above=of s, yshift=10mm] 					{inner}		{left:Product} {} {}; %

			\node[latent, above=of inner, yshift=-5mm] 			(n)			{$\mathbf{n_i}$}; %
			\node[latent, right=of inner]                		(l)			{$\mathbf{l}$}; %

			\factor[above=of l] 								{pl}		{} {} {}; %

			\factor[above=of n] 								{pn}		{} {} {}; %

			\factoredge {z} 		{noise1} 		{x}; %
			\factoredge {s} 		{times} 		{z}; %
			\factoredge {r} 		{times} 		{}; %
			\factoredge {n} 		{inner} 		{s}; %
			\factoredge {l} 		{inner} 		{}; %
			\factoredge {} 			{pn} 			{n}; %
			\factoredge {} 			{pl} 			{l}; %
			\factoredge {} 			{pr} 			{r}; %

			\plate {} {(pn) (x) (r)} {}; %

			\fill (r.south) ++ (0, -0.52) circle (3pt) [fill=red] { };

			\fill (l.south) ++ (0, -0.52) circle (3pt) [fill=red] { };

			\draw [-stealth, red] (-1.45, 2.85) -- (-1.45, 3.15);

			\draw [-stealth, red] (1.45, 5.65) -- (1.45, 5.95);

			\node[yshift=-0.9cm] at (r.south) { $\textcolor{red}{\Delta_i^\mathbf{r}}$ };

			\node[yshift=-0.9cm] at (l.south) { $\textcolor{red}{\Delta^\mathbf{l}}$ };

			\node[yshift=0.35cm, xshift=-0.83cm] at (inner) { Inner };

		\end{tikzpicture}
		\label{fig:shading-model}
	}
	\mycaption{The face problem}{(a)~We observe an image and wish to infer the corresponding reflectance map and normal map (visualized here as 3D shape). (b)~A graphical model for this problem. Symmetry priors not shown.}
	\label{fig:shading}
	\vspace*{-4mm}
\end{figure}

In this sectin, we also investigate a more realistic application: face modeling. The estimation of reflectance and shape from a single image of a human face is a well-studied problem in computer vision (see \eg \citep{Georghiades2001, Lee2005, Wang2009, Kemelmacher2011, Tang2012}). A primary motivation for this task is that reflectance and shape are invariant to confounding light effects, and are therefore useful for downstream tasks such as recognition. The problem is ill-posed and modern approaches make use of prior knowledge in order to obtain good solutions, \eg in the form of average reflectance and normal statistics \citep{Biswas2009, Biswas2010} or morphable 3D models \citep{Zhang2006, Wang2009}.

\textbf{Model.} Given an observation of image pixels $\mathbf{x} = \{x_i\}$, the aim is to infer the reflectance value $r_i$ and normal vector $\mathbf{n_i}$ for each pixel $i$ (see Fig.~\ref{fig:shading-data}). In Fig.~\ref{fig:shading-model}, a model is shown for these variables that represents the following image formation process: $x_i = (\mathbf{n_i} \cdot \mathbf{l}) \times r_i + \epsilon$, thereby assuming Lambertian reflection and an infinitely distant directional light source with variable intensity. We place Gaussian priors over reflectances $\{ r_i \}$, normals $\{ \mathbf{n_i} \}$, and the light $\mathbf{l}$; and set the parameters of the priors using training data. We additionally place a soft symmetry prior on the $\{ r_i \}$ (the reflectance value on one side of the face should be close to its value on the other side) and on the $\{ \mathbf{n_i} \}$ (normal vectors on each side should be approximately symmetric), reflecting our prior knowledge about faces. These symmetry priors can be added to the model in just a few lines of code, illustrating the way in which model-based methods lend themselves to rapid prototyping and experimentation.

Although this model is only a crude approximation to the true image formation process (\eg it does not account for shadows or specularities), similar approximations have been found to be useful in prior work \citep{Biswas2009, Biswas2010, Kemelmacher2011}. Additionally, if we can successfully develop algorithms that perform accurate and reliable inference in this class of models, we would then be able to increase its usefulness by updating it to reflect the true image formation process more accurately. Note that even for a relatively small image of size $96 \times 84$, the model contains over 48,000 latent variables and 56,000 factors, and as we will show below, standard message passing in the model routinely fails to converge to accurate solutions.

\begin{figure*}
\begin{center}
\centerline{\includegraphics[width=1.0\columnwidth]{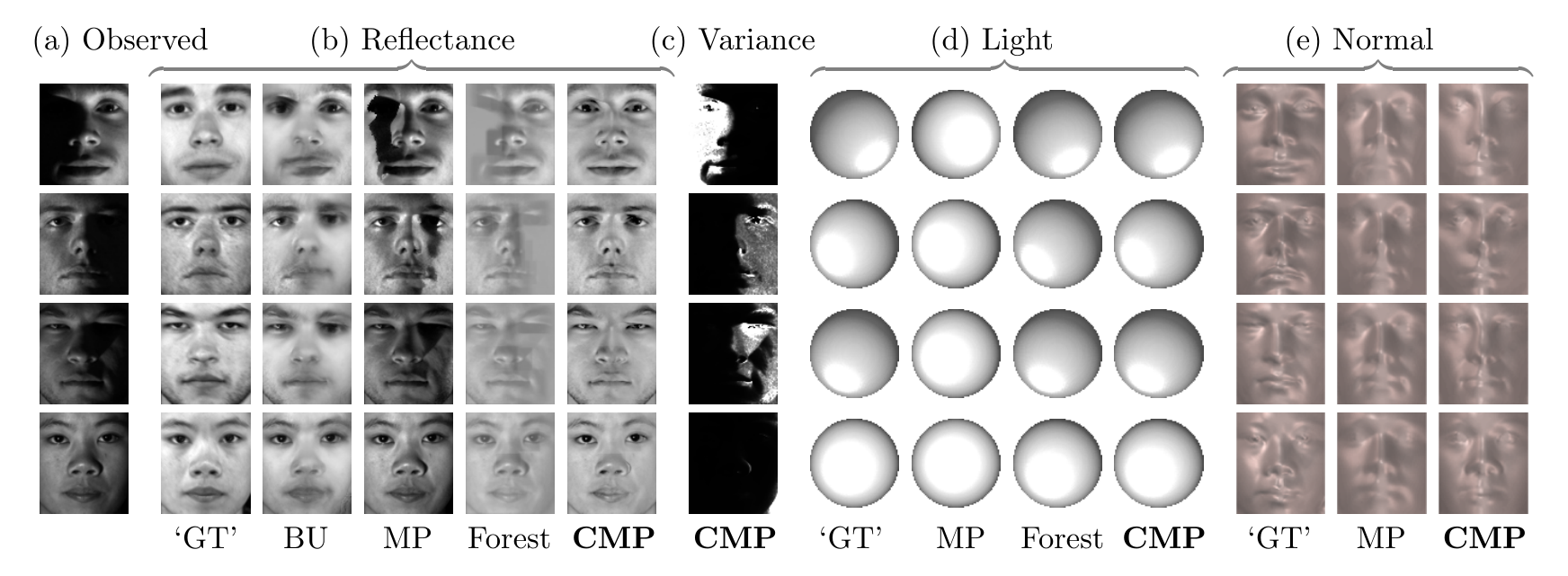}}
\vspace{-1.2cm}
\end{center}
	\mycaption{A visual comparison of inference results for the face problem}{For 4 randomly chosen test images, we show inference results obtained by competing methods. (a)~Observed images. (b)~Inferred reflectance maps. \textit{GT} is the stereo estimate which we use as a proxy for ground-truth, \textit{BU} is the bottom-up reflectance estimate of Biswas \etal (2009), \textit{MP} refers to standard variational message passing, \textit{Forest} is the consensus prediction and \textit{CMP} is the proposed consensus message passing technique. (c)~The variance of the inferred reflectance estimate produced by \MTD (normalized across rows). High variance regions correlate strongly with cast shadows. (d)~Visualization of inferred light. (e)~Inferred normal maps.}
	\label{fig:shading-qualitative-multiple-subjects}
\end{figure*}

\textbf{Consensus Message Passing.} We use predictors at two levels in the model (see Fig.~\ref{fig:shading-model}) to tackle this problem. The first sends consensus messages to \textit{each} reflectance pixel $r_i$, making it an instance of type B of \MTD as described in Fig.~\ref{fig:types-b}. Here, each consensus message is predicted using information from all the contextual messages from the $z_i$. We denote each of these predictors by $\Delta_i^\mathbf{r}$. The second predictor sends a consensus message to $\mathbf{l}$ using information from all the messages from the $s_i$ and is denoted by $\Delta^\mathbf{l}$. The first level of predictors effectively make a guess of the reflectance image from the denoised observation, and the second layer predictor produces an estimate of the light from the shading image (which is likely to be easier to do than directly from the observation). The reflectance predictors $\{ \Delta_i^\mathbf{r} \}$ are all powered by a single random forest, however the pixel position $i$ is used as a feature that it can exploit to create location specific behaviour. The tree parameterization of the contextual messages $\mathbf{c}$ for use in the reflectance predictor $\Delta_i^\mathbf{r}$ also includes 16 features such as mean, median, max, min and gradients of a $21 \times 21$ patch around the pixel. The tree parameterization of the contextual messages for use in the lighting predictor $\Delta^\mathbf{l}$ consists of means of the mean of the shading messages in $12 \times 12$ blocks. We deliberately use simple features to maintain generality but one could imagine the use of more specialized regressors for maximal performance.

\begin{figure}
	\centering
	\setlength\fboxsep{0.2mm}
	\setlength\fboxrule{0pt}
	\begin{tikzpicture}

		\matrix at (0, 0) [matrix of nodes, nodes={anchor=east}, column sep=-0.05cm, row sep=-0.2cm]
		{
			\fbox{\includegraphics[width=1cm]{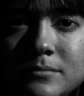}} &
			\fbox{\includegraphics[width=1cm]{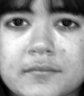}} &
			\fbox{\includegraphics[width=1cm]{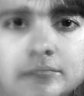}}  &
			\fbox{\includegraphics[width=1cm]{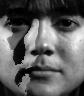}}  &
			\fbox{\includegraphics[width=1cm]{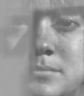}}  &
			\fbox{\includegraphics[width=1cm]{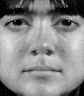}}  &
			\fbox{\includegraphics[width=1cm]{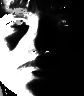}}
			 \\

			\fbox{\includegraphics[width=1cm]{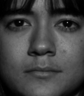}} &
			\fbox{\includegraphics[width=1cm]{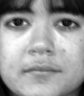}} &
			\fbox{\includegraphics[width=1cm]{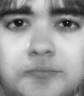}}  &
			\fbox{\includegraphics[width=1cm]{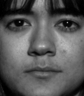}}  &
			\fbox{\includegraphics[width=1cm]{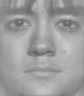}}  &
			\fbox{\includegraphics[width=1cm]{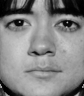}}  &
			\fbox{\includegraphics[width=1cm]{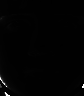}}
			 \\

			\fbox{\includegraphics[width=1cm]{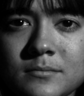}} &
			\fbox{\includegraphics[width=1cm]{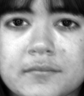}} &
			\fbox{\includegraphics[width=1cm]{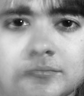}} &
			\fbox{\includegraphics[width=1cm]{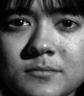}}  &
			\fbox{\includegraphics[width=1cm]{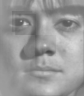}}  &
			\fbox{\includegraphics[width=1cm]{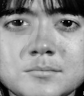}}  &
			\fbox{\includegraphics[width=1cm]{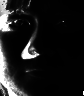}}
			 \\
	     };

	     \node at (-3.85, -2.0) {\small Observed};
	     \node at (-2.55, -2.0) {\small `GT'};
	     \node at (-1.27, -2.0) {\small BU};
	     \node at (0.0, -2.0) {\small MP};
	     \node at (1.27, -2.0) {\small Forest};
	     \node at (2.55, -2.0) {\small \textbf{CMP}};
	     \node at (3.85, -2.0) {\small Variance};

	\end{tikzpicture}
	\mycaption{Robustness to varying illumination}{Left to right: observed image, photometric stereo estimate (proxy for ground-truth), \cite{Biswas2009} estimate, VMP result, consensus forest estimate, CMP mean, and CMP variance.}
	\label{fig:shading-qualitative-same-subject}
  \vspace{-0.3cm}
\end{figure}

\begin{figure}[t]
	\centering
	\subfigure[Without shadows]{
		\includegraphics[width=0.4\linewidth]{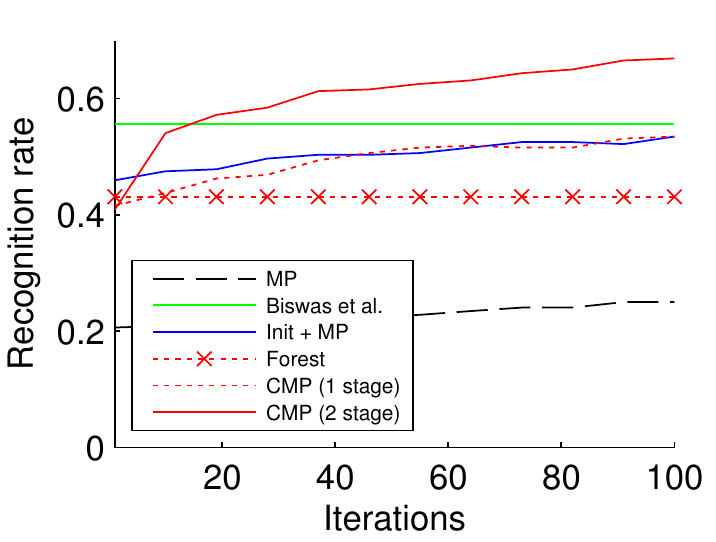}
	}
	\subfigure[With shadows]{
		\includegraphics[width=0.4\linewidth]{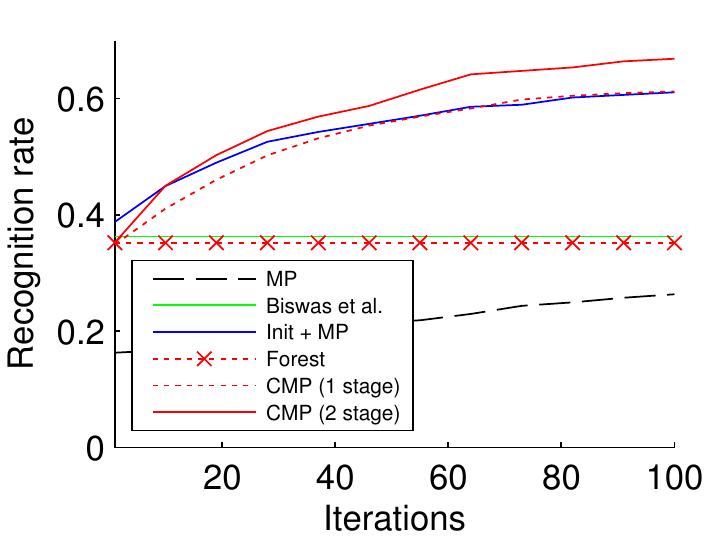}
	}
	\mycaption{Reflectance inference accuracy demonstrated through recognition accuracy}{\MTD allows us to make use of the full potential of the generative model, thereby outperforming the competitive bottom-up method of \cite{Biswas2009}.}
	\label{fig:shading-quantitative-reflectance}

	\subfigure[Without shadows]{
		\includegraphics[width=0.4\linewidth]{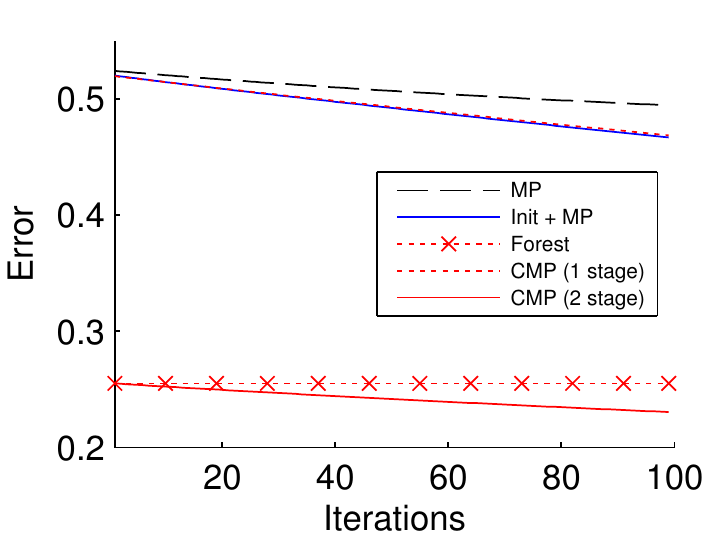}
	}
	\subfigure[With shadows]{
		\includegraphics[width=0.4\linewidth]{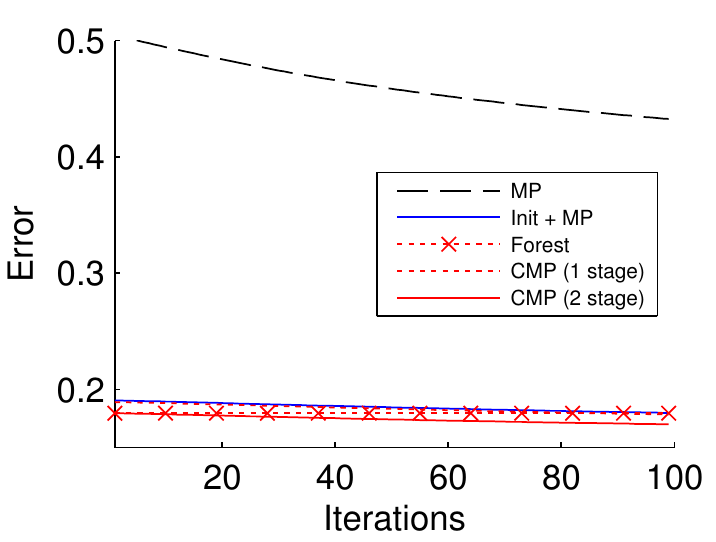}
	}
	\mycaption{Light inference accuracy}{The presence of cast shadows makes the direct prediction task easier, however \MTD is accurate even in their absence.}
	\label{fig:shading-quantitative-light}
\end{figure}

\textbf{Datasets.} We experiment with the `Yale B' and `Extended Yale B' datasets~\citep{Georghiades2001, Lee2005}. Together, they contain images of 38 subjects each with 64 illumination directions. We remove images taken with extreme light angles (azimuth or elevation $\ge 85$ degrees) that are almost entirely in shadow, leaving around 45 images for each subject. Images are down-sampled to $96 \times 84$. There are no ground-truth normals or reflectances for this dataset, however it is common practice to create proxy ground-truths using photometric stereo, which we obtain using the code of~\cite{Queau2013}. We use images from 22 subjects for training and test on the remaining 16 subjects.

\textbf{Results.} We begin by qualitatively assessing the different inference schemes. In Fig.~\ref{fig:shading-qualitative-multiple-subjects} we show inference results for reflectance maps, normal maps and lights that are obtained after 100 iterations of message passing (VMP). For reflectance (Fig.~\ref{fig:shading-qualitative-multiple-subjects}b), we would like inference to produce estimates that match closely the ground-truth produced by photometric stereo (GT). We also display the reflectance estimates produced by the strong baseline (BU) of \cite{Biswas2009} for reference. We note that the baseline achieves excellent accuracy in regions with strong lighting, however it produces blurry estimates in regions under shadow.

As can be seen in Fig.~\ref{fig:shading-qualitative-multiple-subjects}b (MP), standard variational message passing finds solutions that are highly inaccurate with continued presence of illumination and artifacts in areas of cast show. In contrast, inference using \MTD produces artefact-free results that much more closely resemble the stereo ground-truths. Arguably \MTD also improves over the baseline \citep{Biswas2009}, since its estimates are not blurry in regions with cast shadows. This can be attributed to the presence of symmetry priors in the model. Additionally, we note that the variance of the \MTD inference for reflectance (Fig.~\ref{fig:shading-qualitative-multiple-subjects}c) correlates strongly with cast shadows in the observed images (\ie the model is uncertain where it should be) suggesting that in future work it would be fruitful to have the notion of cast shadows explicitly built into the model. Figs.~\ref{fig:shading-qualitative-multiple-subjects}d and \ref{fig:shading-qualitative-multiple-subjects}e show analogous results for lighting and normal maps, and Fig.~\ref{fig:shading-qualitative-same-subject} demonstrates \MTD's ability to robustly infer reflectance maps for images of a single subject taken under varying lighting conditions. More visual results are shown in the supplementary at the end of this thesis
(Figs.~\ref{fig:shading-qualitative-multiple-subjects-supp},~\ref{fig:shading-qualitative-same-subject}).

We use the task of subject recognition (using estimated reflectance) as a quantitative measure of inference accuracy, as it can be difficult to measure in more direct ways (\eg RMSE strongly favors blurry predictions). The reflectance estimate produced by each algorithm is compared to all training subjects' ground-truth reflectances and is assigned the label of its closest match. We have found this evaluation to reflect the quality of inference estimates. Fig.~\ref{fig:shading-quantitative-reflectance} shows the result of this experiment, both for real images and also synthetic images that were produced by taking the stereo ground-truths and adding artificial lighting (but with no cast shadows). We show analogous results for light in Fig.~\ref{fig:shading-quantitative-light}, where error is defined to be the cosine angle distance between the estimated light and the photometric stereo reference. First, we note that standard variational message passing (MP) performs poorly, producing reflectance estimates that are much less useful for recognition than those from \cite{Biswas2009}. Second, we note that \MTD in the same model (both 1 stage and 2 stage versions) produces inferences that are significantly more useful downstream. The horizontal line labelled `Forest' represents the accuracy of the consensus messages without any message passing, showing that the model-based fine-tuning provides a significant benefit. Finally, we highlight the fact that initializing light directly from the image and running message passing (Fig.~\ref{fig:shading-quantitative-reflectance}, Init+MP) leads to worse estimates than \MTD demonstrating the use of layered predictions as opposed to direct predictions from the observations. These results demonstrate that \MTD helps message passing find better fixed points even in the presence of model mis-match (shadows) and make use of the full potential of the generative model.

\section{Discussion and Conclusions}
\label{sec:discussion-chap4}

We have presented \METHOD and shown that it is a computationally efficient technique that can be used to improve the accuracy of message passing inference in a variety of vision models. The crux of the approach is to recognize the importance of global variables, and to take advantage of layered model structures commonly seen in vision to make rough estimates of their values.

The success of \MTD depends on the accuracy of the random forest predictors. The design of forest features is not yet completely automated, but we took care in this work to use generic features that can be applied to a broad class of problems. Our forests are implemented in an extensible manner, and we envisage building a library of them that one can choose from, simply by inspecting the data types of the contextual and target variables.

In future work, we would like to exploit the benefits of the \MTD framework by applying it to more challenging problems from computer vision. Each of the examples in Section~\ref{sec:experiments-chap4} can be extended in various ways, \eg by making considerations for multiple objects, incorporating occlusion in the squares example and cast shadows in the faces example, or by developing more realistic priors. We are also seeking to understand in what other domains the application of our ideas may be fruitful.

More broadly, a major challenge in machine learning is that of enriching models in a scalable way. We continually seek to ask our models to provide interpretations of increasingly complicated, heterogeneous data sources. Graphical models provide an appealing framework to manage this complexity, but the difficulty of inference has long been a barrier to achieving these goals. The \MTD framework takes us one step in the direction of overcoming this barrier.

\part{Inference in Discriminative Vision Models}
\chapter{Learning Sparse High Dimensional Filters}
\label{chap:bnn}

\newcommand{\bv}{\mathbf{x}}
\newcommand{\bw}{\mathbf{w}}

In the Part-II of this thesis, we focus on inference in discriminative CNN models. Since \emph{inference}
amounts to simple evaluation of the \emph{model} in discriminative models,
we propose modifications to the original model itself for better inference.
This is unlike the inference strategies proposed in Part-I of this
thesis, where we proposed to learn a separate inference model that helps in the Bayesian inference
of a given generative model. In this chapter, we propose a learning technique for general sparse
high-dimensional filters and show how this can be used for generalizing standard spatial
convolutions in CNNs to learnable bilateral convolutions with long-range image-adaptive connections.

Bilateral filters have wide spread use due to their edge-preserving properties. The common
use case is to manually choose a parametric filter type, usually a Gaussian filter. In this chapter,
we will generalize the parametrization and in particular derive a gradient descent algorithm
so the filter parameters can be learned from data. This derivation allows to learn
high dimensional linear filters that operate in sparsely populated feature spaces. We build on
the permutohedral lattice construction for efficient filtering.
The ability to learn more general forms of high-dimensional filters can be used in
several diverse applications.
First, we demonstrate the use in applications where single filter applications are desired for runtime reasons.
Further, we show how this algorithm can be used to learn the pairwise potentials in
densely connected conditional random fields and apply these to different image segmentation tasks.
Finally, we introduce layers of bilateral filters in CNNs and propose
\emph{bilateral neural networks} for the use of high-dimensional
sparse data. This view provides new ways to encode model structure into network architectures. A diverse
set of experiments empirically validates the usage of general forms of filters.

\section{Introduction}
\label{sec:intro-bnn}

Image convolutions are basic operations for many image processing and computer vision applications. In this
chapter, we will study the class of bilateral filter convolutions and
propose a general image adaptive convolution that can be learned from data.

The bilateral filter~\cite{aurich1995non, smith97ijcv, tomasi1998bilateral} was originally introduced for the task of image denoising as an edge preserving filter. Since the bilateral filter contains the spatial
convolution as a special case, in the following, we will directly state the general case. Given an image $\bv=(\bv_1,\ldots,\bv_n), \bv_i\in\mathbb{R}^c$
with $n$ pixels and $c$ channels,
and for every pixel $i$, a $d$ dimensional feature vector $\f_i\in\mathbb{R}^d$ (\eg\ the $(x,y)$ position in the image $\f_i=(x_i,y_i)^{\top}$).
The bilateral filter then computes
\begin{equation}
  \bv'_i = \sum_{j=1}^n \bw_{\f_i,\f_j} \bv_j.\label{eq:bilateral}
\end{equation}
for all $i$. Almost the entire literature refers to the bilateral filter as a synonym of the Gaussian parametric form
$\bw_{\f_i,\f_j} = \exp{(-\frac{1}{2}(\f_i-\f_j)^{\top}\Sigma^{-1}(\f_i-\f_j))}$. The features $\f_i$ are most commonly chosen to be
position $(x_i,y_i)$ and color $(r_i,g_i,b_i)$ or pixel intensity. To appreciate the edge-preserving effect of
the bilateral filter, consider the five-dimensional feature $\f={(x,y,r,g,b)}^{\top}$.
Two pixels $i,j$ have a strong influence $\bw_{\f_i,\f_j}$ on each other only if they are close in
position \emph{and} color. At edges, the color changes, therefore pixels lying on opposite sides have low
influence and thus this filter does not blur across edges. This behavior is
sometimes referred to as \emph{image adaptive}, since the filter has a different shape when evaluated at
different locations in the image. More precisely, it is the projection of the filter to the two-dimensional
image plane that changes, the filter values $\bw_{\f,\f'}$ do not change. The filter itself can
be of $c$ dimensions $\bw_{\f_i,\f_j}\in\mathbb{R}^c$,
in which case the multiplication in Eq.~\ref{eq:bilateral} becomes an inner product. For the Gaussian case,
the filter can be applied independently per channel. For an excellent review of image filtering, we refer
to~\cite{milanfar2011tour}.

The filter operation of Eq.~\ref{eq:bilateral} is a sparse high-dimensional convolution,
a view advocated in~\cite{barash2002fundamental,paris2006fast}. An image $\bv$ is
not sparse in the spatial domain, we observe pixels values for all locations $(x,y)$. However, when
pixels are understood in a higher dimensional feature space, \eg $(x,y,r,g,b)$, the image becomes a sparse signal,
since the $r,g,b$ values lie scattered in this five-dimensional space.
This view on filtering is the key difference of the bilateral filter compared to the common spatial convolution.
An image edge is not \emph{visible} for a filter in the spatial domain alone, whereas in the 5D space it is.
The edge-preserving behavior is possible due to the higher dimensional operation. Other data can naturally
be understood as sparse signals, \eg 3D surface points.

The main contribution of this work is to propose a general and learnable sparse high dimensional convolution.
Our technique builds on efficient algorithms
that have been developed to approximate the Gaussian bilateral filter and re-uses them for
more general high-dimensional filter operations. Due to its practical importance (see related work in
Section~\ref{sec:related-bnn}), several efficient algorithms for computing Eq.~\ref{eq:bilateral}
have been developed, including the bilateral grid~\cite{paris2006fast}, Gaussian
KD-trees~\cite{adams2009gaussian}, and the permutohedral
lattice~\cite{adams2010fast}. The design goal for these algorithms was to provide a) fast runtimes and b) small
approximation errors for
the Gaussian filter case. The key insight of this work is to use the permutohedral lattice and use it not as
an approximation of a predefined kernel but to freely parametrize its values. We relax the separable Gaussian
filter case from~\cite{adams2010fast} and show
how to compute gradients of the convolution (Section~\ref{sec:learning}) in lattice space. This enables learning
the filter from data.

This insight has several useful consequences. We discuss applications where the bilateral filter has been used
before: image filtering (Section~\ref{sec:filtering}) and CRF inference (Section~\ref{sec:densecrf}). Further we will demonstrate how the
free parametrization of the filters enables us to use them in deep convolutional neural networks (CNN) and allow convolutions that go beyond the regular
spatially connected receptive fields (Section~\ref{sec:bnn}). For all domains, we present various empirical evaluations with
a wide range of applications.

\section{Related Work}
\label{sec:related-bnn}

We categorize the related work according to the three different generalizations of this work.

\paragraph{Image Adaptive Filtering:} The literature in this area is rich
and we can only provide a brief overview. Important classes of image adaptive filters
include the bilateral filters~\cite{aurich1995non, tomasi1998bilateral,
smith97ijcv}, non-local means~\cite{buades2005non, awate2005higher}, locally
adaptive regressive kernels~\cite{takeda2007kernel}, guided image
filters~\cite{he2013guided} and propagation filters~\cite{chang2015propagated}.
The kernel least-squares regression problem can serve as a unified view of many
of them~\cite{milanfar2011tour}. In contrast to the present work that learns
the filter kernel using supervised learning, all these filtering schemes use a
predefined kernel.
Because of the importance of bilateral filtering to many applications in
image processing, much effort has been devoted to derive fast algorithms; most
notably~\cite{paris2006fast, adams2010fast, adams2009gaussian, gastal2012adaptive}.
Surprisingly, the
only attempt to learn the bilateral filter we found is~\cite{hu2007trained} that
casts the learning problem in the spatial domain by rearranging pixels. However,
the learned filter does not necessarily obey the full region of influence of a
pixel as in the case of a bilateral filter.
The bilateral filter also has been proposed to regularize a large set of
applications in~\cite{barron2015bilateral, barron2015defocus} and the respective
optimization problems are parametrized in a bilateral space.
In these works the filters are part of a learning system but unlike this work
restricted to be Gaussian.

\paragraph{Dense CRF:} The key observation of~\cite{krahenbuhl2012efficient} is
that mean-field inference update steps in densely connected CRFs with Gaussian
edge potentials require Gaussian bilateral filtering operations. This enables
tractable inference through the application of a fast filter implementation
from~\cite{adams2010fast}. This quickly found wide-spread use, \eg the combination of CNNs with
a dense CRF is among the best performing segmentation
models~\cite{chen2014semantic, zheng2015conditional, bell2015minc}. These works
combine structured prediction frameworks on top of CNNs, to model the
relationship between the desired output variables thereby significantly
improving upon the CNN result. Bilateral neural networks, that are presented in
this work, provide a principled framework for encoding the output relationship,
using the feature transformation inside the network itself thereby alleviating
some of the need for later processing. Several works~\cite{krahenbuhl2013parameter,
domke2013learning,kiefel2014human,zheng2015conditional,schwing2015fully} demonstrate how to learn free parameters of
the dense CRF model. However, the parametric form of the pairwise term always remains a
Gaussian. Campbell \etal~\cite{campbell2013fully} embed complex pixel
dependencies into an Euclidean space and use a Gaussian filter for pairwise connections.
This embedding is a pre-processing step and can not directly be learned. In
Section~\ref{sec:densecrf} we will discuss how to learn the pairwise
potentials, while retaining the efficient inference strategy of~\cite{krahenbuhl2012efficient}.

\paragraph{Neural Networks:} In recent years, the use of CNNs enabled tremendous progress in a wide
range of computer vision applications. Most CNN architectures use
spatial convolution layers, which have fixed local receptive fields. This work
suggests to replace these layers with bilateral filters, which have a varying
spatial receptive field depending on the image content. The equivalent
representation of the filter in a higher dimensional space leads to sparse
samples that are handled by a permutohedral lattice data structure. Similarly,
Bruna \etal~\cite{bruna2013spectral} propose convolutions on irregularly sampled
data. Their graph construction is closely related to the high-dimensional
convolution that we propose and defines weights on local neighborhoods of nodes.
However, the structure of the graph is bound to be fixed and it is not
straightforward to add new samples. Furthermore, re-using the same filter
among neighborhoods is only possible with their costly spectral construction.
Both cases are handled naturally by our sparse convolution.
Jaderberg \etal~\cite{jaderberg2015spatial} propose a spatial
transformation of signals within the neural network to learn invariances for a
given task. The work of~\cite{ionescu2015matrix} propose matrix back-propagation
techniques which can be used to build specialized structural layers such as normalized-cuts.
Graham \etal~\cite{graham2015sparse} propose extensions from 2D CNNs
to 3D sparse signals. Our work enables sparse 3D filtering as a special case,
since we use an algorithm that allows for even higher dimensional data.

\section{Learning Sparse High Dimensional Filters}\label{sec:learning}
In this section, we describe the main technical contribution of
this work, we generalize the permutohedral convolution~\cite{adams2010fast} and show how
the filter can be learned from data.

Recall the form of the bilateral convolution from Eq.~\ref{eq:bilateral}. A naive
implementation would compute for every pixel $i$ all associated filter values $\bw_{\f_i,\f_j}$ and
perform the summation independently. The view of $\bw$ as a linear filter in a higher dimensional
space, as proposed by~\cite{paris2006fast}, opened the way for new algorithms. Here,
we will build on the permutohedral lattice convolution developed
in Adams \etal~\cite{adams2010fast} for approximate Gaussian filtering.
The most common application of bilateral filters use photometric features (XYRGB).
We chose the permutohedral lattice as it is particularly designed for this dimensionality,
see Fig.~7 in~\cite{adams2010fast} for a speed comparison.

\subsection{Permutohedral Lattice Convolutions}

We first review the permutohedral lattice convolution for
Gaussian bilateral filters from Adams \etal~\cite{adams2010fast} and
describe its most general case.

\begin{figure}[t]
\begin{center}
\centerline{\includegraphics[width=\columnwidth]{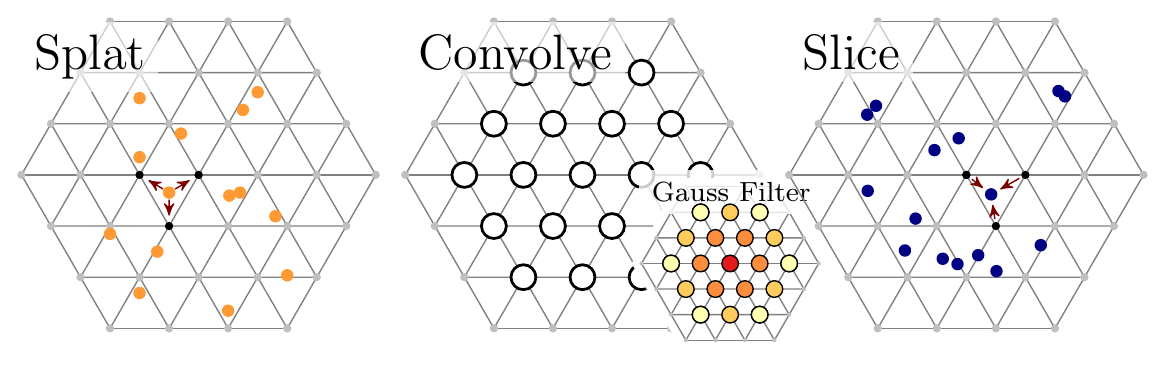}}
  \mycaption{Schematic of the permutohedral convolution}
  {Left: splatting the input
  points (orange) onto the lattice corners (black); Middle: The extent of a filter on the lattice
   with a $s=2$ neighborhood (white circles), for reference we show a Gaussian filter,
   with its values color coded. The general case has a free scalar/vector parameter per circle.
    Right: The result of the convolution at the lattice corners (black) is projected back
    to the output points (blue). Note that in general the output and input points may be different. \label{fig:ops}}
\end{center}
\end{figure}

\begin{figure}[t]
  \centering
  \subfigure[\scriptsize Sample Image]{%
    \includegraphics[width=.23\columnwidth]{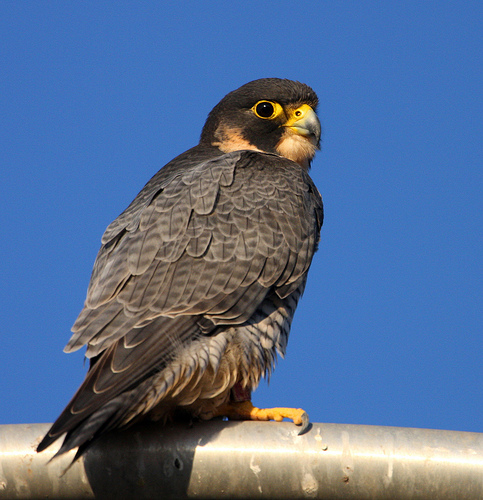}\label{fig:bird}
  }
  \subfigure[\scriptsize Position]{%
    \includegraphics[width=.23\columnwidth]{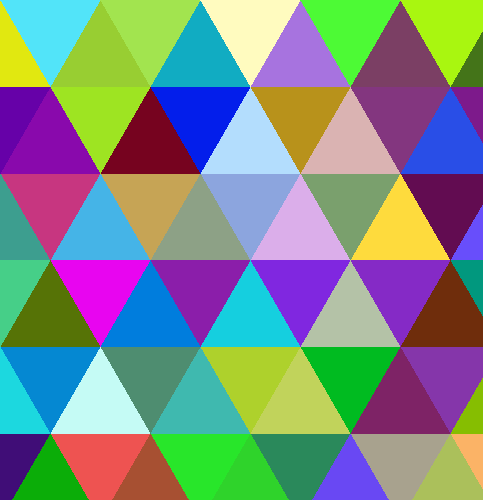}\label{fig:bird_lattice_1}
  }
  \subfigure[\scriptsize Color]{%
    \includegraphics[width=.23\columnwidth]{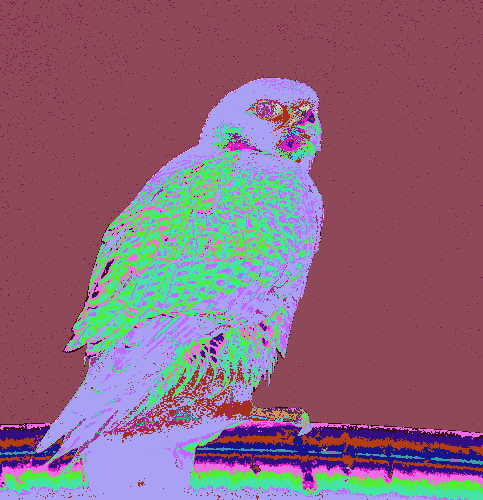}\label{fig:bird_lattice_2}
  }
  \subfigure[\scriptsize Position, Color]{%
    \includegraphics[width=.23\columnwidth]{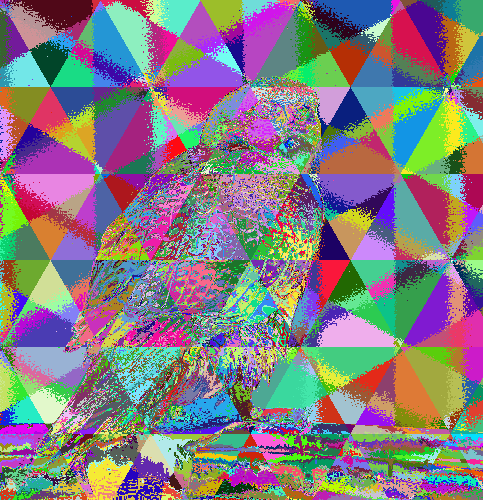}\label{fig:bird_lattice_3}
  }
  \mycaption{Visualization of the permutohedral lattice}
  {(a) Input image; Lattice visualizations for different feature spaces: (b) 2D position features: $0.01(x, y)$,
  (c) color features: $0.01(r, g, b)$ and (d) position and color features: $0.01(x, y, r, g, b)$. All pixels
  falling in the same simplex cell are shown with the same color.}
\label{fig:lattice}
\end{figure}

As before, we assume that every image pixel $i$ is associated with a $d$-dimensional
feature vector $\f_i$. Gaussian bilateral filtering using a permutohedral lattice
approximation involves 3 steps. We begin with an overview of the algorithm, then discuss each step
in more detail in the next paragraphs. Figure~\ref{fig:ops} schematically shows
the three operations for 2D features. First, interpolate the image signal on the
$d$-dimensional grid plane of the permutohedral lattice, which is called
\emph{splatting}. A permutohedral lattice is the tessellation of space into permutohedral simplices.
We refer to~\cite{adams2010fast} for details of the lattice construction and its properties.
In Fig.~\ref{fig:lattice}, we visualize
the permutohedral lattice in the image plane, where every simplex cell receives a different color. All pixels of the same
lattice cell have the same color. Second, \emph{convolve} the signal on the lattice.
And third, retrieve the result by interpolating the signal at the original
$d$-dimensional feature locations, called \emph{slicing}. For example, if
the features used are a combination of position and color $\f_i = (x_i, y_i, r_i, g_i,
b_i)^{\top}$, the input signal is mapped into the 5D cross product space of position and
color and then convolved with a 5D tensor. Afterwards, the filtered result is mapped back to the
original space. In practice we use a feature scaling $\Lambda \f$ with a diagonal matrix~$\Lambda$
and use separate scales
for position and color features. The scale determines the distance of points and thus the size
of the lattice cells.
More formally, the computation is written by $\bv' =
S_{\text{slice}}BS_{\text{splat}} \bv$ and all involved matrices are
defined below. For notational convenience we will assume scalar input signals $x_i$, the vector valued case
is analogous, the lattice convolution changes from scalar multiplications to inner products.

\paragraph{Splat:} The splat operation (cf.\ left-most image in
Fig.~\ref{fig:ops}) finds the enclosing simplex in $\mathcal{O}(d^2)$ on the
lattice of a given pixel feature $\f_i$ and distributes its value $v_i$ onto the corners of the
simplex. How strong a pixel contributes to a corner $j$ is defined by its
barycentric coordinate $t_{i,j}\in\mathbb{R}$ inside the simplex.
Thus, the value $\mathbf{\ell}_j \in \mathbb{R}$ at a lattice point
$j$ is computed by summing over all enclosed input points; more precisely, we
define an index set $J_i$ for a pixel $i$, which contains all the lattice points
$j$ of the enclosing simplex
\vspace{-0.7em}
\begin{equation}
  \label{eq:splat}
  \mathbf{\ell} = S_{\text{splat}} \bv;%
  {(S_{\text{splat}})}_{j, i} = t_{i, j}, \text{ if } j\in J_i, \text{ otherwise } 0.
\vspace{-0.5em}
\end{equation}

\vspace{-0.3cm}
\paragraph{Convolve:} The permutohedral convolution is defined on the
lattice neighborhood $N_s(j)$ of lattice point $j$, \eg\ only $s$ grid hops away. More formally
\vspace{-0.7em}
\begin{equation}
  \label{eq:blur}
  \mathbf{\ell'} = B \mathbf{\ell}; {(B)}_{j', j} =  w_{j, j'},%
  \text{ if } j' \in N_s(j), \text{ otherwise } 0.
\vspace{-0.5em}
\end{equation}
An illustration of a two-dimensional permutohedral filter is shown in Fig.~\ref{fig:ops} (middle).
Note that we already presented the convolution in the general form that we will make
use of. The work of~\cite{adams2010fast} chooses the filter weights such that
the resulting operation approximates a Gaussian blur, which is illustrated in Fig.~\ref{fig:ops}.
Further, the algorithm of~\cite{adams2010fast} takes advantage of the
separability of the Gaussian kernel. Since we are interested in the most general case, we
extended the convolution to include non-separable filters $B$.

\vspace{-0.3cm}
\paragraph{Slice:} The slice operation (cf.\ right-most image in
Fig.~\ref{fig:ops}) computes an output value $x'_{i'}$ for an output pixel $i'$
again based on its barycentric coordinates $t_{i, j}$ and sums over the corner
points $j$ of its lattice simplex
\vspace{-0.7em}
\begin{equation}
  \label{eq:slice}
  \bv' = S_{\text{slice}} \mathbf{\ell'}; %
  {(S_{\text{slice}})}_{i, j} = t_{i, j}, \text{ if } j\in J_i, \text{ otherwise } 0
\vspace{-0.5em}
\end{equation}

The splat and slice operations take a role of an interpolation between the
different signal representations: the irregular and sparse distribution of
pixels with their associated feature vectors and the regular structure of the
permutohedral lattice points. Since high-dimensional spaces are usually sparse,
performing the convolution densely on all lattice points is inefficient. So, for
speed reasons, we keep track of the populated lattice points using a hash table
and only convolve at those locations.

\subsection{Learning Permutohedral Filters}\label{sec:backprop}
The \emph{fixed} set of filter weights $\mathbf{w}$ from~\cite{adams2010fast} in
Eq.~\ref{eq:blur} is designed to approximate a Gaussian filter. However, the
convolution kernel $\mathbf{w}$ can naturally be understood as a general filtering
operation in the permutohedral lattice space with free parameters. In the exposition above we
already presented this general case. As we will show in more detail later,
this modification has non-trivial consequences
for bilateral filters, CNNs and probabilistic graphical models.

The size of the neighborhood $N_s(k)$ for the blur in Eq.~\ref{eq:blur} compares
to the filter size of a spatial convolution. The filtering kernel of a common
spatial convolution that considers $s$ points to either side in all dimensions
has ${(2s + 1)}^d\in\mathcal{O}(s^d)$ parameters. A comparable filter on the
permutohedral lattice with an $s$ neighborhood is specified by ${(s+1)}^{d+1} -
s^{d+1} \in\mathcal{O}(s^d)$ elements (cf. Appendix~\ref{sec:permconv}).
Thus, both share the same asymptotic size.

By computing the gradients of the filter elements, we enable the use
of gradient based optimizers, \eg back-propagation for CNN in the same
way that spatial filters in a CNN are learned. The gradients
with respect to $\bv$ and the filter weights in $B$ of a
scalar loss $L$ are:

\begin{eqnarray}
  \frac {\partial L} {\partial \bv} &=&
  S'_{\text{splat}} %
  B' %
  S'_{\text{slice}} %
  \frac {\partial L} {\partial \bv'}, \label{eq:dv}\\
  \frac {\partial L} {\partial {(B)}_{i, j}} &=&
  {\left(S'_{\text{slice}} \frac {\partial L} {\partial \bv}\right)}_i
  {(S_{\text{splat}} \bv)}_j.\label{eq:db}
\end{eqnarray}

Both gradients are needed during back-propagation and in experiments,
we use stochastic back-propagation for learning the filter kernel.
The permutohedral lattice convolution is parallelizable, and scales linearly with the
filter size. Specialized implementations run at interactive speeds in image
processing applications~\cite{adams2010fast}.
Our implementation in the Caffe deep learning framework~\cite{jia2014caffe}
allows arbitrary filter parameters and the computation
of the gradients on both CPU and GPU.
The code is available at http://bilateralnn.is.tuebingen.mpg.de.

\section{Single Bilateral Filter Applications}\label{sec:filtering}

In this section, we will consider the problems of joint bilateral up-sampling~\cite{kopf2007joint}
and 3D body mesh denoising as prominent instances of single bilateral filter applications.
See~\cite{paris2009bilateral} for a recent overview of other bilateral filter applications.
Further experiments on image denoising are included in Appendix~\ref{sec:appendix-bnn},
together with details about exact experimental protocols and more visualizations.

\subsection{Joint Bilateral Upsampling}

A typical technique to speed up computer vision algorithms is to compute results on a
lower scale and up-sample the result to the full resolution. This up-sampling
step may use the original resolution image as a guidance image. A joint bilateral up-sampling approach
for this problem setting was developed in~\cite{kopf2007joint}. We describe the procedure
for the example of up-sampling a color image. Given a high resolution gray scale image (the guidance image) and
the same image on a lower resolution but with colors, the task is to up-sample the color image to the
same resolution as the guidance image. Using the permutohedral lattice, joint bilateral up-sampling proceeds by
splatting the color image into the lattice, using 2D position and 1D intensity as features and the 3D RGB values
as the signal. A convolution is applied in the lattice and the result is read out at the features of the high
resolution image, that is using the 2D position and intensity of the guidance image. The possibility of
reading out (slicing) points that are not necessarily the input points is an appealing feature of the permutohedral lattice
convolution.

\begin{figure*}[t!]
\centering
\subfigure{%
 \raisebox{2.0em}{
  \includegraphics[width=.08\columnwidth]{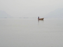}\label{fig:color_given}
 }
}
\subfigure{%
  \includegraphics[width=.16\columnwidth]{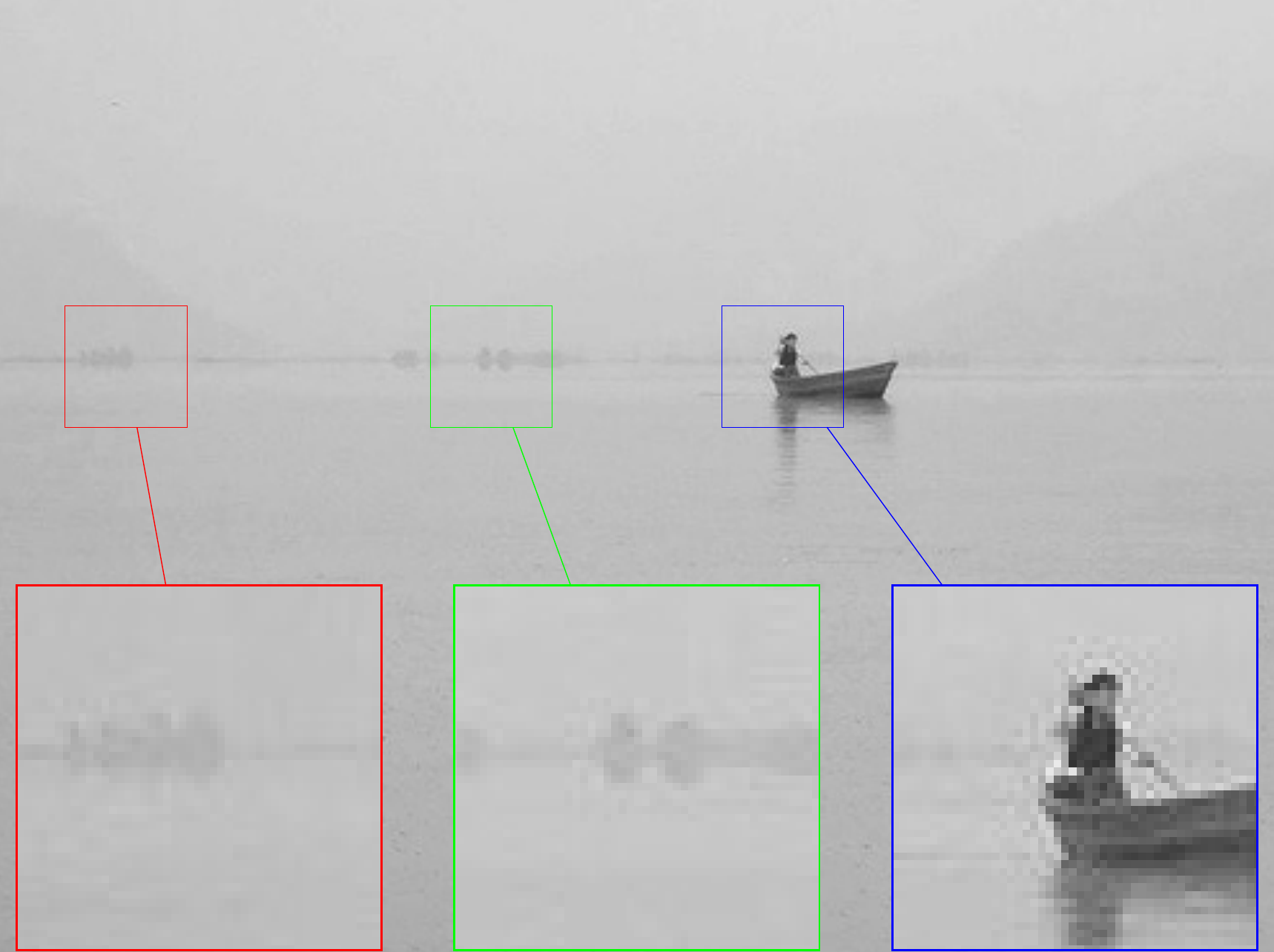}\label{fig:color_guidance}
}
\subfigure{%
  \includegraphics[width=.16\columnwidth]{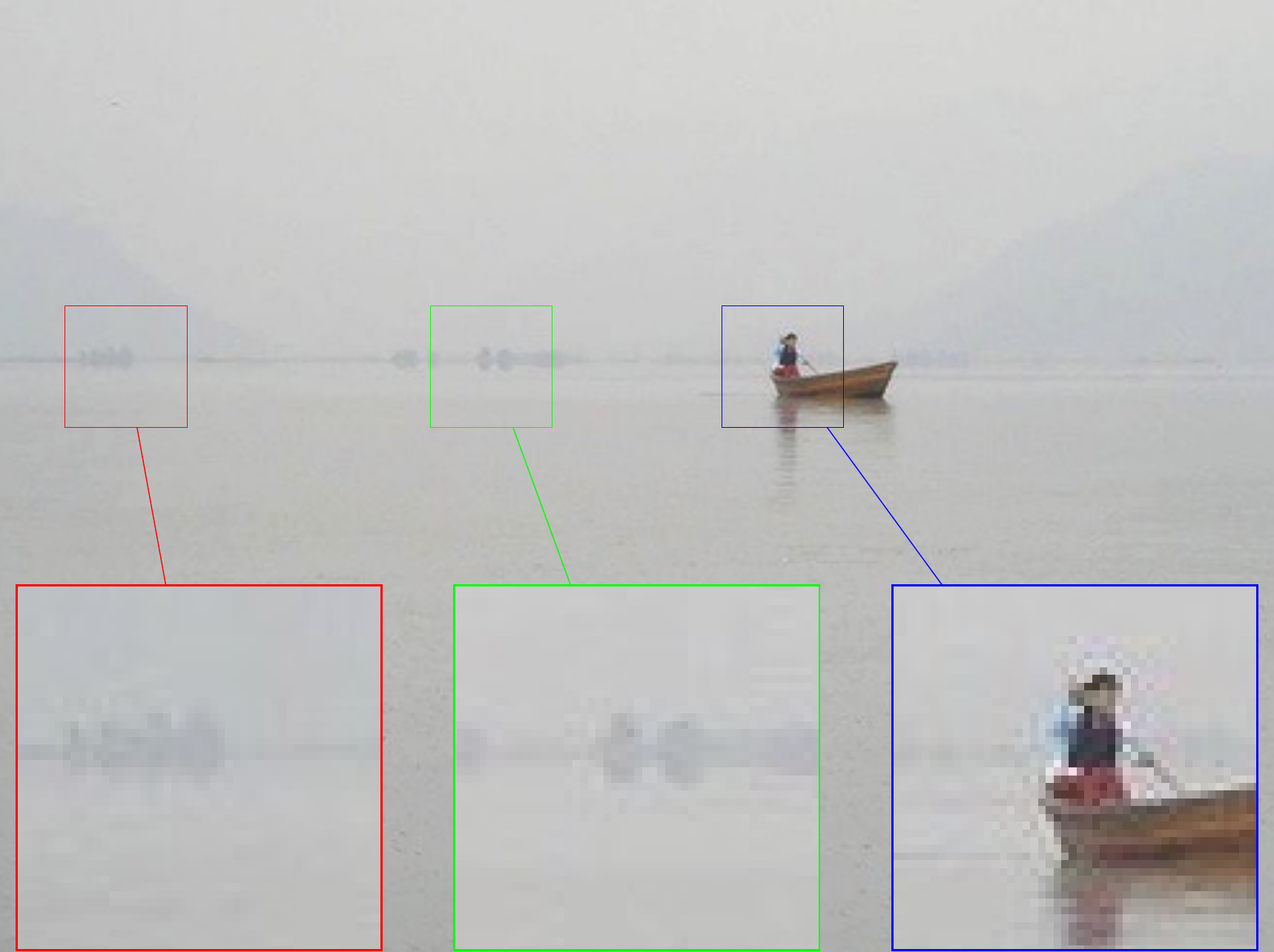}\label{fig:color_original}
}
\subfigure{%
  \includegraphics[width=.16\columnwidth]{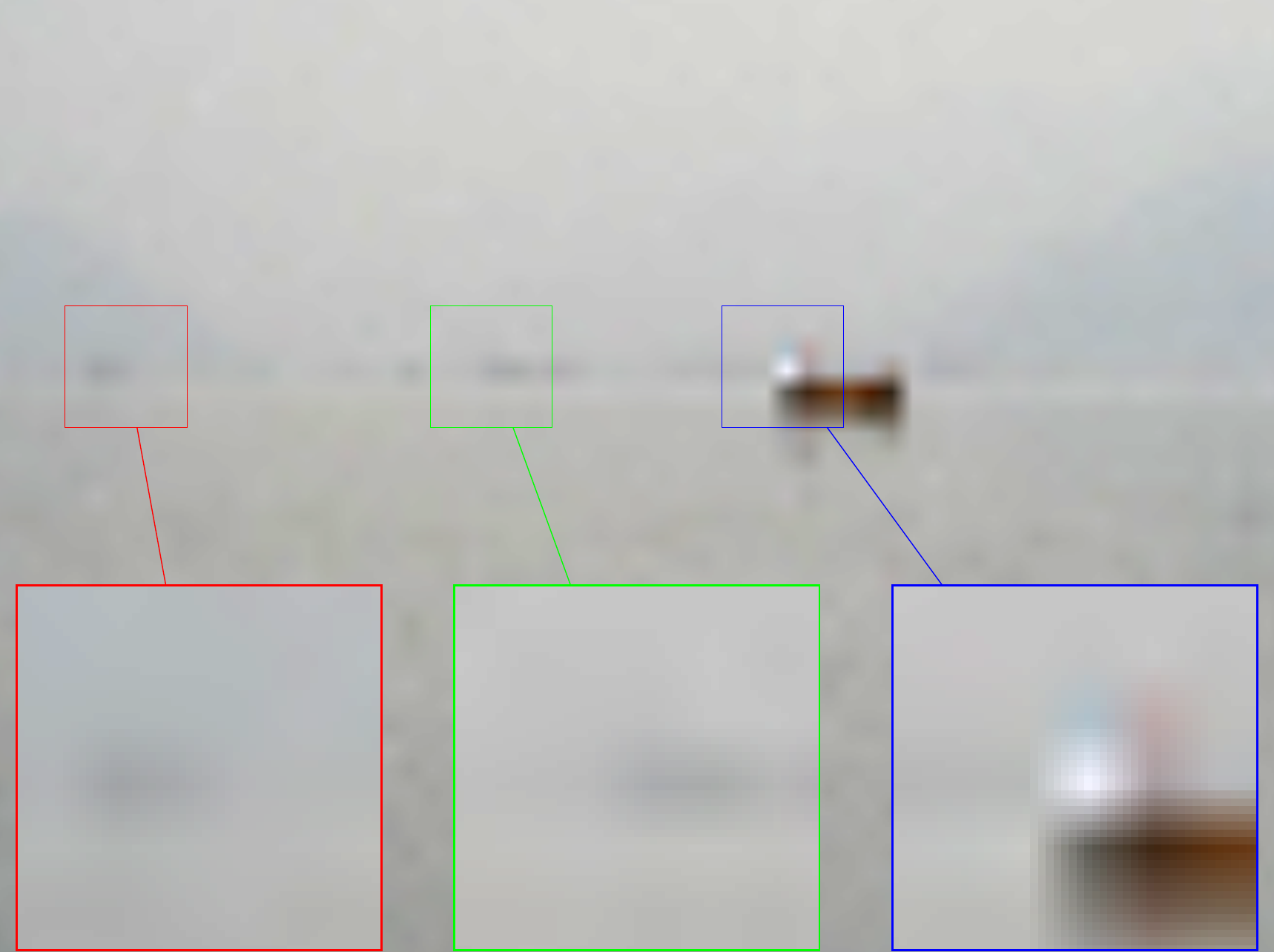}\label{fig:color_bicubic}
}
\subfigure{%
  \includegraphics[width=.16\columnwidth]{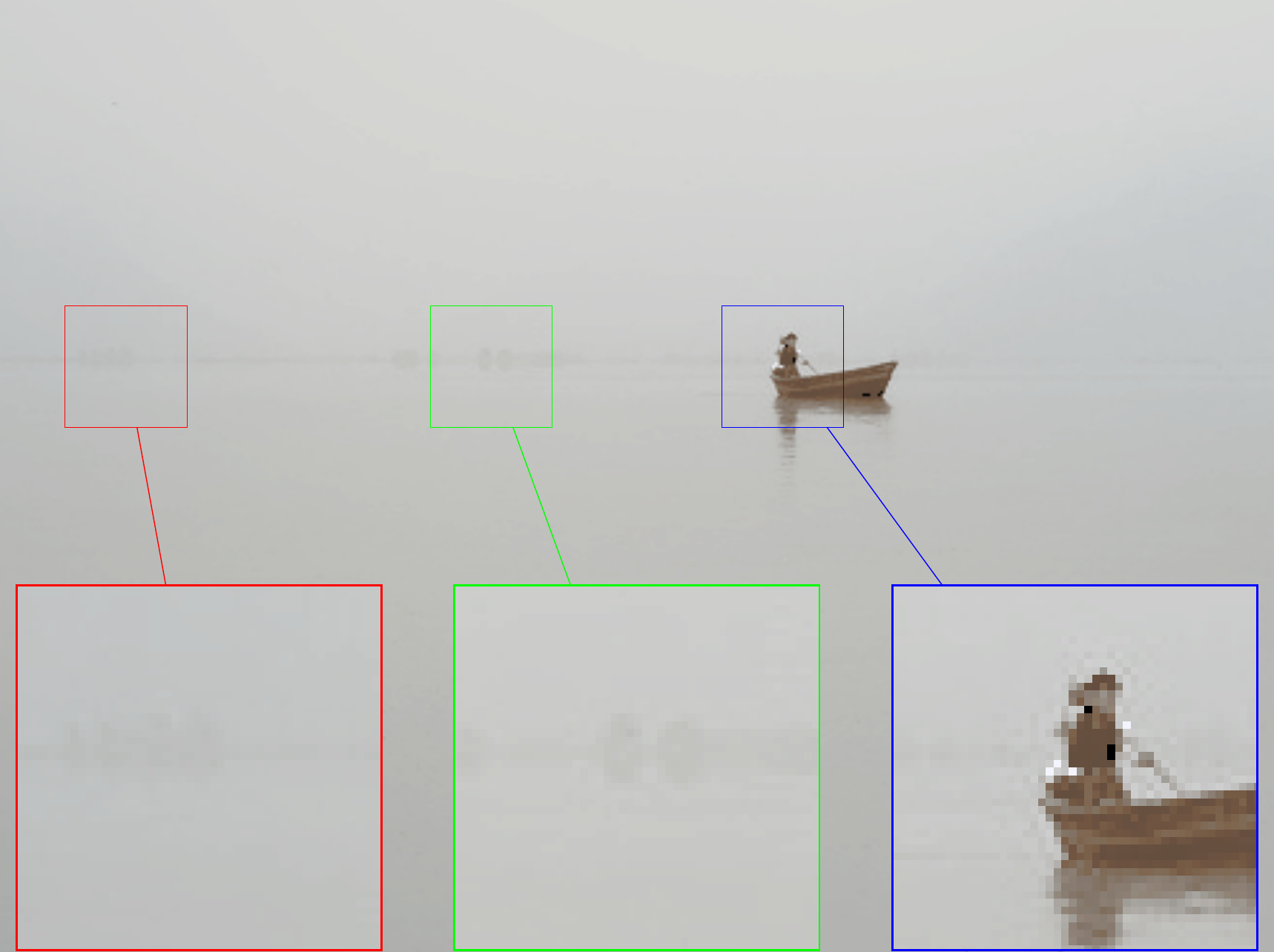}\label{fig:color_gauss}
}
\subfigure{%
  \includegraphics[width=.16\columnwidth]{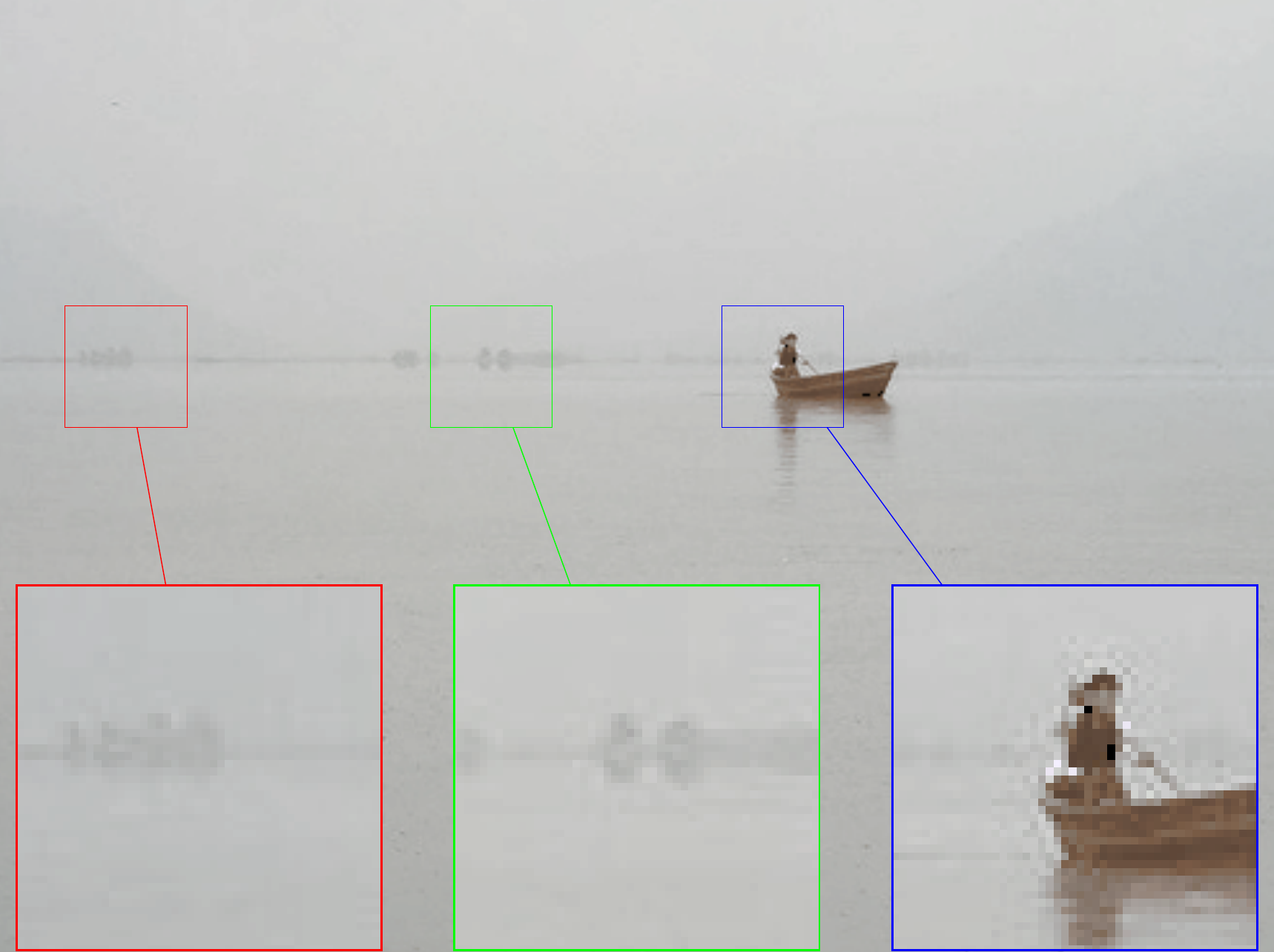}\label{fig:color_learnt}
}\\
\setcounter{subfigure}{0}
\subfigure[Input]{%
\raisebox{2.0em}{
  \includegraphics[width=.08\columnwidth]{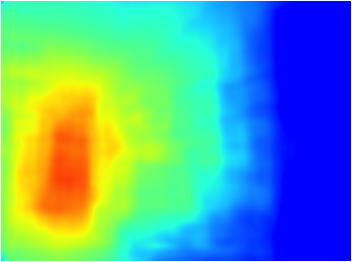}\label{fig:depth_given}
 }
}
\subfigure[Guidance]{%
  \includegraphics[width=.16\columnwidth]{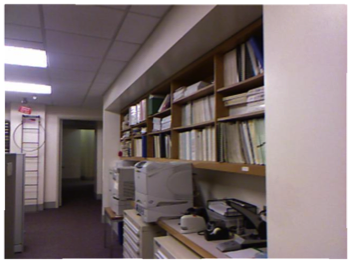}\label{fig:depth_guidance}
}
 \subfigure[Ground Truth]{%
  \includegraphics[width=.16\columnwidth]{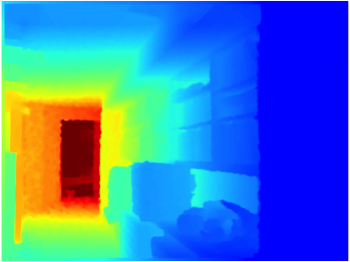}\label{fig:depth_gt}
}
\subfigure[Bicubic]{%
  \includegraphics[width=.16\columnwidth]{figures/depth_bicubic.png}\label{fig:depth_bicubic}
}
\subfigure[Gauss-BF]{%
  \includegraphics[width=.16\columnwidth]{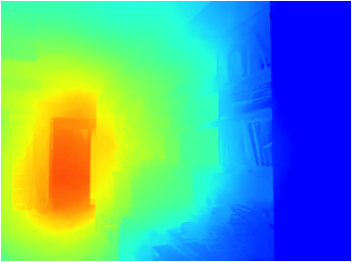}\label{fig:depth_gauss}
}
\subfigure[Learned-BF]{%
  \includegraphics[width=.16\columnwidth]{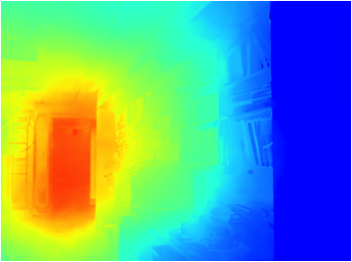}\label{fig:depth_learnt}
}
\mycaption{Guided up-sampling}{Color (top) and depth (bottom) $8\times$ up-sampling results
using different methods: Bicubic - Bicubic interpolation; Gauss-BF - Gaussian bilateral
upsampling; Learned-BF - Learned bialteral up-sampling (best viewed on screen).}
\label{fig:upsample_visuals}
\end{figure*}

\begin{table}[t]
  \scriptsize
  \centering
    \begin{tabular}{l c c c c c}
      \toprule
     & \textbf{Upsampling factor} & \textbf{Bicubic} & \textbf{Gaussian} & \textbf{Learned} \\ [0.1cm]
      \midrule
     \multicolumn{2}{l}\textbf{Color Upsampling (PSNR)} & & &\\
     & \textbf{2x} & 24.19 / 30.59 & 33.46 / 37.93  & \textbf{34.05 / 38.74}  \\
     & \textbf{4x} & 20.34 / 25.28 & 31.87 / 35.66  & \textbf{32.28 / 36.38}  \\
     & \textbf{8x} & 17.99 / 22.12 & 30.51 / 33.92  & \textbf{30.81 / 34.41}  \\
     & \textbf{16x} & 16.10 / 19.80 & 29.19 / 32.24  & \textbf{29.52 / 32.75}  \\
      \midrule
     \multicolumn{2}{l}\textbf{Depth Upsampling (RMSE)} & & &\\
     & \textbf{8x} & 0.753 & 0.753 & \textbf{0.748}  \\
      \bottomrule
      \\
    \end{tabular}
      \vspace{-0.2cm}
    \mycaption{Joint bilateral up-sampling} {(top) PSNR values corresponding to various up-sampling factors and
    up-sampling strategies on the test images of the Pascal VOC12 segmentation / high-resolution 2MP dataset;
    (bottom) RMSE error values corresponding to up-sampling depth images estimated using~\cite{eigen2014depth}
    computed on the test images from the NYU depth dataset~\cite{silberman2012indoor}.}
\label{tbl:upsample}
\end{table}

\vspace{-0.2cm}
\subsubsection{Color Up-sampling}

For the task of color up-sampling, we compare the Gaussian bilateral filter~\cite{kopf2007joint} against
a learned generalized filter. We experimented with two different datasets: Pascal VOC2012 segmentation~\cite{voc2012segmentation}
using train, validation and test splits, and 200 higher resolution (2MP) images from
Google image search~\cite{google_images} with 100 train, 50 validation and 50 test images.
For training, we use the mean
squared error (MSE) criterion and perform stochastic gradient descent with a
momentum term of $0.9$, and weight decay of $0.0005$, found using the validation set.
In Table~\ref{tbl:upsample} we report result in terms of Peak-Signal-to-Noise ratio (PSNR)
for the up-sampling factors $2\times, 4\times, 8\times$ and $16\times$.
We compare a standard bicubic interpolation, that does not use a guidance image, the Gaussian
bilateral filter case (with feature scales optimized on the validation set), and the learned filter.
All filters have the same support.
For all up-sampling factors, joint bilateral Gaussian up-sampling outperforms bicubic interpolation and is in turn improved using a learned filter. A result of the up-sampling is shown in Fig.~\ref{fig:upsample_visuals} and
more results are included in Section~\ref{sec:col_upsample_extra}.
The learned filter recovers finer details in the images.

We also performed the cross-factor analysis of training and testing at different up-sampling
factors. Table~\ref{tbl:crossupsample} shows the PSNR results for this
analysis. Although, in terms of PSNR, it is optimal to train and test at the
same up-sampling factor, the differences are small when training and testing
up-sampling factors are different.

\begin{table}[t]
  \scriptsize
  \centering
    \begin{tabular}{c c c c c c}
      \toprule
      & & \multicolumn{4}{c}{Test Factor} \\ [0.1cm]
     & & \textbf{2$\times$} & \textbf{4$\times$} & \textbf{8$\times$} & \textbf{16$\times$} \\ [0.15cm]
    \parbox[t]{3mm}{\multirow{4}{*}{\rotatebox[origin=c]{90}{Train Factor}}} & \textbf{2$\times$} & \textbf{38.45} & 36.12  & 34.06 & 32.43 \\ [0.1cm]
     & \textbf{4$\times$} & 38.40 & \textbf{36.16} & \textbf{34.08} & 32.47 \\ [0.1cm]
     & \textbf{8$\times$} & 38.40 & 36.15  & \textbf{34.08} & 32.47 \\ [0.1cm]
     & \textbf{16$\times$} & 38.26 & 36.13  & 34.06 & \textbf{32.49} \\
      \bottomrule
      \\
    \end{tabular}
      \vspace{-0.2cm}
    \mycaption{Color upsampling with different train and test up-sampling factors} {PSNR values corresponding to different up-sampling factors used at train and test times on the 2 megapixel image dataset, using our learned bilateral filters.}
\label{tbl:crossupsample}
\vspace{-0.4cm}
\end{table}

\subsubsection{Depth Up-sampling}
We experimented with depth up-sampling as another joint up-sampling task. We use the dataset
of~\cite{silberman2012indoor} that comes with pre-defined
train, validation and test splits. The approach of~\cite{eigen2014depth} is a CNN
model that produces a result at 1/4th of the input resolution due to down-sampling operations in
max-pooling layers. Furthermore, the authors down-sample the $640\times 480$ images to $320\times 240$
as a pre-processing step before CNN convolutions. The final depth result is bicubic interpolated to the
original resolution. It is this interpolation that we replace with a Gaussian and learned joint
bilateral up-sampling. The features are five-dimensional position
and color information from the high resolution input image. The filter is learned using the same protocol
as for color up-sampling minimizing MSE prediction error. The quantitative results
are shown in Table~\ref{tbl:upsample}, the Gaussian filter performs equal to the bicubic interpolation,
the learned filter is better. Qualitative results are shown in Fig~\ref{fig:upsample_visuals},
both joint bilateral up-sampling respect image edges in the result. For this~\cite{ferstl2015b} and other tasks
specialized interpolation algorithms exist, \eg deconvolution networks~\cite{zeiler2010deconvolutional}. Part of future work is to equip these approaches with bilateral filters. More qualitative results are
presented in Appendix~\ref{sec:depth_upsample_extra}.

\begin{figure}[h!]
  \centering
    \includegraphics[width=0.7\columnwidth]{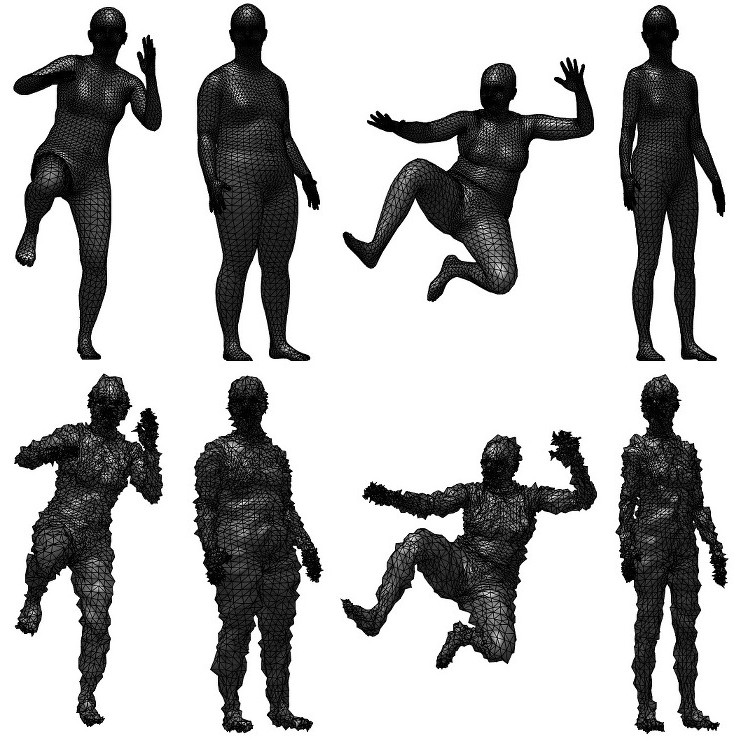}
  \mycaption{Sample data for 3D mesh denoising}
  {(top) Some 3D body meshes sampled from~\cite{SMPL:2015} and (bottom) the corresponding noisy meshes used in denoising experiments.}
\label{fig:samplebody}
\end{figure}

\subsection{3D Mesh Denoising}\label{sec:mesh_denoising}

Permutohedral convolutions can naturally be extended to higher ($>2$)
dimensional data. To highlight this, we use the proposed convolution for the task
of denoising 3D meshes.

\begin{figure}[t!]
  \centering
    \includegraphics[width=0.8\columnwidth]{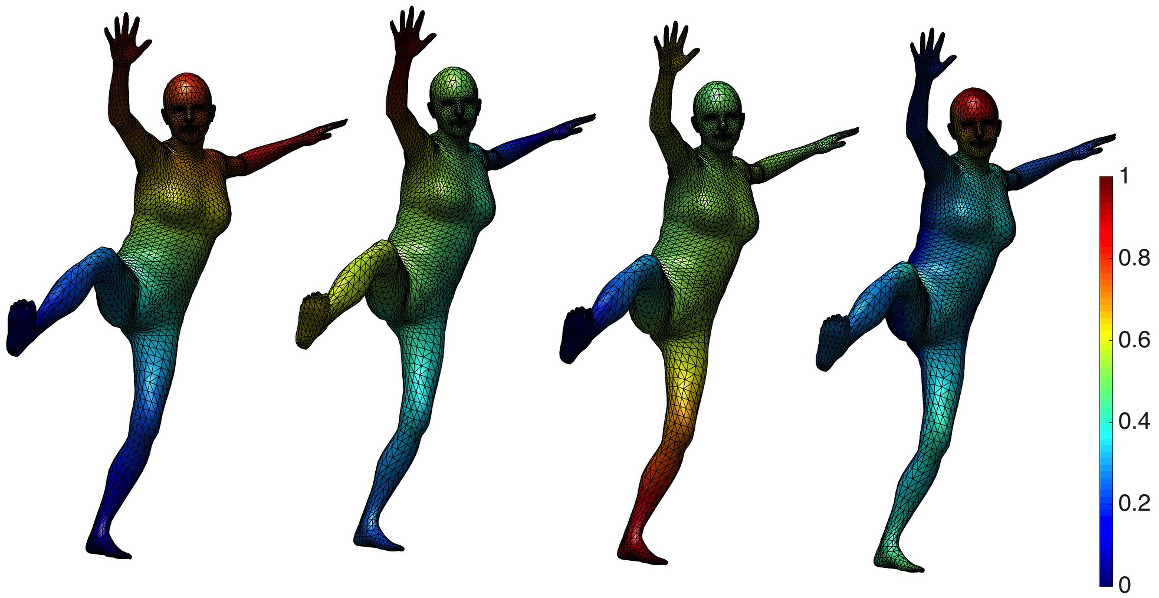}
  \mycaption{4D isomap features for 3D human bodies}
  {Visualization of 4D isomap features for a sample 3D mesh. Isomap feature values are overlaid onto mesh vertices.}
  \label{fig:isomap}
\end{figure}

We sample 3D human body meshes using a
generative 3D body model from~\cite{SMPL:2015}. To the clean meshes, we add Gaussian random
noise displacements along the surface normal at each vertex location.
Figure~\ref{fig:samplebody} shows some sample 3D meshes sampled
from~\cite{SMPL:2015} and corresponding noisy meshes. The task is to take the noisy meshes as inputs and
recover the original 3D body meshes. We create 1000 training, 200 validation and another 500
testing examples for the experiments. Although we use synthetically generated meshes and noise for
our experiments for the sake of training, our technique could be potentially used for
denoising the noisy meshes arising from 3D scanning devices.

\paragraph{Mesh Representation:} The 3D human body meshes from~\cite{SMPL:2015} are represented with
3D vertex locations and the edge connections between the vertices. We found that this signal representation
using global 3D coordinates is not suitable for denoising with bilateral filtering. Therefore, we first
smooth the noisy mesh using mean smoothing applied to the face normals~\cite{yagou2002mesh} and represent the noisy
mesh vertices as 3D vector displacements with respect to the corresponding smoothed mesh. Thus, the
task becomes denoising the 3D vector displacements with respect to the smoothed mesh.

\paragraph{Isomap Features:} To apply permutohedral convolution, we need to define features at each input
vertex point. We use a 4 dimensional isomap embedding~\cite{tenenbaum2000global} of the given 3D mesh graph as features. The
given 3D mesh is converted into a weighted edge graph with edge weights set to the Euclidean distance between the
connected vertices and to infinity between the non-connected vertices. Then the 4 dimensional isomap embedding is
computed for this weighted edge graph using a publicly available implementation~\cite{isomap_code}.
Figure~\ref{fig:isomap} shows the visualization of isomap features on a sample 3D mesh.

\setlength{\tabcolsep}{2pt}
\newcolumntype{L}[1]{>{\raggedright\let\newline\\\arraybackslash\hspace{0pt}}b{#1}}
\newcolumntype{C}[1]{>{\centering\let\newline\\\arraybackslash\hspace{0pt}}b{#1}}
\newcolumntype{R}[1]{>{\raggedleft\let\newline\\\arraybackslash\hspace{0pt}}b{#1}}
\begin{table}[t]
  \scriptsize
  \centering
    \begin{tabular}{C{3.0cm} C{1.5cm} C{1.5cm} C{1.5cm} C{1.5cm}}
      \toprule
& \textbf{Noisy Mesh} & \textbf{Normal Smoothing} & \textbf{Gauss Bilateral} & \textbf{Learned Bilateral} \\ [0.1cm]
      \midrule
\textbf{Vertex Distance (RMSE)} &  5.774 & 3.183 & 2.872 & \textbf{2.825}  \\
\textbf{Normal Angle Error} & 19.680  & 19.707 & 19.357 & \textbf{19.207}  \\
      \bottomrule
      \\
    \end{tabular}
      \vspace{-0.2cm}
    \mycaption{Body denoising} {Vertex distance RMSE values and normal angle error (in degrees)
    corresponding to different denoising strategies averaged over 500 test meshes.}
\label{tbl:bodydenoise}
\end{table}

\begin{figure}[t!]
  \centering
    \includegraphics[width=0.7\columnwidth]{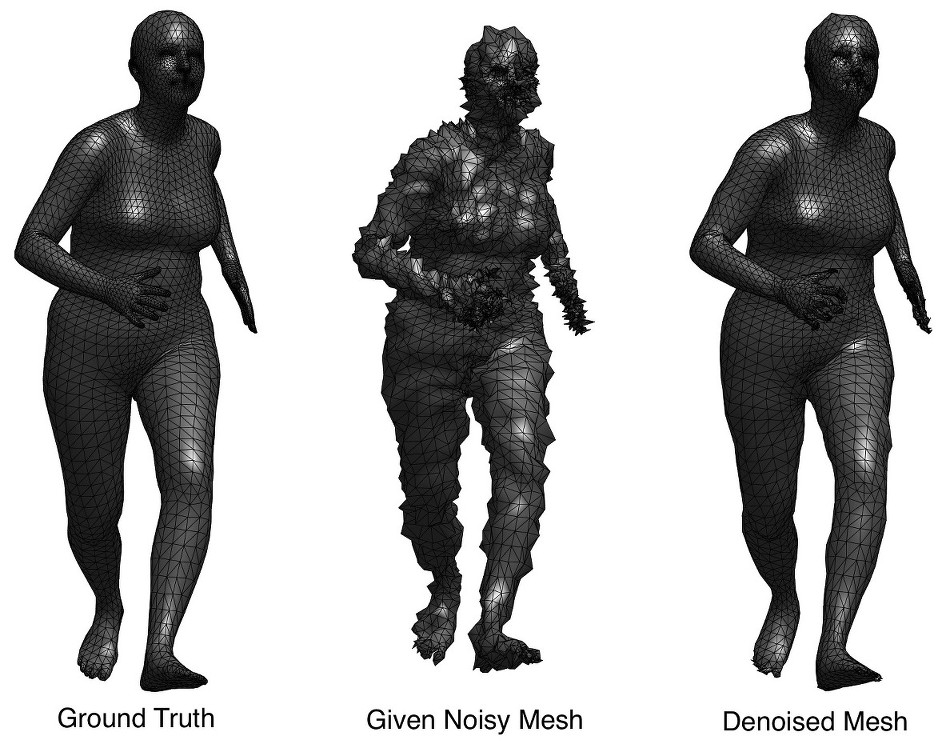}
  \mycaption{Sample denoising result}
  {Ground truth mesh (left), corresponding given noisy mesh (middle) and the denoised result (right) using the learned bilateral filter.}
  \label{fig:bodyresult}
\end{figure}

\paragraph{Experimental Results:} Mesh denoising with a bilateral filter proceeds by
splatting the input 3D mesh vectors (displacements with respect to the smoothed mesh) into
the 4D isomap feature space, filtering the signal in this 4D space and then slicing back
into original 3D input space. The Table~\ref{tbl:bodydenoise} shows quantitative results as RMSE
for different denoising strategies. The normal smoothing~\cite{yagou2002mesh} already reduces the RMSE. The Gauss bilateral filter
results in significant improvement over normal
smoothing and learning the filter weights again improves the result.
A visual result is shown in Fig.~\ref{fig:bodyresult}.

\section{Learning Pairwise Potentials in Dense CRFs}\label{sec:densecrf}
The bilateral convolution from Section~\ref{sec:learning} generalizes
the class of DenseCRF models for which the mean-field inference from~\cite{krahenbuhl2012efficient}
applies. The DenseCRF models have found wide-spread use in various
computer vision applications~\cite{sun2013fully, bell2014intrinsic, zheng2015conditional, vineet12eccv, vineet12emmcvpr, bell2015minc}. Recall from Section~\ref{sec:example_modesl_crf},
for a DenseCRF model with unary potentials $\psi_u$ and pairwise potentials $\psi^{ij}_p$,
the mean-field inference results in a fixed point equation which
can be solved iteratively to update the marginal distributions $Q_i$. In iteration $t$, we have:

\begin{equation}
Q^{t+1}_i(x_i) = \frac{1}{Z_i} \exp\{-\psi_u(x_i) - \sum_{l \in \mathcal{L}}\underbrace{\sum_{j \ne i}
\psi^{ij}_p(x_i,l) Q^{t}_j(l)}_\text{bilateral filtering}\}.
\label{eq:mfupdate-2}
\end{equation}

Thus bilateral filtering is used for fast mean-field inference in DenseCRF models.
One of the fundamental limitations with the existing use of DenseCRFs is the
confinement of pairwise potentials $\psi^{ij}_p(y_i,y_j)$ to be Gaussian as
bilateral filtering is traditionally applied with Gaussian kernel.

\subsection{Learning Pairwise Potentials}

The proposed bilateral convolution generalizes the class of potential functions $\psi^{ij}_p$, since
they allow a richer class of kernels $k(\f_i,\f_j)$ that furthermore can be learned from data. So far, all
dense CRF models have used Gaussian potential functions $k$, we replace it with the
general bilateral convolution and learn the parameters of kernel $k$, thus in effect
learn the pairwise potentials of the dense CRF. This retains the desirable properties of this
model class -- efficient inference through mean-field and the feature dependency of the pairwise
potential.
In order to learn the form of the pairwise potentials $k$ we make use of the
gradients for filter parameters in $k$ and
use back-propagation through the mean-field iterations~\cite{domke2013learning, li2014mean} to learn them.

The work of~\cite{krahenbuhl2013parameter} derived gradients to learn the feature scaling $\Lambda$ but not the
form of the kernel $k$, which still was Gaussian. In~\cite{campbell2013fully}, the features $\f_i$ were
derived using a non-parametric embedding into a Euclidean space and again a Gaussian
kernel was used. The computation of the embedding was a pre-processing step, not integrated in a
end-to-end learning framework. Both aforementioned works are generalizations that are orthogonal
to our development and can be used in conjunction.

\subsection{Experimental Evaluation}

We evaluate the effect of learning more general forms of potential functions on two
pixel labeling tasks, semantic segmentation of VOC data~\cite{voc2012segmentation} and material
classification~\cite{bell2015minc}. We use pre-trained models from the literature
and compare the relative change when learning the pairwise potentials, as in the last section. For both
the experiments, we use multinomial logistic classification loss and learn the filters via
back-propagation~\cite{domke2013learning}. This has also been understood as recurrent neural
network variants~\cite{zheng2015conditional},
the following experiments demonstrate the learnability of bilateral filters.

\begin{figure}[t]
  \centering
  \subfigure{%
    \includegraphics[width=.23\columnwidth]{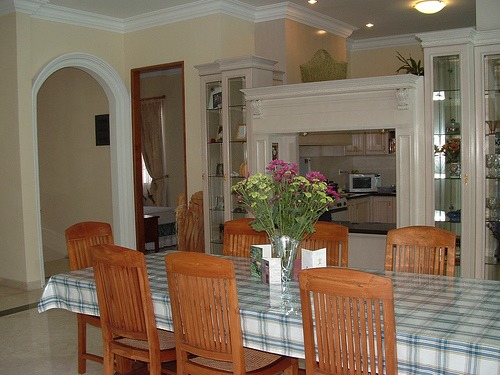} \label{fig:seg_given}
  }
  \subfigure{%
    \includegraphics[width=.23\columnwidth]{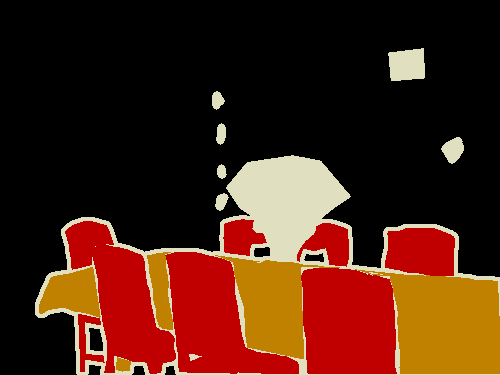} \label{fig:seg_gt}
  }
  \subfigure{%
    \includegraphics[width=.23\columnwidth]{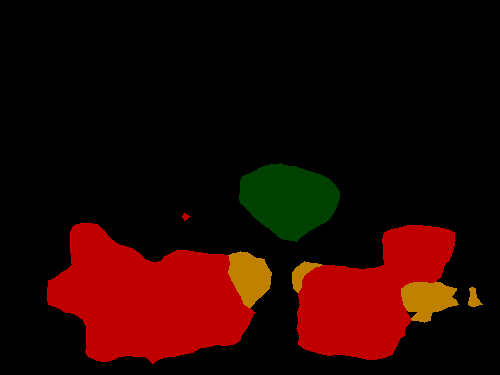} \label{fig:seg_cnn}
  }
  \subfigure{%
    \includegraphics[width=.23\columnwidth]{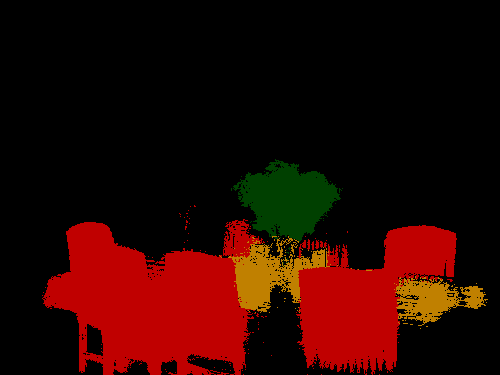} \label{fig:seg_mf2steps}
  }\\
  \setcounter{subfigure}{0}
  \subfigure[Input]{%
    \includegraphics[width=.23\columnwidth]{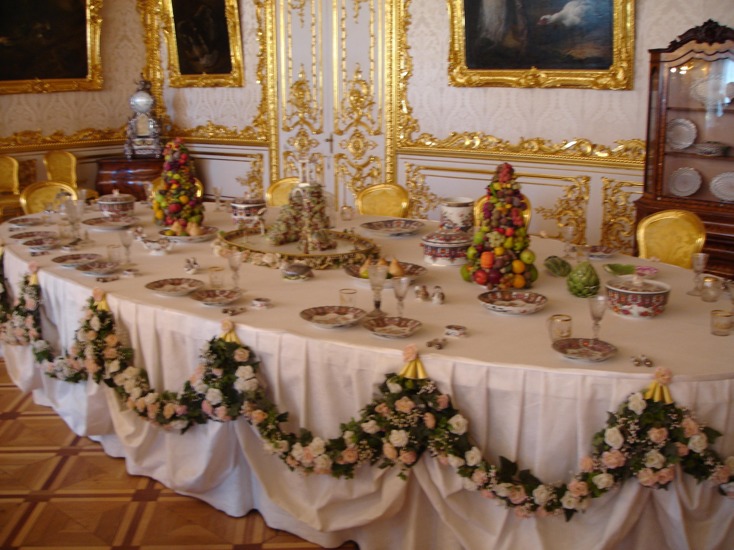} \label{fig:minc_given}
  }
  \subfigure[Ground Truth]{%
    \includegraphics[width=.23\columnwidth]{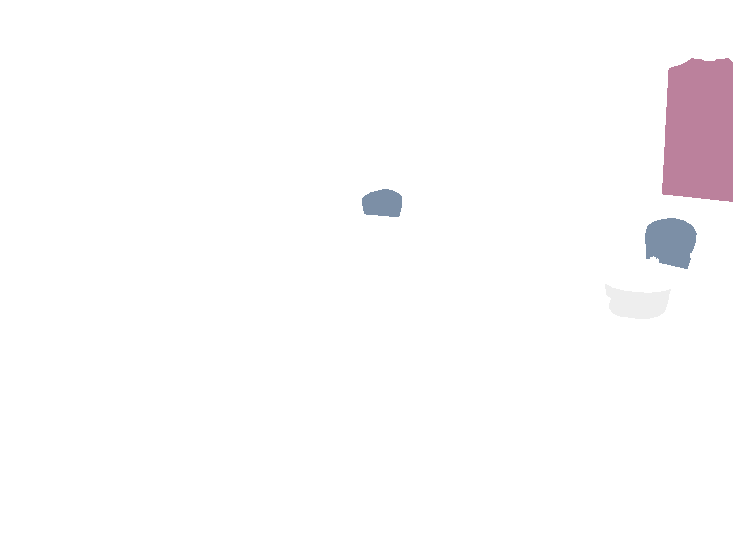} \label{fig:minc_gt}
  }
  \subfigure[CNN]{%
    \includegraphics[width=.23\columnwidth]{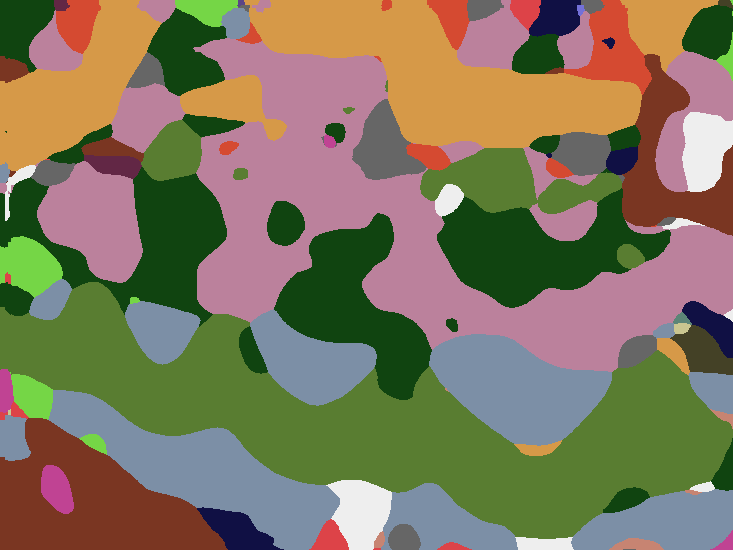} \label{fig:minc_cnn}
  }
  \subfigure[+\textit{loose}MF]{%
    \includegraphics[width=.23\columnwidth]{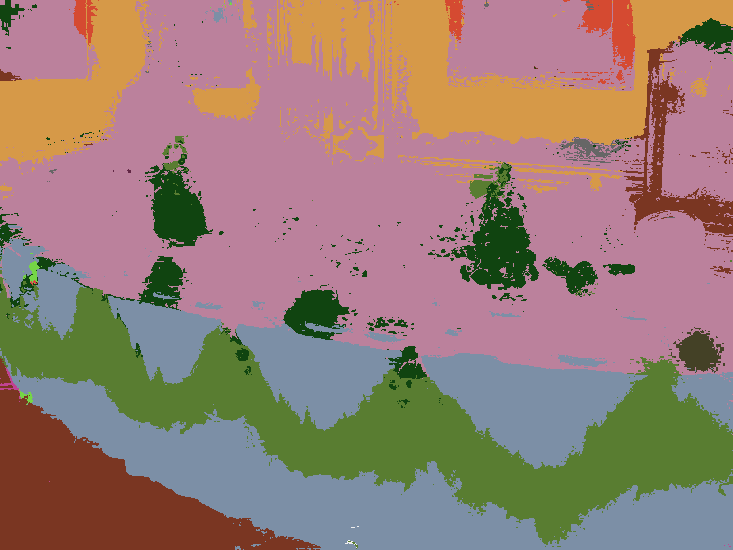} \label{fig:minc_mf2steps}
  }
  \mycaption{Segmentation results}{An example result for semantic (top) and material (bottom) segmentation.
  (c) depicts the unary results before application of MF, (d) after two steps of \textit{loose}-MF with
  a learned CRF. More examples with comparisons to Gaussian pairwise potentials can be found in the supplementary material.}
    \label{fig:seg_visuals}
\end{figure}

\subsubsection{Semantic Segmentation}
Semantic segmentation is the task of assigning a semantic label
to every pixel. We choose the DeepLab network~\cite{chen2014semantic}, a
variant of the VGGnet~\cite{simonyan2014very} for obtaining unaries.
The DeepLab architecture runs a CNN
model on the input image to obtain a result that is down-sampled by a factor of
8. The result is then bilinear interpolated to the desired resolution and serves
as unaries $\psi_u(x_i)$ in a dense CRF. We use the same Pott's label
compatibility function $\mu$, and also use two kernels $k^1(\f_i,\f_j) +
k^2(p_i,p_j)$ with the same features $\f_i=(x_i,y_i,r_i,g_i,b_i)^{\top}$ and $\mathbf{p}_i=(x_i,y_i)^{\top}$
as in~\cite{chen2014semantic}. Thus, the two filters operate in parallel on
color \& position, and spatial domain respectively. We also initialize the
mean-field update equations with the CNN unaries. The only change in the model
is the type of the pairwise potential function from Gauss to a generalized form.

We evaluate the result after 1 step and 2 steps of mean-field inference and
compare the Gaussian filter versus the learned version (cf.
Tab.~\ref{tbl:seg_results}). First, as in~\cite{chen2014semantic} we observe
that one step of mean field improves the performance by 2.48\% in Intersection
over Union (IoU) score. However, a learned potential increases the score by
2.93\%. The same behavior is observed for 2 steps: the learned result again
adds on top of the raised Gaussian mean field performance. Further, we tested a
variant of the mean-field model that learns a separate kernel for the first and
second step~\cite{li2014mean}. This `loose' mean-field model
leads to further improvement of the performance. It is not obvious
how to take advantage
of a loose model in the case of Gaussian potentials.

\begin{table}[t]
\setlength{\tabcolsep}{2pt}
  \scriptsize
  \centering
    \begin{tabular}{l c c c c}
      \toprule
&  & + \textbf{MF-1step} & + \textbf{MF-2 step} & + \textbf{\textit{loose} MF-2 step} \\ [0.1cm]
      \midrule
      \multicolumn{4}{l}{Semantic segmentation (IoU) - CNN~\cite{chen2014semantic}: 72.08 / 66.95} & \\
      & Gauss CRF & +2.48 & +3.38  & +3.38 / +3.00   \\
      & Learned CRF & +2.93 & +3.71  &  \textbf{+3.85 / +3.37} \\
     \midrule
     \multicolumn{4}{l}{Material segmentation (Pixel Accuracy) - CNN~\cite{bell2015minc}: 67.21 / 69.23 } & \\
      & Gauss CRF & +7.91 / +6.28 & +9.68 / +7.35  &  +9.68 / +7.35  \\
      & Learned CRF & +9.48 / +6.23 & +11.89 / +6.93  &  \textbf{+11.91 / +6.93} \\
      \\
    \end{tabular}
    \mycaption{Improved mean-field inference with learned potentials} {(top) Average IoU score on Pascal VOC12
    validation/test data~\cite{voc2012segmentation} for semantic segmentation; (bottom) Accuracy for
    all pixels / averaged over classes on the MINC test data~\cite{bell2015minc} for material segmentation.}
  \label{tbl:seg_results}
\end{table}

\subsubsection{Material Segmentation}

We adopt the method and dataset from~\cite{bell2015minc} for the material segmentation task.
Their approach
proposes the same architecture as in the previous section; a CNN to predict the material labels (\eg\ wool, glass, sky, etc.)
followed by a densely connected CRF using Gaussian potentials and mean-field inference.
We re-use the pre-trained CNN and choose the
CRF parameters and Lab color/position features as in~\cite{bell2015minc}.
Results for pixel accuracy and class-averaged pixel accuracy are shown in Table~\ref{tbl:seg_results}.
Following the CRF validation in~\cite{bell2015minc}, we ignored the label `other' for both the training and
evaluation. For this dataset, the availability of training data is small, 928 images with
only sparse segment annotations. While this is enough to cross-validate few hyper-parameters, we would
expect the general bilateral convolution to benefit from
more training data.
Visual results are shown in Fig.~\ref{fig:seg_visuals} and more are included in
Appendix~\ref{sec:semantic_bnn_extra}.

\section{Bilateral Neural Networks}\label{sec:bnn}

Probably the most promising opportunity for the generalized bilateral filter is its
use in Convolutional Neural Networks. Since we are not restricted to the Gaussian case,
we can stack several filters in both parallel and sequential manner in the same way as filters are ordered in
layers in typical spatial CNN architectures. Having the gradients available allows for end-to-end training
with back-propagation, without the need for any change in CNN training protocols. We refer to the
layers of bilateral filters as `bilateral convolution layers' (BCL). As discussed in the introduction,
these can be understood as either linear filters in a high dimensional space or a filter with an
image adaptive receptive field. In the remainder, we will refer to CNNs that include at least one bilateral
convolutional layer as a bilateral neural network (BNN).

\setlength{\tabcolsep}{4pt}
\begin{table}[h]
  \scriptsize
  \centering
    \begin{tabular}{l c c}
      \toprule
\textbf{Dim.-Features} & \textbf{d-dim caffe} & \textbf{BCL} \\ [0.1cm]
      \midrule
      2D-$(x, y)$ & \textbf{3.3 $\pm$ 0.3 / 0.5$ \pm$ 0.1} & 4.8 $\pm$ 0.5 / 2.8 $\pm$ 0.4  \\
      3D-$(r,g,b)$ & 364.5 $\pm$ 43.2 / 12.1 $\pm$ 0.4 & \textbf{5.1 $\pm$ 0.7 / 3.2 $\pm$ 0.4}  \\
      4D-$(x,r,g,b)$ & 30741.8 $\pm$ 9170.9 / 1446.2 $\pm$ 304.7 & \textbf{6.2 $\pm$ 0.7 / 3.8 $\pm$ 0.5}  \\
      5D-$(x,y,r,g,b)$ & out of memory & \textbf{7.6 $\pm$ 0.4 / 4.5 $\pm$ 0.4}  \\
      \\
    \end{tabular}
    \mycaption{Runtime comparison: BCL vs. spatial convolution} {Average CPU/GPU runtime (in ms)
    of 50 1-neighborhood filters averaged over 1000 images from Pascal VOC. All scaled
    features $(x,y,r,g,b)\in[0,50)$. BCL includes splatting and splicing operations which in layered networks
    can be re-used.}
\label{tbl:time}
\end{table}

What are the possibilities of a BCL compared to a standard spatial layer? First, we can define a
feature space $\f_i\in\mathbb{R}^d$ to define proximity between elements to perform
the convolution. This can include color or intensity as in the previous example.
We performed a runtime comparison (Tab.~\ref{tbl:time}) between
our current implementation of a BCL and the caffe~\cite{jia2014caffe} implementation of
a $d$-dimensional convolution. For 2D positional features (first row), the
standard layer is faster since the permutohedral algorithm comes with an overhead. For higher
dimensions $d>2$, the runtime depends on the sparsity; but ignoring the sparsity is quickly
leading to intractable runtimes for the regular $d$-dimensional convolution.
The permutohedral lattice convolution is in effect a sparse matrix-vector product and thus
performs favorably in this case. In the original work~\cite{adams2010fast}, it was presented
as an approximation to the Gaussian case, here we take the viewpoint of it being the definition of
the convolution itself.

Next we illustrate two use cases of BNNs and compare against spatial CNNs.
Appendix~\ref{sec:appendix-bnn} contains further explanatory experiments
with examples on MNIST digit recognition.

\begin{figure}[t!]
  \centering
  \subfigure[Sample tile images]{%
    \includegraphics[width=0.6\columnwidth]{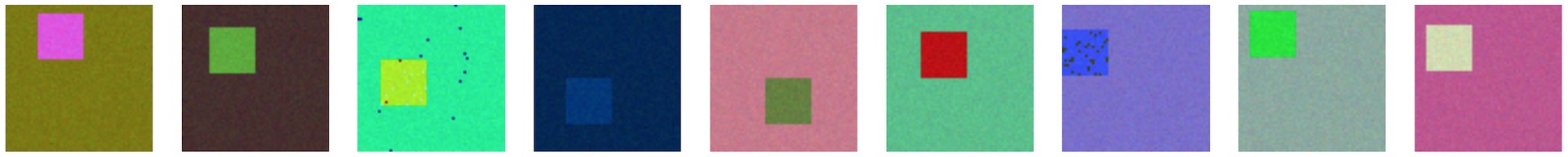}\label{fig:slate_sample}
  }\\
  \subfigure[NN architecture]{%
    \includegraphics[width=.3\columnwidth]{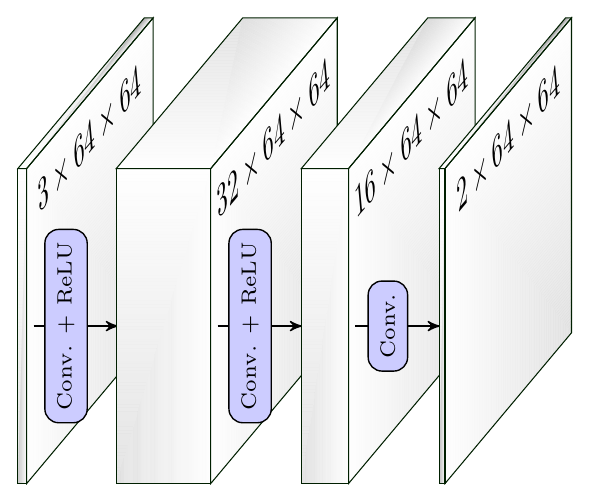}\label{fig:slate_network}
  }
  \subfigure[IoU versus Epochs]{%
    \includegraphics[width=.3\columnwidth]{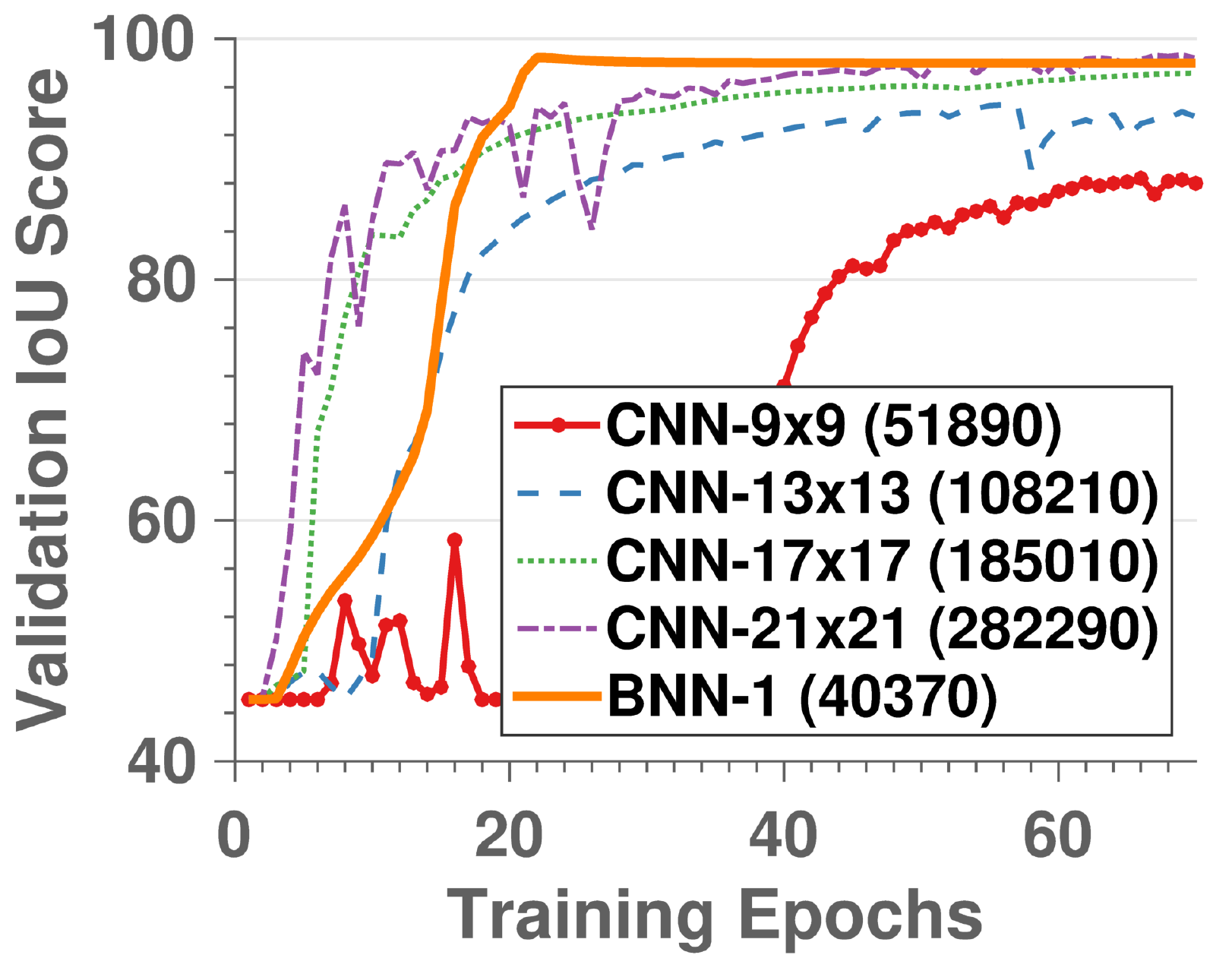}\label{fig:slate_plots}
  }
  \mycaption{Segmenting Tiles}{(a) Example tile input images; (b) the 3-layer NN architecture used
  in experiments. `Conv' stands for spatial convolutions, resp.~bilateral convolutions;
  (c) Training progress in terms of validation IoU versus training epochs.}
  \label{fig:slate_experiment}
\end{figure}

\subsection{An Illustrative Example: Segmenting Tiles}

In order to highlight the model possibilities of using higher dimensional sparse feature
spaces for convolutions through BCLs, we constructed the following illustrative problem. A randomly colored
foreground tile with size $20\times 20$ is placed on a random colored background
of size $64 \times 64$. Gaussian noise with standard deviation of $0.02$ is added and color values
are normalized to $[0,1]$, example images
are shown in Fig.~\ref{fig:slate_sample}.
The task is to segment out the smaller tile. A pixel classifier can
not distinguish foreground from background since the color is random. We train CNNs with three
convolution/ReLU layers and varying
filters of size $n \times n, n \in \{9, 13, 17, 21\}$.
The schematic of the architecture is shown
in Fig~\ref{fig:slate_network} ($32,16,2$ filters).
We create 10k training, 1k validation and 1k test images and, use the
validation set to choose learning rates. In
Fig.~\ref{fig:slate_plots}, we plot the validation IoU against training epochs.

Now, we replace all spatial convolutions with bilateral convolutions for a full BNN.
The features are $\f_i=(x_i,y_i,r_i,g_i,b_i)^{\top}$ and the filter has a neighborhood of $1$.
The total number of
parameters in this network is around $40k$ compared to $52k$ for $9\times 9$ up to $282k$ for a $21\times 21$
CNN. With the same training protocol
and optimizer, the convergence rate of BNN is much faster. In this example
as in semantic segmentation discussed in the last section, color is a discriminative information for
the label. The bilateral convolutions \emph{see} the color difference, the points are already
pre-grouped in the permutohedral lattice and the task remains to assign a label to the two groups.

\vspace{-0.4cm}
\subsection{Character Recognition}
The results for tile, semantic, and material segmentation when using general bilateral
filters mainly improved because the feature space was used to encode useful prior information about the
problem (similar RGB of close-by pixels have the same label). Such prior knowledge is often available when
structured predictions are to be made, but the input signal may also be in a sparse format to begin with.
Let us consider handwritten character recognition, one of the prime cases for CNN use.

\begin{figure}[t]
  \centering
  \subfigure[Sample Assamese character images (9 classes, 2 samples each)]{%
    \includegraphics[width=0.6\columnwidth]{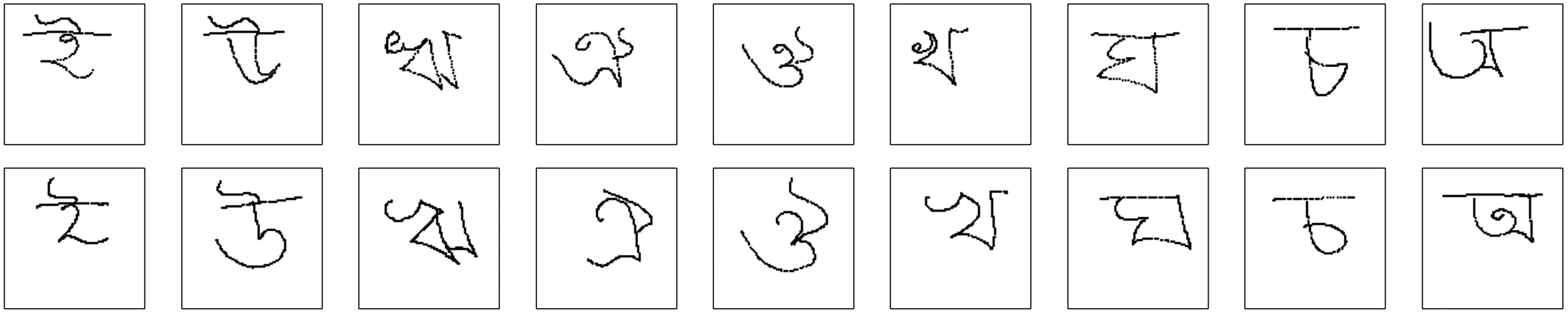}\label{fig:assamese_sample}
  }\\
  \subfigure[LeNet training]{%
    \includegraphics[width=.3\columnwidth]{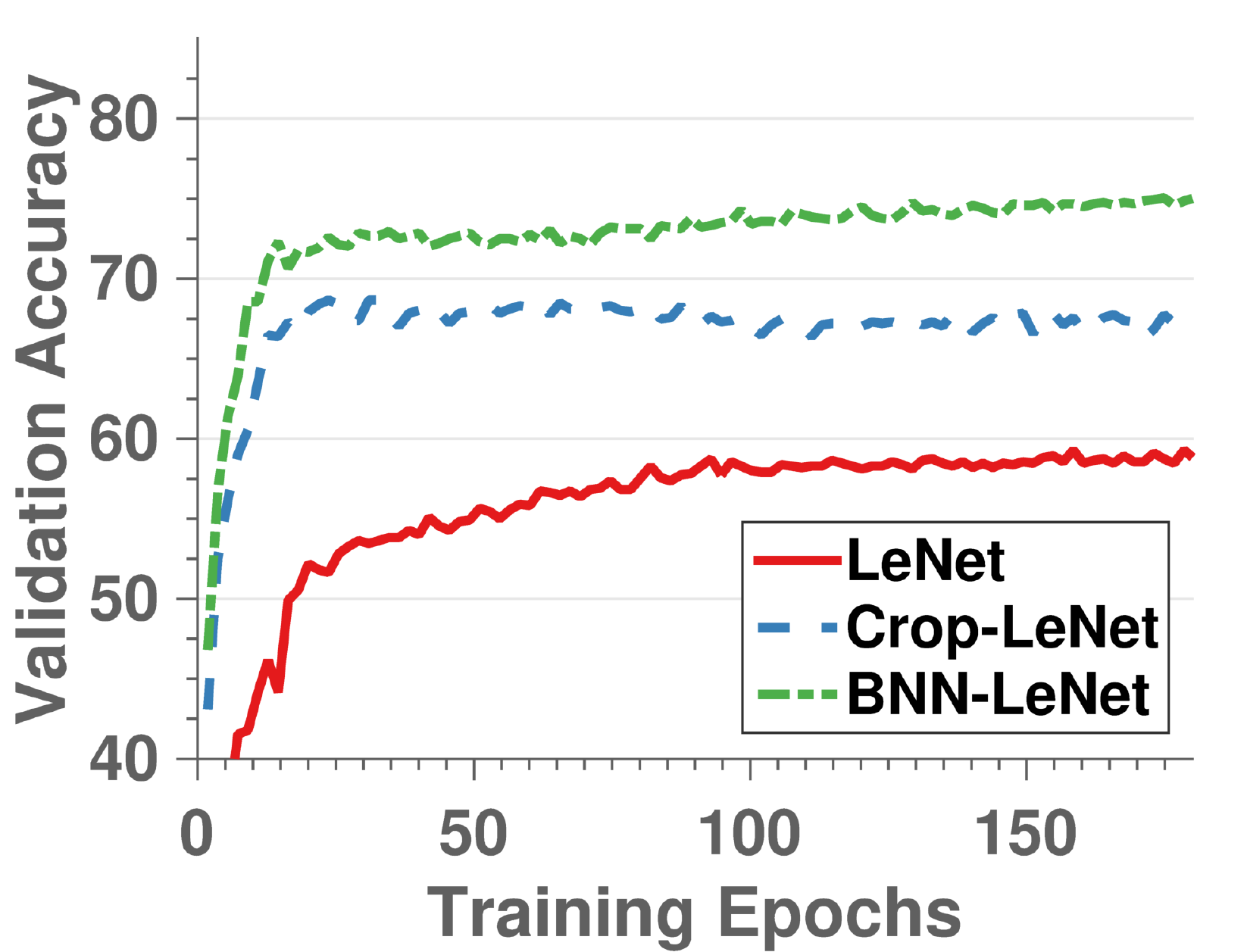}\label{fig:assamese_lenet}
  }
  \subfigure[DeepCNet training]{%
    \includegraphics[width=.3\columnwidth]{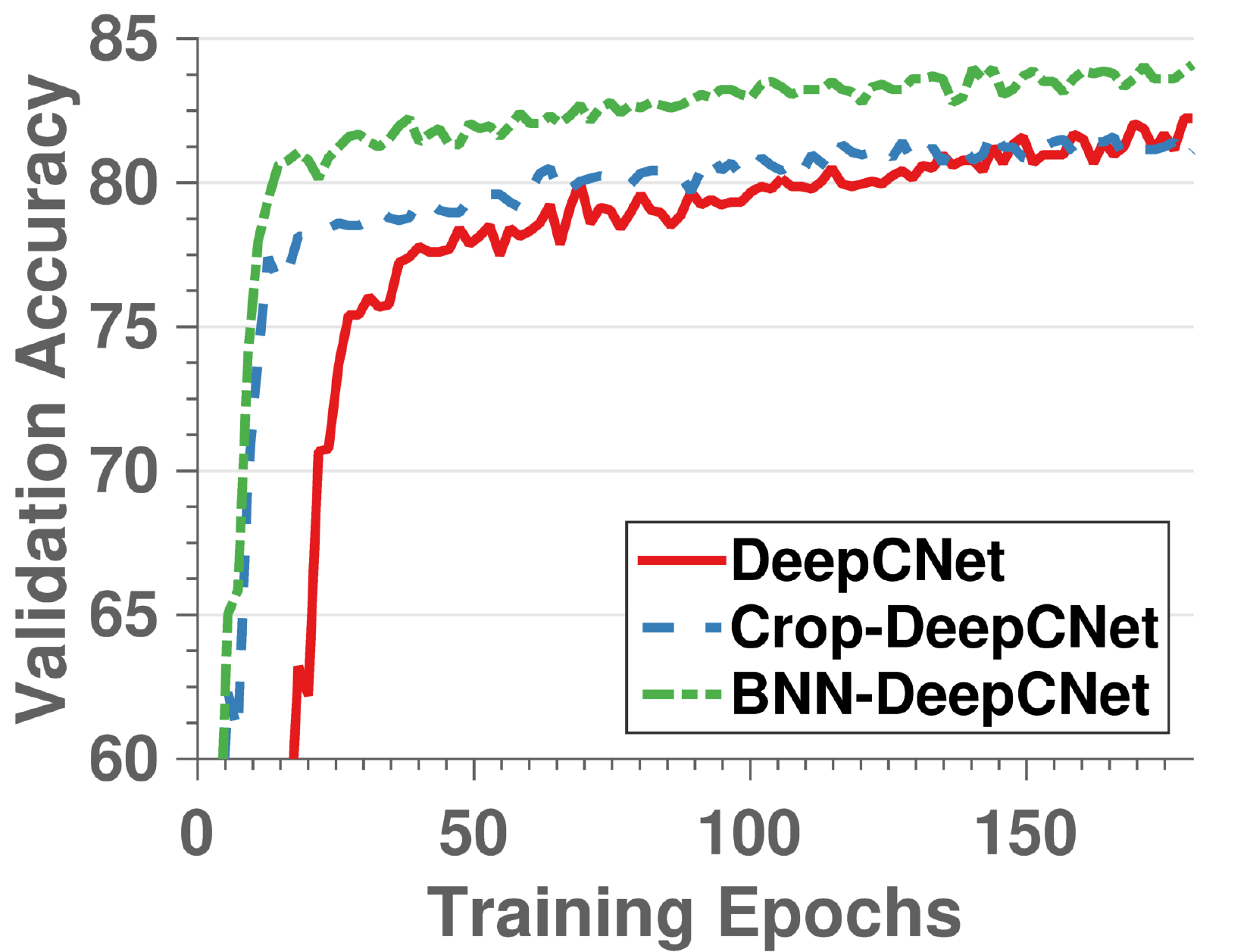}\label{fig:assamese_deepcnet}
  }
  \mycaption{Character recognition}{ (a) Sample Assamese character images~\cite{bache2013uci};
  and training progression of various models with (b) LeNet and (c) DeepCNet base networks.}
\label{fig:assamese_experiment}
  \vspace{-0.3cm}
\end{figure}

The Assamese character dataset~\cite{bache2013uci} contains 183 different Indo-Aryan
symbols with 45 writing samples per class. Some sample character images are shown in
Fig.~\ref{fig:assamese_sample}. This dataset has been collected on a tablet PC using a pen
input device and has been pre-processed to binary images of size $96 \times 96$.
Only about $3\%$ of the pixels contain a pen stroke, which we will denote by $I_i=1$.

A CNN is a natural choice to approach this classification task. We experiment with two CNN
architectures that have been used for this task, LeNet-7 from~\cite{lecun1998gradient}
and DeepCNet~\cite{ciresan2012multi, graham2014spatially}. The LeNet is a shallower network with
bigger filter sizes whereas DeepCNet is deeper with smaller convolutions.
Both networks are fully specified in
Appendix~\ref{sec:addresults}. In order to simplify the task for the networks we cropped the
characters by placing a tight bounding box around them and providing the bounding boxes as input to the
networks. We will call these networks
Crop-LeNet and Crop-DeepCNet. For training, we randomly divided the data into 30 writers for training,
6 for validation and the remaining 9 for
test. Fig.\ref{fig:assamese_lenet} and Fig.~\ref{fig:assamese_deepcnet} show the training progress for various
LeNet and DeepCNet models respectively. DeepCNet is a better choice for this problem and for both cases,
pre-processing the data by cropping improves convergence.

The input is spatially sparse and the BCL provides a natural way to take advantage of this. For
both networks, we create a BNN variant (BNN-LeNet and BNN-DeepCNet) by replacing the
first layer with bilateral convolutions using the features $\f_i = (x_i, y_i)^{\top}$ and we \emph{only}
consider the foreground points $I_i=1$. The values $(x_i,y_i)$ denote the position of the pixel with respect
to the top-left corner of the bounding box around the character. In effect, the lattice is very
sparse which reduces runtime because the convolutions are only performed on $3\%$ of the
points that are actually observed. A bilateral filter has $7$ parameters compared to a receptive field of $3\times 3$ for the first
DeepCNet layer and $5\times 5$ for the first LeNet layer. Thus, a BCL with the same number of filters has fewer parameters.
The result of the BCL convolution is then splatted at all points $(x_i,y_i)$ and
passed on to the remaining spatial layers. The convergence behavior is shown in Fig.\ref{fig:assamese_experiment}
and again we find faster convergence and also better validation accuracy. The empirical results of this
experiment for all tested architectures are summarized in Table~\ref{tbl:assamese}, with BNN variants
clearly outperforming their spatial counterparts.

The absolute results can be vastly improved by making use of virtual examples, \eg
by affine transformations~\cite{graham2014spatially}. The purpose of these experiments is to compare the networks on equal grounds while we
believe that additional data will be beneficial for both networks. We have no reason to believe that a particular
network benefits more.

\setlength{\tabcolsep}{2pt}
\newcolumntype{L}[1]{>{\raggedright\let\newline\\\arraybackslash\hspace{0pt}}b{#1}}
\newcolumntype{C}[1]{>{\centering\let\newline\\\arraybackslash\hspace{0pt}}b{#1}}
\newcolumntype{R}[1]{>{\raggedleft\let\newline\\\arraybackslash\hspace{0pt}}b{#1}}
\begin{table}[t]
  \scriptsize
  \centering
    \begin{tabular}{C{1.65cm} C{1.55cm} C{1.55cm} C{1.55cm} | C{1.55cm} C{1.55cm} C{1.55cm}}
      \toprule
 & \textbf{LeNet} & \textbf{Crop-LeNet} & \textbf{BNN-LeNet}  & \textbf{DeepCNet} & \textbf{Crop-DeepCNet} & \textbf{BNN-DeepCNet} \\ [0.2cm]
      \midrule
      Validation & 59.29 & 68.67 & \textbf{75.05}  & 82.24 & 81.88 & \textbf{84.15} \\
      Test & 55.74 & 69.10 & \textbf{74.98}  & 79.78 & 80.02 & \textbf{84.21}  \\
      \\
    \end{tabular}
    \mycaption{Results on Assamese character images} {Total recognition accuracy for the
    different models.}
  \label{tbl:assamese}
\end{table}

\section{Discussion and Conclusions}
We proposed to learn bilateral filters from data. In
hindsight, it may appear obvious that this leads to performance improvements compared
to a fixed parametric form, \eg the Gaussian. To understand algorithms that
facilitate fast approximate computation of Eq.~\ref{eq:bilateral} as a parameterized
implementation of a bilateral filter with free parameters is the key insight and
enables gradient descent based learning. We relaxed the non-separability in the
algorithm from~\cite{adams2010fast} to allow for more general filter functions.
There is a wide range of possible applications for learned bilateral
filters~\cite{paris2009bilateral} and we discussed some generalizations of
previous work. These include joint bilateral
up-sampling and inference in dense CRFs. We further demonstrated a use case of bilateral
convolutions in neural networks.

The bilateral convolutional layer allows for filters whose
receptive field change given the input image. The feature space view provides a
canonical way to encode similarity between any kind of objects, not only pixels, but \eg bounding boxes,
segmentation, surfaces. The proposed filtering operation is then a natural candidate to
define a filter convolutions on these objects, it takes advantage of sparsity and scales to higher dimensions.
Therefore, we believe that this view will be useful for several problems where CNNs
can be applied. An open research problem is whether the sparse higher dimensional
structure also allows for efficient or compact representations for intermediate layers inside
CNN architectures.

In summary, the proposed technique for learning sparse high-dimensional filters
results in a generalization of bilateral filters that helps in learning task-specific
bilateral filters. In the context of inference, learnable bilateral filters can be used for better
modeling and thus inference in neural network as well as DenseCRF models.

\chapter{Video Propagation Networks}
\label{chap:vpn}

In this chapter, we leverage the learnable bilateral filters developed in the
previous chapter, and develop a novel neural network architecture for inference
in video data. We focus on the task of propagating information across video
frames. Standard CNNs are poor candidates for filtering video data.
Standard spatial CNNs have fixed receptive fields whereas the video content
changes differently in different videos depending on the type of scene and camera
motion. So, filters with video adaptive receptive fields are better candidates
for video filtering.

Based on this observation, we adapt the bilateral convolution layers (BCL) proposed
in the previous chapter for filtering video data.
By stacking several BCL and standard spatial convolutional
layers, we develop a neural network architecture for video information propagation
which we call `Video Propagation Network' (VPN). We evaluate VPN on different
tasks of video object segmentation and semantic video segmentation
and show increased performance comparing to the best previous task-specific methods,
while having favorable runtime. Additionally we demonstrate our approach on an example
regression task of propagating color in a grayscale video.

\section{Introduction}

\begin{figure}[t!]
\begin{center}
\centerline{\includegraphics[width=\columnwidth]{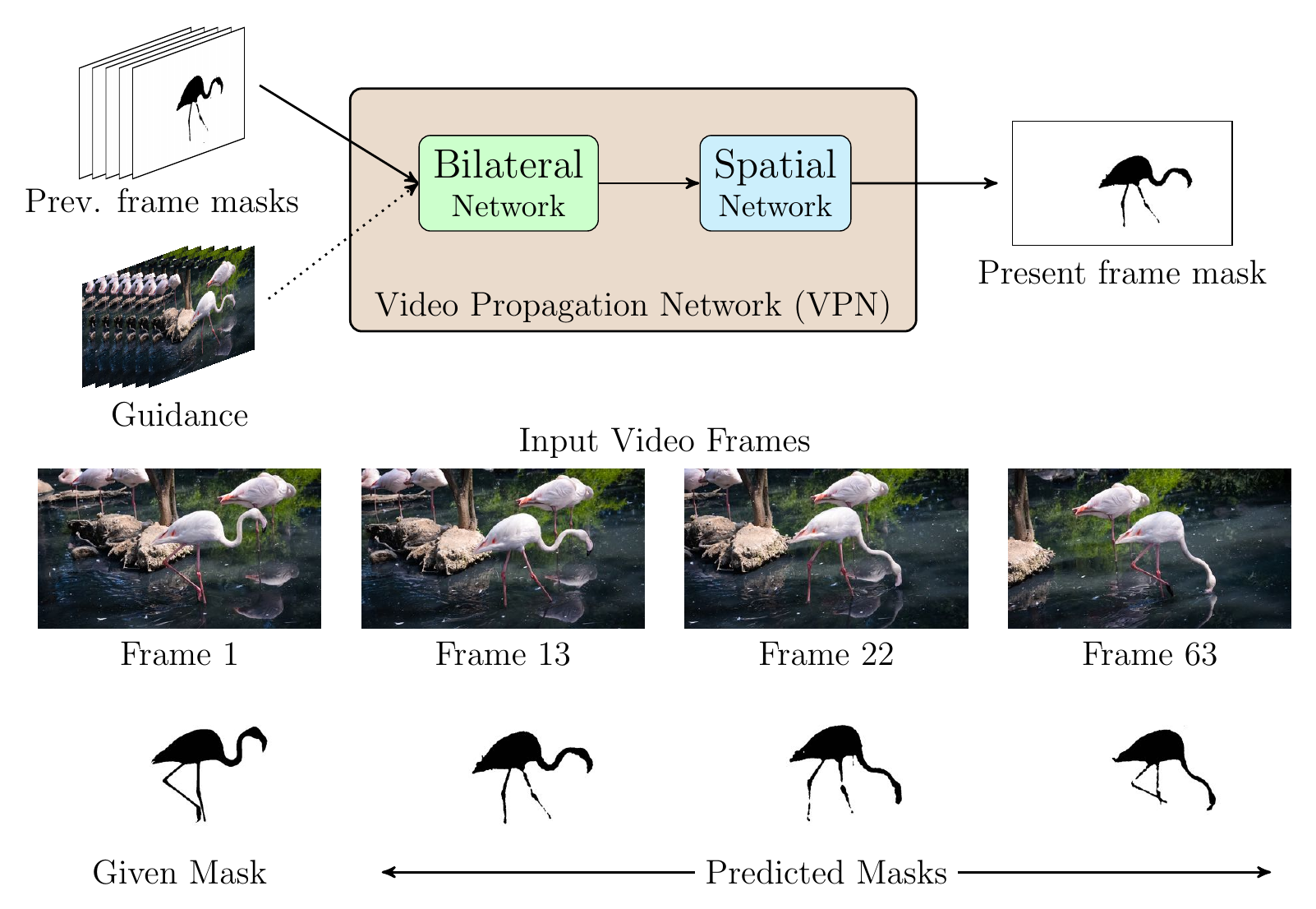}}
  \mycaption{Video Propagation with VPNs} {The end-to-end trained VPN network is composed
  of a bilateral network followed by a standard spatial network and can be used for
  propagating information across frames. Shown here is an example result
  of foreground mask from the 1$^{st}$ frame to other video frames.}
  \label{fig:illustration-vpn}
  \vspace{-0.3cm}
\end{center}
\end{figure}

We focus on the problem of propagating structured information across video frames in this chapter.
~This problem appears in many forms (e.g., semantic segmentation or depth estimation) and is a pre-requisite for many applications.~An example instance is shown in Fig.~\ref{fig:illustration-vpn}.
Given an accurate object mask for the first frame, the problem is to propagate this mask forward
through the entire video sequence.~Propagation of semantic information through time and video
colorization are other problem instances.

Videos pose both technical and representational challenges.
The presence of scene and camera motion lead to the difficult association problem of optical flow.
Video data is computationally more demanding than static images. A naive per-frame approach would scale at least linear with frames.
These challenges complicate the use of standard convolutional neural networks (CNNs) for video processing.
As a result, many previous works for video propagation use slow optimization based techniques.

We propose a generic neural network architecture that propagates information across
video frames. The main innovation is the use of image adaptive convolutional operations that automatically
adapt to
the video stream content.~This allows the network to adapt to the changing content of the video stream.
It can be applied to several types of information, e.g. labels, colors, etc. and runs online, that is, only requiring current and previous frames.

Our architecture is composed of two components (see Fig.~\ref{fig:illustration-vpn}).
A temporal \textit{bilateral network} that performs image-adaptive spatio-temporal dense filtering.
This part allows to connect densely all pixels from current and previous frames and to propagate associated pixel information to the current frame.
The bilateral network allows the specification of a metric between video pixels and allows a straight-forward integration of temporal information.
This is followed by a standard \textit{spatial CNN} on the filter output to refine and predict for the present video frame.
We call this combination a \textit{Video Propagation Network (VPN)}.
In effect we are combining a filtering technique with rather small spatial CNNs which leads to a favorable runtime compared to many previous approaches.

VPNs have the following suitable properties for video processing:

\vspace{-0.5cm}
\paragraph{General applicability:} VPNs can be used for propagating any
type of information content i.e., both discrete
(e.g., semantic labels) and continuous (e.g. color) information across video frames.
\vspace{-0.5cm}
\paragraph{Online propagation:} The method needs no future frames
and so can be used for online video analysis.
\vspace{-0.5cm}
\paragraph{Long-range and image adaptive:} VPNs can efficiently handle a large
number of input frames and are adaptive to the video.
\vspace{-0.5cm}
\paragraph{End-to-end trainable:} VPNs can be trained end-to-end, so they
can be used in other deep network architectures.
\vspace{-0.5cm}
\paragraph{Favorable runtime:} VPNs have favorable runtime in comparison to several current
best methods, also making them amenable for learning with
large datasets.

Empirically we show that VPNs, despite being generic,
perform better or on-par with current best approaches on video object segmentation
and semantic label propagation while being faster.
VPNs can easily be integrated into sequential per-frame approaches
and require only a small fine-tuning step that can be performed separately.

\section{Related Work}
\label{sec:related}

The literature on propagating information across video frames contains a vast and varied number of approaches.
Here, we only discuss those works that are related to our technique and applications.

\vspace{-0.3cm}
\paragraph{General propagation techniques}
Techniques for propagating content across
image or video pixels are predominantly
optimization based or filtering techniques. Optimization
based techniques typically formulate the propagation as an energy minimization problem
on a graph constructed across video pixels or frames.
A classic example is the color propagation technique from~\cite{levin2004colorization} which uses
graph structure that encodes prior knowledge about pixel colors in a local neighborhood.
Although efficient closed-form
solutions~\cite{levin2008closed} exists for certain scenarios,
optimization tends to be slow due to either large graph structures for videos and/or the use of
complex connectivity resulting in the use of iterative optimization schemes. Fully-connected conditional
random fields (CRFs)~\cite{krahenbuhl2012efficient} open a way for incorporating dense
and long-range pixel connections while retaining fast inference.

Filtering techniques~\cite{kopf2007joint,chang2015propagated,he2013guided} aim to propagate
information with the use of image or video filters resulting in fast runtimes compared
to optimization techniques. Bilateral filtering~\cite{aurich1995non,tomasi1998bilateral} is
one of the popular filters for long-range information propagation.
We have already discussed bilateral filtering and its generalization in the previous
chapter. A popular application, that is also discussed in previous chapter, is joint
bilateral up-sampling~\cite{kopf2007joint} that up-samples a low-resolution signal with
the use of a high-resolution guidance image.
Chapter~\ref{chap:bnn} and the works
of~\cite{li2014mean,domke2013learning,zheng2015conditional,schwing2015fully,barron2015bilateral}
showed that one can back-propagate through the bilateral filtering operation for
learning filter parameters (Chapter~\ref{chap:bnn}) or doing optimization in the bilateral
space~\cite{barron2015bilateral,barron2015defocus}.
Recently, several works proposed to do upsampling
in images by learning CNNs that mimics edge-aware filtering~\cite{xu2015deep} or
that directly learns to up-sample~\cite{li2016deep,hui2016depth}.
Most of these works are confined to images and are either not extendible or computationally
too expensive for videos. We leverage some of these previous works and propose a
scalable yet robust neural network based approach for video content propagation.

\vspace{-0.3cm}
\paragraph{Video object segmentation}

Prior work on video object segmentation can be broadly categorized into two types:
Semi-supervised methods that require manual annotation to define what is foreground
object and unsupervised methods that does segmentation completely automatically.
Unsupervised techniques such as
~\cite{faktor2014video,li2013video,lee2011key,papazoglou2013fast,wang2015saliency,
zhang2013video,taylor2015causal,dondera2014interactive}
use some prior information about the foreground objects such as
distinctive motion, saliency etc. And, they typically fail if these
assumptions do not hold in a video.

In this work, we focus on the semi-supervised task of propagating the foreground
mask from the first frame to the entire video. Existing works
predominantly use graph-based optimization frameworks that perform graph-cuts~\cite{boykov2001fast,
boykov2001interactive,shi2000normalized} on video data.
Several of these works~\cite{reso2014interactive,
li2005video,price2009livecut,wang2005interactive,kohli2007dynamic,jain2014supervoxel}
aim to reduce the complexity of graph structure with
clustering techniques such as spatio-temporal superpixels and
optical flow~\cite{tsaivideo}.
Another direction was to estimate correspondence between different frame
pixels~\cite{agarwala2004keyframe,bai2009video,lang2012practical} by using
nearest neighbor fields~\cite{fan2015jumpcut} or optical flow~\cite{chuang2002video}
and then refine the propagated masks with the use of local classifiers.
Closest to our technique are the works of~\cite{perazzi2015fully} and~\cite{marki2016bilateral}.
~\cite{perazzi2015fully} proposed to use fully-connected CRF over the
refined object proposals across the video frames.~\cite{marki2016bilateral} proposed a
graph-cut in the bilateral space. Our approach is similar in the regard that we also
use a bilateral space embedding. Instead of graph-cuts, we learn
propagation filters in the high-dimensional bilateral space with CNNs.
This results in a more generic architecture and allows integration into other deep learning frameworks.

Two contemporary works~\cite{caelles2016one,khoreva2016learning} proposed CNN based
approaches for video object segmentation. Both works rely on fine-tuning a deep network
using the first frame annotation of a given test sequence. This could potentially result
in overfitting to the background.
In contrast, the proposed approach relies only on offline training and thus can be easily adapted
to different problem scenarios.

\vspace{-0.3cm}
\paragraph{Semantic video segmentation}
Earlier methods such as~\cite{brostow2008segmentation,sturgess2009combining} use structure from motion
on video frames to compute geometrical and/or motion features.
More recent works~\cite{ess2009segmentation,chen2011temporally,de2012line,miksik2013efficient,tripathi2015semantic,
kundu2016feature} construct large graphical models on videos and enforce temporal consistency across frames. \cite{chen2011temporally} used dynamic temporal links in their CRF energy formulation.
\cite{de2012line} proposes to use Perturb-and-MAP random field model with spatio-temporal energy terms
based on Potts model and \cite{miksik2013efficient} propagate predictions across time by learning
a similarity function between pixels of consecutive frames.

In the recent years, there is a big leap in the performance of semantic image
segmentation~\cite{long2014fully,chen2014semantic} with the use of CNNs but mostly
applied to images. Recently,~\cite{shelhamer2016clockwork}
proposed to retain the intermediate CNN representations while sliding the image based
CNN across the frames. Another approach, which inspired our work, is to take unary
predictions from CNN and then propagate semantic information across the frames. A recent
prominent approach in this direction is of~\cite{kundu2016feature} which proposes
a technique for optimizing feature spaces for fully-connected CRF.

\section{Video Propagation Networks}
\label{sec:vpn}

We aim to adapt the bilateral filtering operation to predict information forward in time, across video frames.
Formally, we work on a sequence of $n$ (color or grayscale) images $\obs = \{\obs_1, \obs_2, \cdots, \obs_n\}$
and denote with $\target = \{\target_1, \target_2, \cdots, \target_n\}$ a sequence of outputs, one per frame.
Consider as an example, a sequence $\target_1,\ldots,\target_n$ of foreground masks for a moving object in the
scene. Our goal is to develop an online propagation method, that is, a function that has no access to the future frames.
Formally we predict $\target_t$, having observed the video up to frame $t$ and possibly previous $\target_{1,\cdots,t-1}$
\begin{equation}
\mathcal{F}(\target_{t-1}, \target_{t-2}, \cdots; \obs_t, \obs_{t-1}, \obs_{t-2},\cdots) = \target_t.
\end{equation}

\begin{figure*}[t!]
\begin{center}
\centerline{\includegraphics[width=\textwidth]{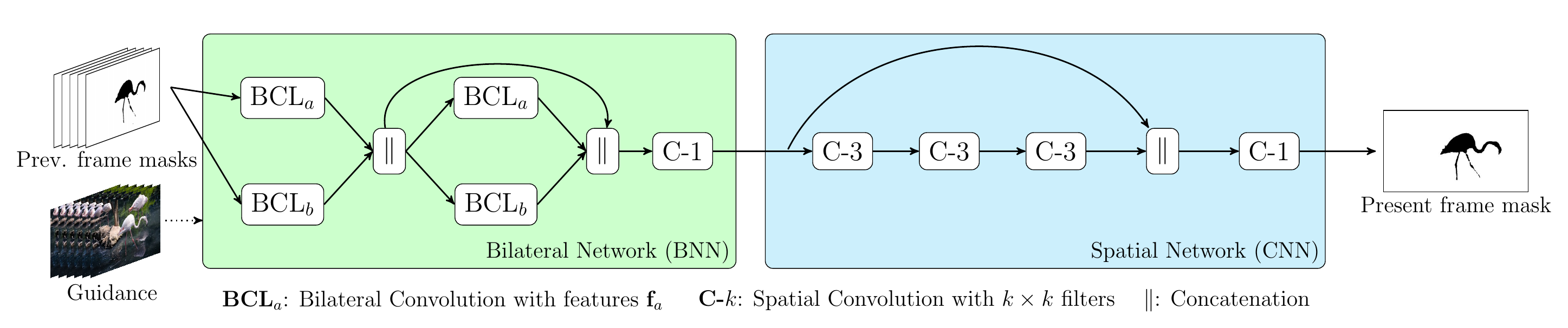}}
  \mycaption{Computation Flow of Video Propagation Network} {Bilateral networks (BNN) consist of a series of bilateral filterings interleaved with ReLU non-linearities. The filtered information from BNN is then passed into a spatial network (CNN) which refines the features with convolution layers interleaved with ReLU
  non-linearities, resulting in the prediction for the current frame.}
  \label{fig:net_illustration}
\end{center}
\end{figure*}

If training examples $(\obs,\target)$ with full or partial knowledge of $\target$ are available, it is possible to learn $\mathcal{F}$ and for a complex and unknown relationship between input and output, a deep CNN is a natural design
choice. However, any learning based method has to face the main challenge: the scene and camera motion and its
effect on $\target$. Since no motion in two different videos is the same, fixed-sized static receptive fields of CNN units are insufficient.
We propose to resolve this with video-adaptive convolutional component, an adaption of the bilateral filtering to videos.
Our Bilateral Network (Section~\ref{sec:bilateralnetwork}) has a connectivity that adapts to video sequences, its output is then fed into a common Spatial Network (Section~\ref{sec:spatialcnn}) that further refines the desired output.
The combined network layout of this Video Propagation Network is depicted in Fig.~\ref{fig:net_illustration}.
It is a sequence of learnable bilateral and spatial filters that is efficient, trainable end-to-end and adaptive to the video input.

\begin{figure*}[t!]
\begin{center}
  \centerline{\includegraphics[width=\textwidth]{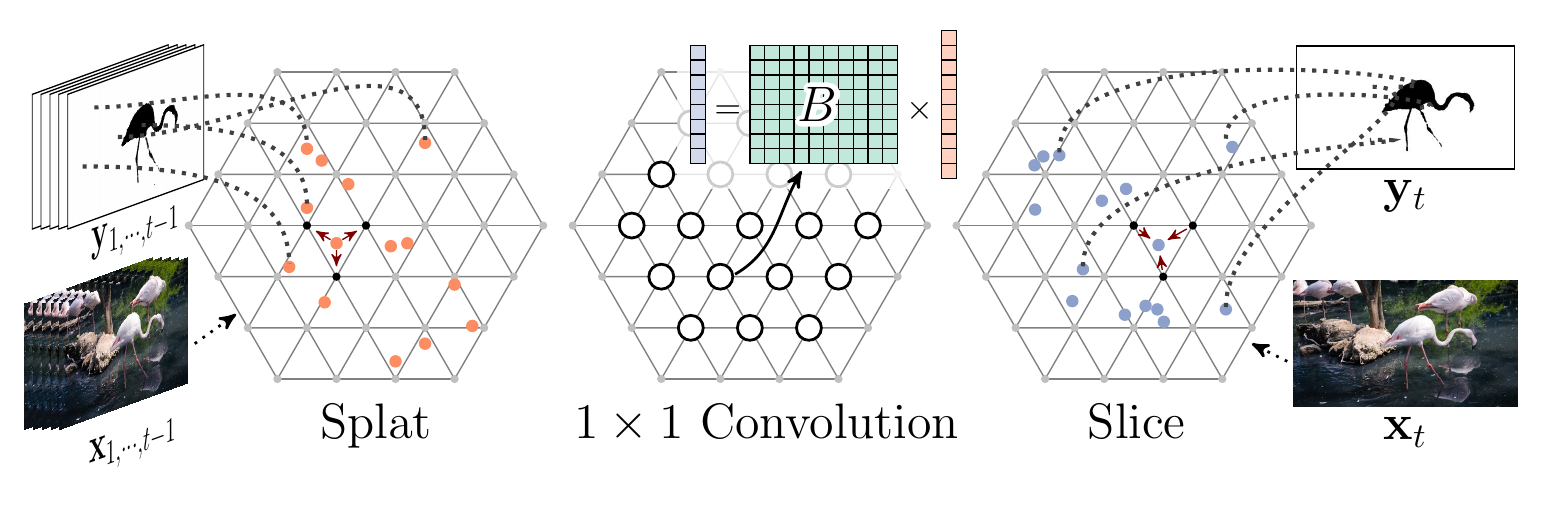}}
    \mycaption{Schematic of Fast Bilateral Filtering for Video Processing}
    {Mask probabilities from previous frames $V_{1,\cdots,t-1}$ are splatted on to the
    lattice positions defined by the image features $f_{I_{1}},f_{I_2},\cdots,f_{I_{t-1}}$.
    The splatted result is convolved with a $1 \times 1$ filter $B$, and the filtered
    result is sliced back to the original image space to get $V_t$ for the present frame.
    Input and output need not be $V_t$, but can also be an intermediate neural network representation.
    $B$ is learned via back-propagation through these operations.}
    \label{fig:filter_illustration}
\end{center}
\end{figure*}

\subsection{Bilateral Network (BNN)}\label{sec:bilateralnetwork}
In this section, we describe the extension of the learnable bilateral filtering, proposed in
Chapter~\ref{chap:bnn} to video data.
Several properties of bilateral filtering make it a perfect candidate for information propagation in videos.
In particular, our method is inspired by two main ideas that we extend in this work: joint bilateral up-sampling~\cite{kopf2007joint} and learnable bilateral filters (Chapter~\ref{chap:bnn}). Although,
bilateral filtering has been used for filtering video data before~\cite{paris2008edge},
its use has been limited to fixed filter weights (say, Gaussian).

{\bf Fast Bilateral Up-sampling across Frames} The idea of joint bilateral up-sampling~\cite{kopf2007joint}
is to view up-sampling as a filtering operation.
A high resolution guidance image is used to up-sample a low-resolution result.
In short, a smaller number of input points $\target_i$ and the corresponding features $\f_i$
are given $\target_i,\f_i; i=1,\ldots,N_{in}$, for example a segmentation result $\target_i$ at a lower resolution.
This is then scaled to a larger number of output points $\f_j;j=1,\ldots,N_{out}$ using the
bilateral filtering operation, that is to compute the following bilateral filtering equation:

\begin{equation}
  \target'_i = \sum_{j=1}^n \bw_{\f_i,\f_j} \target_j
  \label{eq:bilateral2}
\end{equation}

where the sum runs over all $N_{in}$ points and the output is computed for all $N_{out}$ positions.
We will use this idea to propagate content from previous frames to the current frame (all of which have the same dimensions), using the current frame as a guidance image.
This is illustrated in Fig.~\ref{fig:filter_illustration}. We take all previous frame results $\target_{1,\cdots,t-1}$
 and splat them into a lattice using the features computed on video frames $\obs_{1,\cdots,t-1}$.
A filtering (described below) is applied to every lattice point and the result is then sliced back using the
current frame $\obs_t$.
This result need not be the final $\target_t$, in fact we compute a filter bank of responses and continue with further
processing as will be discussed.

For videos, we need to extend bilateral filtering to temporal data, and there are two natural choices.
First, one can simply attach a frame index $t$ as an additional time dimension to the input data, yielding a six dimensional feature vector $\f=(x,y,r,g,b,t)^{\top}$ for every pixel in every frame.
The summation in Eq.~\ref{eq:bilateral2} now runs over \emph{all} previous frames and pixels.
Imagine a video where an object moves to reveal some background.
Pixels of the object and background will be close spatially $(x,y)$ and temporally $(t)$ but likely be of different color $(r,g,b)$.
Therefore they will have no strong influence on each other (being splatted to distant positions in the six-dimensional bilateral space).
In summary, one can understand the filter to be adaptive to color changes across frames, only pixels that are static and have similar color have a strong influence on each other (end up nearby in the lattice space).
The second possibility is to use optical flow. If the perfect flow is available, the video frames could be warped into a common frame of reference. This would resolve the corresponding problem and make information propagation much easier.
We can make use of an optical flow estimate by warping pixel positions $(x,y)$ by their displacement vector $(u_x,u_y)$ to $(x+u_x,y+u_y)$.

Another property of permutohedral filtering that we exploit
is that the \emph{inputs points need not lie on a regular grid} since the filtering
is done in the high-dimensional lattice. Instead of splatting millions of pixels on to the
lattice, we randomly sample or use superpixels and perform filtering using these sampled
points as input to the filter. In practice, we observe that this results in big computational
gains with minor drop in performance (more in Sec.~\ref{sec:videoseg}).

{\bf Learnable Bilateral Filters} The property of propagating information forward using a guidance image through filtering solves the problem of pixel association.
But a Gaussian filter may be insufficient and further, we would like to increase the capacity by using a filter bank instead of a single fixed filter.
We propose to use the technique proposed in previous chapter
to learn the filter values in the permutohedral lattice using back-propagation.

The process works as follows.
A input video is used to determine the positions in the bilateral space to splat the input points
$\target(i)\in\mathbb{R}^D$ i.e. the features $\f$ (e.g. $(x,y,r,g,b,t)$) define the splatting matrix $S_{splat}$.~This leads to a number of vectors $\target_{splatted} = S_{splat}\target$, that lie on the permutohedral lattice, with dimensionality $\target_{splatted}\in\mathbb{R}^D$.
In effect, the splatting operation groups points that are close together, that is, they have similar $\f_i,\f_j$.
All lattice points are now filtered using a filter bank $B\in\mathbb{R}^{F\times D}$ which results in $F$ dimensional vectors on the lattice points.
These are sliced back to the $N_{out}$ points of interest (present video frame).
The values of $B$ are learned by back-propagation.
General parameterization of $B$ from previous chapter allows to have
any neighborhood size for the filters. Since constructing the neighborhood structure in
high-dimensions is time consuming, we choose to use $1 \times 1$ filters for speed reasons.
This makes up one \emph{Bilateral Convolution Layer (BCL)} which we will stack and concatenate to form a Bilateral Network. See Fig.~\ref{fig:filter_illustration} for an illustration of a BCL.

{\bf BNN Architecture} The Bilateral Network (BNN) is illustrated in the green box of
Fig.~\ref{fig:net_illustration}.
The input is a video sequence $\obs$ and the corresponding predictions $\target$ up to frame
$t$. Those are filtered using two BCLs with $32$ filters each.
For both BCLs, we use the same features $\f$ but scale them with different diagonal matrices
$\f_a=\Lambda_a\f,\f_b=\Lambda_b\f$. The feature scales are found by cross-validation.
The two $32$ dimensional outputs are concatenated, passed through a ReLU non-linearity and passed to a
second layer of two separate BCL filters that uses same feature spaces $\f_a,\f_b$.
The output of the second filter bank is then reduced using a $1\times 1$ spatial filter (C-1) to map to
the original dimension of $\target$.
We investigated scaling frame inputs with an exponential time decay and found that, when processing
frame $t$, a re-weighting with $(\alpha \target_{t-1}, \alpha^2 \target_{t-2}, \alpha^3 \target_{t-3} \cdots)$ with
$0\le\alpha\le 1$ improved the performance a little bit.

In the experiments, we also included a simple BNN variant,
where no filters are applied inside the permutohedral space, just splatting and slicing
with the two layers $BCL_a$ and $BCL_b$ and adding the results.
We will refer to this model as \emph{BNN-Identity},
it corresponds to an image adaptive smoothing of the inputs $\target$.
We found this filtering to have a positive effect and include it as a baseline in our experiments.

\vspace{-0.1cm}
\subsection{Spatial Network}\label{sec:spatialcnn}

The BNN was designed to propagate the information from the previous frames, respecting the scene and object motion.
We then add a small spatial CNN with 3 layers, each with $32$ filters of size $3\times 3$,
interleaved with ReLU non-linearities.
The final result is then mapped to the desired output of $\target_t$ using a $1\times 1$
convolution.
The main role of this spatial CNN is to refine the information in frame $t$.
Depending on the problem and the size of the available training data, other network designs are
conceivable. We use the same network architecture shown in Fig.~\ref{fig:net_illustration}
for all the experiments to demonstrate the generality of VPNs.

\vspace{-0.1cm}
\section{Experiments}
\label{sec:exps}

We evaluated VPN on three different propagation tasks: foreground masks, semantic
labels and color information in videos. Our implementation runs in Caffe~\cite{jia2014caffe} using standard settings. We used Adam~\cite{kingma2014adam} stochastic optimization for training VPNs, multinomial-logistic loss for label propagation networks and Euclidean loss for training
color propagation networks. Runtime computations were performed using a
Nvidia TitanX GPU and a 6 core Intel i7-5820K CPU clocked at 3.30GHz machine.
We will make available all the code and experimental results.

\subsection{Video Object Segmentation}
\label{sec:videoseg}

The task of class-agnostic video object segmentation aims to segment foreground objects in
videos. Since the semantics of the foreground object is not pre-defined, this problem is
usually addressed in a semi-supervised manner. The goal is to propagate a
given foreground mask of the first frame to the entire video frames.
Object segmentation in videos is useful for several high level tasks such
as video editing, summarization, rotoscoping etc.

\vspace{-0.5cm}
\paragraph{Dataset} We use the recently published DAVIS dataset~\cite{Perazzi2016}
for experiments on this task. The
DAVIS dataset consists of 50 high-quality (1080p resolution) unconstrained videos
with number of frames in each video ranging from 25 to 104. All the frames come with
high-quality per-pixel annotation of the foreground object. The videos for this
dataset are carefully chosen to contain motion blur, occlusions, view-point changes
and other occurrences of object segmentation challenges.
For robust evaluation and to get results on all the dataset videos,
we evaluate our technique using 5-fold cross-validation.
We randomly divided the data into
5 folds, where in each fold, we used 35 images for training, 5 for validation and
the remaining 10 for the testing. For the evaluation, we used the 3 metrics that
are proposed in~\cite{Perazzi2016}: Intersection over Union (IoU) score, Contour
accuracy ($\mathcal{F}$) score and temporal instability ($\mathcal{T}$) score. The widely
used IoU score is defined as $TP/(TP+FN+FP)$, where TP: True positives; FN: False negatives
and FP: False positives. Please refer to~\cite{Perazzi2016} for the definition of the contour
accuracy and temporal instability scores. We are aware of some other datasets for
this task such as JumpCut~\cite{fan2015jumpcut} and
SegTrack~\cite{tsai2012motion}, but we note that the number of videos in these datasets is too
small for a learning based approach.

\begin{table}[t]
    \centering
    \begin{tabular}{p{3.0cm}>{\centering\arraybackslash}p{1.2cm}>{\centering\arraybackslash}
      p{1.2cm}>{\centering\arraybackslash}p{1.2cm}>{\centering\arraybackslash}p{1.2cm}>{\centering\arraybackslash}p{1.2cm}
      >{\centering\arraybackslash}p{0.6cm}}
        \toprule
        \scriptsize
        & Fold-1 & Fold-2 & Fold-3 & Fold-4 & Fold-5 & All\\ [0.1cm]
        \midrule
        BNN-Identity & 56.4 & 74.0 & 66.1 & 72.2 & 66.5 & 67.0 \\
        VPN-Stage1 & 58.2 & 77.7 & 70.4 & 76.0 & 68.1 & 70.1 \\
        VPN-Stage2 & \textbf{60.9} & \textbf{78.7} & \textbf{71.4} & \textbf{76.8} & \textbf{69.0} & \textbf{71.3} \\

        \bottomrule
        \\
    \end{tabular}
    \mycaption{5-Fold Validation on DAVIS Video Segmentation Dataset}
    {Average IoU scores for different models on the 5 folds.}
    \label{tbl:davis-folds}
\end{table}

\begin{table}[t]
    \centering
    \begin{tabular}{p{3.0cm}>{\centering\arraybackslash}p{1.2cm}>{\centering\arraybackslash}
      p{1.2cm}>{\centering\arraybackslash}p{1.2cm}>{\centering\arraybackslash}p{2.3cm}}
      \toprule
      \scriptsize
      & \textit{IoU$\uparrow$} & $\mathcal{F}\uparrow$ & $\mathcal{T}\downarrow$ & \textit{Runtime}(s) \\ [0.1cm]
      \midrule
      BNN-Identity & 67.0 & 67.1 & 36.3 & 0.21\\
      VPN-Stage1 & 70.1 & 68.4 & 30.1 & 0.48\\
      VPN-Stage2 & 71.3 & 68.9 & 30.2 & 0.75\\
      \midrule
      \multicolumn{4}{l}{\emph{With pre-trained models}} & \\
      DeepLab & 57.0 & 49.9 & 47.8 & 0.15 \\
      VPN-DeepLab & \textbf{75.0} & \textbf{72.4} & 29.5 & 0.63 \\
      \midrule
      OFL~\cite{tsaivideo} & 71.1 & 67.9 & 22.1 & $>$60\\
      BVS~\cite{marki2016bilateral} & 66.5 & 65.6 & 31.6 &  0.37\\
      NLC~\cite{faktor2014video} & 64.1 & 59.3 & 35.6 & 20\\
      FCP~\cite{perazzi2015fully} & 63.1 & 54.6 & 28.5 & 12\\
      JMP~\cite{fan2015jumpcut} & 60.7 & 58.6 & \textbf{13.2} & 12\\
      HVS~\cite{grundmann2010efficient} & 59.6 & 57.6 & 29.7 & 5\\
      SEA~\cite{ramakanth2014seamseg} & 55.6 & 53.3 & 13.7 & 6\\
      \bottomrule
        \\
    \end{tabular}
    \mycaption{Results of Video Object Segmentation on DAVIS dataset}
    {Average IoU score, contour accuracy ($\mathcal{F}$),
    temporal instability ($\mathcal{T}$) scores, and average runtimes (in seconds)
    per frame for different VPN models along with recent published
    techniques for this task. VPN runtimes also include superpixel computation (10ms).
    Runtimes of other methods are taken from~\cite{marki2016bilateral,perazzi2015fully,tsaivideo}
    and only indicative and are not directly comparable to our runtimes. Runtime of VPN-Stage1 includes
    the runtime of BNN-Identity which is in-turn included in the runtime of VPN-Stage2. Runtime
    of VPN-DeepLab model includes the runtime of DeepLab.}
    \label{tbl:davis-main}
    \vspace{-0.7cm}
\end{table}

\begin{figure}[t!]
\begin{center}
  \centerline{\includegraphics[width=0.5\columnwidth]{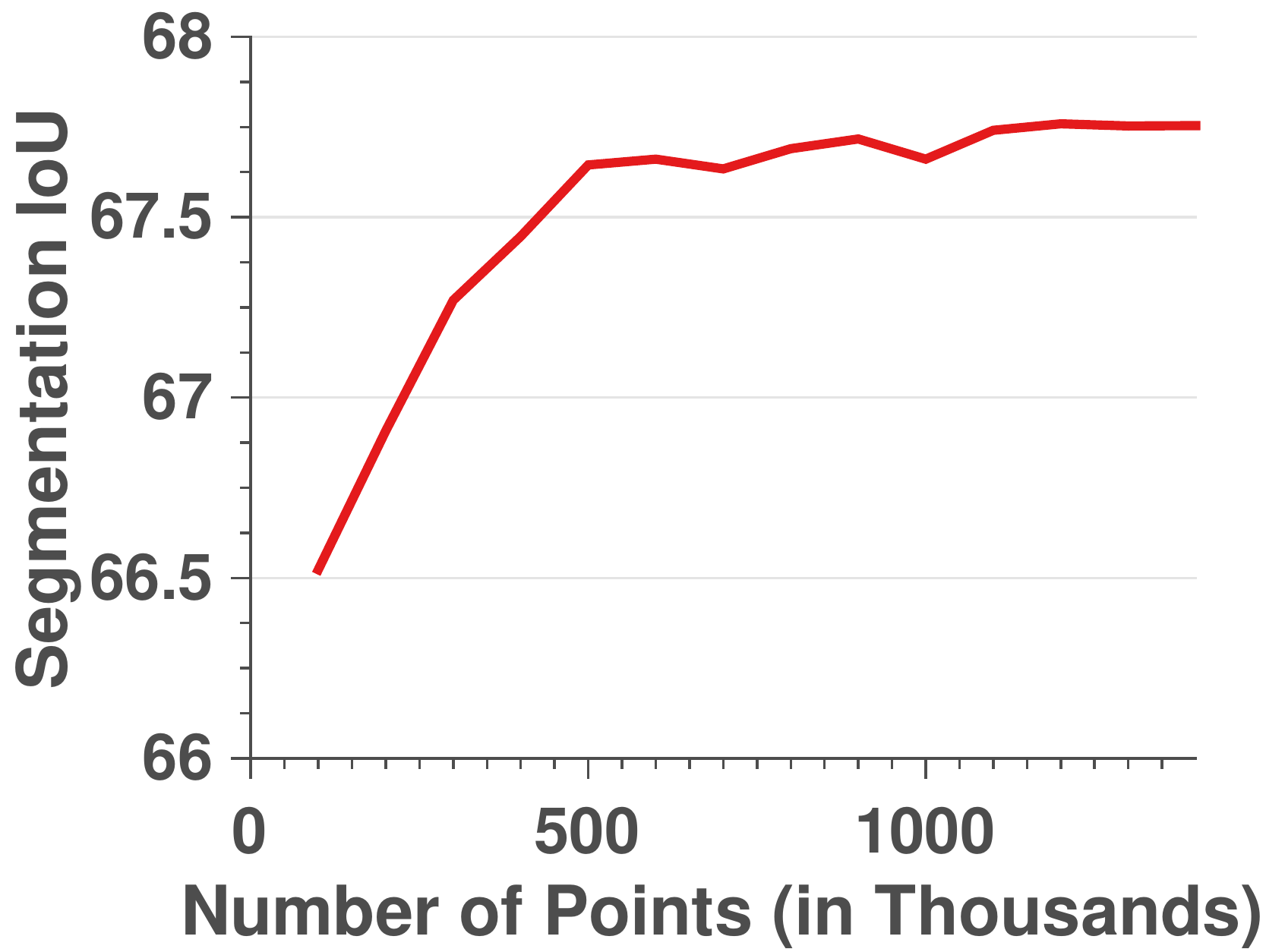}}
    \mycaption{Random Sampling of Input Points vs. IoU}
    {The effect of randomly sampling points from input video frames on object
    segmentation IoU of BNN-Identity on DAVIS dataset.
    The points sampled are out of $\approx$2 million points from the previous 5 frames.}
    \label{fig:acc_vs_points}
\end{center}
\vspace{-0.8cm}
\end{figure}

\begin{figure}[th!]
\begin{center}
  \centerline{\includegraphics[width=0.65\columnwidth]{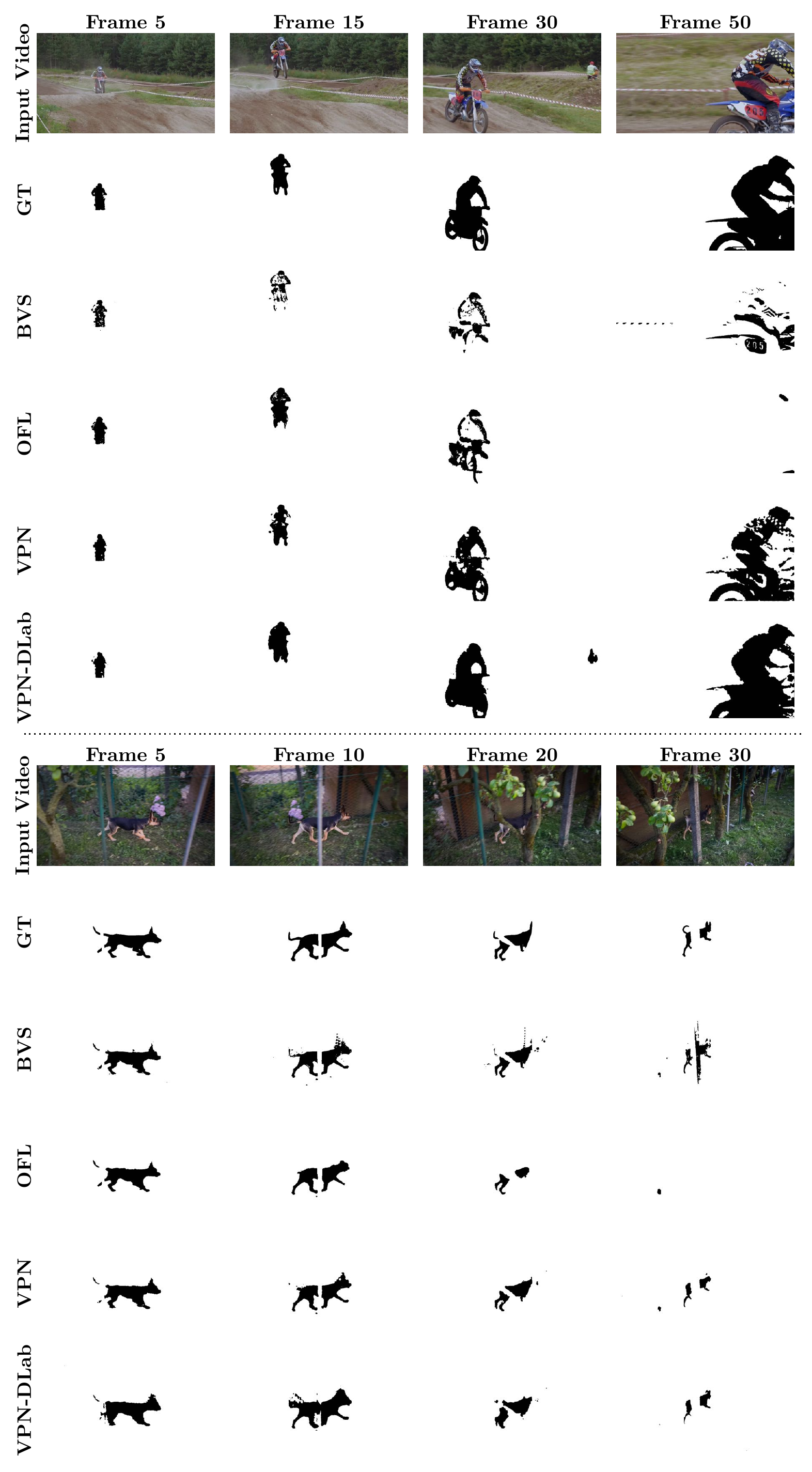}}
    \mycaption{Video Object Segmentation}
    {Shown are the different frames in example videos with the corresponding
    ground truth (GT) masks, predictions from BVS~\cite{marki2016bilateral},
    OFL~\cite{tsaivideo}, VPN (VPN-Stage2) and VPN-DLab (VPN-DeepLab) models.}
    \label{fig:video_seg_visuals}
\end{center}
\vspace{-1.0cm}
\end{figure}

\vspace{-0.5cm}
\paragraph{VPN and Results} In this task, we only have access to foregound mask
for the first frame $V_1$.
For the ease of training VPN, we obtain initial set of predictions with
\emph{BNN-Identity}. We sequentially apply \emph{BNN-Identity} at each frame
and obtain an initial set of foreground masks for the entire video.
These BNN-Identity propagated masks are then used as inputs to train a VPN to
predict the refined masks at each frame. We refer to this
VPN model as \emph{VPN-Stage1}. Once VPN-Stage1 is trained, its refined training
mask predictions are in-turn used as inputs to train another VPN model which we
refer to as \emph{VPN-Stage2}. This resulted in further refinement of foreground
masks. Training further stages did not result in any improvements.

Following the recent work of~\cite{marki2016bilateral} on video object segmentation,
we used scaled features $\f=(x,y,Y,Cb,Cr,t)$ with YCbCr color features for bilateral filtering.
To be comparable with the one of the fastest state-of-the-art technique~\cite{marki2016bilateral},
we do not use any optical flow information. First, we analyze the performance of BNN-Identity by changing the number of randomly sampled input points. Figure~\ref{fig:acc_vs_points} shows how the segmentation IoU changes with the increase
in the number of sampled points (out of 2 million points) from the previous frames.
The IoU levels out after sampling 25\% of points. For
further computational efficiency, we used superpixel sampling instead of random
sampling. Usage of superpixels reduced the IoU slightly (0.5\%), while reducing the
number of input points by a factor of 10 in comparison to a large number
of randomly sampled points. We used 12000 SLIC~\cite{achanta2012slic} superpixels from each frame
computed using the fast GPU implementation from~\cite{gSLICr_2015}. For predictions
at each frame, we input mask probabilities of previous 9 frames into VPN as we observe
no significant improvements with more frames. We set $\alpha$ to $0.5$ and the
feature scales for bilateral filtering are presented in Tab.~\ref{tbl:parameters_supp}.

Table~\ref{tbl:davis-folds} shows the IoU scores for each of the 5 folds and
Tab.~\ref{tbl:davis-main} shows the overall scores and runtimes of different VPN
models along with the best performing segmentation techniques.
The performance improved consistently across all 5 folds with the addition of new VPN stages.~BNN-Identity already performed reasonably well.
And with 1-stage and 2-stage VPNs, we outperformed the present fastest
BVS method~\cite{marki2016bilateral} by a significant margin on all
the performance measures of IoU, contour accuracy and temporal instability scores,
while being comparable in runtime. We perform marginally better than OFL method~\cite{tsaivideo}
while being at least 80$\times$ faster and OFL relies on optical flow whereas we
obtain similar performance without using any optical flow.
~Further, VPN has the advantage of doing online processing
as it looks only at previous frames whereas BVS processes entire video at once.
One can obtain better VPN performance with using better superpixels and
also incorporating optical flow, but this increases runtime as well.
Figure~\ref{fig:video_seg_visuals} shows some qualitative results and more are present
in Figs.~\ref{fig:video_seg_pos_supp}. A couple of
failure cases are shown in Fig.~\ref{fig:video_seg_neg_supp}. Visual results indicate that learned VPN is able to retain foreground masks even with large variations in viewpoint and object size.

\paragraph{Augmenation of Pre-trained Models:} One of the main advantages of the proposed
VPN architecture is that it is end-to-end trainable and can be easily integrated into
other deep neural network architectures. To demonstrate this, we augmented VPN architecture
with standard DeepLab segmentation architecture from~\cite{chen2014semantic}.
We replaced the last classification layer of DeepLab-LargeFOV model
from~\cite{chen2014semantic} to output 2 classes (foreground and background)
in our case and bi-linearly up-sampled the resulting low-resolution probability map to
the original image dimension. 5-fold fine-tuning of the DeepLab model on DAVIS dataset
resulted in the IoU of 57.0 and other scores are shown in Tab.~\ref{tbl:davis-main}.
Then, we combine the VPN and DeepLab models in the following way: The output from
the DeepLab network and the bilateral network are concatenated and then passed on to the spatial network.
In other words, the bilateral network propagates label information from previous frames to the present
frame, whereas the DeepLab network does the prediction for the present frame. The results
of both are then combined and refined by the spatial network in the VPN architecture.
We call this `VPN-DeepLab' model. We trained this model end-to-end and observed big
improvements in performance. As shown in Tab.~\ref{tbl:davis-main}, the VPN-DeepLab
model has the IoU score of 75.0 and contour accuracy score of 72.4 resulting
in significant improvements over the published results. Since DeepLab has also fast
runtime, the total runtime of VPN-DeepLab is only 0.63s which makes this also one of
the fastest video segmentation systems. A couple of visual results of
VPN-DeepLab model are shown in Fig.~\ref{fig:video_seg_visuals}
and more are present in
Figs.~\ref{fig:video_seg_pos_supp} and~\ref{fig:video_seg_neg_supp}.

\vspace{-0.2cm}
\subsection{Semantic Video Segmentation}

A semantic video segmentation assigns a semantic label to every video pixel.
Since the semantics between adjacent frames does not change
radically, intuitively, propagating semantic information across frames should improve
the segmentation quality of each frame. Unlike mask propagation in the previous
section where the ground-truth mask for the first frame is given, we approach
semantic video segmentation in a fully automatic fashion. Specifically, we start
with the unary predictions of standard CNNs and use VPN for propagating semantics across the frames.

\begin{table}[t]
    \centering
    \begin{tabular}{p{5.0cm}>{\centering\arraybackslash}p{2.4cm}>{\centering\arraybackslash}p{3.5cm}}
        \toprule
        \scriptsize
        & \textit{IoU} & \textit{Runtime}(s) \\ [0.1cm]
        \midrule
        CNN from ~\cite{yu2015multi} & 65.3 & 0.38\\
        + FSO-CRF~\cite{kundu2016feature} & 66.1 & \textbf{$>$}10\\
        + BNN-Identity  & 65.3 & 0.31\\
        + BNN-Identity-Flow  & 65.5 & 0.33\\
        + VPN (Ours) & 66.5 & 0.35\\
        + VPN-Flow (Ours) & \textbf{66.7} & 0.37\\
        \midrule
        CNN from ~\cite{richter2016playing} & 68.9 & 0.30\\
        + VPN-Flow (Ours) & \textbf{69.5} & 0.38\\
        \bottomrule
        \\
    \end{tabular}
    \mycaption{Results of Semantic Segmentation on the CamVid Dataset}{
    Average IoU and runtimes (in seconds)
    per frame of different models on \textit{test} split.
    Runtimes exclude CNN computations which are shown separately.
    VPN and BNN-Identity runtimes include superpixel computation which
    takes up large portion of computation time (0.23s).}
    \label{tbl:camvid}
    \vspace{-0.5cm}
\end{table}

\vspace{-0.4cm}
\paragraph{Dataset} We use the CamVid dataset~\cite{brostow2009semantic} that contains 4 high
quality videos captured at 30Hz while the semantically labelled 11-class ground truth is
provided at 1Hz. While the original dataset comes at a resolution of 960$\times$720, similar to
previous works~\cite{yu2015multi,kundu2016feature}, we operate on a resolution of 640$\times$480.
We use the same splits proposed in~\cite{sturgess2009combining} resulting in
367, 100 and 233 frames with ground-truth for training, validation and testing.
Following common practice, we report the IoU scores for evaluation.

\begin{figure}[th!]
\begin{center}
  \centerline{\includegraphics[width=0.9\columnwidth]{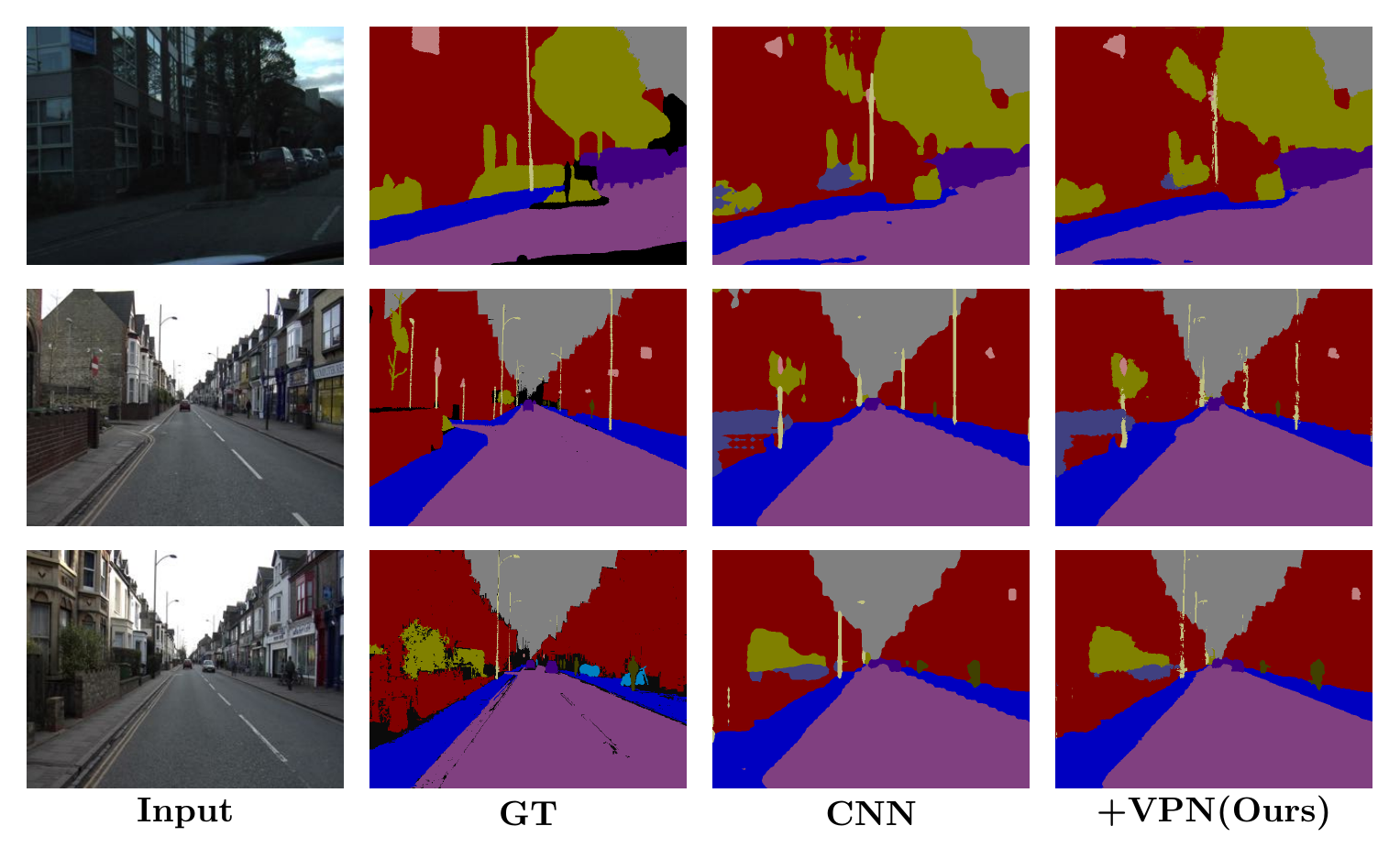}}
    \mycaption{Semantic Video Segmentation}
    {Input video frames and the corresponding ground truth (GT)
    segmentation together with the predictions of CNN~\cite{yu2015multi} and with
    VPN-Flow.}
    \label{fig:semantic_visuals}
\end{center}
\vspace{-0.7cm}
\end{figure}

\vspace{-0.5cm}
\paragraph{VPN and Results} Since we already have CNN predictions for every
frame, we train a VPN that takes the CNN predictions of previous \emph{and} present
frames as input and predicts the refined predictions for the present frame.
We compare with the state-of-the-art CRF approach for this problem~\cite{kundu2016feature}
which we refer to as `FSO-CRF'. Following~\cite{kundu2016feature}, we also experimented with
optical flow in our framework and refer that model as \emph{VPN-Flow}.
We used the fast optical flow method that uses dense inverse search
~\cite{kroeger2016fast} to compute flows and modify the positional features of previous frames.
We used the superpixels method of Dollar et al.~\cite{DollarICCV13edges} for this dataset as
gSLICr~\cite{gSLICr_2015} has introduced artifacts.

We experimented with predictions from two different CNNs:
One is with dilated convolutions~\cite{yu2015multi} (CNN-1) and another one~\cite{richter2016playing} (CNN-2)
is trained with the additional data obtained from a video game,
which is the present state-of-the-art on this dataset.
For CNN-1 and CNN-2, using 2 and 3 previous frames respectively as input
to VPN is found to be optimal. Other parameters of the bilateral network are presented
in Tab.~\ref{tbl:parameters_supp}. Table~\ref{tbl:camvid} shows quantitative results on this dataset.
Using BNN-Identity only slightly improved the CNN performance whereas training the
entire VPN significantly improved the CNN performance by over 1.2\% IoU, with both
VPN and VPN-Flow networks. Moreover, VPN is at least 25$\times$ faster, and simpler to use
compared to the optimization based FSO-CRF which relies on
LDOF optical flow~\cite{brox2009large}, long-term tacks~\cite{sundaram2010dense} and
edges~\cite{dollar2015fast}.
We further improved the performance of the state-of-the-art CNN~\cite{richter2016playing}
with the use of VPN-Flow model. Using better optical flow estimation
might give even better results. Figure~\ref{fig:semantic_visuals} shows some qualitative
results and more are presented in Fig.~\ref{fig:semantic_visuals_supp}.

\subsection{Video Color Propagation}

We also evaluate VPNs on a different kind of information and
experimented with propagating color information in a grayscale video. Given the
color image for the first video frame, the task is to propagate the color to the entire
video. Note that this task is fundamentally different from automatic colorization of images
for which recent CNN based based methods have become popular.
For experiments on this task, we again used the DAVIS dataset~\cite{Perazzi2016} with the
first 25 frames from each video. We randomly divided the dataset into 30 train,
5 validation and 15 test videos.

\begin{table}[t]
    \centering
    \begin{tabular}{p{4.0cm}>{\centering\arraybackslash}p{2.6cm}>{\centering\arraybackslash}p{3.5cm}}
        \toprule
        \scriptsize
        & \textit{PSNR} & \textit{Runtime}(s) \\ [0.1cm]
        \midrule
        BNN-Identity & 27.89 & 0.29\\
        VPN-Stage1 & \textbf{28.15} & 0.90\\
        \midrule
        Levin et al.~\cite{levin2004colorization} & 27.11 & 19\\
        \bottomrule
        \\
    \end{tabular}
    \mycaption{Results of Video Color Propagation}{Average PSNR results and runtimes of
    different methods for video color propagation on images from DAVIS dataset.}
    \label{tbl:color}
    \vspace{-0.5cm}
\end{table}

\begin{figure}[th!]
\begin{center}
  \centerline{\includegraphics[width=0.9\columnwidth]{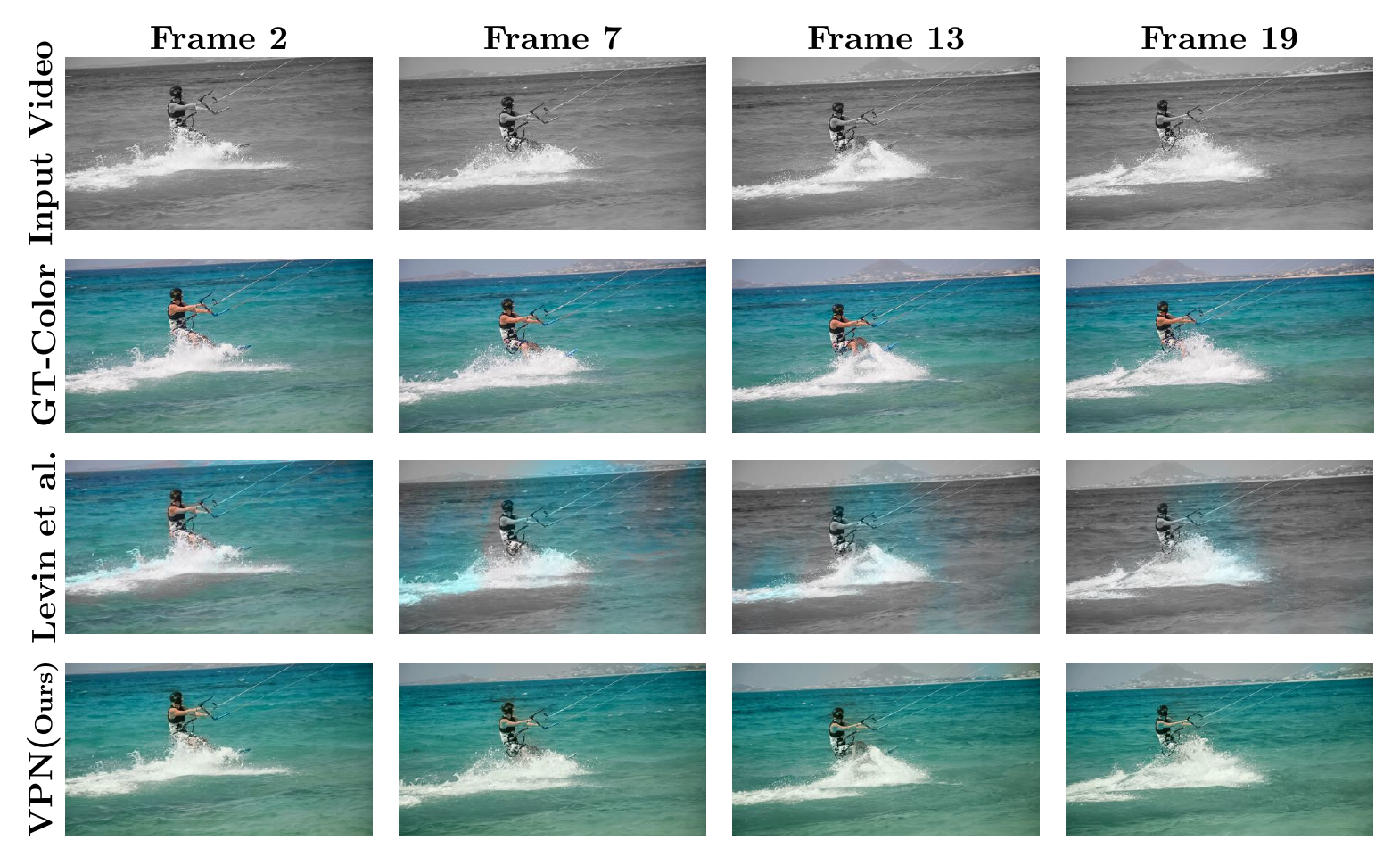}}
    \mycaption{Video Color Propagation}
    {Input grayscale video frames and corresponding ground-truth (GT) color images
    together with color predictions of Levin et al.~\cite{levin2004colorization} and VPN-Stage1 models.}
    \label{fig:color_visuals}
\end{center}
\vspace{-1.0cm}
\end{figure}

We work with YCbCr representation of images and propagate CbCr values from previous
frames with pixel intensity, position and time features as guidance for VPN.
The same strategy as in object segmentation is used, where an initial
set of color propagated results was obtained with BNN-Identity and then used to trained a VPN-Stage1 model.
Training further VPN stages did not improve the performance.
Table~\ref{tbl:color} shows the PSNR results.
We use 300K radomly sampled points from previous 3 frames as input
to the VPN network. We also show a baseline result of~\cite{levin2004colorization} that
does graph based optimization and uses optical flow. We used fast
DIS optical flow~\cite{kroeger2016fast} in the baseline method~\cite{levin2004colorization}
and we did not observe significant differences with using LDOF optical flow~\cite{brox2009large}.
Figure~\ref{fig:color_visuals} shows a visual result with more
in Fig.~\ref{fig:color_visuals_supp}.
From the results,
VPN works reliably better than~\cite{levin2004colorization} while being 20$\times$ faster.
The method of~\cite{levin2004colorization} relies heavily on optical flow
and so the color drifts away with incorrect flow. We observe that our method also bleeds color
in some regions especially when there are large viewpoint changes.
We could not compare against recent video color propagation techniques such as
~\cite{heu2009image,sheng2014video} as their codes are not available online.
This application shows general applicability of VPNs in propagating different
kinds of information.

\vspace{-0.3cm}
\section{Discussion and Conclusions}
\label{sec:conclusion}

We proposed a fast, scalable and generic neural network based learning approach
for propagating information across video frames.~The video propagation network uses
bilateral network for long-range video-adaptive propagation of information from previous
frames to the present frame which is then refined by a standard spatial network.
Experiments on diverse tasks show that VPNs, despite being generic, outperformed
the current state-of-the-art task-specific methods. At the core of our technique
is the exploitation and modification of learnable bilateral filtering for the use
in video processing. We used a simple and fixed network architecture for all the
tasks for showcasing the generality of the approach. Depending on the type of
problems and the availability of data, using more filters and deeper layers
would result in better performance. In this work, we manually tuned the feature scales which
could be amendable to learning. Finding optimal yet fast-to-compute bilateral features for
videos together with the learning of their scales is an important future
research direction.

\chapter{Bilateral Inception Networks}
\label{chap:binception}

\newcommand{\bff}{\mathbf{f}}
\newcommand{\bz}{\mathbf{z}}
\newcommand{\mF}{\mathcal{F}}
\newcommand{\mbF}{\mathcal{\mathbf{F}}}

\newcommand{\deconv}{DeconvNet}
\newcommand{\deconvcrf}{DeconvNet-CRF}
\newcommand{\mincnet}{AlexNet~}
\newcommand{\mincnetcrf}{AlexNet-CRF~}

\newcommand{\deeplablargefov}{DeepLab}
\newcommand{\deeplablargefovcrf}{DeepLab-CRF}
\newcommand{\deeplabmsclargefovcrf}{DeepLab-MSc-CRF}

\newcommand{\deeplab}{DeepLab}
\newcommand{\deeplabcrf}{DeepLab-CRF}
\newcommand{\deeplabstrong}{DeepLab-COCO-Strong}
\newcommand{\deeplabstrongcrf}{DeepLab-COCO-Strong-CRF}
\newcommand{\deeplabmsclargefov}{DeepLab-LargeFOV}

\newcommand{\bi}[2]{BI$_{#1}(#2)$}
\newcommand{\Bi}[1]{BI$_{#1}$}
\newcommand{\gi}[1]{G$(#1)$}
\newcommand{\fc}[1]{FC$_{#1}$}

\hypersetup{
  linkcolor  = black,
  citecolor = black,
  urlcolor   = green!20!black,
  colorlinks = true,
}

\DeclareRobustCommand\onedot{\futurelet\@let@token\@onedot}
\def\onedot{\ifx\let@token.\else.\null\fi\xspace}
\def\eg{\emph{e.g}\onedot} \def\Eg{\emph{E.g}\onedot}
\def\ie{\emph{i.e}\onedot} \def\Ie{\emph{I.e}\onedot}
\def\cf{\emph{c.f}\onedot} \def\Cf{\emph{C.f}\onedot}
\def\wrt{w.r.t\onedot} \def\dof{d.o.f\onedot}

\newcommand{\fix}{\marginpar{FIX}}
\newcommand{\new}{\marginpar{NEW}}

Following up on previous chapters, where we introduced learnable bilateral
filters and their application to a wide range of problems, in this chapter,
we construct a CNN module which we call `Bilateral Inception' that can be
inserted into \emph{existing} CNN architectures for better inference in
pixel prediction tasks.
The bilateral inception module performs bilateral
filtering, at multiple feature-scales, between superpixels in an image.
The feature spaces for bilateral
filtering and other parameters of the module are learned end-to-end using
standard back-propagation techniques.
Instead of using learnable bilateral filtering proposed in Chapter~\ref{chap:bnn},
here, we explicitly construct the Gaussian filter kernel between input
and output superpixels. We show how this explicit Gaussian filtering
results in fast runtimes and also enables the learning of bilateral features.

We focus on the problem of semantic segmentation.
The bilateral inception module addresses two issues that
arise with general CNN segmentation architectures. First, this module propagates
information between (super)pixels while respecting image edges, thus
using the structured information of the problem for improved results.
Second, the layer recovers a full resolution segmentation result from the
lower resolution solution of a CNN.

In the experiments, we modify several existing CNN architectures by inserting
our inception module between the last CNN ($1\times1$ convolution) layers. Empirical results
on three different datasets show reliable improvements not only in comparison
to the baseline networks, but also in comparison to several dense-pixel prediction
techniques such as CRFs, while being competitive in time.

\section{Introduction}

In this work, we propose a CNN architecture for semantic image segmentation.
Given an image $\obs=(x_1,\ldots,x_N$) with $N$ pixels $x_i$, the task of semantic segmentation is to infer a labeling $\target=(y_1,\ldots,y_N)$ with a label $y_i\in\mathcal{Y}$ for every pixel.
This problem can be naturally formulated as a structured prediction problem $g:\obs\rightarrow \target$.
Empirical performance is measured by comparing $\target$ to a human labeled $\target^*$ via a
loss function $\Delta(\target,\target^*)$, \eg with the Intersection over Union (IoU) or pixel-wise
Hamming Loss.

A direct way to approach this problem would be to ignore the structure of the
output variable $\target$ and train a classifier that predicts the class membership of the center
pixel of a given image patch. This procedure reduces the problem to a standard multi-class
classification problem and allows the use of standard learning algorithms. The resulting
classifier is then evaluated at every possible patch in a sliding window fashion (or using
coarse-to-fine strategies) to yield a full segmentation of the image.
With high capacity models and large amounts of training data, this approach would
be sufficient, given that the loss decomposes over the pixels.
Such a per-pixel approach ignores the relationship between the variables $(y_1,\ldots,y_N)$, which are not i.i.d.~since there is an underlying common image. Therefore, besides learning
discriminative per-pixel classifiers, most segmentation approaches further encode the output
relationship of $\target$. A dominating approach is to use Conditional Random Fields
(CRF)~\cite{lafferty2001crf}, which allows an elegant and principled way
to combine single pixel predictions and shared structure through unary, pairwise and
higher order factors.

\begin{figure}[t]
\begin{center}
\centerline{\includegraphics[width=\textwidth]{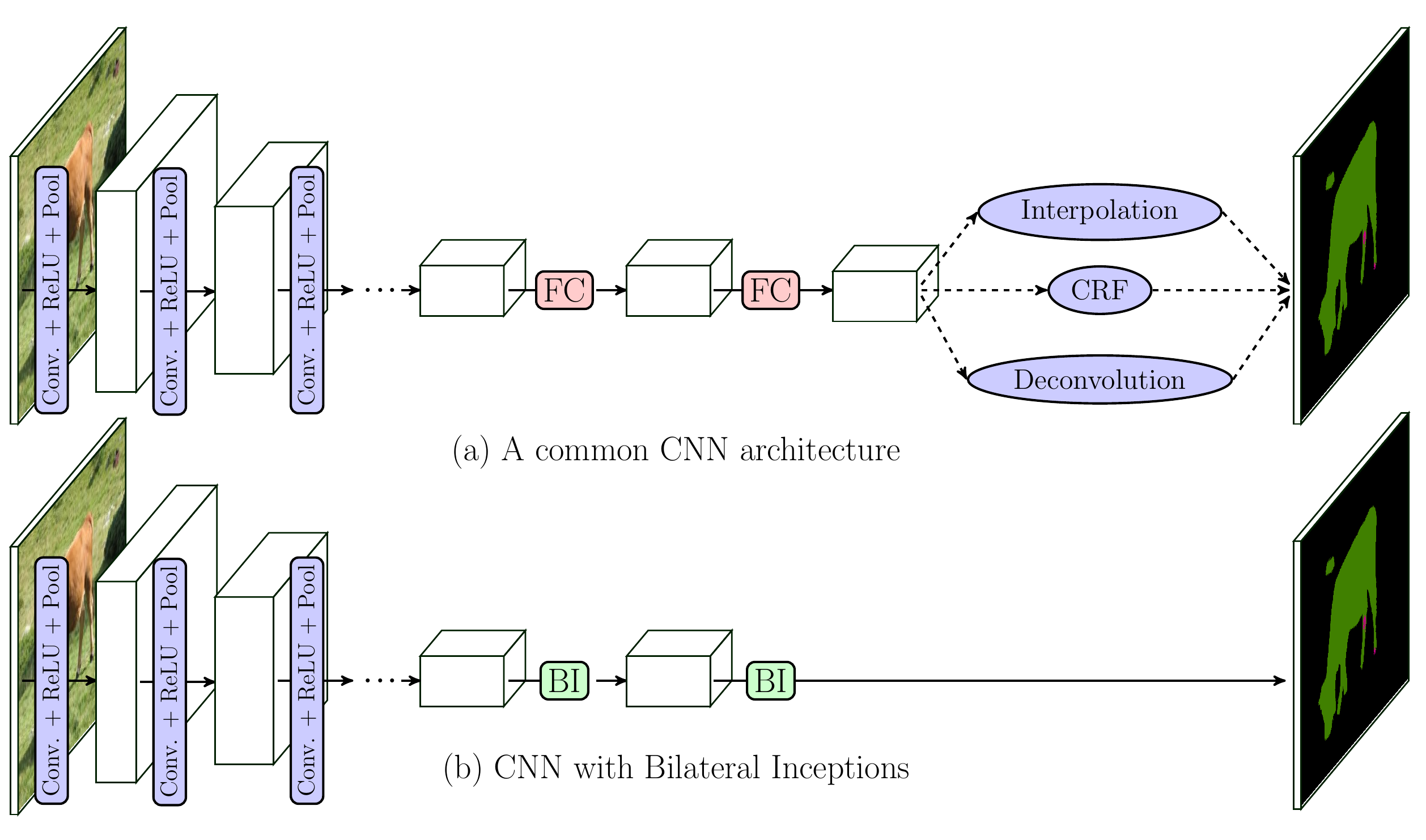}}
  \mycaption{Illustration of CNN layout} {We insert the \emph{Bilateral Inception (BI)} modules
  between the \emph{FC} ($1\times1$ convolution) layers found in
  most networks thus removing the necessity
  of further up-scaling algorithms. Bilateral Inception modules also propagate information
  between distant pixels based on their spatial and color similarity and work
  better than other label propagation approaches.}\label{fig:illustration}
  \vspace{-0.3cm}
\end{center}
\end{figure}

What relates the outputs $(y_1,\ldots,y_N$)? The common hypothesis that
we use in this chapter could be summarized as:
\emph{Pixels that are spatially and photometrically similar are more likely to have the same label.}
Particularly if two pixels $x_i,x_j$ are close in the image and have similar
$RGB$ values, then their corresponding labels $y_i,y_j$ will most likely be the same.
The most prominent example of spatial similarity encoded in a CRF is the Potts model (Ising model for
the binary case).
The work of~\cite{krahenbuhl2012efficient} described a densely connected pairwise CRF (DenseCRF) that includes
pairwise factors encoding both spatial \emph{and} photometric similarity.
The DenseCRF has been used in many recent works on image segmentation
which find also empirically improved results over pure pixel-wise CNN
classifiers~\cite{chen2014semantic,bell2015minc,zheng2015conditional,chen2015semantic}.

In this chapter, we implement the above-mentioned hypothesis of nearby pixels which are photometrically
similar sharing a common label, by designing a new
`Bilateral Inception' (BI) module that can be inserted before/after the last $1\times1$ convolution
layers (which we refer to as `FC' layers - `Fully-Connected' in the original image classification network) of the standard segmentation CNN architectures. The bilateral inception module does edge-aware information propagation across different spatial CNN units of the previous FC layer. Instead of using the spatial grid-layout that is common in CNNs,
we incorporate the superpixel-layout for information propagation. The information
propagation is performed using standard bilateral filters with Gaussian kernels, at
different feature scales. This construction is inspired by~\cite{szegedy2014googlenet,lin2014network}. Feature spaces and other parameters of
the modules can be learned end-to-end using standard back-propagation techniques.
The application of superpixels reduces the number of necessary computations and implements a long-range edge-aware inference between different superpixels. Moreover, since superpixels
provides an output at the full image resolution, it removes the need
for any additional post-processing step.

We introduce BI modules in the CNN segmentation models
of~\cite{chen2014semantic,zheng2015conditional,bell2015minc}.
See Fig.~\ref{fig:illustration} for an illustration. This
achieves better segmentation results than the proposed
interpolation/inference techniques of DenseCRF~\cite{bell2015minc,chen2014semantic},
on all three datasets that we experimented with,
while being faster. Moreover, the results compare favorably against some recently
proposed dense pixel prediction techniques.
As illustrated in Fig.~\ref{fig:illustration}, the BI modules
provide an alternative approach to commonly used up-sampling and CRF
techniques.


\vspace{-0.3cm}
\section{Related Work}\label{sec:related}

The literature on semantic segmentation is large and therefore we
will limit our discussion to those works that perform segmentation with
CNNs and discuss the different ways to encode the output structure.

A natural combination of CNNs and CRFs is to use the CNN as unary potential
and combine it with a CRF that also includes pairwise or higher order factors.
For instance~\cite{chen2014semantic,bell2015minc}
observed large improvements in pixel accuracy when combining a DenseCRF~\cite{krahenbuhl2012efficient}
with a CNN. The mean-field steps of the DenseCRF can
be learned and back-propagated as noted by~\cite{domke2013learning} and implemented
by~\cite{zheng2015conditional,arxivpaper,li2014mean,schwing2015fully} for semantic segmentation and~\cite{kiefel2014human} for human pose estimation.
The works of~\cite{chen2014learning,lin2015efficient,liu2015semantic}
use CNNs also in pairwise and higher order factors for more expressiveness.
The recent work of~\cite{chen2015semantic} replaced the costly
DenseCRF with a faster domain transform performing smoothing filtering while
predicting the image edge maps at the same time.
Our work was inspired by DenseCRF approaches but with the aim to replace the
expensive mean-field inference. Instead of propagating information across unaries
obtained by a CNN, we aim to do the edge-aware information propagation across
\textit{intermediate} representations of the CNN. Experiments on different datasets indicate that
the proposed approach generally gives better results in comparison to DenseCRF
while being faster.

A second group of works aims to inject the structural knowledge in intermediate
CNN representations by using structural layers among CNN internal layers.
The deconvolution layers model from~\cite{zeiler2010deconvolutional}
are being widely used for local propagation of information.
They are computationally efficient and are used in segmentation networks, \eg~\cite{long2014fully}.
They are however limited to small receptive fields. Another
architecture proposed in~\cite{he2014spatial} uses spatial pyramid pooling layers to max-pool
over different spatial scales. The work of~\cite{ionescu2015matrix} proposed specialized
structural layers such as normalized-cut layers with matrix back-propagation techniques.
All these works have either fixed local receptive fields and/or have their complexity increasing exponentially with longer range pixel connections.
Our technique allows for modeling long range (super)pixel dependencies without compromising the computational efficiency.
A very recent work~\cite{yu2015multi} proposed the use of dilated convolutions
for propagating multi-scale contextual information among CNN units.

A contribution of this work is to define convolutions over superpixels by defining
connectivity among them. In~\cite{he2015supercnn}, a method to use superpixels
inside CNNs has been proposed by re-arranging superpixels based on their
features. The technique proposed here is more generic and alleviates the need
for rearranging superpixels. A method to filter irregularly sampled data has been developed
in~\cite{bruna2013spectral} which may be applicable to superpixel convolutions. The difference
being that their method requires a pre-defined graph structure for every example/image
separately while our approach directly works on superpixels.
We experimented with Isomap embeddings~\cite{tenenbaum2000global} of superpixels but for speed reasons
opted for the more efficient kernels presented in this chapter. The work of~\cite{mostajabi2014feedforward}
extracted multi-scale features at each superpixel and performs semantic segmentation
by classifying each superpixel independently. In contrast, we propagate
information across superpixels by using bilateral filters with learned feature
spaces.

Another core contribution of this work is the end-to-end trained bilateral filtering
module.
Several recent works on bilateral filtering~\cite{barron2015bilateral,barron2015defocus}
(including ours in Chapter~\ref{chap:bnn})
back-propagate through permutohedral lattice approximation~\cite{adams2010fast}, to either learn the
filter parameters (Chapter~\ref{chap:bnn}) or do optimization in the
bilateral space~\cite{barron2015bilateral,barron2015defocus}. Most of the existing works
on bilateral filtering use pre-defined feature spaces. In~\cite{campbell2013fully}, the feature
spaces for bilateral filtering are obtained via a non-parametric embedding into
an Euclidean space. In contrast, by explicitly computing the bilateral filter kernel,
we are able to back-propagate through features, thereby learning
the task-specific feature spaces for bilateral filters through integration into
end-to-end trainable CNNs.


\section{Bilateral Inception Networks}

We first formally introduce superpixels in Sec.~\ref{sec:superpixels} before we
describe the bilateral inception modules in Sec.~\ref{sec:inception}.

\subsection{Superpixels}\label{sec:superpixels}

The term \emph{superpixel} refers to a set of $n_i$ pixels $s_i=\{t_1,\ldots,t_{n_i}\}$ with
$t_k\in\{1,\ldots,N\}$ pixels. We use a set of $M$ superpixels $S=\{s_1,\ldots,s_M\}$
that are disjoint $s_i\cap s_j=\emptyset, \forall i,j$ and decompose the image,
$\cup_i s_i = \mathcal{I}$.

\begin{figure}[t]
  \centering
\includegraphics[width=0.5\textwidth]{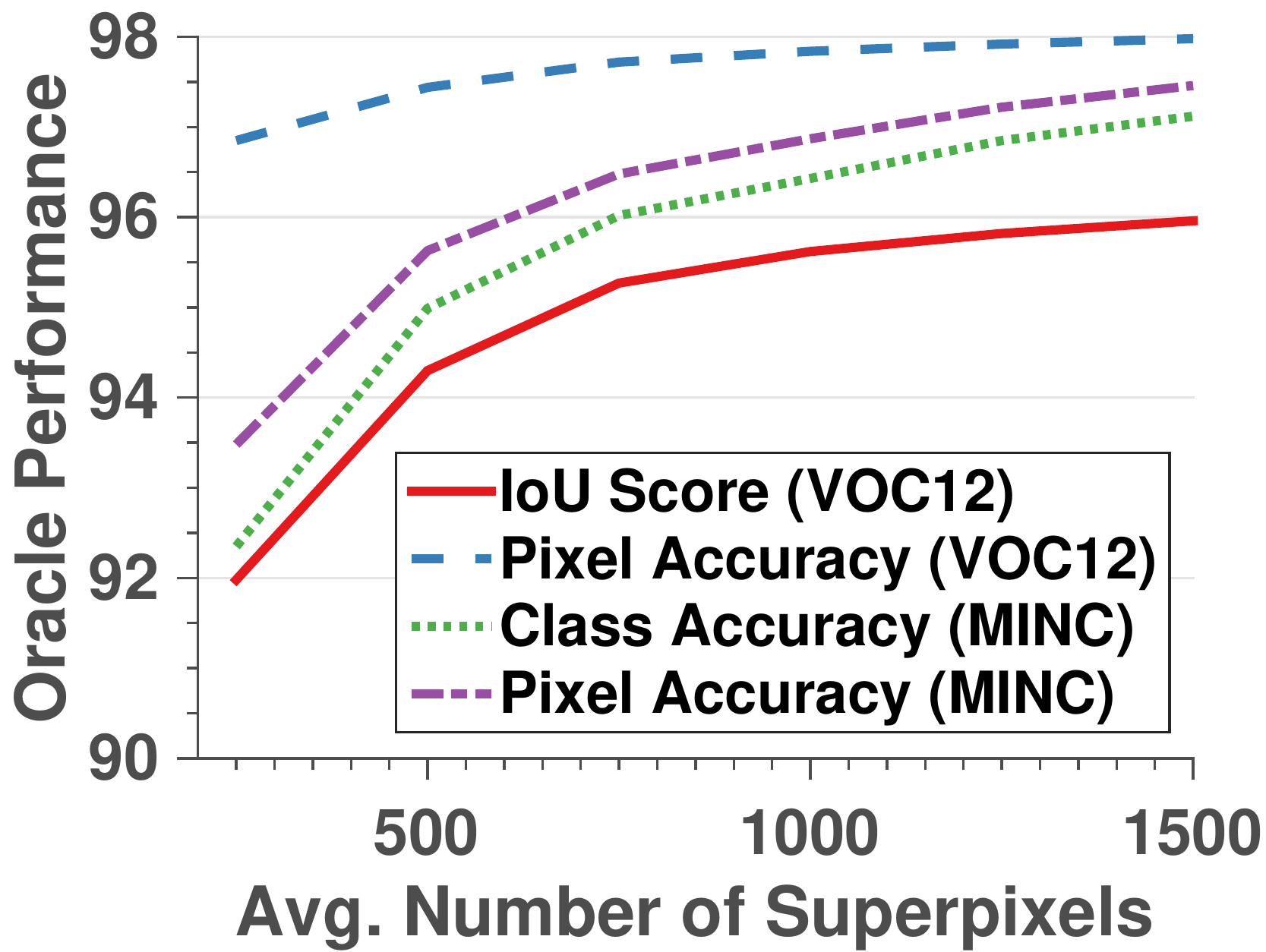}
  \mycaption{Superpixel Quantization Error} {Best achievable segmentation
    performance with a varying number
    of SLIC superpixels~\cite{achanta2012slic} on Pascal VOC12 segmentation~\cite{voc2012segmentation} and MINC material
    segmentation~\cite{bell2015minc} datasets.}
    \label{fig:quantization}
\end{figure}

Superpixels have long been used for image segmentation in many previous
works, \eg~\cite{Gould:ECCV2014,gonfaus2010harmony,nowozin2010parameter,mostajabi2014feedforward},
as they provide a reduction of the problem size.
Instead of predicting a label $y_i$ for every pixel $x_i$, the classifier predicts a label $y_i$
per superpixel $S_i$ and extends this label to all pixels within.
A superpixel algorithm can pre-group pixels based on spatial and photometric similarity,
reducing the number of elements and also thereby regularizing the problem in a
meaningful way. The downside is that superpixels introduce a quantization error whenever
pixels within one segment have different ground truth label assignments.

Figure~\ref{fig:quantization} shows the superpixel quantization effect with
the best achievable performance
as a function in the number of superpixels, on two different segmentation datasets:
PascalVOC~\cite{voc2012segmentation} and Materials in Context~\cite{bell2015minc}.
We find that the quantization effect is small compared to the current best segmentation
performance. Practically, we use the SLIC superpixels~\cite{achanta2012slic} for their runtime and~\cite{DollarICCV13edges} for their lower quantization error to decompose the image into superpixels. For details of the algorithms, please refer to the respective papers. We use the publicly-available
real-time GPU implementation of SLIC, called gSLICr~\cite{gSLICr_2015}, which
runs at over 250 frames per second. And the publicly available Dollar superpixels code~\cite{DollarICCV13edges} computes a super-pixelization for a $400\times 500$ image in about 300ms using an Intel Xeon 3.33GHz CPU.

\subsection{Bilateral Inceptions}\label{sec:inception}

Next, we describe the \emph{Bilateral Inception} (BI) module that performs Gaussian bilateral
filtering on multiple scales of the representations within a CNN.
The BI module can be inserted in between layers of existing CNN architectures.

\begin{figure}[t]
\begin{center}
\centerline{\includegraphics[width=\textwidth]{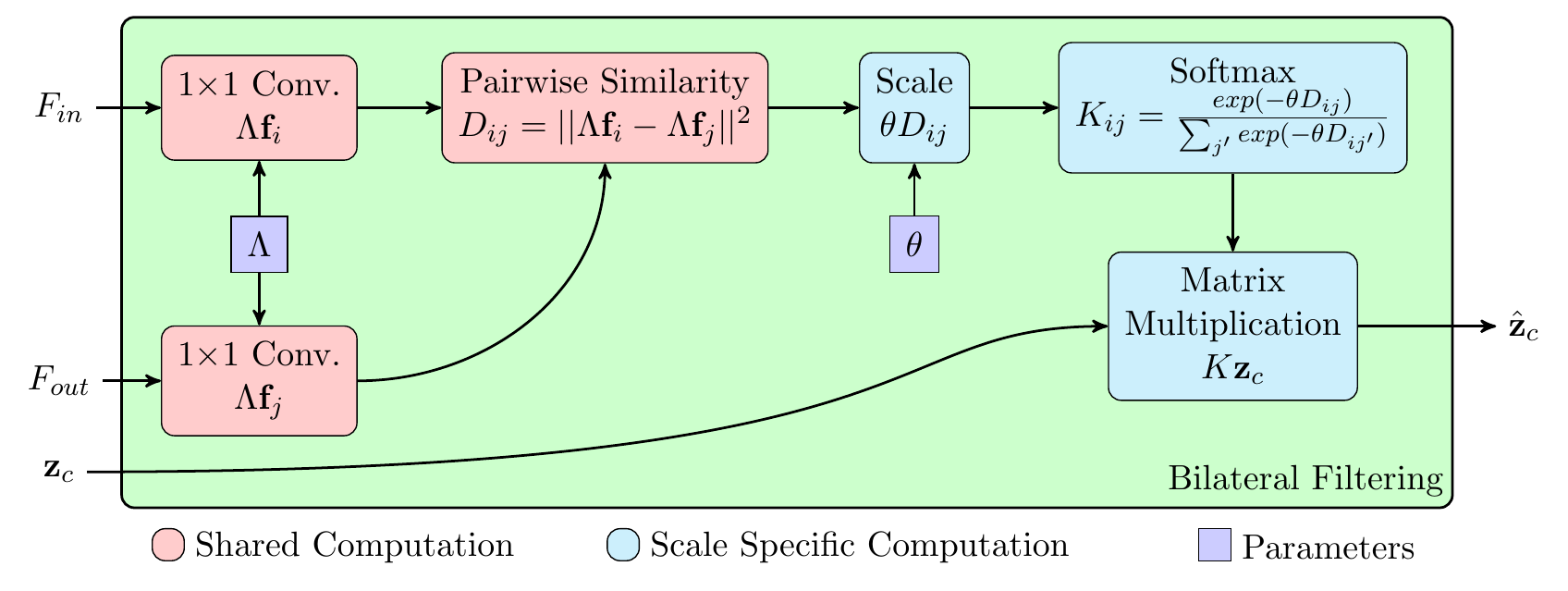}}
  \mycaption{Computation flow of the Gaussian bilateral filtering} {
We implemented the bilateral convolution with five separate computation blocks. $\Lambda$ and $\theta$
are the free parameters.}\label{fig:bi_module}
\end{center}
\end{figure}

{\bfseries Bilateral Filtering:} We first describe the Gaussian bilateral filtering,
the building block of the BI module.
A visualization of the necessary computations is shown in Fig.~\ref{fig:bi_module}.
Let the previous layer CNN activations be $\bz\in\mathbb{R}^{P\times C}$, that is $P$ points and $C$ filter responses.
With $\bz_c\in\mathbb{R}^P$ we denote the vector of activations of filter $c$.
Additionally we have for every point $j$ a feature vector $\bff_j\in\mathbb{R}^D$.
This denotes its spatial position ($D=2$, not necessarily a grid), position and RGB color ($D=5$), or others.
Separate from the input points with features $F_{in}=\{\bff_1,\ldots,\bff_P\}$, we have $Q$ output points with features $F_{out}$.
These can be the same set of points, but also fewer ($Q<P$), equal ($Q=P$), or more ($Q>P$) points.
For example we can filter a $10\times 10$ grid ($P=100$) and produce the result on a $50\times 50$ grid ($Q=2500$) or vice versa.

\begin{figure}[t]
\begin{center}
\centerline{\includegraphics[width=\textwidth]{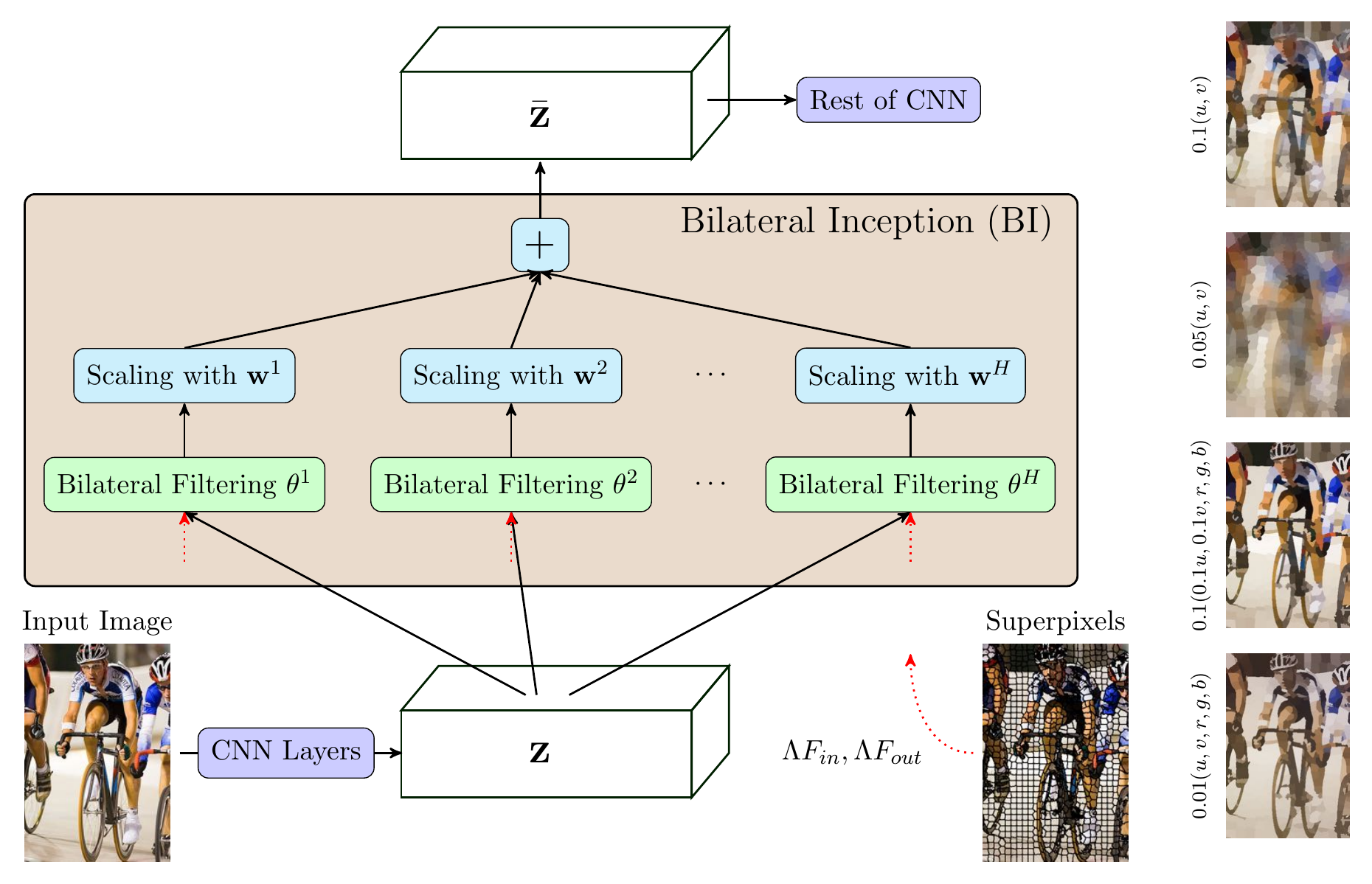}}
  \mycaption{Visualization of a Bilateral Inception (BI) Module} {The unit activations $\bz$
    are passed through several bilateral filters defined over different feature spaces. The result
  is linearly combined to $\bar{\bz}$ and passed on to the next network layer. Also shown are sample
  filtered superpixel images using bilateral filters defined over different example feature spaces. $(u,v)$ correspond to position and $(r,g,b)$ correspond to color features.}\label{fig:inception}
\end{center}
\end{figure}

The bilateral filtered result will be denoted as $\hat{\bz}\in\mathbb{R}^{Q\times C}$.
We apply the same Gaussian bilateral filter to every channel $c$ separately.
A filter has two free parameters: the filter specific scale $\theta\in\mathbb{R}_+$ and the
global feature transformation parameters $\Lambda\in\mathbb{R}^{D\times D}$.
For~$\Lambda$, a more general scaling could be applied using more features or a separate CNN.
Technically the bilateral filtering amounts to a matrix-vector multiplication for all $c$:

\begin{equation}
\hat{\bz}_c = K(\theta, \Lambda, F_{in}, F_{out}) \bz_c,
\label{eq:filter}
\end{equation}
where $K\in\mathbb{R}^{Q\times P}$ and values for $f_i\in F_{out}, f_j\in F_{in}$:
\begin{equation}
K_{i,j} = \frac{\exp(-\theta\|\Lambda \bff_i- \Lambda \bff_j\|^2)}{\sum_{j'}\exp(-\theta\|\Lambda \bff_i- \Lambda \bff_{j'}\|^2)}.
\label{eq:filter}
\end{equation}

From a kernel learning terminology, $K$ is nothing but a Gaussian Gram matrix and it is symmetric if $F_{in}=F_{out}$.
We implemented this filtering in Caffe~\cite{jia2014caffe} using different layers as depicted in Fig.~\ref{fig:bi_module}.
While approximate computations of $K\bz_c$ exist and have improved
runtime~\cite{adams2010fast,paris2006fast,gastal2011domain,adams2009gaussian}, we chose an explicit and exact computation of $K$ due to its small size.
Our implementation makes use of the GPU and the intermediate pairwise similarity computations are re-used across different
modules. The entire runtime is only a fraction of the CNN runtime, but of course applications to larger values of $P$ and $Q$
would require aforementioned algorithmic speed-ups.

{\bfseries Bilateral Inception Module:}
The \textit{bilateral inception module} (BI) is a weighted combination of different bilateral filters.
We combine the output of $H$ different filter kernels~$K$, with different scales $\theta^1,\ldots,\theta^H$.
All kernels use the same feature transformation~$\Lambda$ which allows for easier pre-computation of pairwise difference and avoids an over-parametrization of the filters.
The outputs of different filters $\hat{\bz}^h$ are combined linearly to produce $\bar{\bz}$:
\begin{equation}
\bar{\bz}_c = \sum_{h=1}^H \bw_c^h \hat{\bz}_c^h,
  \label{eq:module}
\end{equation}
using individual weights $\bw_c^h$ per scale $\theta^h$ and channel $c$.
The weights $\bw \in \mathbb{R}^{H\times C}$ are learned using error back-propagation.
The result of the inception module has $C$ channels for every of its $Q$ points, thus $\bar{\bz} \in \mathbb{R}^{Q \times C}$.
The inception module is schematically illustrated in Fig.~\ref{fig:inception}.
In short, information from CNN layers below is filtered using bilateral filters defined in a transformed feature space ($\Lambda \mathbf{f}$).
Most operations in the inception module are parallelizable, resulting in fast runtimes on a GPU.
In this work, inspired from the DenseCRF architecture from~\cite{krahenbuhl2012efficient},
we used pairs of BI modules: one with position features $(u,v)$ and another with both position and
color features $(u,v,r,g,b)$, each with multiple scales $\{\theta^h\}$.

{\bfseries Motivation and Comparison to DenseCRF:}
A BI module filters the activations of a CNN layer. Contrast this with the use of a DenseCRF on the CNN
output. At that point the fine-grained information that intermediate CNN layers represent has been condensed already to a low-dimensional vector representing beliefs over labels. Using a mean-field update is propagating information between these beliefs. Similar behavior is obtained using the BI modules but on different scales (using multiple different filters $K(\theta^h)$) and on the intermediate CNN activations $\bz$. Since in the end, the to-be-predicted pixels are not i.i.d., this blurring leads to better performance both when using a bilateral filter as an approximate message passing step of a DenseCRF as well as in the system outlined here. Both attempts are encoding prior knowledge about the problem, namely that pixels close in position and color are likely to have the same label. Therefore such pixels can also have the same intermediate representation. Consider one would average CNN representations for all pixels that have the same ground truth label. This would result in an intermediate CNN representation that would be very easy to classify for the later layers.

\subsection{Superpixel Convolutions}
The bilateral inception module allows to change how information is stored in the higher level of a CNN. This is where the superpixels are used.
Instead of storing information on a fixed grid, we compute for every image, superpixels $S$ and use the mean color and position of their included pixels as features.
We can insert bilateral inception modules to change from grid representations to superpixel representations and vice versa.
Inception modules in between superpixel layers convolve the unit activations between all superpixels depending on their distance in the feature space.
This retains all properties of the bilateral filter, superpixels that are spatially close and have a similar mean color will have a stronger influence on each other.

Superpixels are not the only choice, in principle one can also sample random points from the image and use them as intermediate representations.
We are using superpixels for computational reasons, since they can be used to propagate label information to the full image resolution.
Other interpolation techniques are possible, including the well known bilinear interpolation, up-convolution networks~\cite{zeiler2010deconvolutional}, and DenseCRFs~\cite{krahenbuhl2012efficient}.
The quantization error mentioned in Sec.~\ref{sec:superpixels} only enters because the superpixels are used for interpolation.
Also note that a fixed grid, that is independent of the image is a hard choice of where information should be stored.
One could in principle evaluate the CNN densely, at all possible spatial locations,
but we found that this resulted in poor performance compared to interpolation methods.

\subsubsection{Back-propagation and Training.}

All free parameters of the inception module $\bw$, $\{\theta^h\}$ and $\Lambda$ are learned
via back-propagation. We also back-propagate the error with respect to the module inputs
thereby enabling the integration of our inception modules inside CNN frameworks without
breaking the end-to-end learning paradigm.
As shown in Fig.~\ref{fig:bi_module}, the bilateral filtering
can be decomposed into 5 different sub-layers.
Derivatives with respect to the open parameters are obtained by the
corresponding layer and standard back-propagation through the directed
acyclic graph.
For example, $\Lambda$ is optimized by back-propagating gradients through $1\times1$ convolution.
Derivatives for non-standard layers (pairwise similarity, matrix
multiplication) are straight forward to obtain using matrix calculus.
To let different filters learn the information propagation at
different scales, we initialized $\{\theta^h\}$ with well separated
scalar values (\eg $\{1, 0.7, 0.3,...\}$).
The learning is performed using the stochastic optimization
method of Adam~\cite{kingma2014adam}.
The implementation is done in the Caffe neural network framework~\cite{jia2014caffe},
and the code is available at http://segmentation.is.tuebingen.mpg.de.

\definecolor{voc_1}{RGB}{0, 0, 0}
\definecolor{voc_2}{RGB}{128, 0, 0}
\definecolor{voc_3}{RGB}{0, 128, 0}
\definecolor{voc_4}{RGB}{128, 128, 0}
\definecolor{voc_5}{RGB}{0, 0, 128}
\definecolor{voc_6}{RGB}{128, 0, 128}
\definecolor{voc_7}{RGB}{0, 128, 128}
\definecolor{voc_8}{RGB}{128, 128, 128}
\definecolor{voc_9}{RGB}{64, 0, 0}
\definecolor{voc_10}{RGB}{192, 0, 0}
\definecolor{voc_11}{RGB}{64, 128, 0}
\definecolor{voc_12}{RGB}{192, 128, 0}
\definecolor{voc_13}{RGB}{64, 0, 128}
\definecolor{voc_14}{RGB}{192, 0, 128}
\definecolor{voc_15}{RGB}{64, 128, 128}
\definecolor{voc_16}{RGB}{192, 128, 128}
\definecolor{voc_17}{RGB}{0, 64, 0}
\definecolor{voc_18}{RGB}{128, 64, 0}
\definecolor{voc_19}{RGB}{0, 192, 0}
\definecolor{voc_20}{RGB}{128, 192, 0}
\definecolor{voc_21}{RGB}{0, 64, 128}
\definecolor{voc_22}{RGB}{128, 64, 128}

\begin{figure*}[t]
  \small
  \centering

  \subfigure{%
    \includegraphics[width=.15\columnwidth]{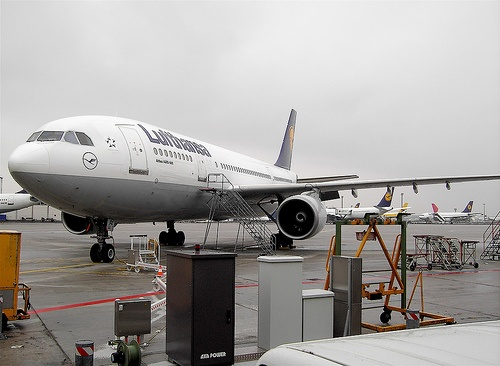}
  }
  \subfigure{%
    \includegraphics[width=.15\columnwidth]{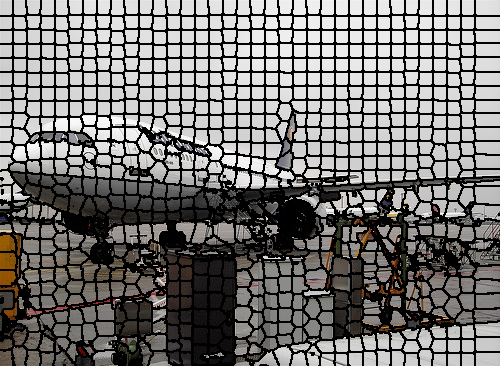}
  }
  \subfigure{%
    \includegraphics[width=.15\columnwidth]{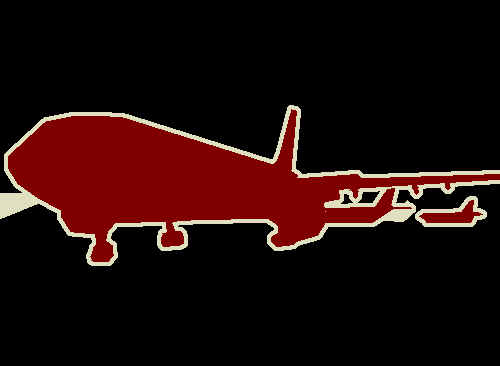}
  }
  \subfigure{%
    \includegraphics[width=.15\columnwidth]{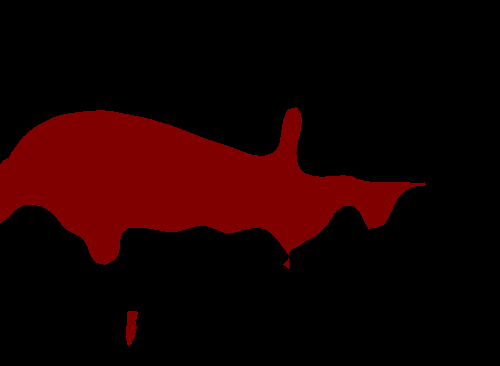}
  }
  \subfigure{%
    \includegraphics[width=.15\columnwidth]{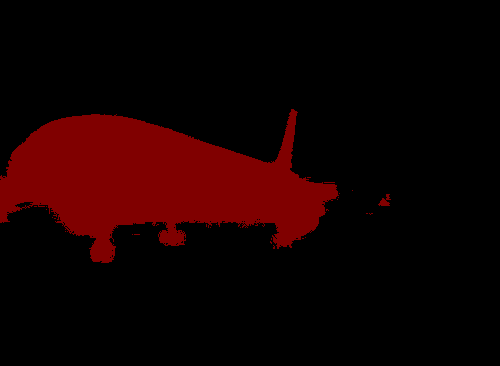}
  }
  \subfigure{%
    \includegraphics[width=.15\columnwidth]{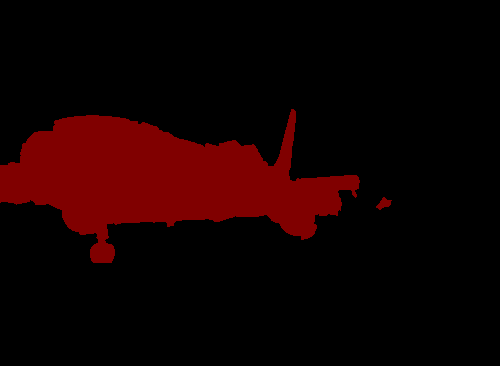}
  }\\[-2ex]
  \setcounter{subfigure}{0}
  \subfigure[\scriptsize Input]{%
    \includegraphics[width=.15\columnwidth]{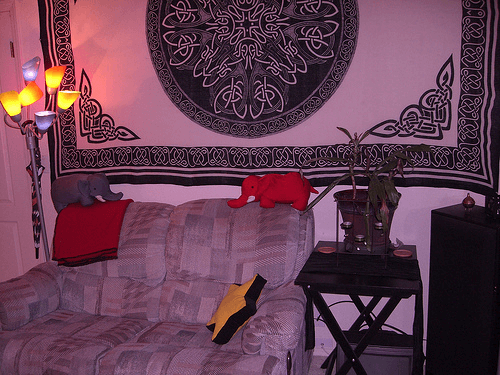}
  }
  \subfigure[\scriptsize Superpixels]{%
    \includegraphics[width=.15\columnwidth]{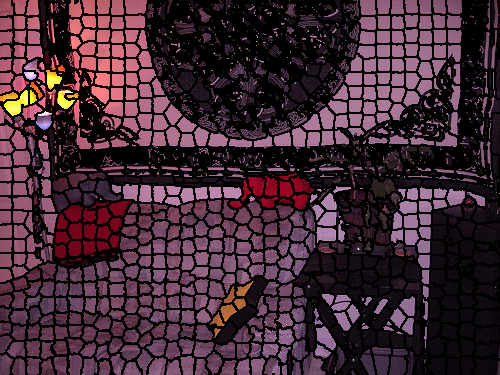}
  }
  \subfigure[\scriptsize GT]{%
    \includegraphics[width=.15\columnwidth]{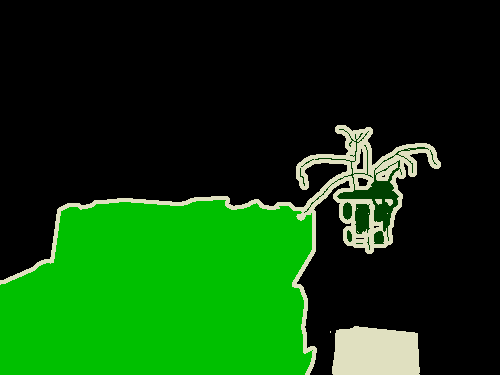}
  }
  \subfigure[\scriptsize Deeplab]{%
    \includegraphics[width=.15\columnwidth]{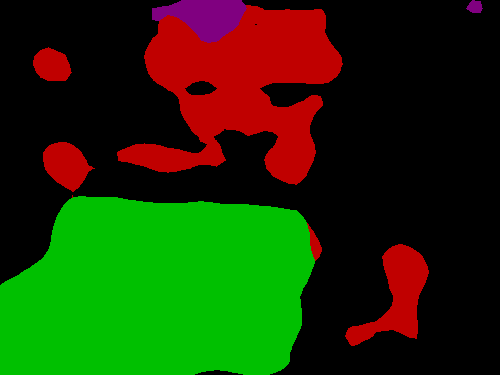}
  }
  \subfigure[\scriptsize +DenseCRF]{%
    \includegraphics[width=.15\columnwidth]{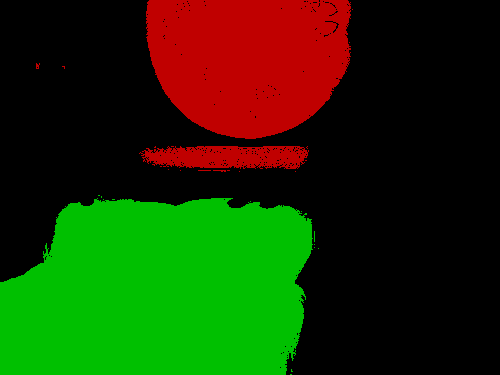}
  }
  \subfigure[\scriptsize Using BI]{%
    \includegraphics[width=.15\columnwidth]{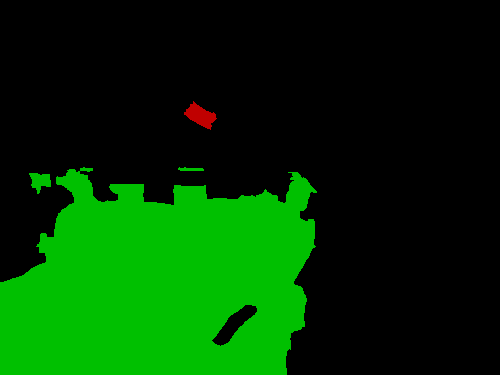}
  }
  \mycaption{Semantic segmentation}{Example results of semantic segmentation
  on Pascal VOC12 dataset.
  (d) depicts the DeepLab CNN result, (e) CNN + 10 steps of mean-field inference,
  (f) result obtained with bilateral inception (BI) modules (\bi{6}{2}+\bi{7}{6}) between \fc~layers.}\label{fig:semantic_visuals}
\end{figure*}

\section{Experiments}

We study the effect of inserting and learning bilateral inception modules
in various existing CNN architectures.
As a testbed, we perform experiments on semantic segmentation
using the Pascal VOC12 segmentation
benchmark dataset~\cite{voc2012segmentation}, Cityscapes street scene dataset~\cite{Cordts2015Cvprw}
and on material segmentation using the Materials in Context (MINC)
dataset from~\cite{bell2015minc}. We take different CNN architectures from the works
of~\cite{chen2014semantic,zheng2015conditional,bell2015minc} and insert the
inception modules before and/or after the spatial FC layers. In Appendix~\ref{sec:binception-app},
we presented some quantitative results with approximate bilateral filtering using the permutohedral
lattice~\cite{adams2010fast}.

\subsection{Semantic Segmentation}

We first use the Pascal VOC12 segmentation dataset~\cite{voc2012segmentation}
with 21 object classes.
For all experiments on VOC12, we train using the extended training set of
10581 images collected by~\cite{hariharan2011moredata}. Following~\cite{zheng2015conditional},
we use a reduced validation set of 346 images for validation.
We experiment on two different network architectures, (a) The DeepLab model from~\cite{chen2014semantic}
which uses a CNN followed by DenseCRF and (b) The CRFasRNN model from~\cite{zheng2015conditional} which uses a
CNN with deconvolution layers followed by DenseCRF trained end-to-end.

\subsubsection{DeepLab}\label{sec:deeplabmodel}
We use the publicly available state-of-the-art pre-trained CNN models from~\cite{chen2014semantic}.
We use the DeepLab-LargeFOV variant as a base architecture and refer to it as `\deeplab'.
The \deeplab~CNN model produces a lower resolution prediction ($\frac{1}{8}\times$)
which is then bilinearly interpolated to the input image resolution.
The original models have been fine-tuned using both the MSCOCO~\cite{lin2014microsoft} and the extended VOC~\cite{hariharan2011moredata} datasets.
Next, we describe modifications to these models and show performance improvements in terms of both IoU and runtimes.

\begin{table}[t]
\centering
  \small
  \begin{tabular}{>{\raggedright\arraybackslash}p{4.8cm}>{\raggedright\arraybackslash}p{1.8cm}>{\centering\arraybackslash}p{1.2cm}>{\centering\arraybackslash}p{1.7cm}}
    \toprule
    \textbf{Model} & Training & \emph{IoU} & \emph{Runtime}(ms)\\
    \midrule
  \deeplablargefov~\cite{chen2014semantic} &  & 68.9 & 145\\
  \midrule
  With BI modules & & & \\
  \bi{6}{2} & only BI & \href{http://host.robots.ox.ac.uk:8080/anonymous/31URIG.html}{70.8} & +20 \\
  \bi{6}{2} & BI+FC & \href{http://host.robots.ox.ac.uk:8080/anonymous/JOB8CE.html}{71.5} & +20\\
  \bi{6}{6} & BI+FC & \href{http://host.robots.ox.ac.uk:8080/anonymous/IB1UAZ.html}{72.9} & +45\\
  \bi{7}{6} & BI+FC & \href{http://host.robots.ox.ac.uk:8080/anonymous/EQB3CR.html}{73.1} & +50\\
  \bi{8}{10} & BI+FC & \href{http://host.robots.ox.ac.uk:8080/anonymous/JR27XL.html}{72.0} & +30\\
  \bi{6}{2}-\bi{7}{6} & BI+FC & \href{http://host.robots.ox.ac.uk:8080/anonymous/VOTV5E.html}{73.6} & +35\\
  \bi{7}{6}-\bi{8}{10} & BI+FC & \href{http://host.robots.ox.ac.uk:8080/anonymous/X7A3GP.html}{73.4} & +55\\
  \bi{6}{2}-\bi{7}{6} & FULL &  \href{http://host.robots.ox.ac.uk:8080/anonymous/CLLB3J.html}{\textbf{74.1}} & +35\\
  \bi{6}{2}-\bi{7}{6}-CRF & FULL & \href{http://host.robots.ox.ac.uk:8080/anonymous/7NGWWU.html}{\textbf{75.1}} & +865\\
  \midrule
  \deeplablargefovcrf~\cite{chen2014semantic} & & 72.7 & +830\\
  \deeplabmsclargefovcrf~\cite{chen2014semantic} & & \textbf{73.6} & +880\\
  DeepLab-EdgeNet~\cite{chen2015semantic} & & 71.7 & +30\\
  DeepLab-EdgeNet-CRF~\cite{chen2015semantic} & & \textbf{73.6} & +860\\
  \bottomrule
\end{tabular}
\mycaption{Semantic segmentation using \deeplablargefov~model}
{IoU scores on Pascal VOC12 segmentation test dataset
and average runtimes (ms) corresponding to different models. Also shown are the results
corresponding to competitive dense pixel prediction techniques that used
the same base DeepLab CNN. Runtimes also include superpixel
computation (6ms). In the second column, `BI', `FC' and `FULL'
correspond to training `BI', `FC' and full model layers respectively.}
\label{tab:deeplabresults}
\end{table}

We add inception modules after different FC layers in the original model and
remove the DenseCRF post processing. For this dataset, we use
1000 SLIC superpixels~\cite{achanta2012slic,gSLICr_2015}.
The inception modules after \fc{6}, \fc{7} and \fc{8} layers are referred to as
\bi{6}{H}, \bi{7}{H} and \bi{8}{H} respectively, where $H$ is the number
of kernels. All results using the~\deeplab~model on Pascal VOC12 dataset
are summarized in Tab.~\ref{tab:deeplabresults}.
We report the `test' numbers without validation numbers, because the released
DeepLab model that we adapted was trained using both train and validation sets.
The~\deeplab~network achieves an IoU of 68.9 after bilinear interpolation.
Experiments with the \bi{6}{2}
module indicate that even only learning the inception module while keeping the remaining
network fixed results in a reliable IoU improvement ($+1.9$).
Additional joint training with \fc{} layers significantly improved the performance.
The results also show that more kernels improve performance.
Next, we add multiple modules to the base DeepLab network at various stages and train them jointly. This
results in further improvement of the performance. The \bi{6}{2}-\bi{7}{6} model with
two inception modules shows significant improvement in IoU by $4.7$ and $0.9$
 in comparison to the baseline model and DenseCRF application respectively.
Finally, fine-tuning the entire network (FULL in Tab.~\ref{tab:deeplabresults})
boosts the performance by $5.2$ and $1.4$ compared to the baseline and DenseCRF application.

Some visual results are shown in Fig.~\ref{fig:semantic_visuals} and more
are included in Appendix~\ref{sec:qualitative-app}.
Several other variants of using BI are conceivable. During our experiments,
we have observed that more kernels and more modules improve the performance,
so we expect that even better results can be achieved.
In Tab.~\ref{tab:deeplabresults}, the runtime in milliseconds is included for several models.
These numbers have been obtained using a
Nvidia Tesla K80 GPU and standard Caffe time benchmarking~\cite{jia2014caffe}.
DenseCRF timings are taken from~\cite{chen2015semantic}. The runtimes indicate
that the overhead with BI modules is quite minimal in comparison to
using Dense CRF.

In addition, we include the results of some other dense pixel prediction methods that
are build on top of the same DeepLab base model. \deeplabmsclargefovcrf~is a multi-scale
version~\cite{chen2014semantic} of \deeplab~with DenseCRF on top. DeepLab-EdgeNet~\cite{chen2015semantic}
is a recently proposed fast and discriminatively trained domain transform technique
for propagating information across pixels. Comparison with these techniques in terms
of performance and runtime indicates that our approach performs on par with latest
dense pixel prediction techniques with significantly less time overhead.
Several
state-of-the-art CNN based systems~\cite{lin2015efficient,liu2015semantic}
have achieved higher results than \deeplab~on Pascal VOC12. These models are not yet
publicly available and so we could not test the use of BI models in them.
A close variant~\cite{barron2015bilateral} of our work, which propose
to do optimization in the bilateral space also has fast runtimes, but reported
lower performance in comparison to the application of DenseCRF.

\begin{table}
\centering
  \small
  \begin{tabular}{>{\raggedright\arraybackslash}p{5.0cm}>{\centering\arraybackslash}p{1.2cm}>{\centering\arraybackslash}p{1.9cm}}
    \toprule
    \textbf{Model} & \emph{IoU} & \emph{Runtime}(ms)\\
    \midrule
  \deconv(CNN+Deconv.) & 72.0 & 190 \\
  \midrule
  With BI modules & & \\
  \bi{3}{2}-\bi{4}{2}-\bi{6}{2}-\bi{7}{2} & \textbf{74.9} & 245 \\
  \midrule
  CRFasRNN (\deconvcrf)& 74.7 & 2700\\
  \bottomrule
\end{tabular}
\mycaption{Semantic segmentation using CRFasRNN model}{IoU scores and runtimes corresponding to different models on Pascal VOC12 test dataset. Note that runtime also includes superpixel computation.}
\label{tab:deconvresults}
\end{table}

\subsubsection{CRFasRNN}
As a second architecture, we modified the CNN architecture
trained by~\cite{zheng2015conditional} that
produces a result at an even lower resolution ($\frac{1}{16} \times$).
Multiple deconvolution steps are employed to obtain the segmentation at input
image resolution. This result is then passed onto the DenseCRF recurrent neural network
to obtain the final segmentation result.
We insert BI modules after score-pool3, score-pool4, \fc{6} and \fc{7} layers, please
see~\cite{long2014fully,zheng2015conditional} for the network architecture details.
Instead of combining outputs from the above layers with deconvolution steps, we introduce
BI modules after them and linearly combined the outputs to obtain a final segmentation result.
Note that we entirely removed both the deconvolution and the DenseCRF parts of the original model~\cite{zheng2015conditional}.
See Tab.~\ref{tab:deconvresults} for results on the DeconvNet model.
Without the DenseCRF part and only evaluating the deconvolutional part of this
model, one obtains an IoU score of $72.0$.
Ten steps of mean field inference increase the IoU to $74.7$~\cite{zheng2015conditional}. Our model, with few additional parameters compared to the base CNN,
achieves a IoU performance of $74.9$, showing an improvement of 0.2 over the CRFasRNN model.
The BI layers lead to better performance than deconvolution and DenseCRF combined while being much faster.

\subsubsection{Hierarchical Clustering Analysis}
\begin{figure}[t]
\begin{tabular}{c}
  \subfigure{%
    \includegraphics[width=0.25\textwidth]{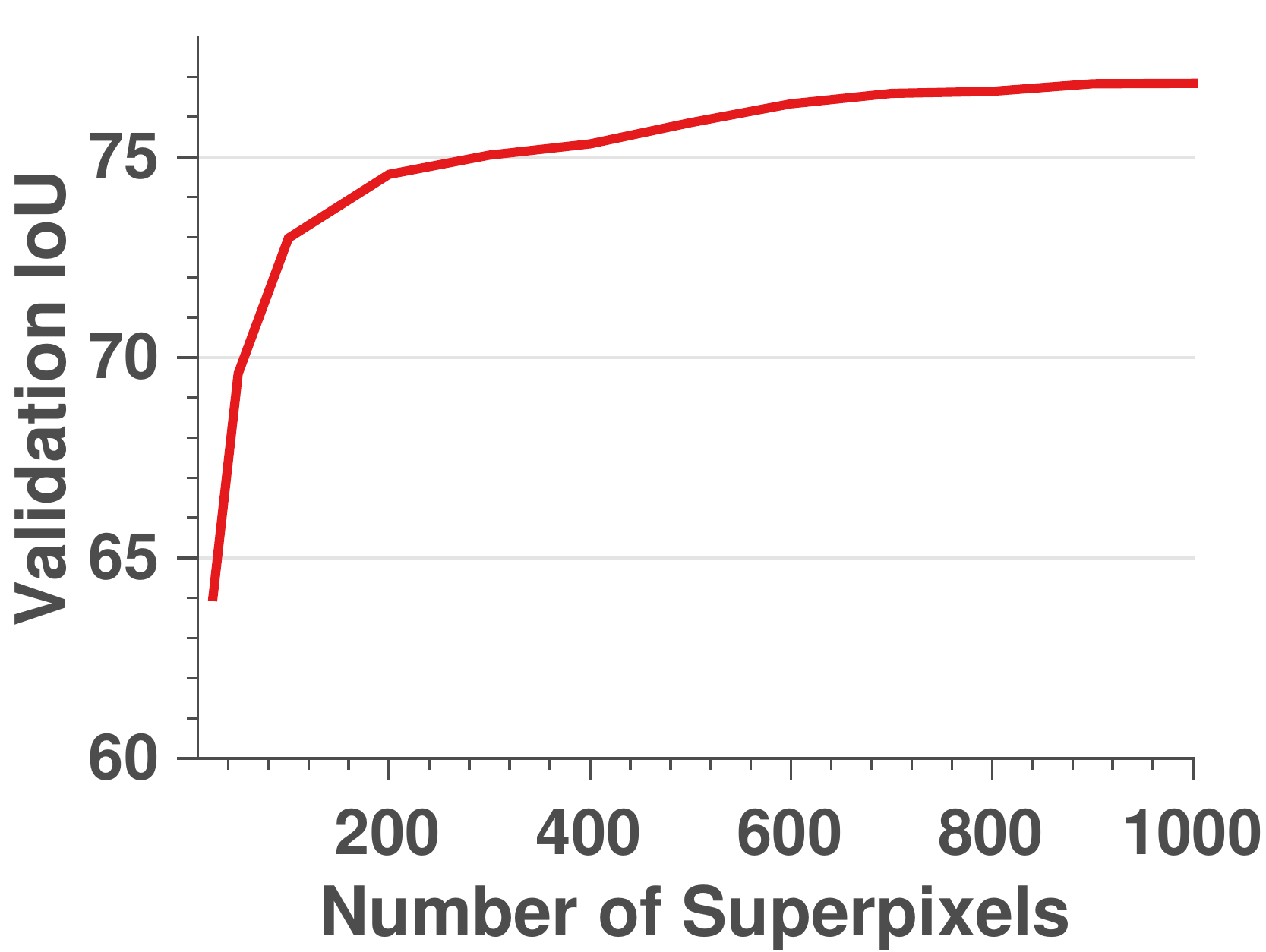}
  }
\end{tabular} \hfill
\begin{tabular}{c}
  \subfigure{%
    \includegraphics[width=0.14\textwidth]{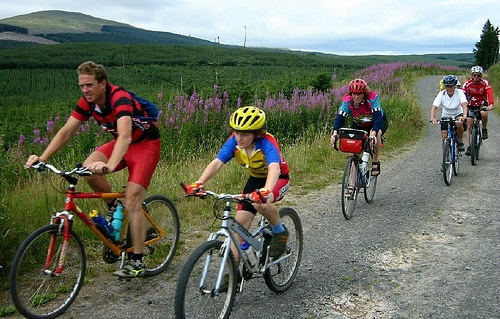}
  }\\
  \subfigure{%
    \includegraphics[width=0.14\textwidth]{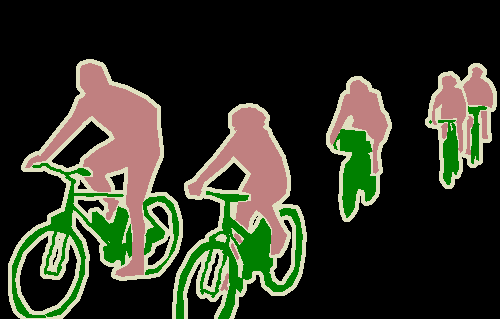}
  }
\end{tabular} \hfill
\begin{tabular}{c}
  \subfigure{%
    \includegraphics[width=0.14\textwidth]{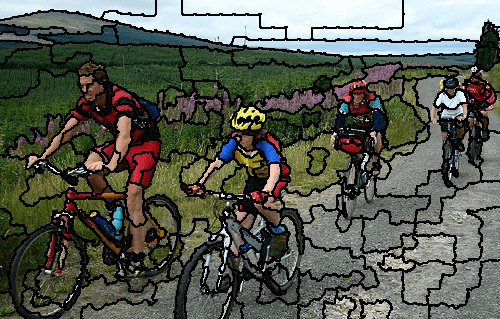}
  }\\
  \subfigure{%
    \includegraphics[width=0.14\textwidth]{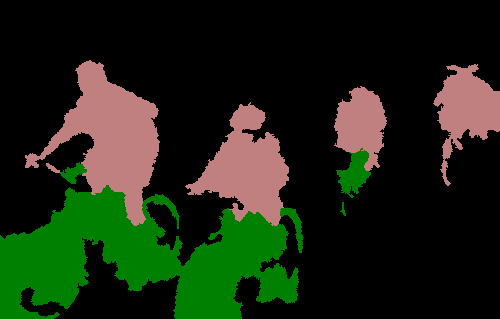}
  }
\end{tabular}\hfill
\begin{tabular}{c}
  \subfigure{%
    \includegraphics[width=0.14\textwidth]{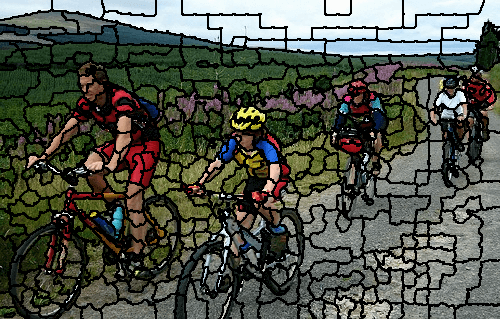}
  }\\
  \subfigure{%
    \includegraphics[width=0.14\textwidth]{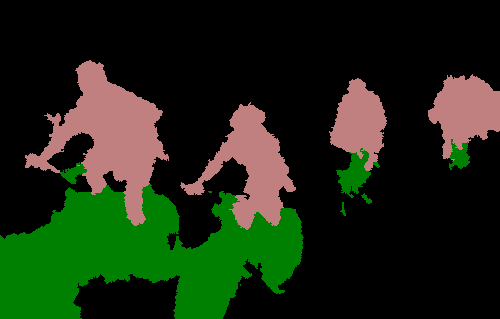}
  }
\end{tabular}\hfill
\begin{tabular}{c}
  \subfigure{%
    \includegraphics[width=0.14\textwidth]{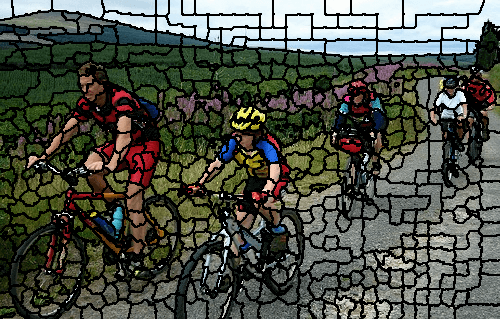}
  }\\
  \subfigure{%
    \includegraphics[width=0.14\textwidth]{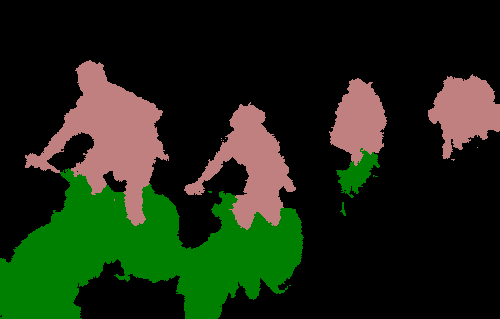}
  }
\end{tabular}
  \mycaption{Hierarchical clustering analysis}{From left to right: Validation performance when using different super-pixel layouts, visualization of an image with ground truth segmentation, and the \bi{6}{2}-\bi{7}{6} result with 200, 600, and 1000 superpixels.}
  \label{fig:clustering}
\end{figure}

We learned the network parameters using 1000 gSLIC superpixels per image, however the inception module allows to change the resolution (a non-square $K$).
To illustrate this, we perform agglomerative clustering of the superpixels, sequentially merging the nearest (in spatial and
photometric features) two superpixels into a single one.
We then evaluated the DeepLab-\bi{6}{2}-\bi{7}{6} network using different levels of the resulting hierarchy re-using all the trained network parameters.
Results in Fig.~\ref{fig:clustering} show that the IoU score on the validation set decreases slowly with decreasing number of points and then drops for less than 200 superpixels.
This validates that the network generalizes to different superpixel layouts and it is sufficient to represent larger regions of similar color by fewer points.
In future, we plan to explore different strategies to allocate the representation to those regions that require more resolution and to remove the superpixelization altogether.
Fig.~\ref{fig:clustering} shows example image with 200, 600, and 1000 superpixels and their obtained segmentation with BI modules.

\subsection{Material Segmentation}

We also experiment on a different pixel prediction task of
material segmentation by adapting a CNN
architecture fine-tuned for Materials in Context (MINC)~\cite{bell2015minc} dataset.
MINC consists of 23 material classes and is available in three different
resolutions with the same aspect ratio: low ($550^2$), mid ($1100^2$) and an original higher
resolution.
The authors of~\cite{bell2015minc} train CNNs on the mid resolution images and then
combine with a DenseCRF to predict and evaluate on low resolution images.
We build our work based on the AlexNet model~\cite{krizhevsky2012imagenet}
released by the authors of~\cite{bell2015minc}. To obtain a
per pixel labeling of a given image, there are several
processing steps that~\cite{bell2015minc} use for good performance. First,
a CNN is applied at several scales with different strides followed by an interpolation of
the predictions to reach the input image resolution and is then followed by a
DenseCRF. For simplicity, we choose to run the CNN network
with single scale. The authors used just one kernel
with $(u, v, L, a, b)$ features in the DenseCRF part.
We used the same features in our inception modules.
We modified the base AlexNet model by inserting BI modules after \fc{7}
and \fc{8} layers. Again, 1000 SLIC superpixels are used for all experiments.
Results on the test set are shown in Table~\ref{tab:mincresults}.
When inserting BI modules, the performance improves both in total pixel accuracy as well as in class-averaged
accuracy. We observe an improvement
of $12\%$ compared to CNN predictions and $2-4\%$ compared to CNN+DenseCRF results.
Qualitative examples are shown in Fig.~\ref{fig:material_visuals} and more are included
in Appendix~\ref{sec:qualitative-app}. The weights to combine outputs in the BI layers are
found by validation on the validation set. For this model we do not provide any
learned setup due to very limited segment training data.

\begin{table}
\centering
  \small
  \begin{tabular}{p{4.2cm}>{\centering\arraybackslash}p{4cm}>{\centering\arraybackslash}p{1.9cm}}
    \toprule
    \textbf{Model} & \emph{Class / Total accuracy} & \emph{Runtime}(ms)\\
    \midrule
    AlexNet CNN & 55.3 / 58.9 & 300 \\
    \midrule
    \bi{7}{2}-\bi{8}{6} & 67.7 / 71.3 & 410 \\
    \bi{7}{6}-\bi{8}{6} & \textbf{69.4 / 72.8} & 470 \\
    \midrule
    AlexNet-CRF & 65.5 / 71.0 & 3400 \\
    \bottomrule
  \end{tabular}
  \mycaption{Material segmentation using AlexNet}{Pixel accuracies and runtimes (in ms)
  of different models on the MINC material segmentation dataset~\cite{bell2015minc}.
  Runtimes also include the time for superpixel extraction (15ms).}
  \label{tab:mincresults}
\end{table}

\definecolor{minc_1}{HTML}{771111}
\definecolor{minc_2}{HTML}{CAC690}
\definecolor{minc_3}{HTML}{EEEEEE}
\definecolor{minc_4}{HTML}{7C8FA6}
\definecolor{minc_5}{HTML}{597D31}
\definecolor{minc_6}{HTML}{104410}
\definecolor{minc_7}{HTML}{BB819C}
\definecolor{minc_8}{HTML}{D0CE48}
\definecolor{minc_9}{HTML}{622745}
\definecolor{minc_10}{HTML}{666666}
\definecolor{minc_11}{HTML}{D54A31}
\definecolor{minc_12}{HTML}{101044}
\definecolor{minc_13}{HTML}{444126}
\definecolor{minc_14}{HTML}{75D646}
\definecolor{minc_15}{HTML}{DD4348}
\definecolor{minc_16}{HTML}{5C8577}
\definecolor{minc_17}{HTML}{C78472}
\definecolor{minc_18}{HTML}{75D6D0}
\definecolor{minc_19}{HTML}{5B4586}
\definecolor{minc_20}{HTML}{C04393}
\definecolor{minc_21}{HTML}{D69948}
\definecolor{minc_22}{HTML}{7370D8}
\definecolor{minc_23}{HTML}{7A3622}
\definecolor{minc_24}{HTML}{000000}

\begin{figure*}[t]
  \tiny 
  \centering
  \subfigure{%
    \includegraphics[width=.15\columnwidth]{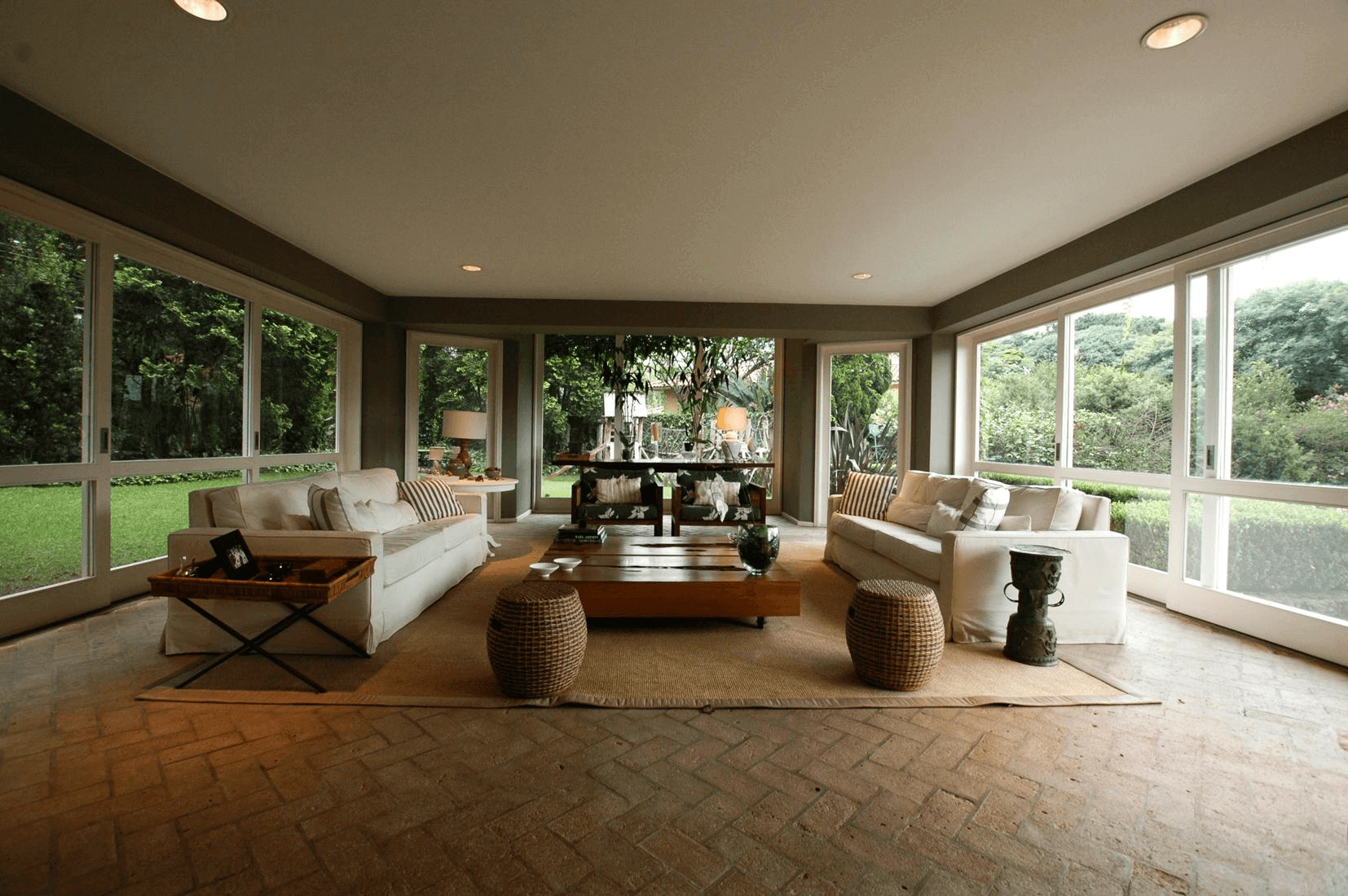}
  }
  \subfigure{%
    \includegraphics[width=.15\columnwidth]{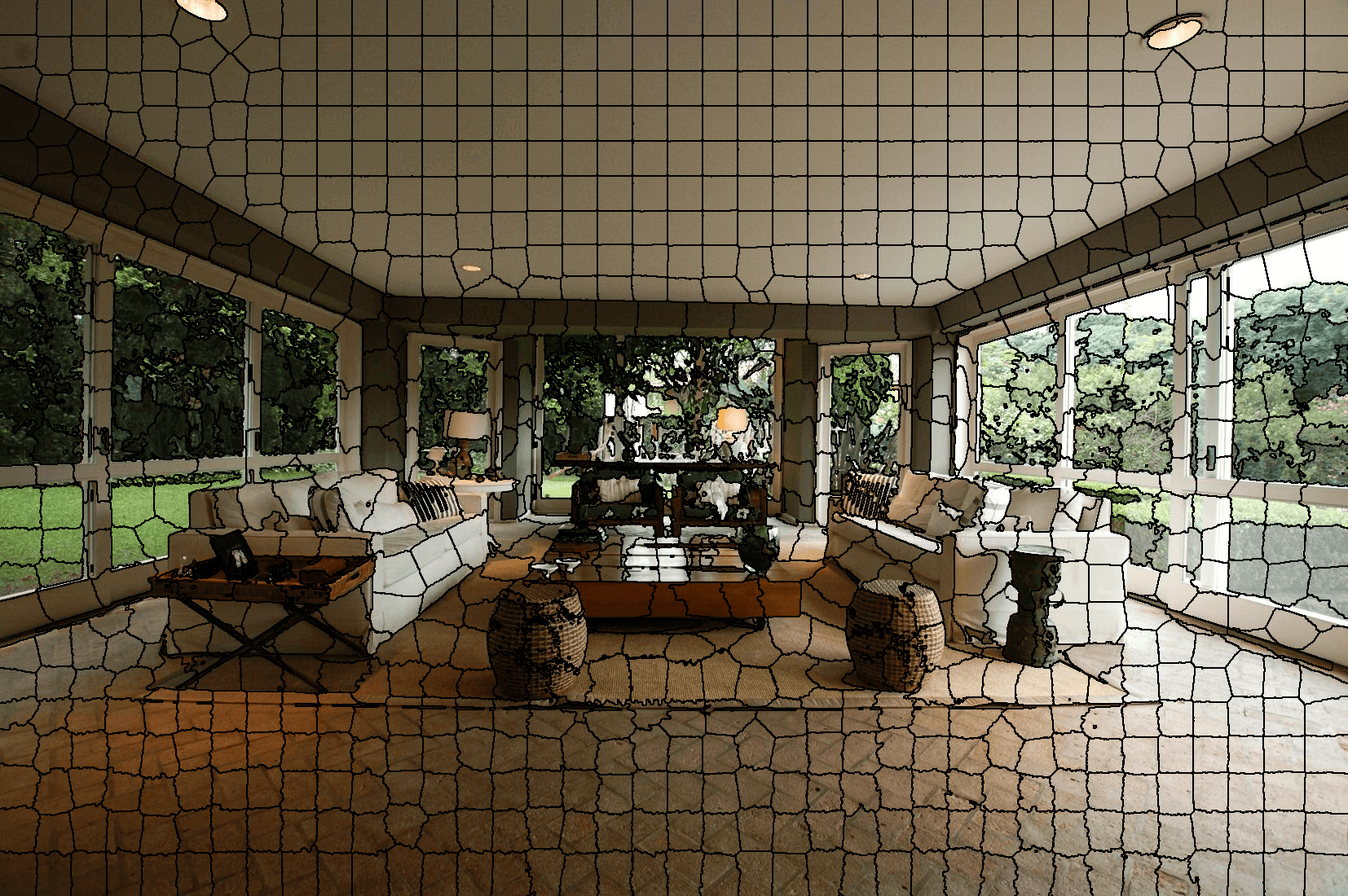}
  }
  \subfigure{%
    \includegraphics[width=.15\columnwidth]{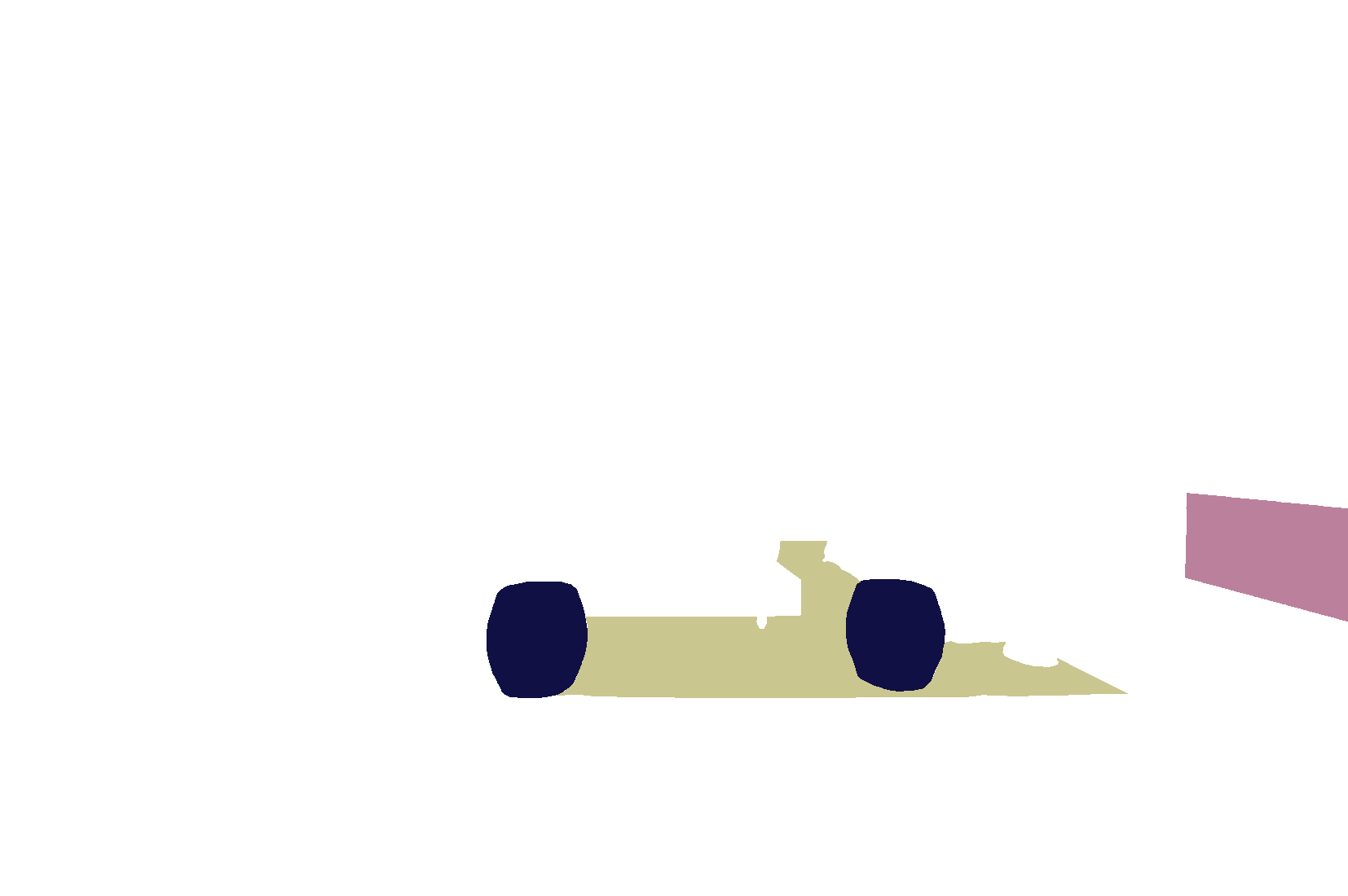}
  }
  \subfigure{%
    \includegraphics[width=.15\columnwidth]{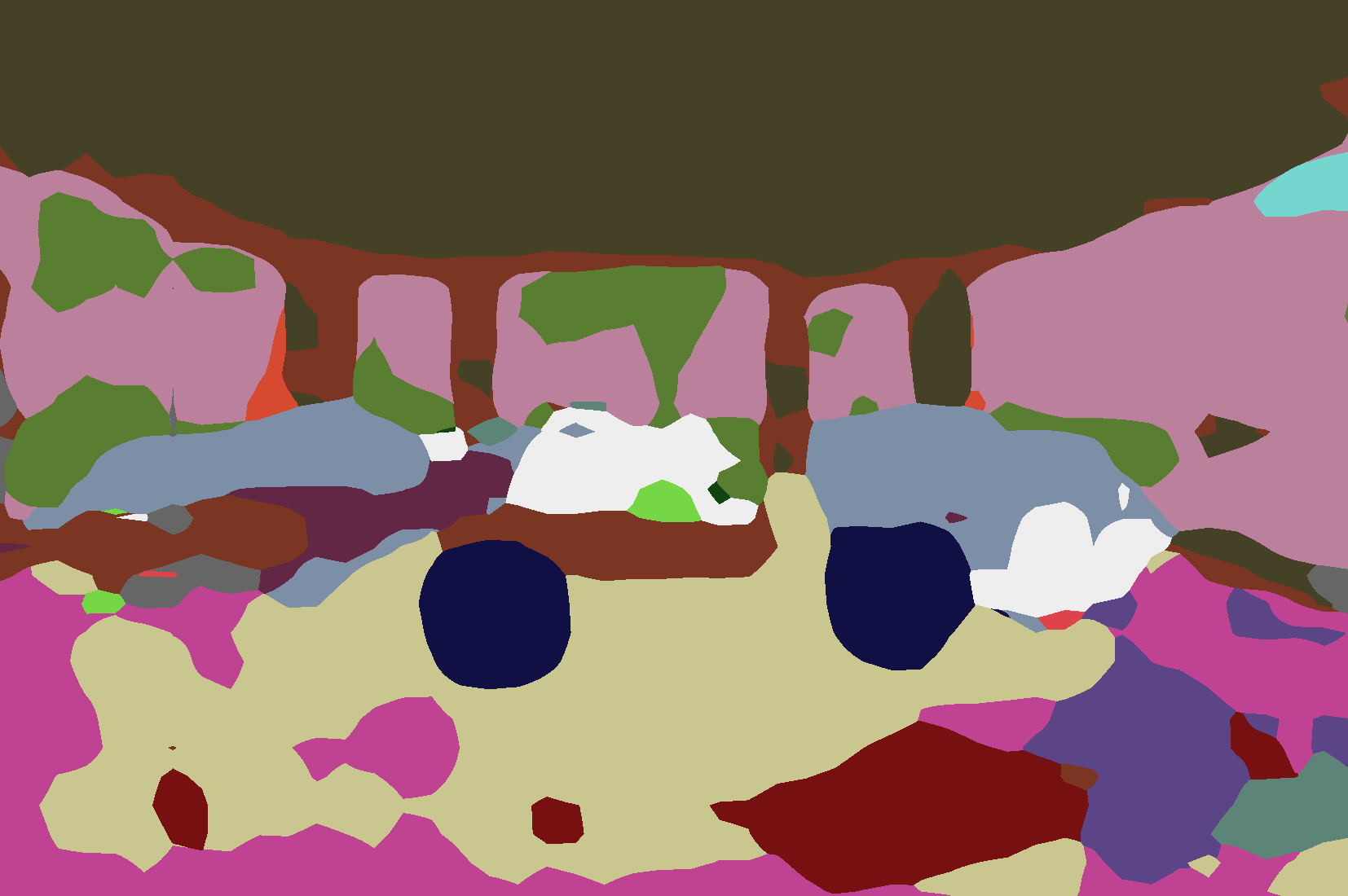}
  }
  \subfigure{%
    \includegraphics[width=.15\columnwidth]{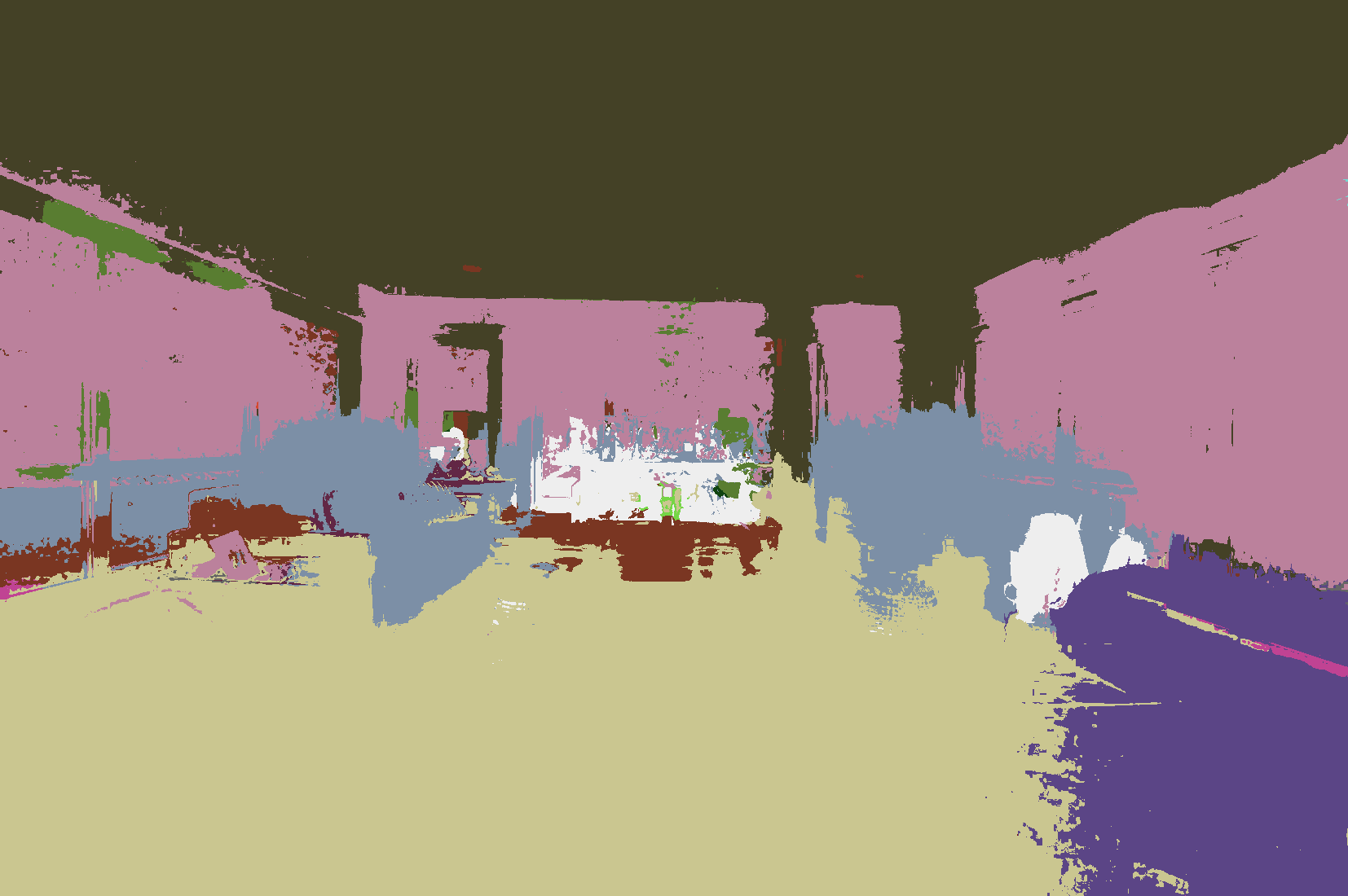}
  }
  \subfigure{%
    \includegraphics[width=.15\columnwidth]{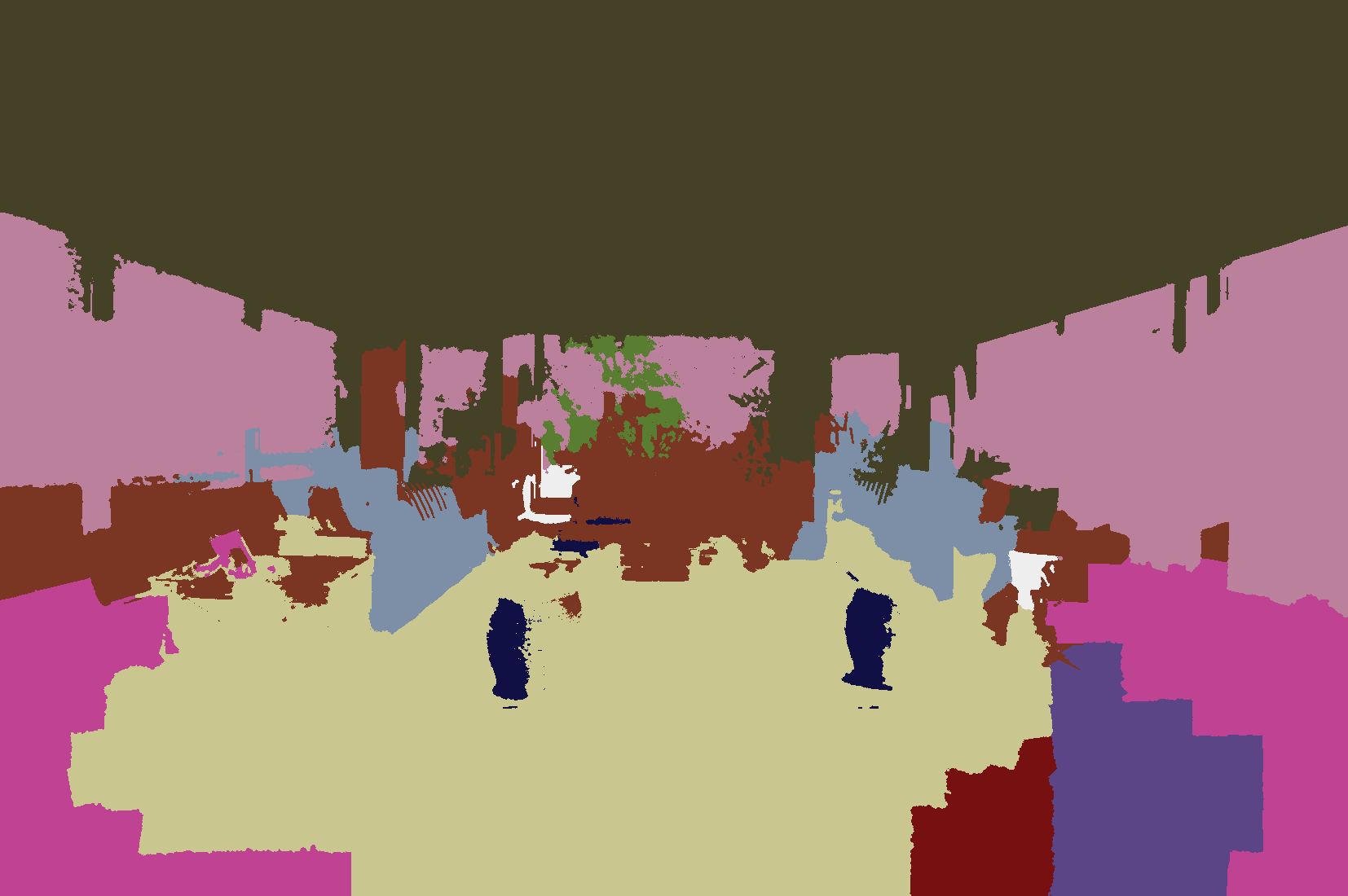}
  }\\[-2ex]
  \setcounter{subfigure}{0}
  \subfigure[\tiny Input]{%
    \includegraphics[width=.15\columnwidth]{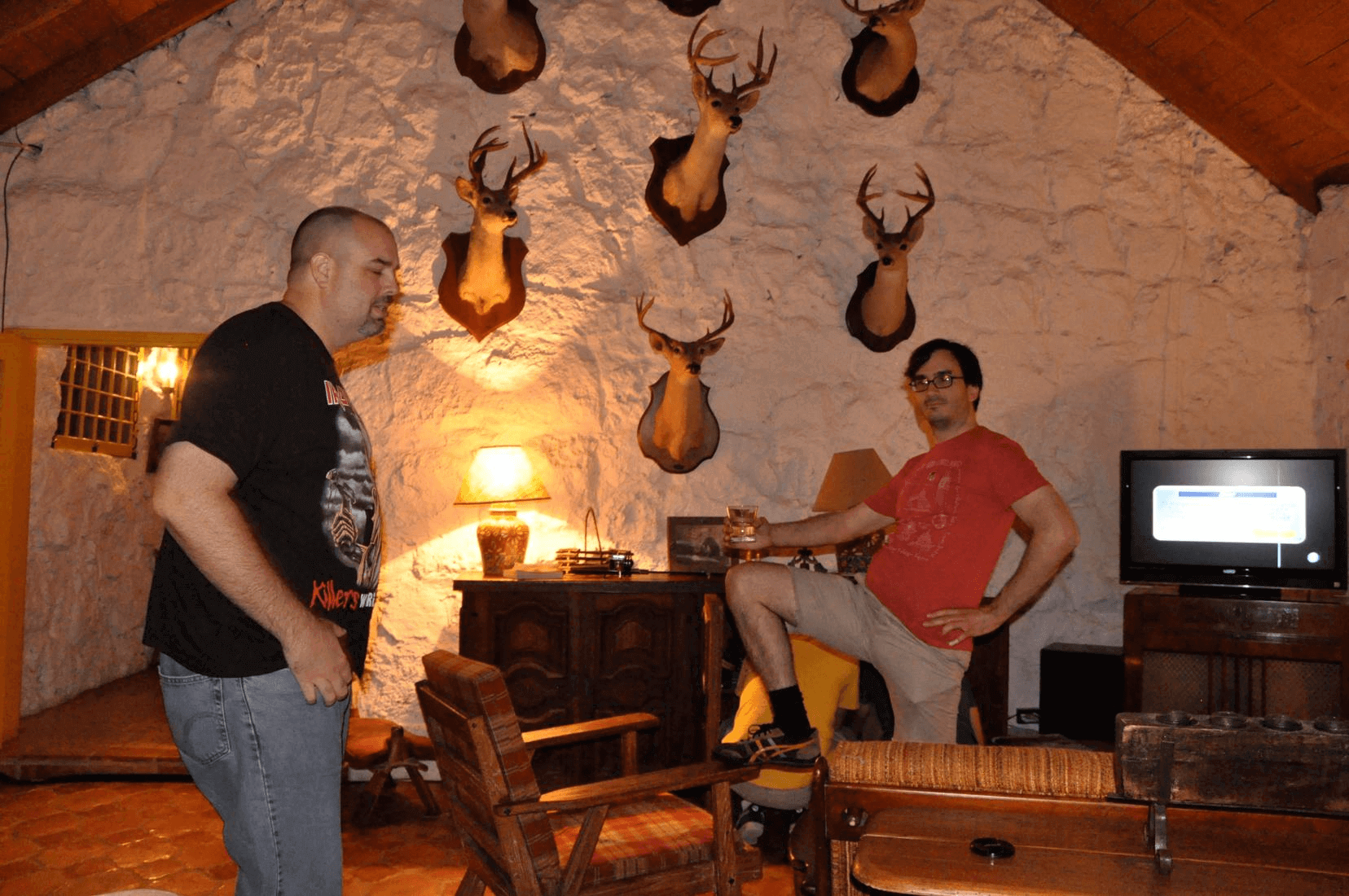}
  }
  \subfigure[\tiny Superpixels]{%
    \includegraphics[width=.15\columnwidth]{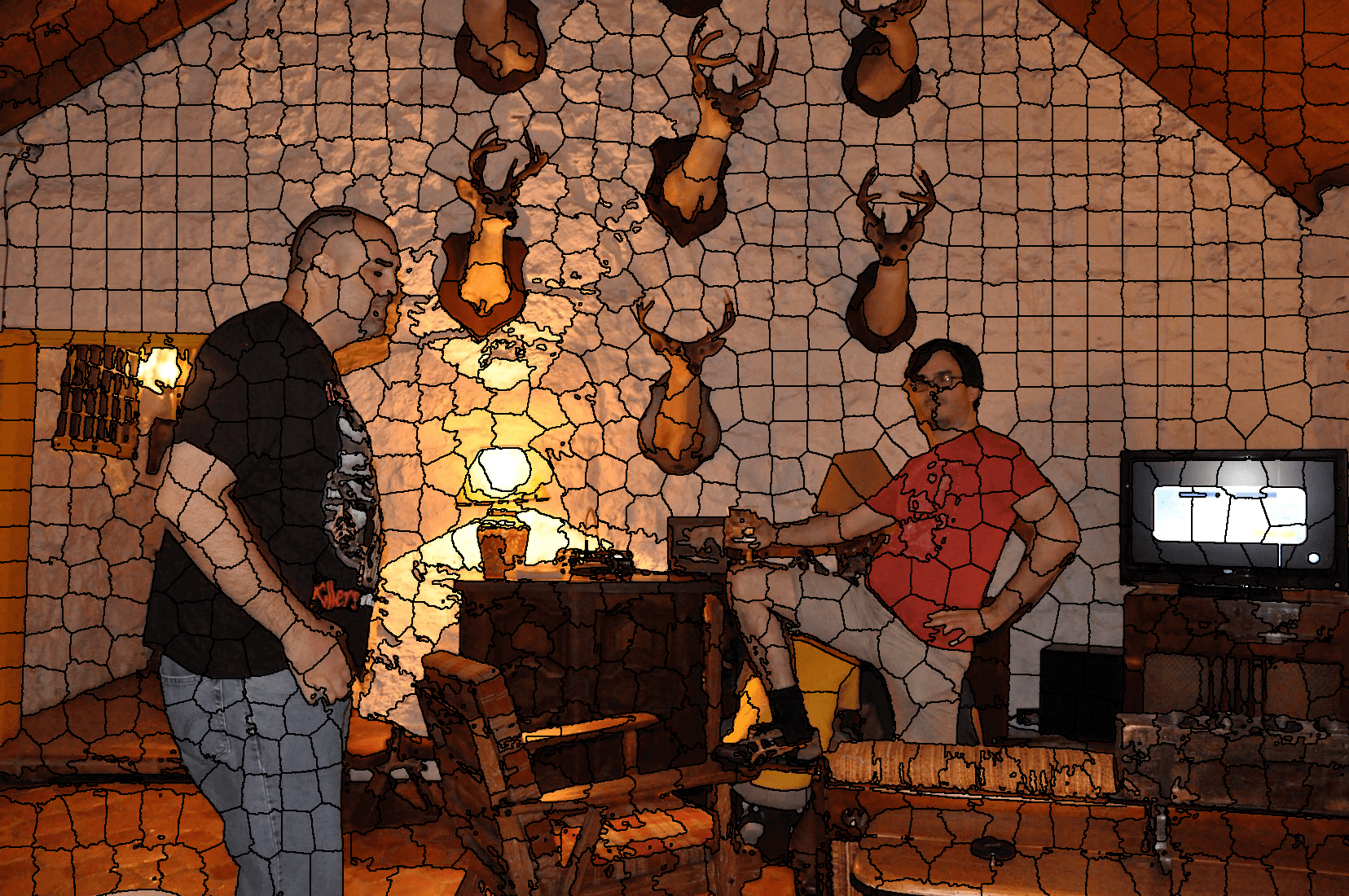}
  }
  \subfigure[\tiny GT]{%
    \includegraphics[width=.15\columnwidth]{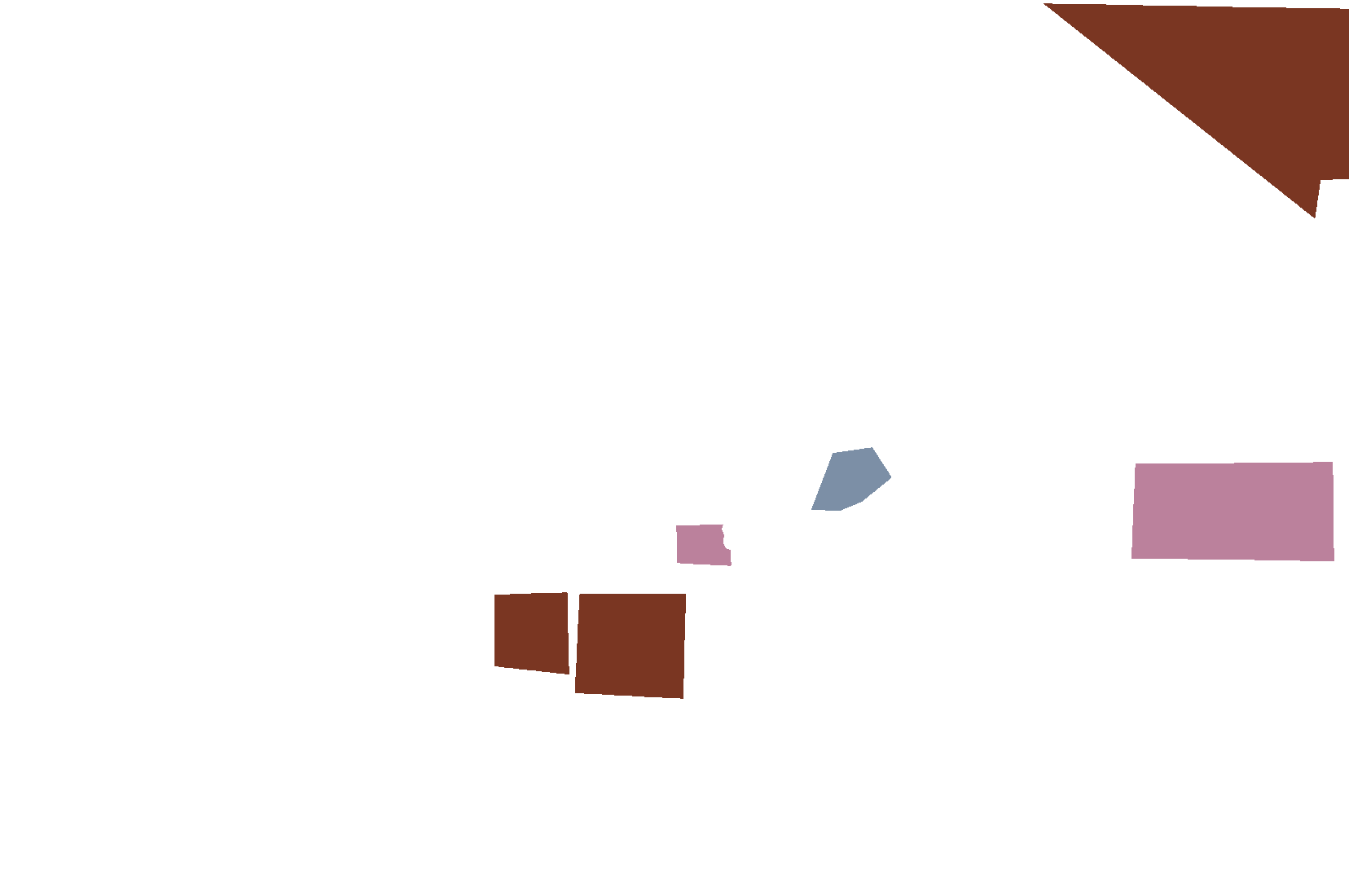}
  }
  \subfigure[\tiny AlexNet]{%
    \includegraphics[width=.15\columnwidth]{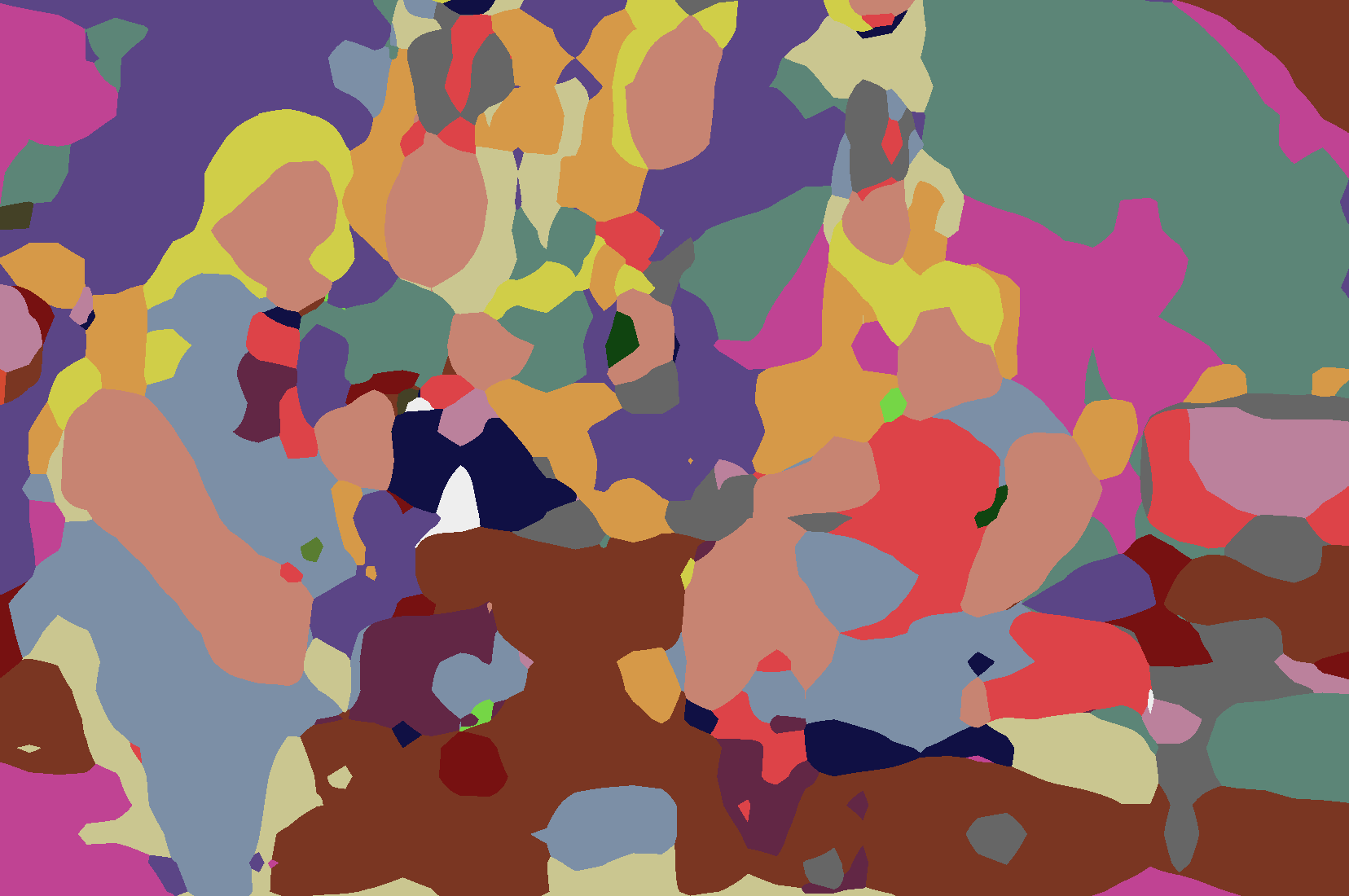}
  }
  \subfigure[\tiny +DenseCRF]{%
    \includegraphics[width=.15\columnwidth]{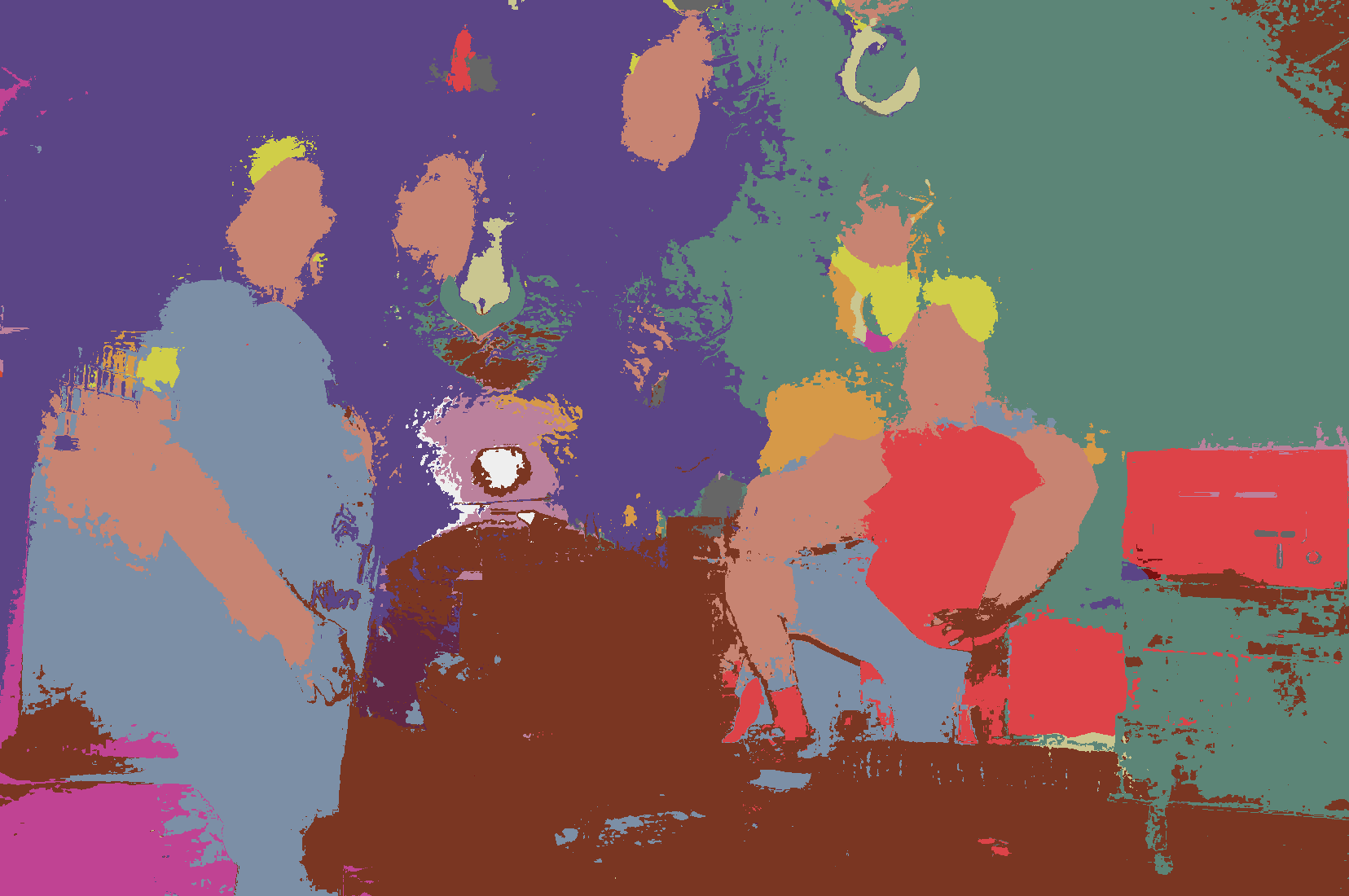}
  }
  \subfigure[\tiny Using BI]{%
    \includegraphics[width=.15\columnwidth]{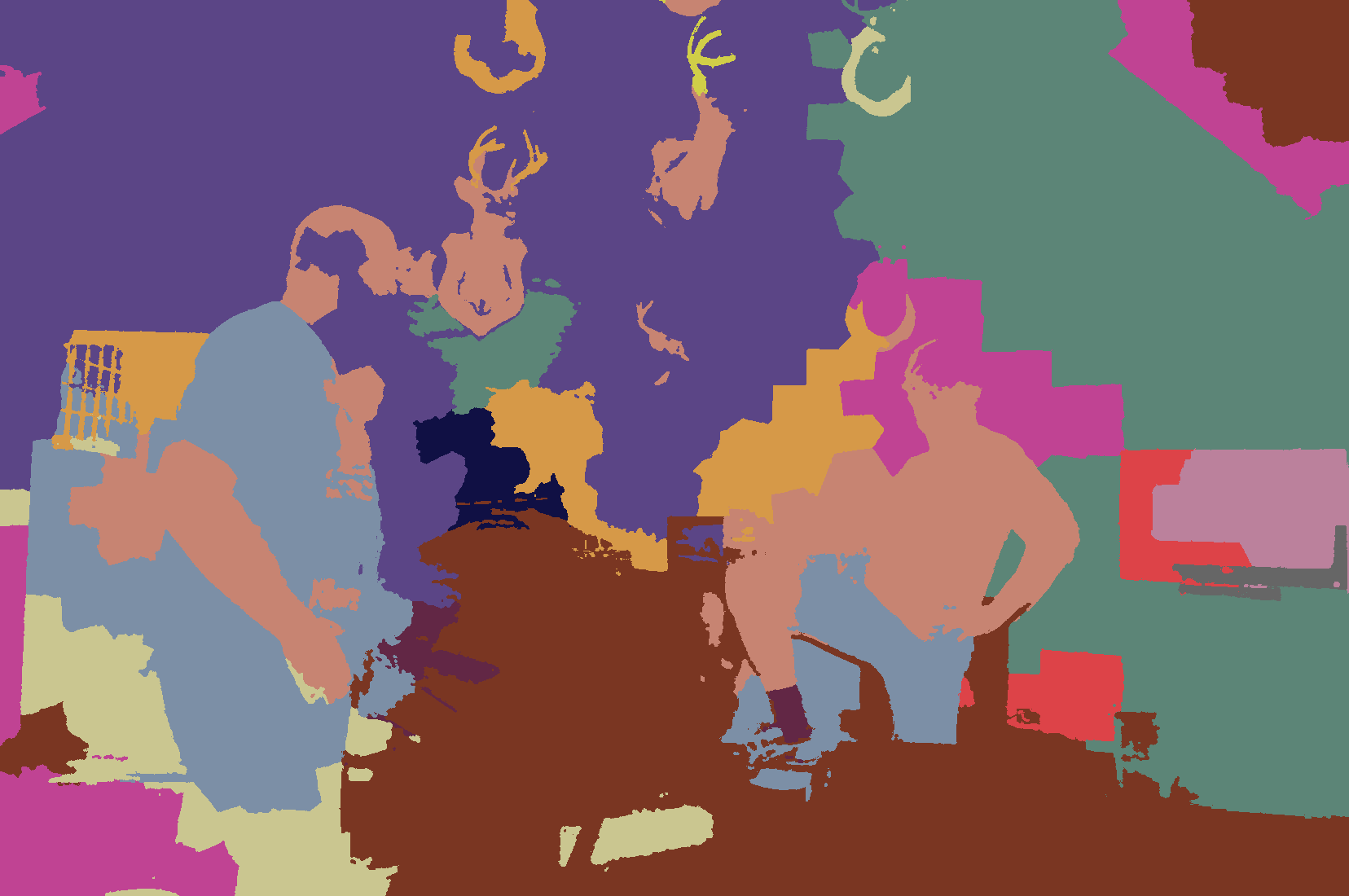}
  }
  \mycaption{Material segmentation}{Example results of material segmentation.
  (d) depicts the AlexNet CNN result, (e) CNN + 10 steps of mean-field inference,
  (f) results obtained with bilateral inception (BI) modules (\bi{7}{2}+\bi{8}{6}) between
  \fc~layers.}
\label{fig:material_visuals}
\end{figure*}

\begin{table}
\centering
  \small
  \begin{tabular}{p{3.2cm}>{\centering\arraybackslash}p{2.8cm}>{\centering\arraybackslash}p{2.8cm}>{\centering\arraybackslash}p{1.9cm}}
    \toprule
    \textbf{Model} & \emph{IoU (Half-res.)} & \emph{IoU (Full-res.)} & \emph{Runtime}(s)\\
    \midrule
    \deeplablargefov~CNN & 62.2 & 65.7 & 0.3 \\
    \midrule
    \bi{6}{2} & 62.7 & 66.5 & 5.7 \\
    \bi{6}{2}-\bi{7}{6} & \textbf{63.1} & \textbf{66.9} & 6.1 \\
    \midrule
    \deeplablargefovcrf & 63.0 & 66.6 & 6.9 \\
    \bottomrule
  \end{tabular}
  \mycaption{Street scene Segmentation using \deeplablargefov~model}
  {IoU scores and runtimes (in sec) of different models on Cityscapes
  segmentation dataset~\cite{Cordts2015Cvprw}, for both half-resolution
  and full-resolution images. Runtime computations also include superpixel
  computation time (5.2s).}
  \label{tab:cityscaperesults}
\end{table}

\subsection{Street Scene Segmentation}

We further evaluate the use of BI modules on the Cityscapes dataset~\cite{Cordts2015Cvprw}.
Cityscapes contains 20K high-resolution ($1024\times2048$) images of street scenes with
coarse pixel annotations and another 5K images with fine annotations, all annotations are from 19 semantic classes.
The 5K images are divided into 2975 train, 500 validation and remaining test images.
Since there are no publicly available pre-trained models for this dataset yet,
we trained a \deeplablargefov~model.
We trained the base \deeplablargefov~model with half resolution images ($512\times1024$) so that the model fits into GPU memory. The result is then interpolated to full-resolution using bilinear interpolation.

\definecolor{city_1}{RGB}{128, 64, 128}
\definecolor{city_2}{RGB}{244, 35, 232}
\definecolor{city_3}{RGB}{70, 70, 70}
\definecolor{city_4}{RGB}{102, 102, 156}
\definecolor{city_5}{RGB}{190, 153, 153}
\definecolor{city_6}{RGB}{153, 153, 153}
\definecolor{city_7}{RGB}{250, 170, 30}
\definecolor{city_8}{RGB}{220, 220, 0}
\definecolor{city_9}{RGB}{107, 142, 35}
\definecolor{city_10}{RGB}{152, 251, 152}
\definecolor{city_11}{RGB}{70, 130, 180}
\definecolor{city_12}{RGB}{220, 20, 60}
\definecolor{city_13}{RGB}{255, 0, 0}
\definecolor{city_14}{RGB}{0, 0, 142}
\definecolor{city_15}{RGB}{0, 0, 70}
\definecolor{city_16}{RGB}{0, 60, 100}
\definecolor{city_17}{RGB}{0, 80, 100}
\definecolor{city_18}{RGB}{0, 0, 230}
\definecolor{city_19}{RGB}{119, 11, 32}
\begin{figure*}[t]
  \tiny 
  \centering
  \subfigure{%
    \includegraphics[width=.18\columnwidth]{figures/frankfurt00000_008206_given.png}
  }
  \subfigure{%
    \includegraphics[width=.18\columnwidth]{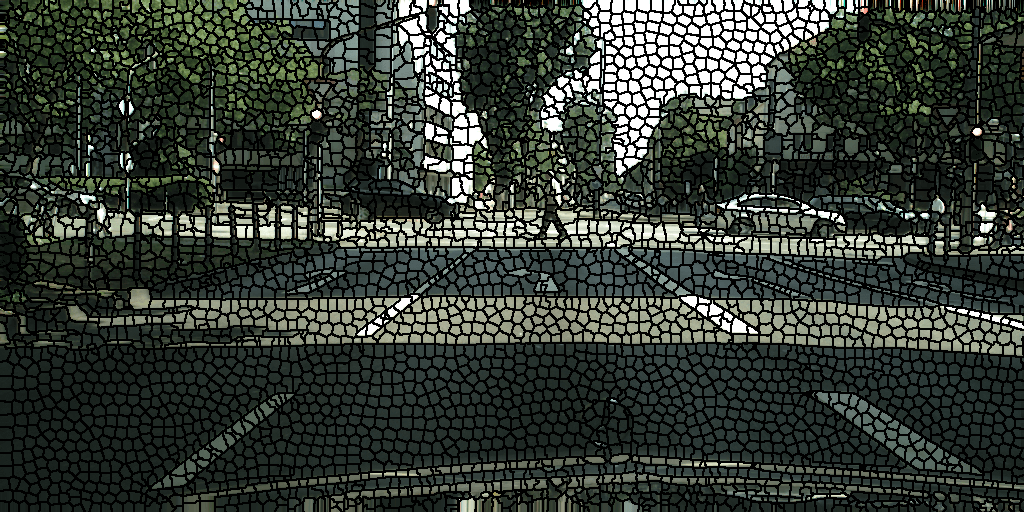}
  }
  \subfigure{%
    \includegraphics[width=.18\columnwidth]{figures/frankfurt00000_008206_gt.png}
  }
  \subfigure{%
    \includegraphics[width=.18\columnwidth]{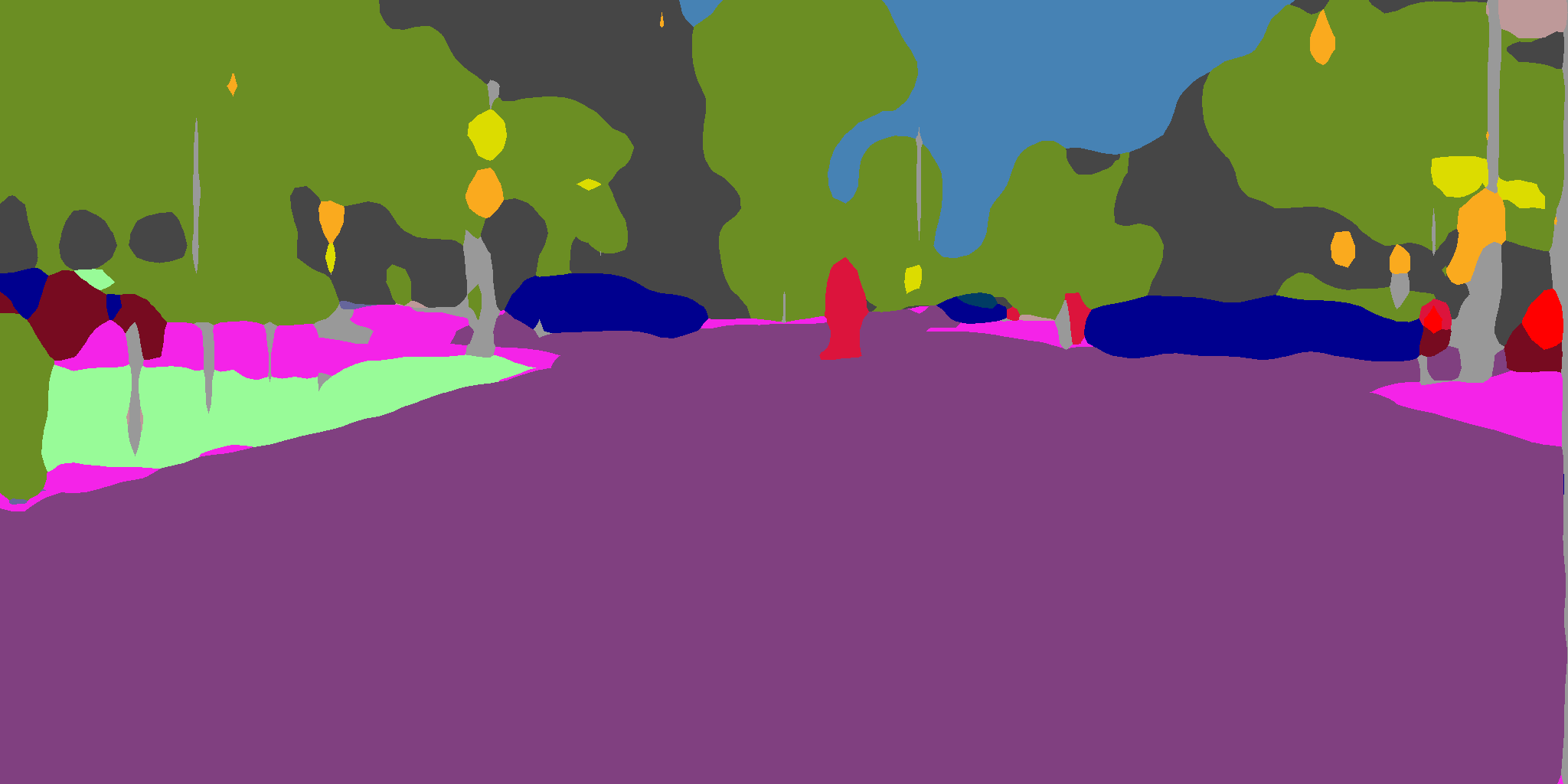}
  }
  \subfigure{%
    \includegraphics[width=.18\columnwidth]{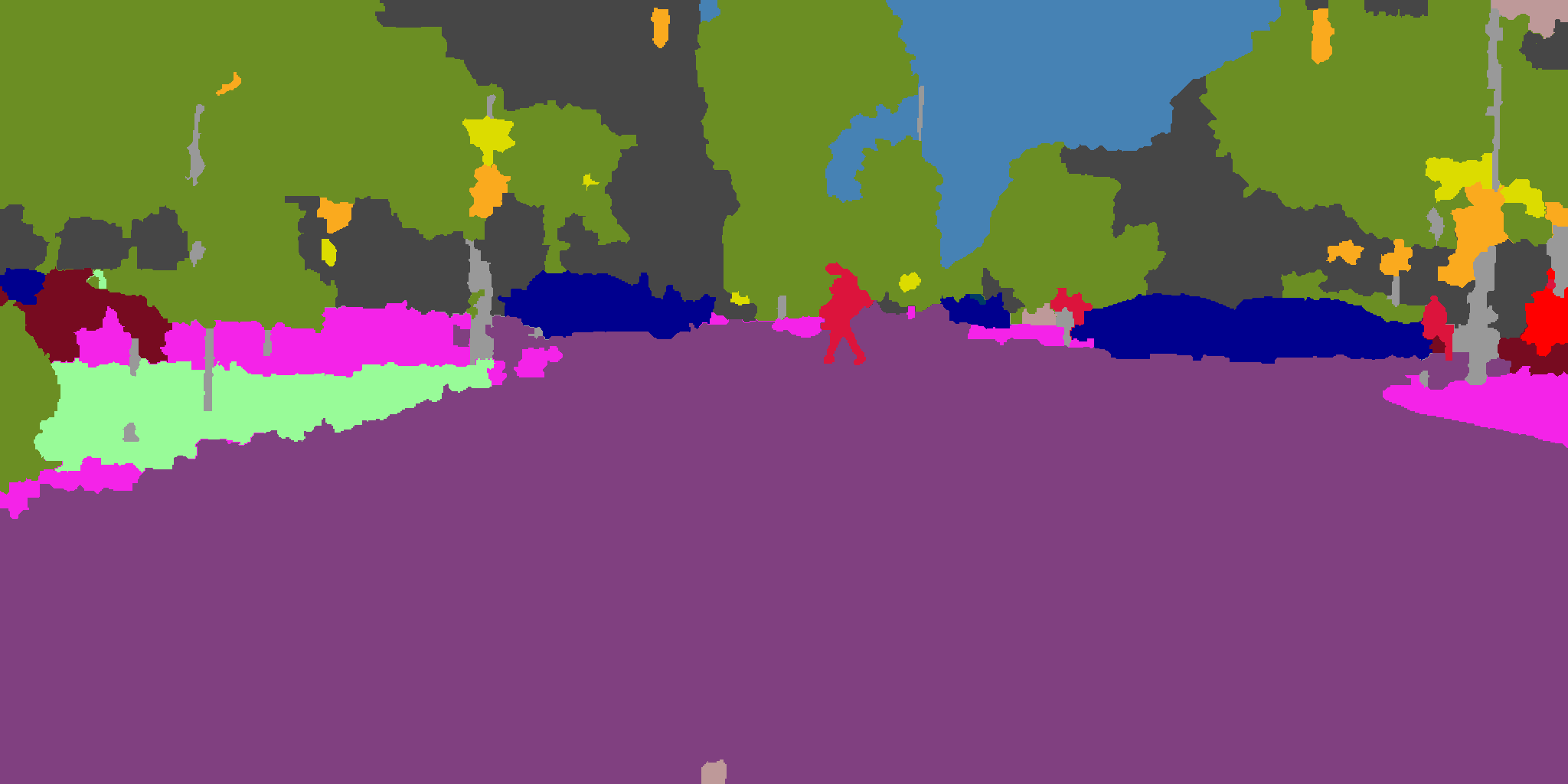}
  }\\[-2ex]
  \setcounter{subfigure}{0}
  \subfigure[\scriptsize Input]{%
    \includegraphics[width=.18\columnwidth]{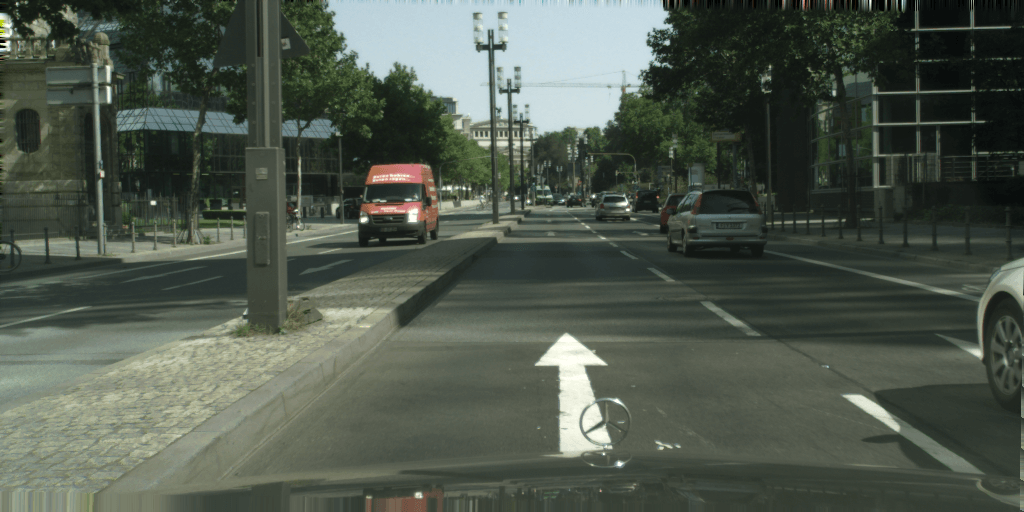}
  }
  \subfigure[\scriptsize Superpixels]{%
    \includegraphics[width=.18\columnwidth]{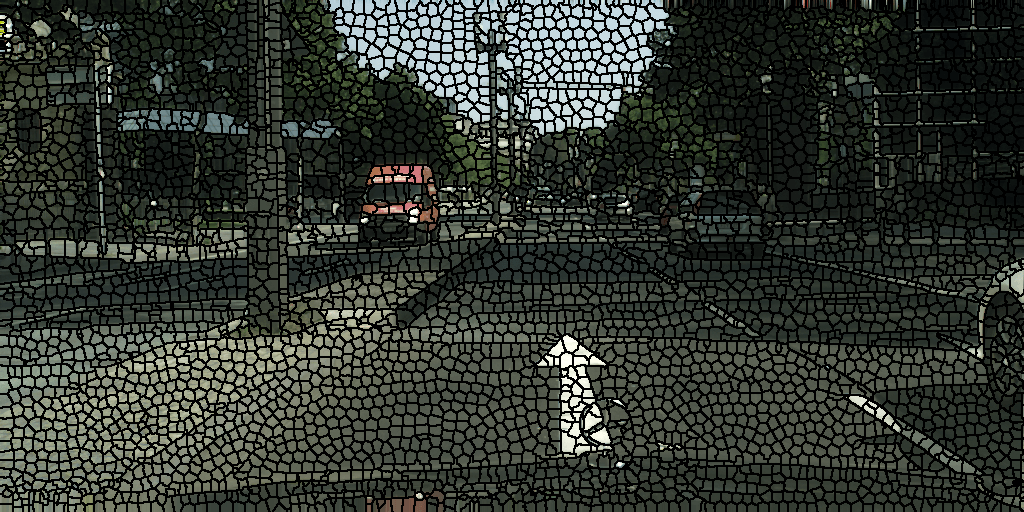}
  }
  \subfigure[\scriptsize GT]{%
    \includegraphics[width=.18\columnwidth]{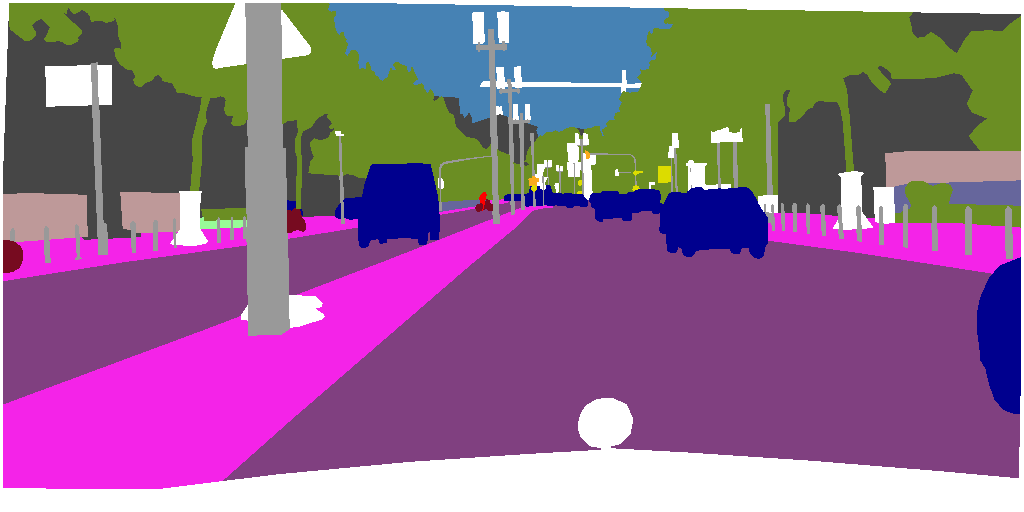}
  }
  \subfigure[\scriptsize Deeplab]{%
    \includegraphics[width=.18\columnwidth]{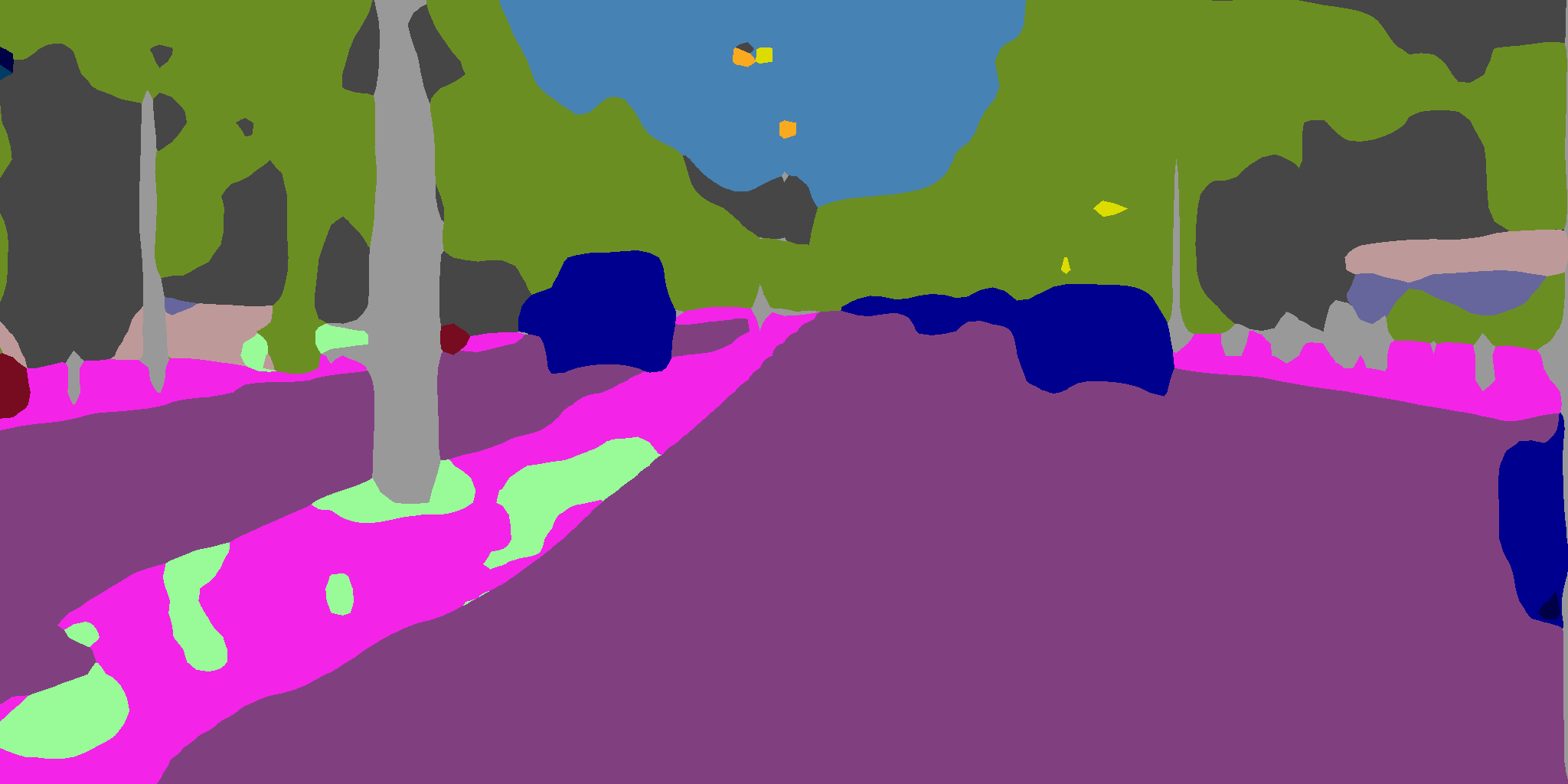}
  }
  \subfigure[\scriptsize Using BI]{%
    \includegraphics[width=.18\columnwidth]{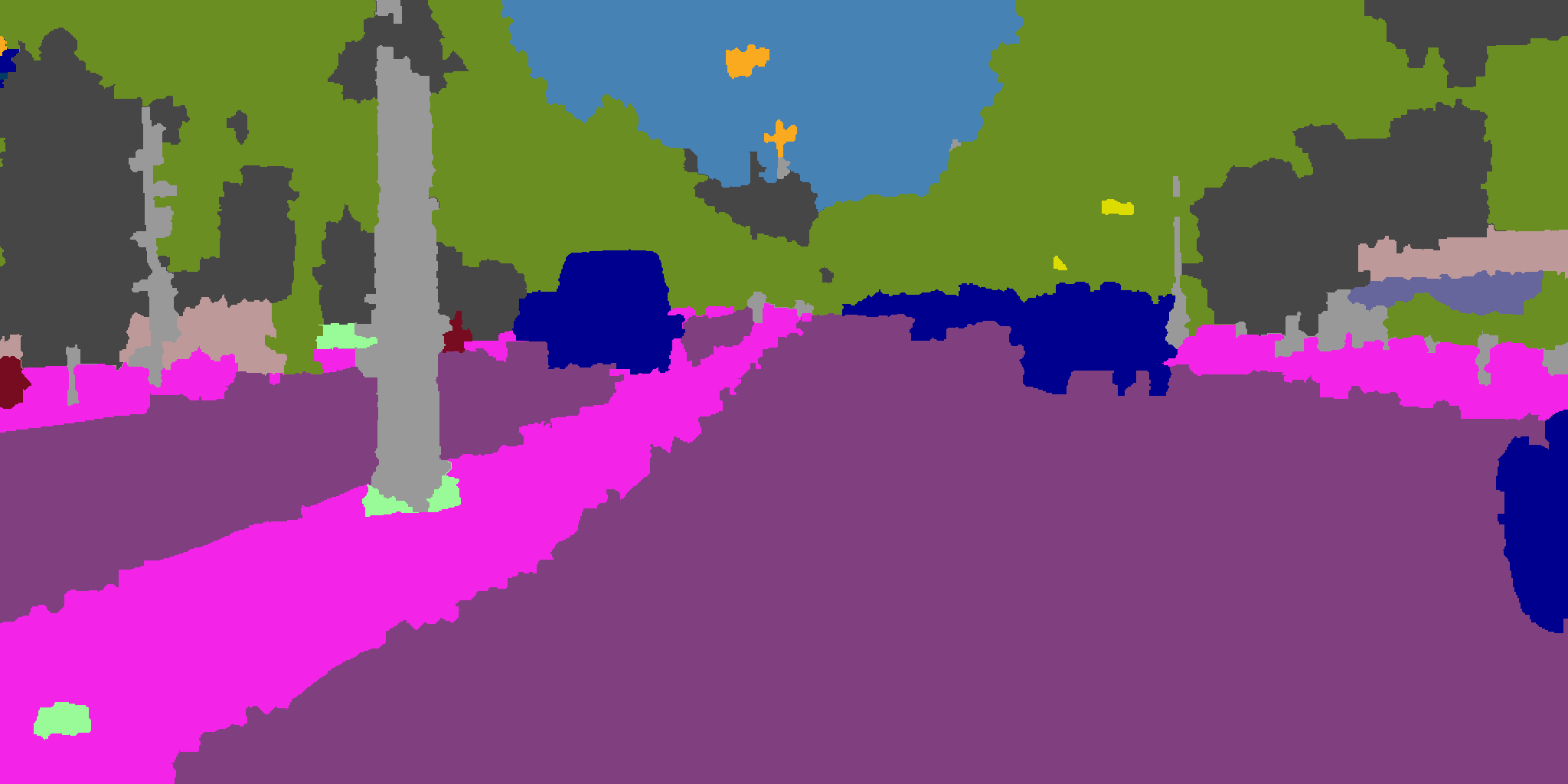}
  }
  \mycaption{Street scene segmentation}{Example results of street scene segmentation.
  (d) depicts the DeepLab results, (e) result obtained by adding bilateral inception (BI) modules (\bi{6}{2}+\bi{7}{6}) between \fc~layers.}
\label{fig:street_visuals}
\end{figure*}

We experimented with two layouts: only a single \bi{6}{2}
and one with two inception \bi{6}{2}-\bi{7}{6} modules. We notice that the
SLIC superpixels~\cite{achanta2012slic} give higher quantization error than on VOC
and thus used 6000 superpixels using~\cite{DollarICCV13edges} for our experiments.
Quantitative results on the validation set are
shown in Tab.~\ref{tab:cityscaperesults}. In contrast to the findings on the previous datasets, we only
observe modest improvements with both DenseCRF and our inception modules in comparison
to the base model. Similar to the previous experiments, the inception modules achieve
better performance than DenseCRF while being faster. The majority of the computation
time in our approach is due to the extraction of superpixels
($5.2s$) using a CPU implementation. Some visual results with \bi{6}{2}-\bi{7}{6} model are shown in Fig.~\ref{fig:street_visuals} with more in Appendix~\ref{sec:qualitative-app}.

\section{Discussion and Conclusions}
The DenseCRF~\cite{krahenbuhl2012efficient} with mean field inference has been used in many CNN segmentation approaches.
Its main ingredient and reason for the improved performance is the use of a bilateral filter applied to the beliefs over labels. We have introduced a CNN approach that uses this key component in a novel way: filtering intermediate representations of higher levels in CNNs while jointly learning the task-specific feature spaces.
This propagates information between earlier and more detailed intermediate representations of the classes instead of beliefs over labels.
Further we show that image adaptive layouts in the higher levels of CNNs can be used to an advantage in the same spirit as CRF graphs have been constructed using superpixels in previous works on semantic segmentation.
The computations in the $1\times1$ convolution layers scales in the number of superpixels which may be an advantage. Further we have shown that the same representation can be used to interpolate the coarser representations to the full image.

The use of image-adaptive convolutions in between the FC layers retains the appealing effect of producing segmentation masks with sharp edges.
This is not a property of the superpixels, using them to represent information in FC layers and their use to interpolate to the full resolution are orthogonal.
Different interpolation steps can be used to propagate the label information to the entire image, including bilinear interpolation, bilateral upsampling, up-convolutions and DenseCRFs.
We plan to investigate the effect of different sampling strategies to represent information in the higher layers of CNNs and apply similar image-adaptive ideas to videos.

We believe that the Bilateral Inception models are an interesting step that aims to directly include the model structure of CRF factors into the forward architecture of CNNs.
The BI modules are easy to implement and are applicable to CNNs that perform structured output prediction.

\chapter{Conclusions and Outlook}
\label{chap:conclusion}

Generative models provide a strong set of tools for modeling the vision world
around us. But their use in computer vision is hampered by the complexity
of inference in them causing the vision community to favor data-hungry discriminative
models. Both generative and discriminative models have complementary advantages and
disadvantages as discussed in Chapter~\ref{chap:models}. Generative models provide
an easy handle for incorporating prior knowledge about the task but inference
is often too complex in them. Discriminative models, on the other hand, have
a straightforward inference scheme as forward evaluation of models, but lack
principled ways of incorporating prior knowledge into them.

This thesis work proposed techniques for alleviating some of the key issues
with prominent computer vision models by improving inference in them.
A common strategy that is followed across several techniques proposed in this
thesis is leveraging the complementary models for better inference in a given
model. That is, we leverage discriminative models for better inference in
generative computer vision models. And we used generative knowledge (in the form
of bilateral filters) and enriched the existing discriminative CNN models.
This way, this thesis made important steps in bridging the gap between
generative and discriminative vision models. The proposed inference techniques
are flexible enough to deal with different task scenarios (e.g.,\ availability
of large or small amounts of data).

\paragraph{Inference in Generative Vision Models} In the case of generative
models, we leverage discriminative clustering or random forests techniques
to accelerate and/or to improve the Bayesian inference. In Chapter~\ref{chap:infsampler},
we proposed a new sampling technique called `Informed Sampler', where discriminative models
help in better exploration of target domain, via \emph{informed} proposals,
while doing MCMC sampling. In Chapter~\ref{chap:cmp}, we proposed a new message passing technique called
`Consensus Message Passing' where random forest predictors are used for predicting
\emph{consensus} messages during standard message passing inference resulting in
convergence to better solutions.

In both `Informed Sampler' and `Consensus Message
Passing' (CMP), we made sure that the theoretical guarantees that come with
the well established inference techniques are not violated with our modified
inference schemes. In the informed sampler, we achieve this by injecting
discriminative knowledge in MCMC sampling via proposal distributions and
adhering to detailed balance condition while sampling. And in consensus
message passing, we used consensus messages from discriminative predictors only
during the first few iterations ensuring that the fixed point reached by our
modified inference is also a fixed point of standard message passing in the
model. We evaluated both the informed sampler and CMP techniques on three
different generative models each, reflecting a wide range of problem scenarios,
where we consistently observed improved inference with the proposed techniques
in comparison to standard sampling and message passing inference techniques.

\paragraph{Inference in Discriminative Vision Models} In this thesis, we focus
on the inference in prominent CNN models. Spatial convolutions form the basic
building block of most CNN architectures.
A key observation in this thesis work is that the bilateral
filters~\cite{aurich1995non,tomasi1998bilateral}
are a generalization of spatial convolutions and do not have many of the limitations
that spatial convolutions have. The key issue with the existing use of bilateral
filters is they are confined to fixed hand-tuned parameterization.

In Chapter~\ref{chap:bnn}, we proposed a generalized bilateral filter and devised
a gradient based technique for
learning the filter parameters. Experiments on wide range of problems showed the
superior performance of learnable bilateral filters with respect to using Gaussian
filter kernel. Learnable bilateral filters enabled us to stack several filters
together and learn all of them via back-propagation. Using this, we proposed
novel neural network architectures which we call `Bilateral Neural Networks' (BNN).

In Chapter~\ref{chap:vpn}, we showed how BNNs can be easily adapted to filter video
data for propagating temporal information across video frames.
In Chapter~\ref{chap:binception}, we proposed new and fast neural network modules, based
on explicit Gaussian bilateral filtering called `Bilateral Inceptions' and showcased
how we can modify existing segmentation CNN architectures for big improvements in
accuracy while adding little time overhead.

Bilateral filters form the core of mean-field
inference in DenseCRF models~\cite{krahenbuhl2012efficient} and provide a way to incorporate
prior knowledge about the scene in the form of dense feature-based connectivity
across the pixels. By integrating learnable bilateral filters into standard CNN architectures,
we brought the worlds of CRF and CNN closer, providing a way to incorporate prior
knowledge into CNNs.

\section{Summary of Contributions}

The following list summarizes the specific contributions of this thesis work:

\begin{itemize}
\item We devised a novel MCMC sampling approach called `Informed Sampler' (Chapter~\ref{chap:infsampler})
for doing Bayesian inference in complex generative vision models.
The Informed sampler leverages discriminative approaches
for improving the efficiency of MCMC sampling. Experiments on a wide range of generative vision
models showed significantly faster convergence of our sampler while maintaining higher acceptance rates.
This opens up possibilities for using complex generative models like graphics engines for addressing vision problems.

\item We devised a novel message passing technique called `Consensus Message Passing’ (CMP),
in Chapter~\ref{chap:cmp}, for doing Bayesian inference in layered graphical models used in vision.
Experiments on diverse
graphical models in vision showed that CMP resulted in significantly better performance compared
to standard message passing techniques such as expectation propagation or variational message passing.
Moreover, CMP is the first instance where the Infer.NET~\cite{InferNET2012} probabilistic programming language is shown to
be useful for addressing vision problems.

\item We parameterized bilateral filters as general sparse high dimensional filters (Chapter~\ref{chap:bnn}) and
devised an approach for learning the filter kernels via standard back-propagation learning
techniques. This resulted in a general technique for learning sparse high-dimensional filters,
which in turn resulted in a generalization of standard bilateral filters. Experiments on wide
range of applications showed improved performance with respect to standard Gaussian bilateral filters.

\item We also show how learning bilateral filters can generalize the fully-connected conditional
random field models (DenseCRF) to arbitrarily learned pairwise potentials (Section~\ref{sec:densecrf})
instead of standard Gaussian pairwise edge potentials.
DenseCRF is one of the widely used CRF techniques in vision and this generalization
carries forward to most of its existing applications and helps in better integration into end-to-end
trained models like convolutional neural networks (CNN).

\item Our technique for learning general sparse high dimensional filters also generalizes standard
spatial convolutions in CNN frameworks. This opens up possibilities for applying CNNs to
sparse high-dimensional data, which is not feasible with many standard CNN techniques. Moreover,
our technique can be used for learning image-adaptive filters inside CNNs instead of standard
image-agnostic 2D filters.

\item We adapted the learnable sparse high dimensional filters for video filtering, in Chapter~\ref{chap:vpn},
and proposed a novel neural network approach for propagating content across video frames.
We call our networks `Video Propagation Networks'~(VPN).
Experiments on video object segmentation and semantic video segmentation showed that VPN
outperformed existing task-specific methods while being faster.

\item In Chapter~\ref{chap:binception}, we devised a new CNN module called `Bilateral Inception'
that can be readily inserted into standard segmentation CNN models resulting in better performance
while producing the result at original image resolution and also alleviating some of the need for
post-processing. Experiments on state-of-the-art CNN models resulted in significantly better
performance with our inception module while being competitive in time.
\end{itemize}

\section{Outlook}

With the diverse range of experiments on each proposed technique, I hope to have
convinced the readers that this thesis work resulted in several important advances
for inference in computer vision models.
Inference in computer vision models is still a very active area of research
and I hope the work presented in this thesis aids in further advances in this
area of research. The following are some of the research topics that could benefit
from the concepts and techniques presented in this thesis.

\vspace{-0.2cm}
\paragraph{Leveraging Photo-Realistic Graphics for Vision:} As discussed in
Section~\ref{sec:inv-graphics}, modern graphics engines leverage dedicated hardware
setups and provide real-time renderings with stunning level of realism.
This is made possible with accurate modeling of image
formation. Such realistic graphics models are seldom utilized in building vision
systems due to the difficulty in posterior inference. The informed sampler
(Chapter~\ref{chap:infsampler}) and consensus message passing (Chapter~\ref{chap:cmp})
techniques presented in this thesis made important steps in making the inference
faster. But the proposed techniques are still not fast enough
for practical purposes. An important research direction is to further improve the
efficiency of inference by making use of the internals of state­-of­-the-­art rendering
mechanisms such as ‘Metropolis Light Transport’~\cite{vorba2014line}.
One way to achieve this is by improving MCMC sampling efficiency
via pre­-rejection using the rejection technique of~\cite{korattikara2013austerity},
without the need for doing full rendering for the rejected samples in MCMC.
This would result in a better coupling of graphics and Bayesian vision systems.

\vspace{-0.2cm}
\paragraph{CNNs for Sparse 3D Data:}
In recent years, there is an increasing need for techniques that efficiently
process 3D data, with the onset of cheaper 3D scanners and the consumer devices
that requires processing 3D data (e.g.~virtual reality devices such as Oculus Rift~\cite{oculus}).
One of the distinguishing characters of high-dimensional data such as 3D point clouds
or videos is that they are sparse. 3D point clouds are inherently sparse and the sparsity
of video data comes from the redundant representation across frames.
Learnable bilateral filters proposed in this thesis
(Chapter~\ref{chap:bnn}), provide a principled way to process sparse high-dimensional
data by enabling long-range data dependent connections. This thesis work also demonstrated
that one could easily integrate these sparse high-dimensional filters into other
CNN architectures and are also shown to be fruitful for video processing (Chapter~\ref{chap:vpn}).
I hope this thesis work inspires future research work for efficiently processing
3D data such as point clouds or meshes. Since the convolutions are performed in
bilateral space with sparsely populated cells, there is no need to voxelize a given
point cloud or meshes for doing 3D convolutions.
One of the limitations of bilateral neural networks
is that the bilateral feature scales are hand-­tuned.
There are other recent works such as~\cite{krahenbuhl2013parameter,kundu2016feature}
trying to optimize feature spaces for bilateral filtering but are not integrated into CNN frameworks.
An interesting future research direction is to bring these works together by learning feature spaces
for bilateral neural networks in an end-­to-­end fashion.

\vspace{-0.2cm}
\paragraph{Structured CNNs:} A​lthough images are represented with independent pixel values,
real­ world is highly structured and prior knowledge therein can be captured by rich
modeling tools like graphical models. For example, in the case of urban scene
understanding, one could leverage the prior knowledge that objects like cars and persons
stand on the ground plane; building facades are vertical etc. Most of the existing CNN
frameworks are agnostic to such explicit prior information. One may argue that the CNNs
are capable of implicitly learning such prior knowledge directly from the training data.
But, in practice, the amount of labelled training data is always limited and thus CNNs
need external prior constraints to be able to perform well on several real­ world problems
especially when the training data is very limited. In this thesis, we developed techniques
for incorporating prior knowledge into CNNs via learnable bilateral filters. Such prior
knowledge is low-level and, ideally I would like to have techniques for integrating
high-level prior knowledge (e.g., the prior knowledge that is encoded in
graphical models). Other existing works in this direction
(e.g.~\cite{jain2015structural,zheng2015conditional,tompson2014joint,chandra2016fast})
are also either limited in the type of prior knowledge they model and/or graphical
model constraints are generally enforced after the main CNN structure.
I believe that a fruitful direction to tackle this problem is developing structured
filters that can be easily integrated into CNN frameworks. The work presented in
this thesis make important steps in this direction.

\vspace{-0.2cm}
\paragraph{On Combining Generative and Discriminative Models:} Hybrid generative
and discriminative models are an active area of research and I believe that the
future of computer vision would be dominated by such hybrid models which are
scalable and generic to be applicable to a wide range of problem scenarios.
In Section~\ref{sec:gen-disc-comb}, we discussed several recent works that aim
to develop such hybrid models. Several inference techniques presented in this
thesis are based on the synergistic combinations of generative and discriminative
models. I hope that this thesis work inspires or aids in the further development
of hybrid generative and discriminative vision models.

\appendix
\chapter{Supplementary Material}
\label{appendix}

In this appendix, we present supplementary material for the techniques and
experiments presented in the main text.

\section{Baseline Results and Analysis for Informed Sampler}
\label{appendix:chap3}

Here, we give an in-depth
performance analysis of the various samplers and the effect of their
hyperparameters. We choose hyperparameters with the lowest PSRF value
after $10k$ iterations, for each sampler individually. If the
differences between PSRF are not significantly different among
multiple values, we choose the one that has the highest acceptance
rate.

\subsection{Experiment: Estimating Camera Extrinsics}
\label{appendix:chap3:room}

\subsubsection{Parameter Selection}
\paragraph{Metropolis Hastings (\MH)}

\begin{figure}[h]
\centering
  \subfigure[MH]{%
    \includegraphics[width=.48\textwidth]{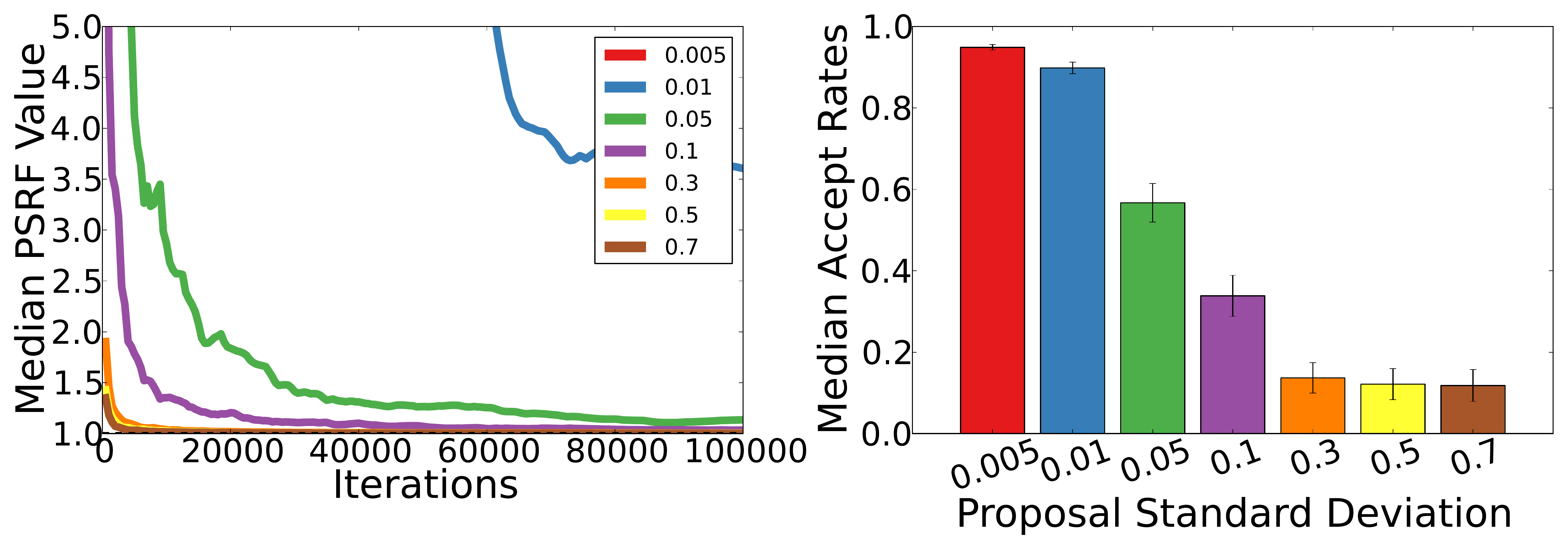} \label{fig:exp1_MH}
  }
  \subfigure[PT]{%
    \includegraphics[width=.48\textwidth]{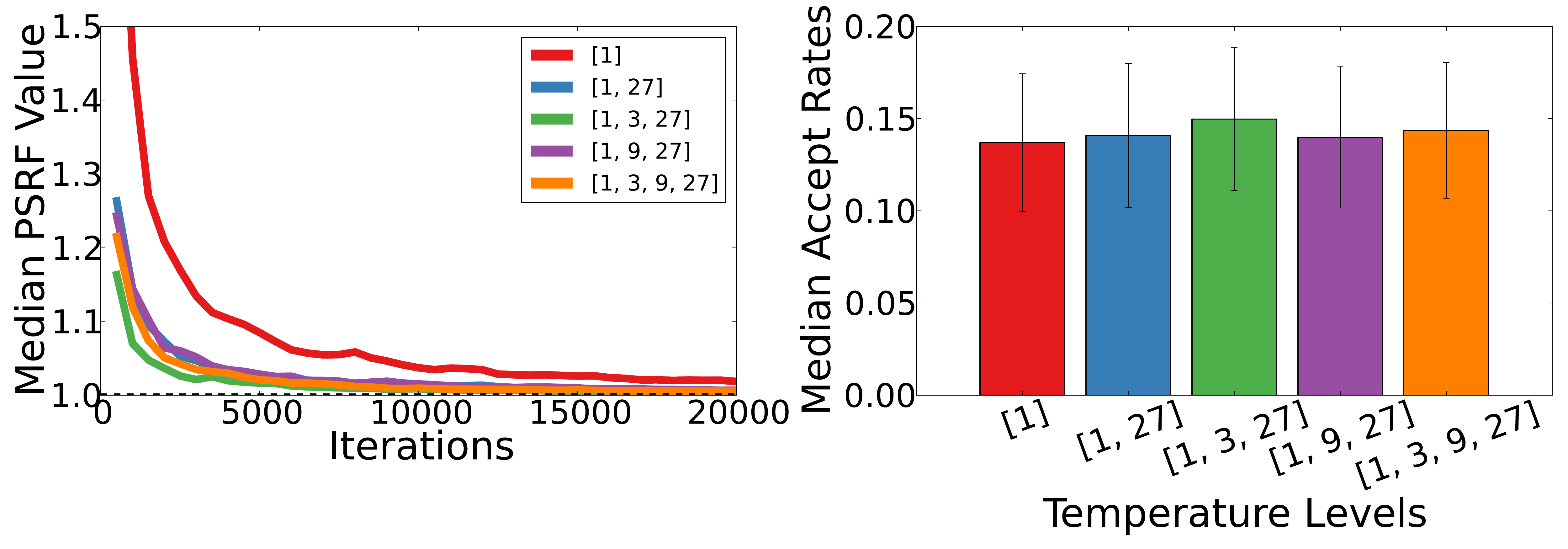} \label{fig:exp1_PT}
  }
\\
  \subfigure[INF-MH]{%
    \includegraphics[width=.48\textwidth]{figures/supplementary/camPose_alpha.pdf} \label{fig:exp1_alpha}
  }
  \mycaption{Results of the `Estimating Camera Extrinsics' experiment}{PRSFs and Acceptance rates corresponding to (a) various standard deviations of \MH, (b) various temperature level combinations of \PT sampling and (c) various mixture coefficients of \MIXLMH sampling.}
\end{figure}

Figure~\ref{fig:exp1_MH} shows the median acceptance rates and PSRF
values corresponding to various proposal standard deviations of plain
\MH~sampling. Mixing gets better and the acceptance rate gets worse as
the standard deviation increases. The value $0.3$ is selected standard
deviation for this sampler.

\paragraph{Metropolis Hastings Within Gibbs (\MHWG)}

As mentioned in Section~\ref{sec:room}, the \MHWG~sampler with one-dimensional
updates did not converge for any value of proposal standard deviation.
This problem has high correlation of the camera parameters and is of
multi-modal nature, which this sampler has problems with.

\paragraph{Parallel Tempering (\PT)}

For \PT~sampling, we took the best performing \MH~sampler and used
different temperature chains to improve the mixing of the
sampler. Figure~\ref{fig:exp1_PT} shows the results corresponding to
different combination of temperature levels. The sampler with
temperature levels of $[1,3,27]$ performed best in terms of both
mixing and acceptance rate.

\paragraph{Effect of Mixture Coefficient in Informed Sampling (\MIXLMH)}

Figure~\ref{fig:exp1_alpha} shows the effect of mixture
coefficient ($\alpha$) on the informed sampling
\MIXLMH. Since there is no significant different in PSRF values for
$0 \le \alpha \le 0.7$, we chose $0.7$ due to its high acceptance
rate.


\begin{figure}[!t]
\centering
  \subfigure[\MH]{%
    \includegraphics[width=.48\textwidth]{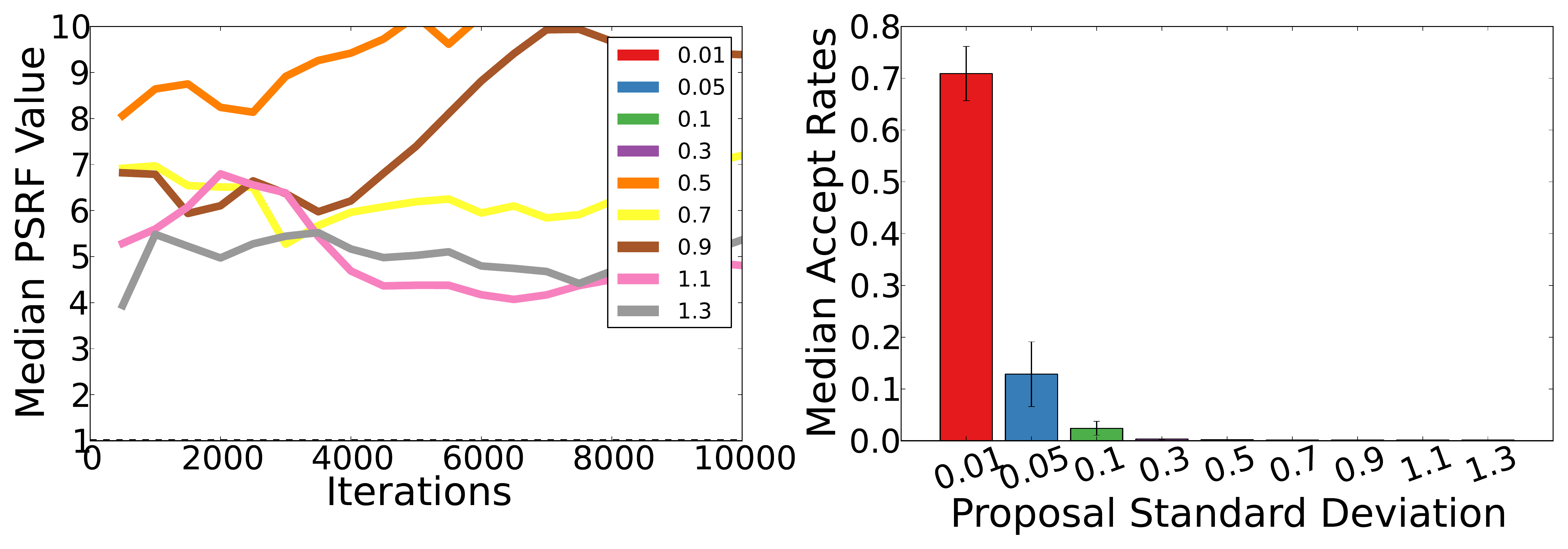} \label{fig:exp2_MH}
  }
  \subfigure[\BMHWG]{%
    \includegraphics[width=.48\textwidth]{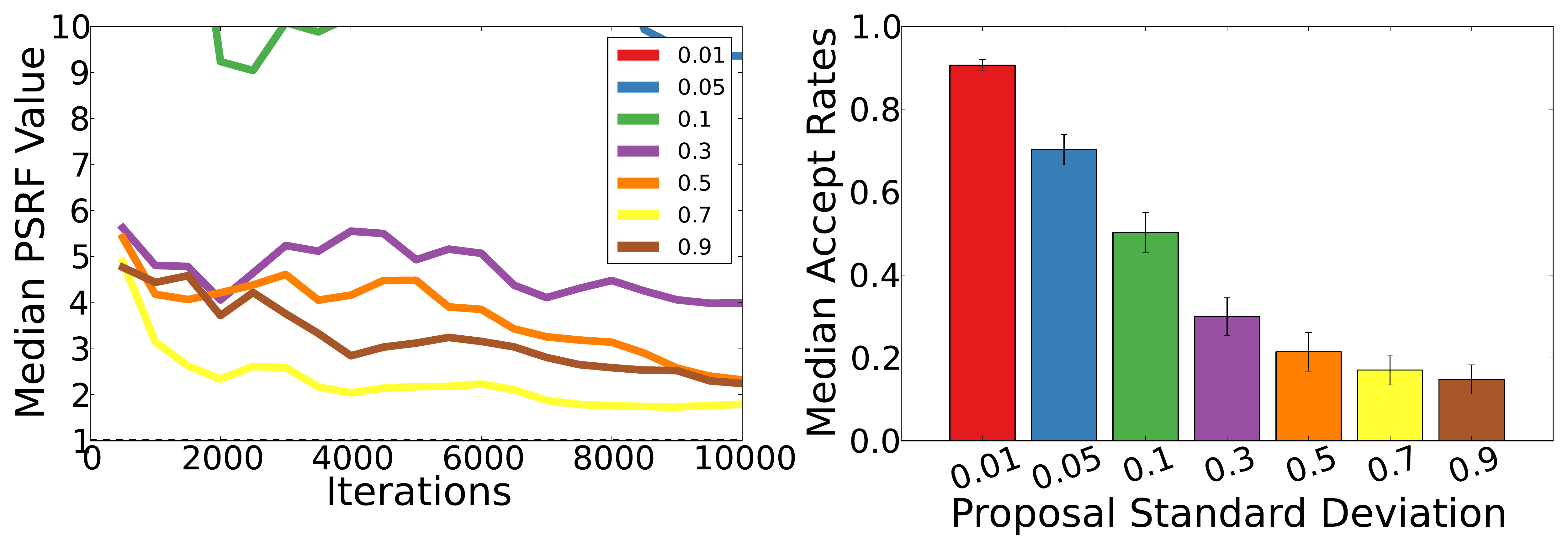} \label{fig:exp2_BMHWG}
  }
\\
  \subfigure[\MHWG]{%
    \includegraphics[width=.48\textwidth]{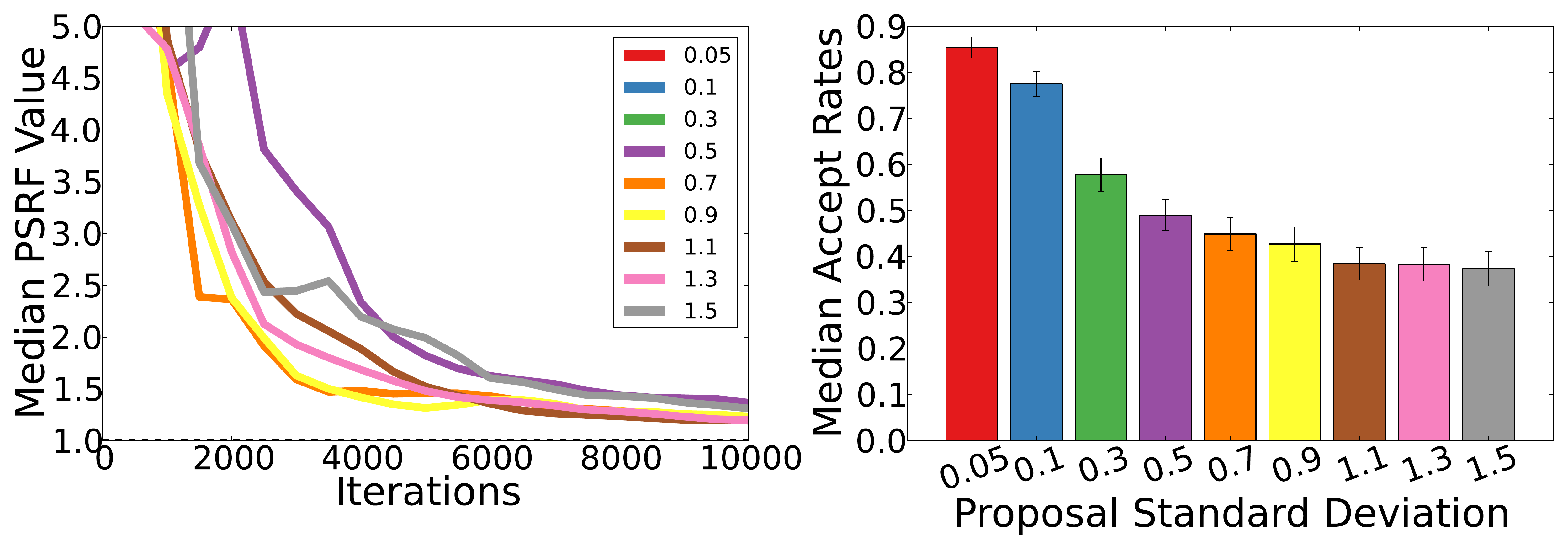} \label{fig:exp2_MHWG}
  }
  \subfigure[\PT]{%
    \includegraphics[width=.48\textwidth]{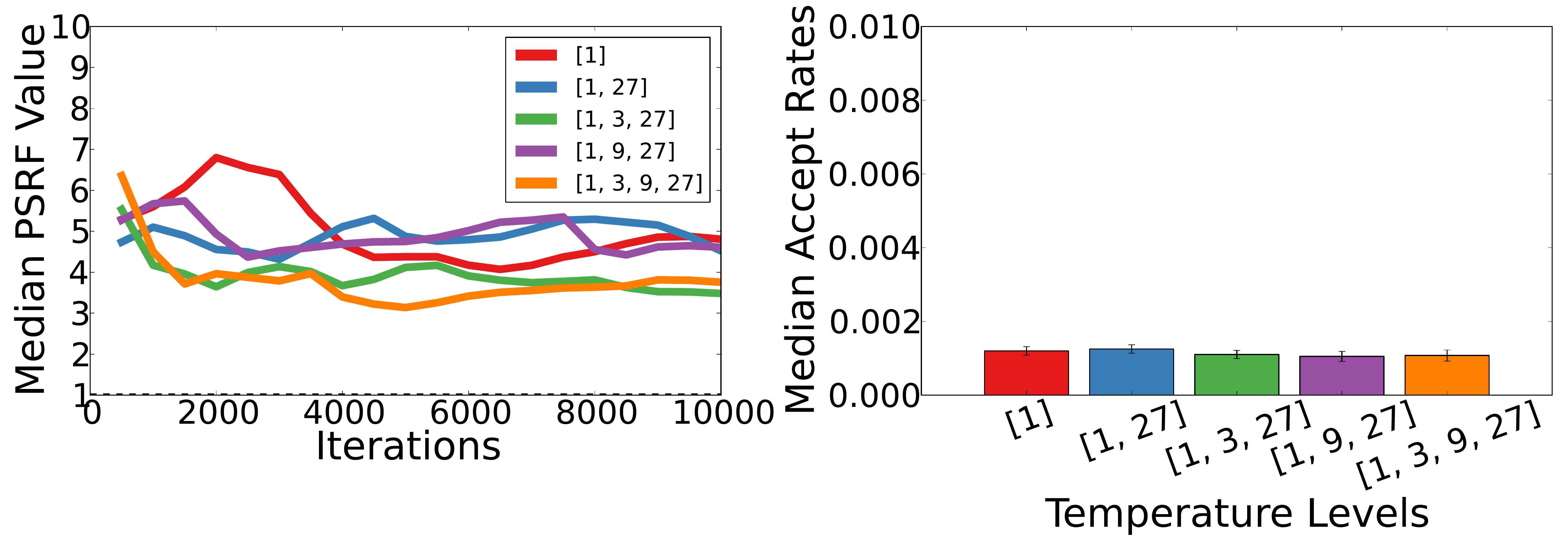} \label{fig:exp2_PT}
  }
\\
  \subfigure[\INFBMHWG]{%
    \includegraphics[width=.5\textwidth]{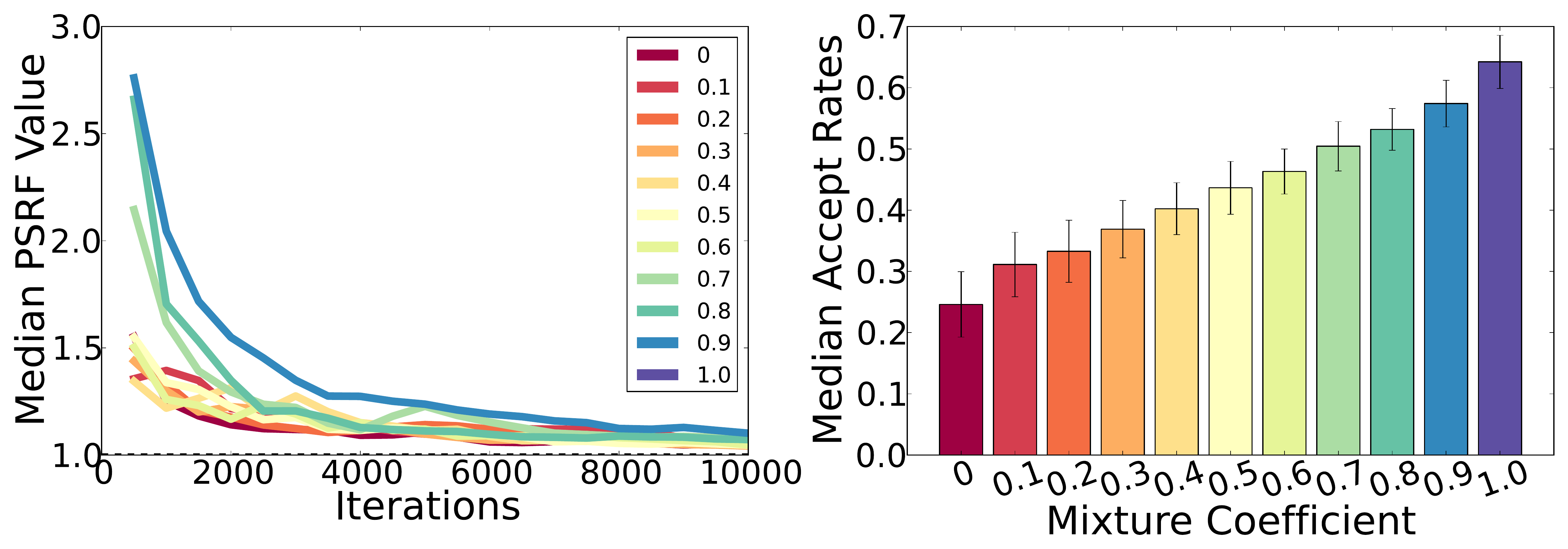} \label{fig:exp2_alpha}
  }
  \mycaption{Results of the `Occluding Tiles' experiment}{PRSF and
    Acceptance rates corresponding to various standard deviations of
    (a) \MH, (b) \BMHWG, (c) \MHWG, (d) various temperature level
    combinations of \PT~sampling and; (e) various mixture coefficients
    of our informed \INFBMHWG sampling.}
\end{figure}

\subsection{Experiment: Occluding Tiles}
\label{appendix:chap3:tiles}

\subsubsection{Parameter Selection}

\paragraph{Metropolis Hastings (\MH)}

Figure~\ref{fig:exp2_MH} shows the results of
\MH~sampling. Results show the poor convergence for all proposal
standard deviations and rapid decrease of AR with increasing standard
deviation. This is due to the high-dimensional nature of
the problem. We selected a standard deviation of $1.1$.

\paragraph{Blocked Metropolis Hastings Within Gibbs (\BMHWG)}

The results of \BMHWG are shown in Figure~\ref{fig:exp2_BMHWG}. In
this sampler we update only one block of tile variables (of dimension
four) in each sampling step. Results show much better performance
compared to plain \MH. The optimal proposal standard deviation for
this sampler is $0.7$.

\paragraph{Metropolis Hastings Within Gibbs (\MHWG)}

Figure~\ref{fig:exp2_MHWG} shows the result of \MHWG sampling. This
sampler is better than \BMHWG and converges much more quickly. Here
a standard deviation of $0.9$ is found to be best.

\paragraph{Parallel Tempering (\PT)}

Figure~\ref{fig:exp2_PT} shows the results of \PT sampling with various
temperature combinations. Results show no improvement in AR from plain
\MH sampling and again $[1,3,27]$ temperature levels are found to be optimal.

\paragraph{Effect of Mixture Coefficient in Informed Sampling (\INFBMHWG)}

Figure~\ref{fig:exp2_alpha} shows the effect of mixture
coefficient ($\alpha$) on the blocked informed sampling
\INFBMHWG. Since there is no significant different in PSRF values for
$0 \le \alpha \le 0.8$, we chose $0.8$ due to its high acceptance
rate.

\subsection{Experiment: Estimating Body Shape}
\label{appendix:chap3:body}

\subsubsection{Parameter Selection}
\paragraph{Metropolis Hastings (\MH)}

\begin{figure}[t]
\centering
  \subfigure[\MH]{%
    \includegraphics[width=.48\textwidth]{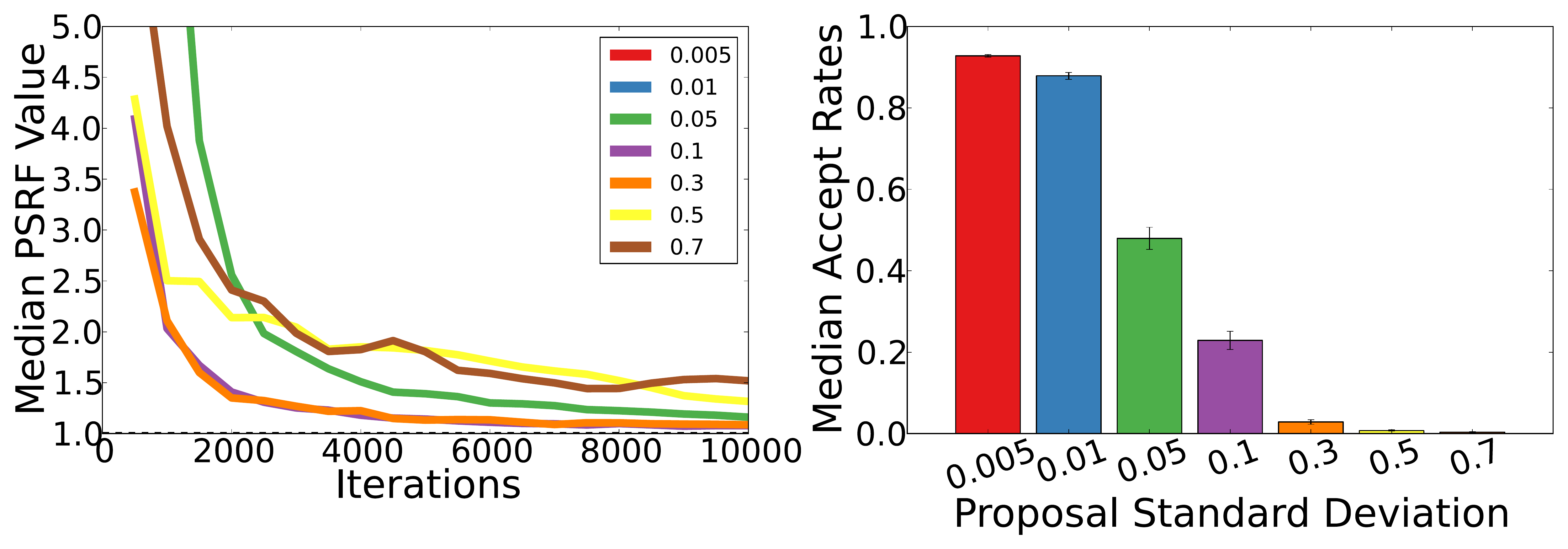} \label{fig:exp3_MH}
  }
  \subfigure[\MHWG]{%
    \includegraphics[width=.48\textwidth]{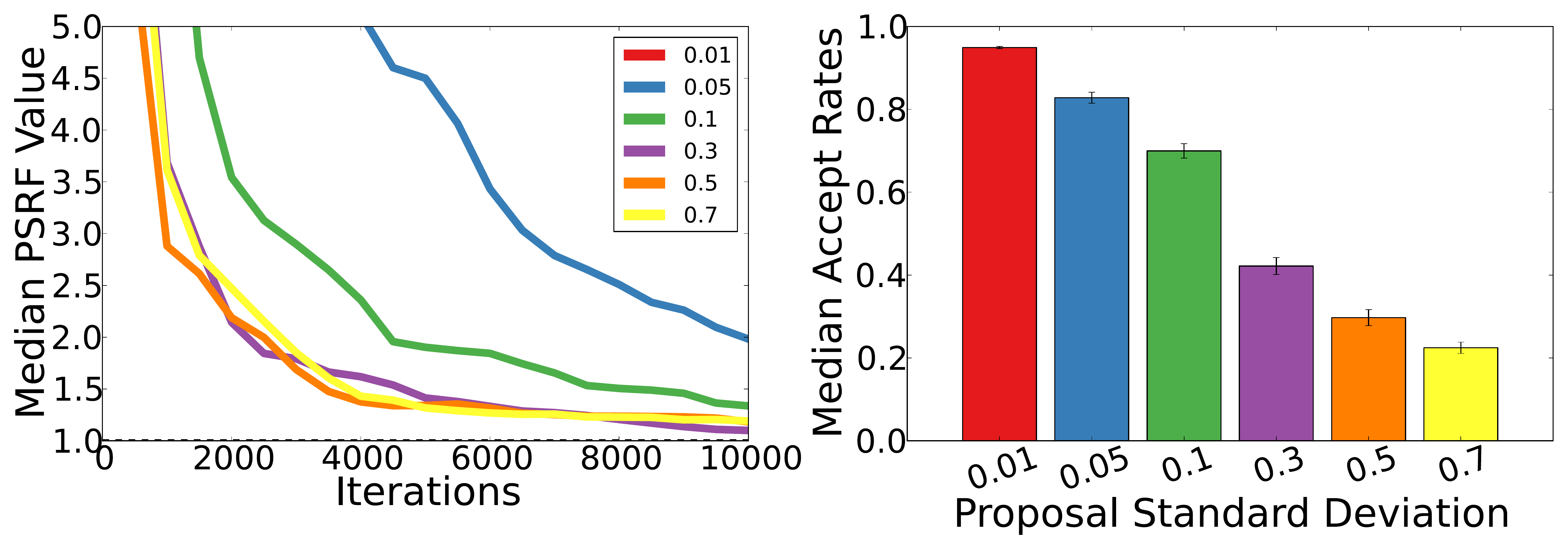} \label{fig:exp3_MHWG}
  }
\\
  \subfigure[\PT]{%
    \includegraphics[width=.48\textwidth]{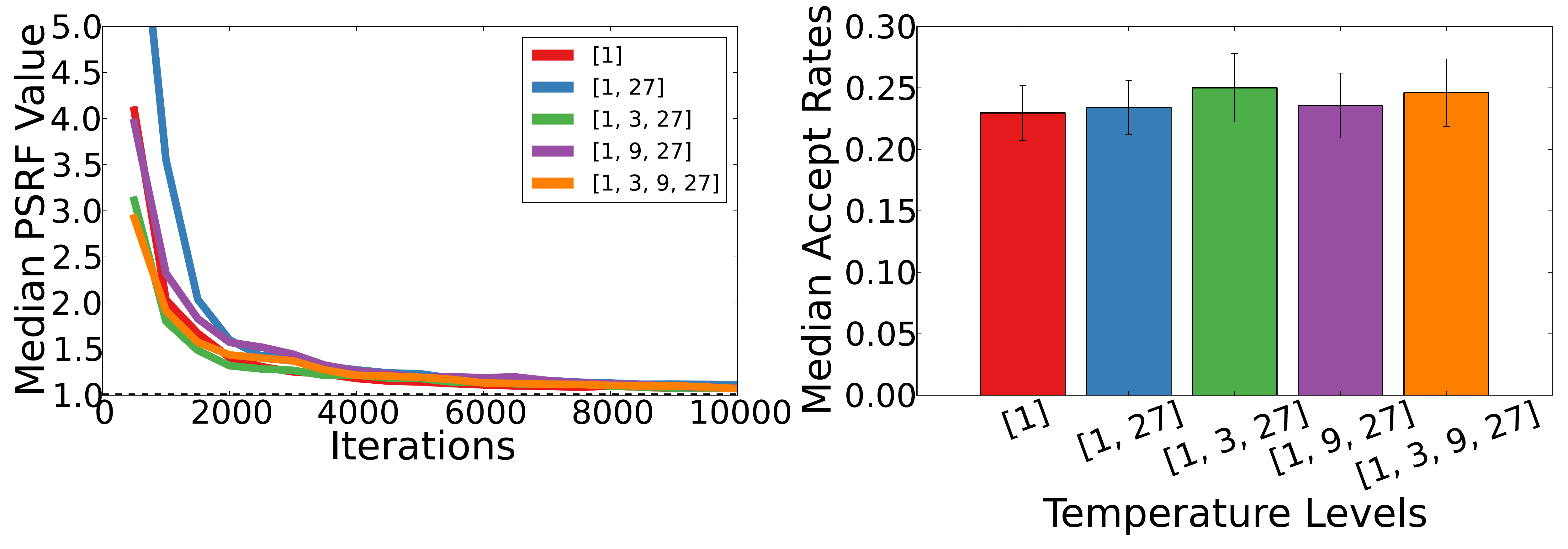} \label{fig:exp3_PT}
  }
  \subfigure[\MIXLMH]{%
    \includegraphics[width=.48\textwidth]{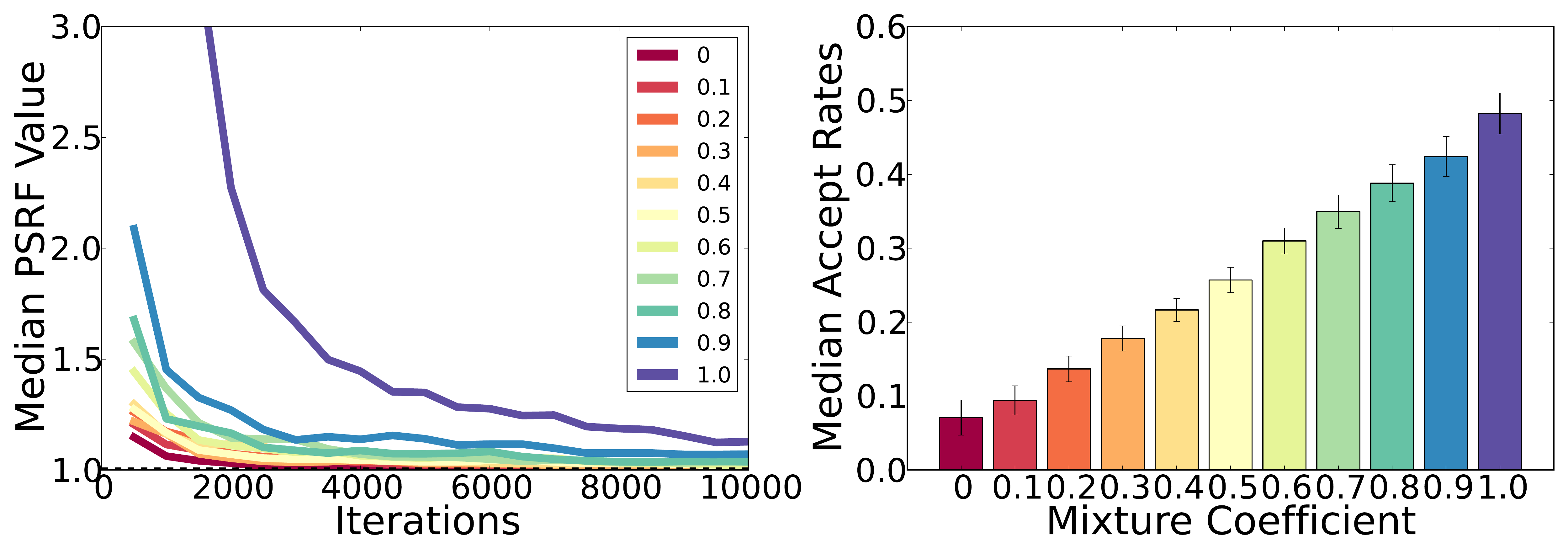} \label{fig:exp3_alpha}
  }
\\
  \mycaption{Results of the `Body Shape Estimation' experiment}{PRSFs and
    Acceptance rates corresponding to various standard deviations of
    (a) \MH, (b) \MHWG; (c) various temperature level combinations
    of \PT sampling and; (d) various mixture coefficients of the
    informed \MIXLMH sampling.}
\end{figure}

Figure~\ref{fig:exp3_MH} shows the result of \MH~sampling with various
proposal standard deviations. The value of $0.1$ is found to be
best.

\paragraph{Metropolis Hastings Within Gibbs (\MHWG)}

For \MHWG sampling we select $0.3$ proposal standard
deviation. Results are shown in Fig.~\ref{fig:exp3_MHWG}.

\paragraph{Parallel Tempering (\PT)}

As before, results in Fig.~\ref{fig:exp3_PT}, the temperature levels
were selected to be $[1,3,27]$ due its slightly higher AR.

\paragraph{Effect of Mixture Coefficient in Informed Sampling (\MIXLMH)}

Figure~\ref{fig:exp3_alpha} shows the effect of $\alpha$ on PSRF and
AR. Since there is no significant differences in PSRF values for $0 \le
\alpha \le 0.8$, we choose $0.8$.

\subsection{Results Overview}
\begin{figure*}[h!]
\centering
  \subfigure[Results for: Estimating Camera Extrinsics]{%
    \includegraphics[width=0.9\textwidth]{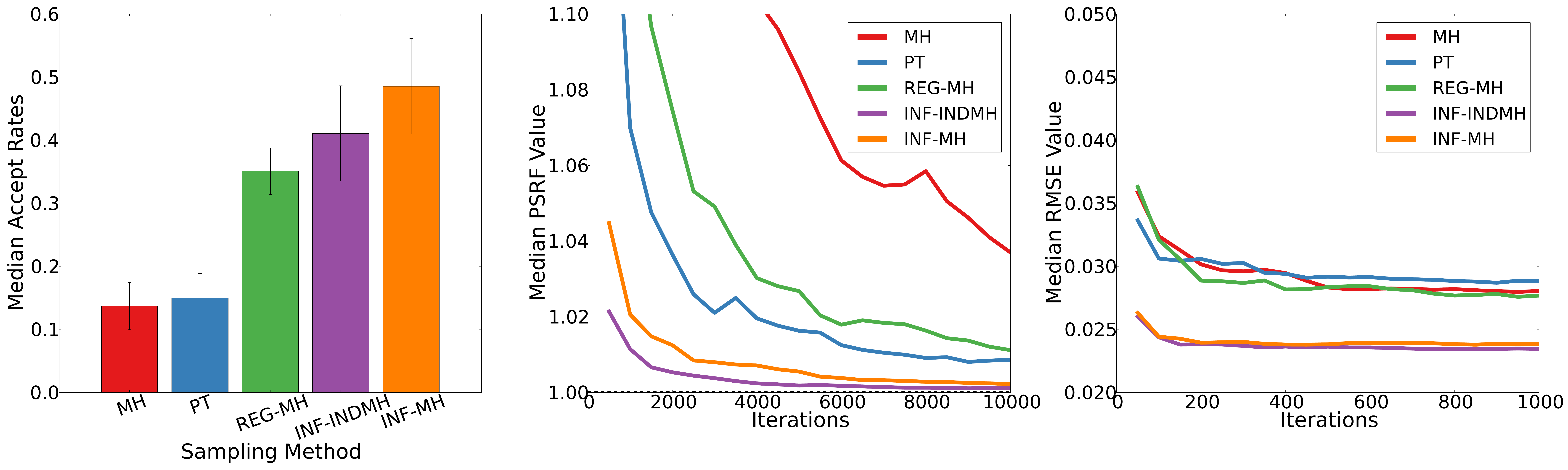} \label{fig:exp1_all}
  }
  \subfigure[Results for: Occluding Tiles]{%
    \includegraphics[width=0.9\textwidth]{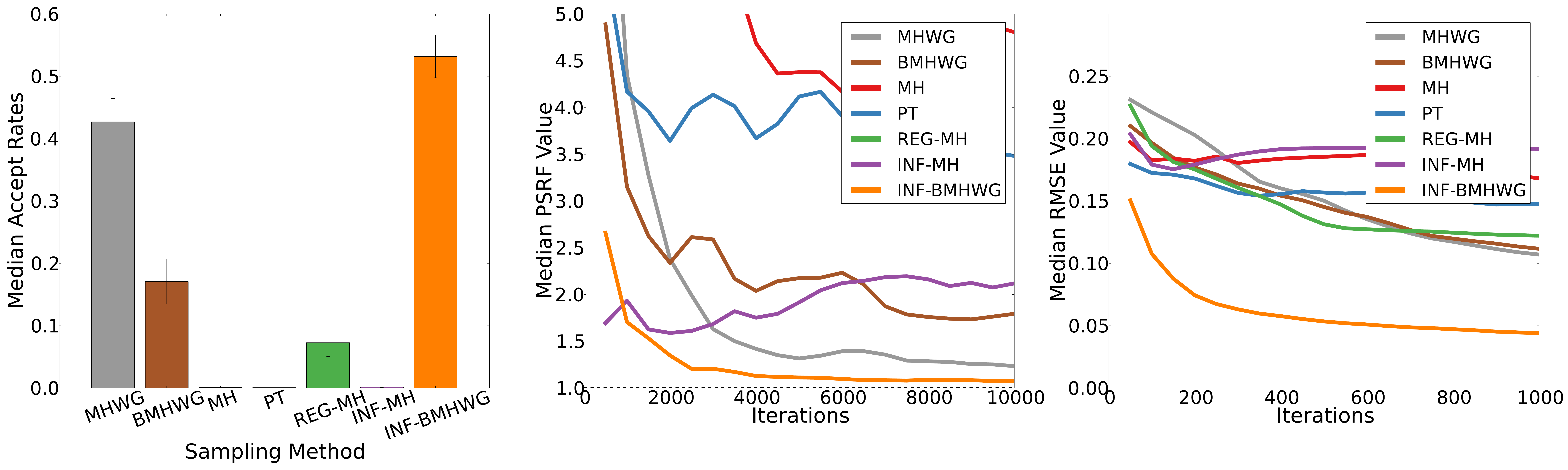} \label{fig:exp2_all}
  }
  \subfigure[Results for: Estimating Body Shape]{%
    \includegraphics[width=0.9\textwidth]{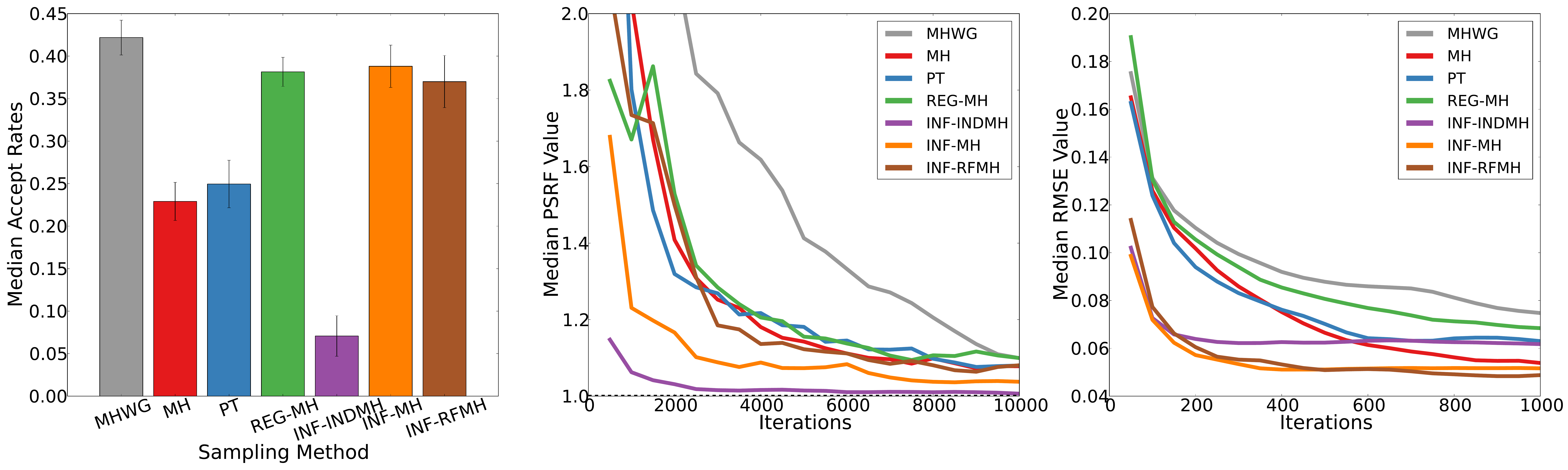} \label{fig:exp3_all}
  }
  \label{fig:exp_summary}
  \mycaption{Summary of the statistics for the three experiments}{Shown are
    for several baseline methods and the informed samplers the
    acceptance rates (left), PSRFs (middle), and RMSE values
    (right). All results are median results over multiple test
    examples.}
\end{figure*}

Figure~\ref{fig:exp_summary} shows the summary results of the all the three
experimental studies related to informed sampler.

\subsection{Additional Qualitative Results}

\subsubsection{Occluding Tiles}
\begin{figure*}[h!]
\begin{center}
\centerline{\includegraphics[width=0.95\textwidth]{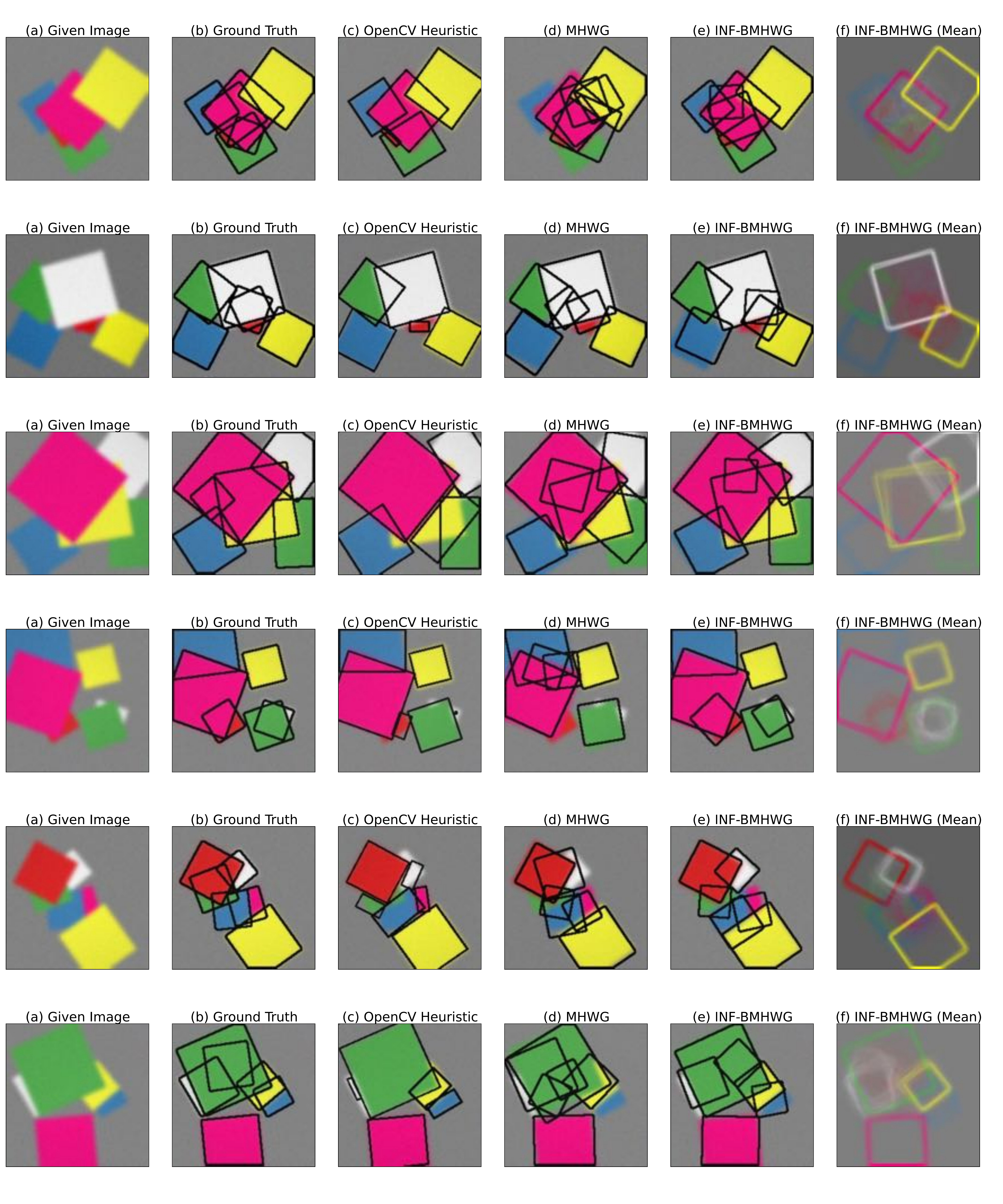}}
\mycaption{Additional qualitative results of the occluding tiles experiment}
  {From left to right: (a)
  Given image, (b) Ground truth tiles, (c) OpenCV heuristic and most probable estimates
  from 5000 samples obtained by (d) MHWG sampler (best baseline) and
  (e) our INF-BMHWG sampler. (f) Posterior expectation of the tiles
  boundaries obtained by INF-BMHWG sampling (First 2000 samples are
  discarded as burn-in).}
\label{fig:exp2_visual_more}
\end{center}
\end{figure*}

In Figure~\ref{fig:exp2_visual_more} more qualitative results of the
occluding tiles experiment are shown. The informed sampling approach
(\INFBMHWG) is better than the best baseline (\MHWG). This still is a
very challenging problem since the parameters for occluded tiles are
flat over a large region. Some of the posterior variance of the
occluded tiles is already captured by the informed sampler.

\subsubsection{Body Shape}
\begin{figure*}[t]
\begin{center}
\centerline{\includegraphics[width=0.75\textwidth]{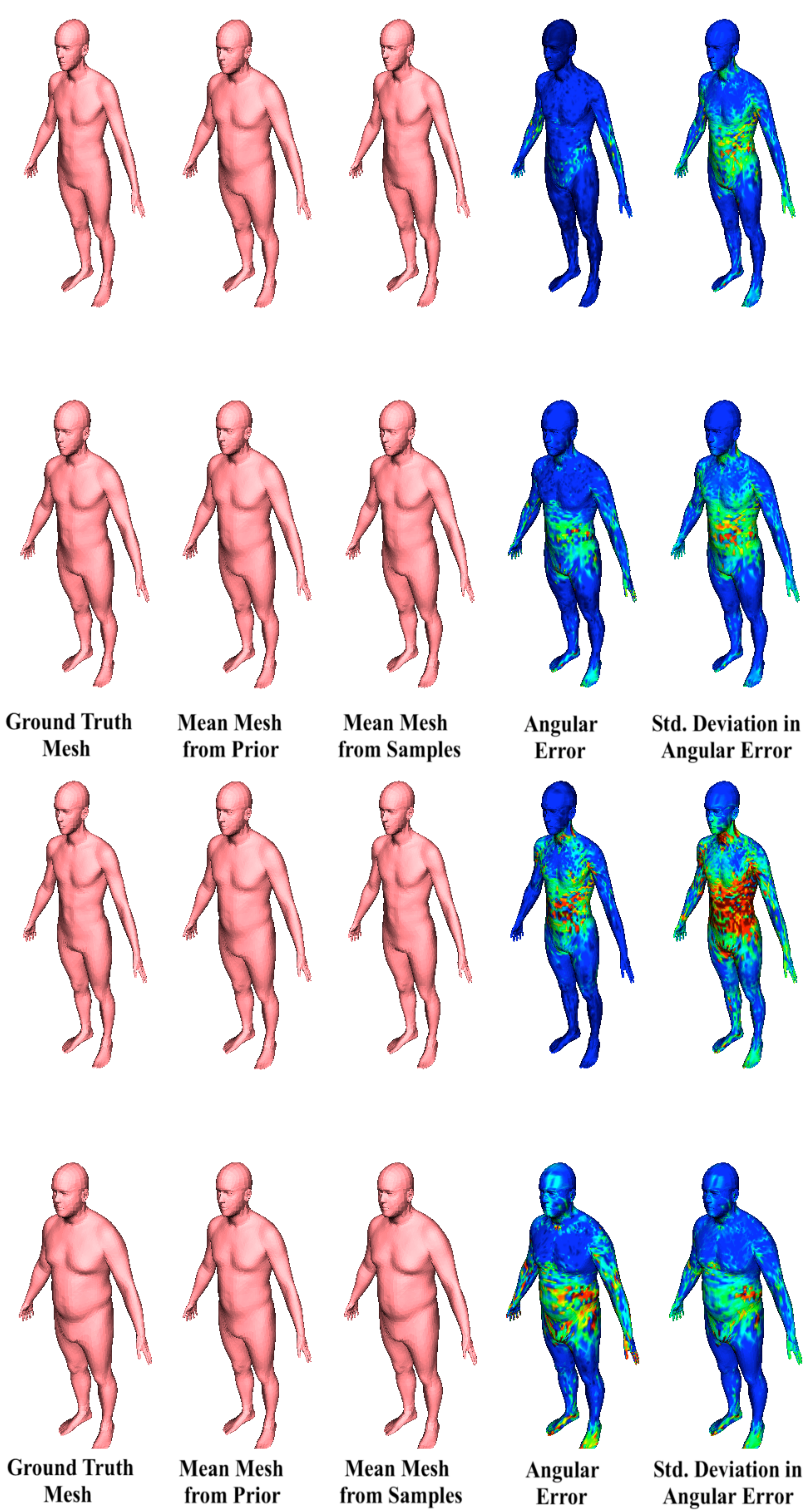}}
\mycaption{Qualitative results for the body shape experiment}
  {Shown is the 3D mesh reconstruction results with first 1000 samples obtained
  using the \MIXLMH informed sampling method. (blue indicates small
  values and red indicates high values)}
\label{fig:exp3_bodyMeshes}
\end{center}
\end{figure*}

Figure~\ref{fig:exp3_bodyMeshes} shows some more results of 3D mesh
reconstruction using posterior samples obtained by our informed
sampling \MIXLMH.

\clearpage

\section{Additional Results on the Face Problem with CMP}

\begin{figure*}[h]
  \begin{center}
  \centerline{\includegraphics[width=1.0\columnwidth]{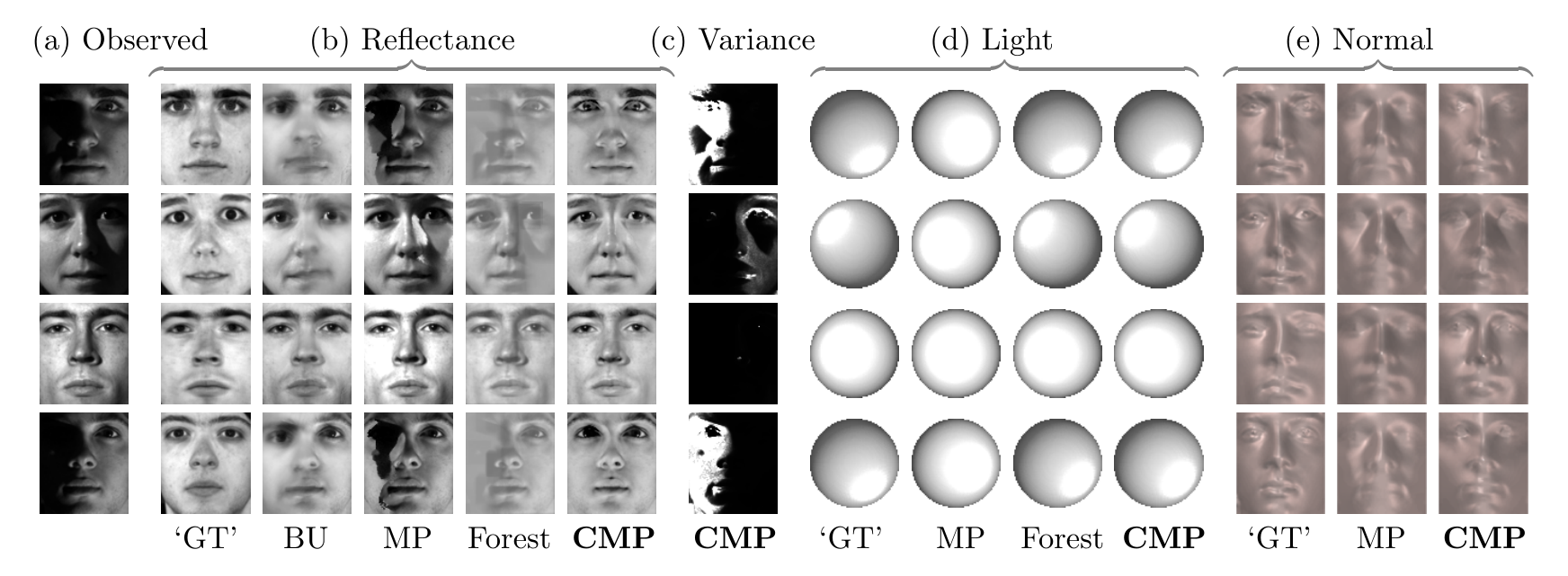}}
  \vspace{-1.2cm}
  \end{center}
	\mycaption{A visual comparison of inference results}{(a)~Observed images. (b)~Inferred reflectance maps. \textit{GT} is the photometric stereo groundtruth, \textit{BU} is the Biswas \etal (2009) reflectance estimate and \textit{Forest} is the consensus prediction. (c)~The variance of the inferred reflectance estimate produced by \MTD (normalized across rows).(d)~Visualization of inferred light directions. (e)~Inferred normal maps.}
	\label{fig:shading-qualitative-multiple-subjects-supp}
\end{figure*}

\begin{figure*}[h]
	\centering
	\setlength\fboxsep{0.2mm}
	\setlength\fboxrule{0pt}
	\begin{tikzpicture}

		\matrix at (0, 0) [matrix of nodes, nodes={anchor=east}, column sep=-0.05cm, row sep=-0.2cm]
		{
			\fbox{\includegraphics[width=1cm]{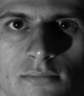}} &
			\fbox{\includegraphics[width=1cm]{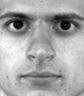}} &
			\fbox{\includegraphics[width=1cm]{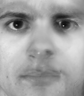}}  &
			\fbox{\includegraphics[width=1cm]{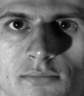}}  &
			\fbox{\includegraphics[width=1cm]{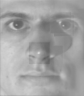}}  &
			\fbox{\includegraphics[width=1cm]{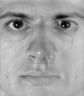}}  &
			\fbox{\includegraphics[width=1cm]{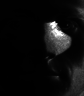}}
			 \\

			\fbox{\includegraphics[width=1cm]{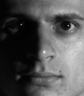}} &
			\fbox{\includegraphics[width=1cm]{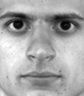}} &
			\fbox{\includegraphics[width=1cm]{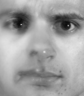}}  &
			\fbox{\includegraphics[width=1cm]{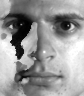}}  &
			\fbox{\includegraphics[width=1cm]{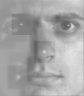}}  &
			\fbox{\includegraphics[width=1cm]{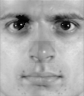}}  &
			\fbox{\includegraphics[width=1cm]{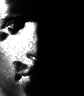}}
			 \\

			\fbox{\includegraphics[width=1cm]{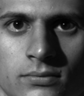}} &
			\fbox{\includegraphics[width=1cm]{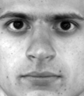}} &
			\fbox{\includegraphics[width=1cm]{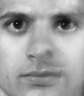}}  &
			\fbox{\includegraphics[width=1cm]{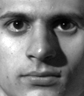}}  &
			\fbox{\includegraphics[width=1cm]{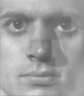}}  &
			\fbox{\includegraphics[width=1cm]{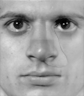}}  &
			\fbox{\includegraphics[width=1cm]{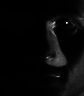}}
			 \\
	     };

       \node at (-3.85, -2.0) {\small Observed};
       \node at (-2.55, -2.0) {\small `GT'};
       \node at (-1.27, -2.0) {\small BU};
       \node at (0.0, -2.0) {\small MP};
       \node at (1.27, -2.0) {\small Forest};
       \node at (2.55, -2.0) {\small \textbf{CMP}};
       \node at (3.85, -2.0) {\small Variance};

	\end{tikzpicture}
	\mycaption{Robustness to varying illumination}{Reflectance estimation on a subject images with varying illumination. Left to right: observed image, photometric stereo estimate (GT)
  which is used as a proxy for groundtruth, bottom-up estimate of \cite{Biswas2009}, VMP result, consensus forest estimate, CMP mean, and CMP variance.}
	\label{fig:shading-qualitative-same-subject-supp}
\end{figure*}

Figure~\ref{fig:shading-qualitative-multiple-subjects-supp} shows inference results for reflectance maps, normal maps and lights for randomly chosen test images, and Fig.~\ref{fig:shading-qualitative-same-subject-supp} shows reflectance estimation results on multiple images of the same subject produced under different illumination conditions. CMP is able to produce estimates that are closer to the groundtruth across different subjects and illumination conditions.

\clearpage

\section{Additional Material for Learning Sparse High Dimensional Filters}
\label{sec:appendix-bnn}

This part of supplementary material contains a more detailed overview of the permutohedral
lattice convolution in Section~\ref{sec:permconv}, more experiments in
Section~\ref{sec:addexps} and additional results with protocols for
the experiments presented in Chapter~\ref{chap:bnn} in Section~\ref{sec:addresults}.

\vspace{-0.2cm}
\subsection{General Permutohedral Convolutions}
\label{sec:permconv}

A core technical contribution of this work is the generalization of the Gaussian permutohedral lattice
convolution proposed in~\cite{adams2010fast} to the full non-separable case with the
ability to perform back-propagation. Although, conceptually, there are minor
differences between Gaussian and general parameterized filters, there are non-trivial practical
differences in terms of the algorithmic implementation. The Gauss filters belong to
the separable class and can thus be decomposed into multiple
sequential one dimensional convolutions. We are interested in the general filter
convolutions, which can not be decomposed. Thus, performing a general permutohedral
convolution at a lattice point requires the computation of the inner product with the
neighboring elements in all the directions in the high-dimensional space.

Here, we give more details of the implementation differences of separable
and non-separable filters. In the following, we will explain the scalar case first.
Recall, that the forward pass of general permutohedral convolution
involves 3 steps: \textit{splatting}, \textit{convolving} and \textit{slicing}.
We follow the same splatting and slicing strategies as in~\cite{adams2010fast}
since these operations do not depend on the filter kernel. The main difference
between our work and the existing implementation of~\cite{adams2010fast} is
the way that the convolution operation is executed. This proceeds by constructing
a \emph{blur neighbor} matrix $K$ that stores for every lattice point all
values of the lattice neighbors that are needed to compute the filter output.

\begin{figure}[t!]
  \centering
    \includegraphics[width=0.6\columnwidth]{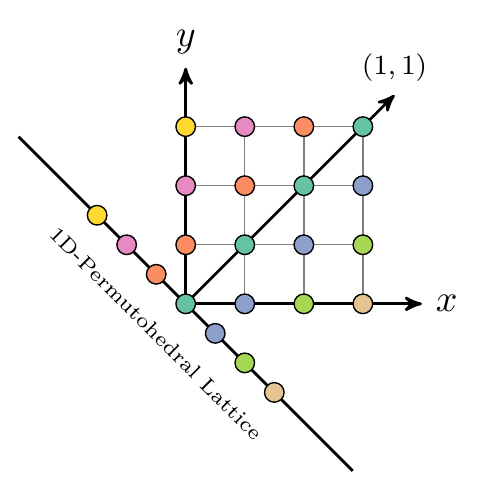}
  \mycaption{Illustration of 1D permutohedral lattice construction}
  {A $4\times 4$ $(x,y)$ grid lattice is projected onto the plane defined by the normal
  vector $(1,1)^{\top}$. This grid has $s+1=4$ and $d=2$ $(s+1)^{d}=4^2=16$ elements.
  In the projection, all points of the same color are projected onto the same points in the plane.
  The number of elements of the projected lattice is $t=(s+1)^d-s^d=4^2-3^2=7$, that is
  the $(4\times 4)$ grid minus the size of lattice that is $1$ smaller at each size, in this
  case a $(3\times 3)$ lattice (the upper right $(3\times 3)$ elements).
  }
\label{fig:latticeconstruction}
\end{figure}

The blur neighbor matrix is constructed by traversing through all the populated
lattice points and their neighboring elements.
This is done recursively to share computations. For any lattice point, the neighbors that are
$n$ hops away are the direct neighbors of the points that are $n-1$ hops away.
The size of a $d$ dimensional spatial filter with width $s+1$ is $(s+1)^{d}$ (\eg, a
$3\times 3$ filter, $s=2$ in $d=2$ has $3^2=9$ elements) and this size grows
exponentially in the number of dimensions $d$. The permutohedral lattice is constructed by
projecting a regular grid onto the plane spanned by the $d$ dimensional normal vector ${(1,\ldots,1)}^{\top}$. See
Fig.~\ref{fig:latticeconstruction} for an illustration of the 1D lattice construction.
Many corners of a grid filter are projected onto the same point, in total $t = {(s+1)}^{d} -
s^{d}$ elements remain in the permutohedral filter with $s$ neighborhood in $d-1$ dimensions.
If the lattice has $m$ populated elements, the
matrix $K$ has size $t\times m$. Note that, since the input signal is typically
sparse, only a few lattice corners are being populated in the \textit{slicing} step.
We use a hash-table to keep track of these points and traverse only through
the populated lattice points for this neighborhood matrix construction.

Once the blur neighbor matrix $K$ is constructed, we can perform the convolution
by the matrix vector multiplication
\begin{equation}
\ell' = BK,
\label{eq:conv}
\end{equation}
where $B$ is the $1 \times t$ filter kernel (whose values we will learn) and $\ell'\in\mathbb{R}^{1\times m}$
is the result of the filtering at the $m$ lattice points. In practice, we found that the
matrix $K$ is sometimes too large to fit into GPU memory and we divided the matrix $K$
into smaller pieces to compute Eq.~\ref{eq:conv} sequentially.

In the general multi-dimensional case, the signal $\ell$ is of $c$ dimensions. Then
the kernel $B$ is of size $c \times t$ and $K$ stores the $c$ dimensional vectors
accordingly. When the input and output points are different, we slice only the
input points and splat only at the output points.

\subsection{Additional Experiments}
\label{sec:addexps}
In this section, we discuss more use-cases for the learned bilateral filters, one
use-case of BNNs and two single filter applications for image and 3D mesh denoising.

\subsubsection{Recognition of subsampled MNIST}\label{sec:app_mnist}

One of the strengths of the proposed filter convolution is that it does not
require the input to lie on a regular grid. The only requirement is to define a distance
between features of the input signal.
We highlight this feature with the following experiment using the
classical MNIST ten class classification problem~\cite{lecun1998mnist}. We sample a
sparse set of $N$ points $(x,y)\in [0,1]\times [0,1]$
uniformly at random in the input image, use their interpolated values
as signal and the \emph{continuous} $(x,y)$ positions as features. This mimics
sub-sampling of a high-dimensional signal. To compare against a spatial convolution,
we interpolate the sparse set of values at the grid positions.

We take a reference implementation of LeNet~\cite{lecun1998gradient} that
is part of the Caffe project~\cite{jia2014caffe} and compare it
against the same architecture but replacing the first convolutional
layer with a bilateral convolution layer (BCL). The filter size
and numbers are adjusted to get a comparable number of parameters
($5\times 5$ for LeNet, $2$-neighborhood for BCL).

The results are shown in Table~\ref{tab:all-results}. We see that training
on the original MNIST data (column Original, LeNet vs. BNN) leads to a slight
decrease in performance of the BNN (99.03\%) compared to LeNet
(99.19\%). The BNN can be trained and evaluated on sparse
signals, and we resample the image as described above for $N=$ 100\%, 60\% and
20\% of the total number of pixels. The methods are also evaluated
on test images that are subsampled in the same way. Note that we can
train and test with different subsampling rates. We introduce an additional
bilinear interpolation layer for the LeNet architecture to train on the same
data. In essence, both models perform a spatial interpolation and thus we
expect them to yield a similar classification accuracy. Once the data is of
higher dimensions, the permutohedral convolution will be faster due to hashing
the sparse input points, as well as less memory demanding in comparison to
naive application of a spatial convolution with interpolated values.

\begin{table}[t]
  \begin{center}
    \footnotesize
    \centering
    \begin{tabular}[t]{lllll}
      \toprule
              &     & \multicolumn{3}{c}{Test Subsampling} \\
       Method  & Original & 100\% & 60\% & 20\%\\
      \midrule
       LeNet &  \textbf{0.9919} & 0.9660 & 0.9348 & \textbf{0.6434} \\
       BNN &  0.9903 & \textbf{0.9844} & \textbf{0.9534} & 0.5767 \\
      \hline
       LeNet 100\% & 0.9856 & 0.9809 & 0.9678 & \textbf{0.7386} \\
       BNN 100\% & \textbf{0.9900} & \textbf{0.9863} & \textbf{0.9699} & 0.6910 \\
      \hline
       LeNet 60\% & 0.9848 & 0.9821 & 0.9740 & 0.8151 \\
       BNN 60\% & \textbf{0.9885} & \textbf{0.9864} & \textbf{0.9771} & \textbf{0.8214}\\
      \hline
       LeNet 20\% & \textbf{0.9763} & \textbf{0.9754} & 0.9695 & 0.8928 \\
       BNN 20\% & 0.9728 & 0.9735 & \textbf{0.9701} & \textbf{0.9042}\\
      \bottomrule
    \end{tabular}
  \end{center}
\vspace{-.2cm}
\caption{Classification accuracy on MNIST. We compare the
    LeNet~\cite{lecun1998gradient} implementation that is part of
    Caffe~\cite{jia2014caffe} to the network with the first layer
    replaced by a bilateral convolution layer (BCL). Both are trained
    on the original image resolution (first two rows). Three more BNN
    and CNN models are trained with randomly subsampled images (100\%,
    60\% and 20\% of the pixels). An additional bilinear interpolation
    layer samples the input signal on a spatial grid for the CNN model.
  }
  \label{tab:all-results}
\vspace{-.5cm}
\end{table}

\subsubsection{Image Denoising}

The main application that inspired the development of the bilateral
filtering operation is image denoising~\cite{aurich1995non}, there
using a single Gaussian kernel. Our development allows to learn this
kernel function from data and we explore how to improve using a \emph{single}
but more general bilateral filter.

We use the Berkeley segmentation dataset
(BSDS500)~\cite{arbelaezi2011bsds500} as a test bed. The color
images in the dataset are converted to gray-scale,
and corrupted with Gaussian noise with a standard deviation of
$\frac {25} {255}$.

We compare the performance of four different filter models on a
denoising task.
The first baseline model (`Spatial' in Table \ref{tab:denoising}, $25$
weights) uses a single spatial filter with a kernel size of
$5$ and predicts the scalar gray-scale value at the center pixel. The next model
(`Gauss Bilateral') applies a bilateral \emph{Gaussian}
filter to the noisy input, using position and intensity features $\f=(x,y,v)^\top$.
The third setup (`Learned Bilateral', $65$ weights)
takes a Gauss kernel as initialization and
fits all filter weights on the train set to minimize the
mean squared error with respect to the clean images.
We run a combination
of spatial and permutohedral convolutions on spatial and bilateral
features (`Spatial + Bilateral (Learned)') to check for a complementary
performance of the two convolutions.

\label{sec:exp:denoising}
\begin{table}[!h]
\begin{center}
  \footnotesize
  \begin{tabular}[t]{lr}
    \toprule
    Method & PSNR \\
    \midrule
    Noisy Input & $20.17$ \\
    Spatial & $26.27$ \\
    Gauss Bilateral & $26.51$ \\
    Learned Bilateral & $26.58$ \\
    Spatial + Bilateral (Learned) & \textbf{$26.65$} \\
    \bottomrule
  \end{tabular}
\end{center}
\vspace{-0.5em}
\caption{PSNR results of a denoising task using the BSDS500
  dataset~\cite{arbelaezi2011bsds500}}
\vspace{-0.5em}
\label{tab:denoising}
\end{table}
\vspace{-0.2em}

The PSNR scores evaluated on full images of the test set are
shown in Table \ref{tab:denoising}. We find that an untrained bilateral
filter already performs better than a trained spatial convolution
($26.27$ to $26.51$). A learned convolution further improve the
performance slightly. We chose this simple one-kernel setup to
validate an advantage of the generalized bilateral filter. A competitive
denoising system would employ RGB color information and also
needs to be properly adjusted in network size. Multi-layer perceptrons
have obtained state-of-the-art denoising results~\cite{burger12cvpr}
and the permutohedral lattice layer can readily be used in such an
architecture, which is intended future work.

\subsection{Additional results}
\label{sec:addresults}

This section contains more qualitative results for the experiments presented in Chapter~\ref{chap:bnn}.

\begin{figure*}[th!]
  \centering
    \includegraphics[width=\columnwidth,trim={5cm 2.5cm 5cm 4.5cm},clip]{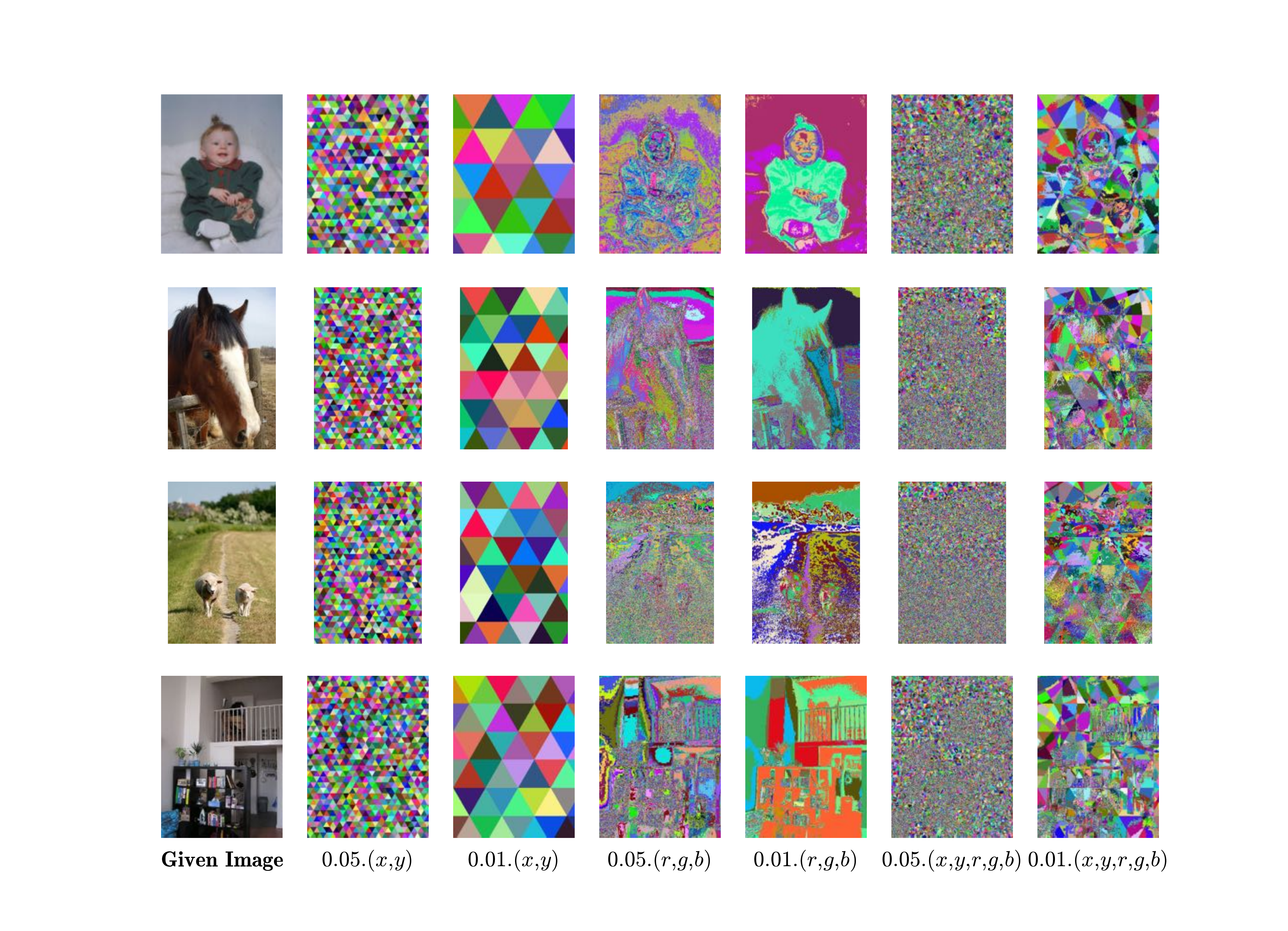}
    \vspace{-0.7cm}
  \mycaption{Visualization of the Permutohedral Lattice}
  {Sample lattice visualizations for different feature spaces. All pixels falling in the same simplex cell are shown with
  the same color. $(x,y)$ features correspond to image pixel positions, and $(r,g,b) \in [0,255]$ correspond
  to the red, green and blue color values.}
\label{fig:latticeviz}
\end{figure*}

\subsubsection{Lattice Visualization}

Figure~\ref{fig:latticeviz} shows sample lattice visualizations for different feature spaces.

\newcolumntype{L}[1]{>{\raggedright\let\newline\\\arraybackslash\hspace{0pt}}b{#1}}
\newcolumntype{C}[1]{>{\centering\let\newline\\\arraybackslash\hspace{0pt}}b{#1}}
\newcolumntype{R}[1]{>{\raggedleft\let\newline\\\arraybackslash\hspace{0pt}}b{#1}}

\subsubsection{Color Upsampling}\label{sec:color_upsampling}
\label{sec:col_upsample_extra}

\begin{figure*}[t!]
  \centering
    \subfigure{%
   \raisebox{2.0em}{
    \includegraphics[width=.06\columnwidth]{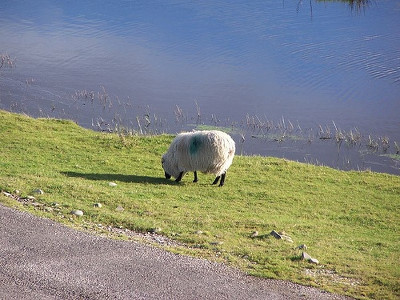}
   }
  }
  \subfigure{%
    \includegraphics[width=.17\columnwidth]{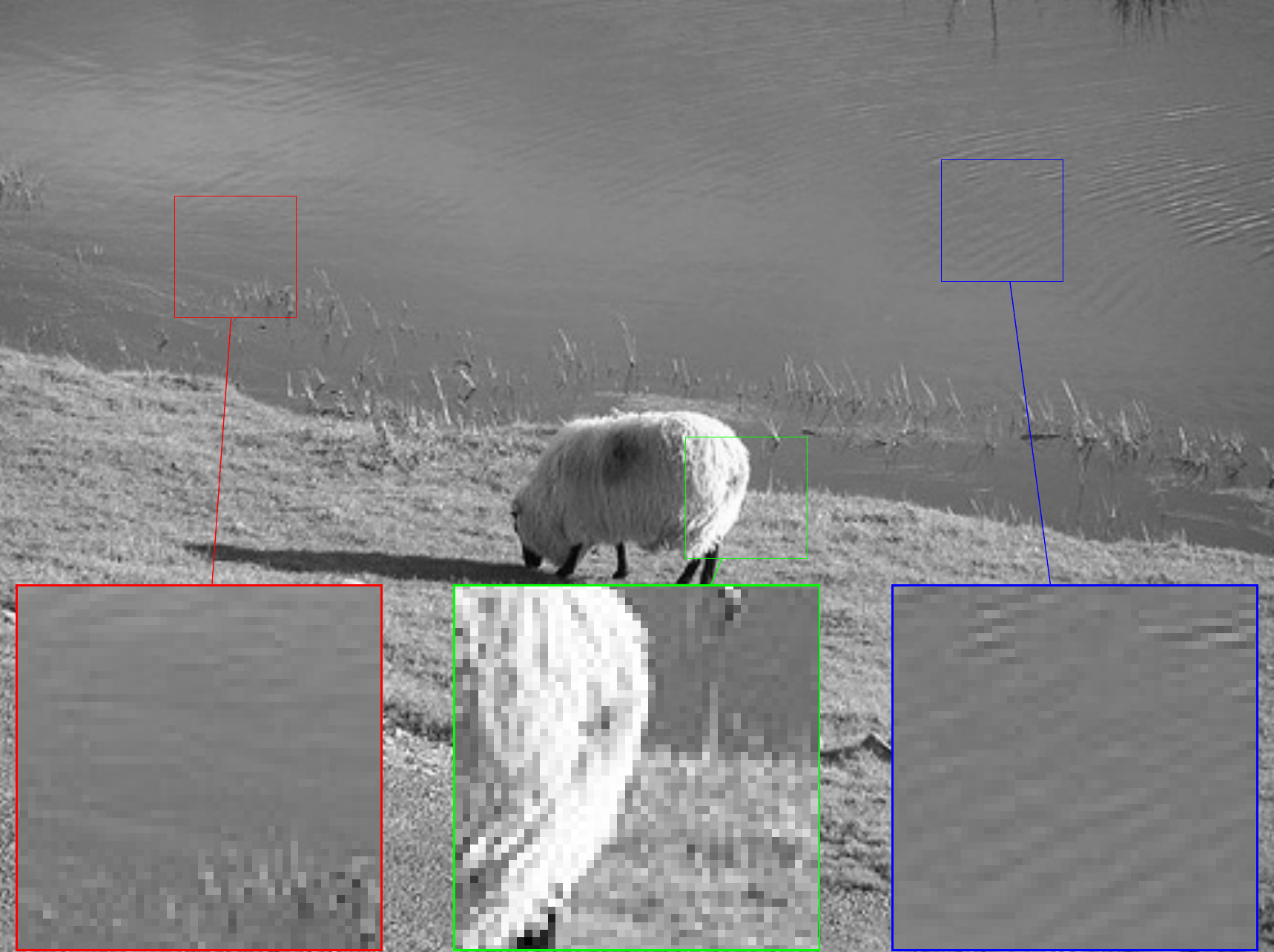}
  }
  \subfigure{%
    \includegraphics[width=.17\columnwidth]{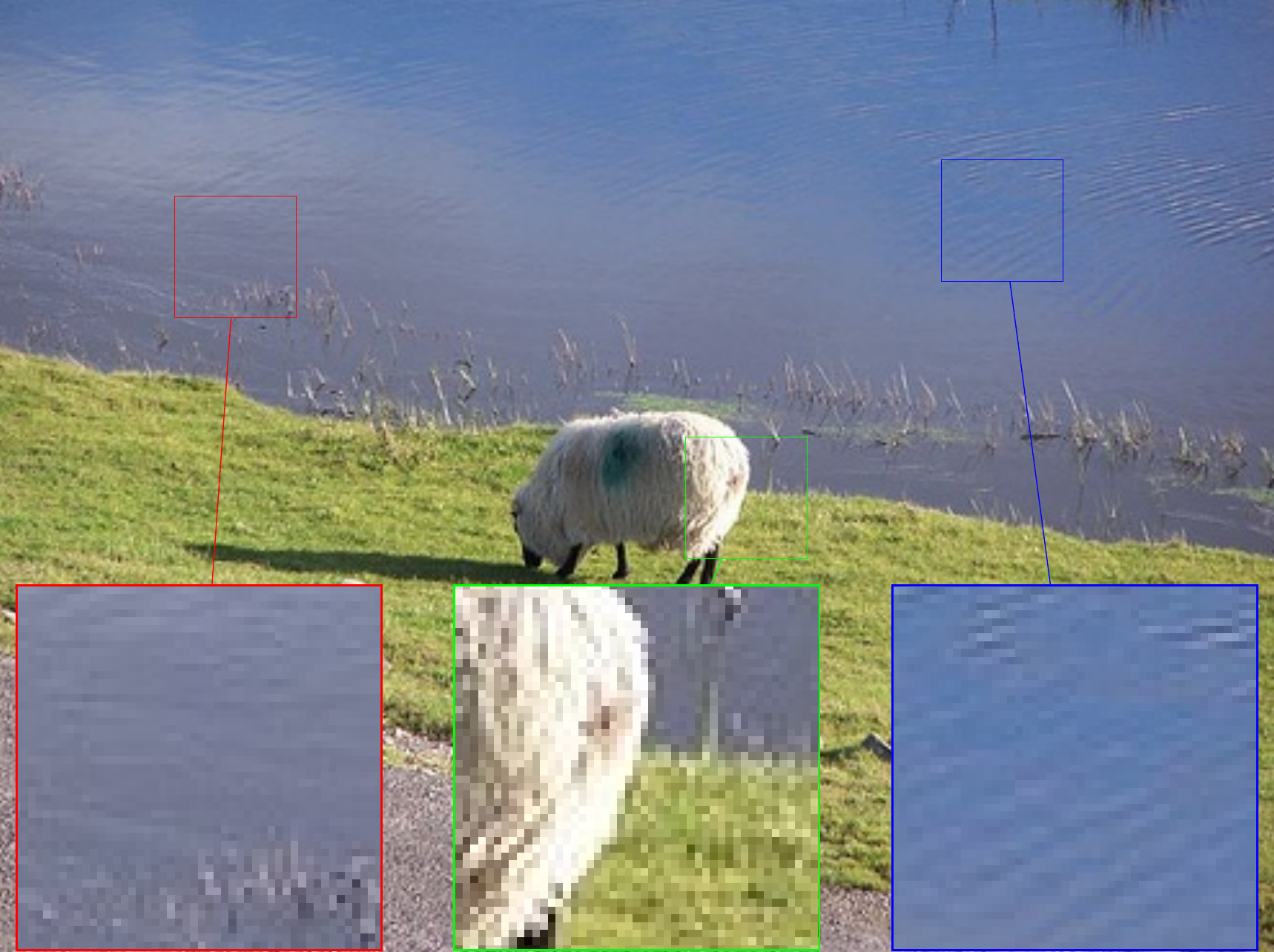}
  }
  \subfigure{%
    \includegraphics[width=.17\columnwidth]{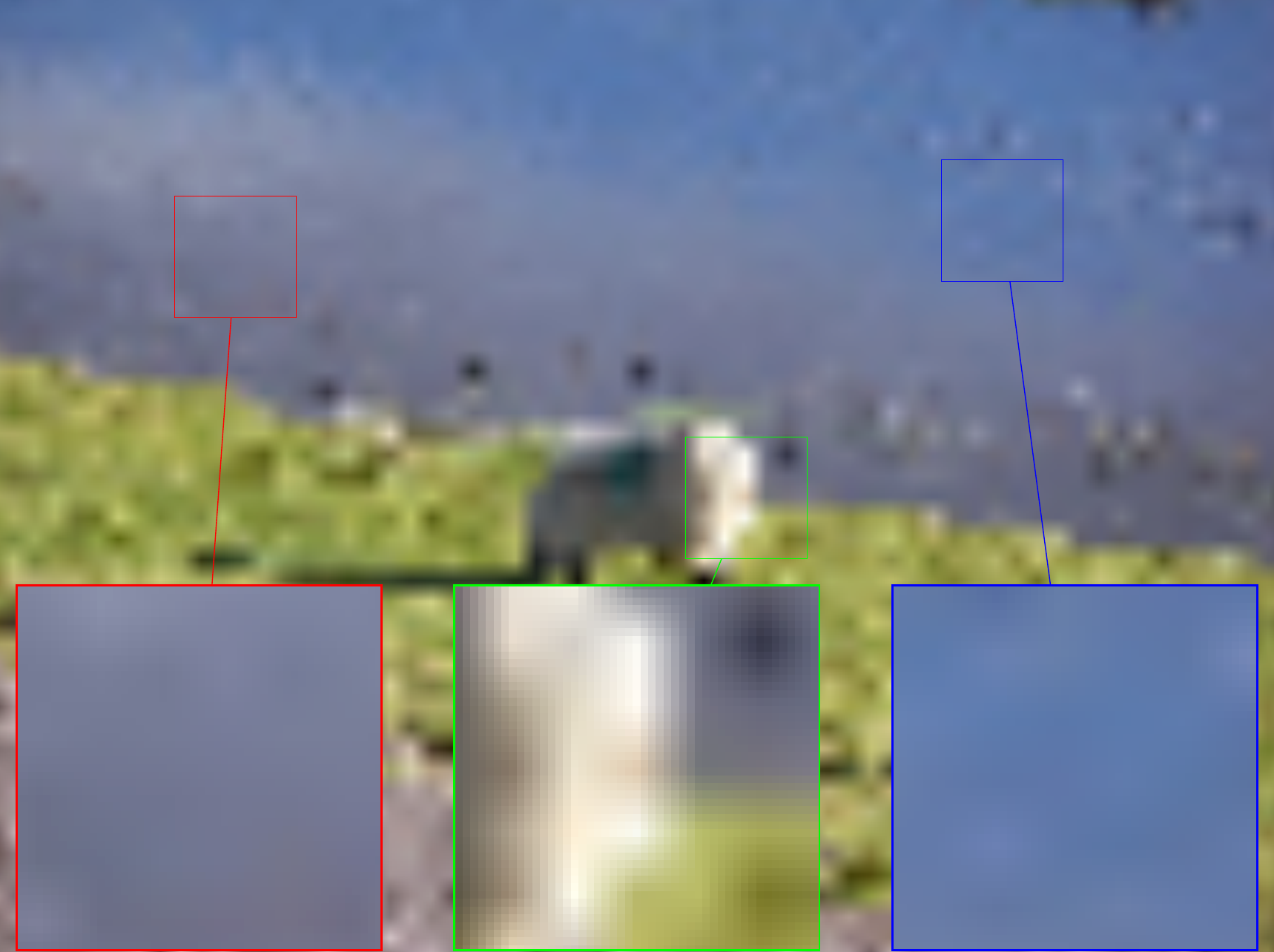}
  }
  \subfigure{%
    \includegraphics[width=.17\columnwidth]{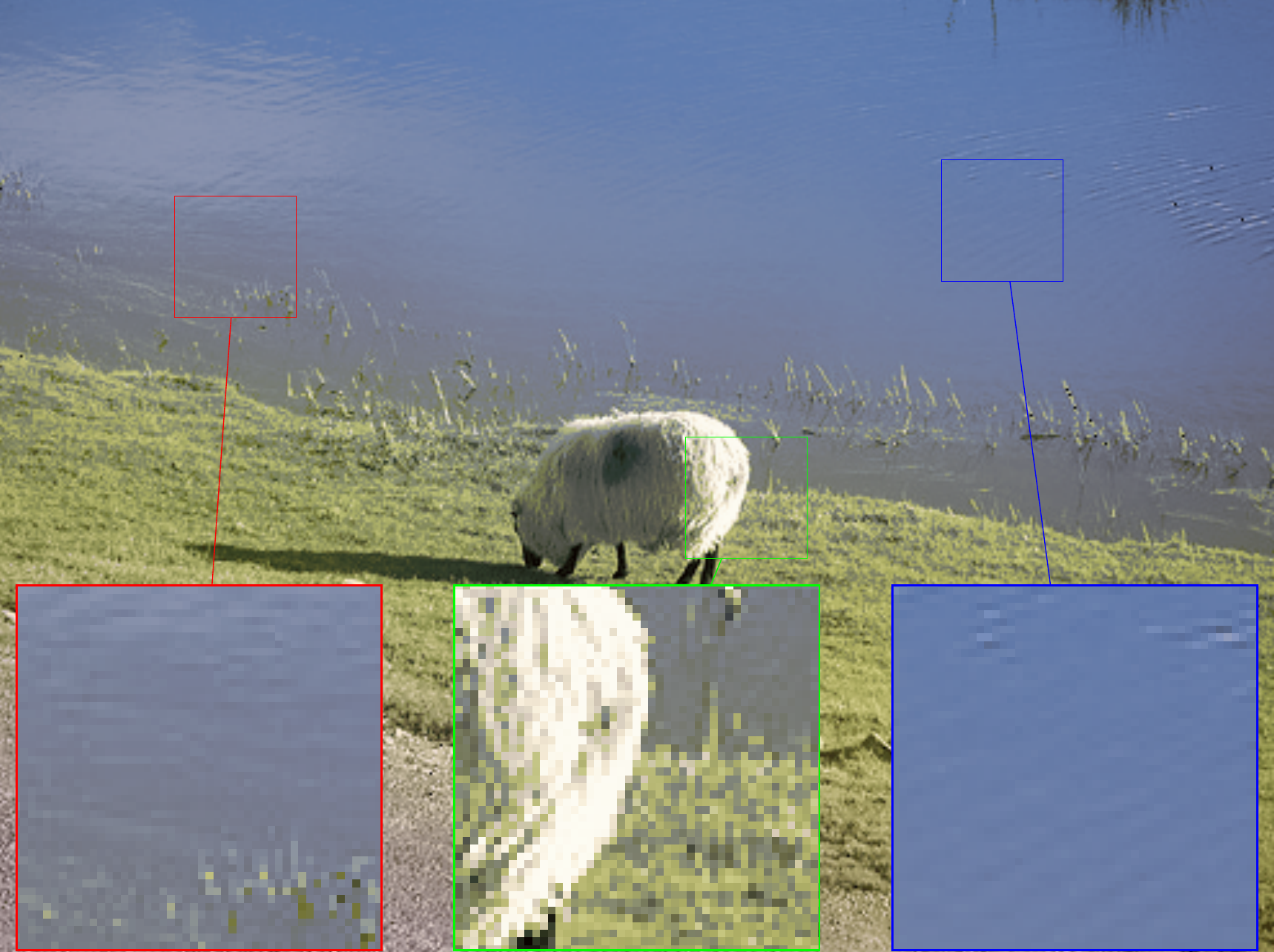}
  }
  \subfigure{%
    \includegraphics[width=.17\columnwidth]{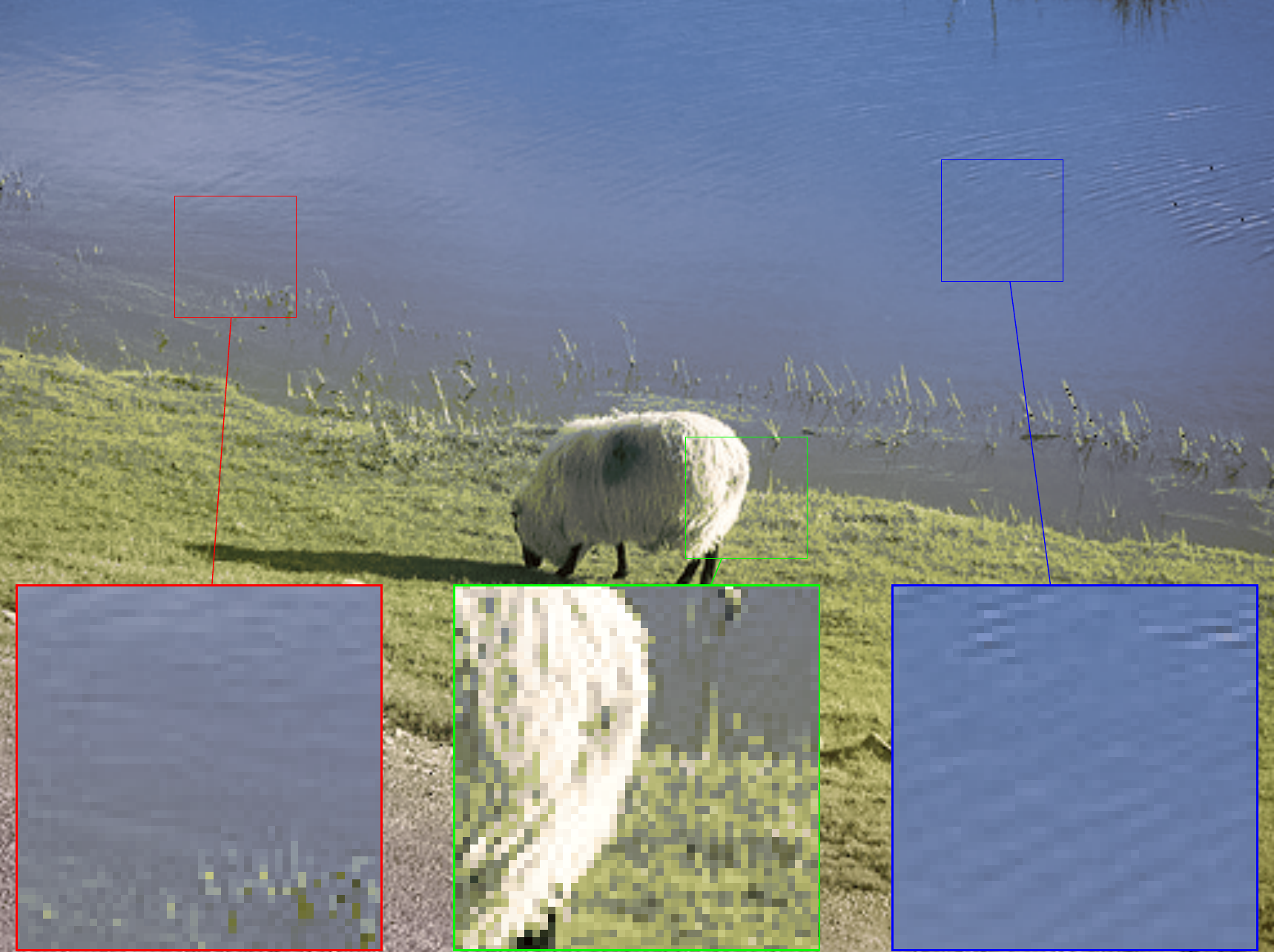}
  }\\
    \subfigure{%
   \raisebox{2.0em}{
    \includegraphics[width=.06\columnwidth]{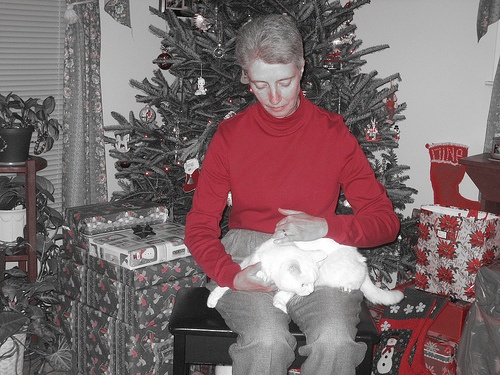}
   }
  }
  \subfigure{%
    \includegraphics[width=.17\columnwidth]{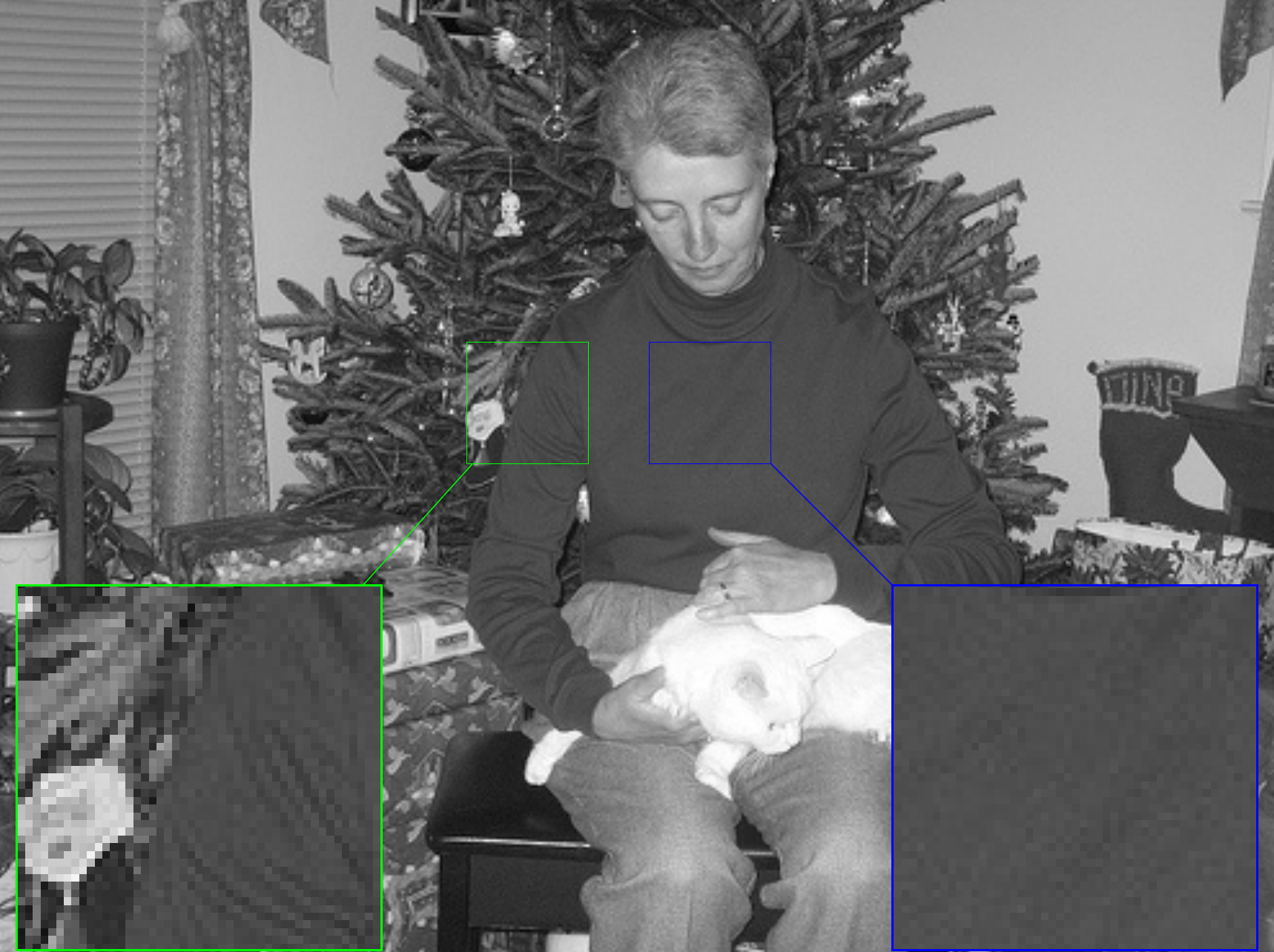}
  }
  \subfigure{%
    \includegraphics[width=.17\columnwidth]{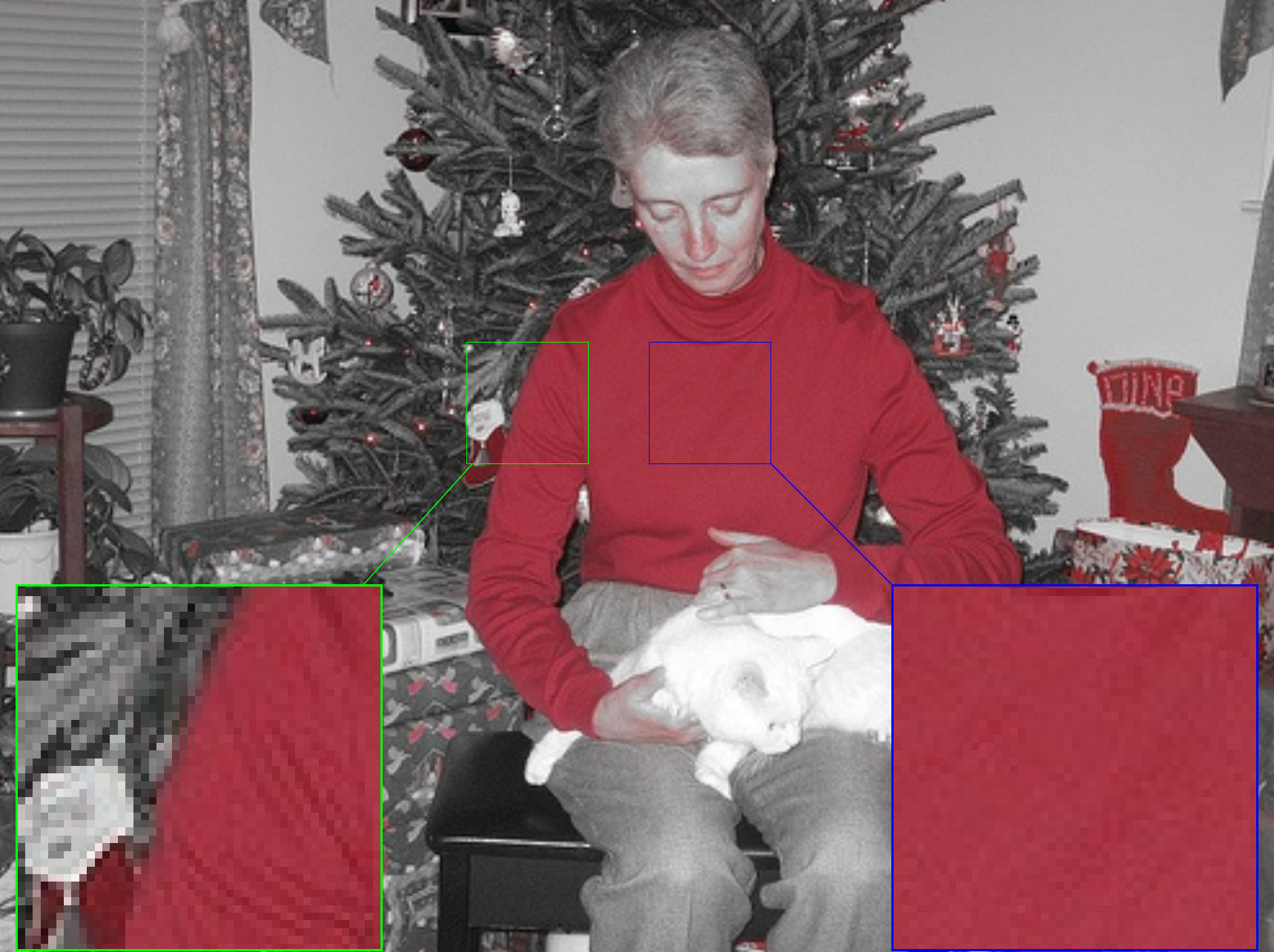}
  }
  \subfigure{%
    \includegraphics[width=.17\columnwidth]{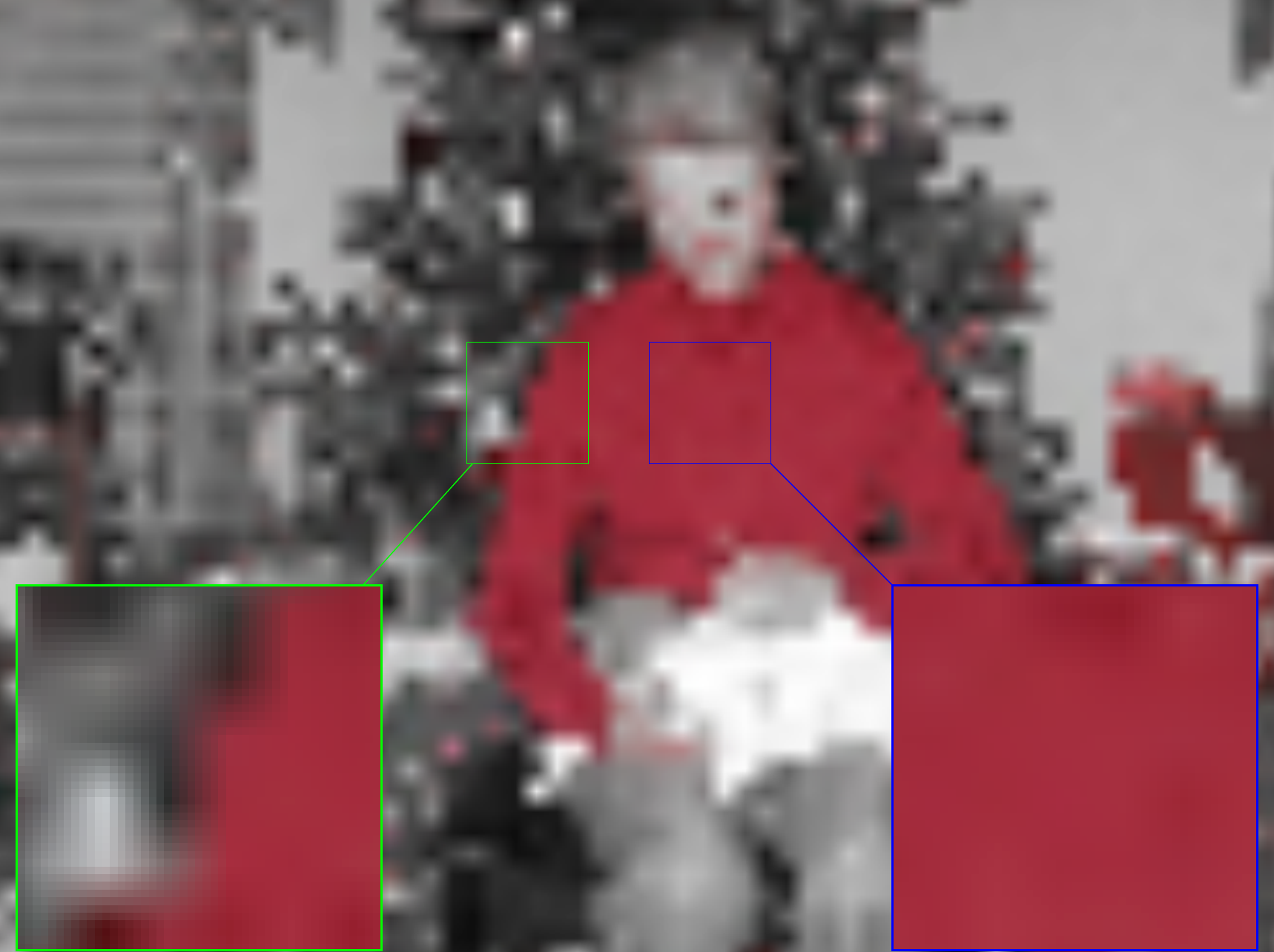}
  }
  \subfigure{%
    \includegraphics[width=.17\columnwidth]{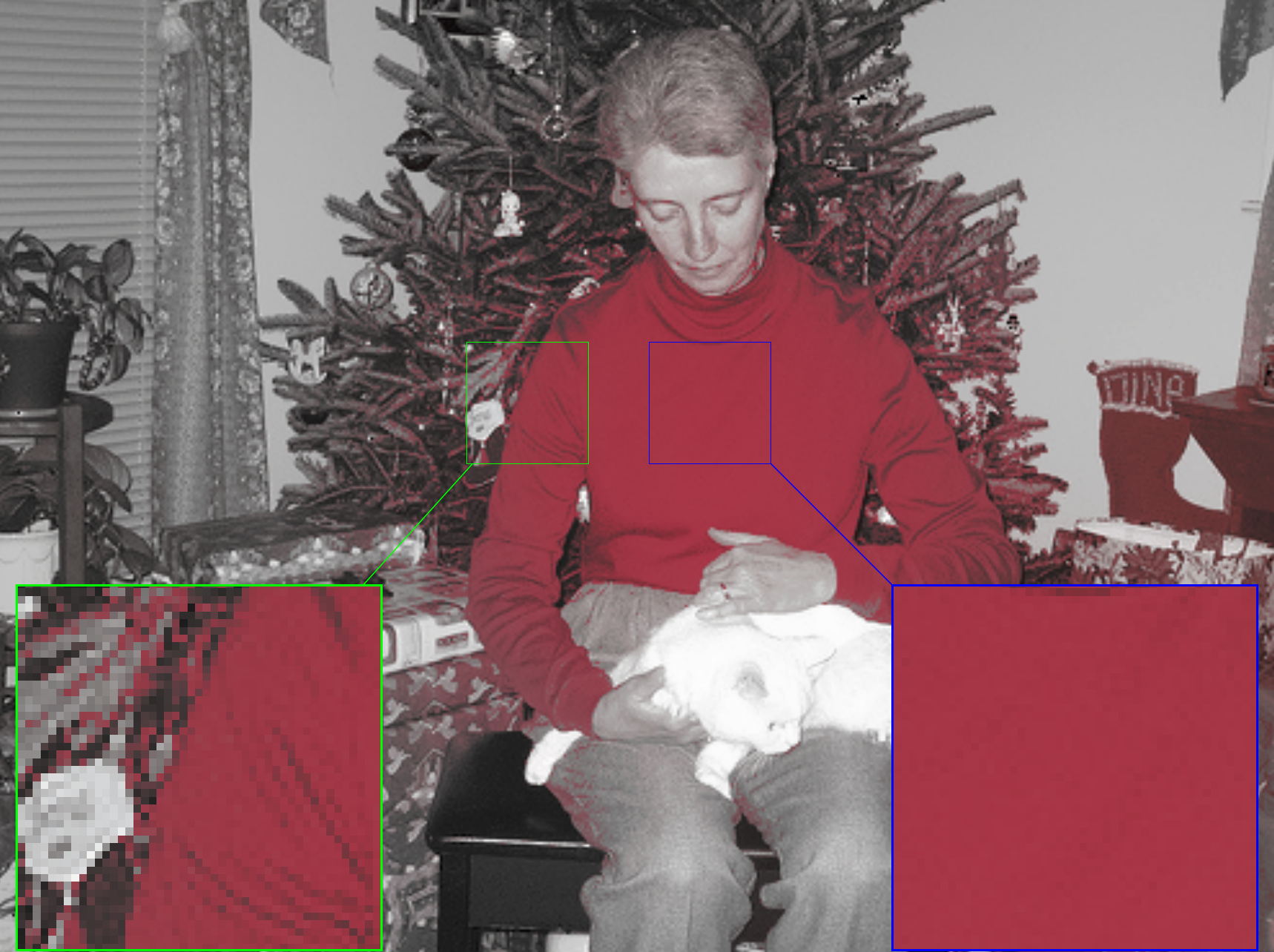}
  }
  \subfigure{%
    \includegraphics[width=.17\columnwidth]{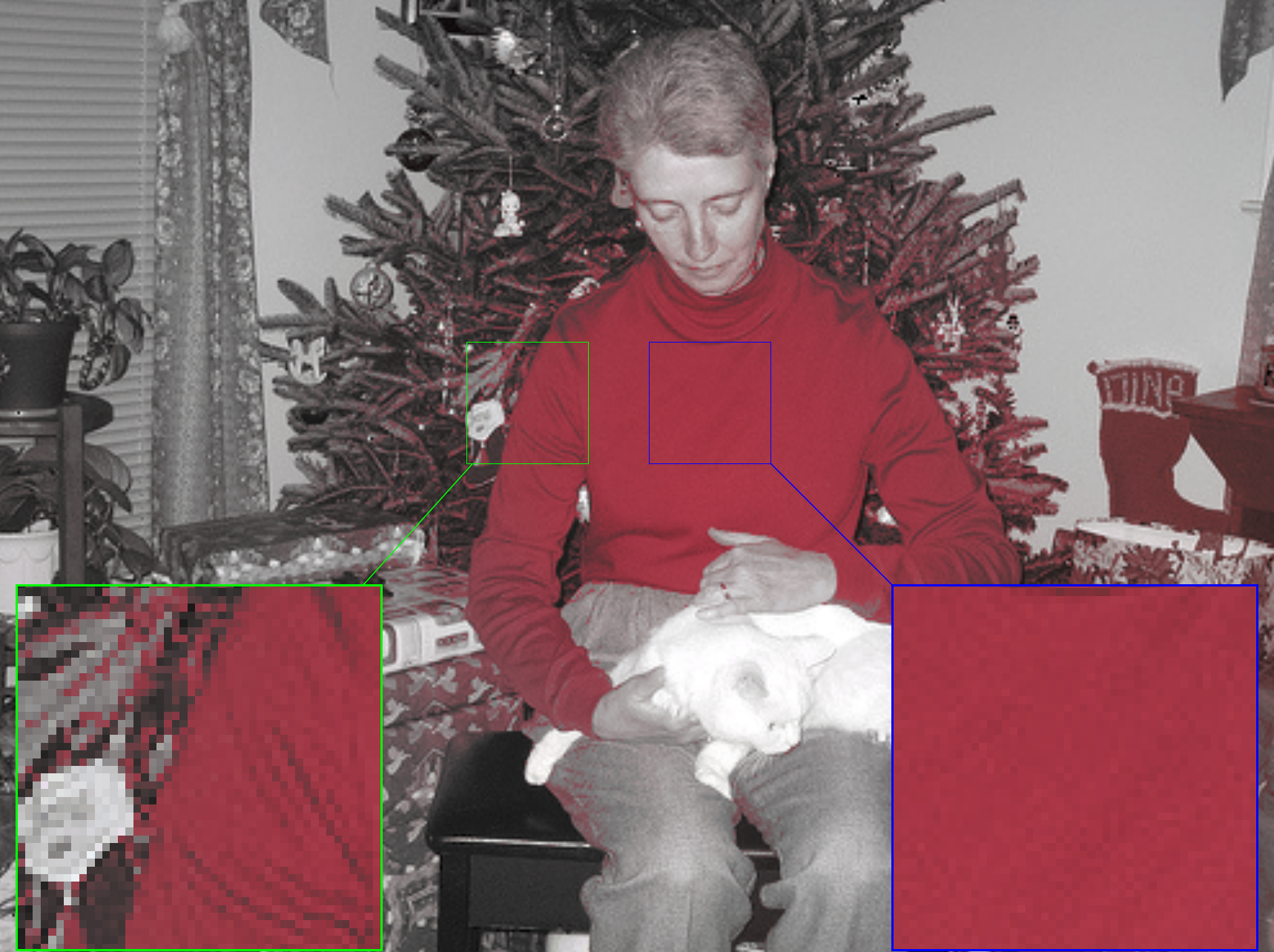}
  }\\
  \setcounter{subfigure}{0}
  \small{
  \subfigure[Inp.]{%
  \raisebox{2.0em}{
    \includegraphics[width=.06\columnwidth]{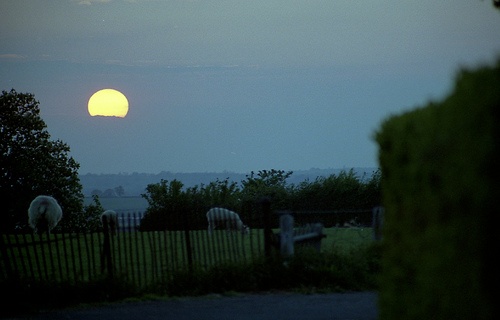}
   }
  }
  \subfigure[Guidance]{%
    \includegraphics[width=.17\columnwidth]{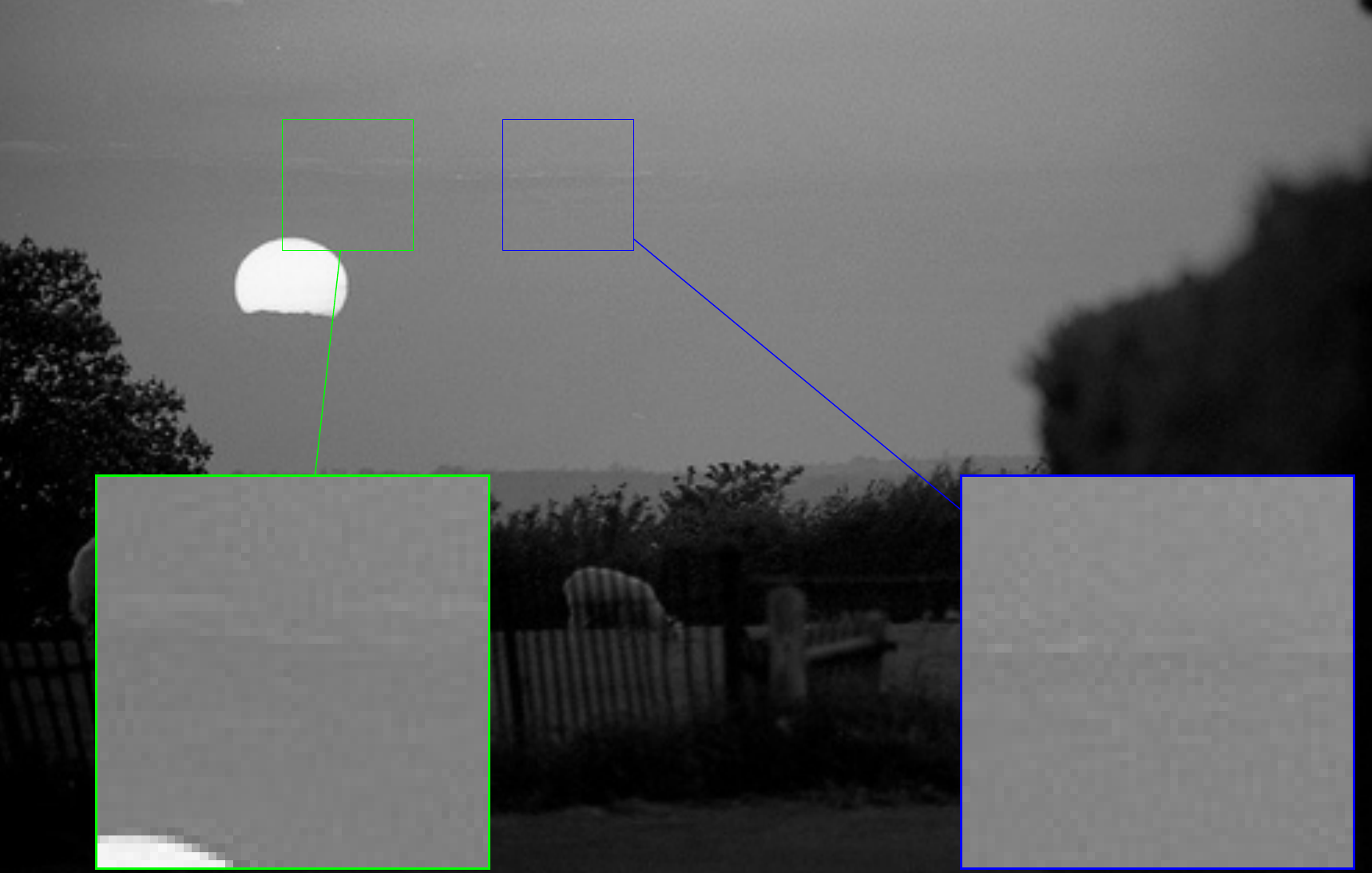}
  }
   \subfigure[GT]{%
    \includegraphics[width=.17\columnwidth]{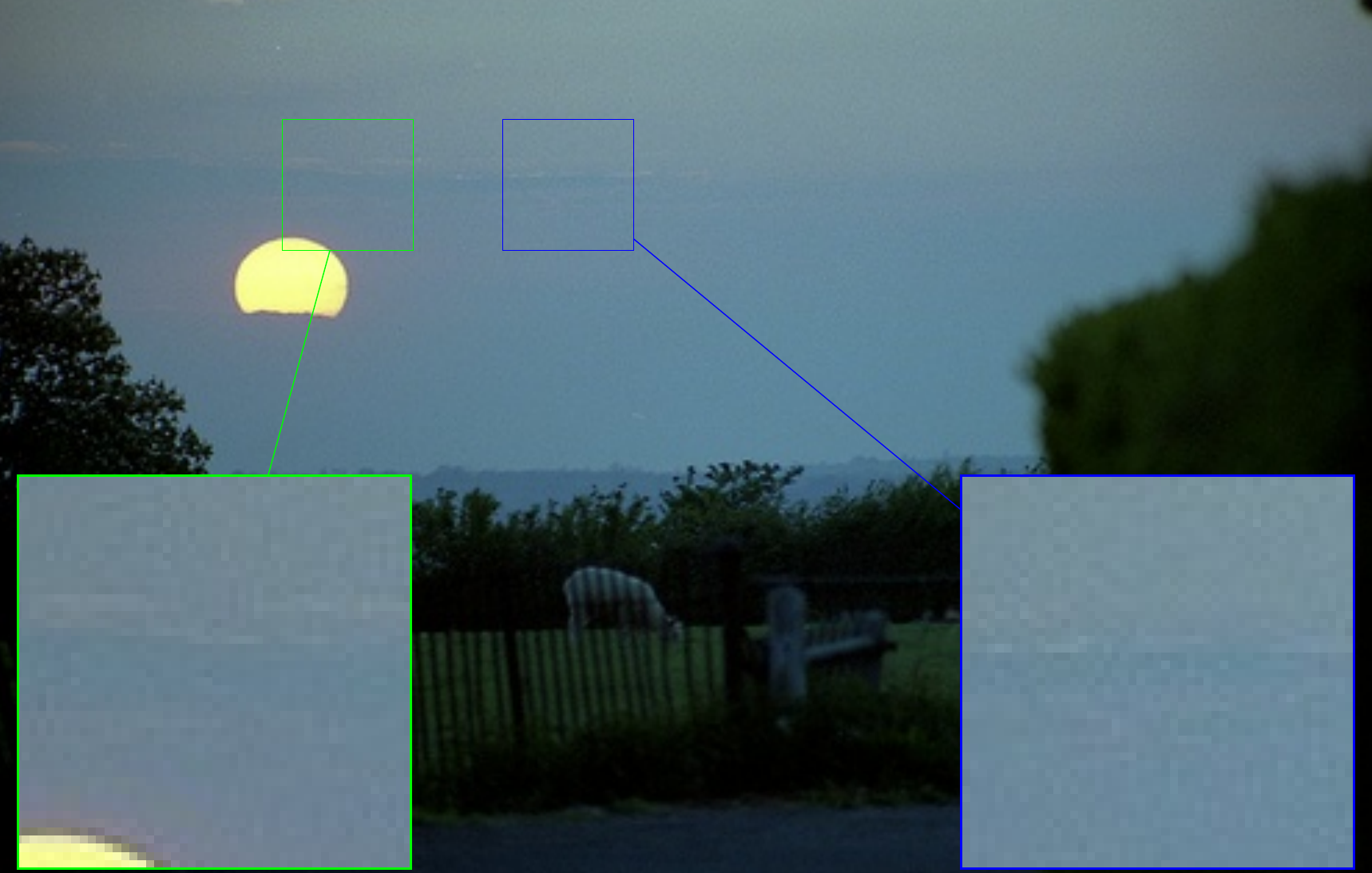}
  }
  \subfigure[Bicubic]{%
    \includegraphics[width=.17\columnwidth]{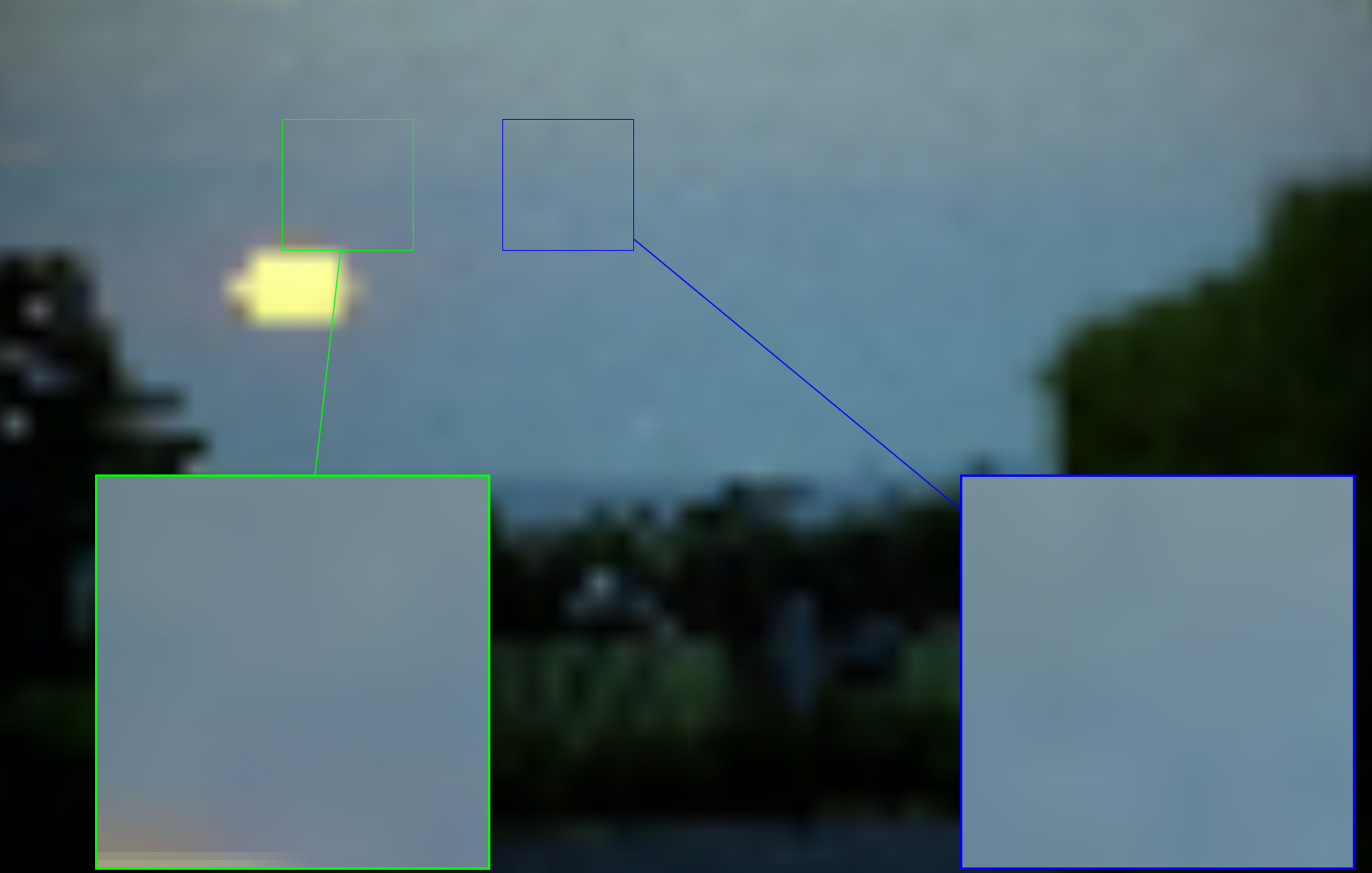}
  }
  \subfigure[Gauss-BF]{%
    \includegraphics[width=.17\columnwidth]{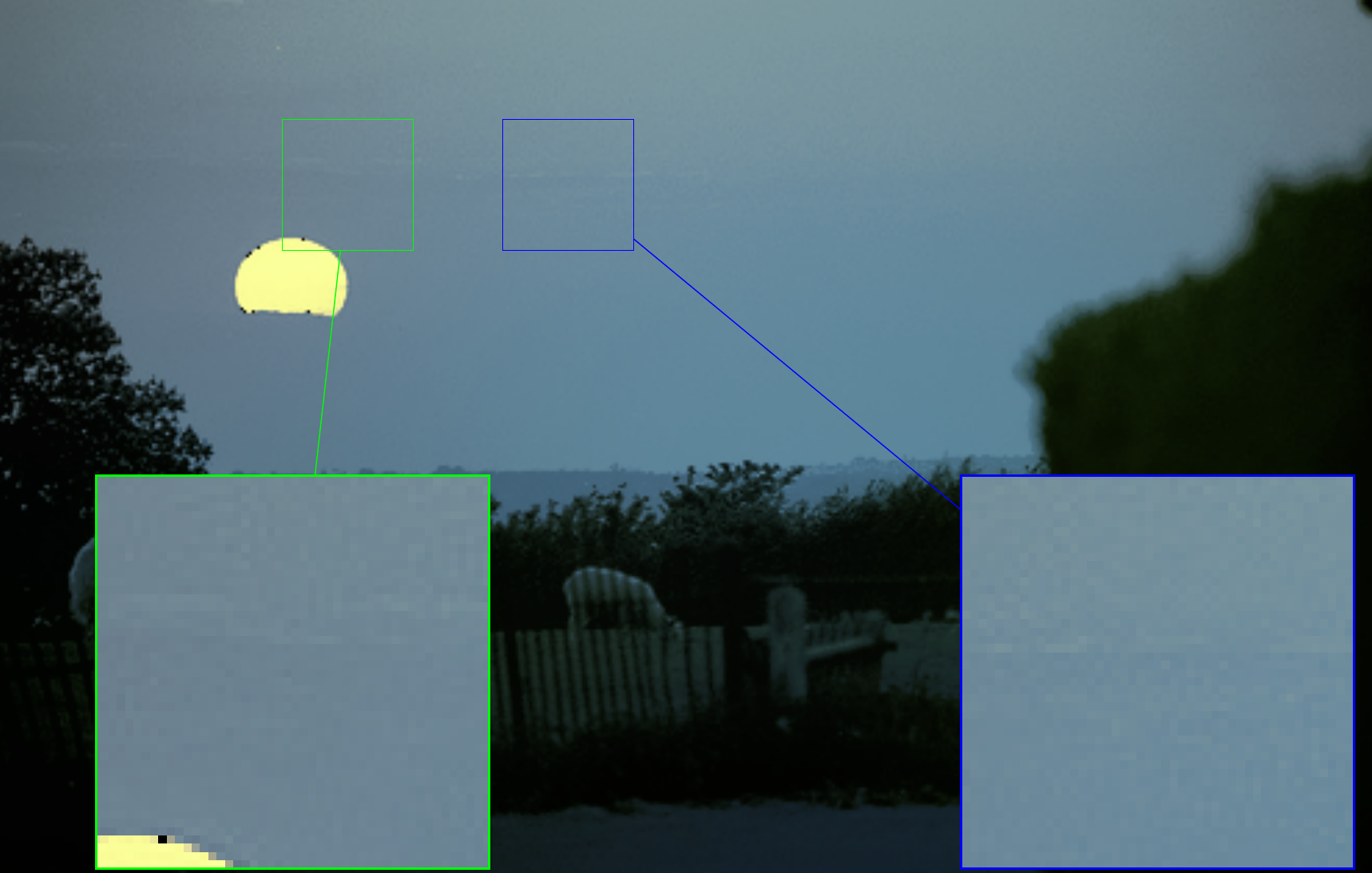}
  }
  \subfigure[Learned-BF]{%
    \includegraphics[width=.17\columnwidth]{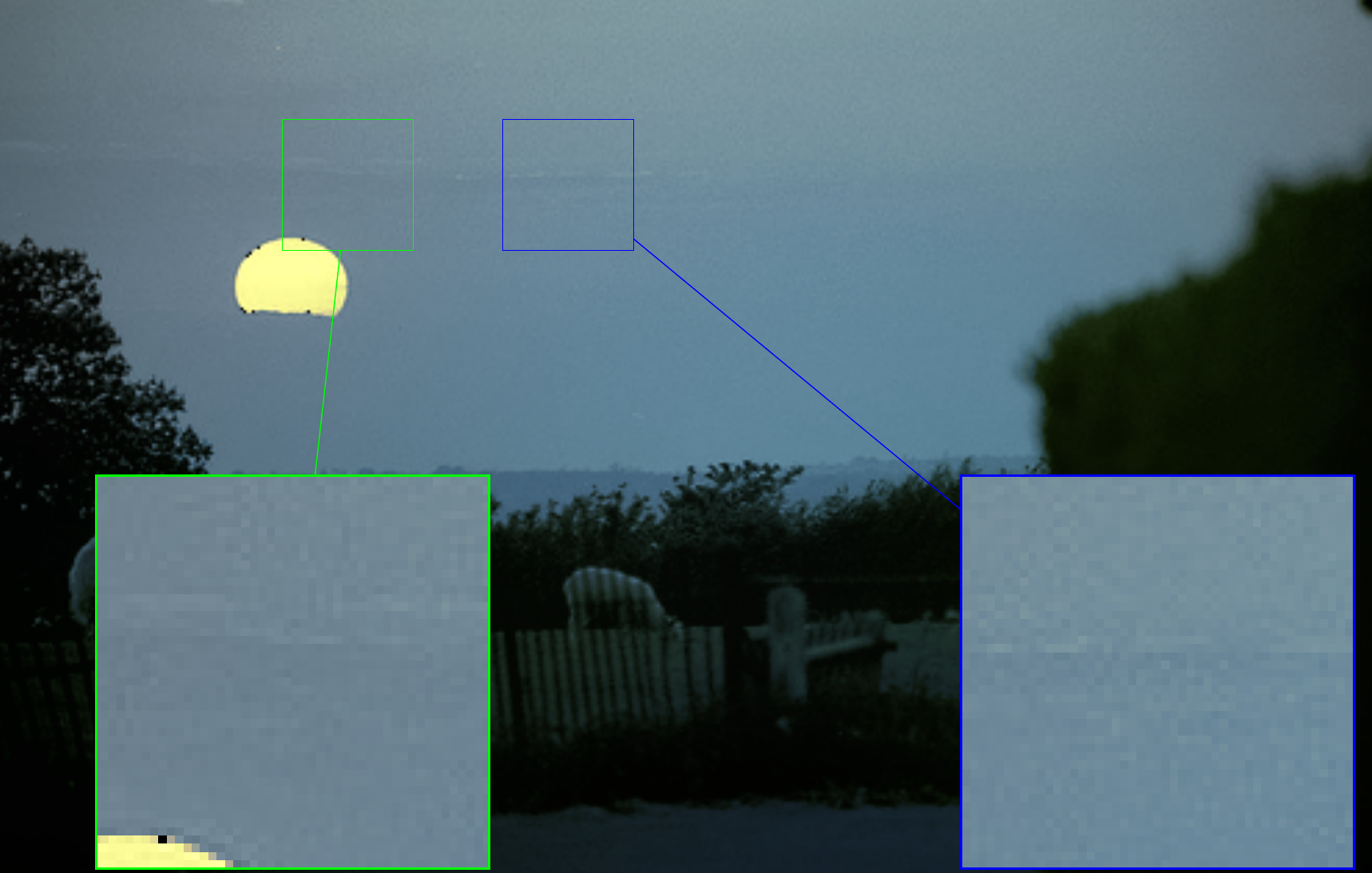}
  }
  }
  \vspace{-0.5cm}
  \mycaption{Color Upsampling}{Color $8\times$ upsampling results
  using different methods, from left to right, (a)~Low-resolution input color image (Inp.),
  (b)~Gray scale guidance image, (c)~Ground-truth color image; Upsampled color images with
  (d)~Bicubic interpolation, (e) Gauss bilateral upsampling and, (f)~Learned bilateral
  updampgling (best viewed on screen).}

\label{fig:Colour_upsample_visuals}
\end{figure*}

Some images of the upsampling for the Pascal
VOC12 dataset are shown in Fig.~\ref{fig:Colour_upsample_visuals}. It is
especially the low level image details that are better preserved with
a learned bilateral filter compared to the Gaussian case.

\subsubsection{Depth Upsampling}
\label{sec:depth_upsample_extra}

\begin{figure*}[t!]
  \centering
    \subfigure{%
   \raisebox{2.0em}{
    \includegraphics[width=.06\columnwidth]{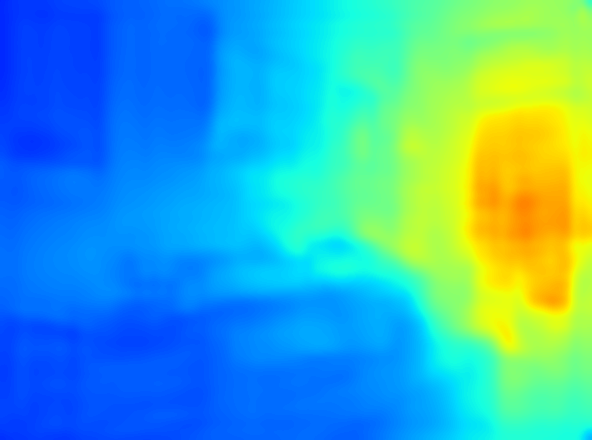}
   }
  }
  \subfigure{%
    \includegraphics[width=.17\columnwidth]{figures/supplementary/2given_image}
  }
  \subfigure{%
    \includegraphics[width=.17\columnwidth]{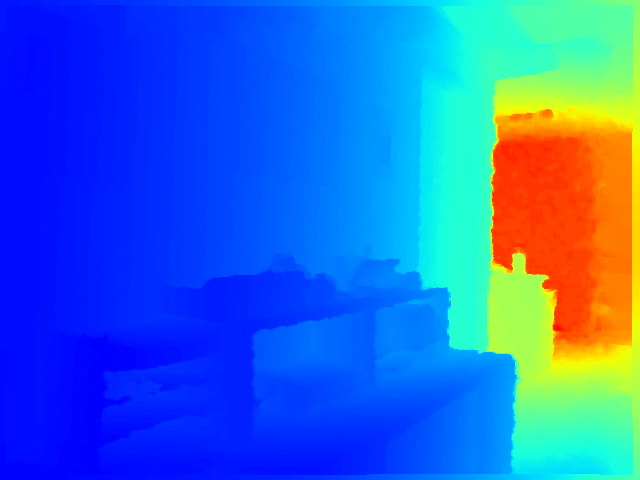}
  }
  \subfigure{%
    \includegraphics[width=.17\columnwidth]{figures/supplementary/2bicubic}
  }
  \subfigure{%
    \includegraphics[width=.17\columnwidth]{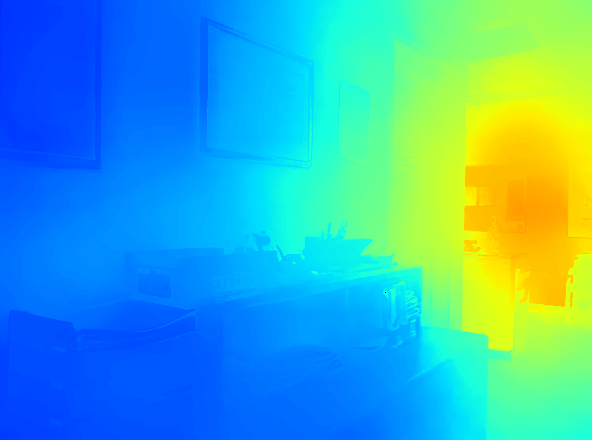}
  }
  \subfigure{%
    \includegraphics[width=.17\columnwidth]{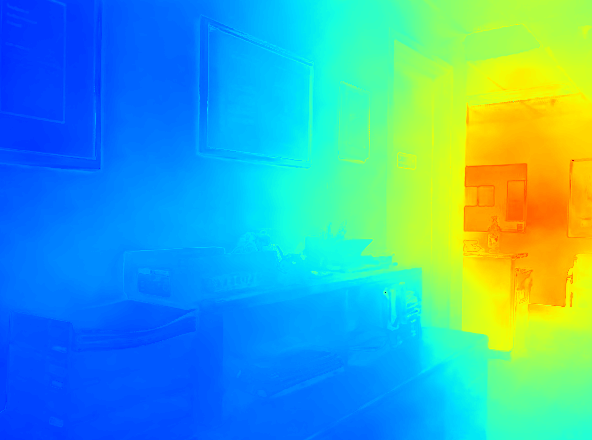}
  }\\
    \subfigure{%
   \raisebox{2.0em}{
    \includegraphics[width=.06\columnwidth]{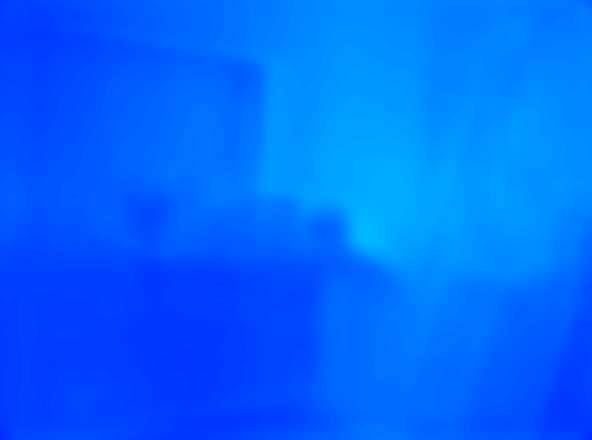}
   }
  }
  \subfigure{%
    \includegraphics[width=.17\columnwidth]{figures/supplementary/32given_image}
  }
  \subfigure{%
    \includegraphics[width=.17\columnwidth]{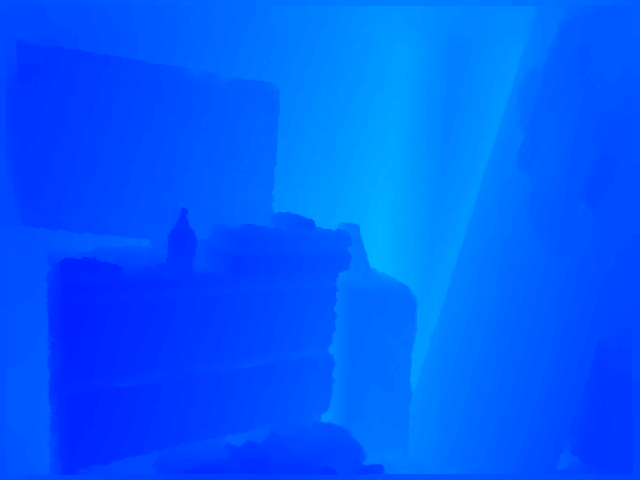}
  }
  \subfigure{%
    \includegraphics[width=.17\columnwidth]{figures/supplementary/32bicubic}
  }
  \subfigure{%
    \includegraphics[width=.17\columnwidth]{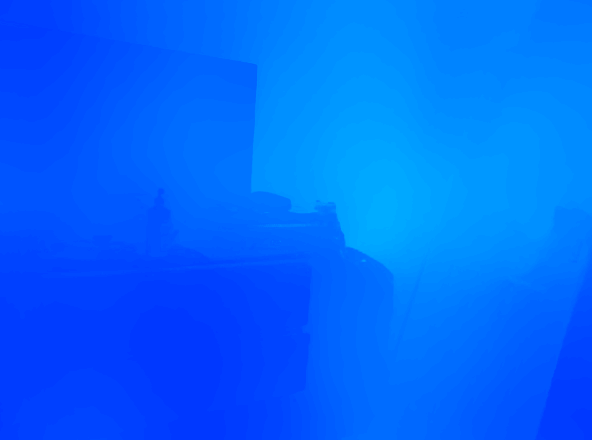}
  }
  \subfigure{%
    \includegraphics[width=.17\columnwidth]{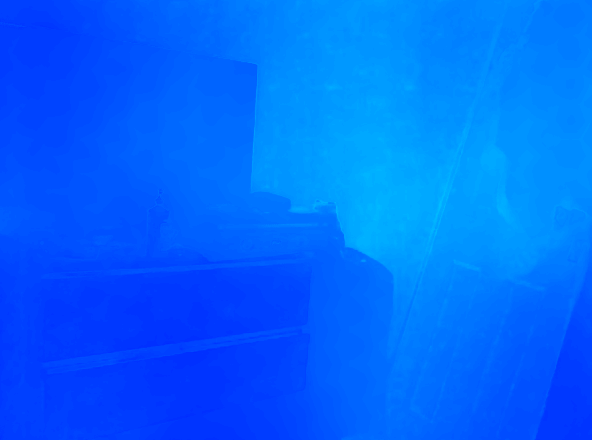}
  }\\
  \setcounter{subfigure}{0}
  \small{
  \subfigure[Inp.]{%
  \raisebox{2.0em}{
    \includegraphics[width=.06\columnwidth]{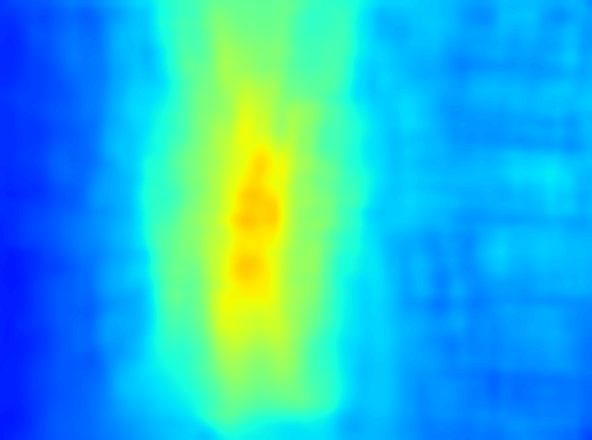}
   }
  }
  \subfigure[Guidance]{%
    \includegraphics[width=.17\columnwidth]{figures/supplementary/41given_image}
  }
   \subfigure[GT]{%
    \includegraphics[width=.17\columnwidth]{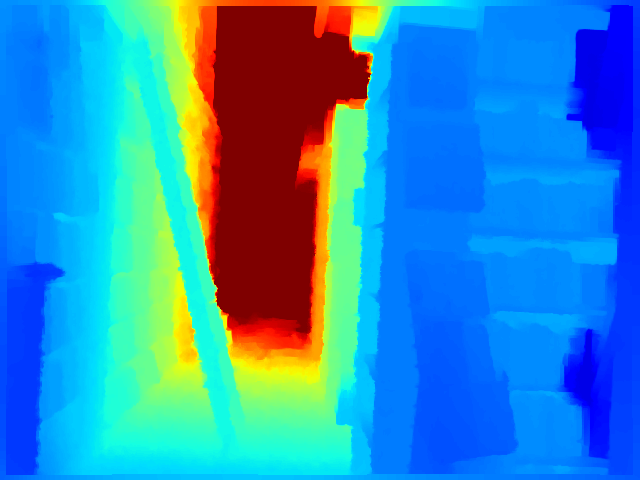}
  }
  \subfigure[Bicubic]{%
    \includegraphics[width=.17\columnwidth]{figures/supplementary/41bicubic}
  }
  \subfigure[Gauss-BF]{%
    \includegraphics[width=.17\columnwidth]{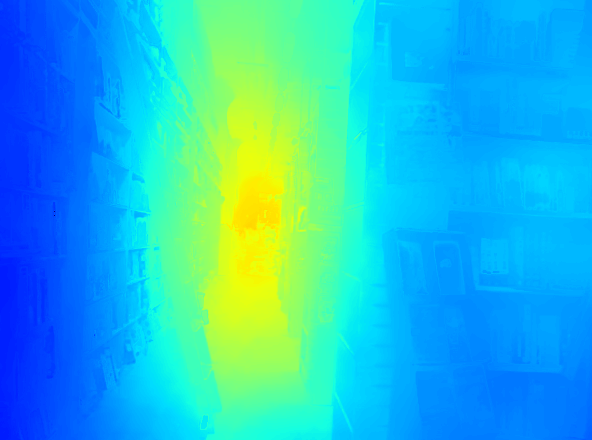}
  }
  \subfigure[Learned-BF]{%
    \includegraphics[width=.17\columnwidth]{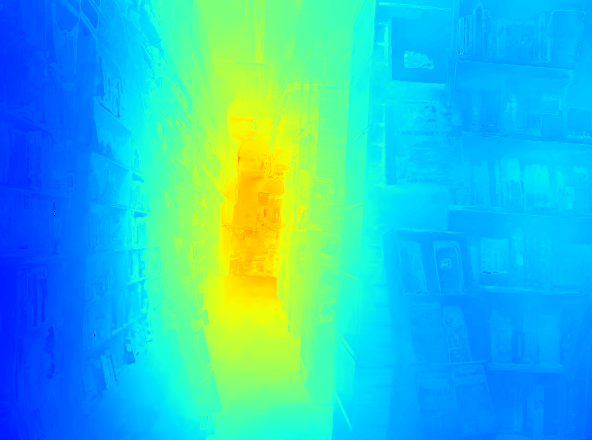}
  }
  }
  \mycaption{Depth Upsampling}{Depth $8\times$ upsampling results
  using different upsampling strategies, from left to right,
  (a)~Low-resolution input depth image (Inp.),
  (b)~High-resolution guidance image, (c)~Ground-truth depth; Upsampled depth images with
  (d)~Bicubic interpolation, (e) Gauss bilateral upsampling and, (f)~Learned bilateral
  updampgling (best viewed on screen).}

\label{fig:depth_upsample_visuals}
\end{figure*}

Figure~\ref{fig:depth_upsample_visuals} presents some more qualitative results comparing bicubic interpolation, Gauss
bilateral and learned bilateral upsampling on NYU depth dataset image~\cite{silberman2012indoor}.

\subsubsection{Character Recognition}\label{sec:app_character}

\begin{figure*}[t!]
  \centering
  \subfigure[LeNet-7]{
    \includegraphics[width=0.7\columnwidth]{figures/supplementary/lenet_cnn_network}
    }\\
    \subfigure[DeepCNet]{
    \includegraphics[width=\columnwidth]{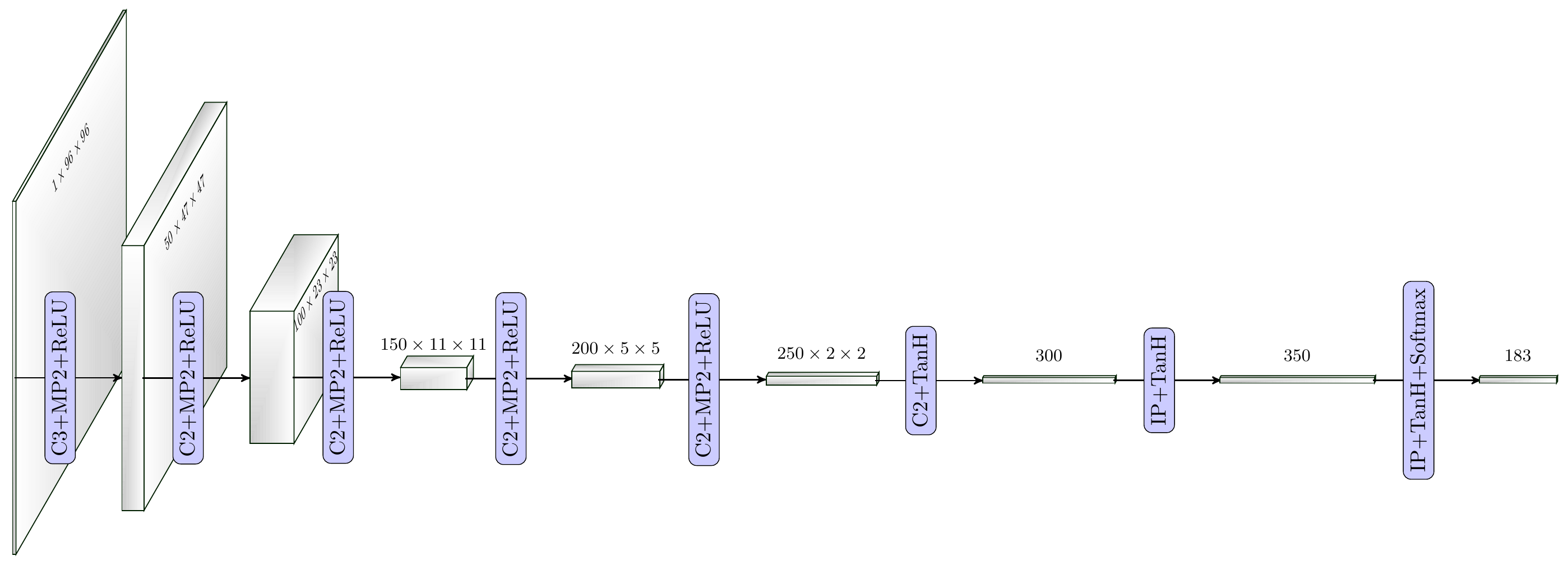}
    }
  \mycaption{CNNs for Character Recognition}
  {Schematic of (top) LeNet-7~\cite{lecun1998mnist} and (bottom) DeepCNet(5,50)~\cite{ciresan2012multi,graham2014spatially} architectures used in Assamese
  character recognition experiments.}
\label{fig:nnrecognition}
\end{figure*}

 Figure~\ref{fig:nnrecognition} shows the schematic of different layers
 of the network architecture for LeNet-7~\cite{lecun1998mnist}
 and DeepCNet(5, 50)~\cite{ciresan2012multi,graham2014spatially}. For the BNN variants, the first layer filters are replaced
 with learned bilateral filters and are learned end-to-end.

\subsubsection{Semantic Segmentation}\label{sec:app_semantic_segmentation}
\label{sec:semantic_bnn_extra}

Some more visual results for semantic segmentation are shown in Figure~\ref{fig:semantic_visuals}.
These include the underlying DeepLab CNN\cite{chen2014semantic} result (DeepLab),
the 2 step mean-field result with Gaussian edge potentials (+2stepMF-GaussCRF)
and also corresponding results with learned edge potentials (+2stepMF-LearnedCRF).
In general, we observe that mean-field in learned CRF leads to slightly dilated
classification regions in comparison to using Gaussian CRF thereby filling-in the
false negative pixels and also correcting some mis-classified regions.

\subsubsection{Material Segmentation}\label{sec:app_material_segmentation}
\label{sec:material_bnn_extra}

In Fig.~\ref{fig:material_visuals-app2}, we present visual results comparing 2 step
mean-field inference with Gaussian and learned pairwise CRF potentials. In
general, we observe that the pixels belonging to dominant classes in the
training data are being more accurately classified with learned CRF. This leads to
a significant improvements in overall pixel accuracy. This also results
in a slight decrease of the accuracy from less frequent class pixels thereby
slightly reducing the average class accuracy with learning. We attribute this
to the type of annotation that is available for this dataset, which is not
for the entire image but for some segments in the image. We have very few
images of the infrequent classes to combat this behaviour during training.

\subsubsection{Experiment Protocols}
\label{sec:protocols}

Table~\ref{tbl:parameters} shows experiment protocols of different experiments.

\definecolor{voc_1}{RGB}{0, 0, 0}
\definecolor{voc_2}{RGB}{128, 0, 0}
\definecolor{voc_3}{RGB}{0, 128, 0}
\definecolor{voc_4}{RGB}{128, 128, 0}
\definecolor{voc_5}{RGB}{0, 0, 128}
\definecolor{voc_6}{RGB}{128, 0, 128}
\definecolor{voc_7}{RGB}{0, 128, 128}
\definecolor{voc_8}{RGB}{128, 128, 128}
\definecolor{voc_9}{RGB}{64, 0, 0}
\definecolor{voc_10}{RGB}{192, 0, 0}
\definecolor{voc_11}{RGB}{64, 128, 0}
\definecolor{voc_12}{RGB}{192, 128, 0}
\definecolor{voc_13}{RGB}{64, 0, 128}
\definecolor{voc_14}{RGB}{192, 0, 128}
\definecolor{voc_15}{RGB}{64, 128, 128}
\definecolor{voc_16}{RGB}{192, 128, 128}
\definecolor{voc_17}{RGB}{0, 64, 0}
\definecolor{voc_18}{RGB}{128, 64, 0}
\definecolor{voc_19}{RGB}{0, 192, 0}
\definecolor{voc_20}{RGB}{128, 192, 0}
\definecolor{voc_21}{RGB}{0, 64, 128}
\definecolor{voc_22}{RGB}{128, 64, 128}

\begin{figure*}[t]
  \centering
  \small{
  \fcolorbox{white}{voc_1}{\rule{0pt}{6pt}\rule{6pt}{0pt}} Background~~
  \fcolorbox{white}{voc_2}{\rule{0pt}{6pt}\rule{6pt}{0pt}} Aeroplane~~
  \fcolorbox{white}{voc_3}{\rule{0pt}{6pt}\rule{6pt}{0pt}} Bicycle~~
  \fcolorbox{white}{voc_4}{\rule{0pt}{6pt}\rule{6pt}{0pt}} Bird~~
  \fcolorbox{white}{voc_5}{\rule{0pt}{6pt}\rule{6pt}{0pt}} Boat~~
  \fcolorbox{white}{voc_6}{\rule{0pt}{6pt}\rule{6pt}{0pt}} Bottle~~
  \fcolorbox{white}{voc_7}{\rule{0pt}{6pt}\rule{6pt}{0pt}} Bus~~
  \fcolorbox{white}{voc_8}{\rule{0pt}{6pt}\rule{6pt}{0pt}} Car~~ \\
  \fcolorbox{white}{voc_9}{\rule{0pt}{6pt}\rule{6pt}{0pt}} Cat~~
  \fcolorbox{white}{voc_10}{\rule{0pt}{6pt}\rule{6pt}{0pt}} Chair~~
  \fcolorbox{white}{voc_11}{\rule{0pt}{6pt}\rule{6pt}{0pt}} Cow~~
  \fcolorbox{white}{voc_12}{\rule{0pt}{6pt}\rule{6pt}{0pt}} Dining Table~~
  \fcolorbox{white}{voc_13}{\rule{0pt}{6pt}\rule{6pt}{0pt}} Dog~~
  \fcolorbox{white}{voc_14}{\rule{0pt}{6pt}\rule{6pt}{0pt}} Horse~~
  \fcolorbox{white}{voc_15}{\rule{0pt}{6pt}\rule{6pt}{0pt}} Motorbike~~
  \fcolorbox{white}{voc_16}{\rule{0pt}{6pt}\rule{6pt}{0pt}} Person~~ \\
  \fcolorbox{white}{voc_17}{\rule{0pt}{6pt}\rule{6pt}{0pt}} Potted Plant~~
  \fcolorbox{white}{voc_18}{\rule{0pt}{6pt}\rule{6pt}{0pt}} Sheep~~
  \fcolorbox{white}{voc_19}{\rule{0pt}{6pt}\rule{6pt}{0pt}} Sofa~~
  \fcolorbox{white}{voc_20}{\rule{0pt}{6pt}\rule{6pt}{0pt}} Train~~
  \fcolorbox{white}{voc_21}{\rule{0pt}{6pt}\rule{6pt}{0pt}} TV monitor~~ \\
  }
  \subfigure{%
    \includegraphics[width=.18\columnwidth]{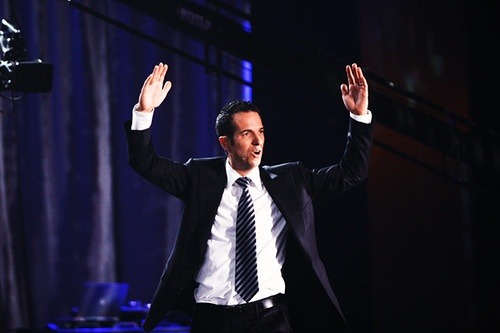}
  }
  \subfigure{%
    \includegraphics[width=.18\columnwidth]{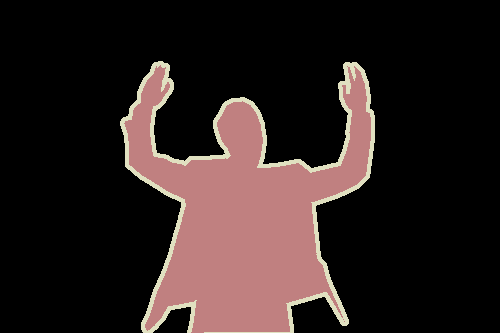}
  }
  \subfigure{%
    \includegraphics[width=.18\columnwidth]{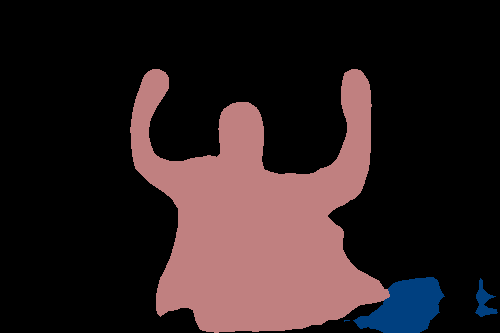}
  }
  \subfigure{%
    \includegraphics[width=.18\columnwidth]{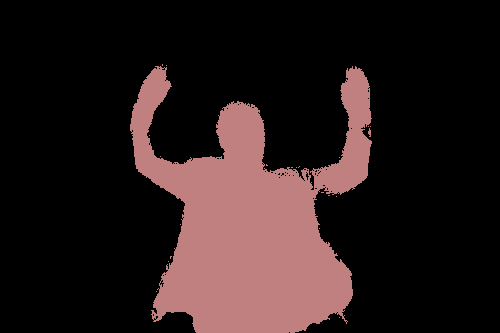}
  }
  \subfigure{%
    \includegraphics[width=.18\columnwidth]{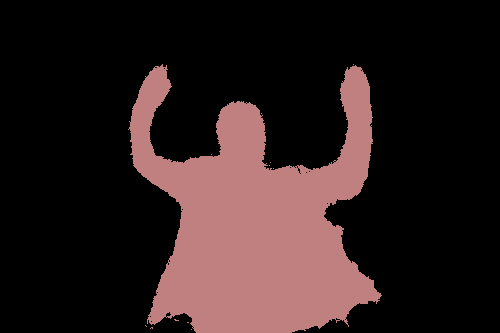}
  }\\
  \subfigure{%
    \includegraphics[width=.18\columnwidth]{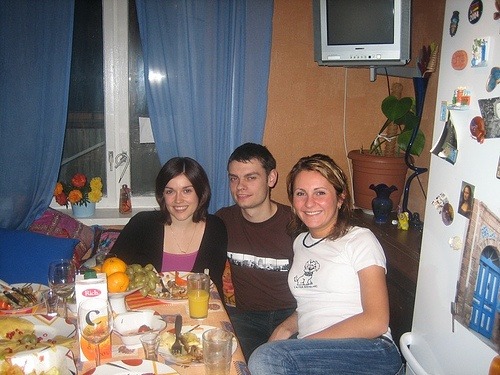}
  }
  \subfigure{%
    \includegraphics[width=.18\columnwidth]{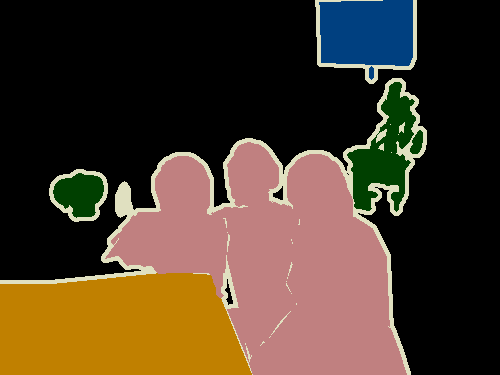}
  }
  \subfigure{%
    \includegraphics[width=.18\columnwidth]{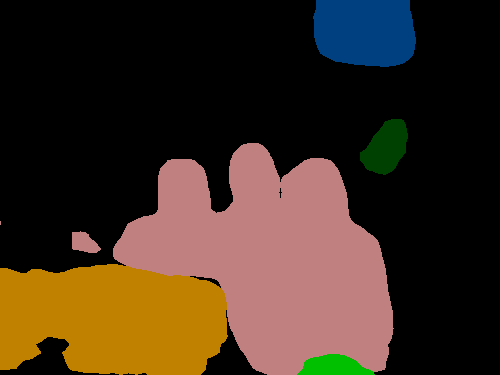}
  }
  \subfigure{%
    \includegraphics[width=.18\columnwidth]{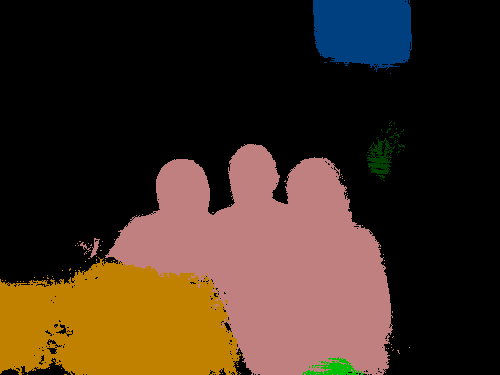}
  }
  \subfigure{%
    \includegraphics[width=.18\columnwidth]{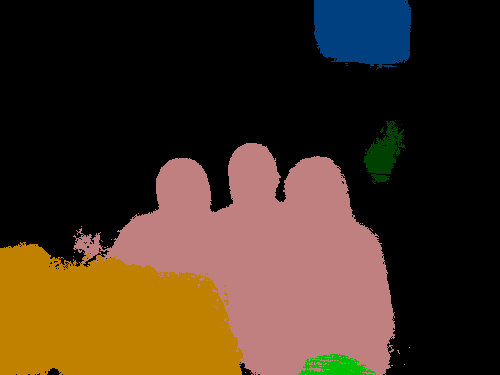}
  }\\
    \subfigure{%
    \includegraphics[width=.18\columnwidth]{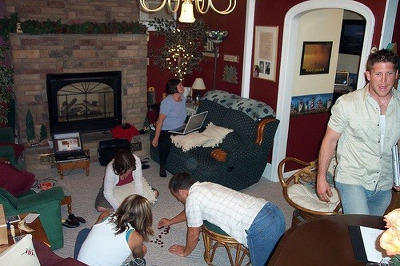}
  }
  \subfigure{%
    \includegraphics[width=.18\columnwidth]{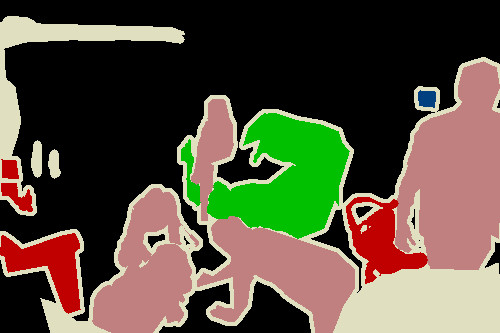}
  }
  \subfigure{%
    \includegraphics[width=.18\columnwidth]{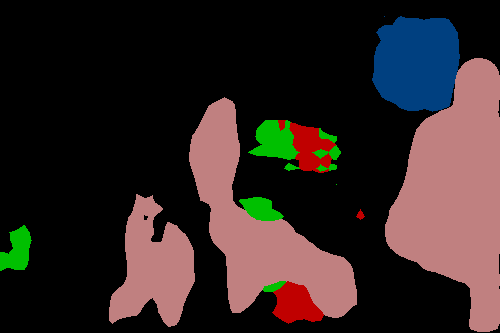}
  }
  \subfigure{%
    \includegraphics[width=.18\columnwidth]{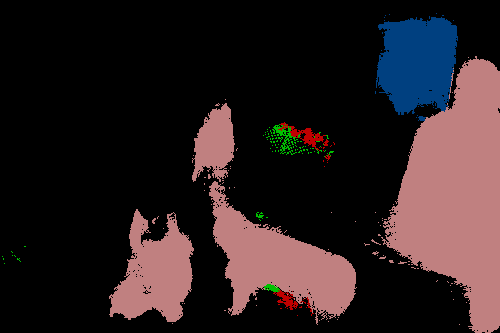}
  }
  \subfigure{%
    \includegraphics[width=.18\columnwidth]{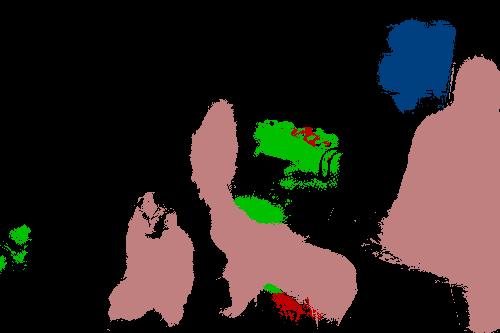}
  }\\
   \subfigure{%
    \includegraphics[width=.18\columnwidth]{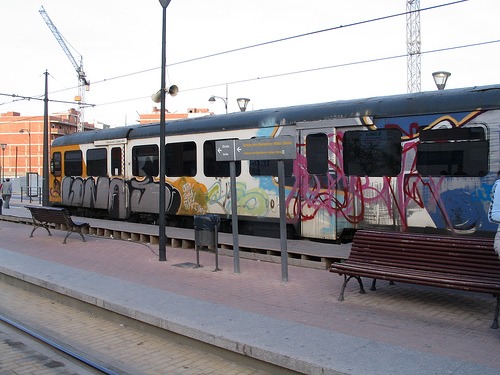}
  }
  \subfigure{%
    \includegraphics[width=.18\columnwidth]{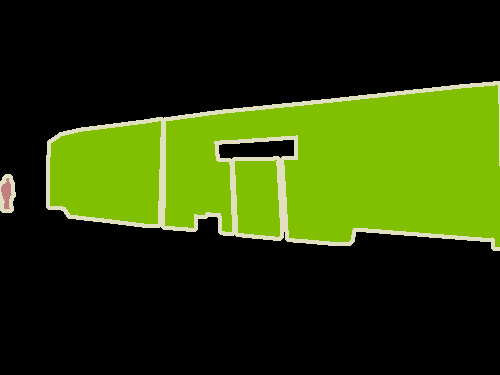}
  }
  \subfigure{%
    \includegraphics[width=.18\columnwidth]{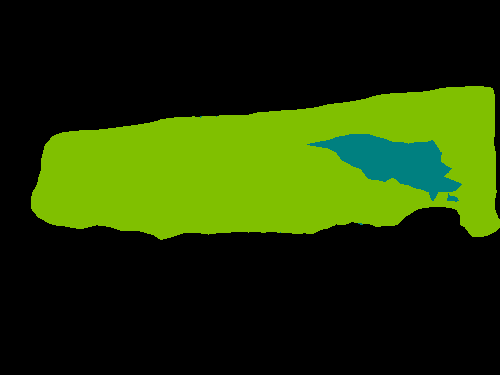}
  }
  \subfigure{%
    \includegraphics[width=.18\columnwidth]{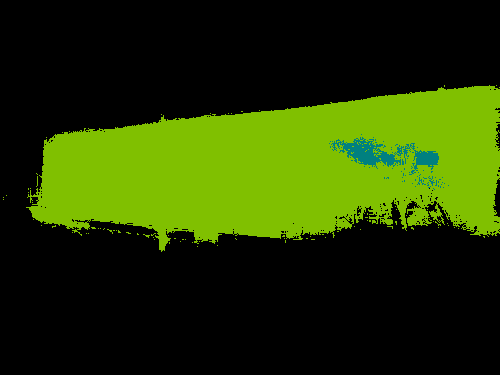}
  }
  \subfigure{%
    \includegraphics[width=.18\columnwidth]{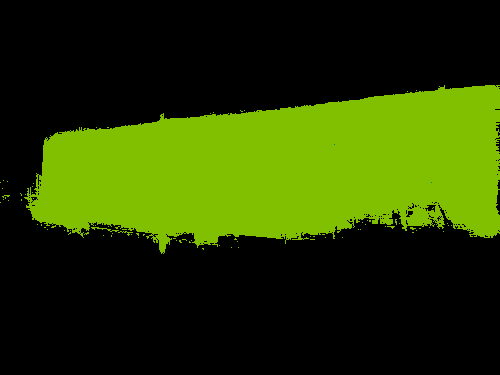}
  }\\
     \subfigure{%
    \includegraphics[width=.18\columnwidth]{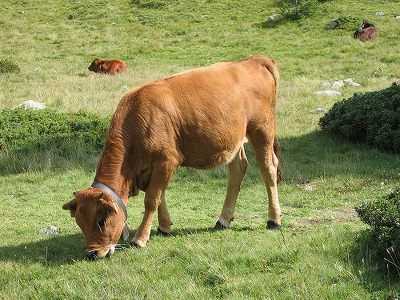}
  }
  \subfigure{%
    \includegraphics[width=.18\columnwidth]{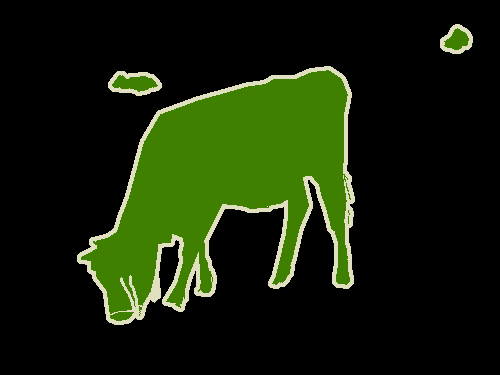}
  }
  \subfigure{%
    \includegraphics[width=.18\columnwidth]{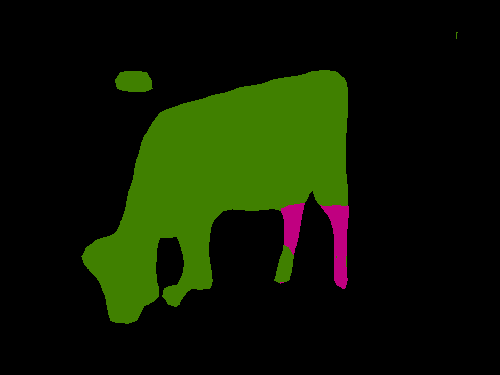}
  }
  \subfigure{%
    \includegraphics[width=.18\columnwidth]{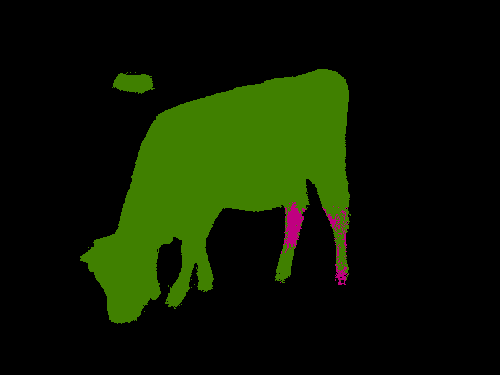}
  }
  \subfigure{%
    \includegraphics[width=.18\columnwidth]{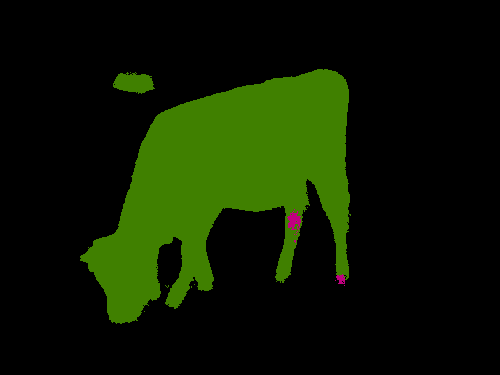}
  }\\
       \subfigure{%
    \includegraphics[width=.18\columnwidth]{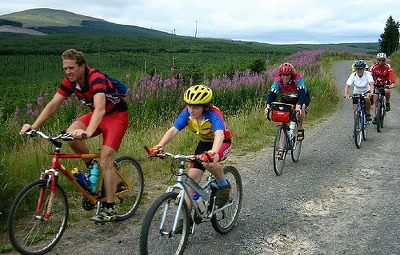}
  }
  \subfigure{%
    \includegraphics[width=.18\columnwidth]{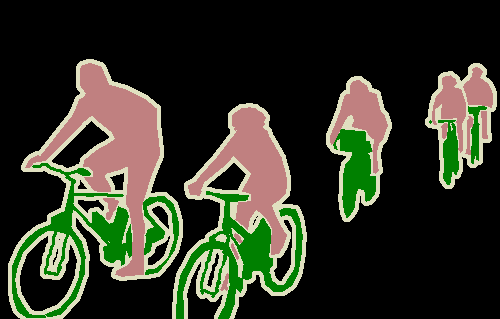}
  }
  \subfigure{%
    \includegraphics[width=.18\columnwidth]{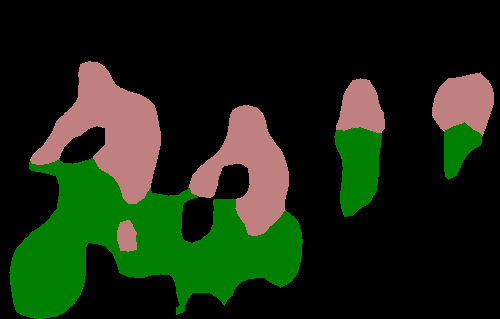}
  }
  \subfigure{%
    \includegraphics[width=.18\columnwidth]{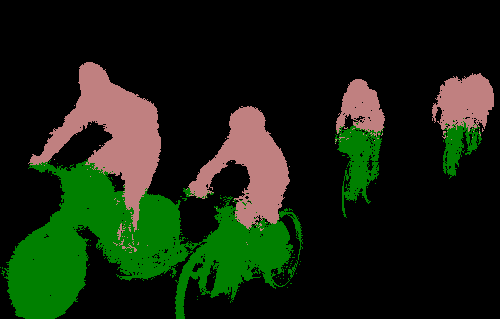}
  }
  \subfigure{%
    \includegraphics[width=.18\columnwidth]{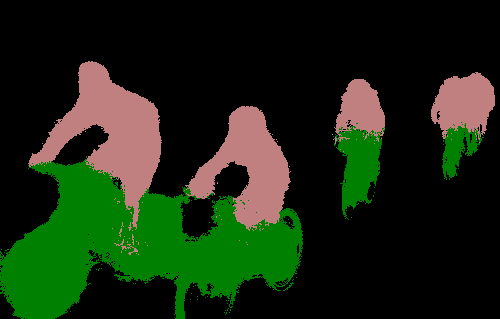}
  }\\
  \setcounter{subfigure}{0}
  \subfigure[Input]{%
    \includegraphics[width=.18\columnwidth]{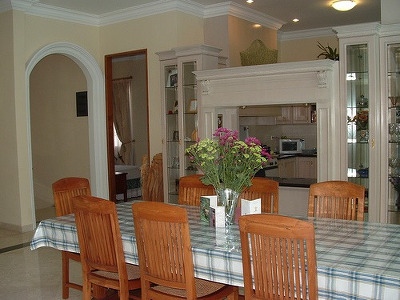}
  }
  \subfigure[Ground Truth]{%
    \includegraphics[width=.18\columnwidth]{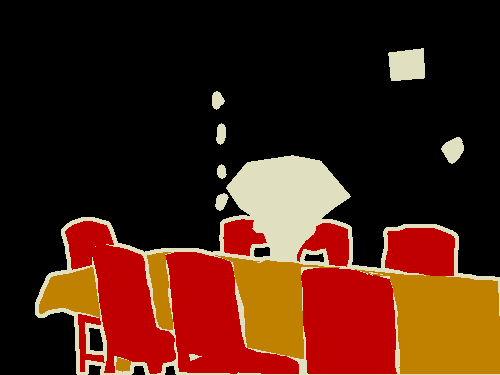}
  }
  \subfigure[DeepLab]{%
    \includegraphics[width=.18\columnwidth]{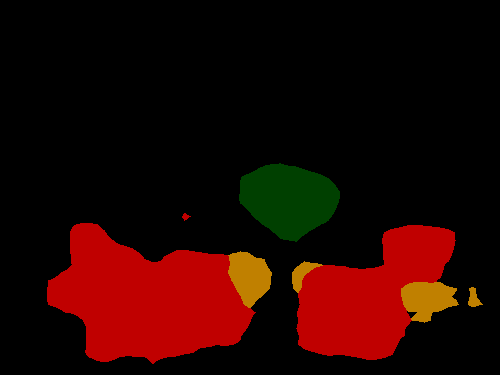}
  }
  \subfigure[+GaussCRF]{%
    \includegraphics[width=.18\columnwidth]{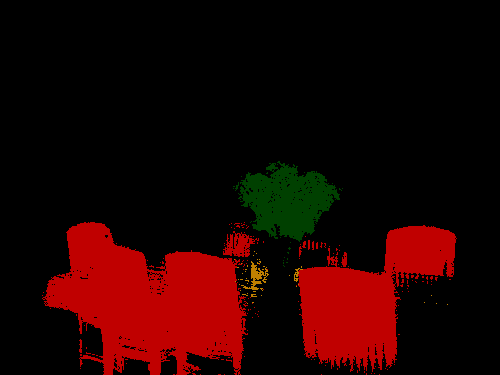}
  }
  \subfigure[+LearnedCRF]{%
    \includegraphics[width=.18\columnwidth]{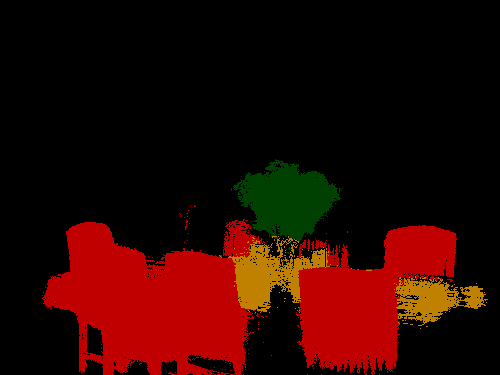}
  }
  \vspace{-0.3cm}
  \mycaption{Semantic Segmentation}{Example results of semantic segmentation.
  (c)~depicts the unary results before application of MF, (d)~after two steps of MF with Gaussian edge CRF potentials, (e)~after
  two steps of MF with learned edge CRF potentials.}
    \label{fig:semantic_visuals}
\end{figure*}

\definecolor{minc_1}{HTML}{771111}
\definecolor{minc_2}{HTML}{CAC690}
\definecolor{minc_3}{HTML}{EEEEEE}
\definecolor{minc_4}{HTML}{7C8FA6}
\definecolor{minc_5}{HTML}{597D31}
\definecolor{minc_6}{HTML}{104410}
\definecolor{minc_7}{HTML}{BB819C}
\definecolor{minc_8}{HTML}{D0CE48}
\definecolor{minc_9}{HTML}{622745}
\definecolor{minc_10}{HTML}{666666}
\definecolor{minc_11}{HTML}{D54A31}
\definecolor{minc_12}{HTML}{101044}
\definecolor{minc_13}{HTML}{444126}
\definecolor{minc_14}{HTML}{75D646}
\definecolor{minc_15}{HTML}{DD4348}
\definecolor{minc_16}{HTML}{5C8577}
\definecolor{minc_17}{HTML}{C78472}
\definecolor{minc_18}{HTML}{75D6D0}
\definecolor{minc_19}{HTML}{5B4586}
\definecolor{minc_20}{HTML}{C04393}
\definecolor{minc_21}{HTML}{D69948}
\definecolor{minc_22}{HTML}{7370D8}
\definecolor{minc_23}{HTML}{7A3622}
\definecolor{minc_24}{HTML}{000000}

\begin{figure*}[t]
  \centering
  \small{
  \fcolorbox{white}{minc_1}{\rule{0pt}{6pt}\rule{6pt}{0pt}} Brick~~
  \fcolorbox{white}{minc_2}{\rule{0pt}{6pt}\rule{6pt}{0pt}} Carpet~~
  \fcolorbox{white}{minc_3}{\rule{0pt}{6pt}\rule{6pt}{0pt}} Ceramic~~
  \fcolorbox{white}{minc_4}{\rule{0pt}{6pt}\rule{6pt}{0pt}} Fabric~~
  \fcolorbox{white}{minc_5}{\rule{0pt}{6pt}\rule{6pt}{0pt}} Foliage~~
  \fcolorbox{white}{minc_6}{\rule{0pt}{6pt}\rule{6pt}{0pt}} Food~~
  \fcolorbox{white}{minc_7}{\rule{0pt}{6pt}\rule{6pt}{0pt}} Glass~~
  \fcolorbox{white}{minc_8}{\rule{0pt}{6pt}\rule{6pt}{0pt}} Hair~~ \\
  \fcolorbox{white}{minc_9}{\rule{0pt}{6pt}\rule{6pt}{0pt}} Leather~~
  \fcolorbox{white}{minc_10}{\rule{0pt}{6pt}\rule{6pt}{0pt}} Metal~~
  \fcolorbox{white}{minc_11}{\rule{0pt}{6pt}\rule{6pt}{0pt}} Mirror~~
  \fcolorbox{white}{minc_12}{\rule{0pt}{6pt}\rule{6pt}{0pt}} Other~~
  \fcolorbox{white}{minc_13}{\rule{0pt}{6pt}\rule{6pt}{0pt}} Painted~~
  \fcolorbox{white}{minc_14}{\rule{0pt}{6pt}\rule{6pt}{0pt}} Paper~~
  \fcolorbox{white}{minc_15}{\rule{0pt}{6pt}\rule{6pt}{0pt}} Plastic~~\\
  \fcolorbox{white}{minc_16}{\rule{0pt}{6pt}\rule{6pt}{0pt}} Polished Stone~~
  \fcolorbox{white}{minc_17}{\rule{0pt}{6pt}\rule{6pt}{0pt}} Skin~~
  \fcolorbox{white}{minc_18}{\rule{0pt}{6pt}\rule{6pt}{0pt}} Sky~~
  \fcolorbox{white}{minc_19}{\rule{0pt}{6pt}\rule{6pt}{0pt}} Stone~~
  \fcolorbox{white}{minc_20}{\rule{0pt}{6pt}\rule{6pt}{0pt}} Tile~~
  \fcolorbox{white}{minc_21}{\rule{0pt}{6pt}\rule{6pt}{0pt}} Wallpaper~~
  \fcolorbox{white}{minc_22}{\rule{0pt}{6pt}\rule{6pt}{0pt}} Water~~
  \fcolorbox{white}{minc_23}{\rule{0pt}{6pt}\rule{6pt}{0pt}} Wood~~ \\
  }
  \subfigure{%
    \includegraphics[width=.18\columnwidth]{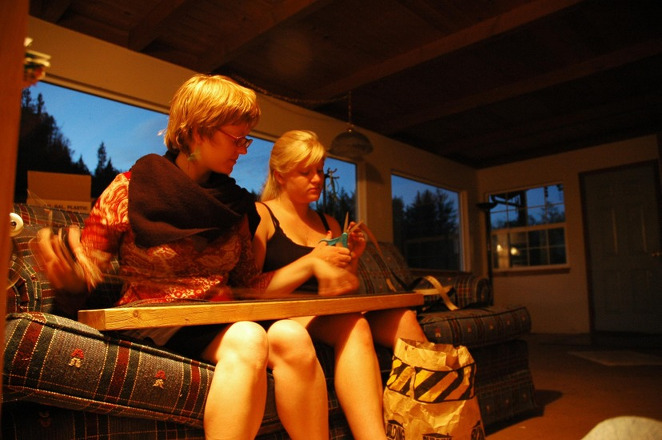}
  }
  \subfigure{%
    \includegraphics[width=.18\columnwidth]{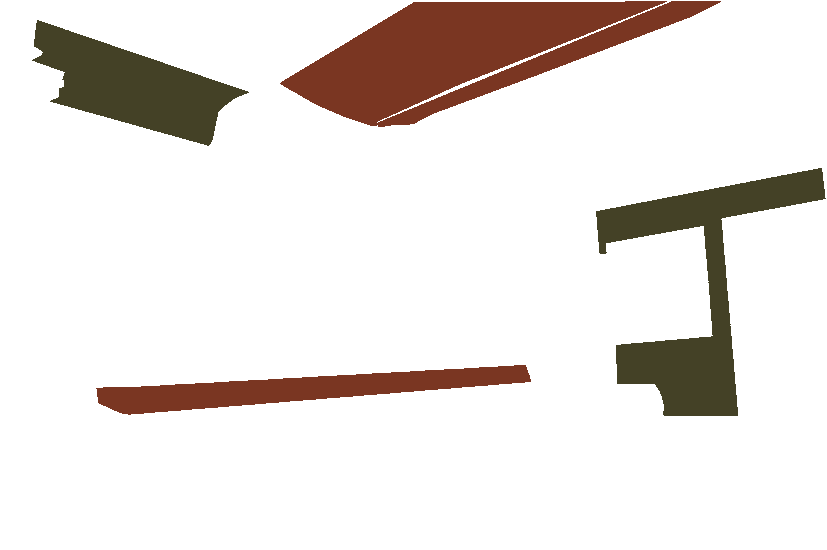}
  }
  \subfigure{%
    \includegraphics[width=.18\columnwidth]{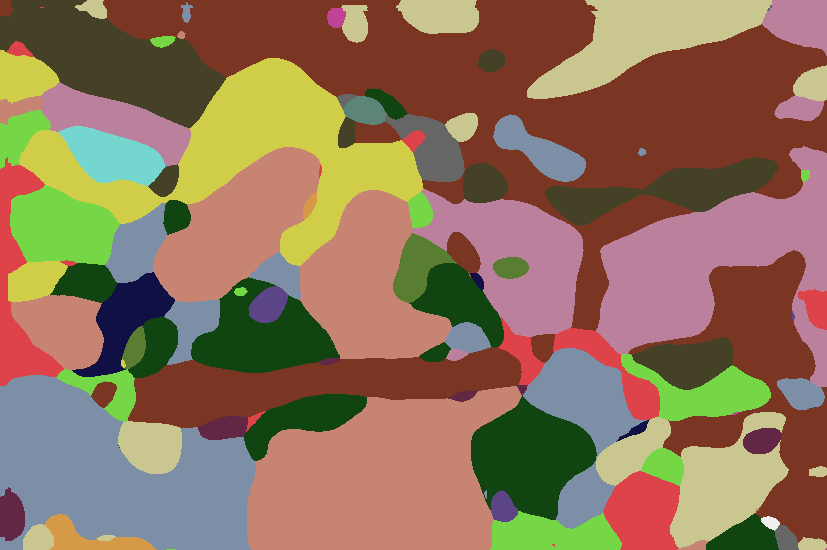}
  }
  \subfigure{%
    \includegraphics[width=.18\columnwidth]{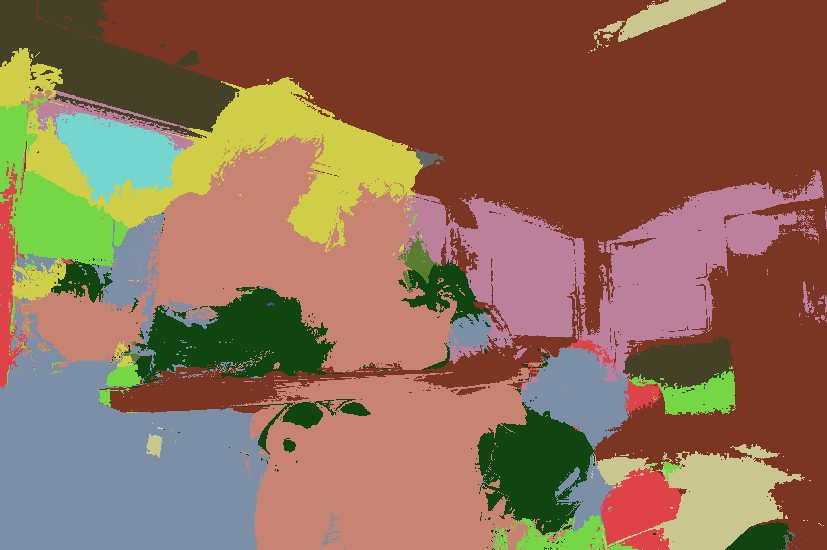}
  }
  \subfigure{%
    \includegraphics[width=.18\columnwidth]{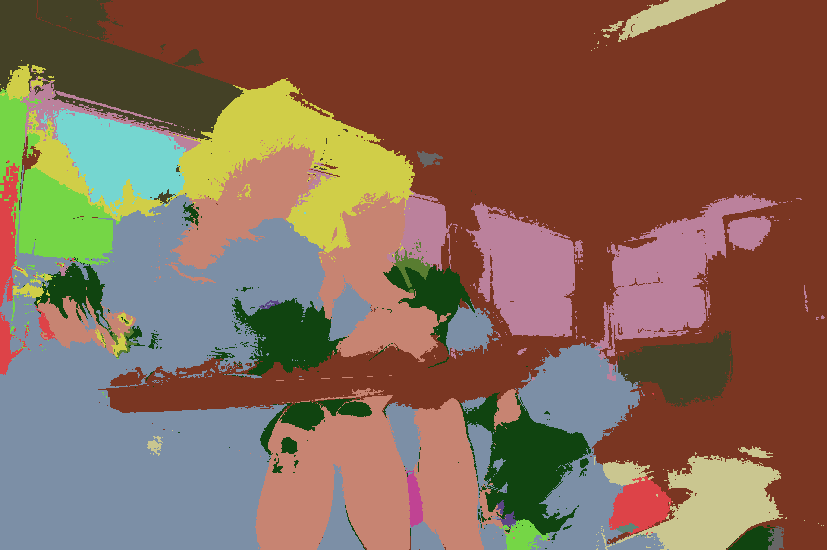}
  }\\[-2ex]
  \subfigure{%
    \includegraphics[width=.18\columnwidth]{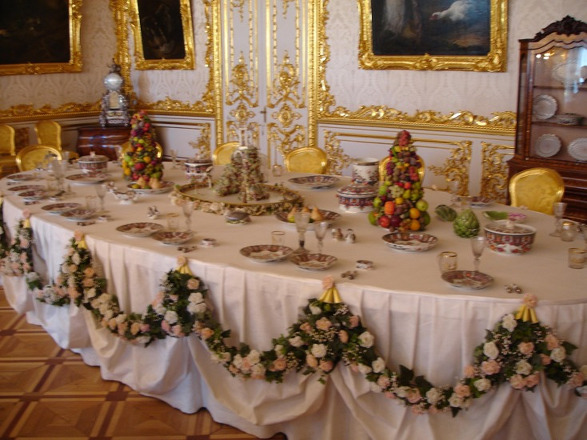}
  }
  \subfigure{%
    \includegraphics[width=.18\columnwidth]{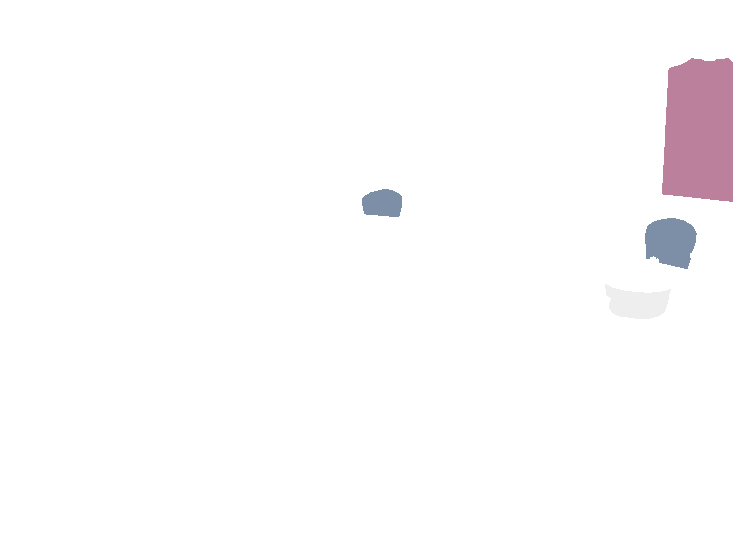}
  }
  \subfigure{%
    \includegraphics[width=.18\columnwidth]{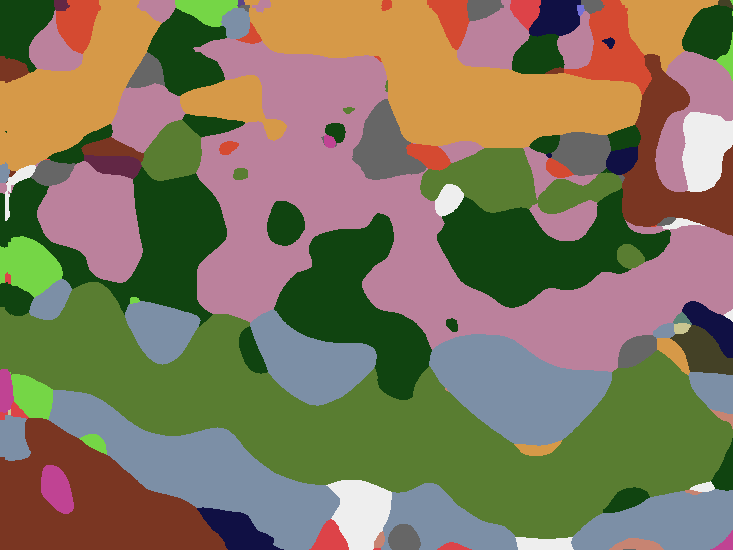}
  }
  \subfigure{%
    \includegraphics[width=.18\columnwidth]{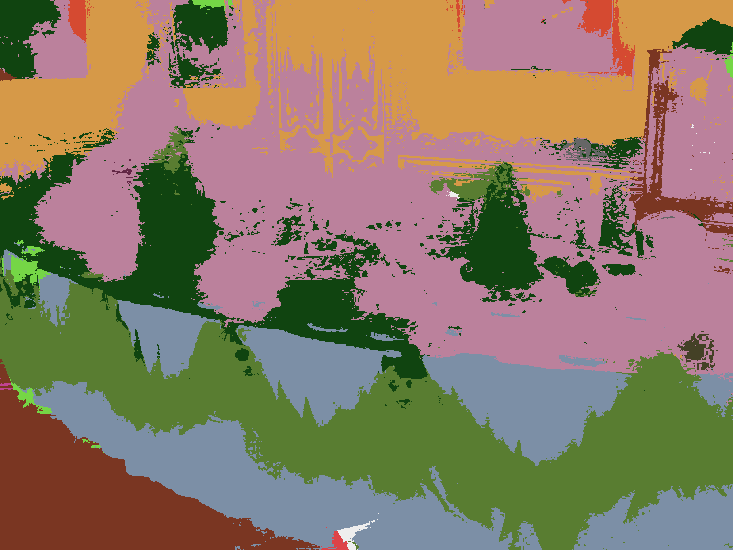}
  }
  \subfigure{%
    \includegraphics[width=.18\columnwidth]{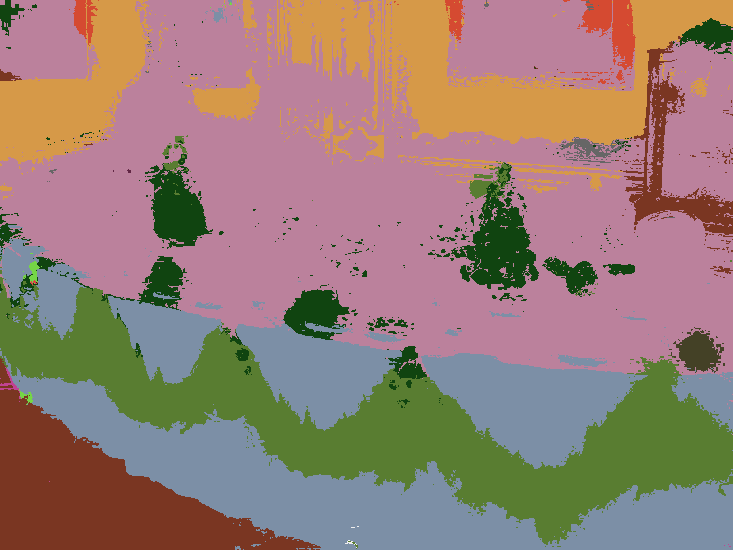}
  }\\[-2ex]
    \subfigure{%
    \includegraphics[width=.18\columnwidth]{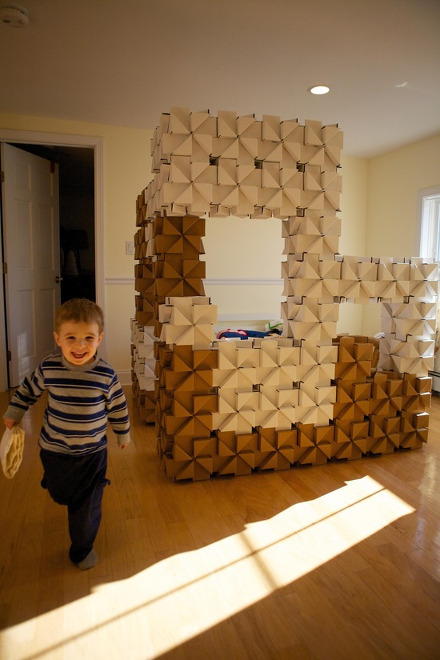}
  }
  \subfigure{%
    \includegraphics[width=.18\columnwidth]{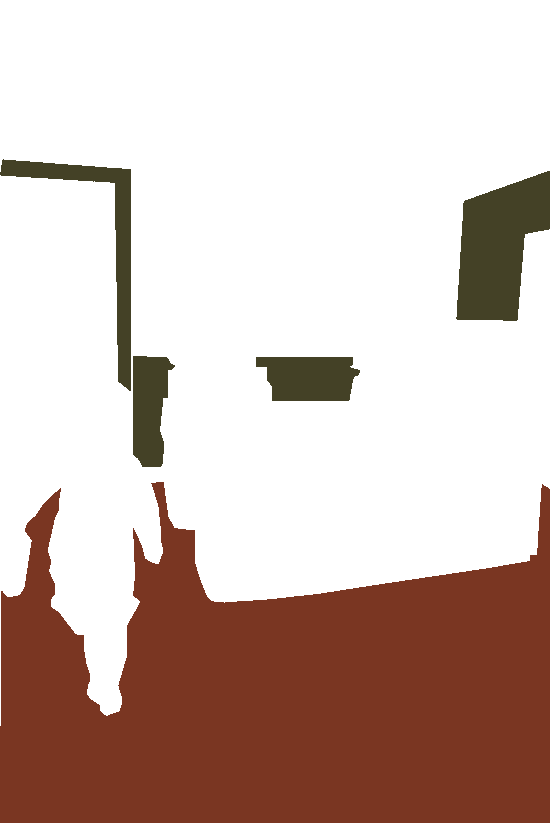}
  }
  \subfigure{%
    \includegraphics[width=.18\columnwidth]{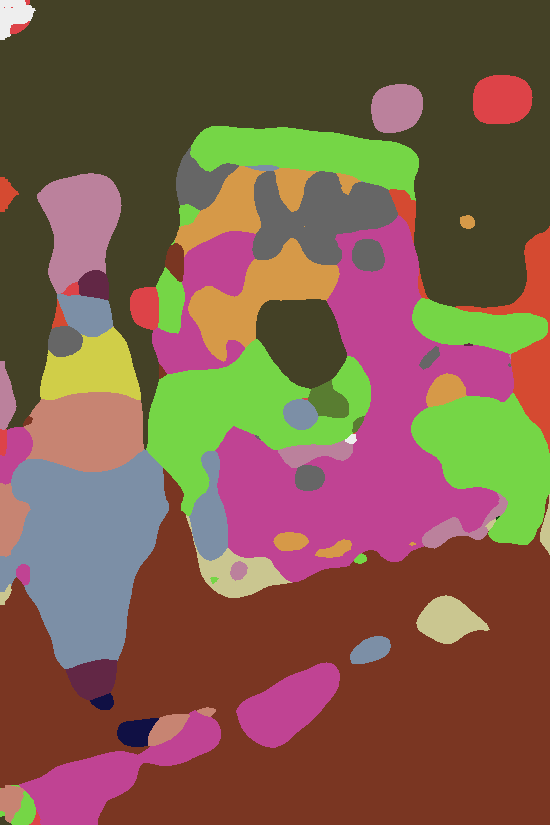}
  }
  \subfigure{%
    \includegraphics[width=.18\columnwidth]{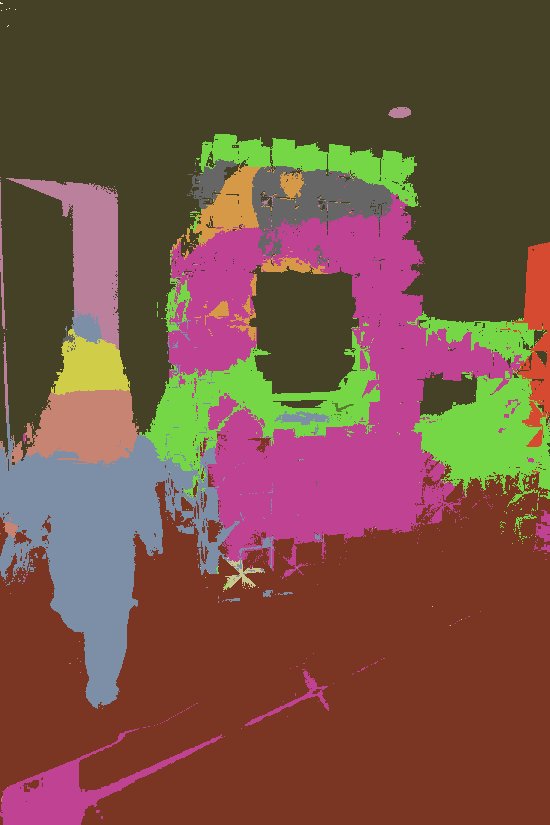}
  }
  \subfigure{%
    \includegraphics[width=.18\columnwidth]{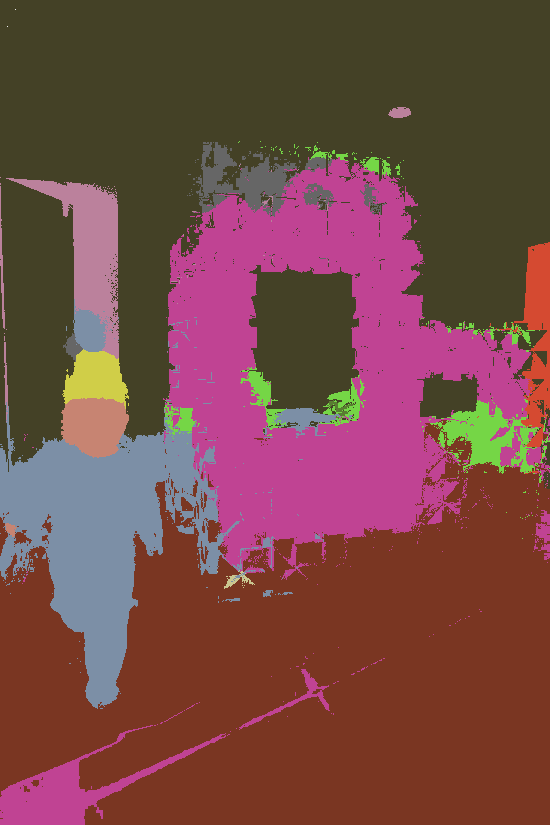}
  }\\[-2ex]
   \subfigure{%
    \includegraphics[width=.18\columnwidth]{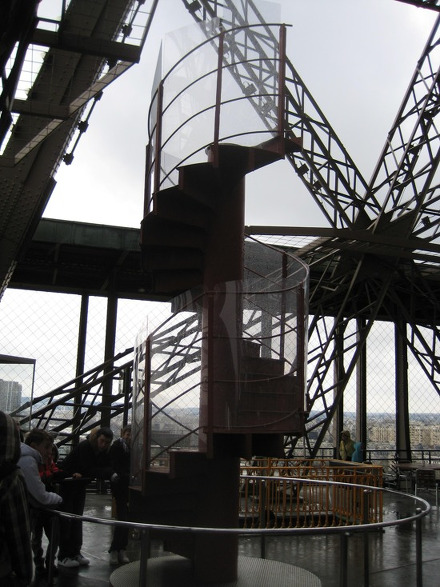}
  }
  \subfigure{%
    \includegraphics[width=.18\columnwidth]{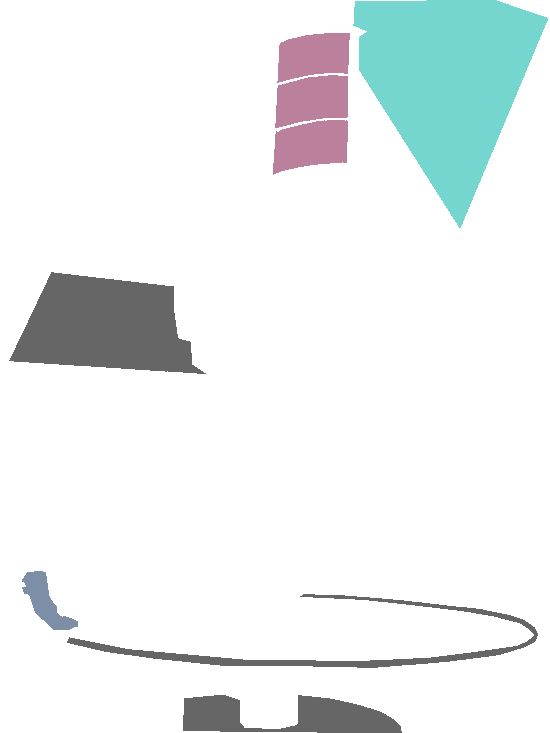}
  }
  \subfigure{%
    \includegraphics[width=.18\columnwidth]{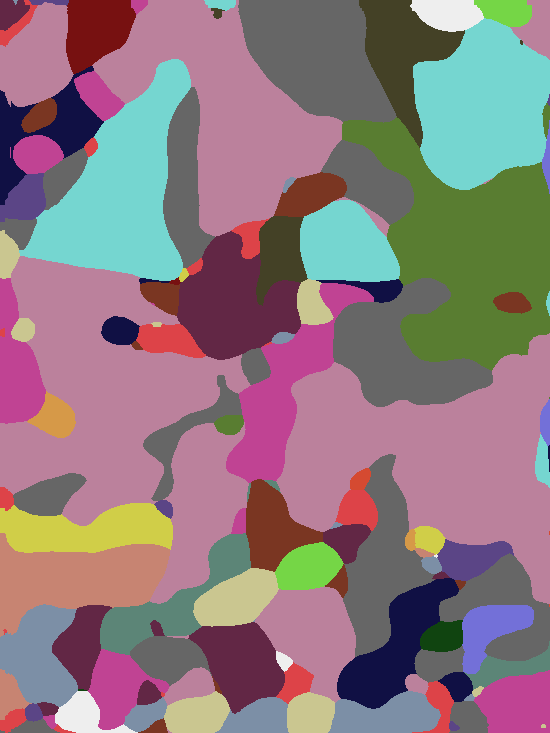}
  }
  \subfigure{%
    \includegraphics[width=.18\columnwidth]{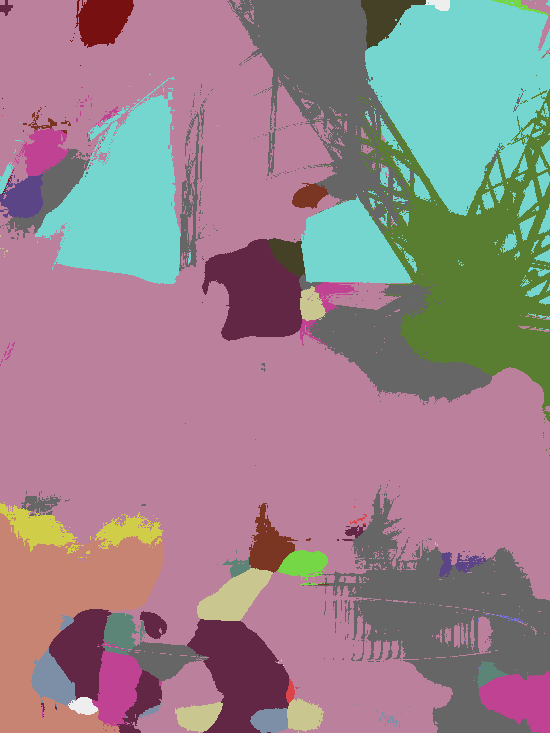}
  }
  \subfigure{%
    \includegraphics[width=.18\columnwidth]{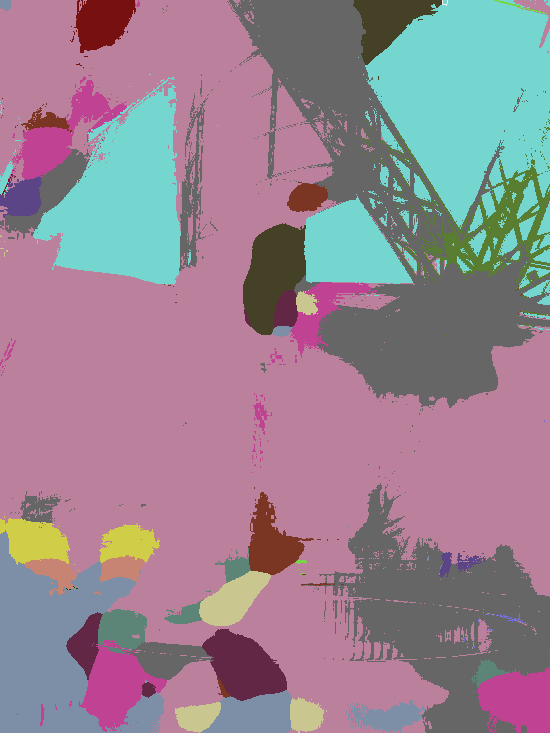}
  }\\[-2ex]
  \setcounter{subfigure}{0}
  \subfigure[Input]{%
    \includegraphics[width=.18\columnwidth]{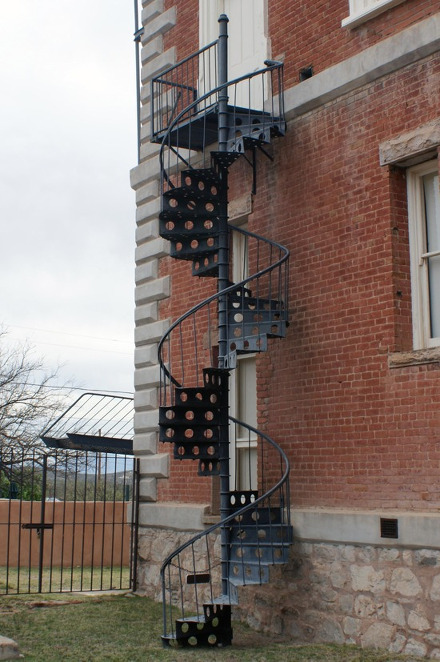}
  }
  \subfigure[Ground Truth]{%
    \includegraphics[width=.18\columnwidth]{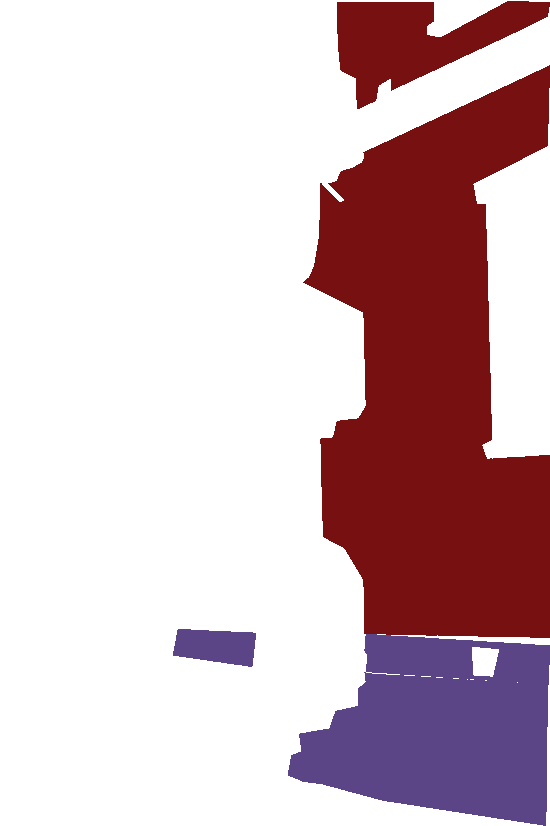}
  }
  \subfigure[DeepLab]{%
    \includegraphics[width=.18\columnwidth]{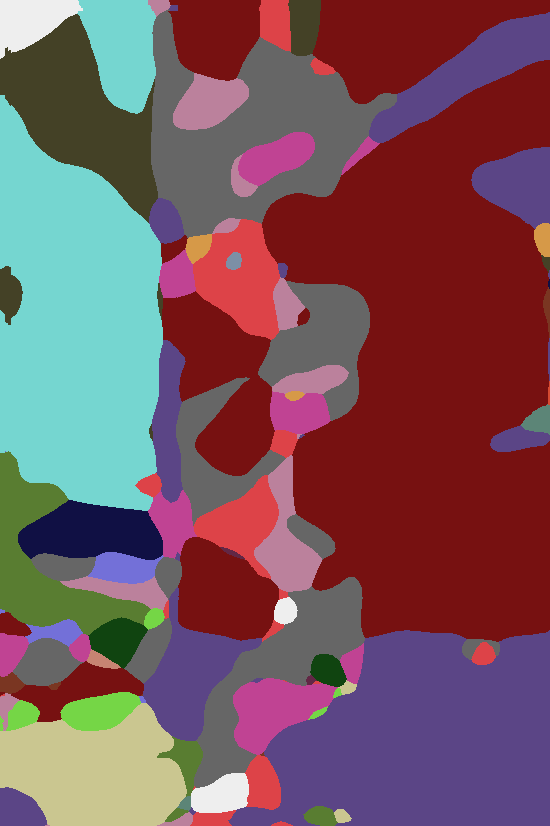}
  }
  \subfigure[+GaussCRF]{%
    \includegraphics[width=.18\columnwidth]{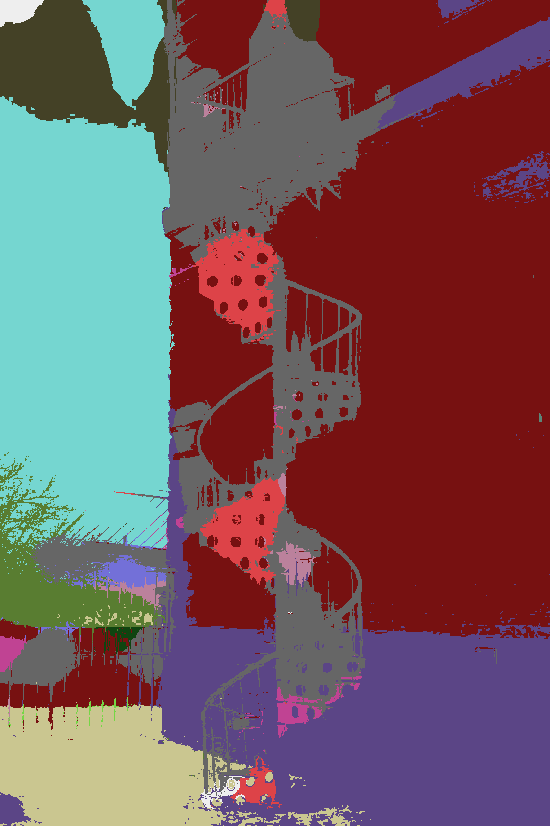}
  }
  \subfigure[+LearnedCRF]{%
    \includegraphics[width=.18\columnwidth]{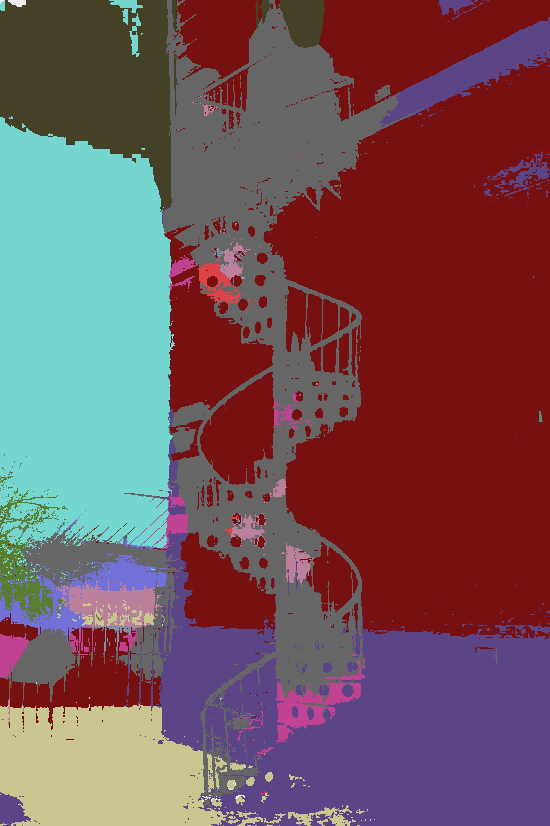}
  }
  \mycaption{Material Segmentation}{Example results of material segmentation.
  (c)~depicts the unary results before application of MF, (d)~after two steps of MF with Gaussian edge CRF potentials, (e)~after two steps of MF with learned edge CRF potentials.}
    \label{fig:material_visuals-app2}
\end{figure*}

\begin{table*}[h]
\tiny
  \centering
    \begin{tabular}{L{2.3cm} L{2.25cm} C{1.5cm} C{0.7cm} C{0.6cm} C{0.7cm} C{0.7cm} C{0.7cm} C{1.6cm} C{0.6cm} C{0.6cm} C{0.6cm}}
      \toprule
& & & & & \multicolumn{3}{c}{\textbf{Data Statistics}} & \multicolumn{4}{c}{\textbf{Training Protocol}} \\

\textbf{Experiment} & \textbf{Feature Types} & \textbf{Feature Scales} & \textbf{Filter Size} & \textbf{Filter Nbr.} & \textbf{Train}  & \textbf{Val.} & \textbf{Test} & \textbf{Loss Type} & \textbf{LR} & \textbf{Batch} & \textbf{Epochs} \\
      \midrule
      \multicolumn{2}{c}{\textbf{Single Bilateral Filter Applications}} & & & & & & & & & \\
      \textbf{2$\times$ Color Upsampling} & Position$_{1}$, Intensity (3D) & 0.13, 0.17 & 65 & 2 & 10581 & 1449 & 1456 & MSE & 1e-06 & 200 & 94.5\\
      \textbf{4$\times$ Color Upsampling} & Position$_{1}$, Intensity (3D) & 0.06, 0.17 & 65 & 2 & 10581 & 1449 & 1456 & MSE & 1e-06 & 200 & 94.5\\
      \textbf{8$\times$ Color Upsampling} & Position$_{1}$, Intensity (3D) & 0.03, 0.17 & 65 & 2 & 10581 & 1449 & 1456 & MSE & 1e-06 & 200 & 94.5\\
      \textbf{16$\times$ Color Upsampling} & Position$_{1}$, Intensity (3D) & 0.02, 0.17 & 65 & 2 & 10581 & 1449 & 1456 & MSE & 1e-06 & 200 & 94.5\\
      \textbf{Depth Upsampling} & Position$_{1}$, Color (5D) & 0.05, 0.02 & 665 & 2 & 795 & 100 & 654 & MSE & 1e-07 & 50 & 251.6\\
      \textbf{Mesh Denoising} & Isomap (4D) & 46.00 & 63 & 2 & 1000 & 200 & 500 & MSE & 100 & 10 & 100.0 \\
      \midrule
      \multicolumn{2}{c}{\textbf{DenseCRF Applications}} & & & & & & & & &\\
      \multicolumn{2}{l}{\textbf{Semantic Segmentation}} & & & & & & & & &\\
      \textbf{- 1step MF} & Position$_{1}$, Color (5D); Position$_{1}$ (2D) & 0.01, 0.34; 0.34  & 665; 19  & 2; 2 & 10581 & 1449 & 1456 & Logistic & 0.1 & 5 & 1.4 \\
      \textbf{- 2step MF} & Position$_{1}$, Color (5D); Position$_{1}$ (2D) & 0.01, 0.34; 0.34 & 665; 19 & 2; 2 & 10581 & 1449 & 1456 & Logistic & 0.1 & 5 & 1.4 \\
      \textbf{- \textit{loose} 2step MF} & Position$_{1}$, Color (5D); Position$_{1}$ (2D) & 0.01, 0.34; 0.34 & 665; 19 & 2; 2 &10581 & 1449 & 1456 & Logistic & 0.1 & 5 & +1.9  \\ \\
      \multicolumn{2}{l}{\textbf{Material Segmentation}} & & & & & & & & &\\
      \textbf{- 1step MF} & Position$_{2}$, Lab-Color (5D) & 5.00, 0.05, 0.30  & 665 & 2 & 928 & 150 & 1798 & Weighted Logistic & 1e-04 & 24 & 2.6 \\
      \textbf{- 2step MF} & Position$_{2}$, Lab-Color (5D) & 5.00, 0.05, 0.30 & 665 & 2 & 928 & 150 & 1798 & Weighted Logistic & 1e-04 & 12 & +0.7 \\
      \textbf{- \textit{loose} 2step MF} & Position$_{2}$, Lab-Color (5D) & 5.00, 0.05, 0.30 & 665 & 2 & 928 & 150 & 1798 & Weighted Logistic & 1e-04 & 12 & +0.2\\
      \midrule
      \multicolumn{2}{c}{\textbf{Neural Network Applications}} & & & & & & & & &\\
      \textbf{Tiles: CNN-9$\times$9} & - & - & 81 & 4 & 10000 & 1000 & 1000 & Logistic & 0.01 & 100 & 500.0 \\
      \textbf{Tiles: CNN-13$\times$13} & - & - & 169 & 6 & 10000 & 1000 & 1000 & Logistic & 0.01 & 100 & 500.0 \\
      \textbf{Tiles: CNN-17$\times$17} & - & - & 289 & 8 & 10000 & 1000 & 1000 & Logistic & 0.01 & 100 & 500.0 \\
      \textbf{Tiles: CNN-21$\times$21} & - & - & 441 & 10 & 10000 & 1000 & 1000 & Logistic & 0.01 & 100 & 500.0 \\
      \textbf{Tiles: BNN} & Position$_{1}$, Color (5D) & 0.05, 0.04 & 63 & 1 & 10000 & 1000 & 1000 & Logistic & 0.01 & 100 & 30.0 \\
      \textbf{LeNet} & - & - & 25 & 2 & 5490 & 1098 & 1647 & Logistic & 0.1 & 100 & 182.2 \\
      \textbf{Crop-LeNet} & - & - & 25 & 2 & 5490 & 1098 & 1647 & Logistic & 0.1 & 100 & 182.2 \\
      \textbf{BNN-LeNet} & Position$_{2}$ (2D) & 20.00 & 7 & 1 & 5490 & 1098 & 1647 & Logistic & 0.1 & 100 & 182.2 \\
      \textbf{DeepCNet} & - & - & 9 & 1 & 5490 & 1098 & 1647 & Logistic & 0.1 & 100 & 182.2 \\
      \textbf{Crop-DeepCNet} & - & - & 9 & 1 & 5490 & 1098 & 1647 & Logistic & 0.1 & 100 & 182.2 \\
      \textbf{BNN-DeepCNet} & Position$_{2}$ (2D) & 40.00  & 7 & 1 & 5490 & 1098 & 1647 & Logistic & 0.1 & 100 & 182.2 \\
      \bottomrule
      \\
    \end{tabular}
    \mycaption{Experiment Protocols} {Experiment protocols for the different experiments presented in this work. \textbf{Feature Types}:
    Feature spaces used for the bilateral convolutions. Position$_1$ corresponds to un-normalized pixel positions whereas Position$_2$ corresponds
    to pixel positions normalized to $[0,1]$ with respect to the given image. \textbf{Feature Scales}: Cross-validated scales for the features used.
     \textbf{Filter Size}: Number of elements in the filter that is being learned. \textbf{Filter Nbr.}: Half-width of the filter. \textbf{Train},
     \textbf{Val.} and \textbf{Test} corresponds to the number of train, validation and test images used in the experiment. \textbf{Loss Type}: Type
     of loss used for back-propagation. ``MSE'' corresponds to Euclidean mean squared error loss and ``Logistic'' corresponds to multinomial logistic
     loss. ``Weighted Logistic'' is the class-weighted multinomial logistic loss. We weighted the loss with inverse class probability for material
     segmentation task due to the small availability of training data with class imbalance. \textbf{LR}: Fixed learning rate used in stochastic gradient
     descent. \textbf{Batch}: Number of images used in one parameter update step. \textbf{Epochs}: Number of training epochs. In all the experiments,
     we used fixed momentum of 0.9 and weight decay of 0.0005 for stochastic gradient descent. ```Color Upsampling'' experiments in this Table corresponds
     to those performed on Pascal VOC12 dataset images. For all experiments using Pascal VOC12 images, we use extended
     training segmentation dataset available from~\cite{hariharan2011moredata}, and used standard validation and test splits
     from the main dataset~\cite{voc2012segmentation}.}
  \label{tbl:parameters}
\end{table*}

\clearpage

\section{Parameters and Additional Results for Video Propagation Networks}

\begin{figure}[th!]
\begin{center}
  \centerline{\includegraphics[width=0.7\textwidth]{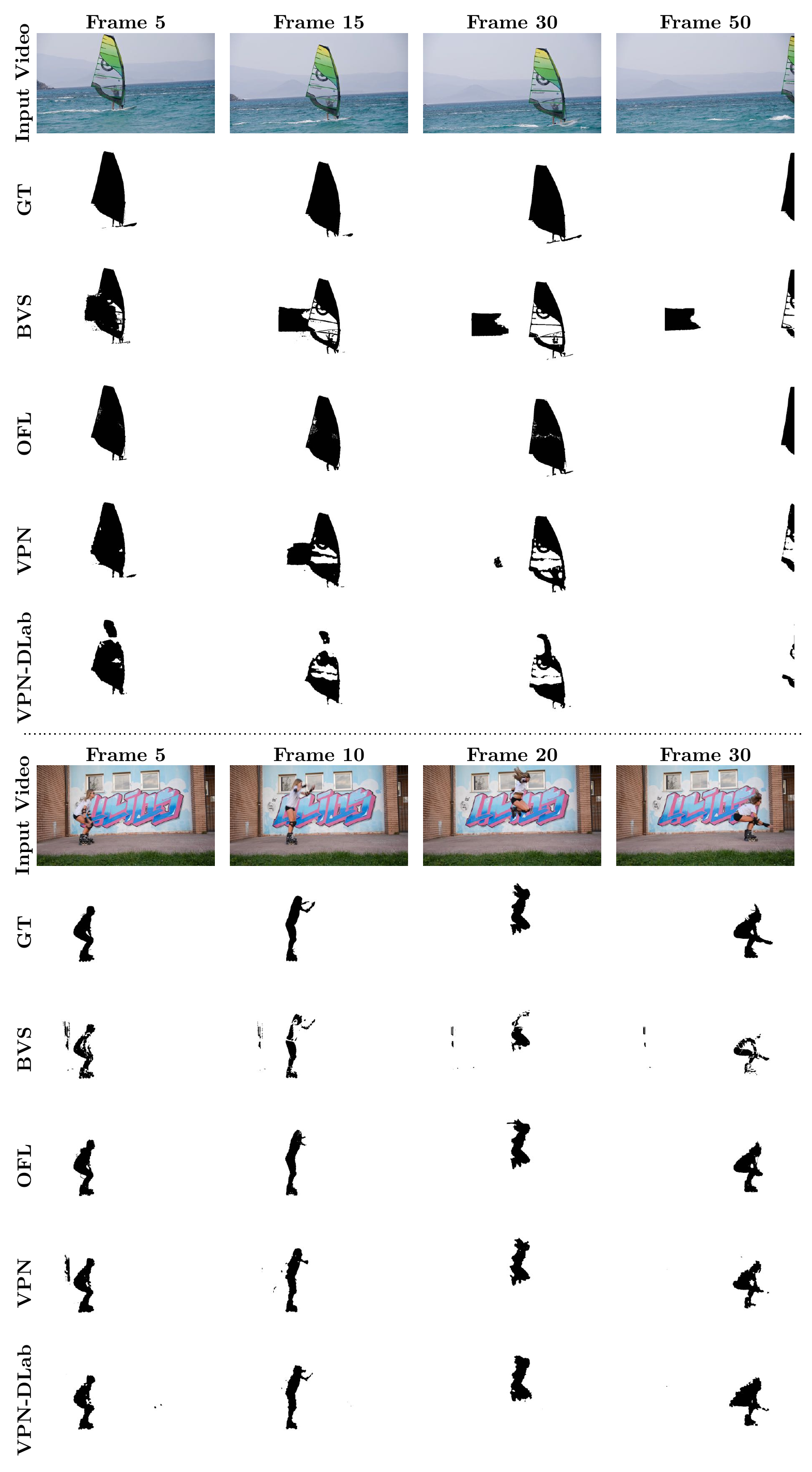}}
    \mycaption{Video Object Segmentation}
    {Shown are the different frames in example videos with the corresponding
    ground truth (GT) masks, predictions from BVS~\cite{marki2016bilateral},
    OFL~\cite{tsaivideo}, VPN (VPN-Stage2) and VPN-DLab (VPN-DeepLab) models.}
    \label{fig:video_seg_pos_supp}
\end{center}
\vspace{-1.0cm}
\end{figure}

\begin{figure}[th!]
\begin{center}
  \centerline{\includegraphics[width=0.7\textwidth]{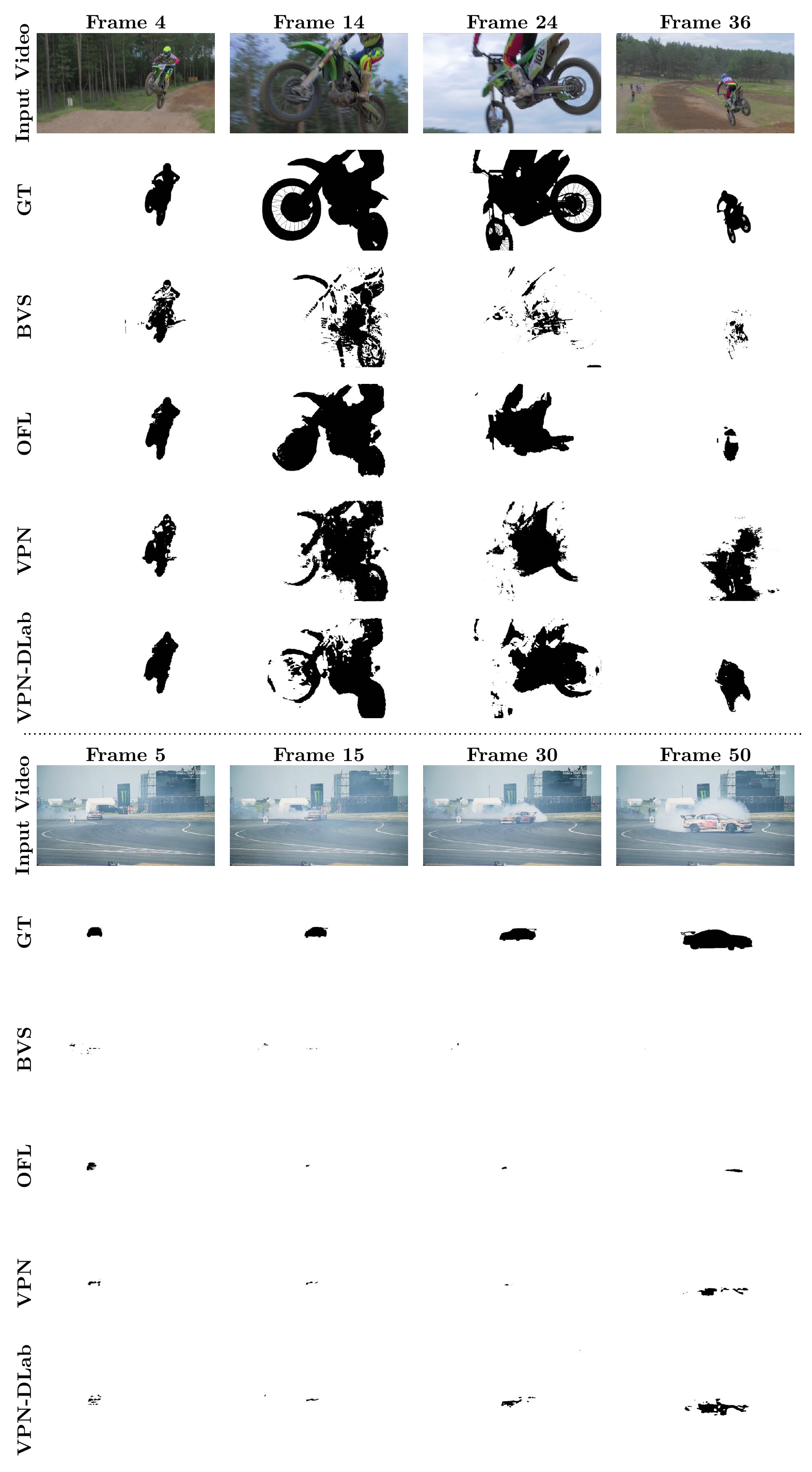}}
    \mycaption{Failure Cases for Video Object Segmentation}
    {Shown are the different frames in example videos with the corresponding
    ground truth (GT) masks, predictions from BVS~\cite{marki2016bilateral},
    OFL~\cite{tsaivideo}, VPN (VPN-Stage2) and VPN-DLab (VPN-DeepLab) models.}
    \label{fig:video_seg_neg_supp}
\end{center}
\vspace{-1.0cm}
\end{figure}

\begin{figure}[th!]
\begin{center}
  \centerline{\includegraphics[width=0.9\textwidth]{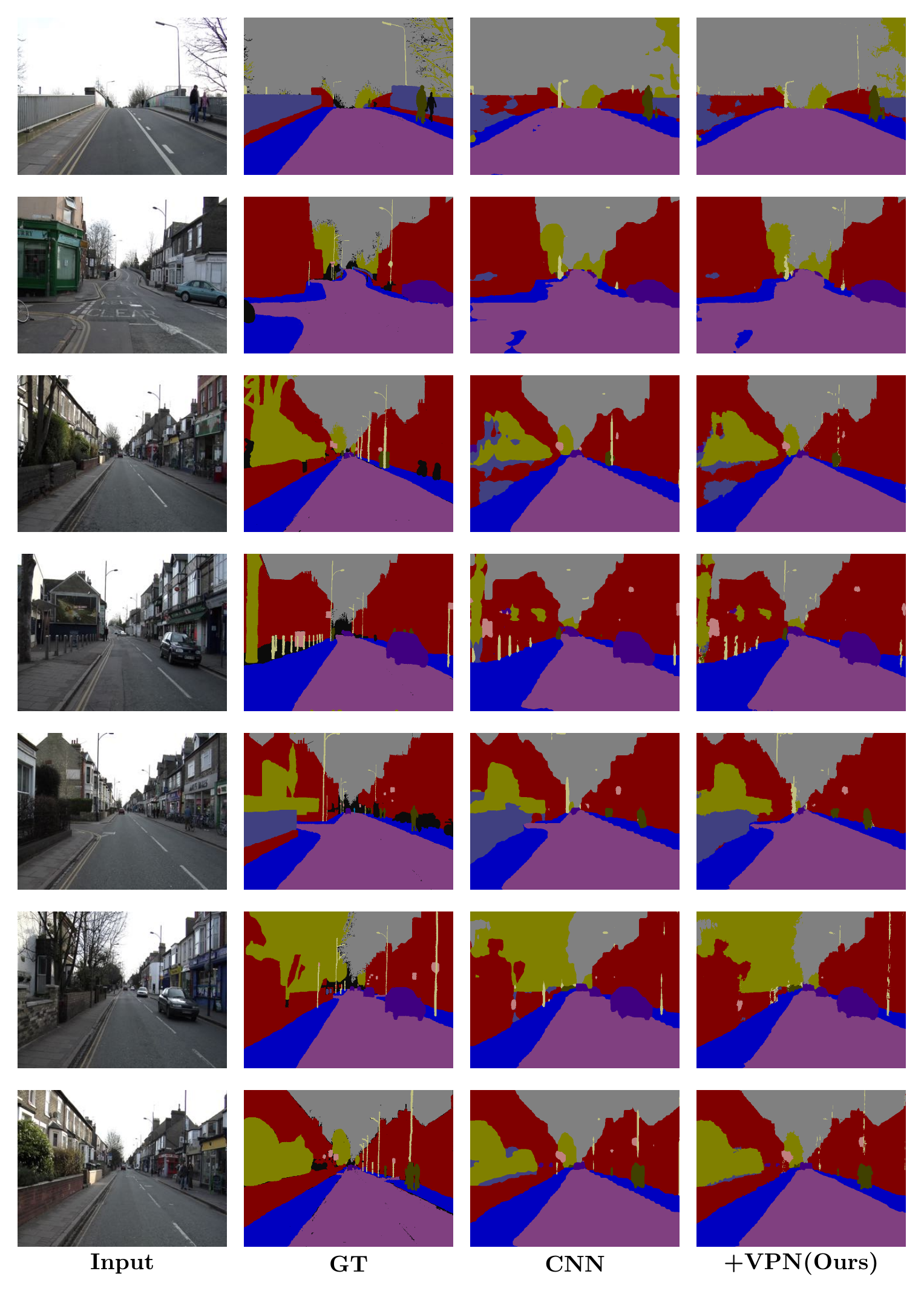}}
    \mycaption{Semantic Video Segmentation}
    {Input video frames and the corresponding ground truth (GT)
    segmentation together with the predictions of CNN~\cite{yu2015multi} and with
    VPN-Flow.}
    \label{fig:semantic_visuals_supp}
\end{center}
\vspace{-0.7cm}
\end{figure}

\begin{figure}[th!]
\begin{center}
  \centerline{\includegraphics[width=\textwidth]{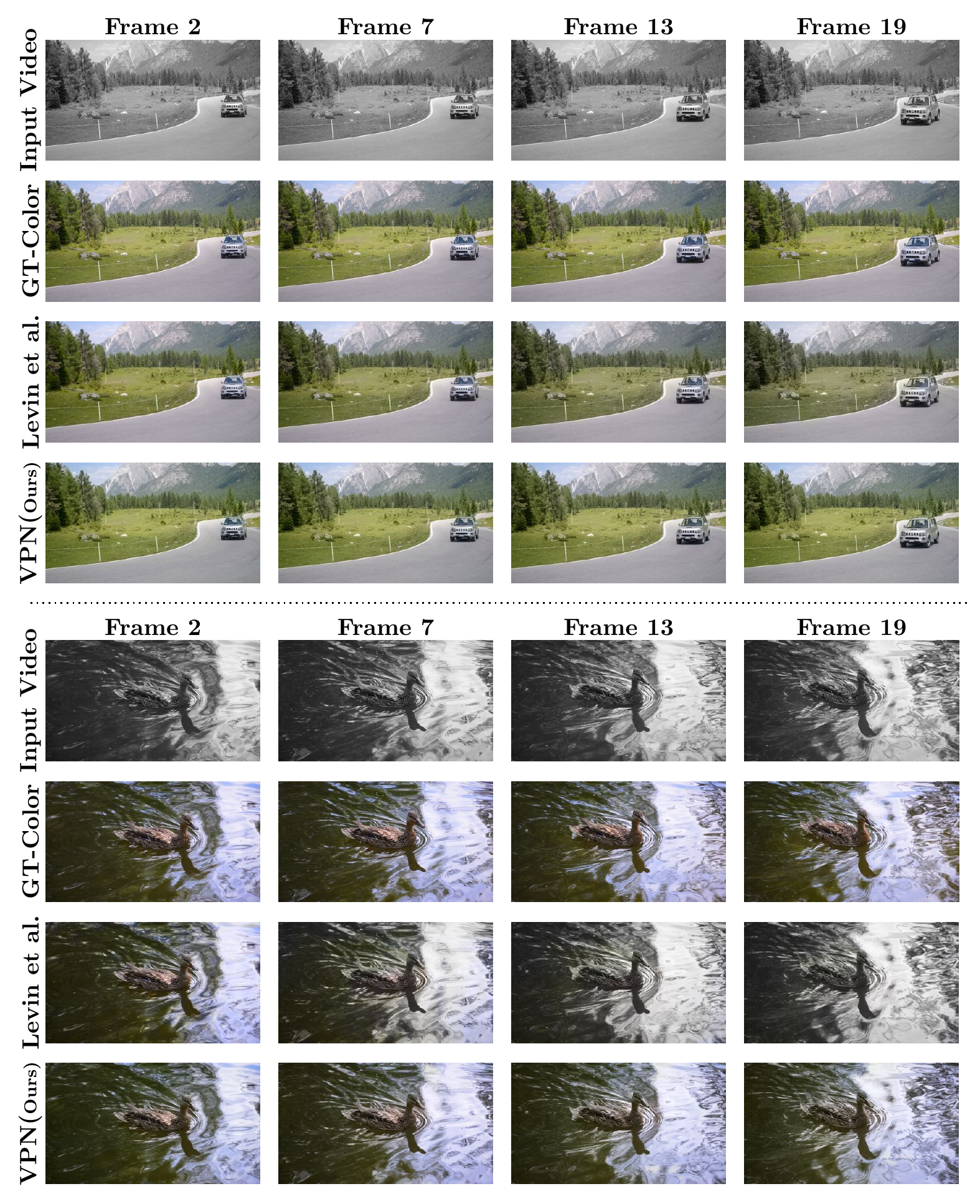}}
  \mycaption{Video Color Propagation}
  {Input grayscale video frames and corresponding ground-truth (GT) color images
  together with color predictions of Levin et al.~\cite{levin2004colorization} and VPN-Stage1 models.}
  \label{fig:color_visuals_supp}
\end{center}
\vspace{-0.7cm}
\end{figure}

In this Section, we present experiment protocols and additional qualitative results for experiments
on video object segmentation, semantic video segmentation and video color
propagation. Table~\ref{tbl:parameters_supp} shows the feature scales and other parameters used in different experiments.
Figures~\ref{fig:video_seg_pos_supp} show some qualitative results on video object segmentation
with some failure cases in Fig.~\ref{fig:video_seg_neg_supp}.
Figure~\ref{fig:semantic_visuals_supp} shows some qualitative results on semantic video segmentation and
Fig.~\ref{fig:color_visuals_supp} shows results on video color propagation.

\newcolumntype{L}[1]{>{\raggedright\let\newline\\\arraybackslash\hspace{0pt}}b{#1}}
\newcolumntype{C}[1]{>{\centering\let\newline\\\arraybackslash\hspace{0pt}}b{#1}}
\newcolumntype{R}[1]{>{\raggedleft\let\newline\\\arraybackslash\hspace{0pt}}b{#1}}

\begin{table*}[h]
\tiny
  \centering
    \begin{tabular}{L{3.0cm} L{2.4cm} L{2.8cm} L{2.8cm} C{0.5cm} C{1.0cm} L{1.2cm}}
      \toprule
\textbf{Experiment} & \textbf{Feature Type} & \textbf{Feature Scale-1, $\Lambda_a$} & \textbf{Feature Scale-2, $\Lambda_b$} & \textbf{$\alpha$} & \textbf{Input Frames} & \textbf{Loss Type} \\
      \midrule
      \textbf{Video Object Segmentation} & ($x,y,Y,Cb,Cr,t$) & (0.02,0.02,0.07,0.4,0.4,0.01) & (0.03,0.03,0.09,0.5,0.5,0.2) & 0.5 & 9 & Logistic\\
      \midrule
      \textbf{Semantic Video Segmentation} & & & & & \\
      \textbf{with CNN1~\cite{yu2015multi}-NoFlow} & ($x,y,R,G,B,t$) & (0.08,0.08,0.2,0.2,0.2,0.04) & (0.11,0.11,0.2,0.2,0.2,0.04) & 0.5 & 3 & Logistic \\
      \textbf{with CNN1~\cite{yu2015multi}-Flow} & ($x+u_x,y+u_y,R,G,B,t$) & (0.11,0.11,0.14,0.14,0.14,0.03) & (0.08,0.08,0.12,0.12,0.12,0.01) & 0.65 & 3 & Logistic\\
      \textbf{with CNN2~\cite{richter2016playing}-Flow} & ($x+u_x,y+u_y,R,G,B,t$) & (0.08,0.08,0.2,0.2,0.2,0.04) & (0.09,0.09,0.25,0.25,0.25,0.03) & 0.5 & 4 & Logistic\\
      \midrule
      \textbf{Video Color Propagation} & ($x,y,I,t$)  & (0.04,0.04,0.2,0.04) & No second kernel & 1 & 4 & MSE\\
      \bottomrule
      \\
    \end{tabular}
    \mycaption{Experiment Protocols} {Experiment protocols for the different experiments presented in this work. \textbf{Feature Types}:
    Feature spaces used for the bilateral convolutions, with position ($x,y$) and color
    ($R,G,B$ or $Y,Cb,Cr$) features $\in [0,255]$. $u_x$, $u_y$ denotes optical flow with respect
    to the present frame and $I$ denotes grayscale intensity.
    \textbf{Feature Scales ($\Lambda_a, \Lambda_b$)}: Cross-validated scales for the features used.
    \textbf{$\alpha$}: Exponential time decay for the input frames.
    \textbf{Input Frames}: Number of input frames for VPN.
    \textbf{Loss Type}: Type
     of loss used for back-propagation. ``MSE'' corresponds to Euclidean mean squared error loss and ``Logistic'' corresponds to multinomial logistic loss.}
  \label{tbl:parameters_supp}
\end{table*}


\clearpage

\section{Additional Material for Bilateral Inception Networks}
\label{sec:binception-app}

In this section of the Appendix, we first discuss the use of approximate bilateral
filtering in BI modules (Sec.~\ref{sec:lattice}).
Later, we present some qualitative results using different models for the approach presented in
Chapter~\ref{chap:binception} (Sec.~\ref{sec:qualitative-app}).

\subsection{Approximate Bilateral Filtering}
\label{sec:lattice}

The bilateral inception module presented in Chapter~\ref{chap:binception} computes a matrix-vector
product between a Gaussian filter $K$ and a vector of activations $\bz_c$.
Bilateral filtering is an important operation and many algorithmic techniques have been
proposed to speed-up this operation~\cite{paris2006fast,adams2010fast,gastal2011domain}.
In the main paper we opted to implement what can be considered the
brute-force variant of explicitly constructing $K$ and then using BLAS to compute the
matrix-vector product. This resulted in a few millisecond operation.
The explicit way to compute is possible due to the
reduction to super-pixels, e.g., it would not work for DenseCRF variants
that operate on the full image resolution.

Here, we present experiments where we use the fast approximate bilateral filtering
algorithm of~\cite{adams2010fast}, which is also used in Chapter~\ref{chap:bnn}
for learning sparse high dimensional filters. This
choice allows for larger dimensions of matrix-vector multiplication. The reason for choosing
the explicit multiplication in Chapter~\ref{chap:binception} was that it was computationally faster.
For the small sizes of the involved matrices and vectors, the explicit computation is sufficient and we had no
GPU implementation of an approximate technique that matched this runtime. Also it
is conceptually easier and the gradient to the feature transformations ($\Lambda \mathbf{f}$) is
obtained using standard matrix calculus.

\subsubsection{Experiments}

We modified the existing segmentation architectures analogous to those in Chapter~\ref{chap:binception}.
The main difference is that, here, the inception modules use the lattice
approximation~\cite{adams2010fast} to compute the bilateral filtering.
Using the lattice approximation did not allow us to back-propagate through feature transformations ($\Lambda$)
and thus we used hand-specified feature scales as will be explained later.
Specifically, we take CNN architectures from the works
of~\cite{chen2014semantic,zheng2015conditional,bell2015minc} and insert the BI modules between
the spatial FC layers.
We use superpixels from~\cite{DollarICCV13edges}
for all the experiments with the lattice approximation. Experiments are
performed using Caffe neural network framework~\cite{jia2014caffe}.

\begin{table}
  \small
  \centering
  \begin{tabular}{p{5.5cm}>{\raggedright\arraybackslash}p{1.4cm}>{\centering\arraybackslash}p{2.2cm}}
    \toprule
		\textbf{Model} & \emph{IoU} & \emph{Runtime}(ms) \\
    \midrule

		\deeplablargefov & 68.9 & 145ms\\
    \midrule
    \bi{7}{2}-\bi{8}{10}& \textbf{73.8} & +600 \\
    \midrule
    \deeplablargefovcrf~\cite{chen2014semantic} & 72.7 & +830\\
    \deeplabmsclargefovcrf~\cite{chen2014semantic} & \textbf{73.6} & +880\\
    DeepLab-EdgeNet~\cite{chen2015semantic} & 71.7 & +30\\
    DeepLab-EdgeNet-CRF~\cite{chen2015semantic} & \textbf{73.6} & +860\\
  \bottomrule \\
  \end{tabular}
  \mycaption{Semantic Segmentation using the DeepLab model}
  {IoU scores on the Pascal VOC12 segmentation test dataset
  with different models and our modified inception model.
  Also shown are the corresponding runtimes in milliseconds. Runtimes
  also include superpixel computations (300 ms with Dollar superpixels~\cite{DollarICCV13edges})}
  \label{tab:largefovresults}
\end{table}

\paragraph{Semantic Segmentation}
The experiments in this section use the Pascal VOC12 segmentation dataset~\cite{voc2012segmentation} with 21 object classes and the images have a maximum resolution of 0.25 megapixels.
For all experiments on VOC12, we train using the extended training set of
10581 images collected by~\cite{hariharan2011moredata}.
We modified the \deeplab~network architecture of~\cite{chen2014semantic} and
the CRFasRNN architecture from~\cite{zheng2015conditional} which uses a CNN with
deconvolution layers followed by DenseCRF trained end-to-end.

\paragraph{DeepLab Model}\label{sec:deeplabmodel}
We experimented with the \bi{7}{2}-\bi{8}{10} inception model.
Results using the~\deeplab~model are summarized in Tab.~\ref{tab:largefovresults}.
Although we get similar improvements with inception modules as with the
explicit kernel computation, using lattice approximation is slower.

\begin{table}
  \small
  \centering
  \begin{tabular}{p{6.4cm}>{\raggedright\arraybackslash}p{1.8cm}>{\raggedright\arraybackslash}p{1.8cm}}
    \toprule
    \textbf{Model} & \emph{IoU (Val)} & \emph{IoU (Test)}\\
    \midrule
    CNN &  67.5 & - \\
    \deconv (CNN+Deconvolutions) & 69.8 & 72.0 \\
    \midrule
    \bi{3}{6}-\bi{4}{6}-\bi{7}{2}-\bi{8}{6}& 71.9 & - \\
    \bi{3}{6}-\bi{4}{6}-\bi{7}{2}-\bi{8}{6}-\gi{6}& 73.6 &  \href{http://host.robots.ox.ac.uk:8080/anonymous/VOTV5E.html}{\textbf{75.2}}\\
    \midrule
    \deconvcrf (CRF-RNN)~\cite{zheng2015conditional} & 73.0 & 74.7\\
    Context-CRF-RNN~\cite{yu2015multi} & ~~ - ~ & \textbf{75.3} \\
    \bottomrule \\
  \end{tabular}
  \mycaption{Semantic Segmentation using the CRFasRNN model}{IoU score corresponding to different models
  on Pascal VOC12 reduced validation / test segmentation dataset. The reduced validation set consists of 346 images
  as used in~\cite{zheng2015conditional} where we adapted the model from.}
  \label{tab:deconvresults-app}
\end{table}

\paragraph{CRFasRNN Model}\label{sec:deepinception}
We add BI modules after score-pool3, score-pool4, \fc{7} and \fc{8} $1\times1$ convolution layers
resulting in the \bi{3}{6}-\bi{4}{6}-\bi{7}{2}-\bi{8}{6}
model and also experimented with another variant where $BI_8$ is followed by another inception
module, G$(6)$, with 6 Gaussian kernels.
Note that here also we discarded both deconvolution and DenseCRF parts of the original model~\cite{zheng2015conditional}
and inserted the BI modules in the base CNN and found similar improvements compared to the inception modules with explicit
kernel computaion. See Tab.~\ref{tab:deconvresults-app} for results on the CRFasRNN model.

\paragraph{Material Segmentation}
Table~\ref{tab:mincresults-app} shows the results on the MINC dataset~\cite{bell2015minc}
obtained by modifying the AlexNet architecture with our inception modules. We observe
similar improvements as with explicit kernel construction.
For this model, we do not provide any learned setup due to very limited segment training
data. The weights to combine outputs in the bilateral inception layer are
found by validation on the validation set.

\begin{table}[t]
  \small
  \centering
  \begin{tabular}{p{3.5cm}>{\centering\arraybackslash}p{4.0cm}}
    \toprule
    \textbf{Model} & Class / Total accuracy\\
    \midrule

    AlexNet CNN & 55.3 / 58.9 \\
    \midrule
    \bi{7}{2}-\bi{8}{6}& 68.5 / 71.8 \\
    \bi{7}{2}-\bi{8}{6}-G$(6)$& 67.6 / 73.1 \\
    \midrule
    AlexNet-CRF & 65.5 / 71.0 \\
    \bottomrule \\
  \end{tabular}
  \mycaption{Material Segmentation using AlexNet}{Pixel accuracy of different models on
  the MINC material segmentation test dataset~\cite{bell2015minc}.}
  \label{tab:mincresults-app}
\end{table}

\paragraph{Scales of Bilateral Inception Modules}
\label{sec:scales}

Unlike the explicit kernel technique presented in the main text (Chapter~\ref{chap:binception}),
we didn't back-propagate through feature transformation ($\Lambda$)
using the approximate bilateral filter technique.
So, the feature scales are hand-specified and validated, which are as follows.
The optimal scale values for the \bi{7}{2}-\bi{8}{2} model are found by validation for the best performance which are
$\sigma_{xy}$ = (0.1, 0.1) for the spatial (XY) kernel and $\sigma_{rgbxy}$ = (0.1, 0.1, 0.1, 0.01, 0.01) for color and position (RGBXY)  kernel.
Next, as more kernels are added to \bi{8}{2}, we set scales to be $\alpha$*($\sigma_{xy}$, $\sigma_{rgbxy}$).
The value of $\alpha$ is chosen as  1, 0.5, 0.1, 0.05, 0.1, at uniform interval, for the \bi{8}{10} bilateral inception module.

\subsection{Qualitative Results}
\label{sec:qualitative-app}

In this section, we present more qualitative results obtained using the BI module with explicit
kernel computation technique presented in Chapter~\ref{chap:binception}. Results on the Pascal VOC12
dataset~\cite{voc2012segmentation} using the DeepLab-LargeFOV model are shown in Fig.~\ref{fig:semantic_visuals-app},
followed by the results on MINC dataset~\cite{bell2015minc}
in Fig.~\ref{fig:material_visuals-app} and on
Cityscapes dataset~\cite{Cordts2015Cvprw} in Fig.~\ref{fig:street_visuals-app}.

\definecolor{voc_1}{RGB}{0, 0, 0}
\definecolor{voc_2}{RGB}{128, 0, 0}
\definecolor{voc_3}{RGB}{0, 128, 0}
\definecolor{voc_4}{RGB}{128, 128, 0}
\definecolor{voc_5}{RGB}{0, 0, 128}
\definecolor{voc_6}{RGB}{128, 0, 128}
\definecolor{voc_7}{RGB}{0, 128, 128}
\definecolor{voc_8}{RGB}{128, 128, 128}
\definecolor{voc_9}{RGB}{64, 0, 0}
\definecolor{voc_10}{RGB}{192, 0, 0}
\definecolor{voc_11}{RGB}{64, 128, 0}
\definecolor{voc_12}{RGB}{192, 128, 0}
\definecolor{voc_13}{RGB}{64, 0, 128}
\definecolor{voc_14}{RGB}{192, 0, 128}
\definecolor{voc_15}{RGB}{64, 128, 128}
\definecolor{voc_16}{RGB}{192, 128, 128}
\definecolor{voc_17}{RGB}{0, 64, 0}
\definecolor{voc_18}{RGB}{128, 64, 0}
\definecolor{voc_19}{RGB}{0, 192, 0}
\definecolor{voc_20}{RGB}{128, 192, 0}
\definecolor{voc_21}{RGB}{0, 64, 128}
\definecolor{voc_22}{RGB}{128, 64, 128}

\begin{figure*}[!ht]
  \small
  \centering
  \fcolorbox{white}{voc_1}{\rule{0pt}{4pt}\rule{4pt}{0pt}} Background~~
  \fcolorbox{white}{voc_2}{\rule{0pt}{4pt}\rule{4pt}{0pt}} Aeroplane~~
  \fcolorbox{white}{voc_3}{\rule{0pt}{4pt}\rule{4pt}{0pt}} Bicycle~~
  \fcolorbox{white}{voc_4}{\rule{0pt}{4pt}\rule{4pt}{0pt}} Bird~~
  \fcolorbox{white}{voc_5}{\rule{0pt}{4pt}\rule{4pt}{0pt}} Boat~~
  \fcolorbox{white}{voc_6}{\rule{0pt}{4pt}\rule{4pt}{0pt}} Bottle~~
  \fcolorbox{white}{voc_7}{\rule{0pt}{4pt}\rule{4pt}{0pt}} Bus~~
  \fcolorbox{white}{voc_8}{\rule{0pt}{4pt}\rule{4pt}{0pt}} Car~~\\
  \fcolorbox{white}{voc_9}{\rule{0pt}{4pt}\rule{4pt}{0pt}} Cat~~
  \fcolorbox{white}{voc_10}{\rule{0pt}{4pt}\rule{4pt}{0pt}} Chair~~
  \fcolorbox{white}{voc_11}{\rule{0pt}{4pt}\rule{4pt}{0pt}} Cow~~
  \fcolorbox{white}{voc_12}{\rule{0pt}{4pt}\rule{4pt}{0pt}} Dining Table~~
  \fcolorbox{white}{voc_13}{\rule{0pt}{4pt}\rule{4pt}{0pt}} Dog~~
  \fcolorbox{white}{voc_14}{\rule{0pt}{4pt}\rule{4pt}{0pt}} Horse~~
  \fcolorbox{white}{voc_15}{\rule{0pt}{4pt}\rule{4pt}{0pt}} Motorbike~~
  \fcolorbox{white}{voc_16}{\rule{0pt}{4pt}\rule{4pt}{0pt}} Person~~\\
  \fcolorbox{white}{voc_17}{\rule{0pt}{4pt}\rule{4pt}{0pt}} Potted Plant~~
  \fcolorbox{white}{voc_18}{\rule{0pt}{4pt}\rule{4pt}{0pt}} Sheep~~
  \fcolorbox{white}{voc_19}{\rule{0pt}{4pt}\rule{4pt}{0pt}} Sofa~~
  \fcolorbox{white}{voc_20}{\rule{0pt}{4pt}\rule{4pt}{0pt}} Train~~
  \fcolorbox{white}{voc_21}{\rule{0pt}{4pt}\rule{4pt}{0pt}} TV monitor~~\\

  \subfigure{%
    \includegraphics[width=.15\columnwidth]{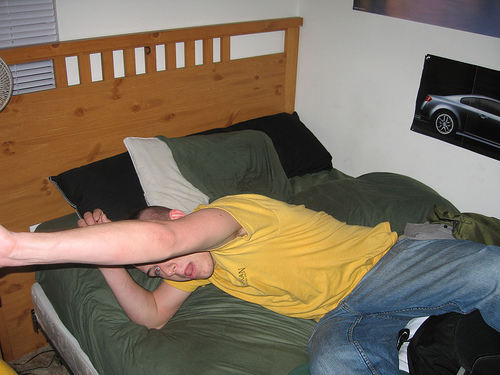}
  }
  \subfigure{%
    \includegraphics[width=.15\columnwidth]{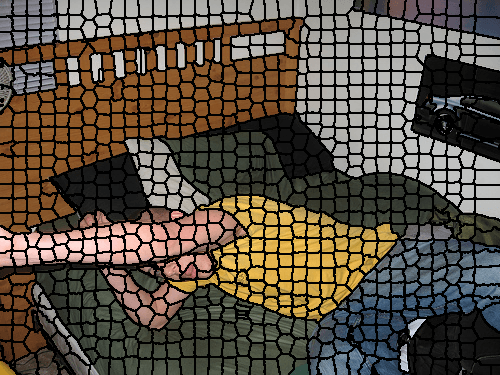}
  }
  \subfigure{%
    \includegraphics[width=.15\columnwidth]{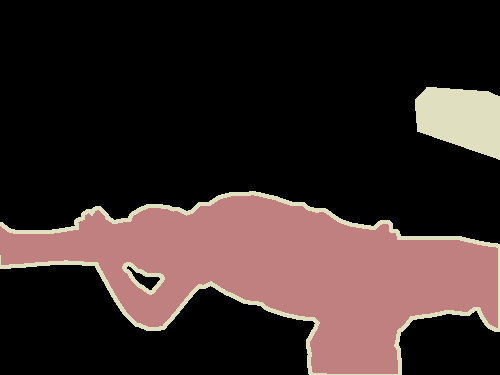}
  }
  \subfigure{%
    \includegraphics[width=.15\columnwidth]{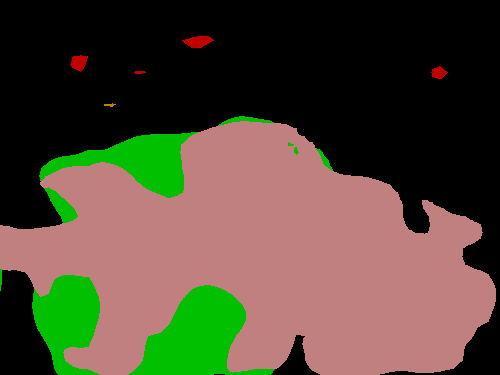}
  }
  \subfigure{%
    \includegraphics[width=.15\columnwidth]{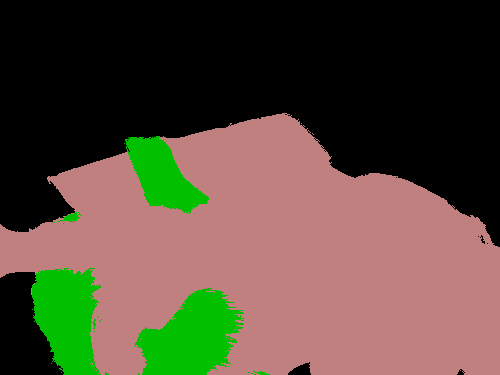}
  }
  \subfigure{%
    \includegraphics[width=.15\columnwidth]{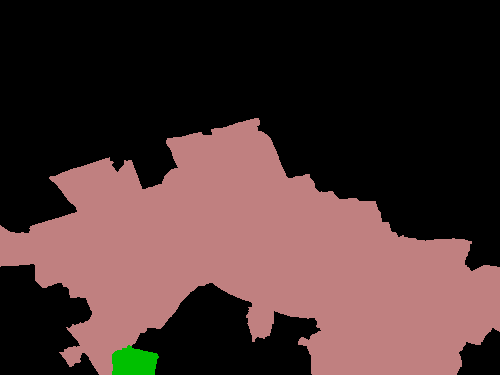}
  }\\[-2ex]

  \subfigure{%
    \includegraphics[width=.15\columnwidth]{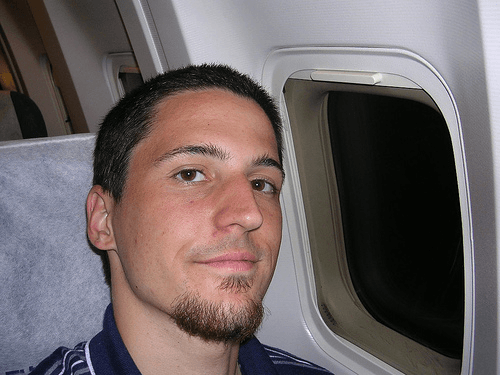}
  }
  \subfigure{%
    \includegraphics[width=.15\columnwidth]{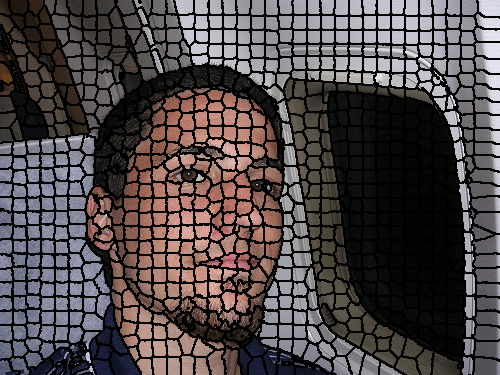}
  }
  \subfigure{%
    \includegraphics[width=.15\columnwidth]{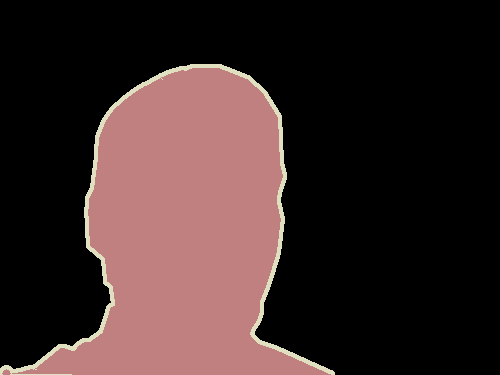}
  }
  \subfigure{%
    \includegraphics[width=.15\columnwidth]{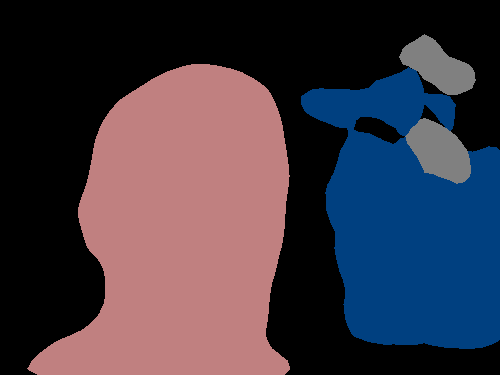}
  }
  \subfigure{%
    \includegraphics[width=.15\columnwidth]{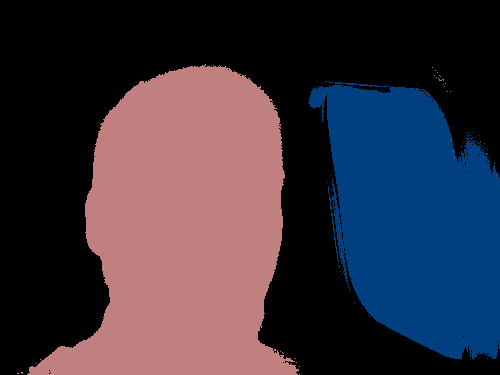}
  }
  \subfigure{%
    \includegraphics[width=.15\columnwidth]{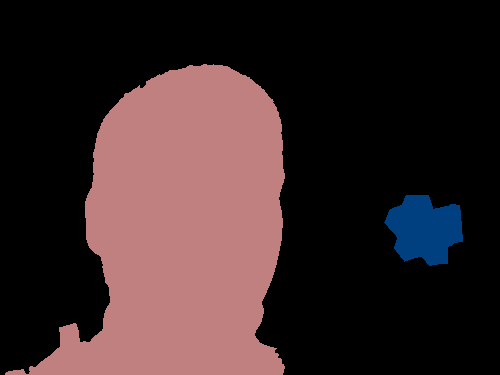}
  }\\[-2ex]

  \subfigure{%
    \includegraphics[width=.15\columnwidth]{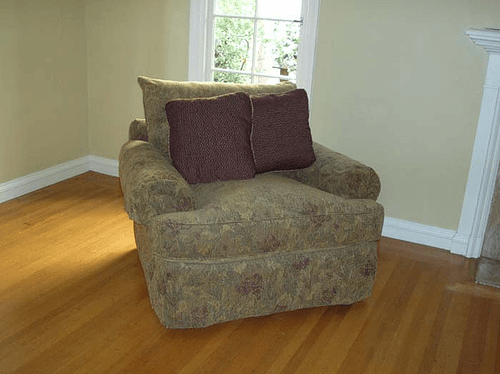}
  }
  \subfigure{%
    \includegraphics[width=.15\columnwidth]{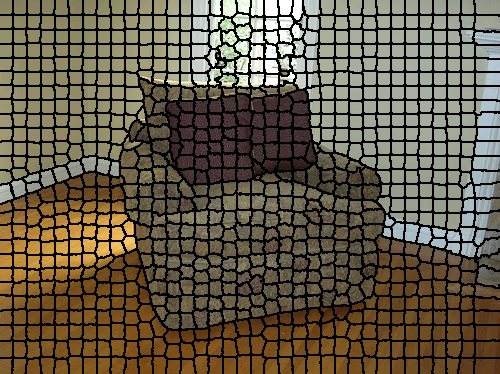}
  }
  \subfigure{%
    \includegraphics[width=.15\columnwidth]{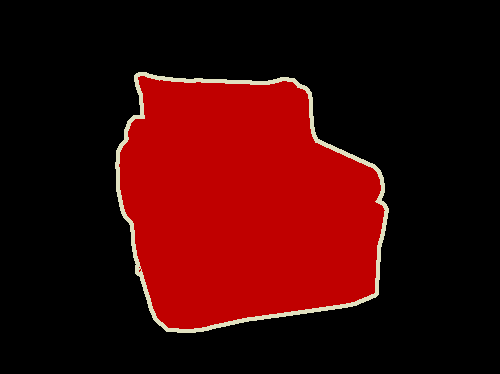}
  }
  \subfigure{%
    \includegraphics[width=.15\columnwidth]{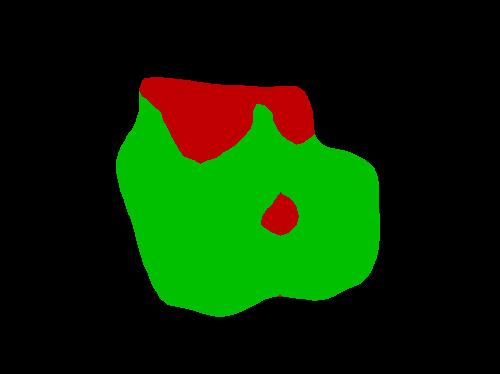}
  }
  \subfigure{%
    \includegraphics[width=.15\columnwidth]{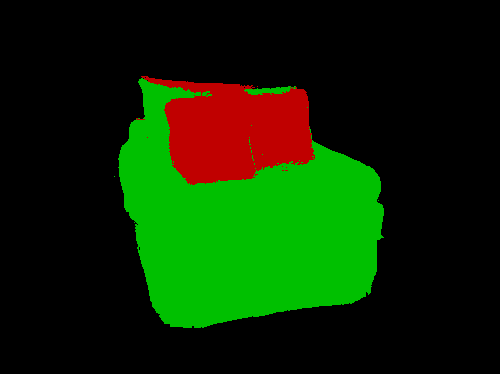}
  }
  \subfigure{%
    \includegraphics[width=.15\columnwidth]{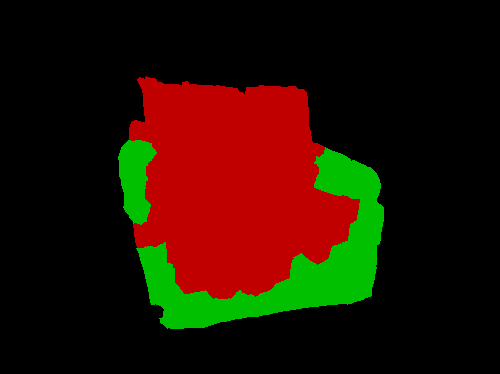}
  }\\[-2ex]

  \subfigure{%
    \includegraphics[width=.15\columnwidth]{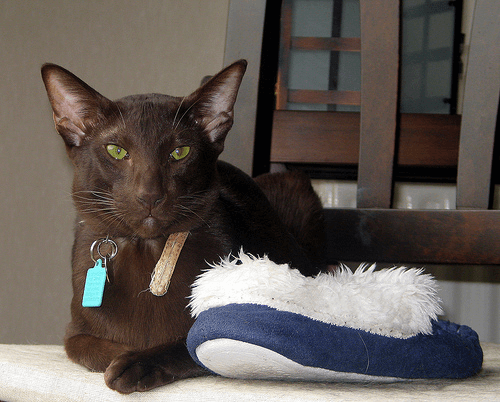}
  }
  \subfigure{%
    \includegraphics[width=.15\columnwidth]{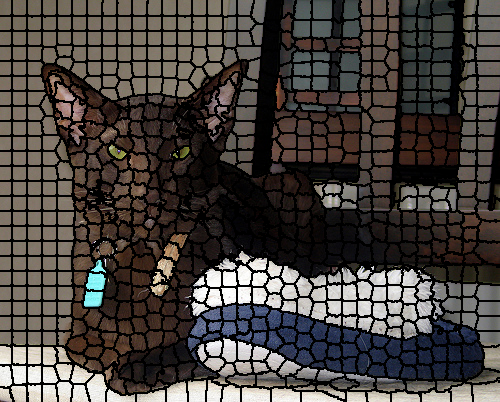}
  }
  \subfigure{%
    \includegraphics[width=.15\columnwidth]{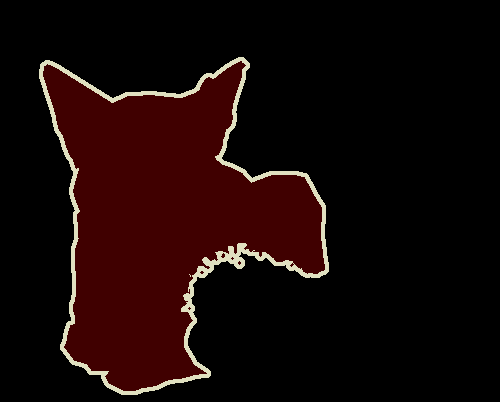}
  }
  \subfigure{%
    \includegraphics[width=.15\columnwidth]{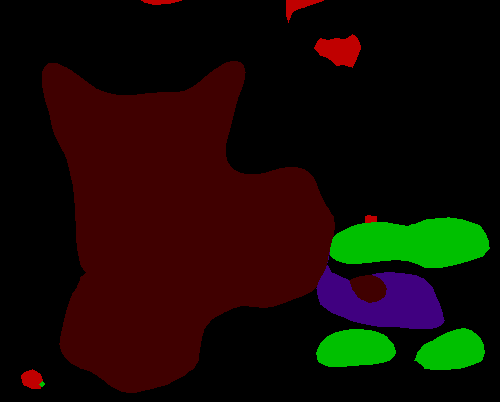}
  }
  \subfigure{%
    \includegraphics[width=.15\columnwidth]{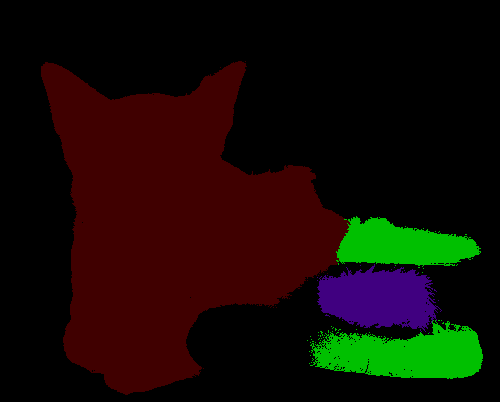}
  }
  \subfigure{%
    \includegraphics[width=.15\columnwidth]{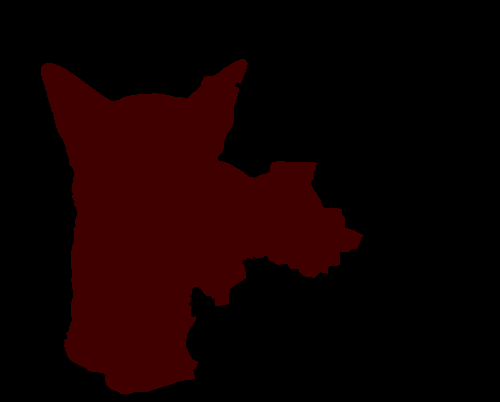}
  }\\[-2ex]

  \subfigure{%
    \includegraphics[width=.15\columnwidth]{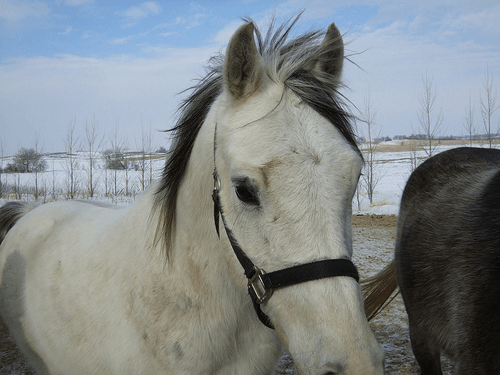}
  }
  \subfigure{%
    \includegraphics[width=.15\columnwidth]{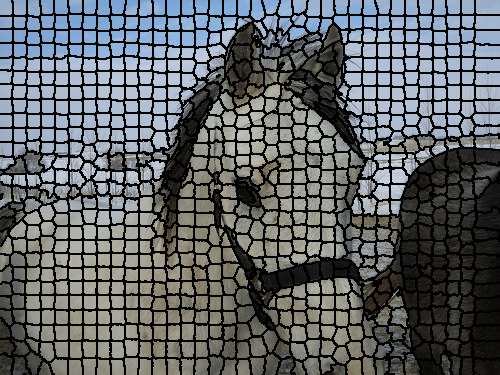}
  }
  \subfigure{%
    \includegraphics[width=.15\columnwidth]{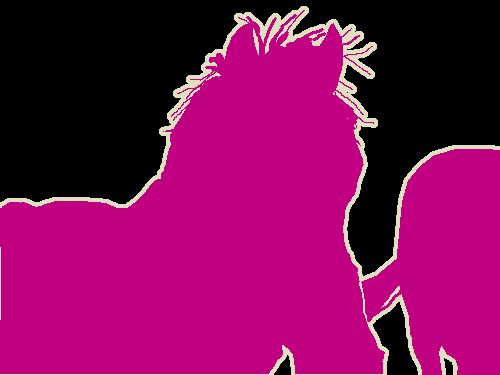}
  }
  \subfigure{%
    \includegraphics[width=.15\columnwidth]{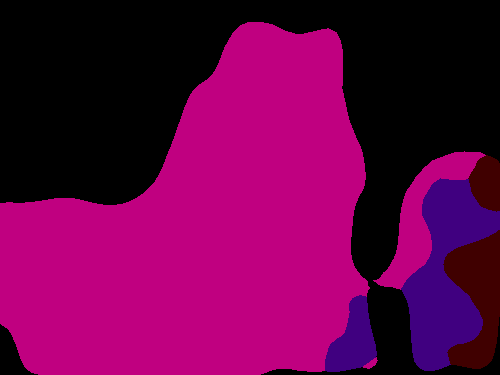}
  }
  \subfigure{%
    \includegraphics[width=.15\columnwidth]{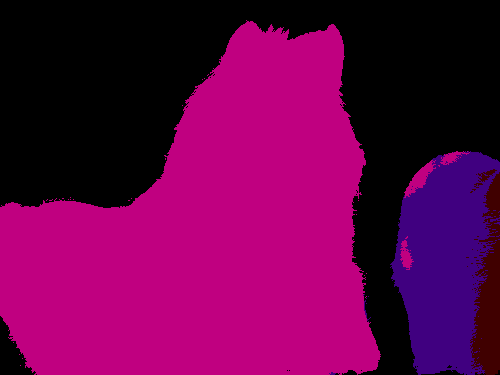}
  }
  \subfigure{%
    \includegraphics[width=.15\columnwidth]{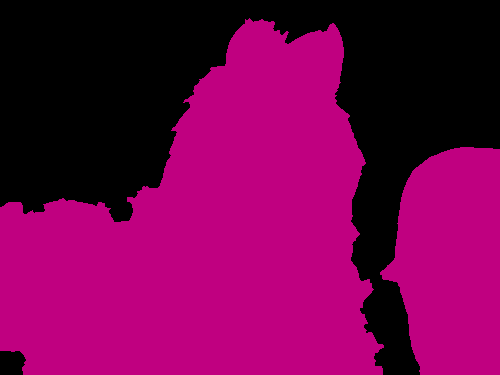}
  }\\[-2ex]

  \setcounter{subfigure}{0}
  \subfigure[\scriptsize Input]{%
    \includegraphics[width=.15\columnwidth]{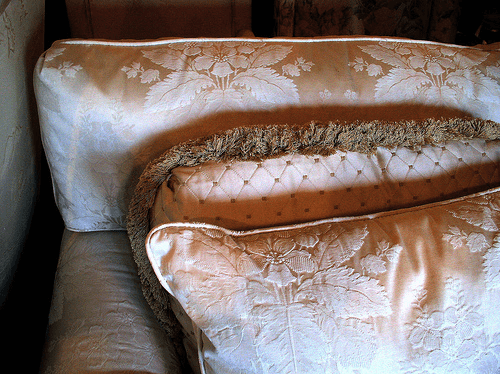}
  }
  \subfigure[\scriptsize Superpixels]{%
    \includegraphics[width=.15\columnwidth]{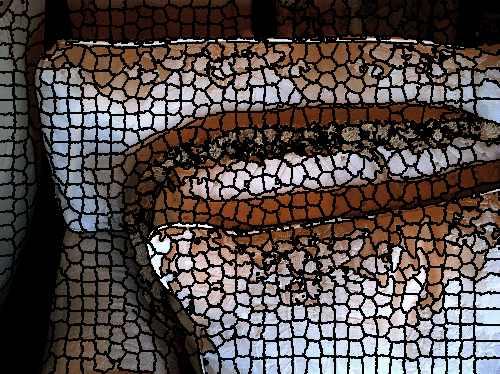}
  }
  \subfigure[\scriptsize GT]{%
    \includegraphics[width=.15\columnwidth]{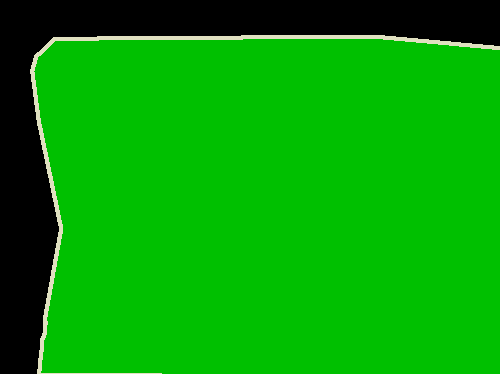}
  }
  \subfigure[\scriptsize Deeplab]{%
    \includegraphics[width=.15\columnwidth]{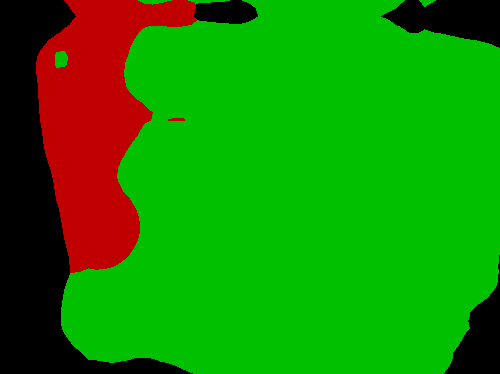}
  }
  \subfigure[\scriptsize +DenseCRF]{%
    \includegraphics[width=.15\columnwidth]{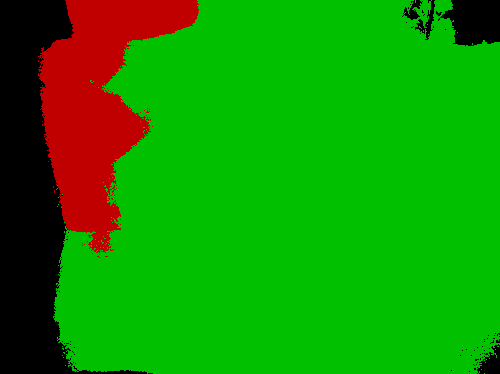}
  }
  \subfigure[\scriptsize Using BI]{%
    \includegraphics[width=.15\columnwidth]{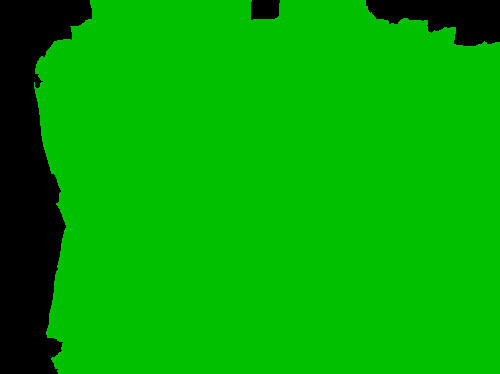}
  }
  \mycaption{Semantic Segmentation}{Example results of semantic segmentation
  on the Pascal VOC12 dataset.
  (d)~depicts the DeepLab CNN result, (e)~CNN + 10 steps of mean-field inference,
  (f~result obtained with bilateral inception (BI) modules (\bi{6}{2}+\bi{7}{6}) between \fc~layers.}
  \label{fig:semantic_visuals-app}
\end{figure*}

\definecolor{minc_1}{HTML}{771111}
\definecolor{minc_2}{HTML}{CAC690}
\definecolor{minc_3}{HTML}{EEEEEE}
\definecolor{minc_4}{HTML}{7C8FA6}
\definecolor{minc_5}{HTML}{597D31}
\definecolor{minc_6}{HTML}{104410}
\definecolor{minc_7}{HTML}{BB819C}
\definecolor{minc_8}{HTML}{D0CE48}
\definecolor{minc_9}{HTML}{622745}
\definecolor{minc_10}{HTML}{666666}
\definecolor{minc_11}{HTML}{D54A31}
\definecolor{minc_12}{HTML}{101044}
\definecolor{minc_13}{HTML}{444126}
\definecolor{minc_14}{HTML}{75D646}
\definecolor{minc_15}{HTML}{DD4348}
\definecolor{minc_16}{HTML}{5C8577}
\definecolor{minc_17}{HTML}{C78472}
\definecolor{minc_18}{HTML}{75D6D0}
\definecolor{minc_19}{HTML}{5B4586}
\definecolor{minc_20}{HTML}{C04393}
\definecolor{minc_21}{HTML}{D69948}
\definecolor{minc_22}{HTML}{7370D8}
\definecolor{minc_23}{HTML}{7A3622}
\definecolor{minc_24}{HTML}{000000}

\begin{figure*}[!ht]
  \small 
  \centering
  \fcolorbox{white}{minc_1}{\rule{0pt}{4pt}\rule{4pt}{0pt}} Brick~~
  \fcolorbox{white}{minc_2}{\rule{0pt}{4pt}\rule{4pt}{0pt}} Carpet~~
  \fcolorbox{white}{minc_3}{\rule{0pt}{4pt}\rule{4pt}{0pt}} Ceramic~~
  \fcolorbox{white}{minc_4}{\rule{0pt}{4pt}\rule{4pt}{0pt}} Fabric~~
  \fcolorbox{white}{minc_5}{\rule{0pt}{4pt}\rule{4pt}{0pt}} Foliage~~
  \fcolorbox{white}{minc_6}{\rule{0pt}{4pt}\rule{4pt}{0pt}} Food~~
  \fcolorbox{white}{minc_7}{\rule{0pt}{4pt}\rule{4pt}{0pt}} Glass~~
  \fcolorbox{white}{minc_8}{\rule{0pt}{4pt}\rule{4pt}{0pt}} Hair~~\\
  \fcolorbox{white}{minc_9}{\rule{0pt}{4pt}\rule{4pt}{0pt}} Leather~~
  \fcolorbox{white}{minc_10}{\rule{0pt}{4pt}\rule{4pt}{0pt}} Metal~~
  \fcolorbox{white}{minc_11}{\rule{0pt}{4pt}\rule{4pt}{0pt}} Mirror~~
  \fcolorbox{white}{minc_12}{\rule{0pt}{4pt}\rule{4pt}{0pt}} Other~~
  \fcolorbox{white}{minc_13}{\rule{0pt}{4pt}\rule{4pt}{0pt}} Painted~~
  \fcolorbox{white}{minc_14}{\rule{0pt}{4pt}\rule{4pt}{0pt}} Paper~~
  \fcolorbox{white}{minc_15}{\rule{0pt}{4pt}\rule{4pt}{0pt}} Plastic~~\\
  \fcolorbox{white}{minc_16}{\rule{0pt}{4pt}\rule{4pt}{0pt}} Polished Stone~~
  \fcolorbox{white}{minc_17}{\rule{0pt}{4pt}\rule{4pt}{0pt}} Skin~~
  \fcolorbox{white}{minc_18}{\rule{0pt}{4pt}\rule{4pt}{0pt}} Sky~~
  \fcolorbox{white}{minc_19}{\rule{0pt}{4pt}\rule{4pt}{0pt}} Stone~~
  \fcolorbox{white}{minc_20}{\rule{0pt}{4pt}\rule{4pt}{0pt}} Tile~~
  \fcolorbox{white}{minc_21}{\rule{0pt}{4pt}\rule{4pt}{0pt}} Wallpaper~~
  \fcolorbox{white}{minc_22}{\rule{0pt}{4pt}\rule{4pt}{0pt}} Water~~
  \fcolorbox{white}{minc_23}{\rule{0pt}{4pt}\rule{4pt}{0pt}} Wood~~\\
  \subfigure{%
    \includegraphics[width=.15\columnwidth]{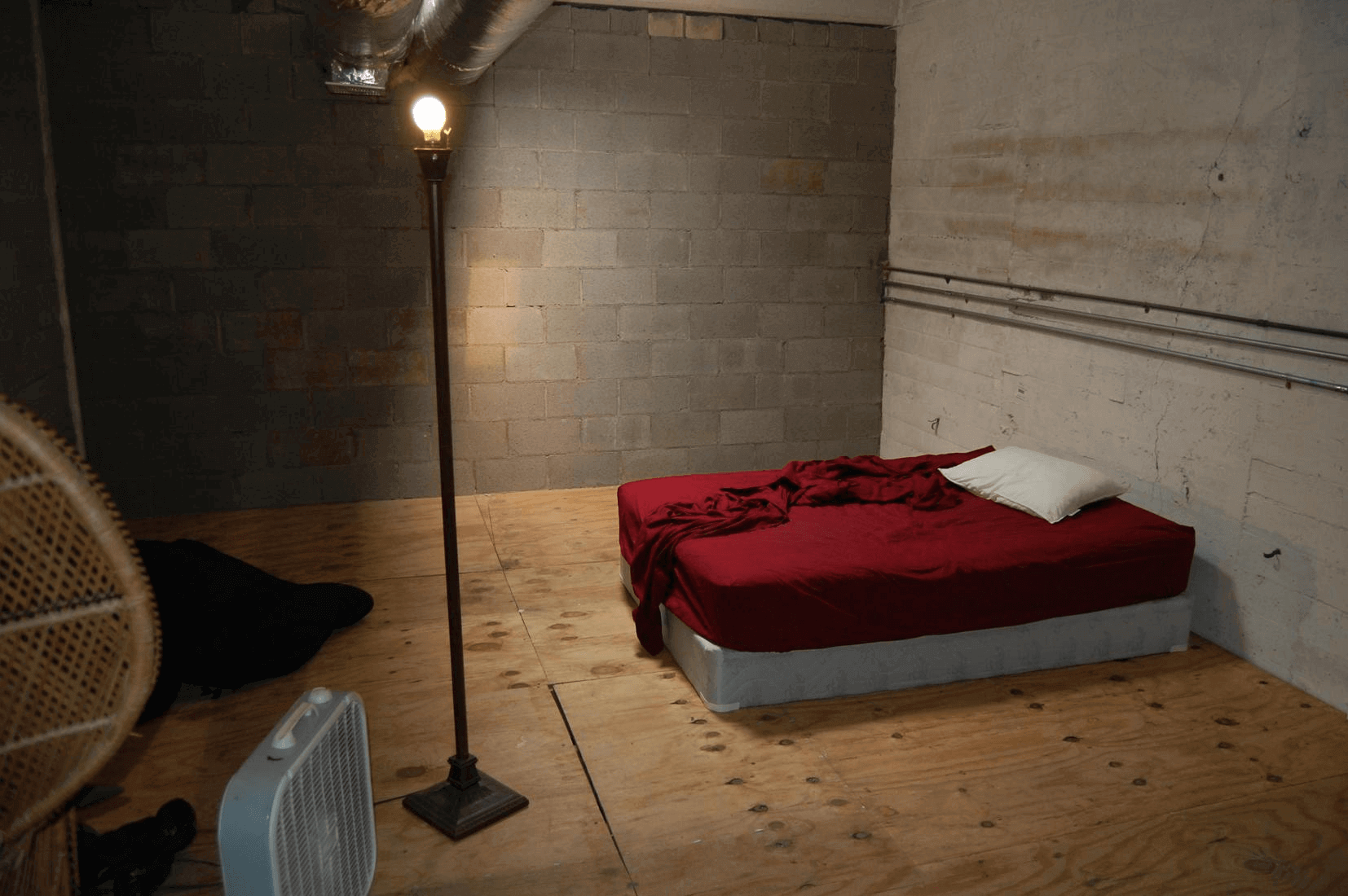}
  }
  \subfigure{%
    \includegraphics[width=.15\columnwidth]{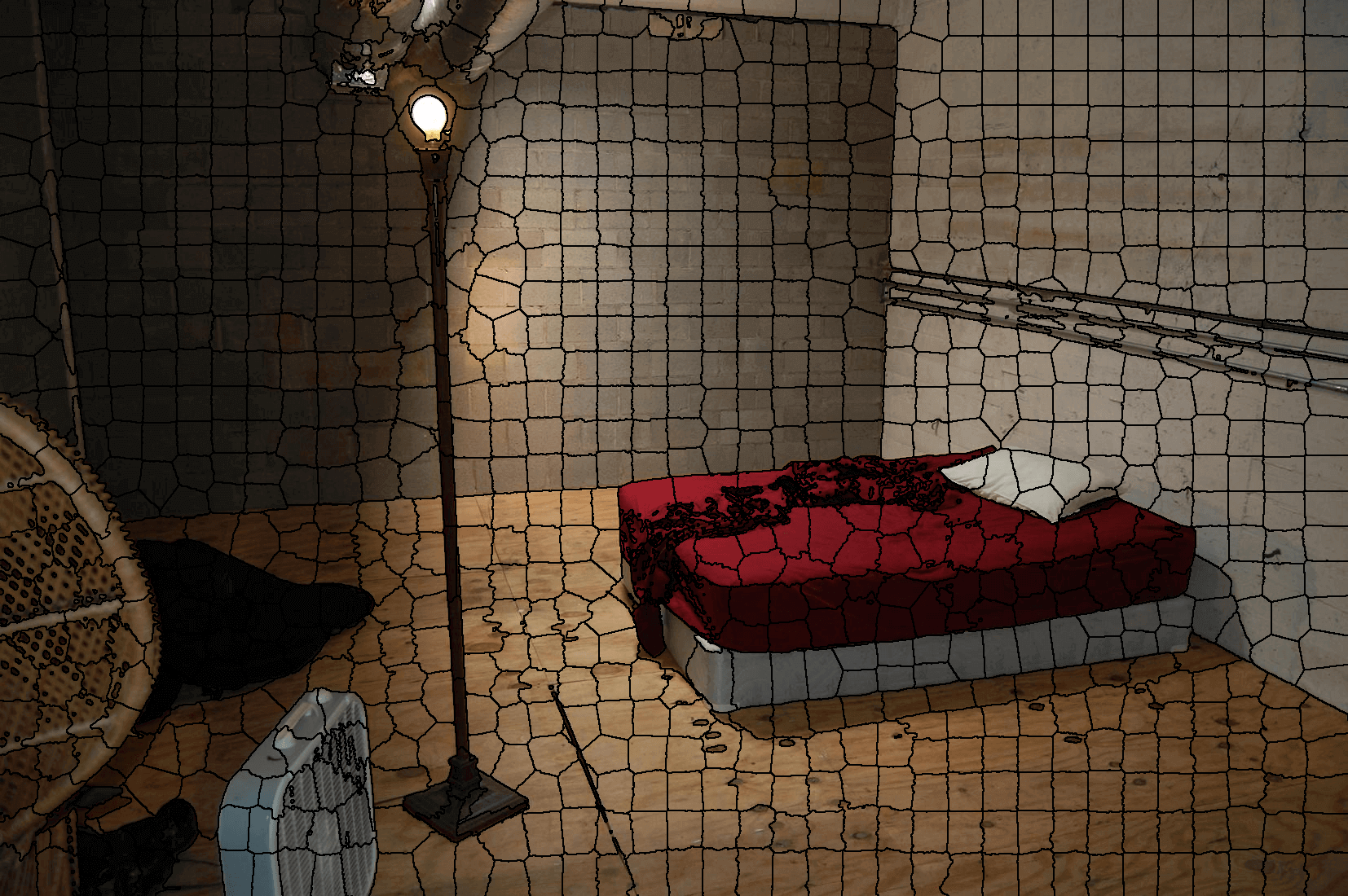}
  }
  \subfigure{%
    \includegraphics[width=.15\columnwidth]{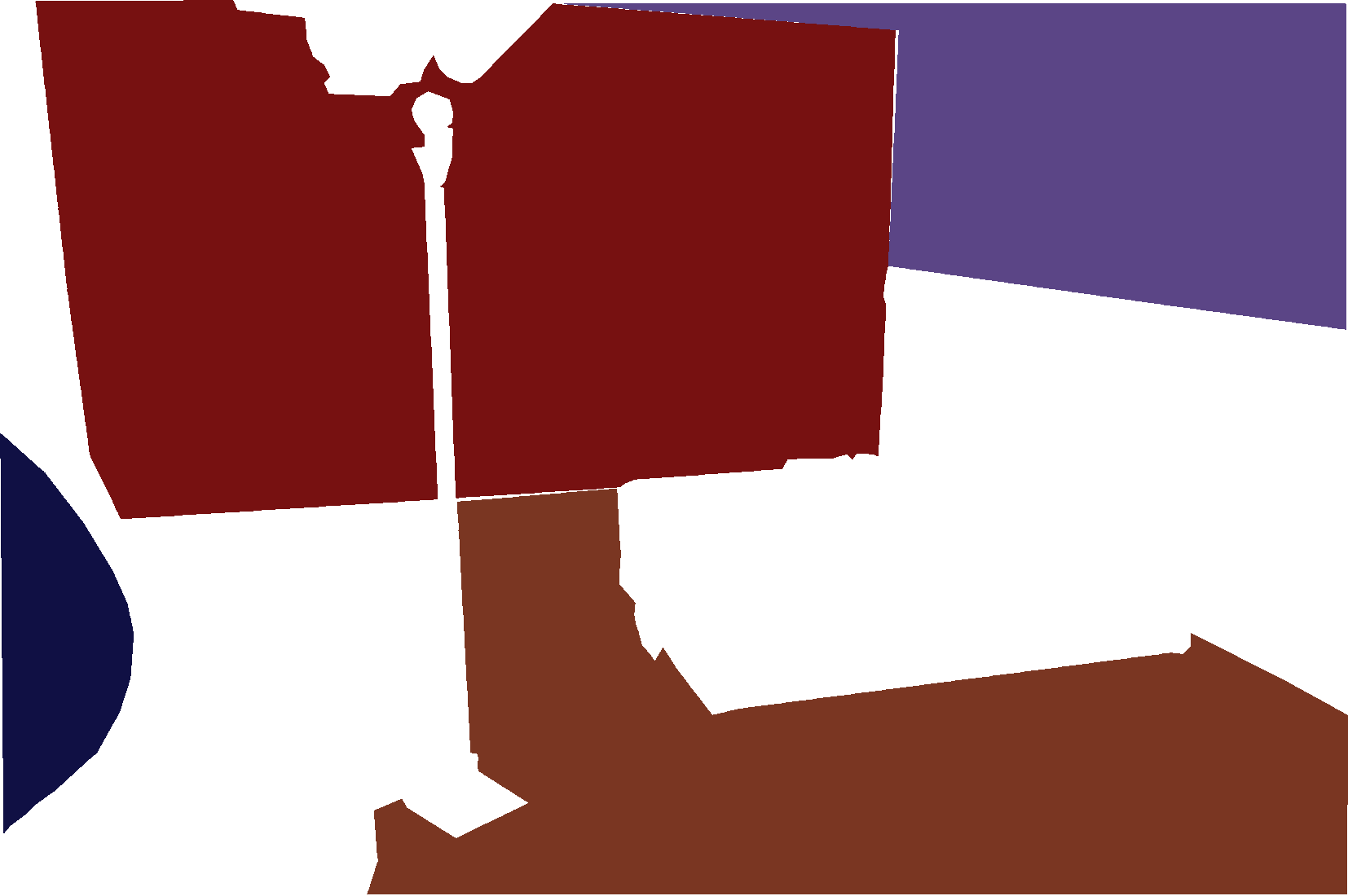}
  }
  \subfigure{%
    \includegraphics[width=.15\columnwidth]{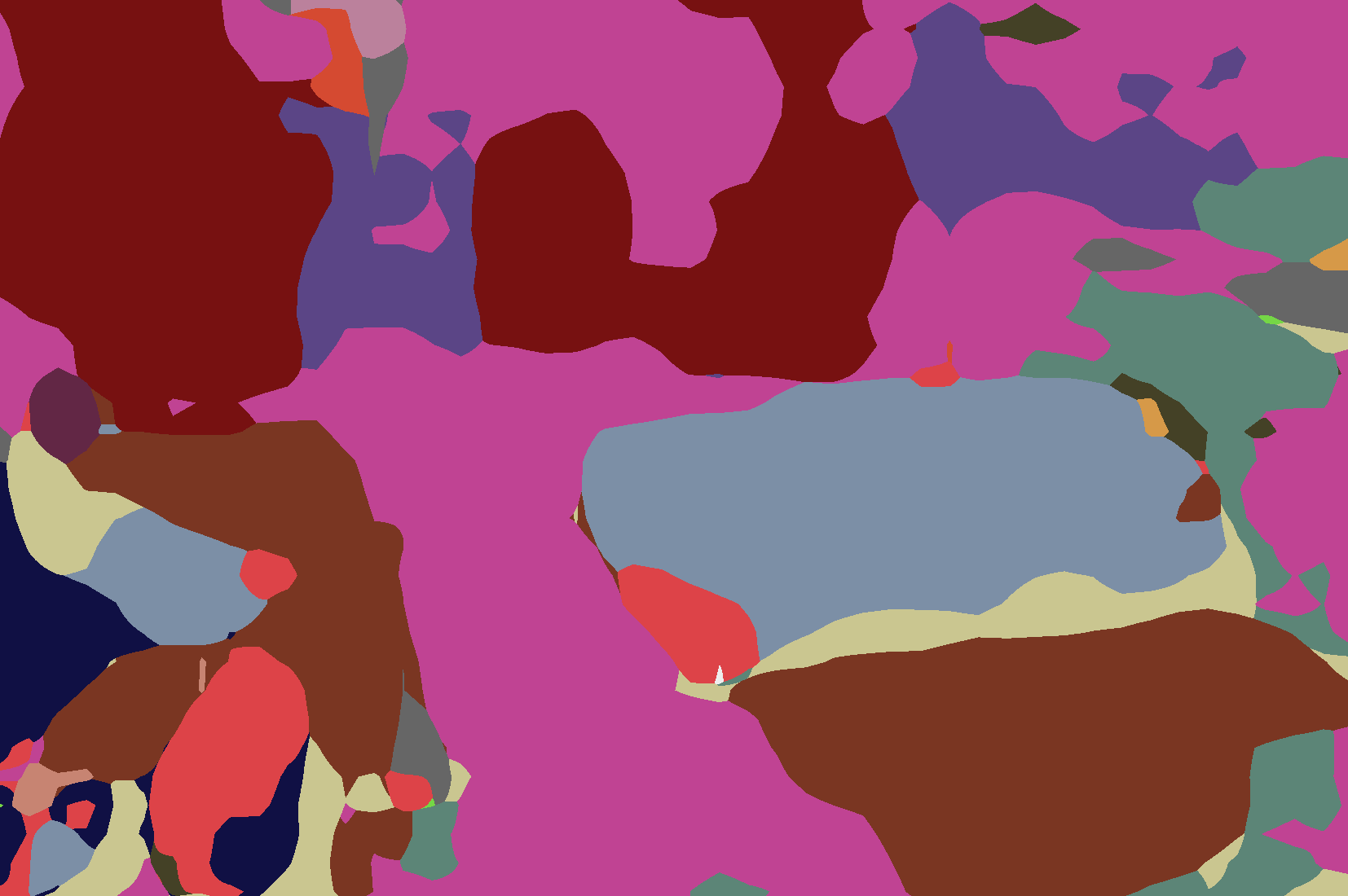}
  }
  \subfigure{%
    \includegraphics[width=.15\columnwidth]{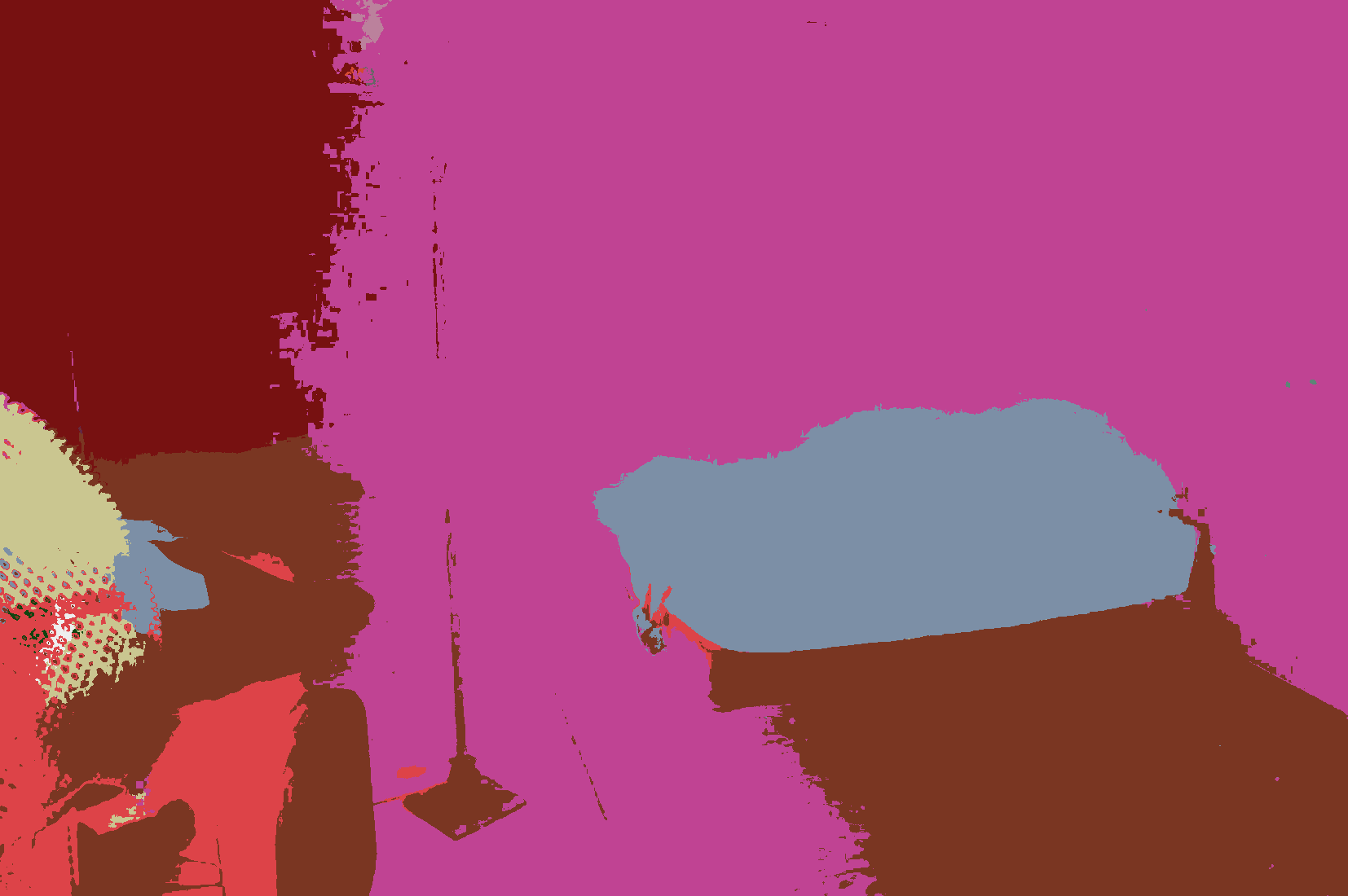}
  }
  \subfigure{%
    \includegraphics[width=.15\columnwidth]{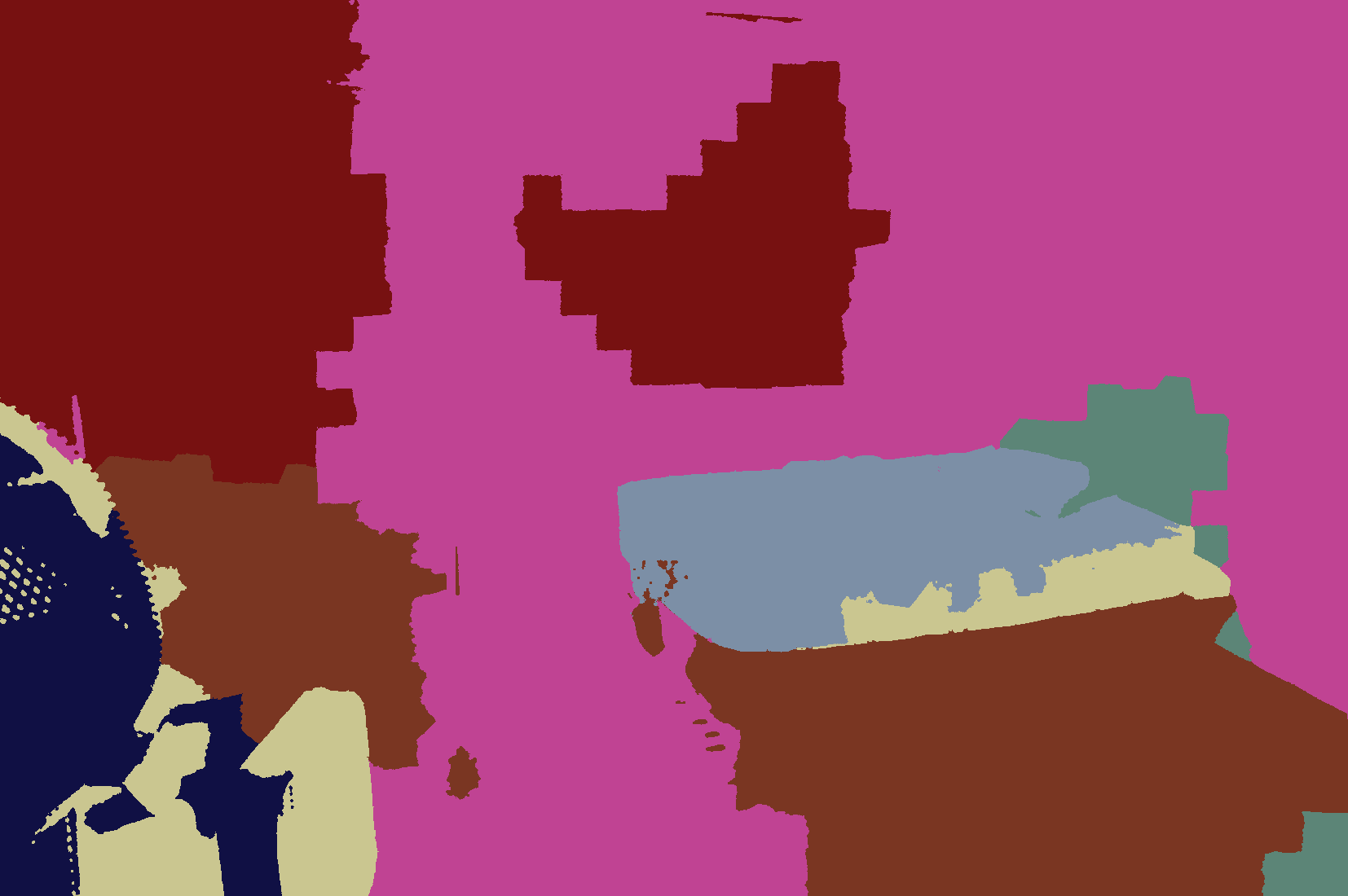}
  }\\[-2ex]

  \subfigure{%
    \includegraphics[width=.15\columnwidth]{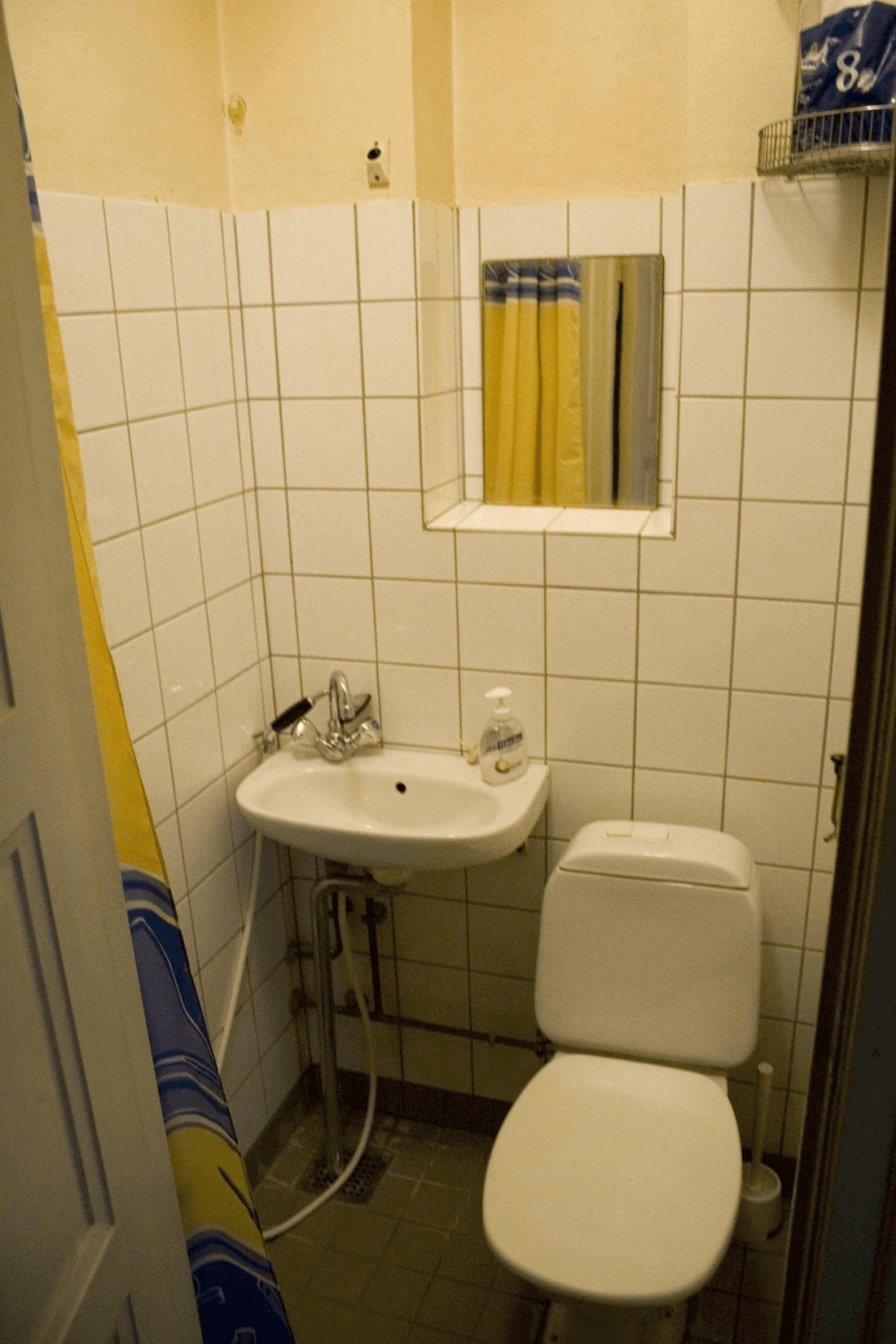}
  }
  \subfigure{%
    \includegraphics[width=.15\columnwidth]{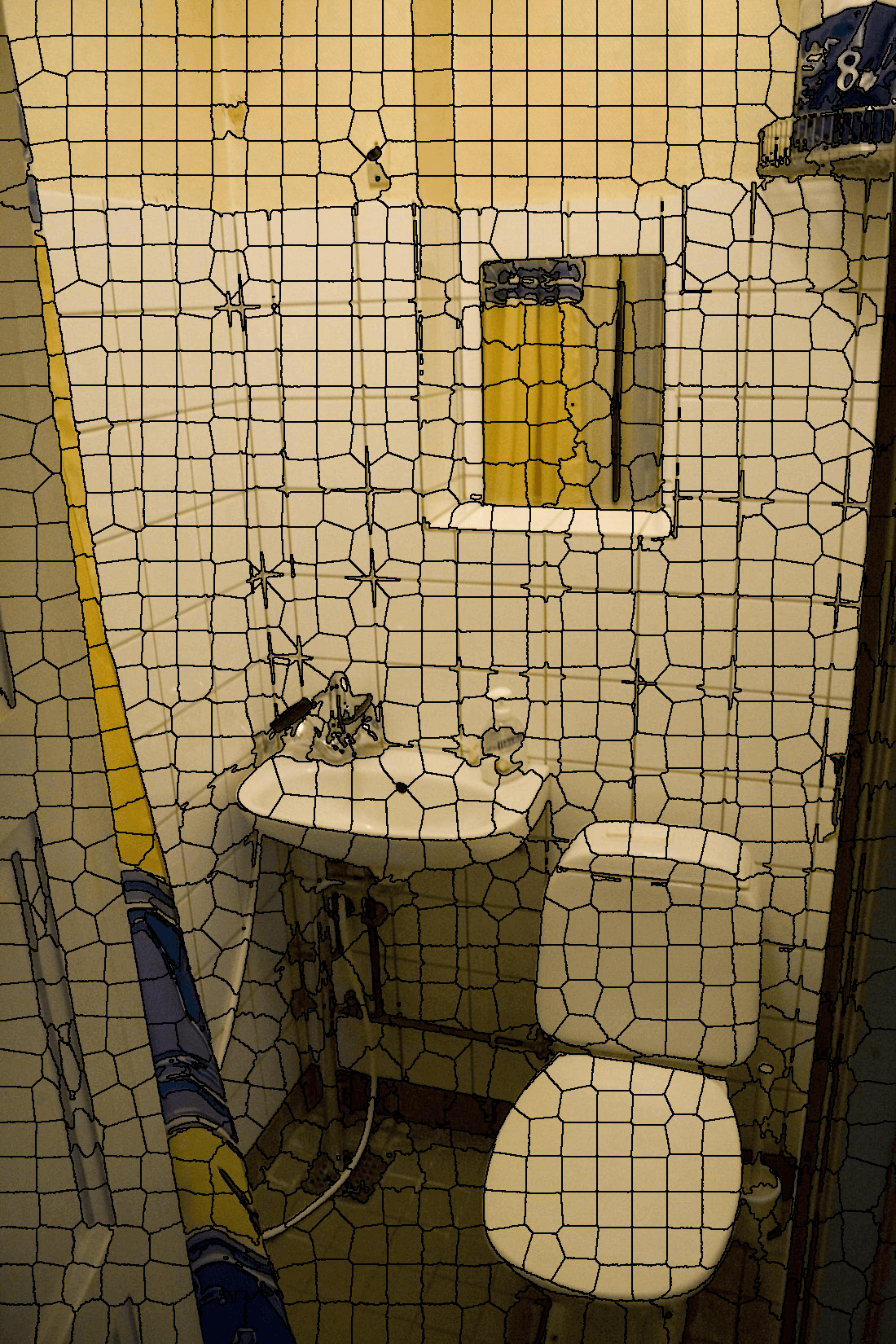}
  }
  \subfigure{%
    \includegraphics[width=.15\columnwidth]{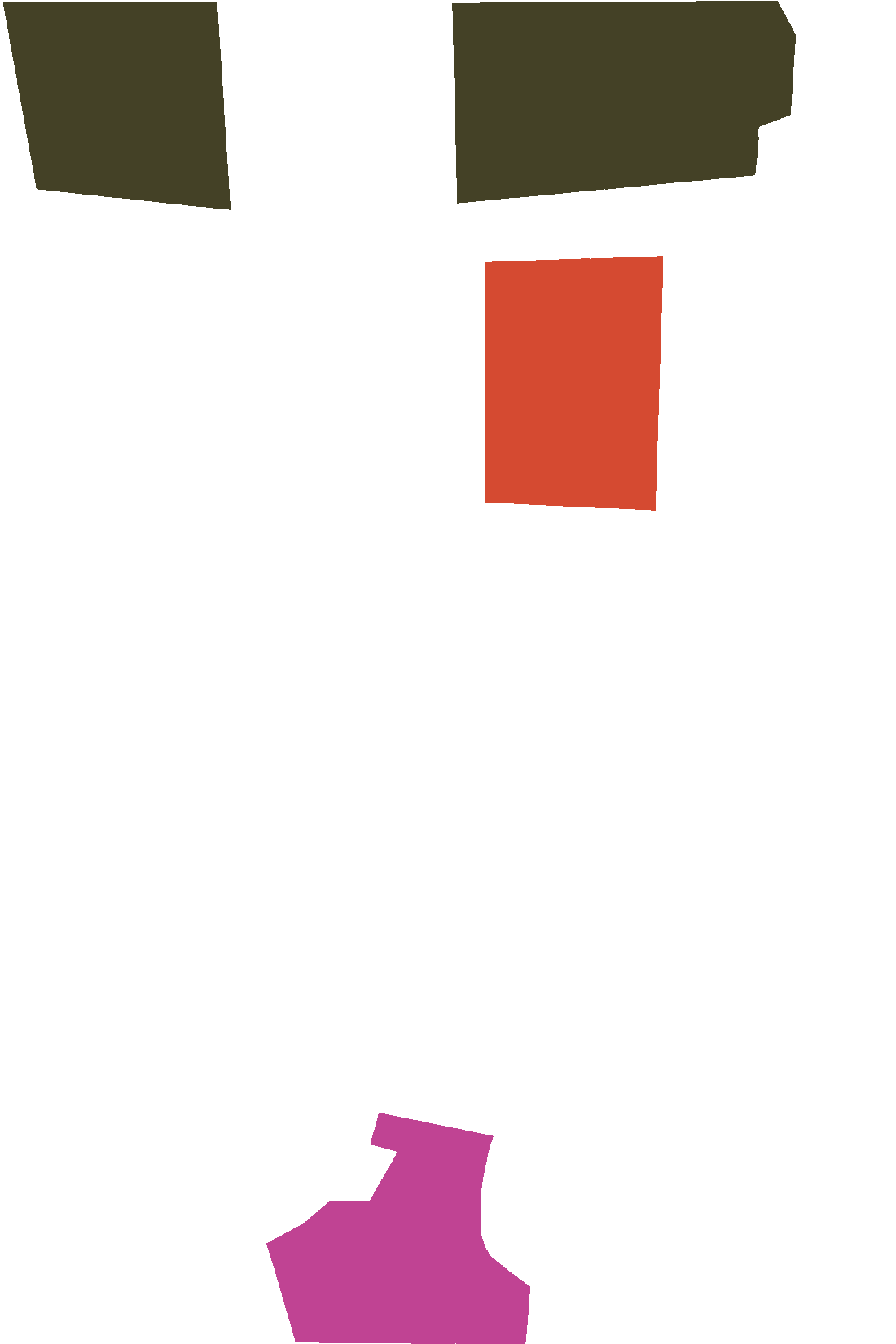}
  }
  \subfigure{%
    \includegraphics[width=.15\columnwidth]{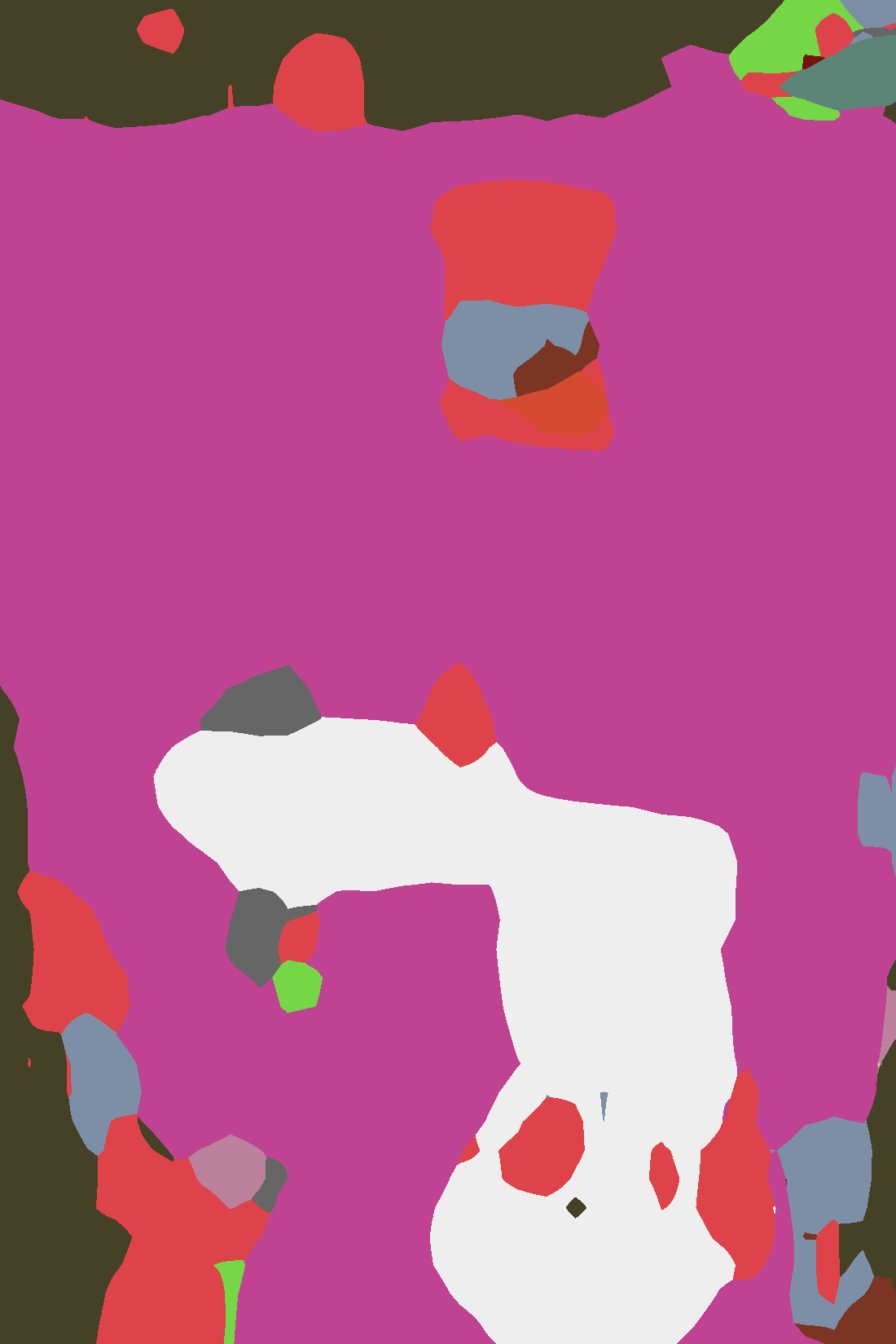}
  }
  \subfigure{%
    \includegraphics[width=.15\columnwidth]{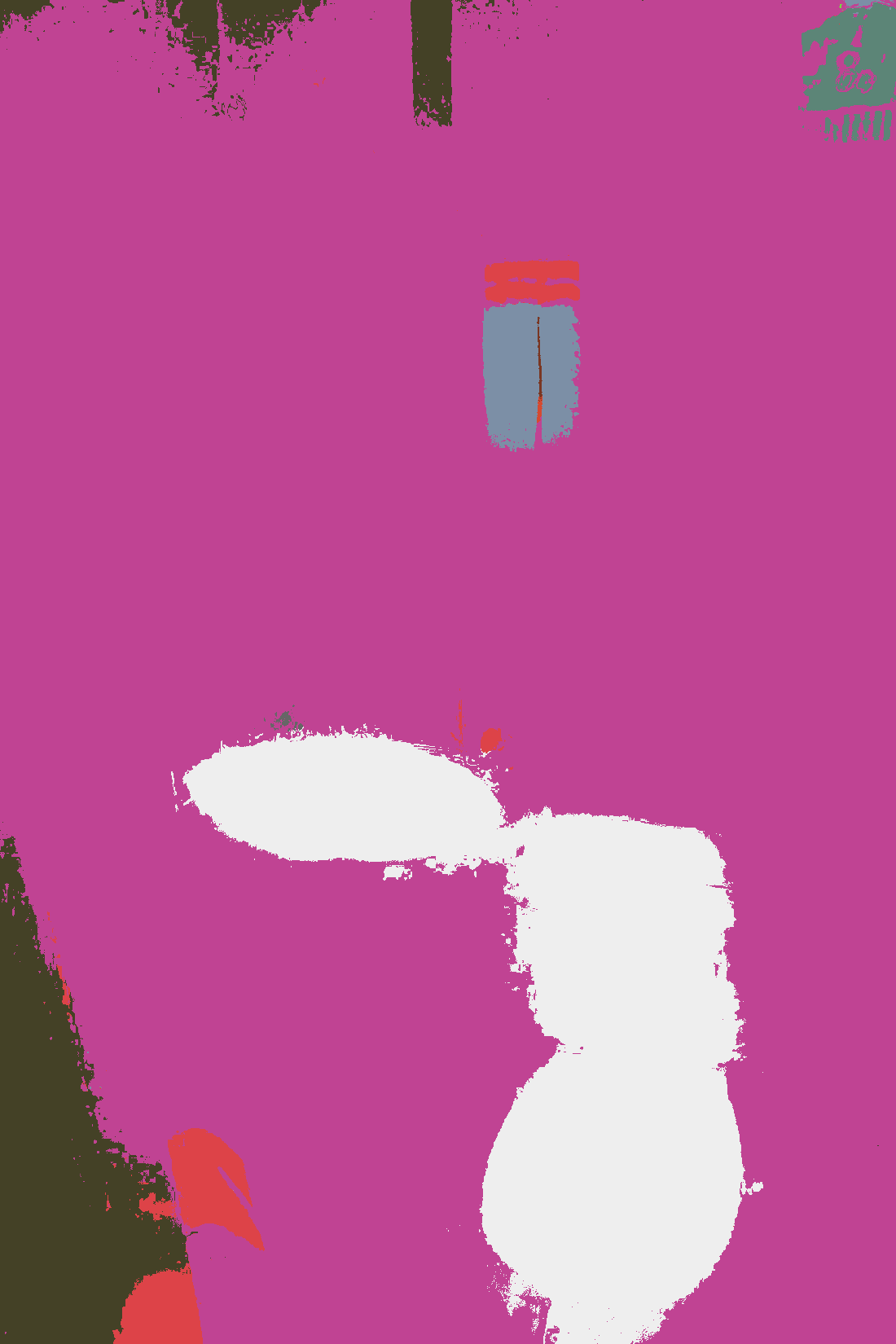}
  }
  \subfigure{%
    \includegraphics[width=.15\columnwidth]{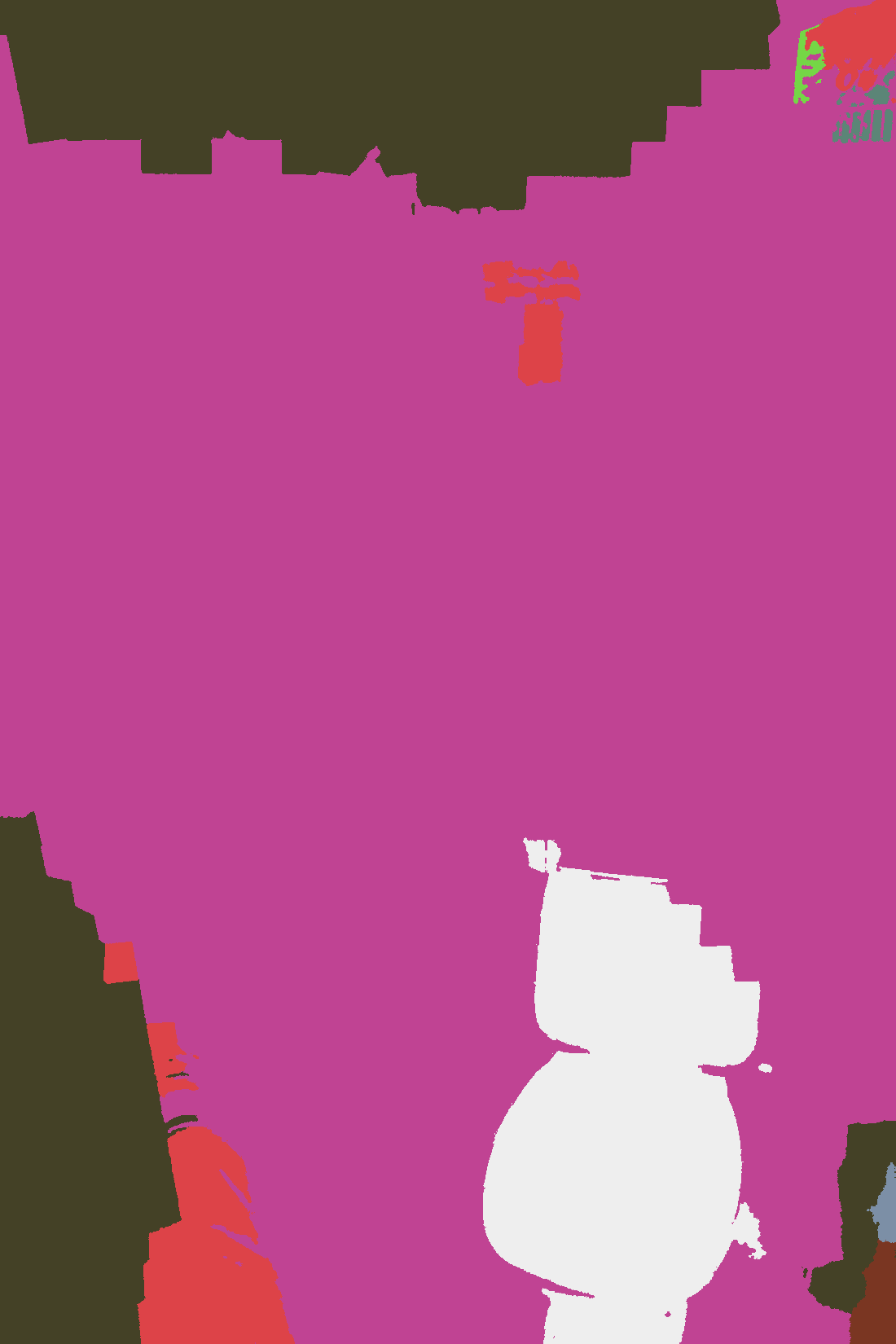}
  }\\[-2ex]

  \subfigure{%
    \includegraphics[width=.15\columnwidth]{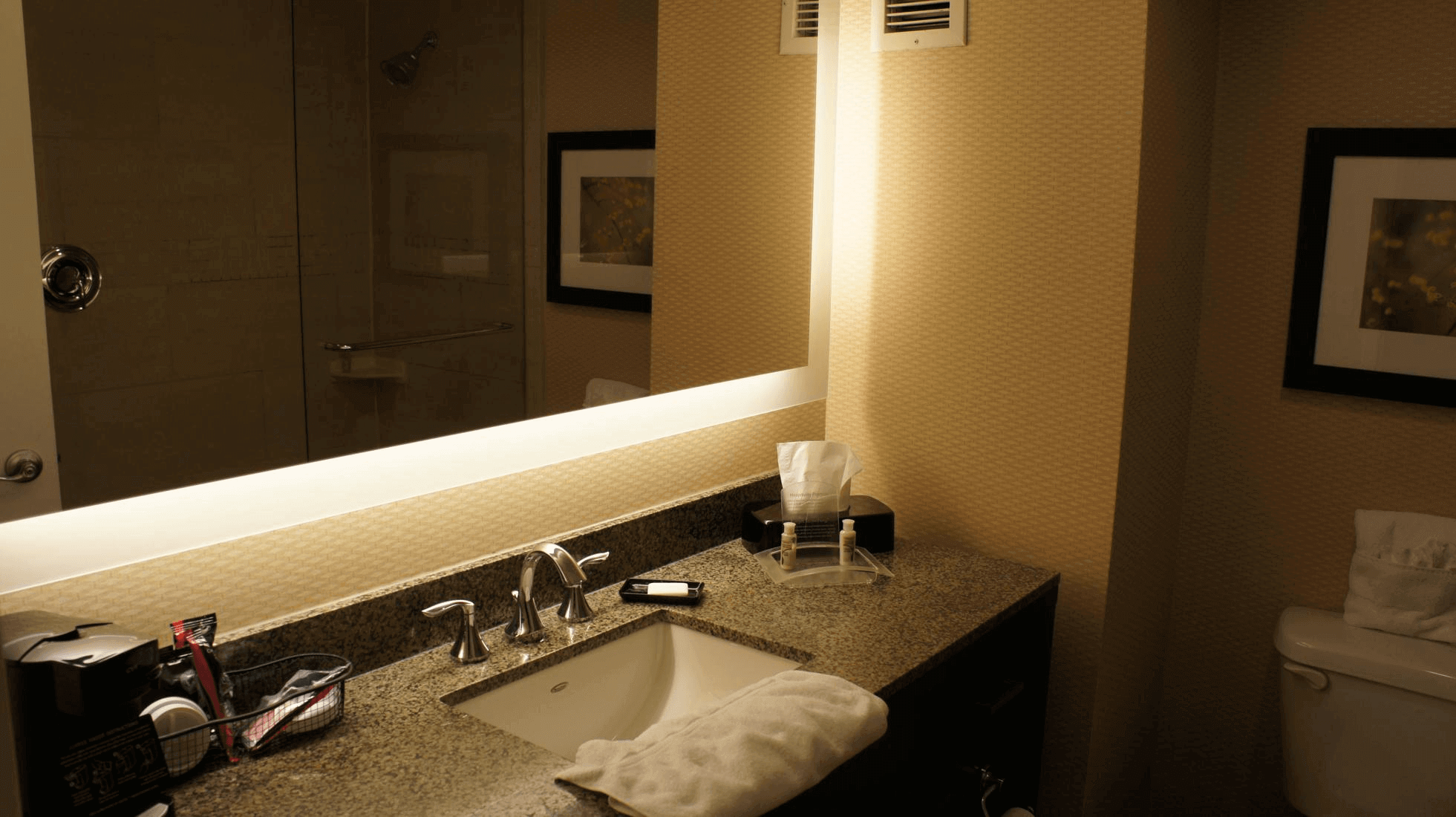}
  }
  \subfigure{%
    \includegraphics[width=.15\columnwidth]{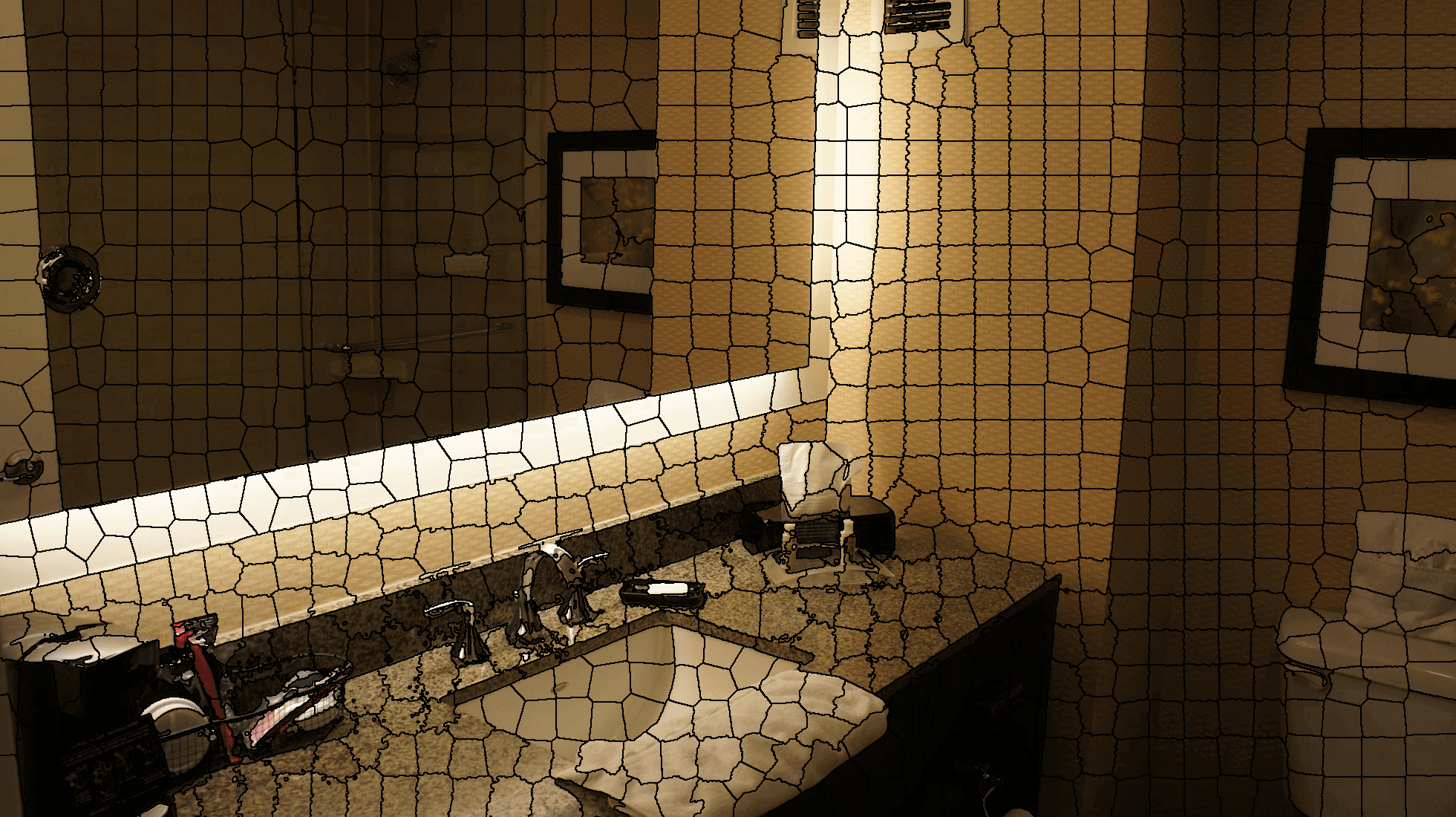}
  }
  \subfigure{%
    \includegraphics[width=.15\columnwidth]{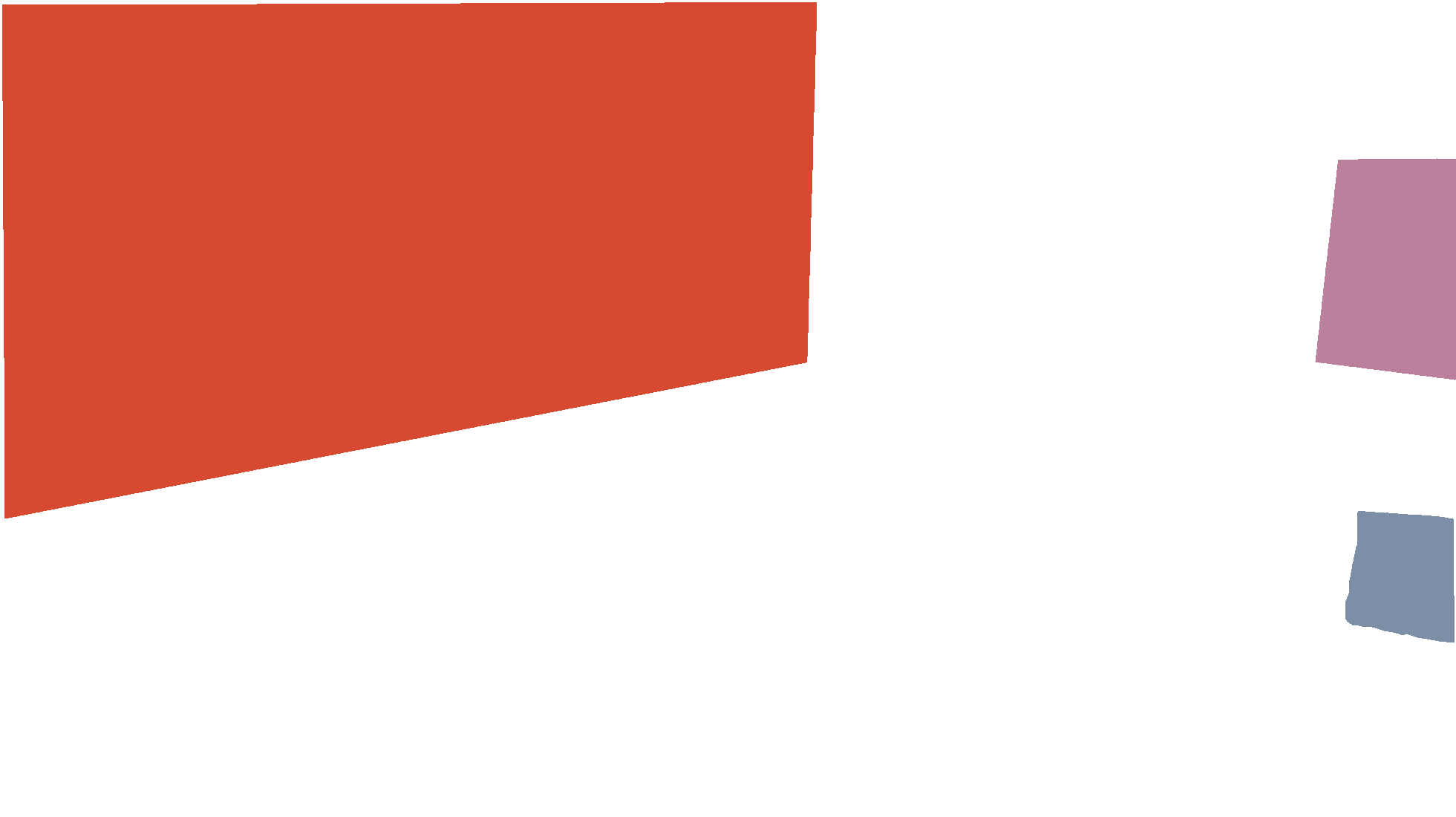}
  }
  \subfigure{%
    \includegraphics[width=.15\columnwidth]{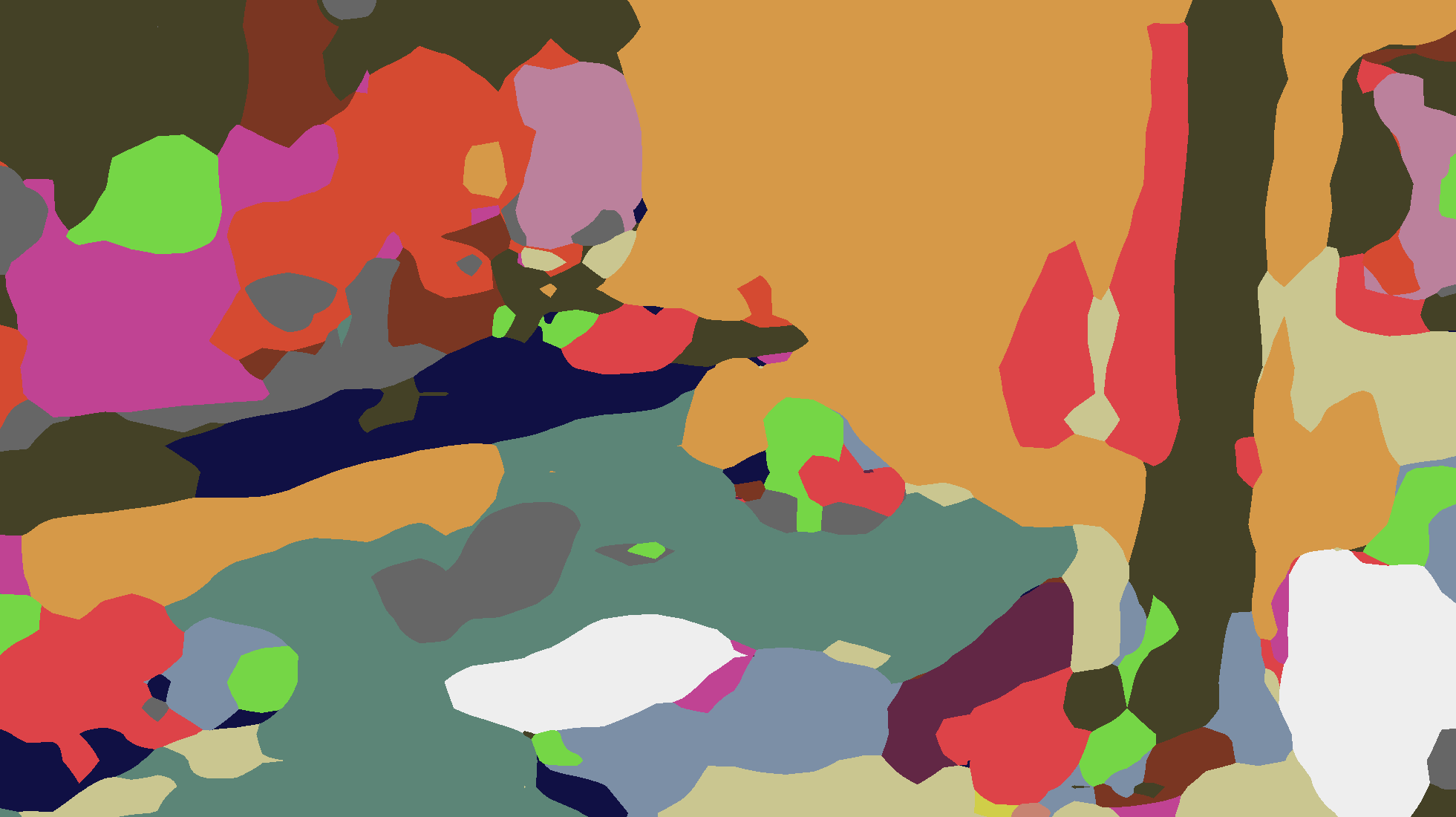}
  }
  \subfigure{%
    \includegraphics[width=.15\columnwidth]{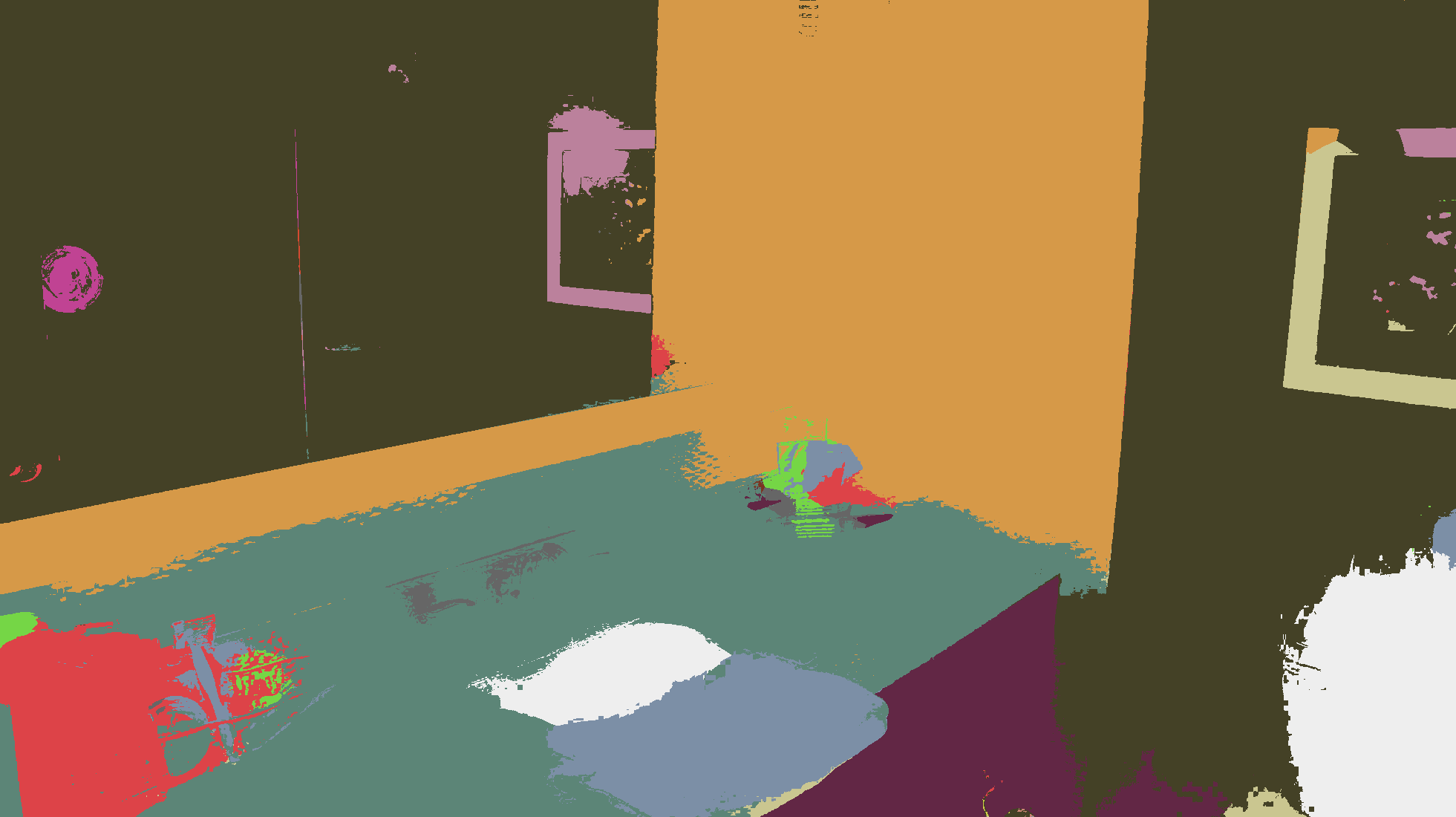}
  }
  \subfigure{%
    \includegraphics[width=.15\columnwidth]{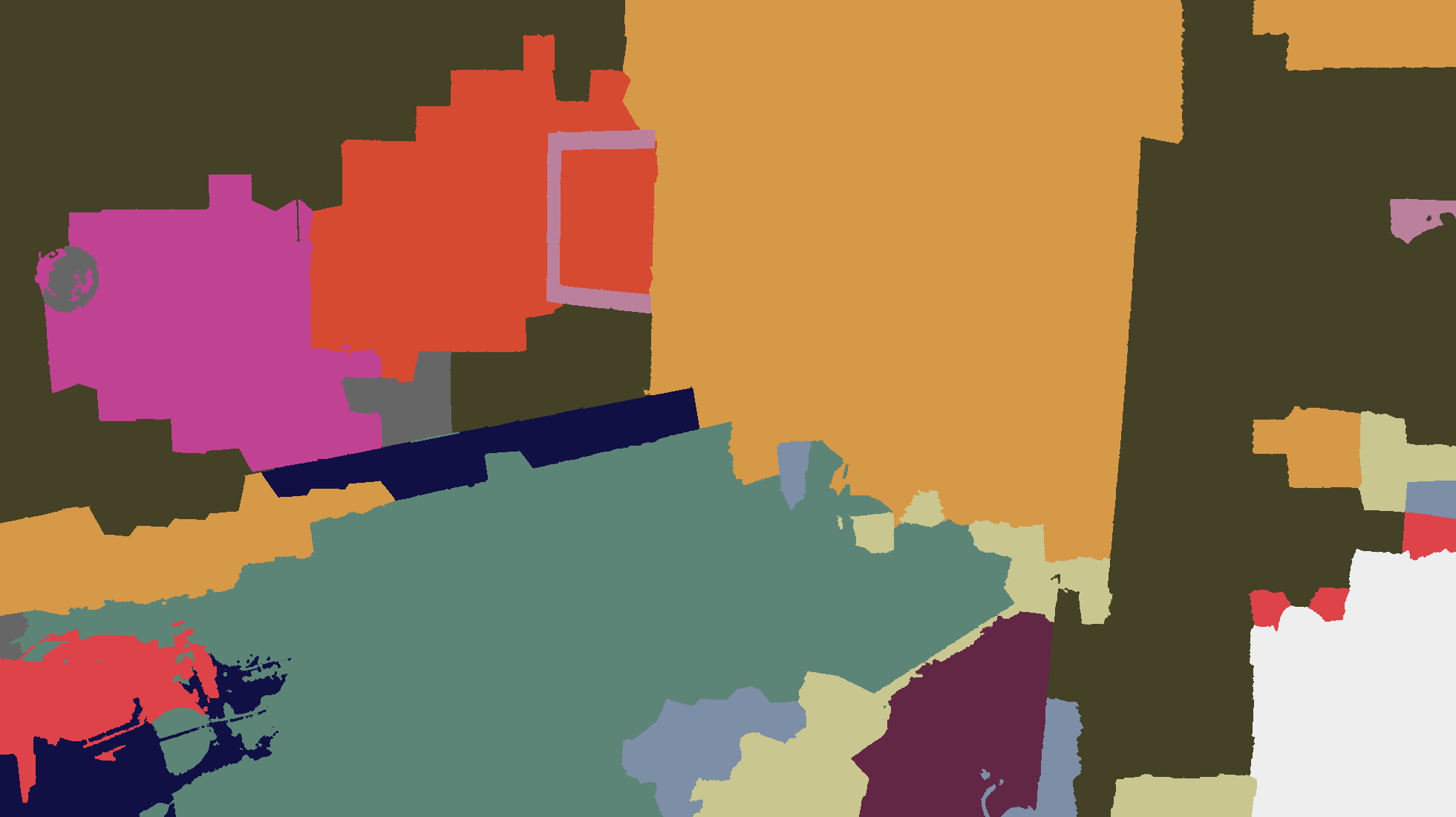}
  }\\[-2ex]

  \subfigure{%
    \includegraphics[width=.15\columnwidth]{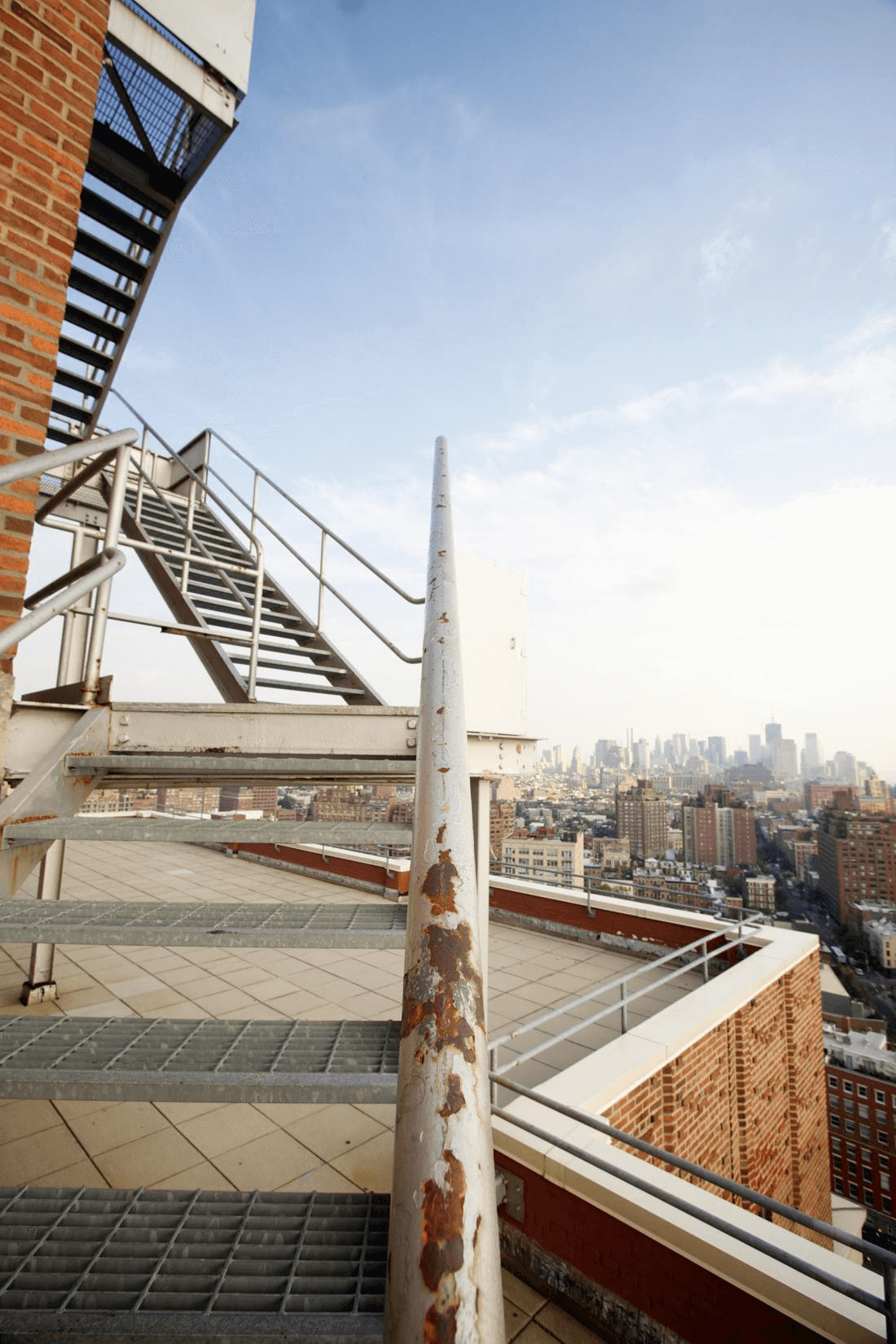}
  }
  \subfigure{%
    \includegraphics[width=.15\columnwidth]{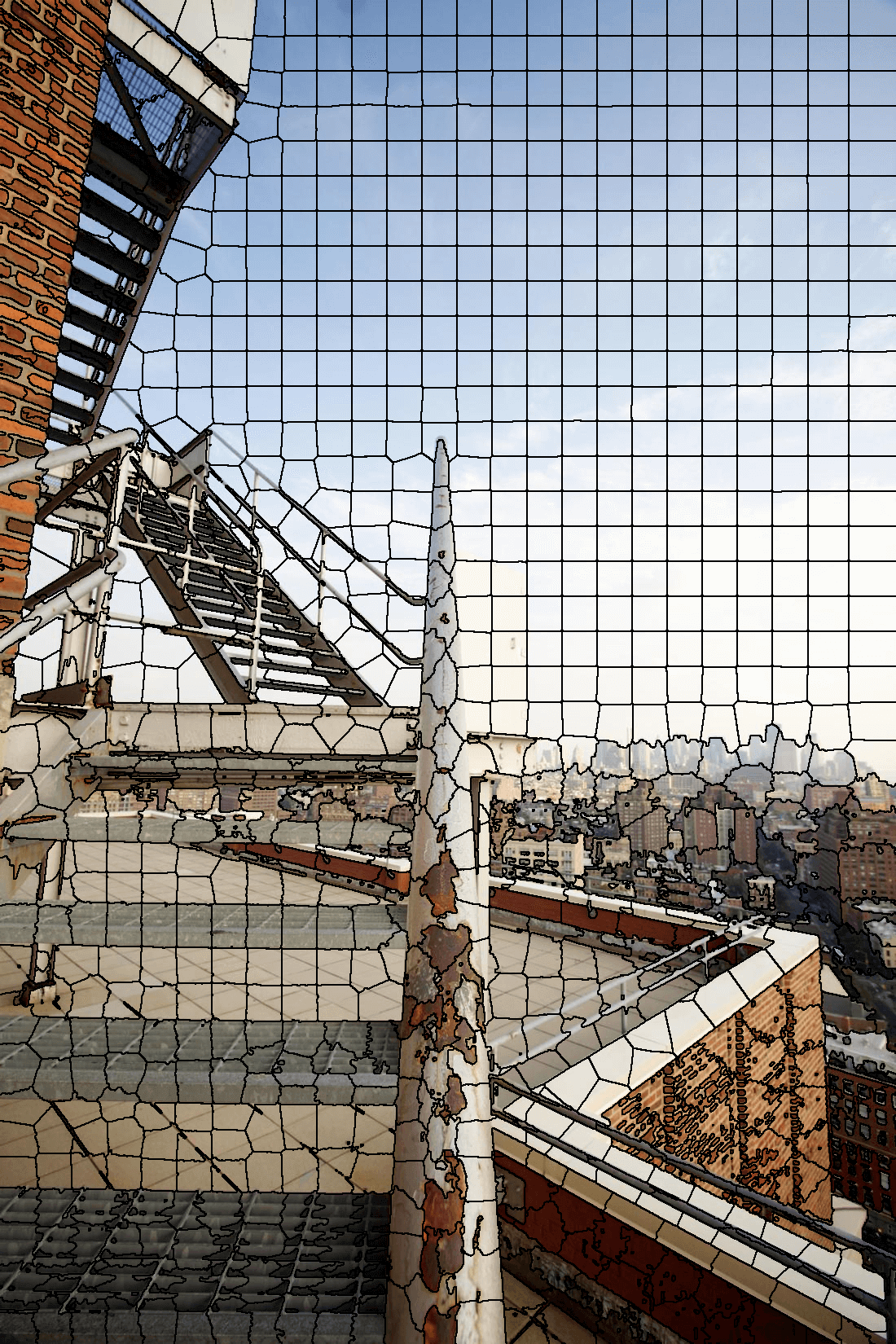}
  }
  \subfigure{%
    \includegraphics[width=.15\columnwidth]{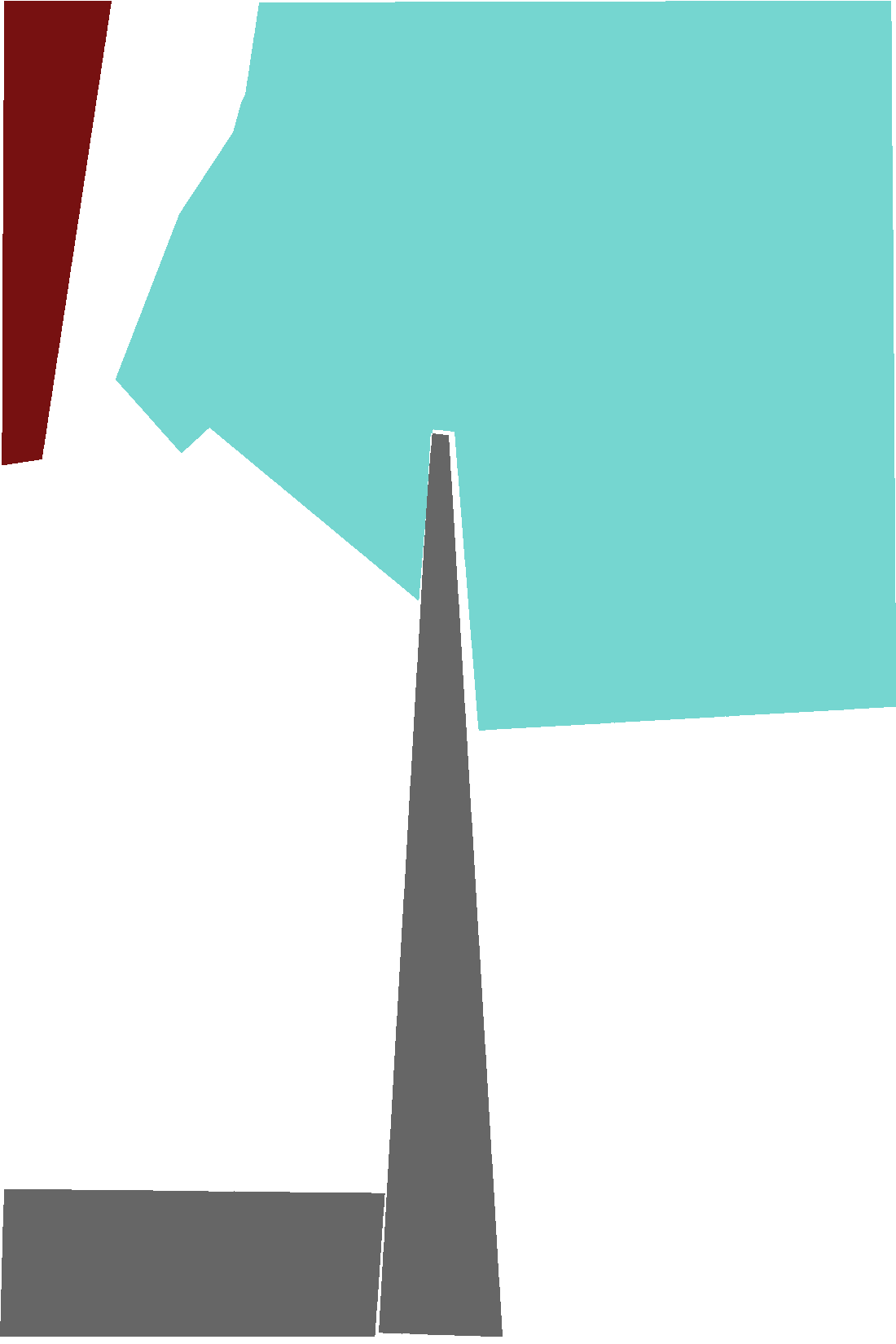}
  }
  \subfigure{%
    \includegraphics[width=.15\columnwidth]{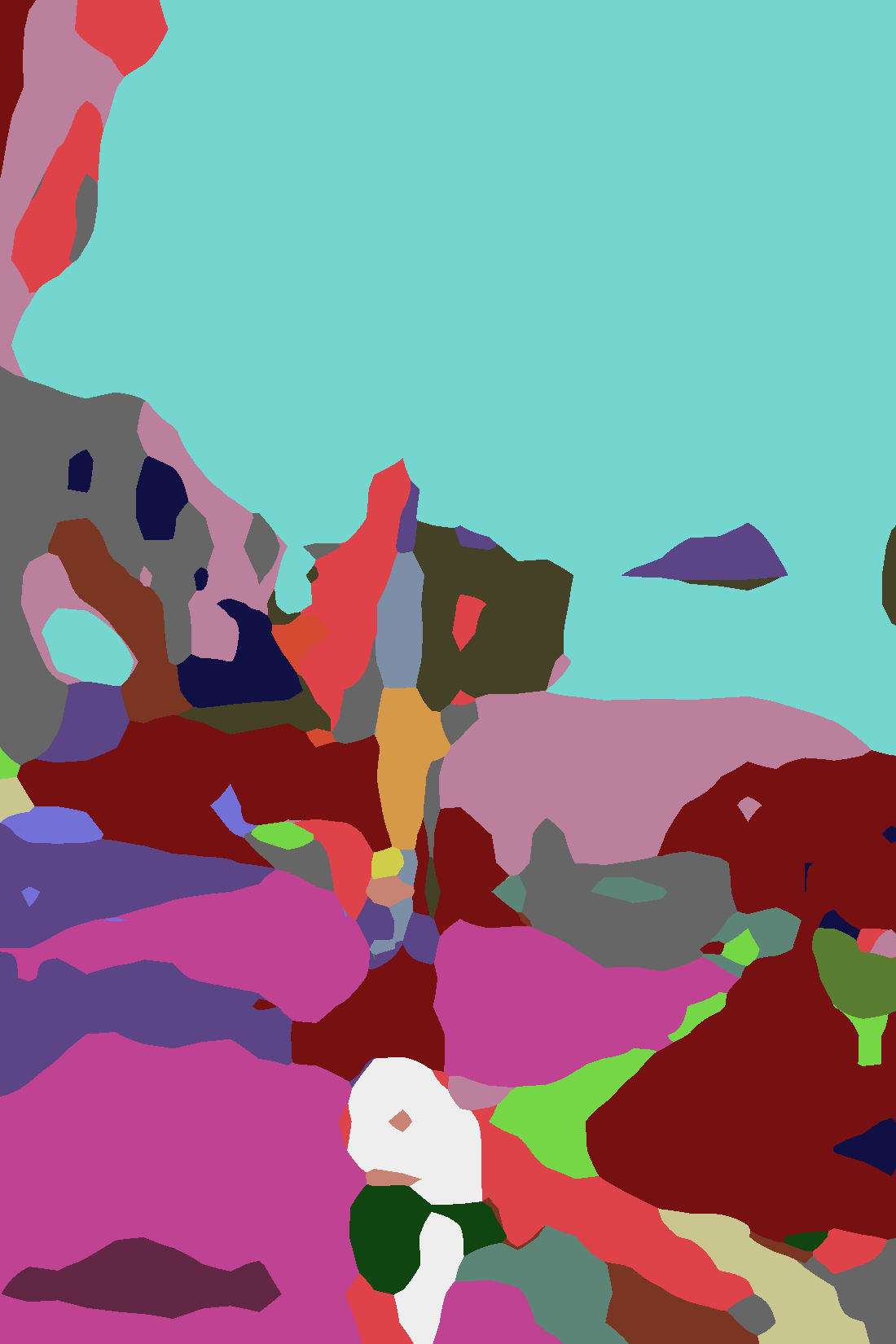}
  }
  \subfigure{%
    \includegraphics[width=.15\columnwidth]{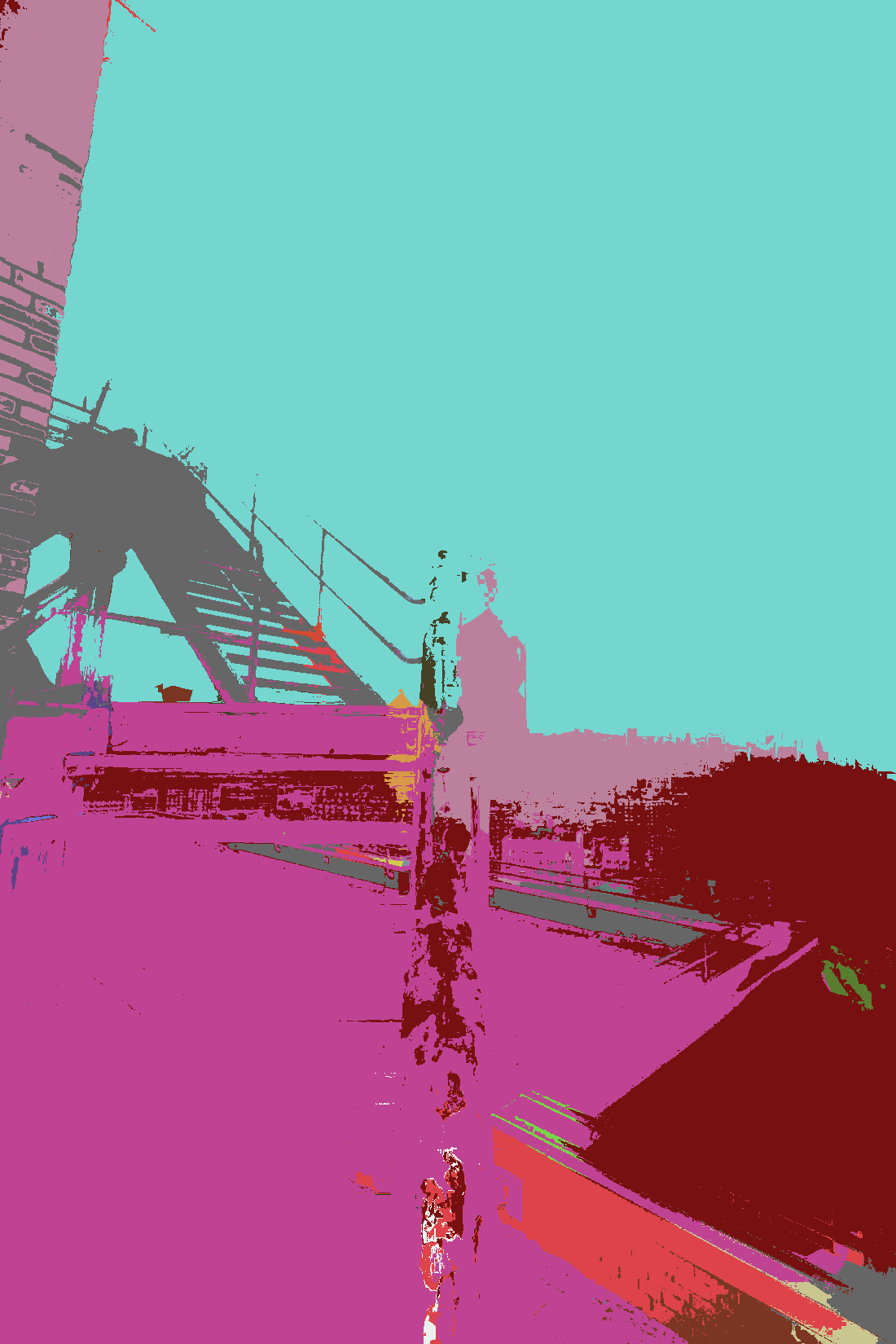}
  }
  \subfigure{%
    \includegraphics[width=.15\columnwidth]{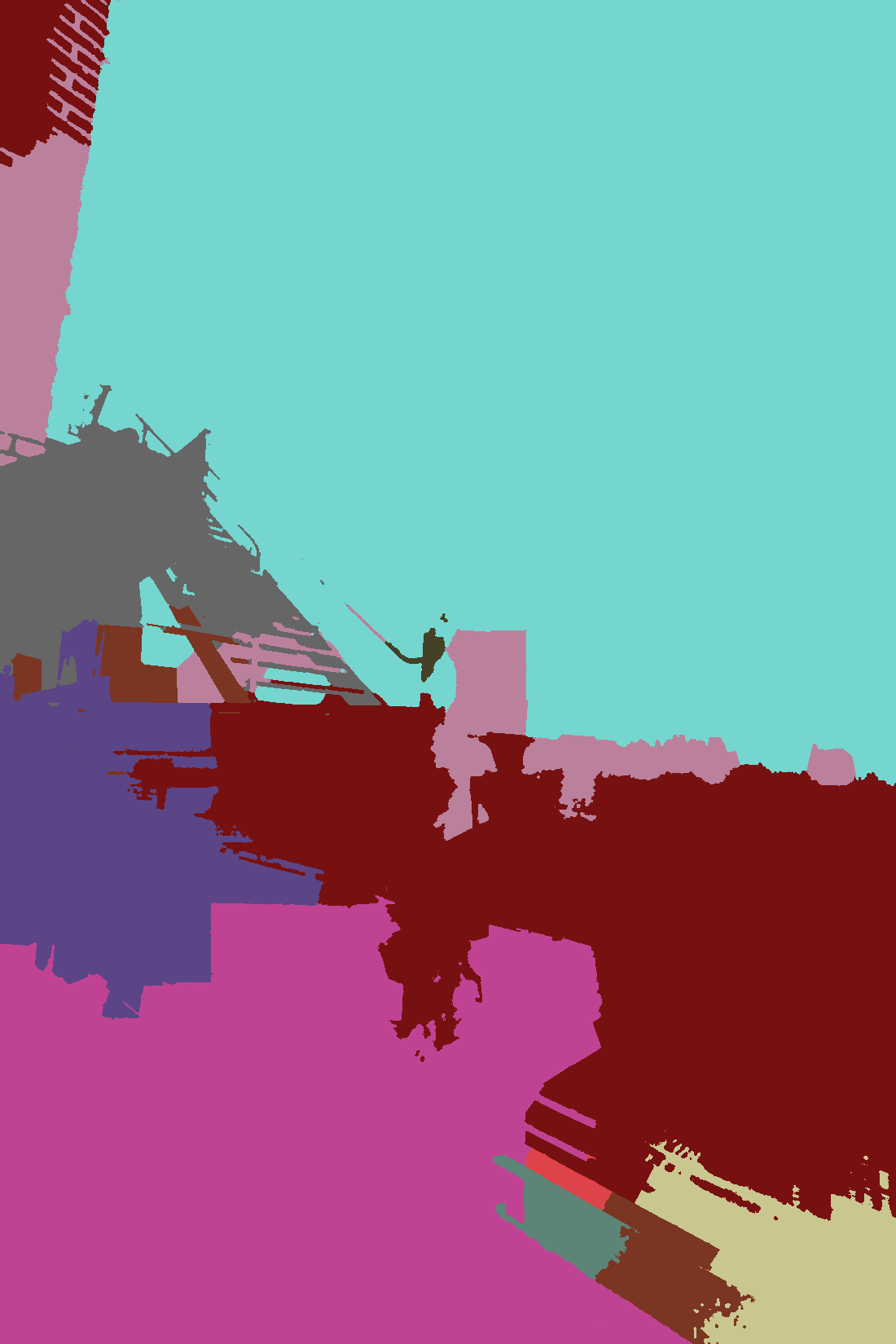}
  }\\[-2ex]

  \subfigure{%
    \includegraphics[width=.15\columnwidth]{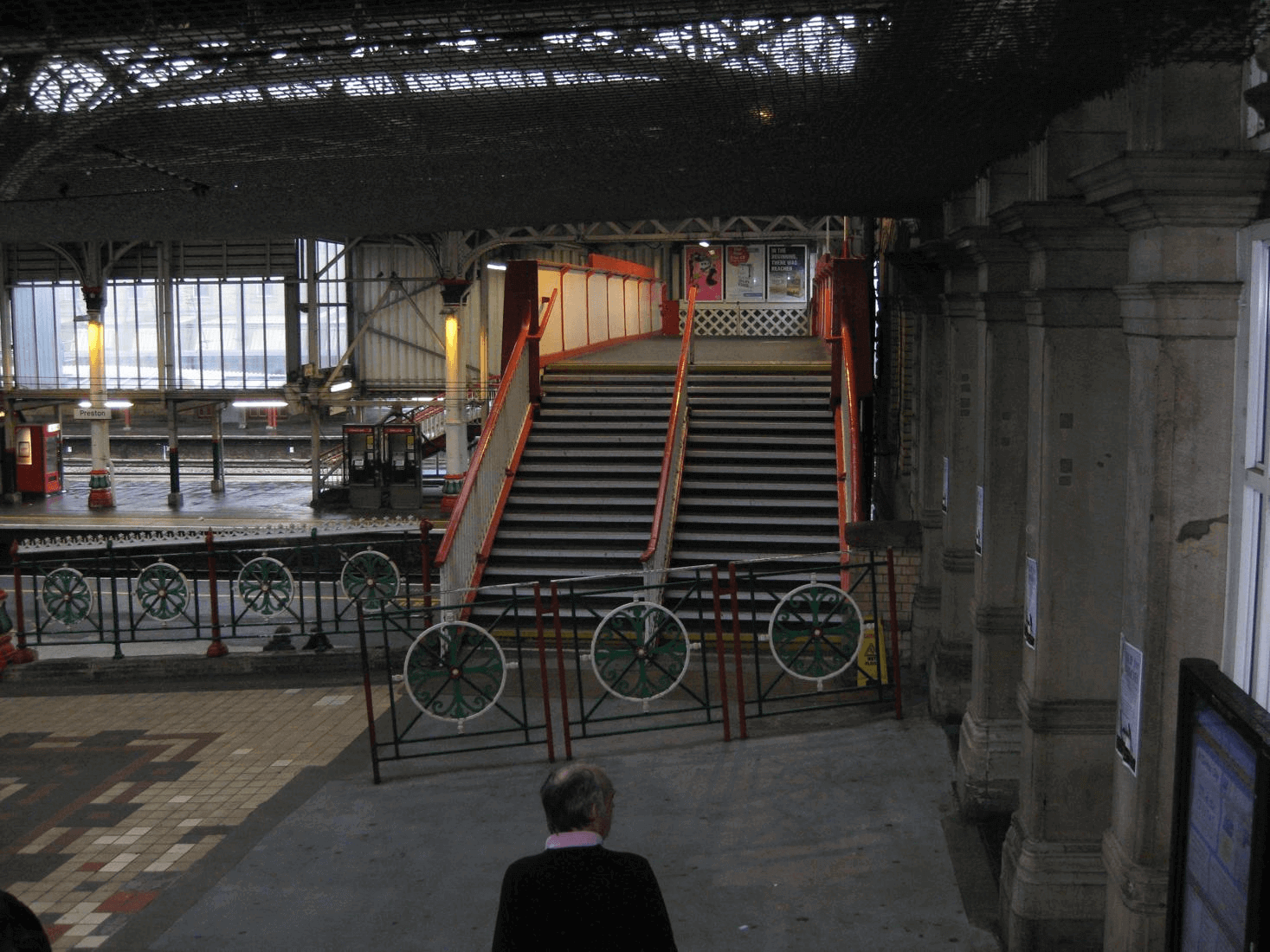}
  }
  \subfigure{%
    \includegraphics[width=.15\columnwidth]{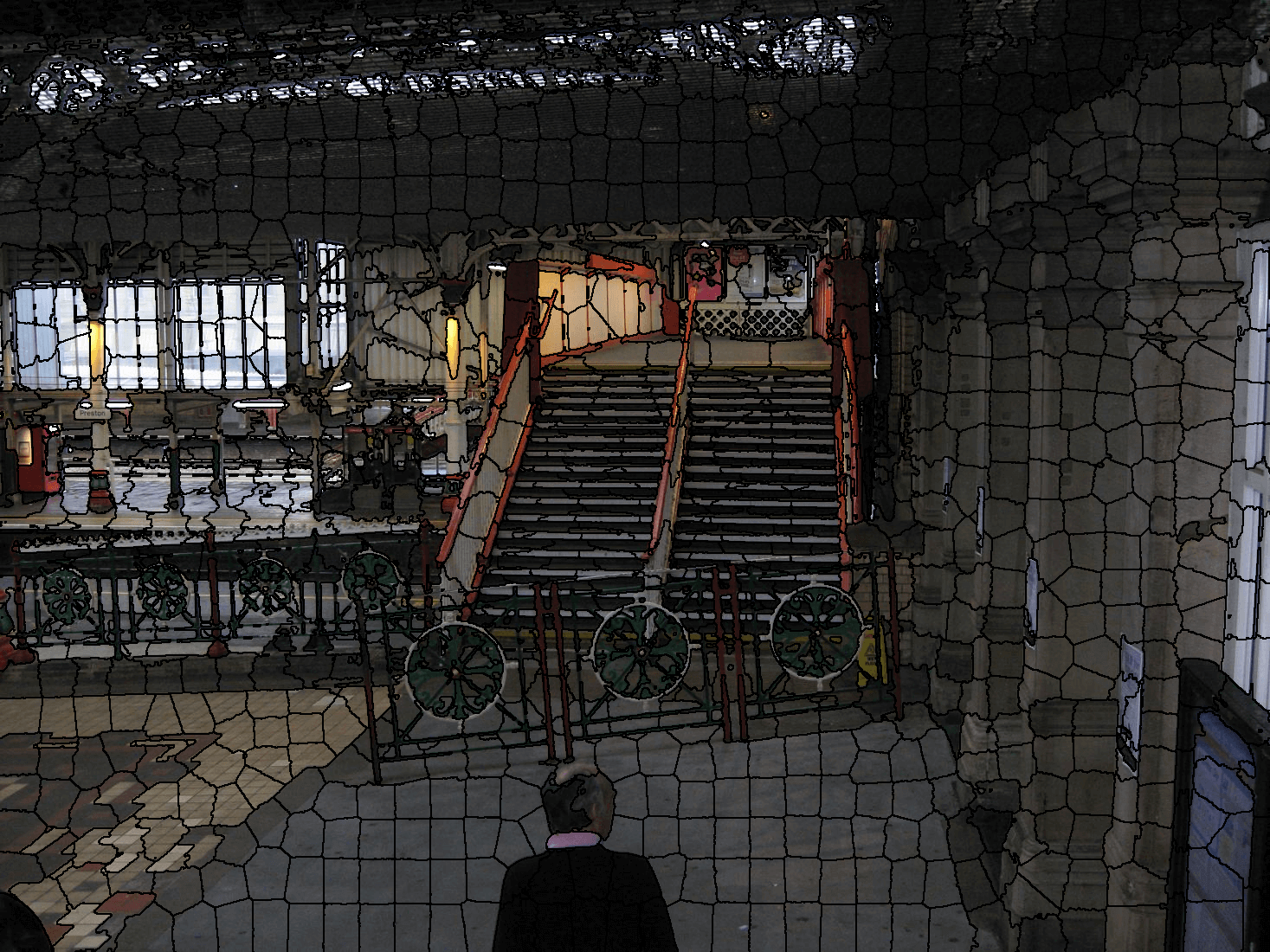}
  }
  \subfigure{%
    \includegraphics[width=.15\columnwidth]{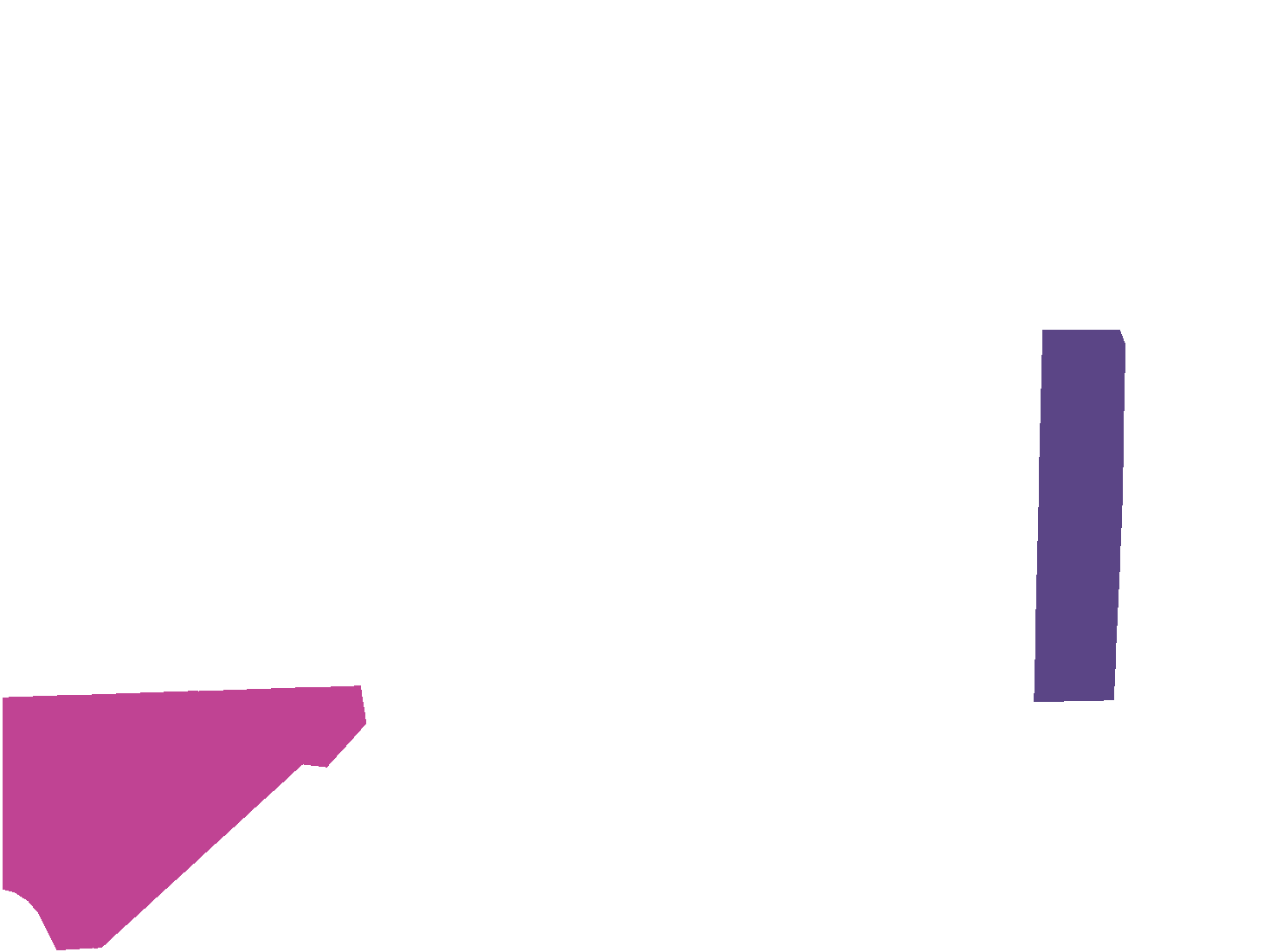}
  }
  \subfigure{%
    \includegraphics[width=.15\columnwidth]{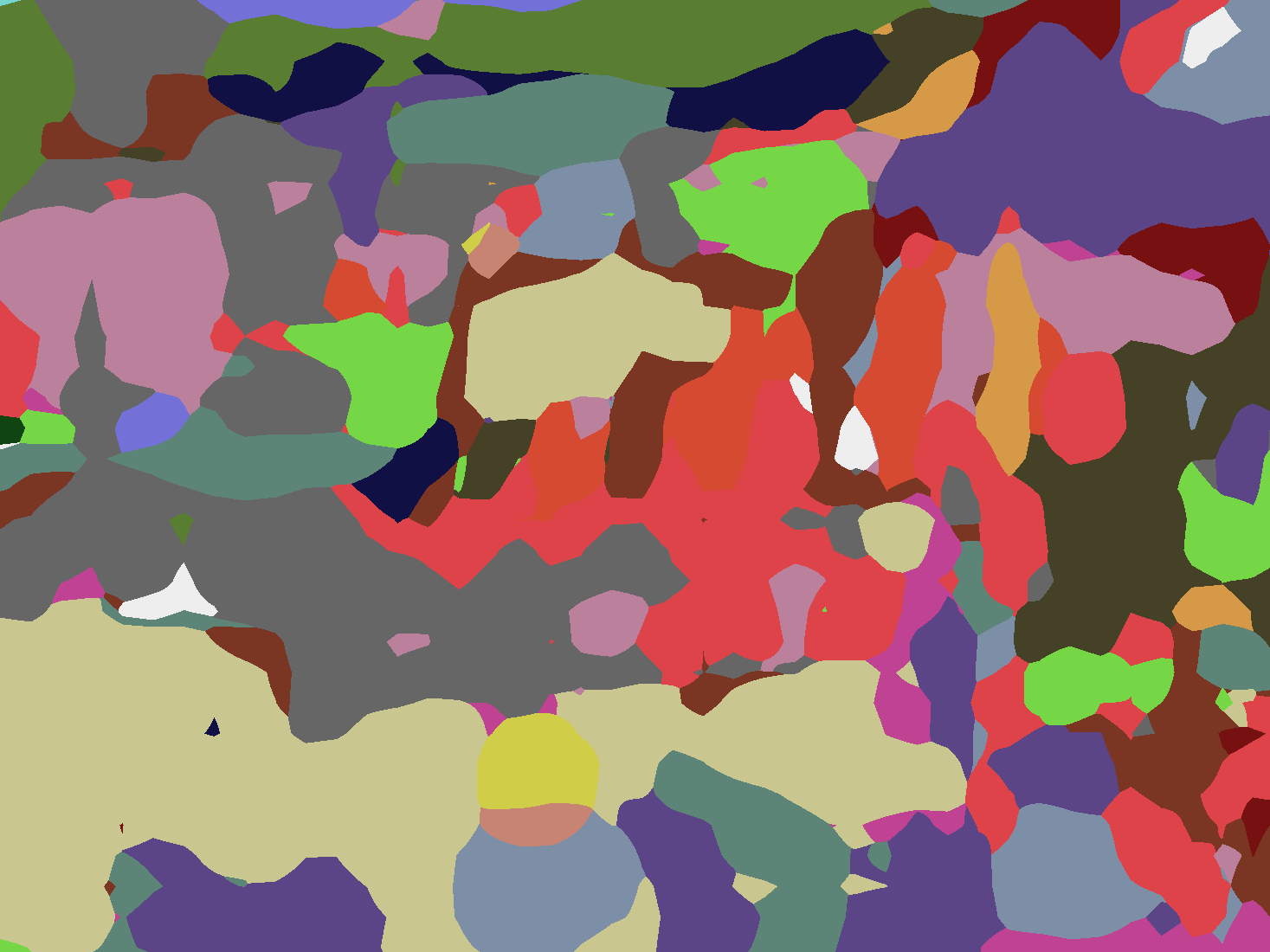}
  }
  \subfigure{%
    \includegraphics[width=.15\columnwidth]{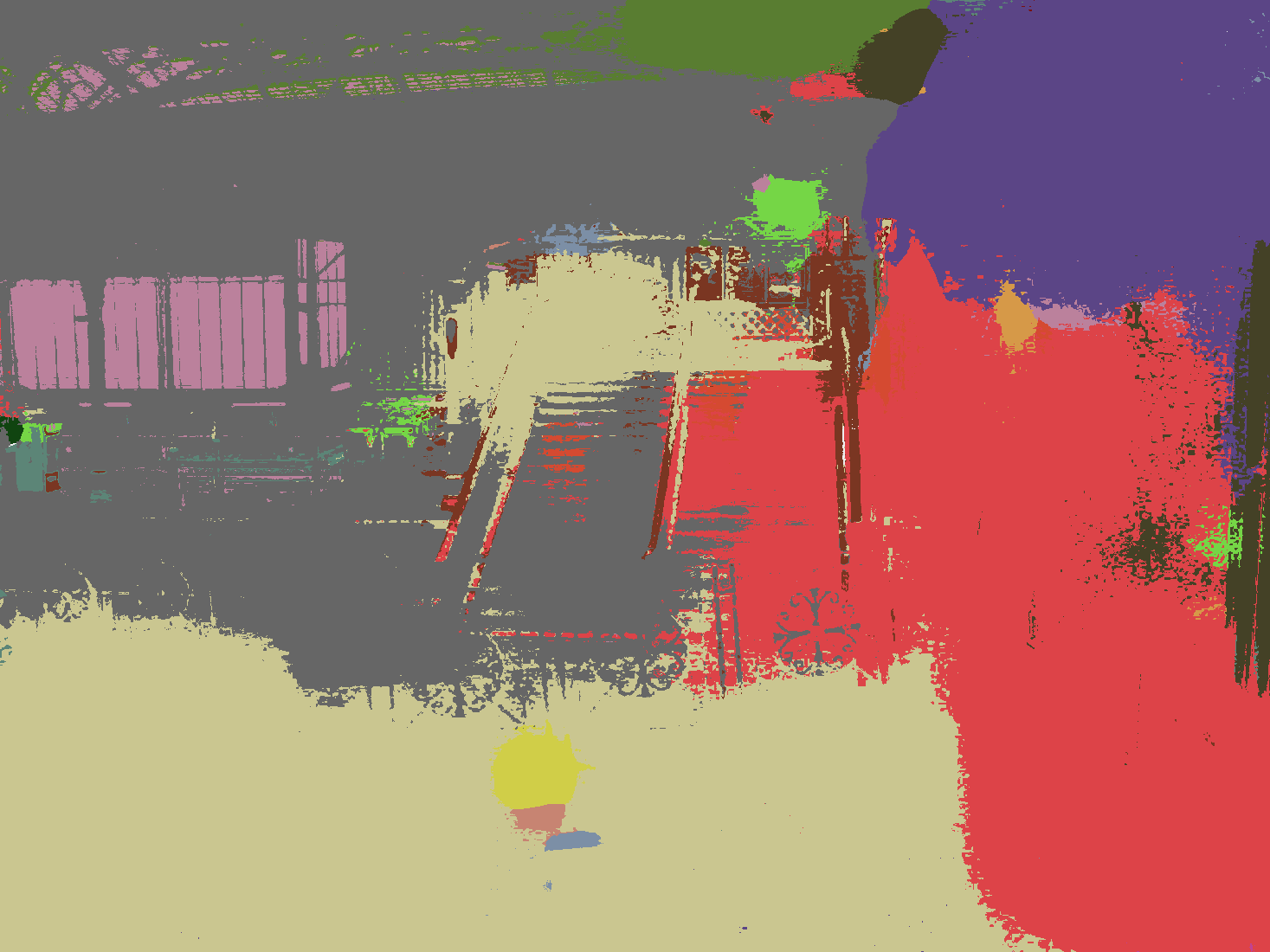}
  }
  \subfigure{%
    \includegraphics[width=.15\columnwidth]{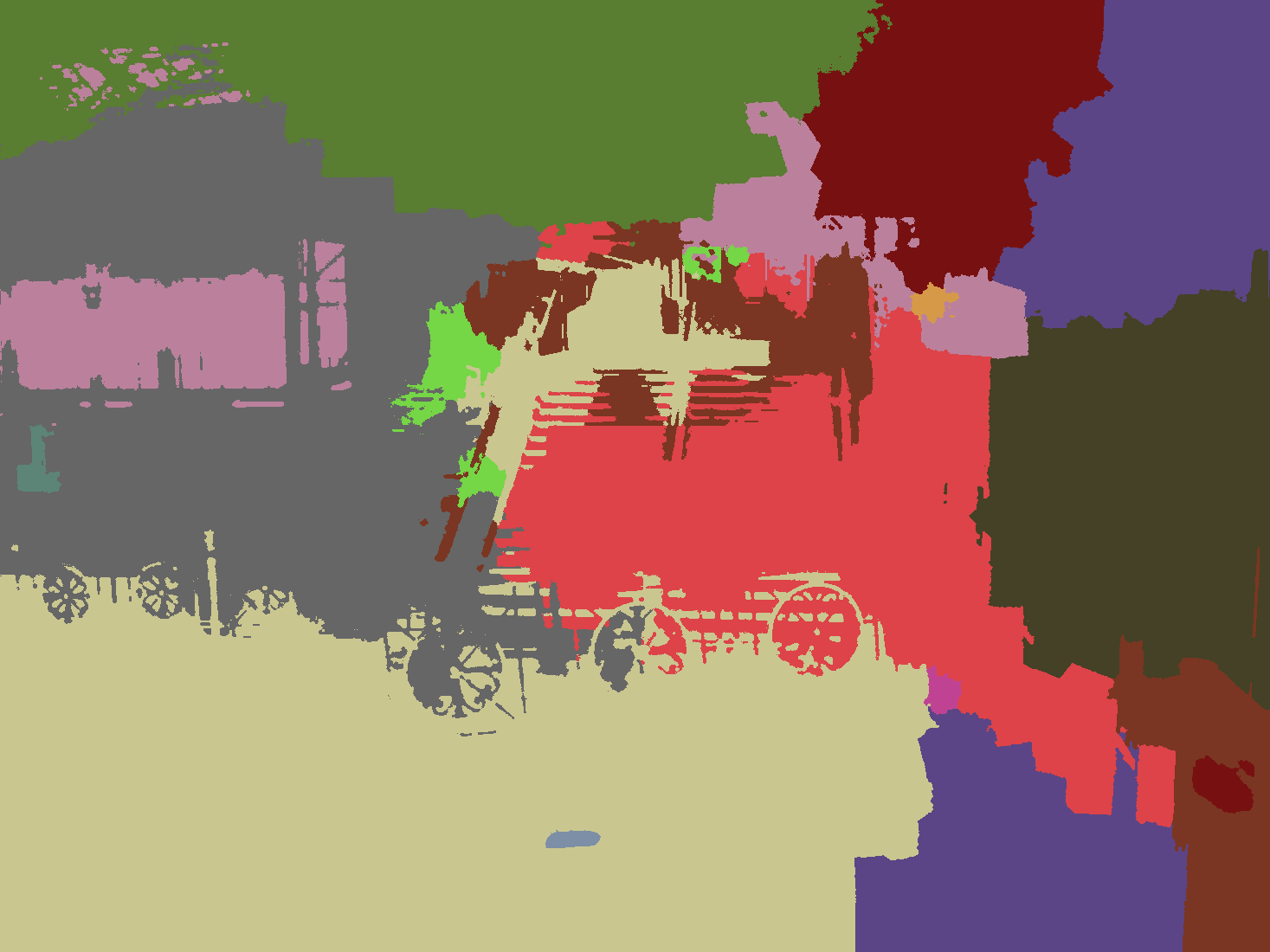}
  }\\[-2ex]

  \subfigure{%
    \includegraphics[width=.15\columnwidth]{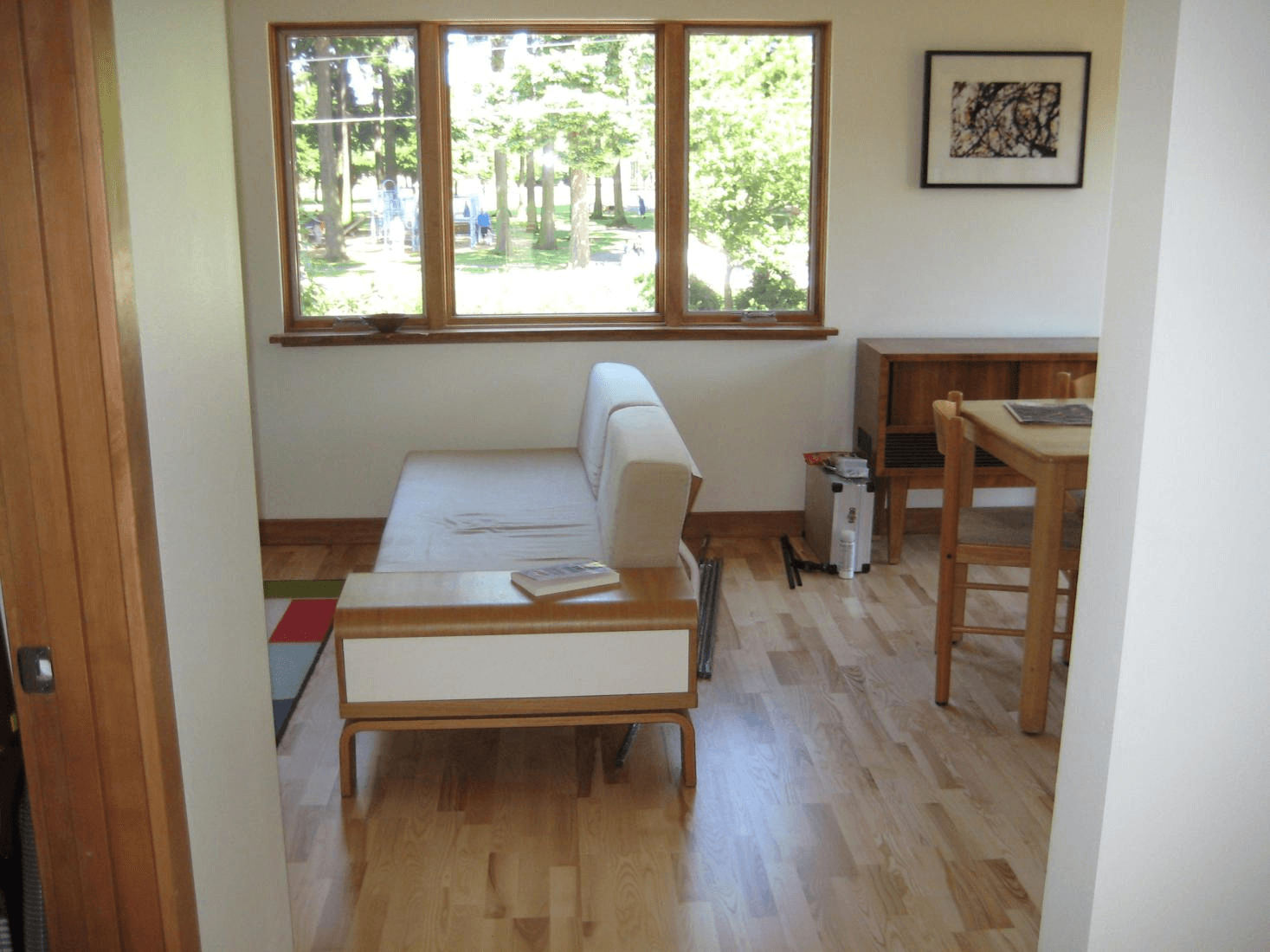}
  }
  \subfigure{%
    \includegraphics[width=.15\columnwidth]{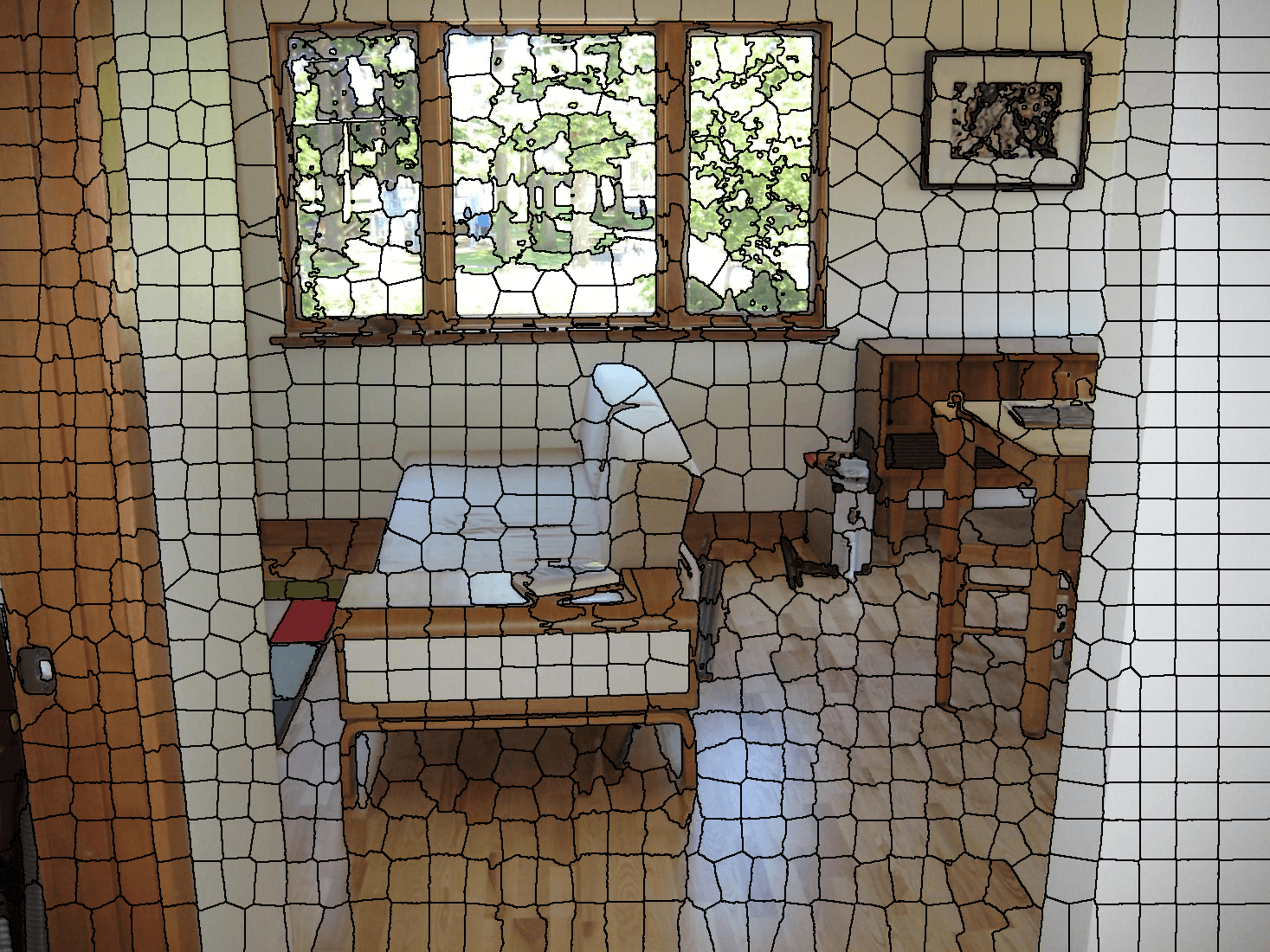}
  }
  \subfigure{%
    \includegraphics[width=.15\columnwidth]{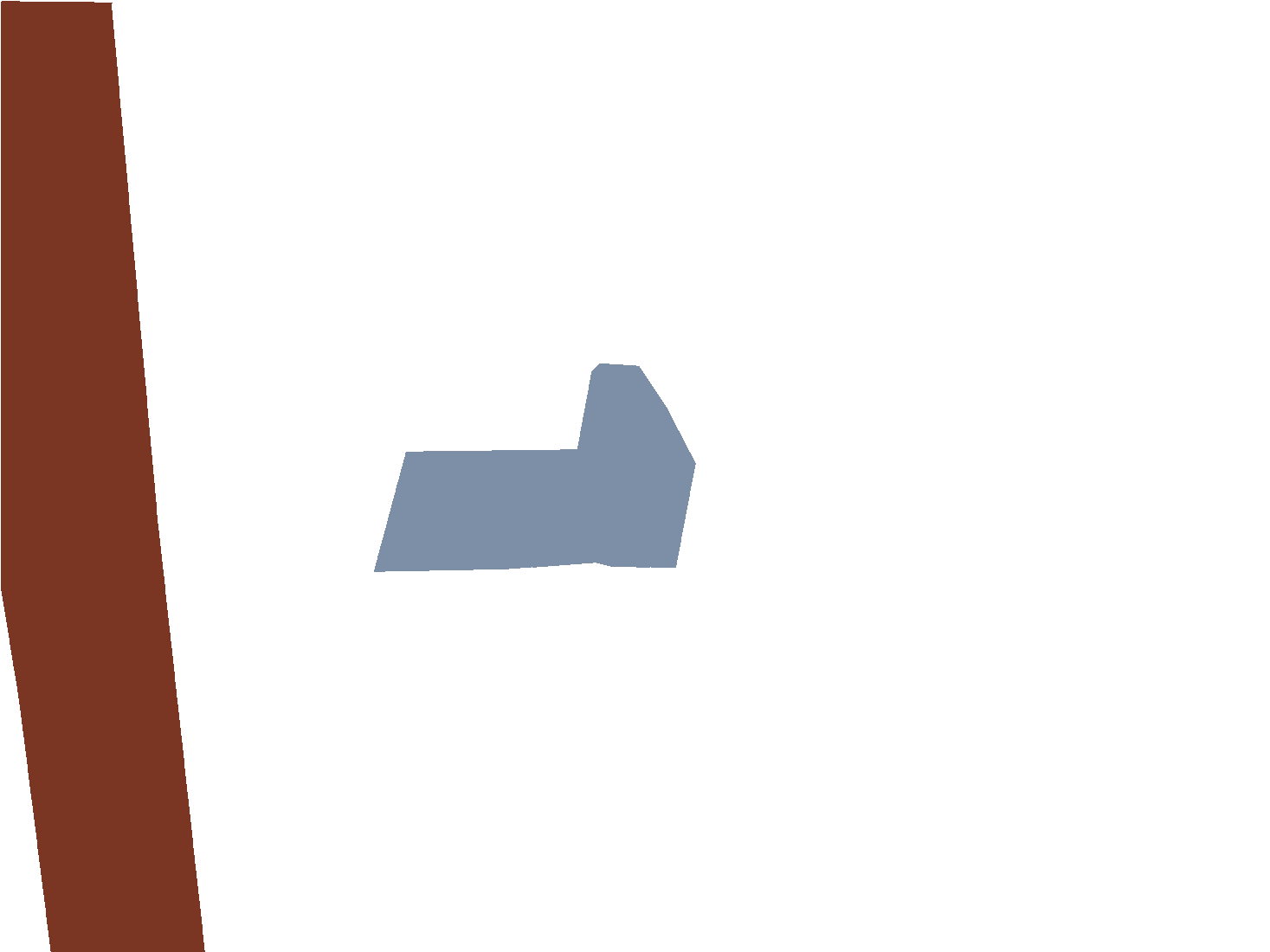}
  }
  \subfigure{%
    \includegraphics[width=.15\columnwidth]{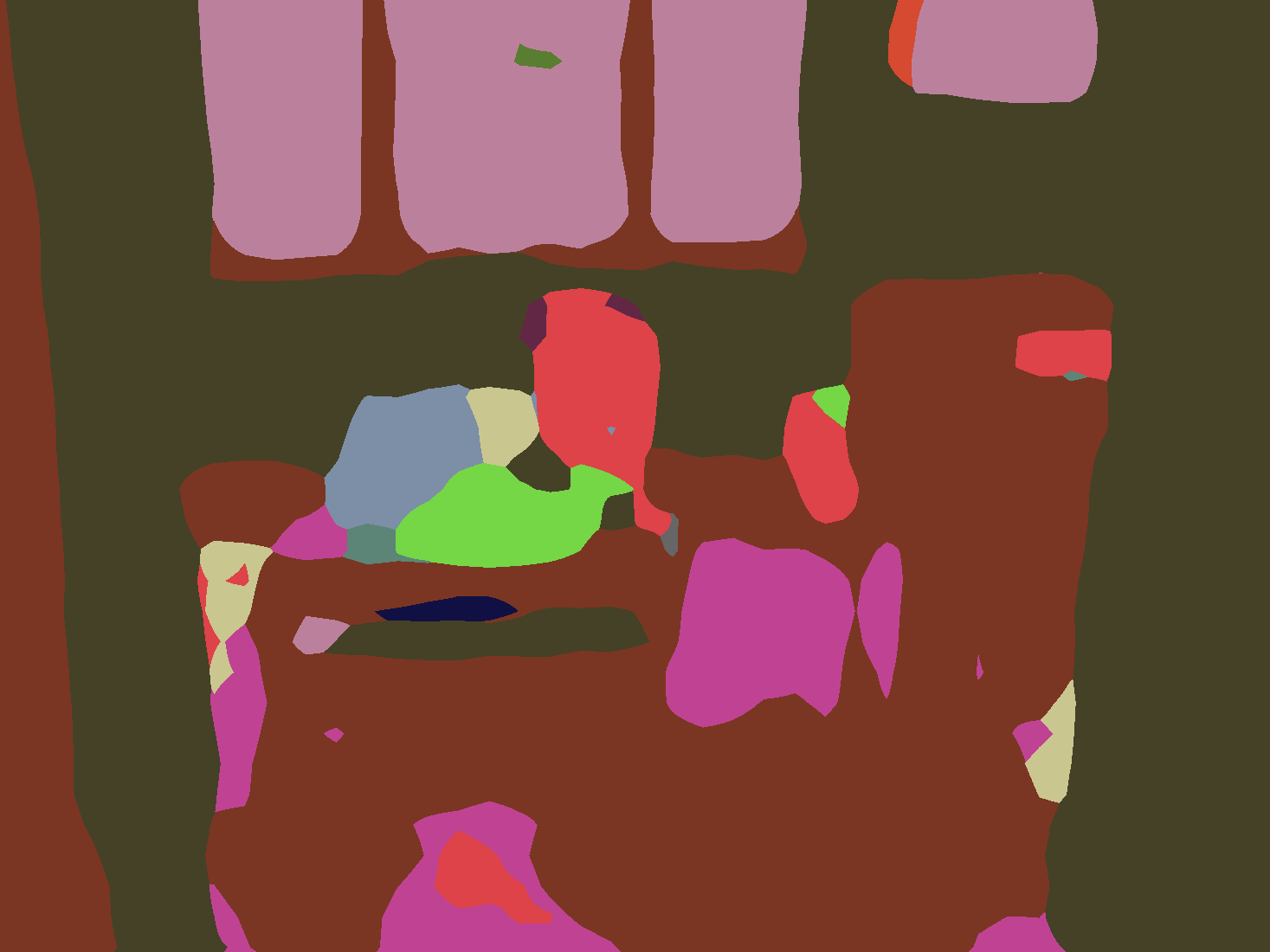}
  }
  \subfigure{%
    \includegraphics[width=.15\columnwidth]{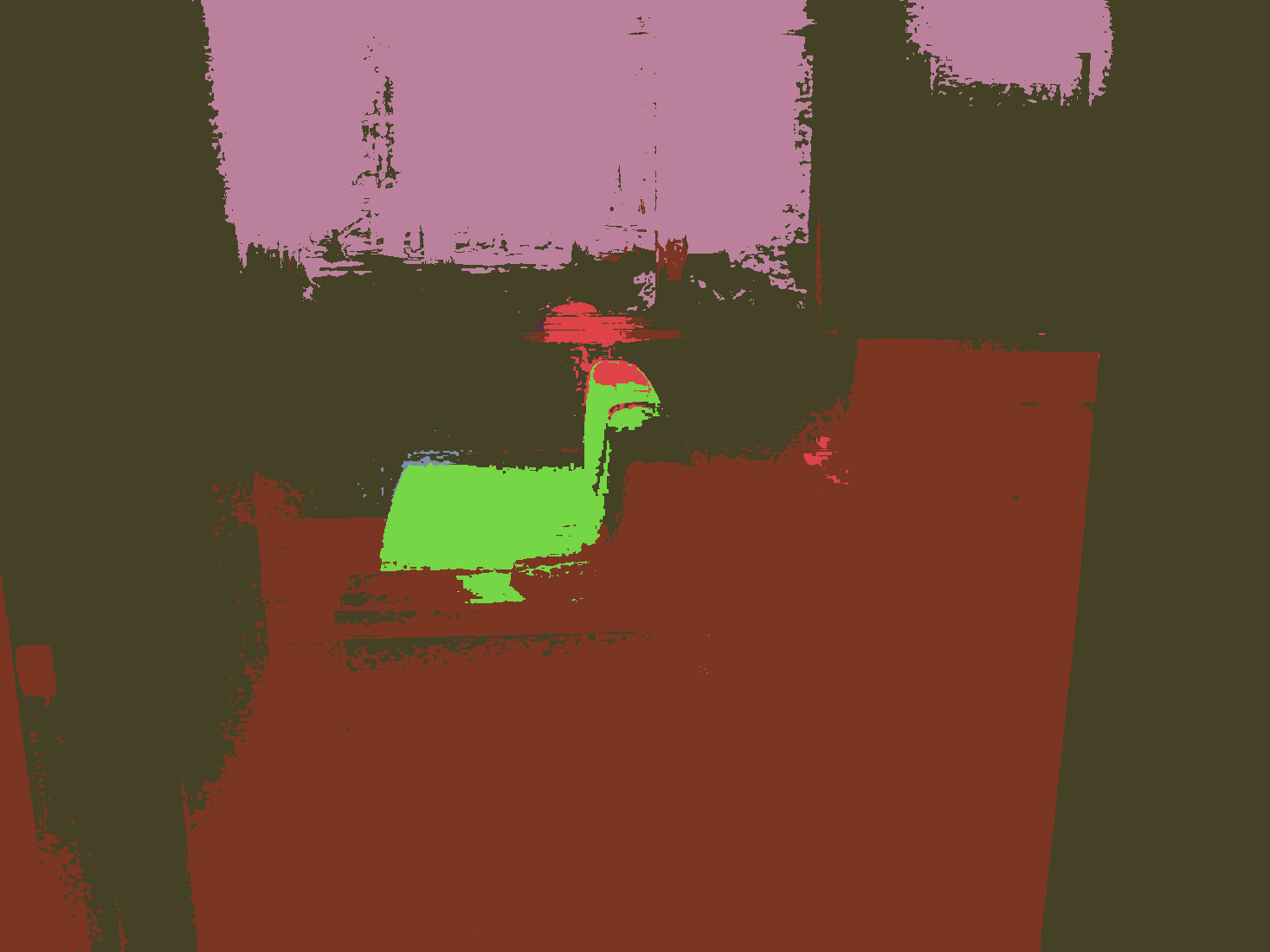}
  }
  \subfigure{%
    \includegraphics[width=.15\columnwidth]{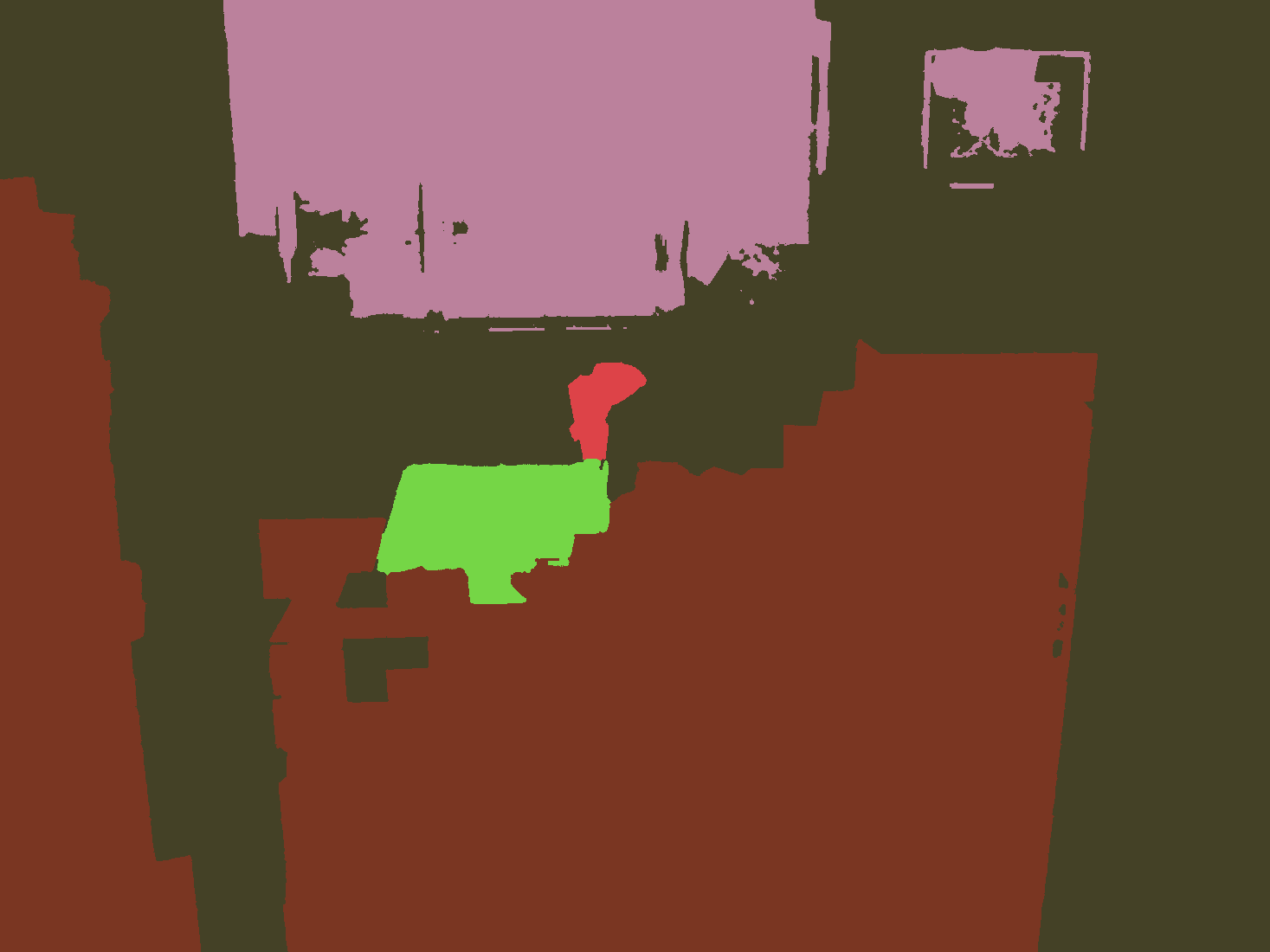}
  }\\[-2ex]

  \subfigure{%
    \includegraphics[width=.15\columnwidth]{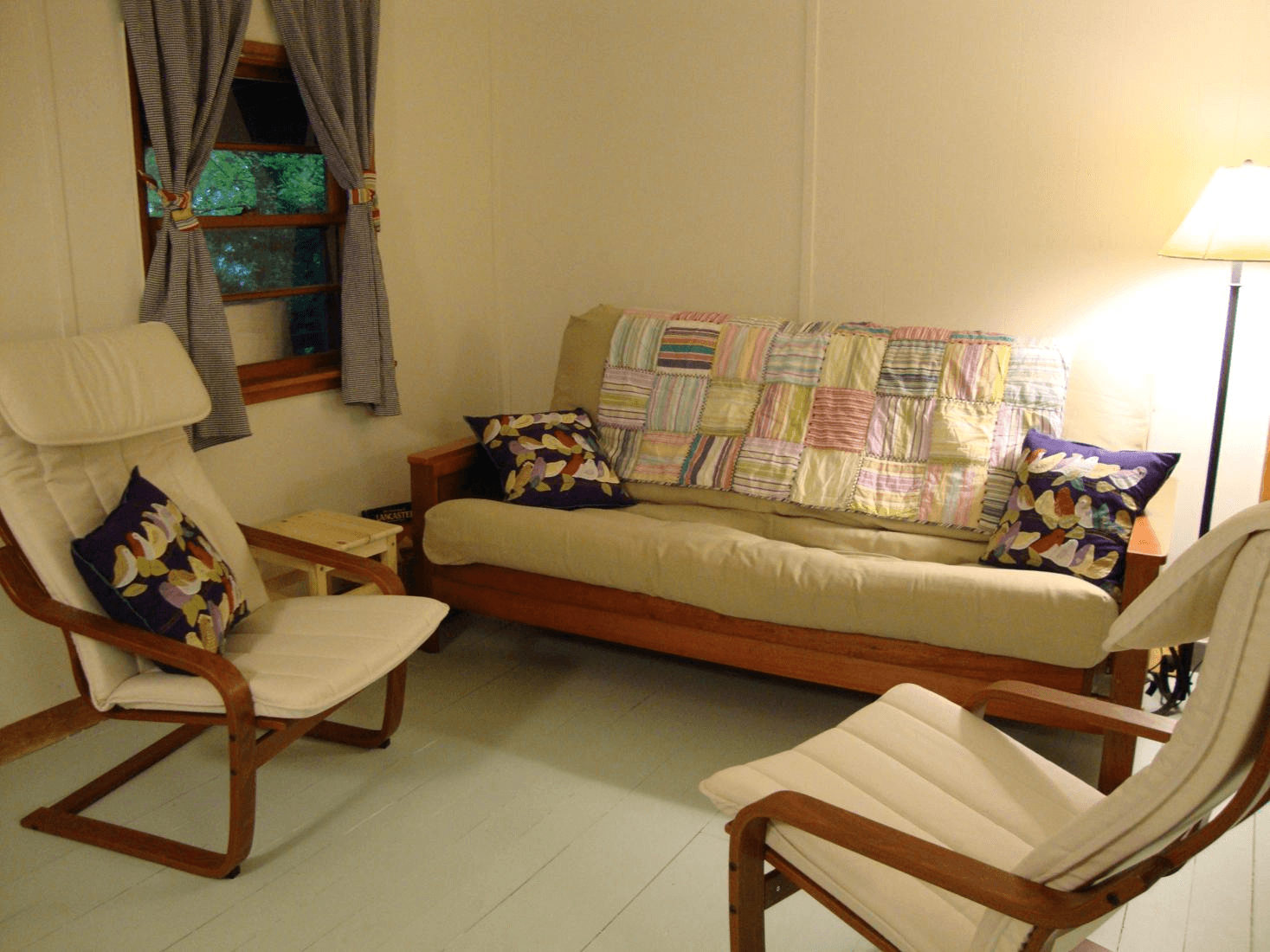}
  }
  \subfigure{%
    \includegraphics[width=.15\columnwidth]{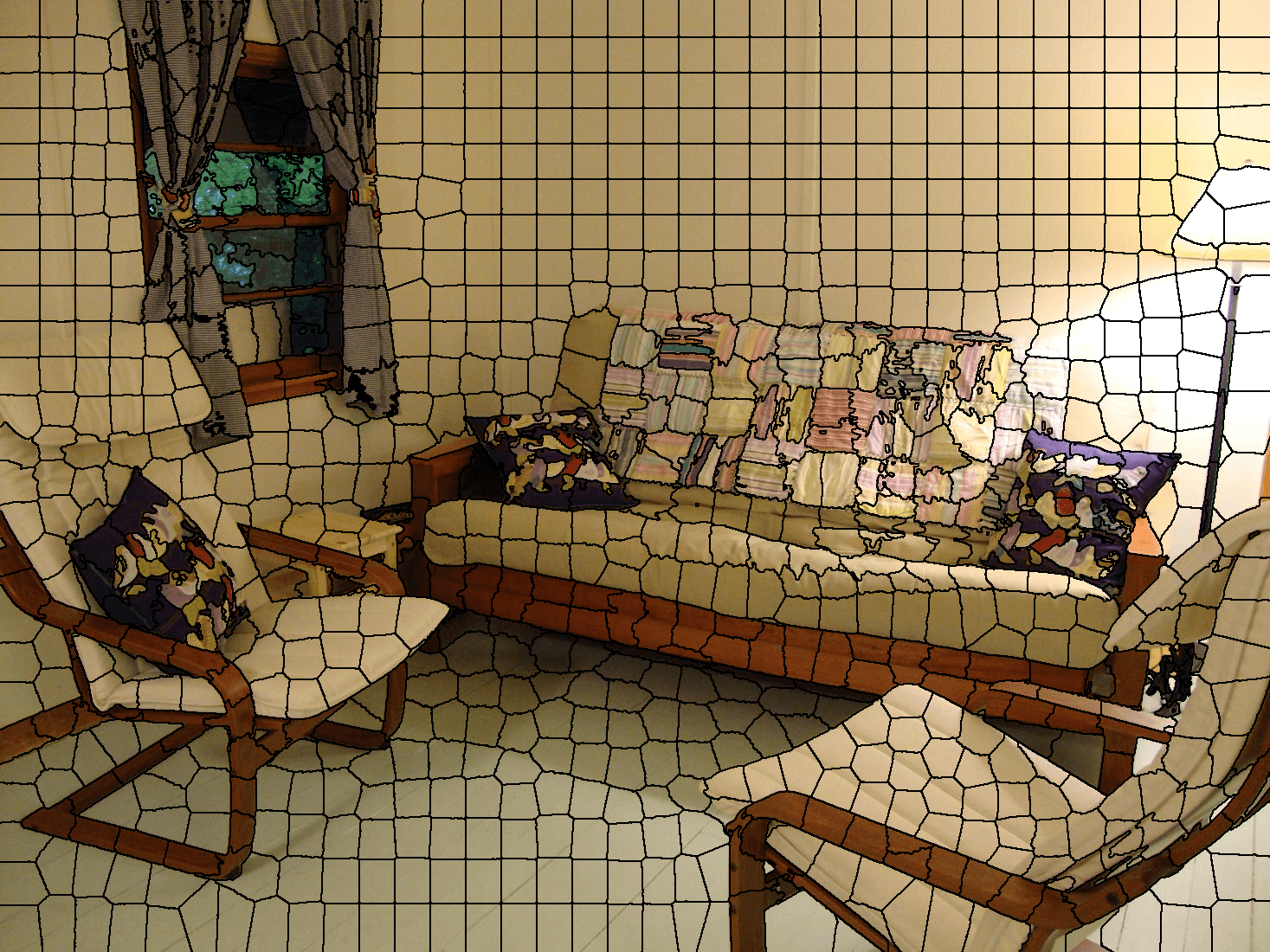}
  }
  \subfigure{%
    \includegraphics[width=.15\columnwidth]{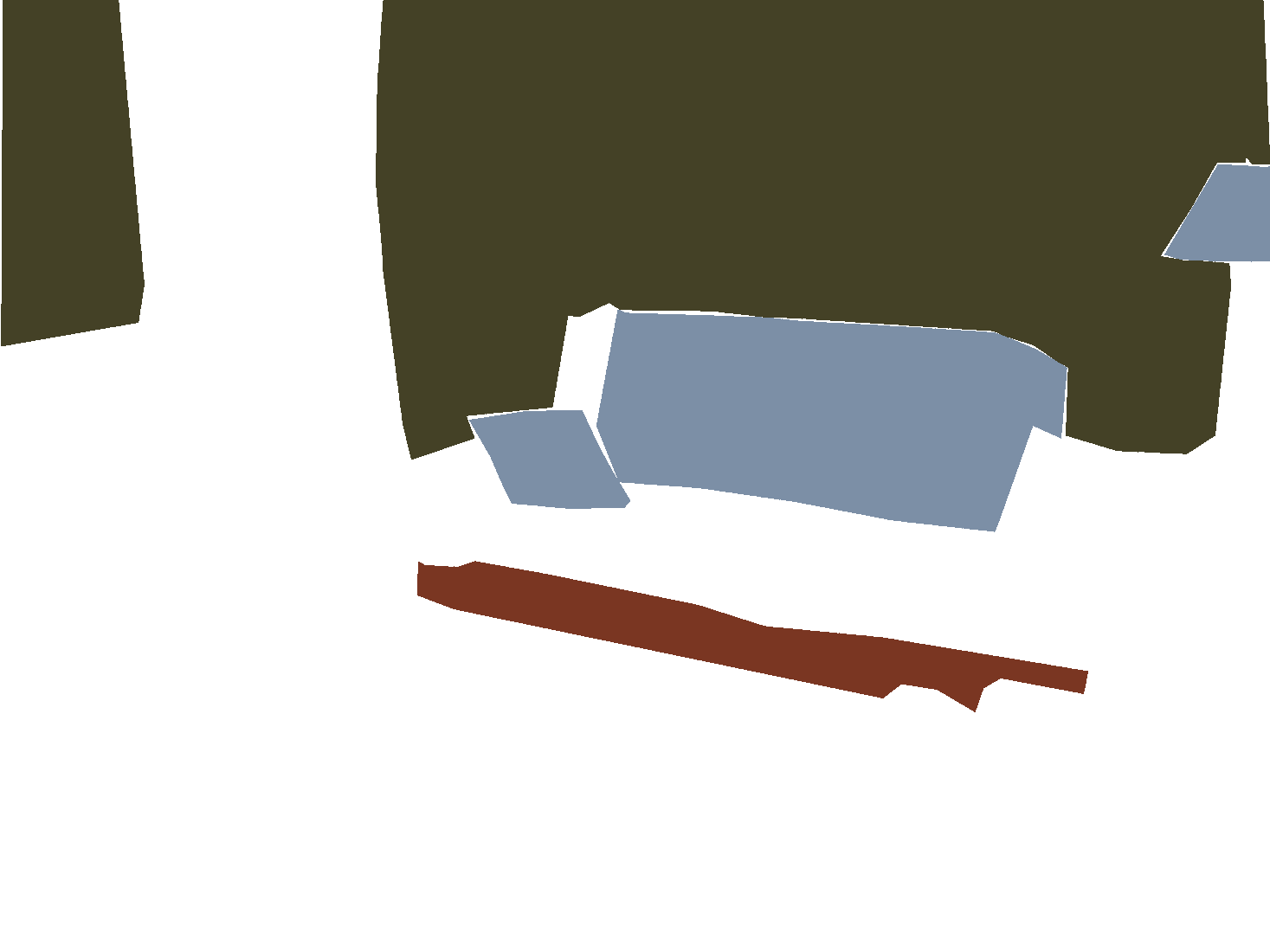}
  }
  \subfigure{%
    \includegraphics[width=.15\columnwidth]{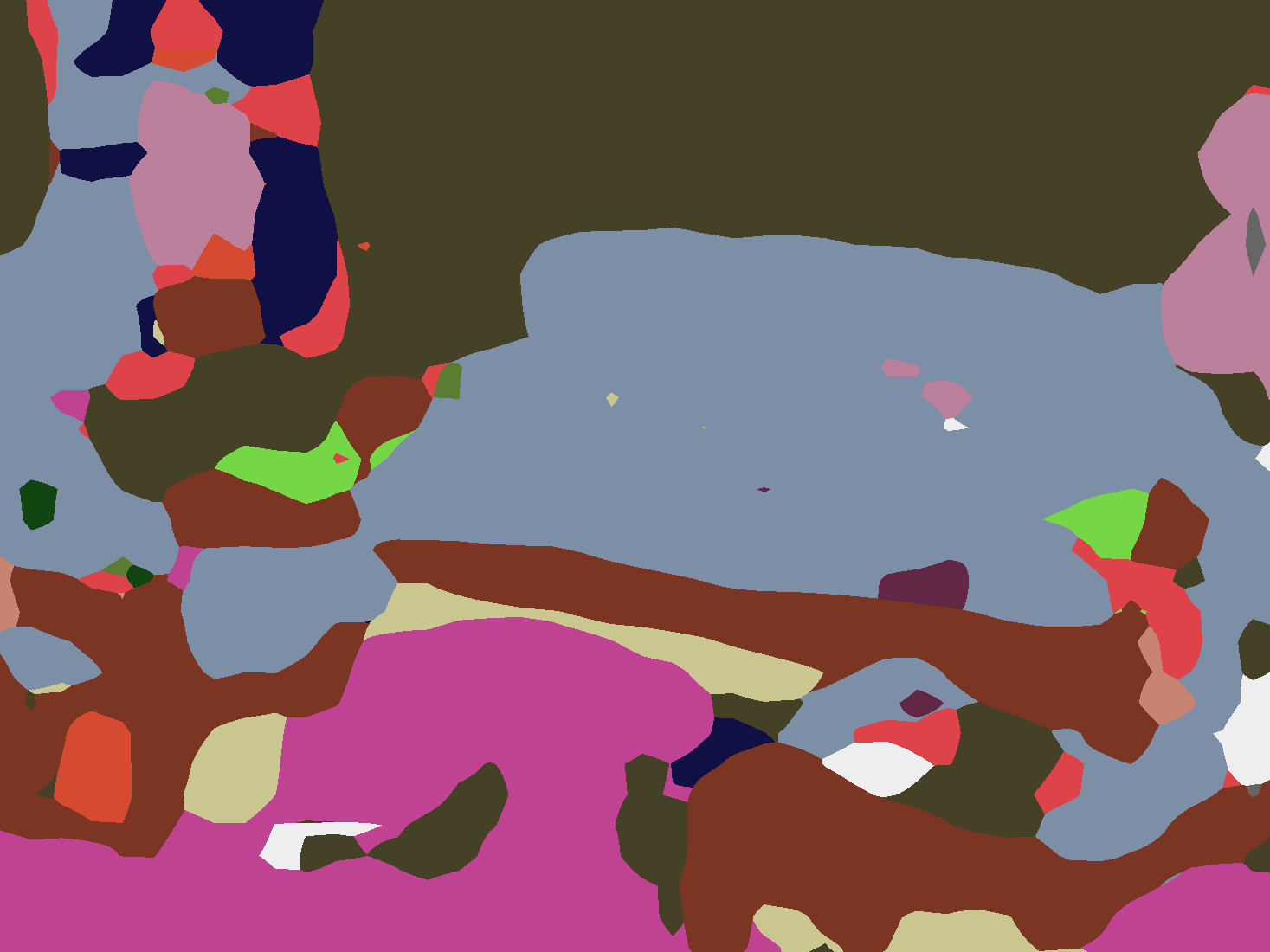}
  }
  \subfigure{%
    \includegraphics[width=.15\columnwidth]{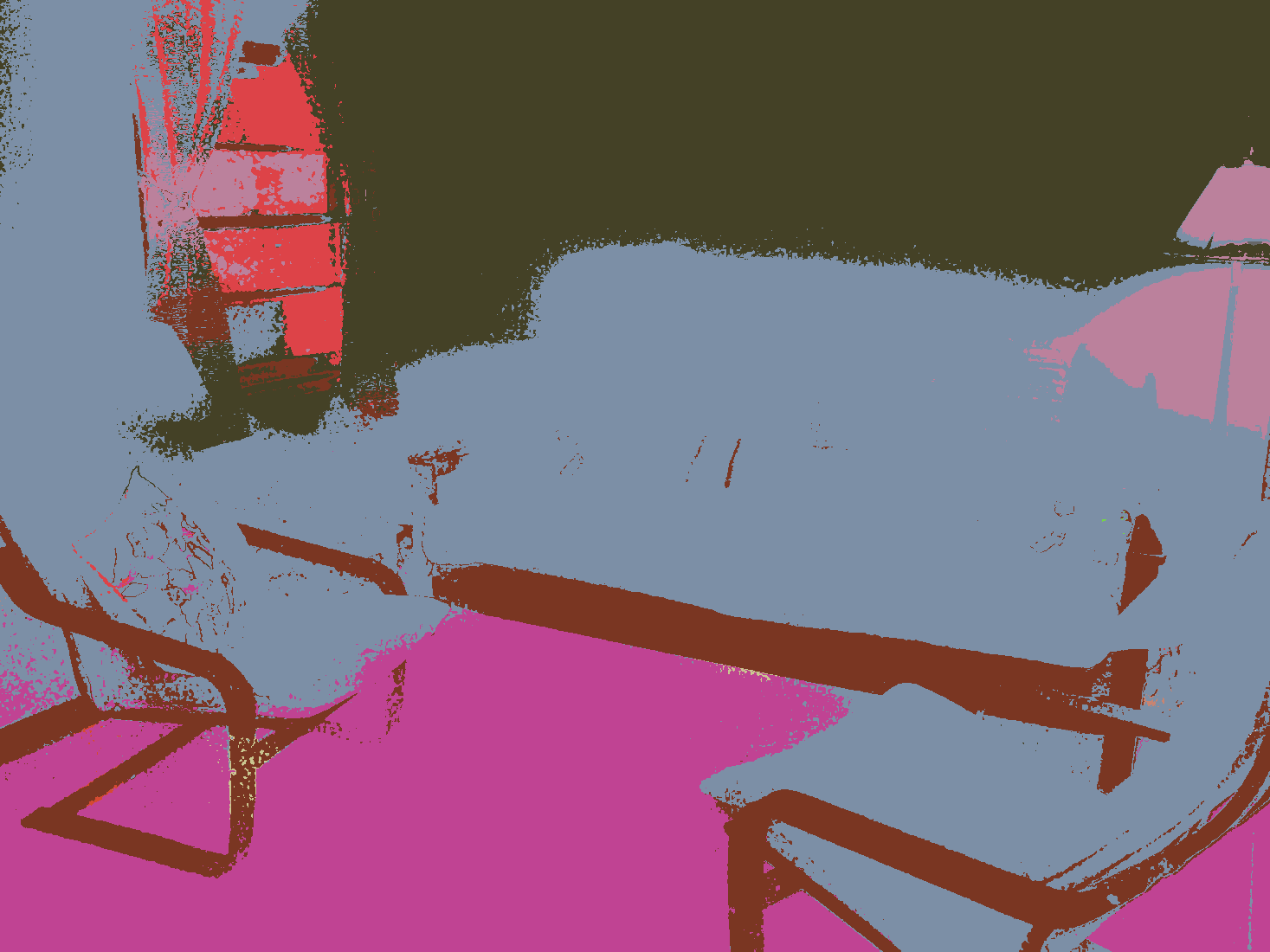}
  }
  \subfigure{%
    \includegraphics[width=.15\columnwidth]{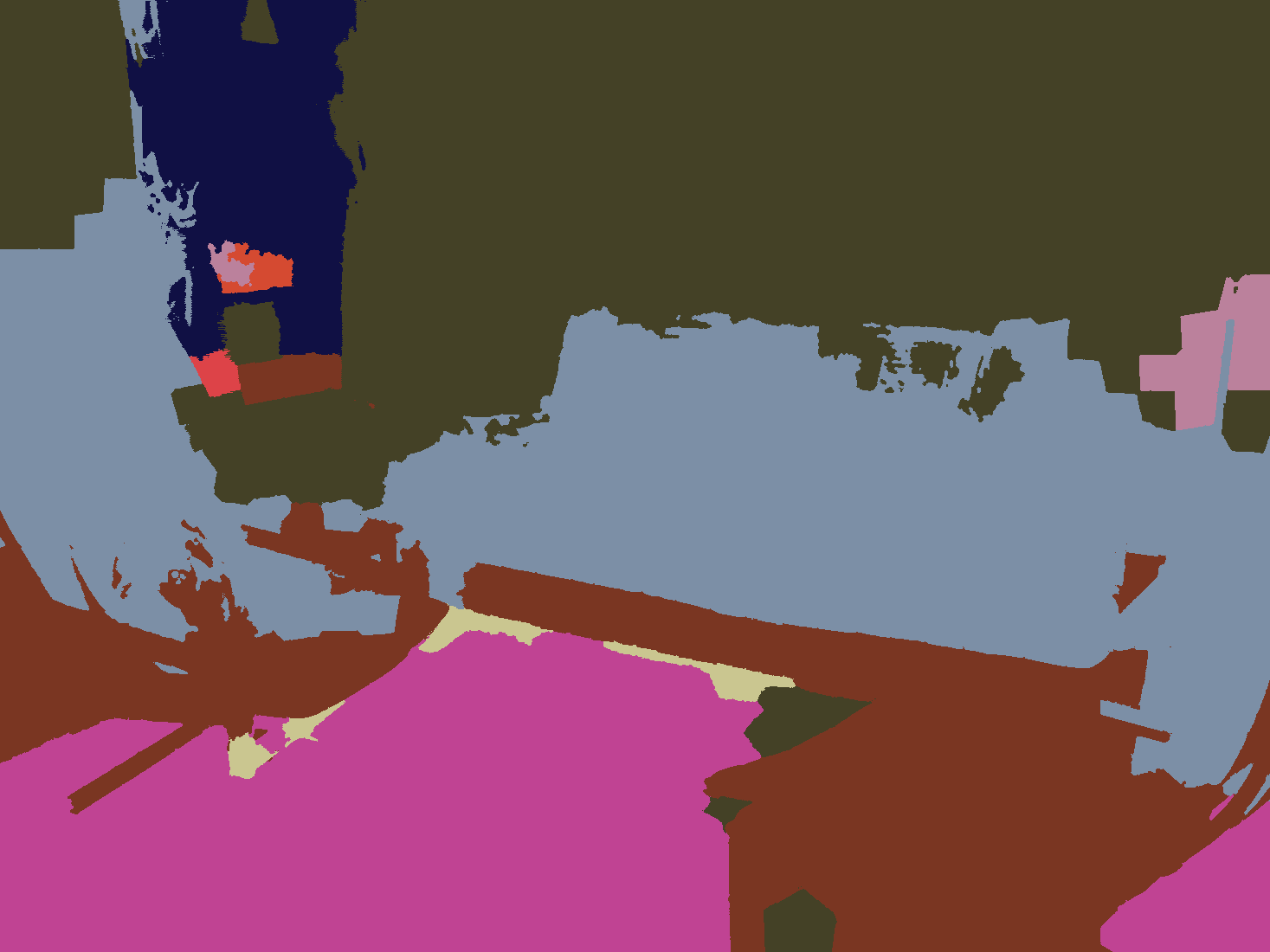}
  }\\[-2ex]

  \setcounter{subfigure}{0}
  \subfigure[\scriptsize Input]{%
    \includegraphics[width=.15\columnwidth]{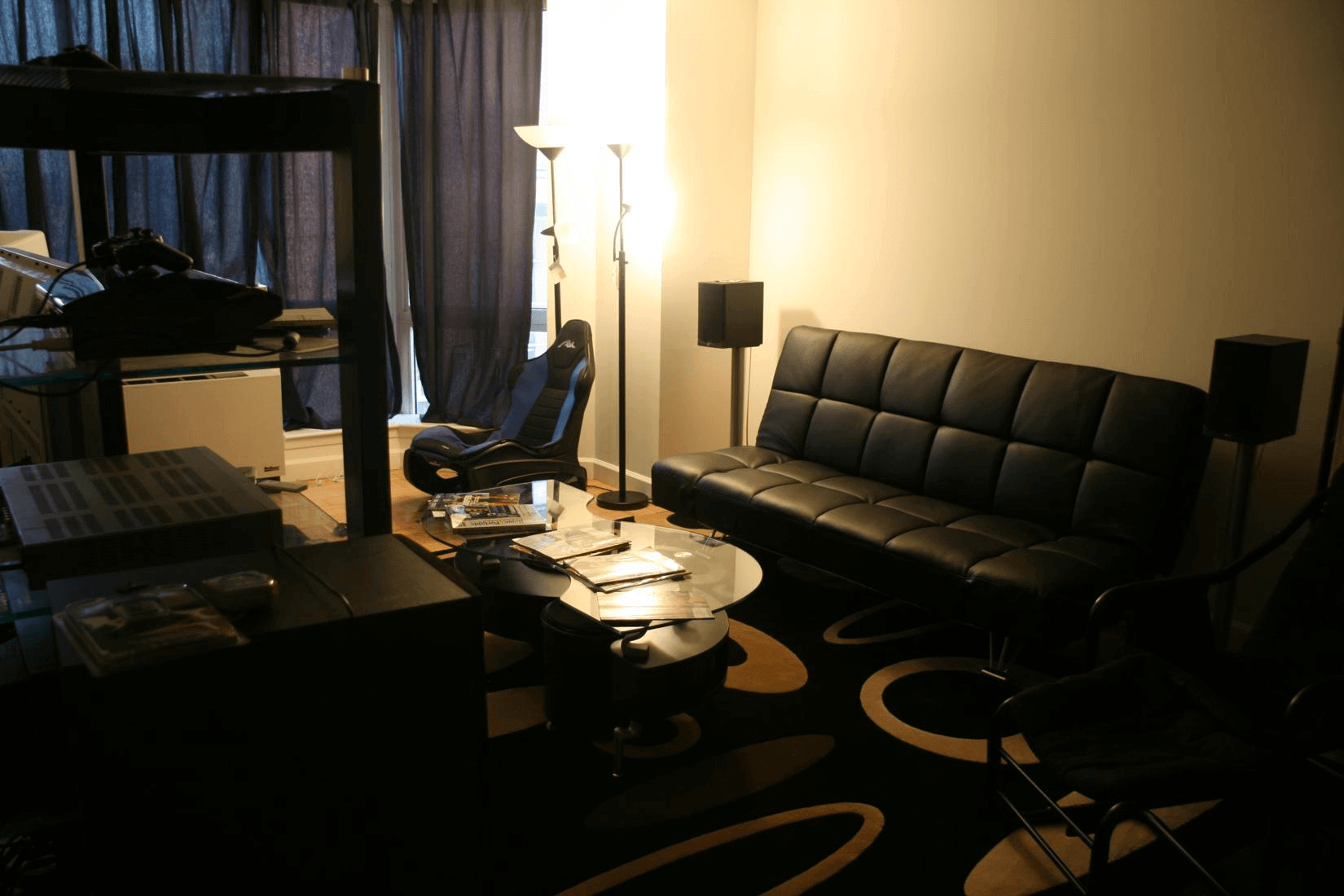}
  }
  \subfigure[\scriptsize Superpixels]{%
    \includegraphics[width=.15\columnwidth]{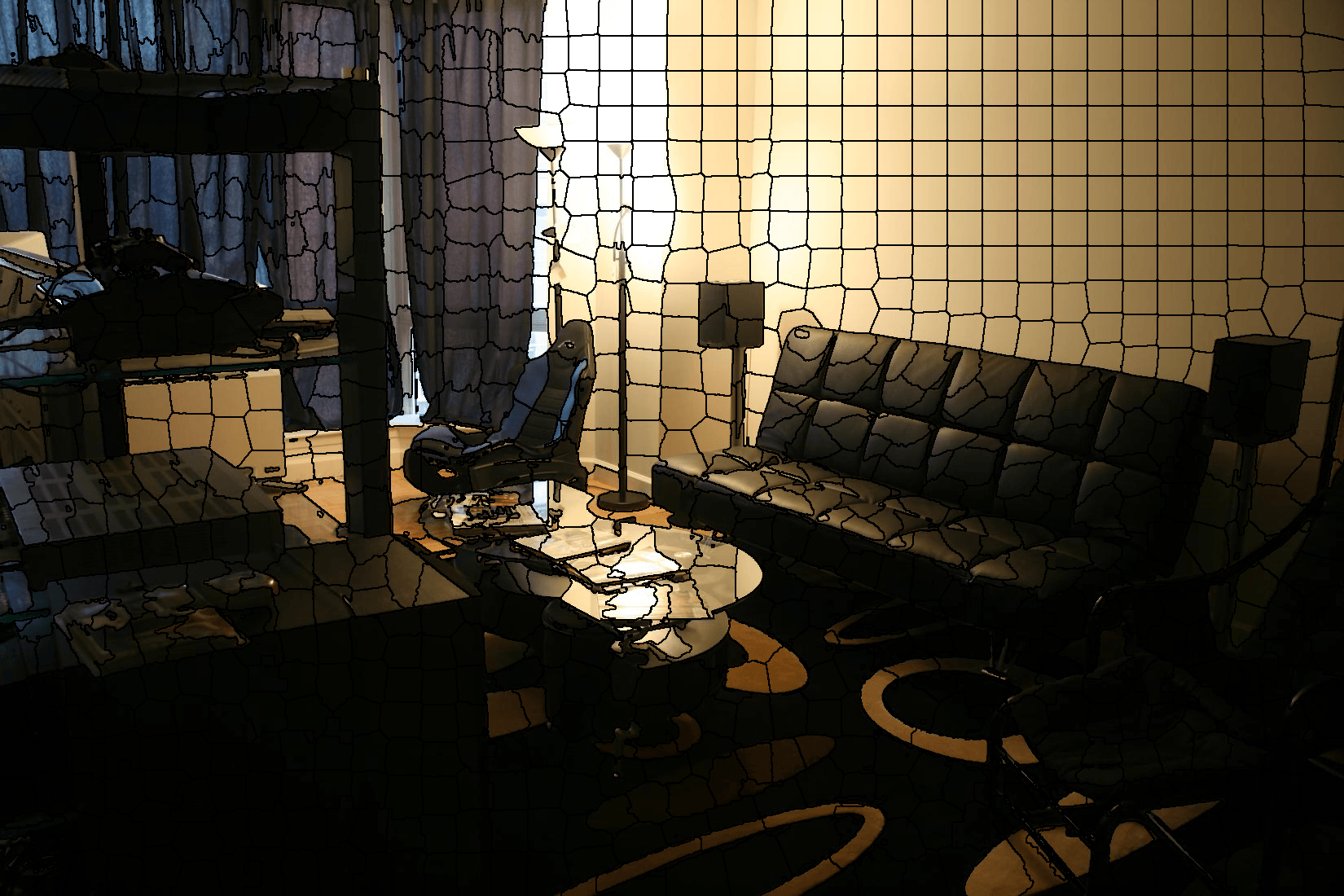}
  }
  \subfigure[\scriptsize GT]{%
    \includegraphics[width=.15\columnwidth]{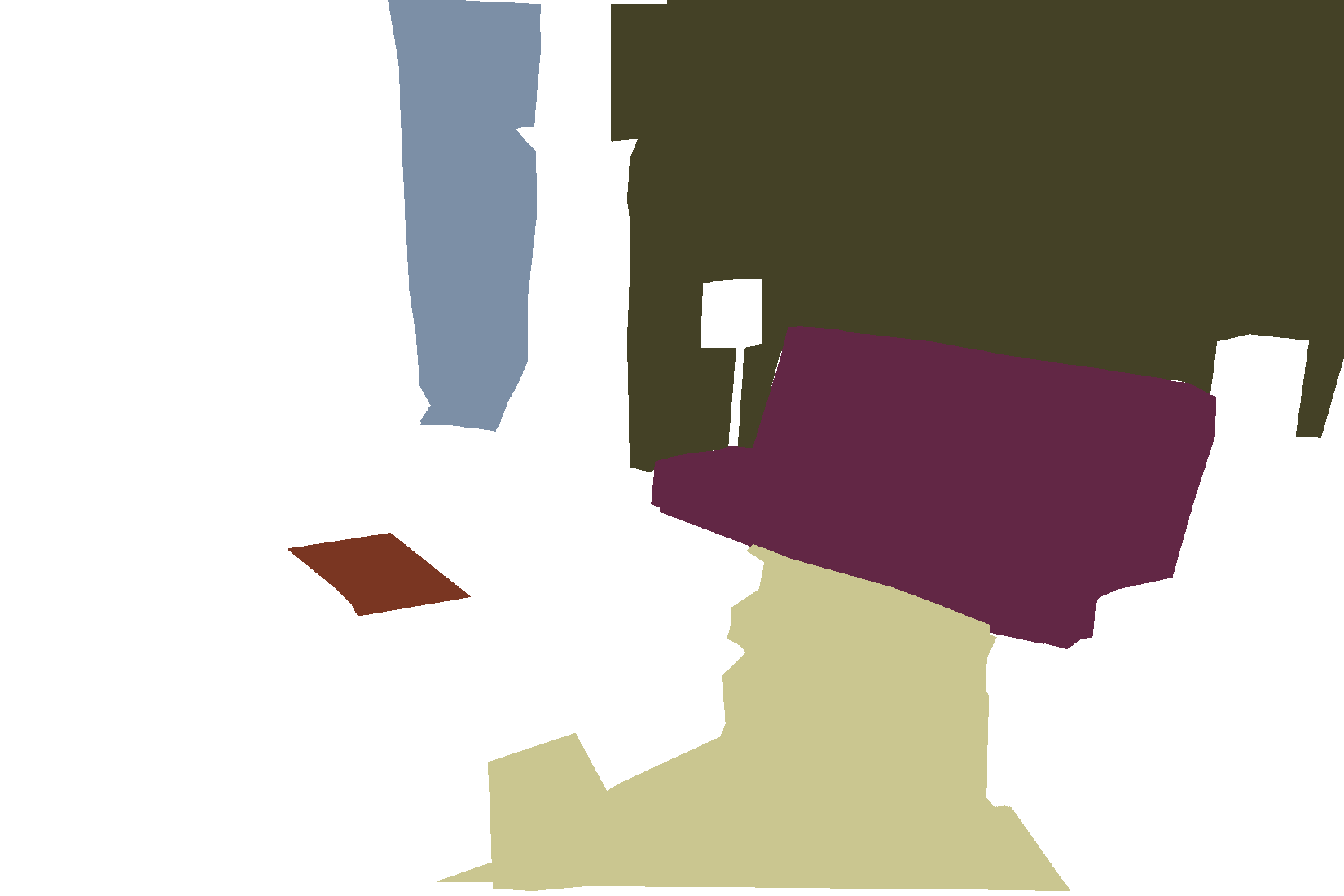}
  }
  \subfigure[\scriptsize AlexNet]{%
    \includegraphics[width=.15\columnwidth]{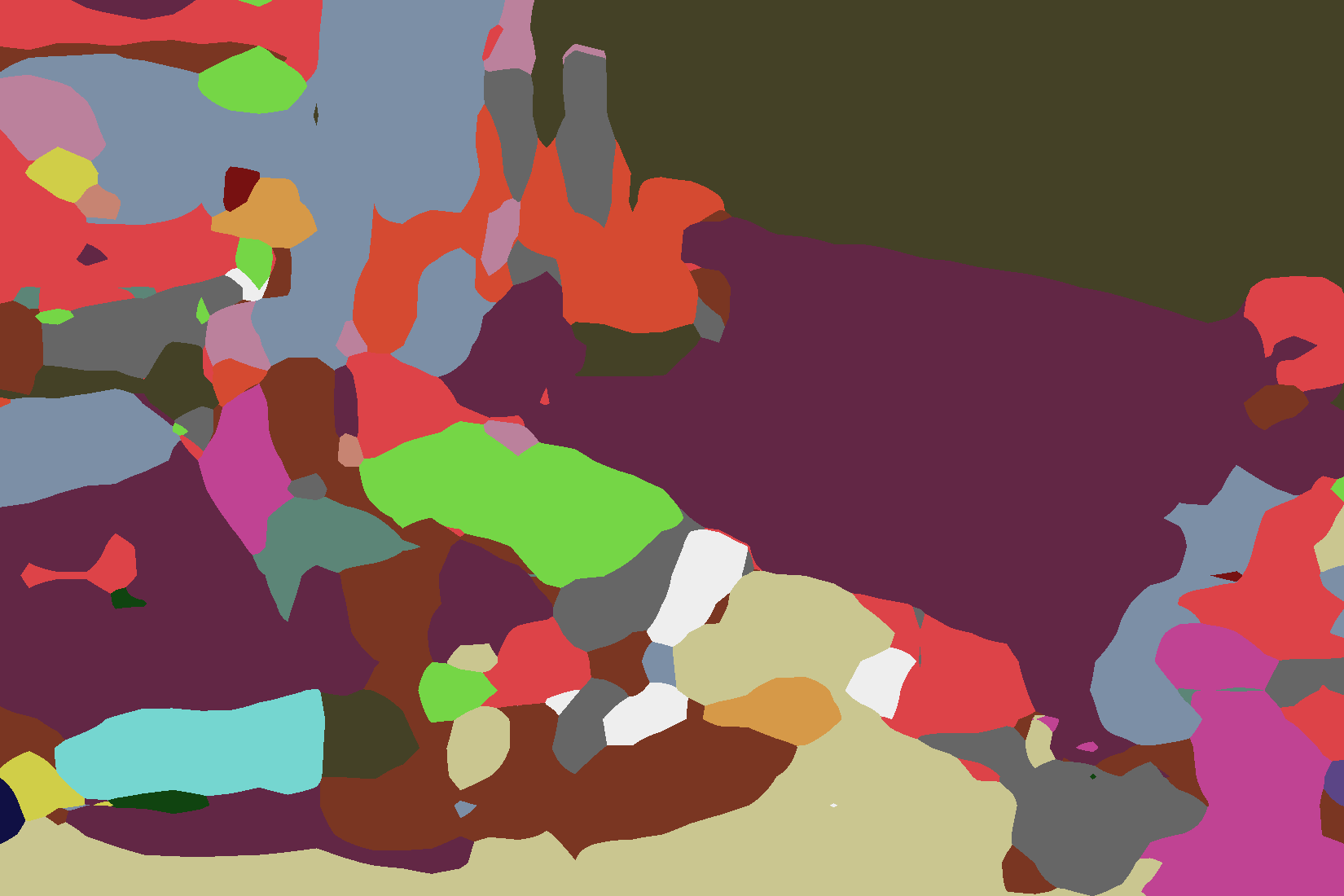}
  }
  \subfigure[\scriptsize +DenseCRF]{%
    \includegraphics[width=.15\columnwidth]{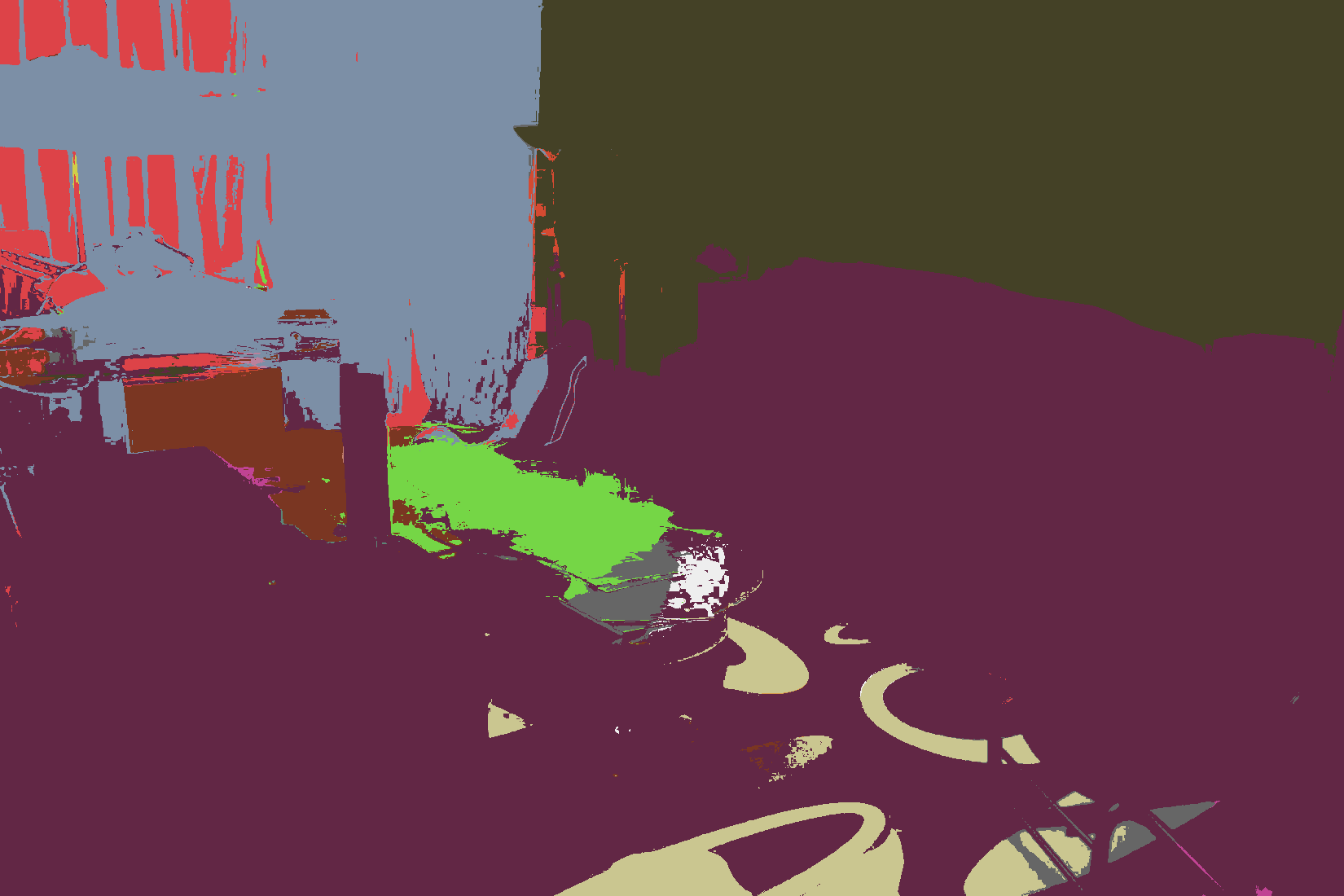}
  }
  \subfigure[\scriptsize Using BI]{%
    \includegraphics[width=.15\columnwidth]{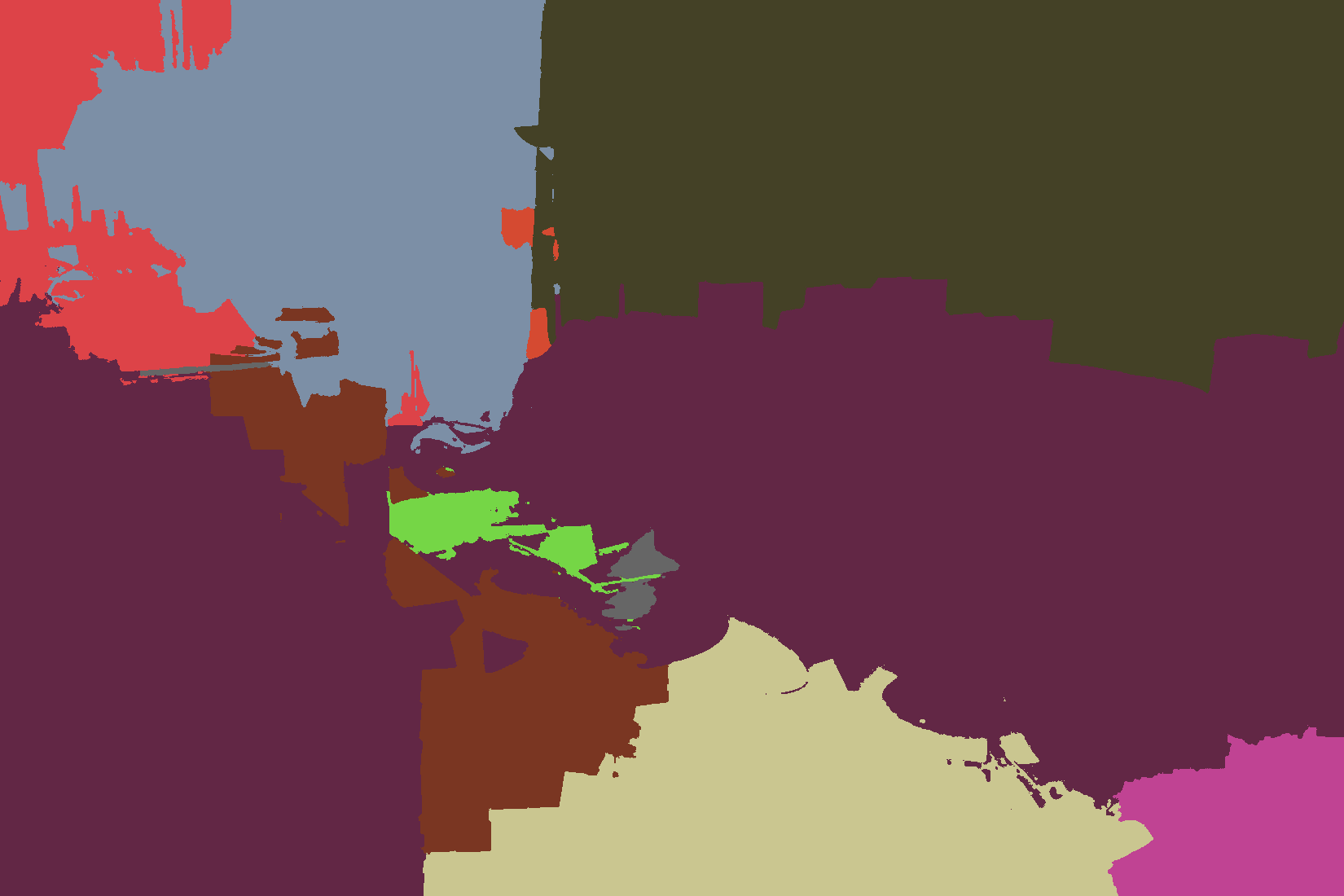}
  }
  \mycaption{Material Segmentation}{Example results of material segmentation.
  (d)~depicts the AlexNet CNN result, (e)~CNN + 10 steps of mean-field inference,
  (f)~result obtained with bilateral inception (BI) modules (\bi{7}{2}+\bi{8}{6}) between
  \fc~layers.}
\label{fig:material_visuals-app}
\end{figure*}

\definecolor{city_1}{RGB}{128, 64, 128}
\definecolor{city_2}{RGB}{244, 35, 232}
\definecolor{city_3}{RGB}{70, 70, 70}
\definecolor{city_4}{RGB}{102, 102, 156}
\definecolor{city_5}{RGB}{190, 153, 153}
\definecolor{city_6}{RGB}{153, 153, 153}
\definecolor{city_7}{RGB}{250, 170, 30}
\definecolor{city_8}{RGB}{220, 220, 0}
\definecolor{city_9}{RGB}{107, 142, 35}
\definecolor{city_10}{RGB}{152, 251, 152}
\definecolor{city_11}{RGB}{70, 130, 180}
\definecolor{city_12}{RGB}{220, 20, 60}
\definecolor{city_13}{RGB}{255, 0, 0}
\definecolor{city_14}{RGB}{0, 0, 142}
\definecolor{city_15}{RGB}{0, 0, 70}
\definecolor{city_16}{RGB}{0, 60, 100}
\definecolor{city_17}{RGB}{0, 80, 100}
\definecolor{city_18}{RGB}{0, 0, 230}
\definecolor{city_19}{RGB}{119, 11, 32}
\begin{figure*}[!ht]
  \small 
  \centering

  \subfigure{%
    \includegraphics[width=.18\columnwidth]{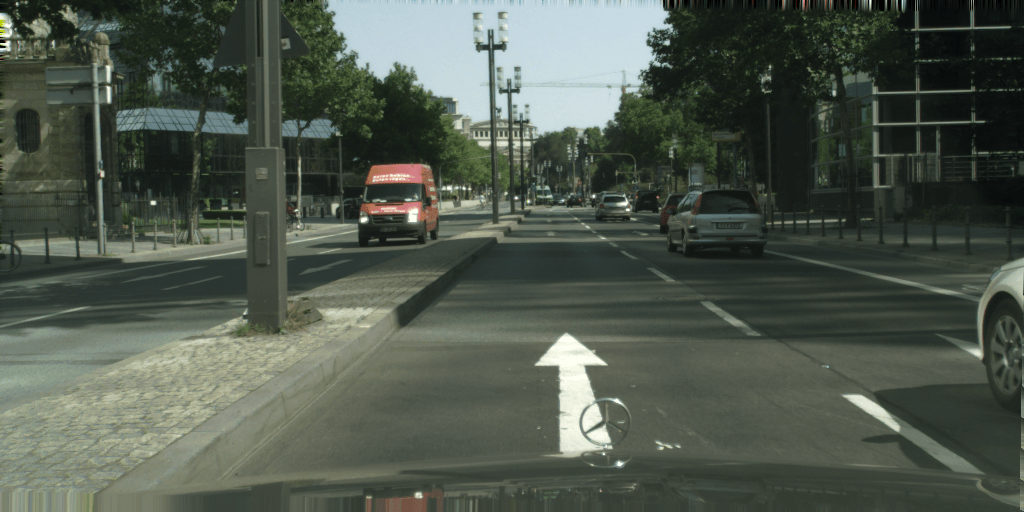}
  }
  \subfigure{%
    \includegraphics[width=.18\columnwidth]{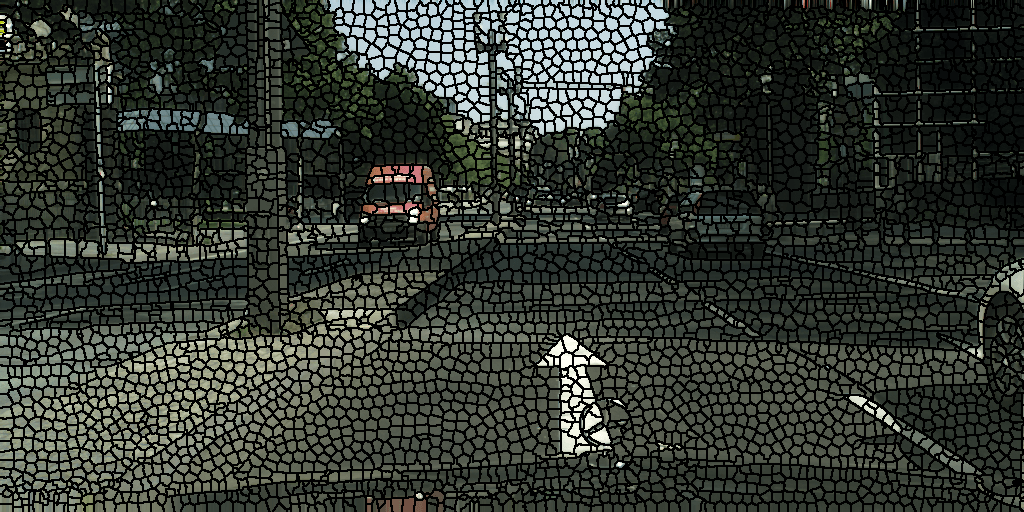}
  }
  \subfigure{%
    \includegraphics[width=.18\columnwidth]{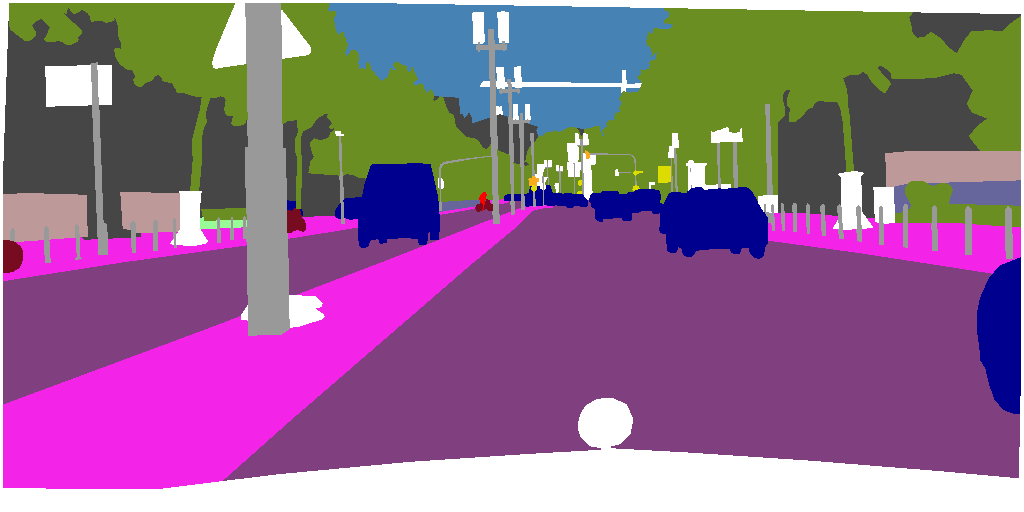}
  }
  \subfigure{%
    \includegraphics[width=.18\columnwidth]{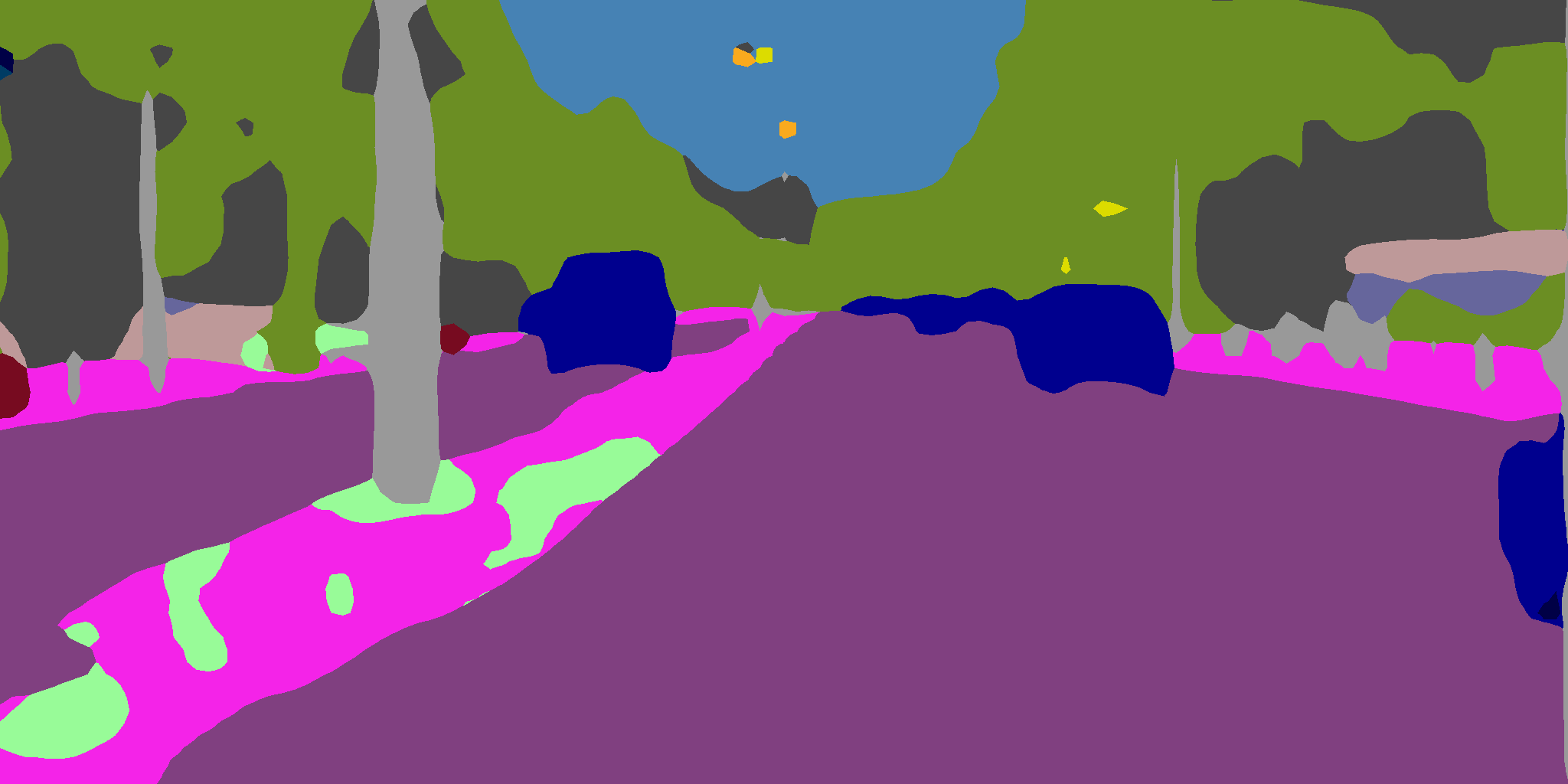}
  }
  \subfigure{%
    \includegraphics[width=.18\columnwidth]{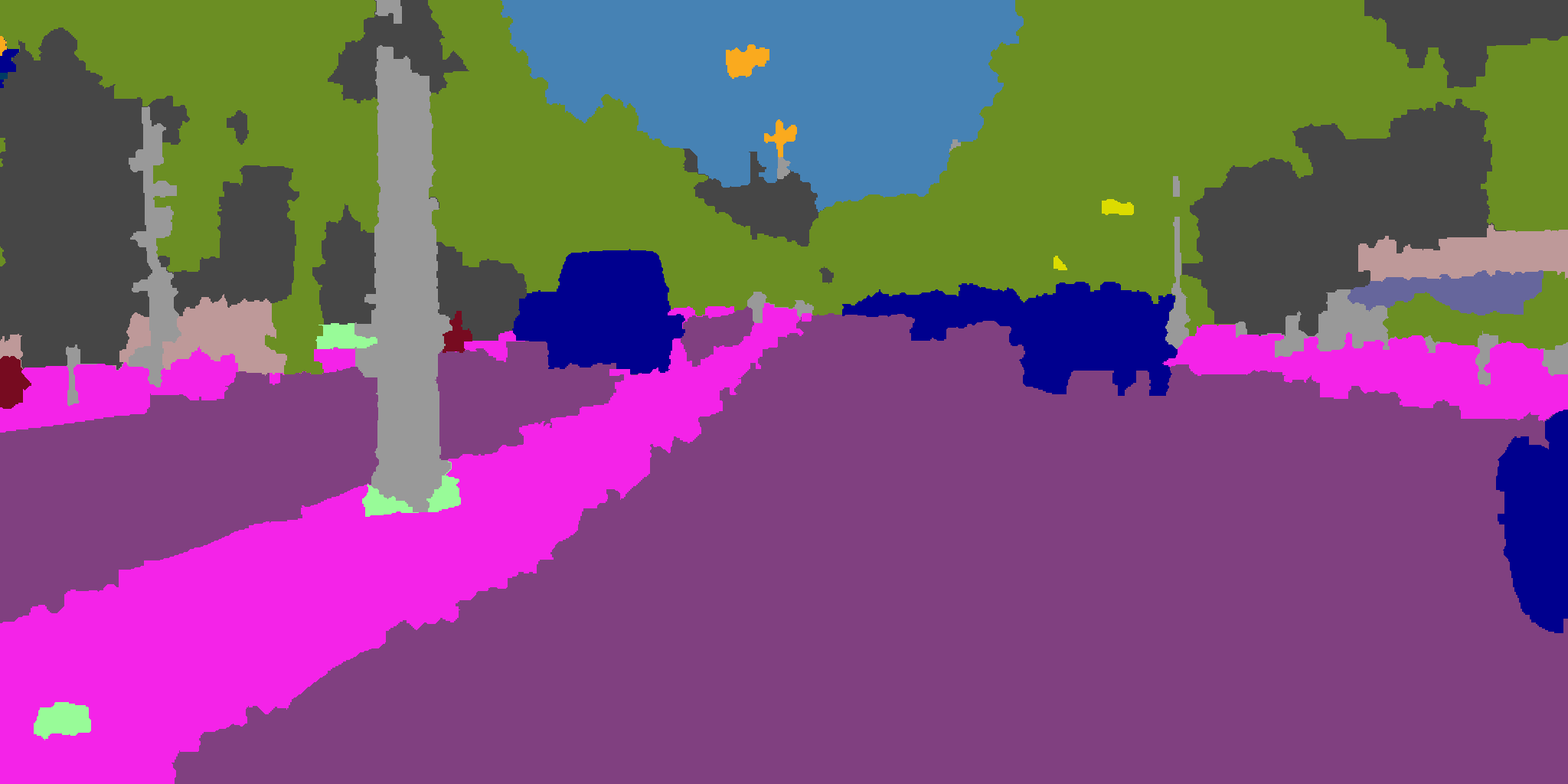}
  }\\[-2ex]

  \subfigure{%
    \includegraphics[width=.18\columnwidth]{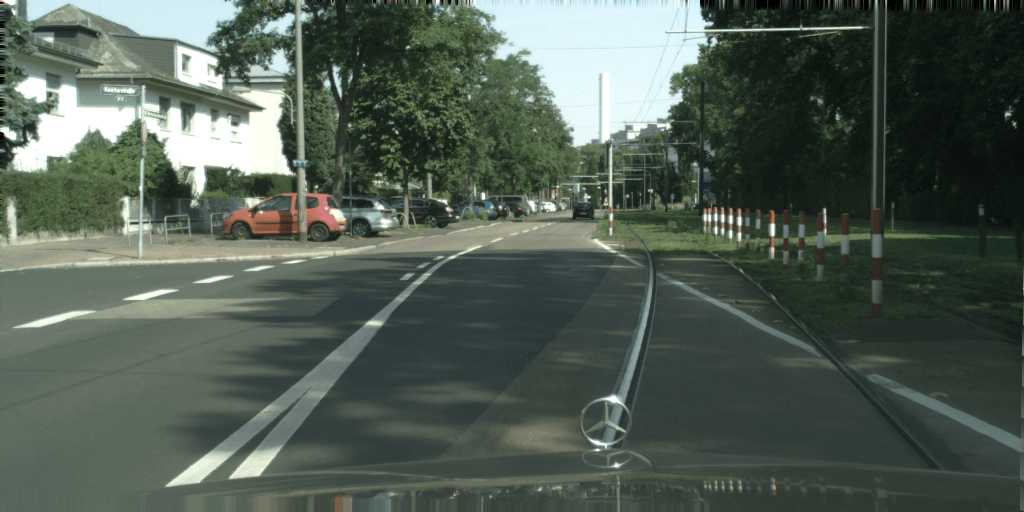}
  }
  \subfigure{%
    \includegraphics[width=.18\columnwidth]{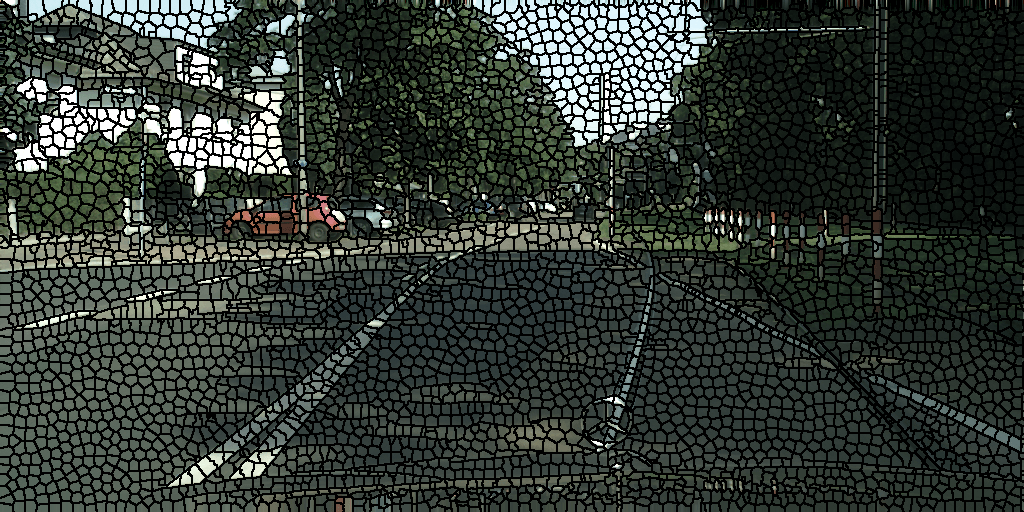}
  }
  \subfigure{%
    \includegraphics[width=.18\columnwidth]{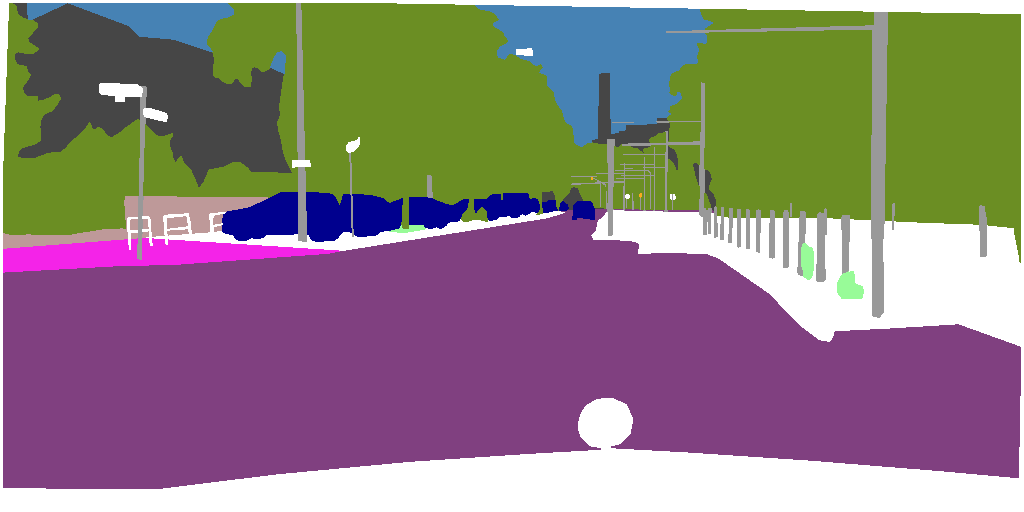}
  }
  \subfigure{%
    \includegraphics[width=.18\columnwidth]{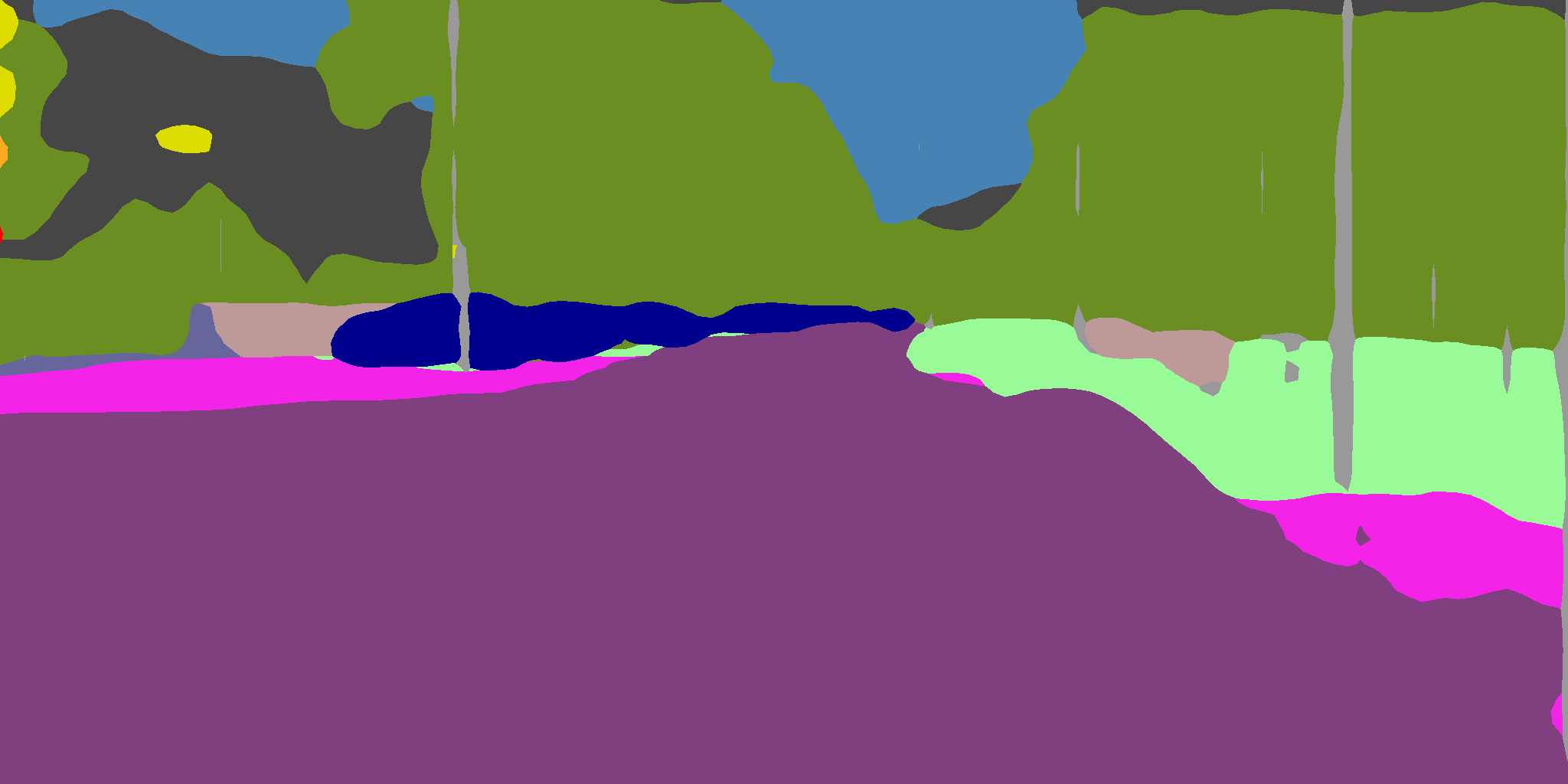}
  }
  \subfigure{%
    \includegraphics[width=.18\columnwidth]{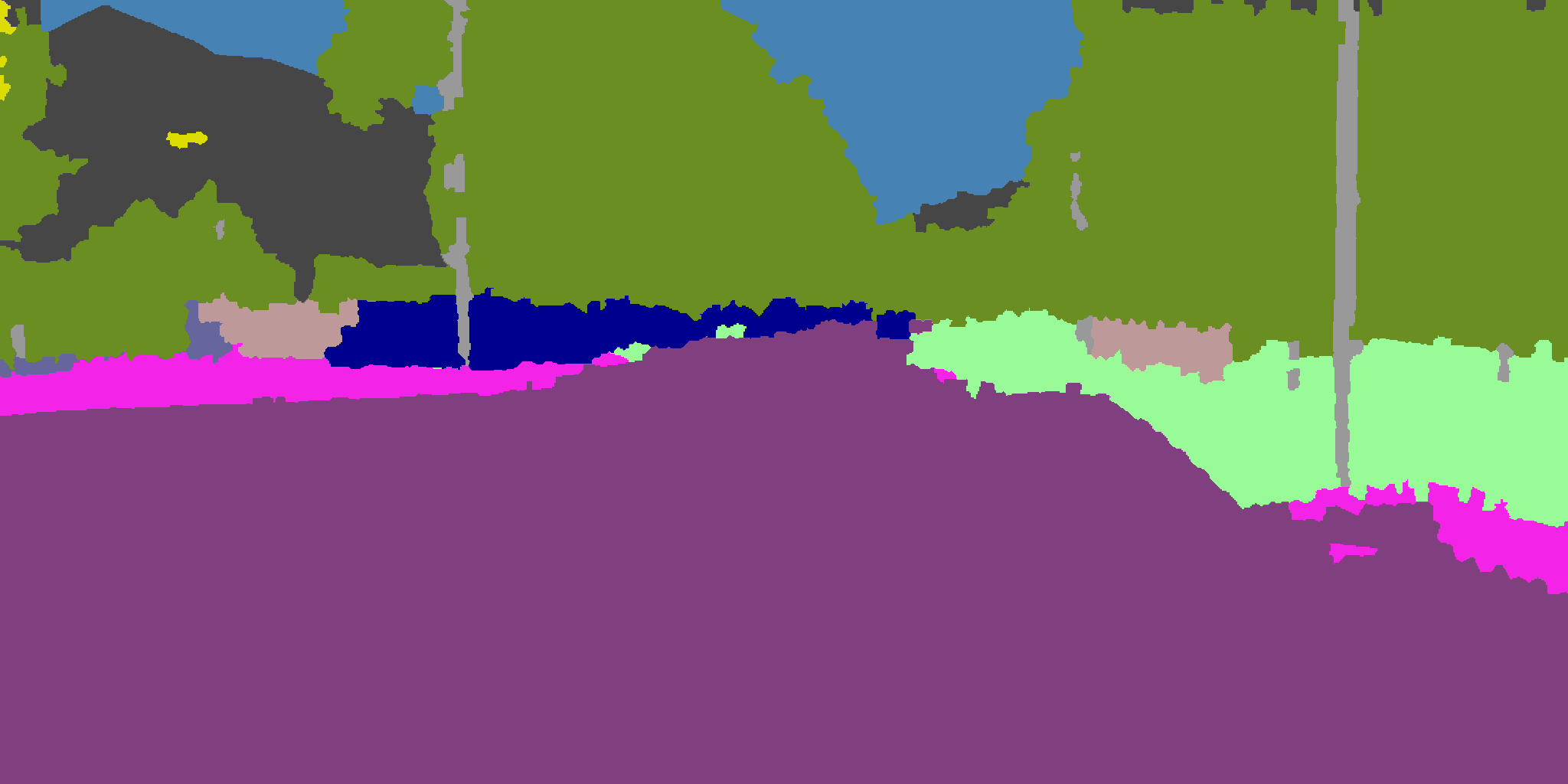}
  }\\[-2ex]

  \subfigure{%
    \includegraphics[width=.18\columnwidth]{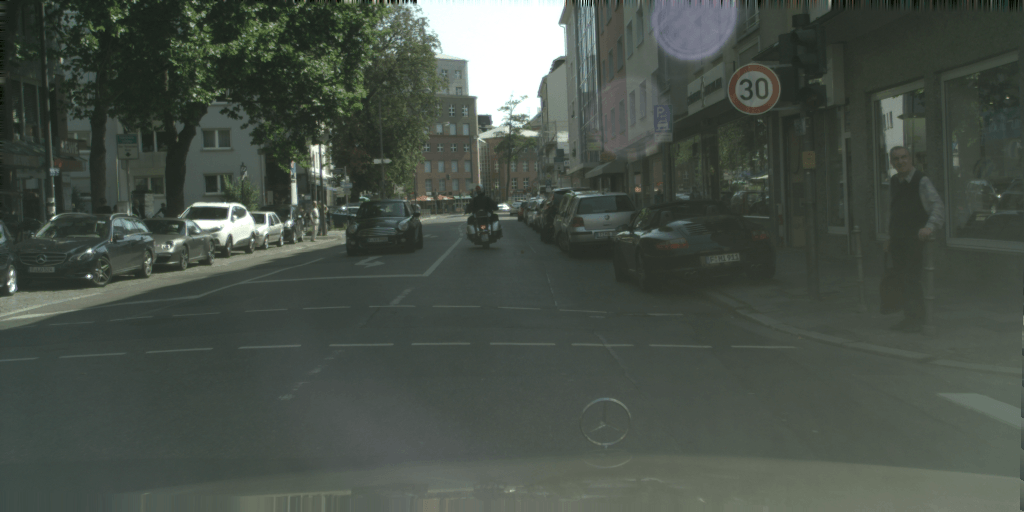}
  }
  \subfigure{%
    \includegraphics[width=.18\columnwidth]{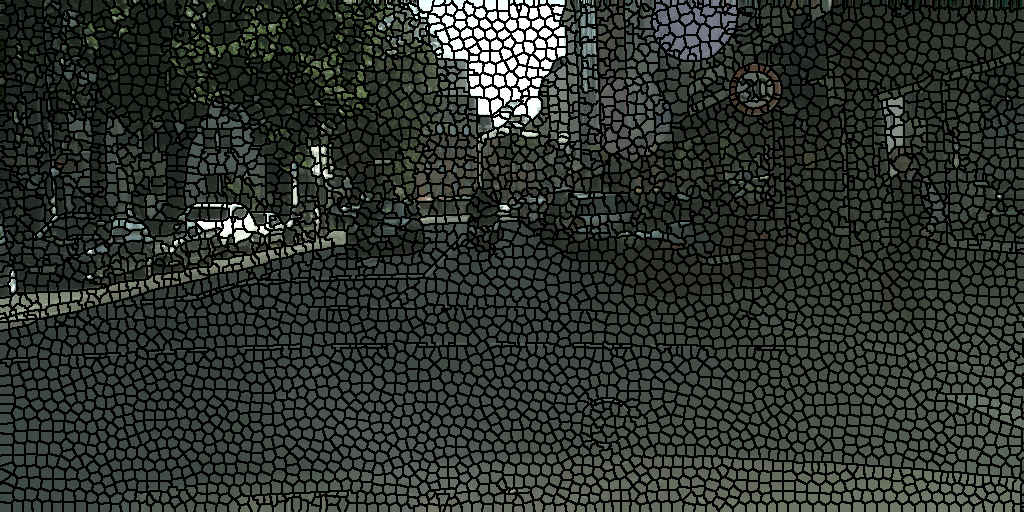}
  }
  \subfigure{%
    \includegraphics[width=.18\columnwidth]{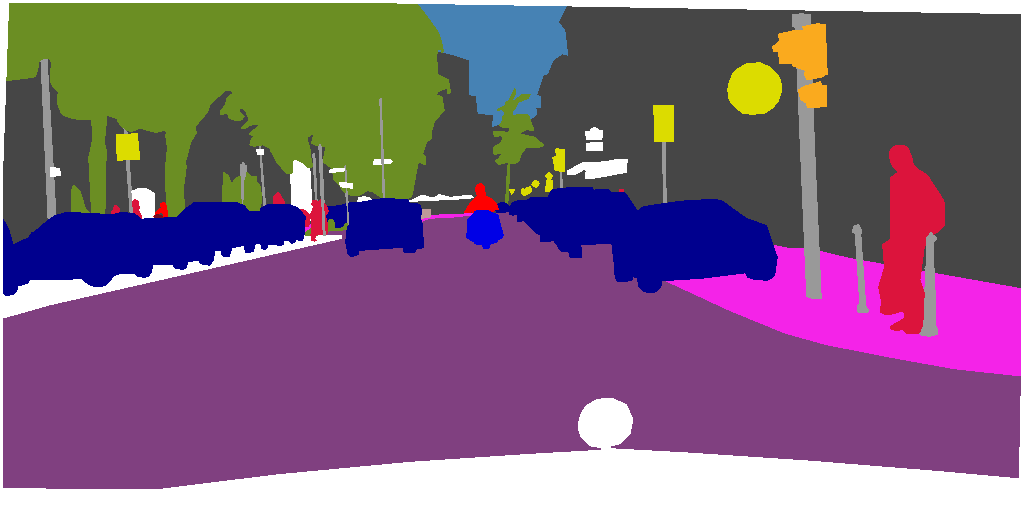}
  }
  \subfigure{%
    \includegraphics[width=.18\columnwidth]{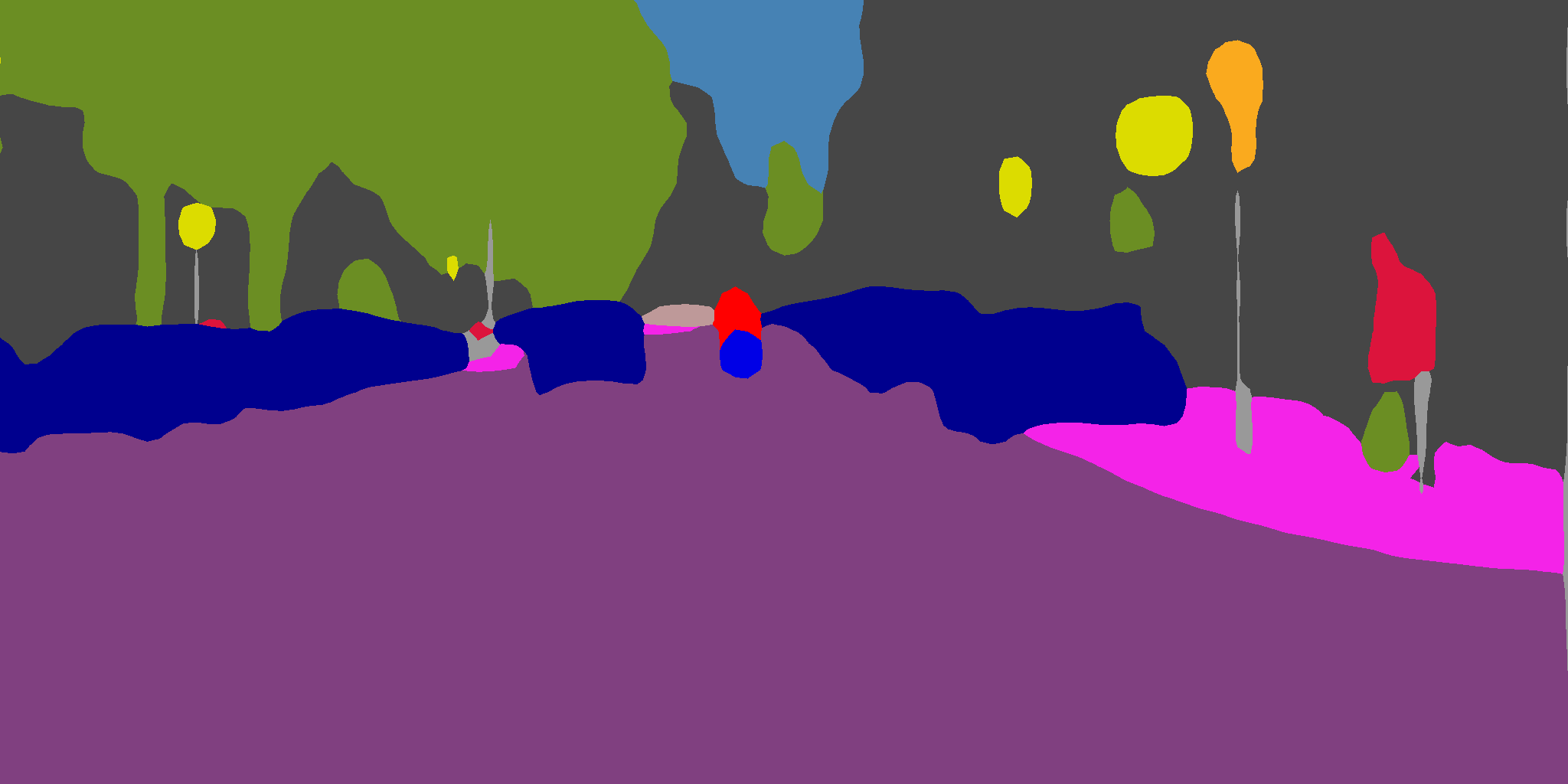}
  }
  \subfigure{%
    \includegraphics[width=.18\columnwidth]{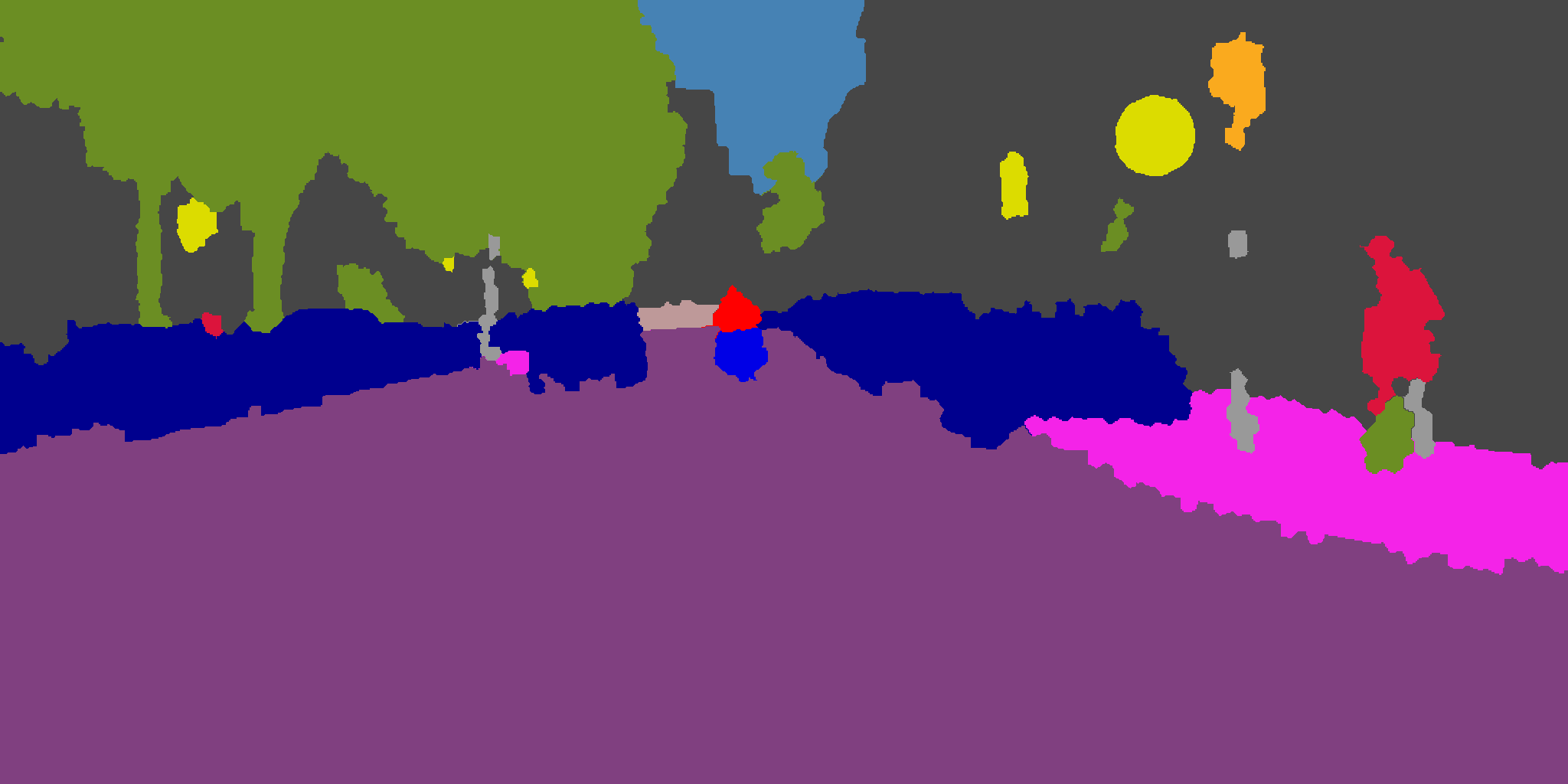}
  }\\[-2ex]

  \subfigure{%
    \includegraphics[width=.18\columnwidth]{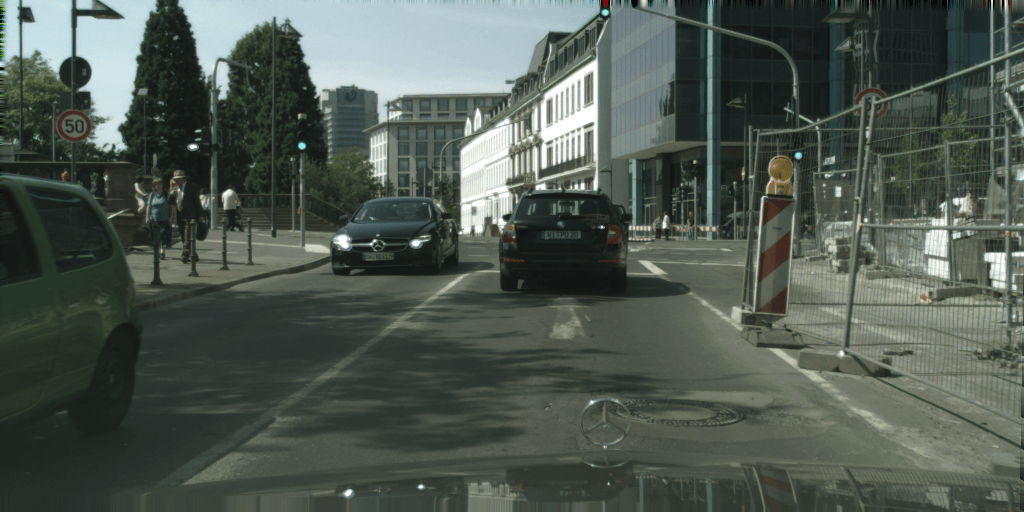}
  }
  \subfigure{%
    \includegraphics[width=.18\columnwidth]{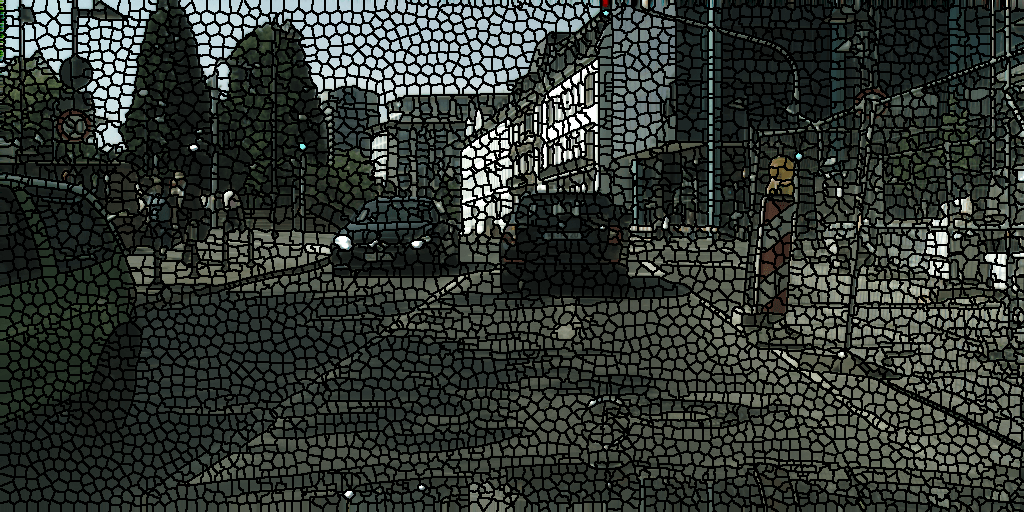}
  }
  \subfigure{%
    \includegraphics[width=.18\columnwidth]{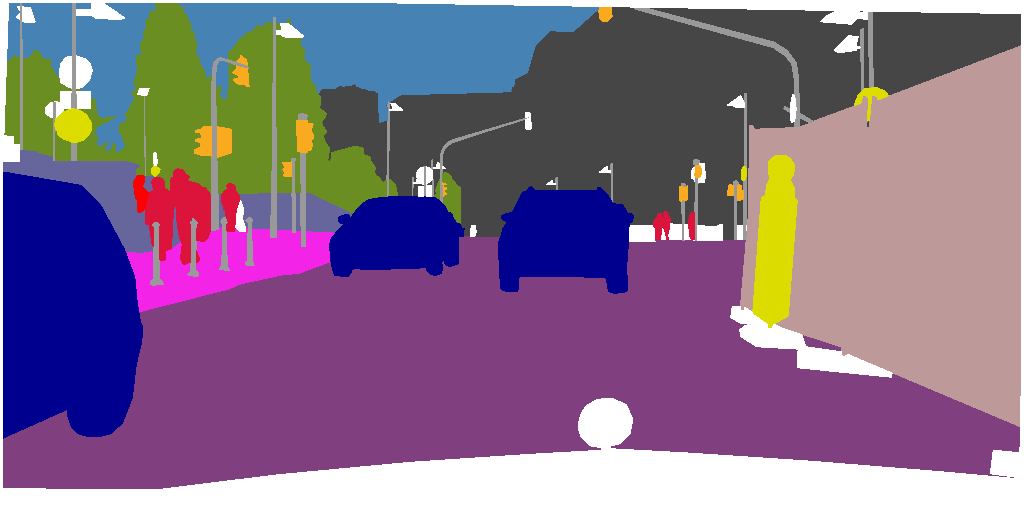}
  }
  \subfigure{%
    \includegraphics[width=.18\columnwidth]{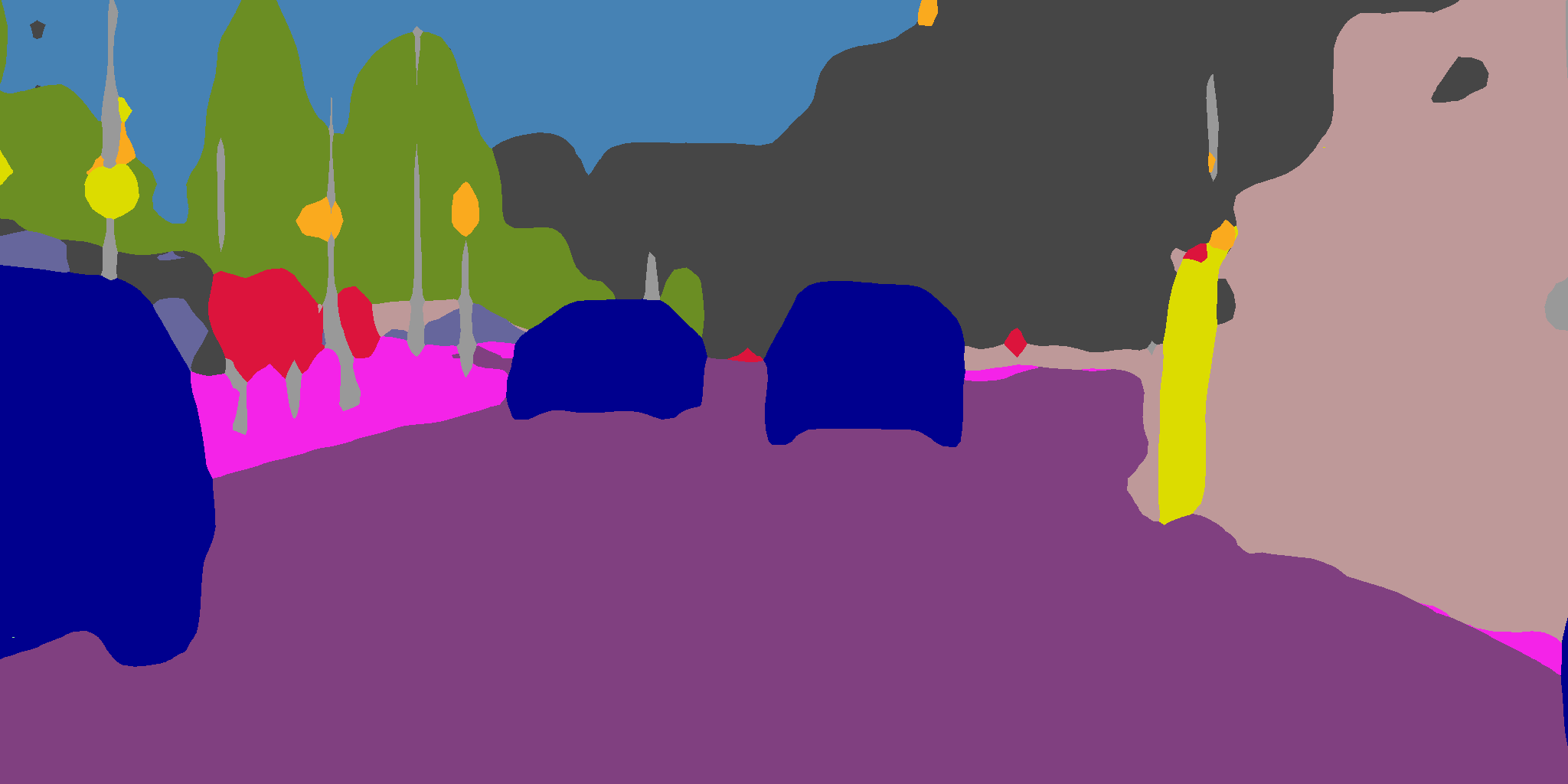}
  }
  \subfigure{%
    \includegraphics[width=.18\columnwidth]{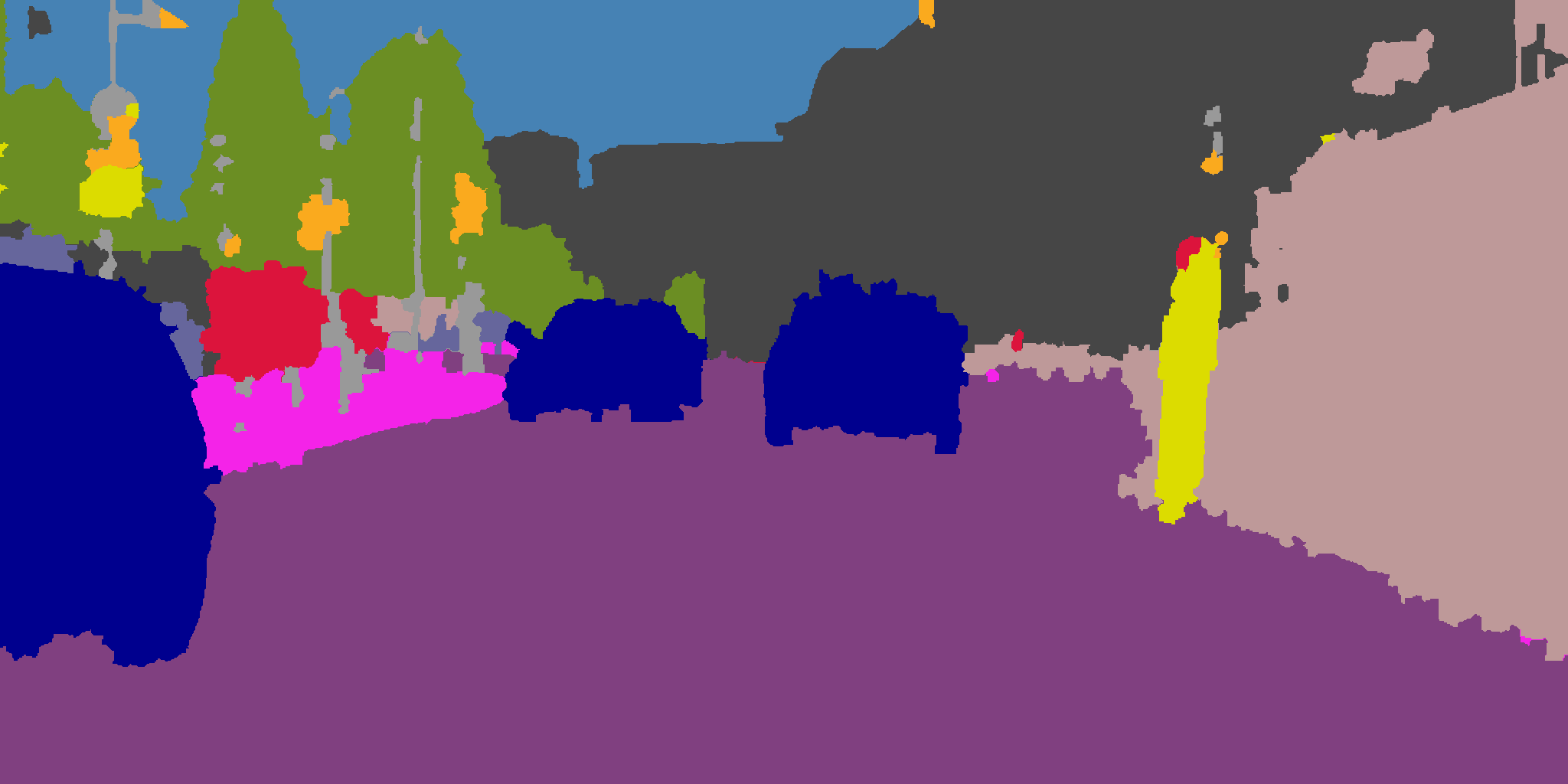}
  }\\[-2ex]

  \subfigure{%
    \includegraphics[width=.18\columnwidth]{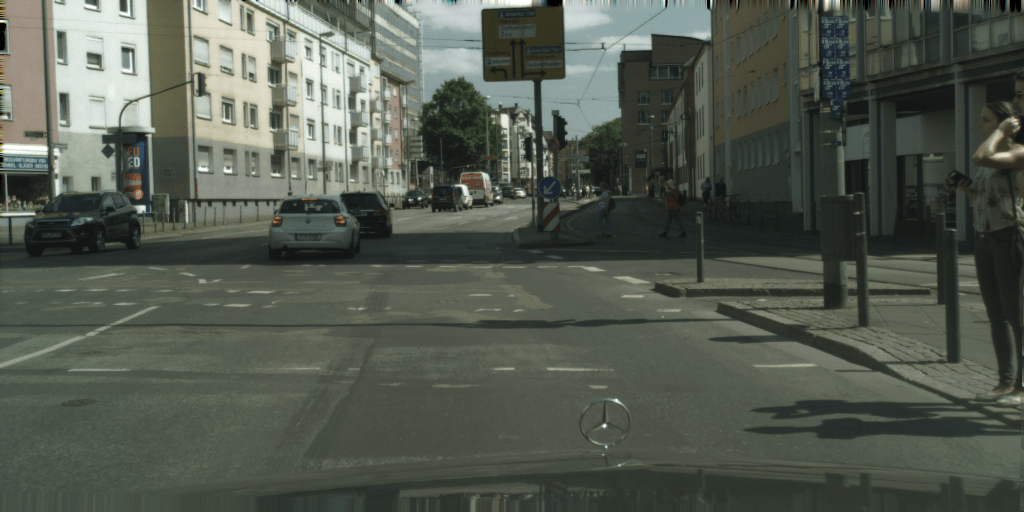}
  }
  \subfigure{%
    \includegraphics[width=.18\columnwidth]{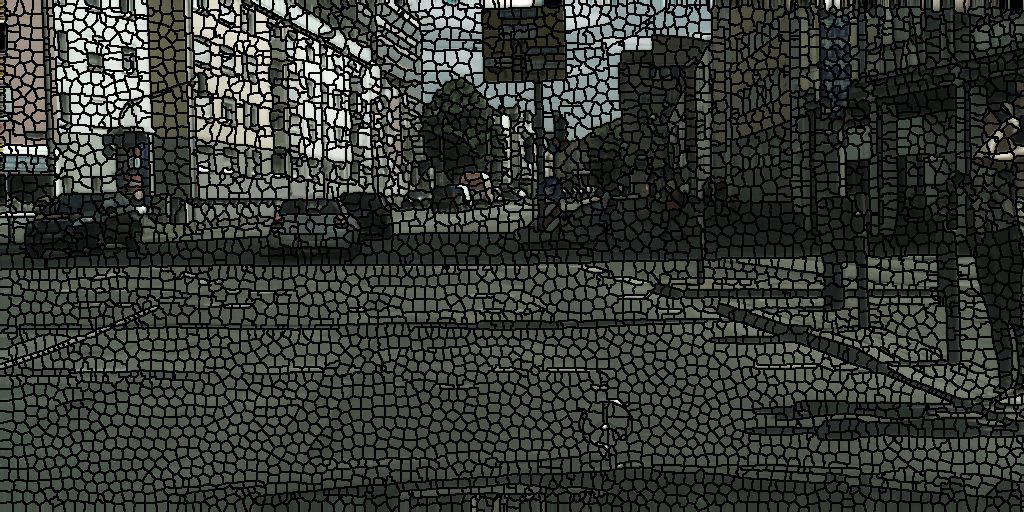}
  }
  \subfigure{%
    \includegraphics[width=.18\columnwidth]{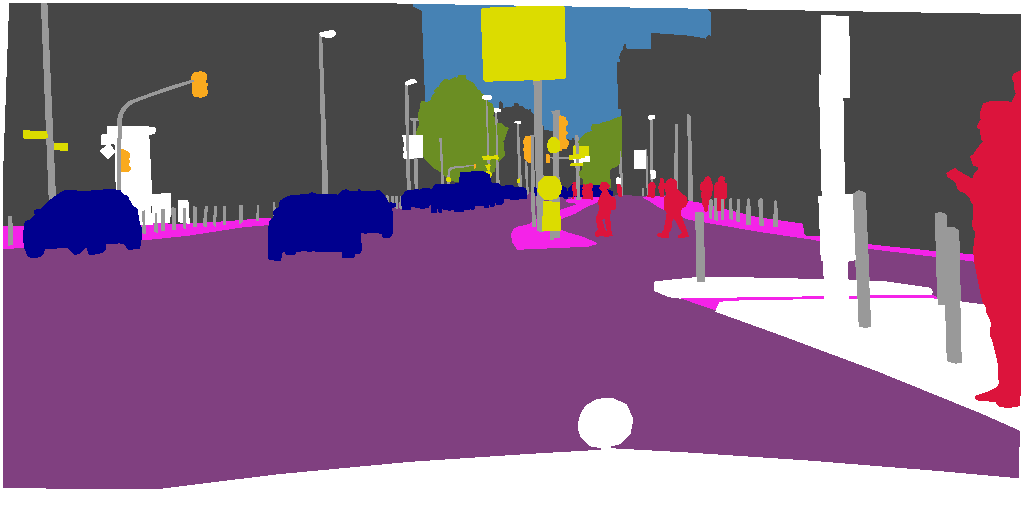}
  }
  \subfigure{%
    \includegraphics[width=.18\columnwidth]{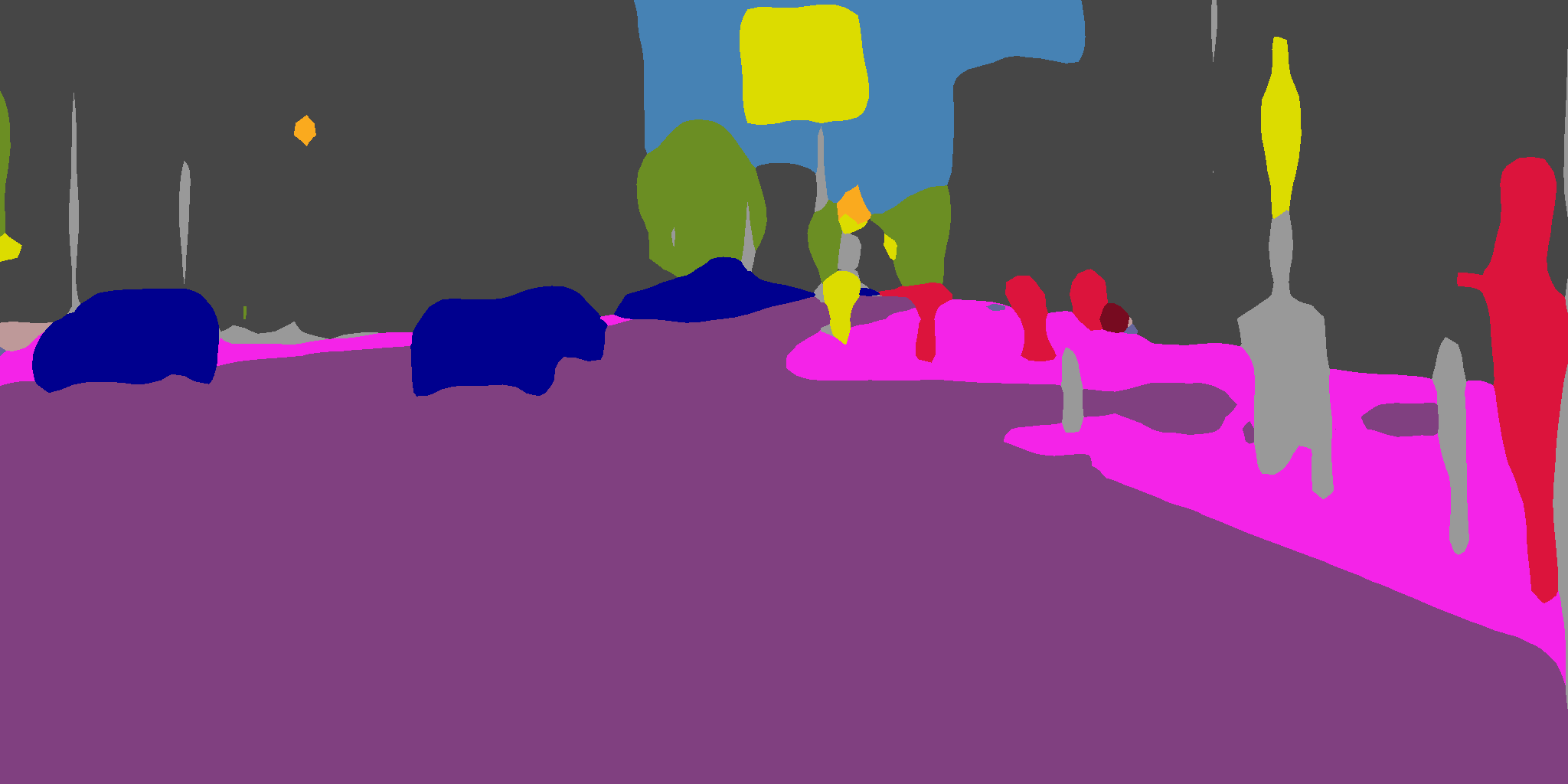}
  }
  \subfigure{%
    \includegraphics[width=.18\columnwidth]{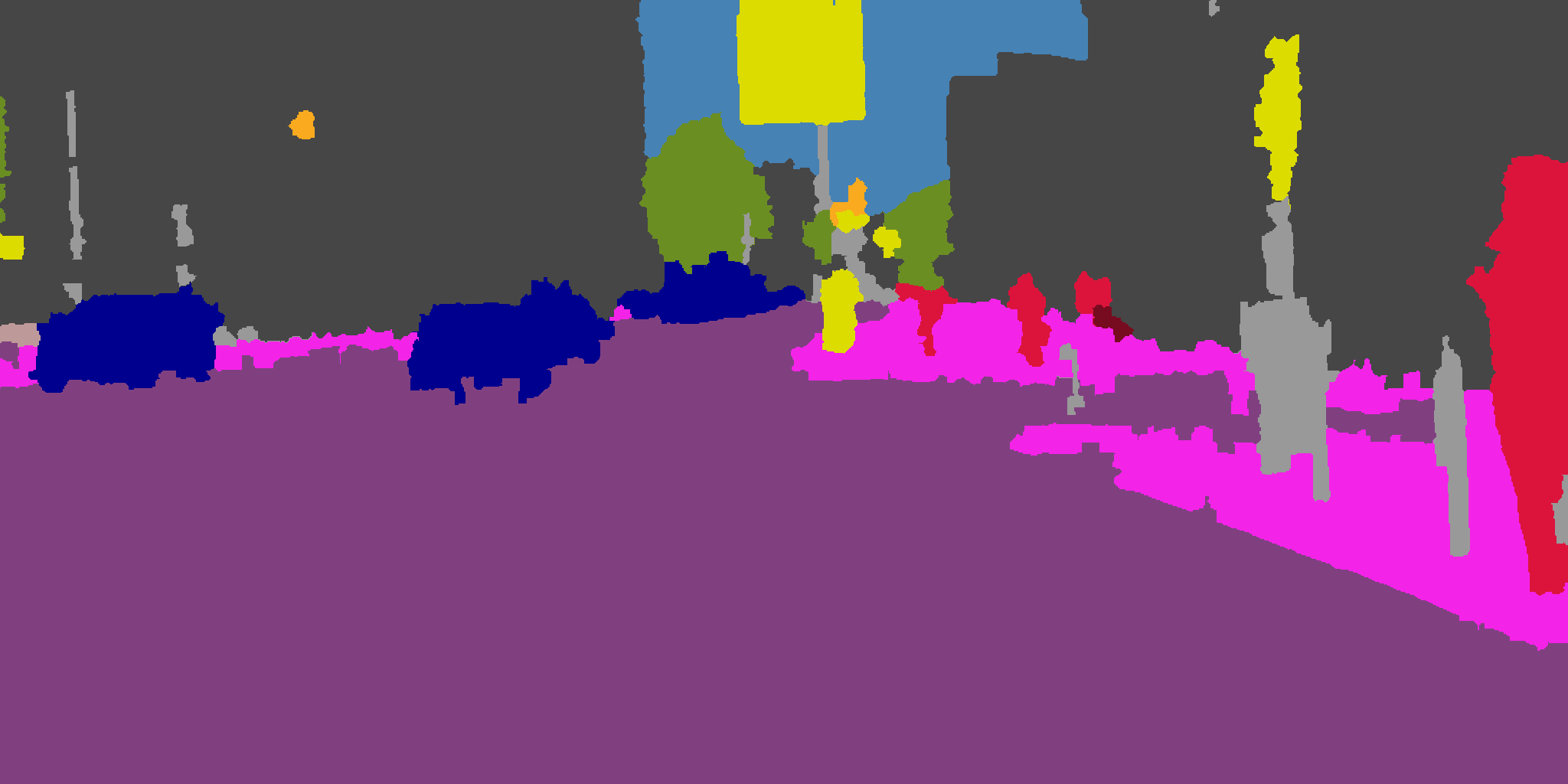}
  }\\[-2ex]

  \subfigure{%
    \includegraphics[width=.18\columnwidth]{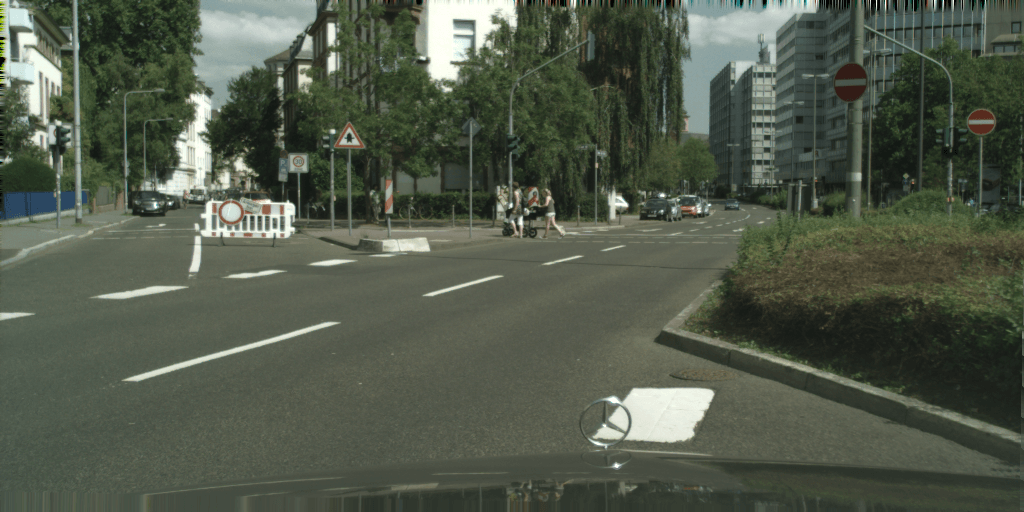}
  }
  \subfigure{%
    \includegraphics[width=.18\columnwidth]{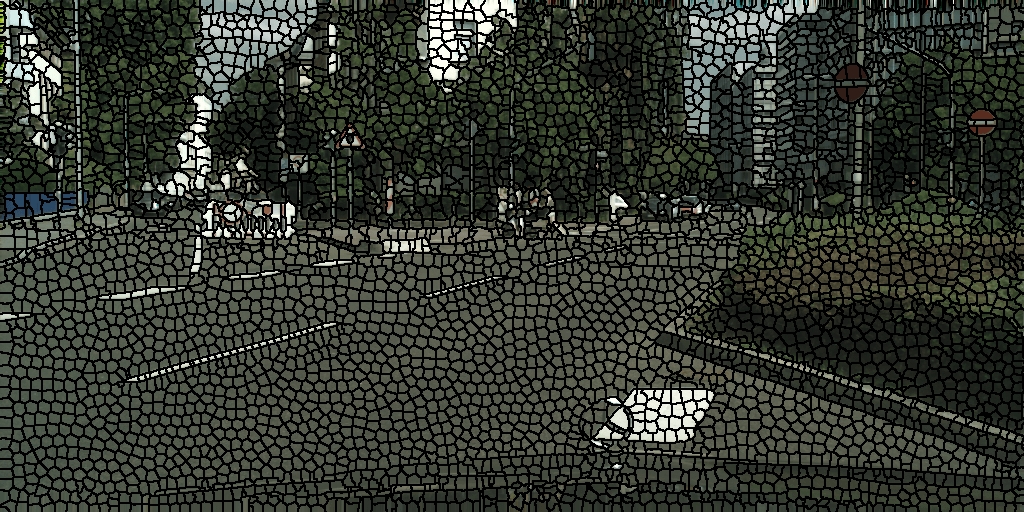}
  }
  \subfigure{%
    \includegraphics[width=.18\columnwidth]{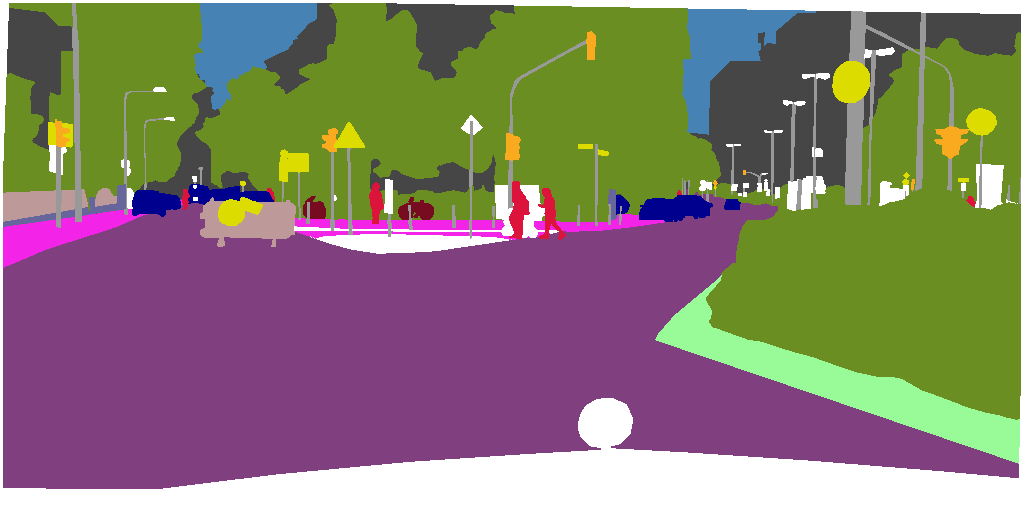}
  }
  \subfigure{%
    \includegraphics[width=.18\columnwidth]{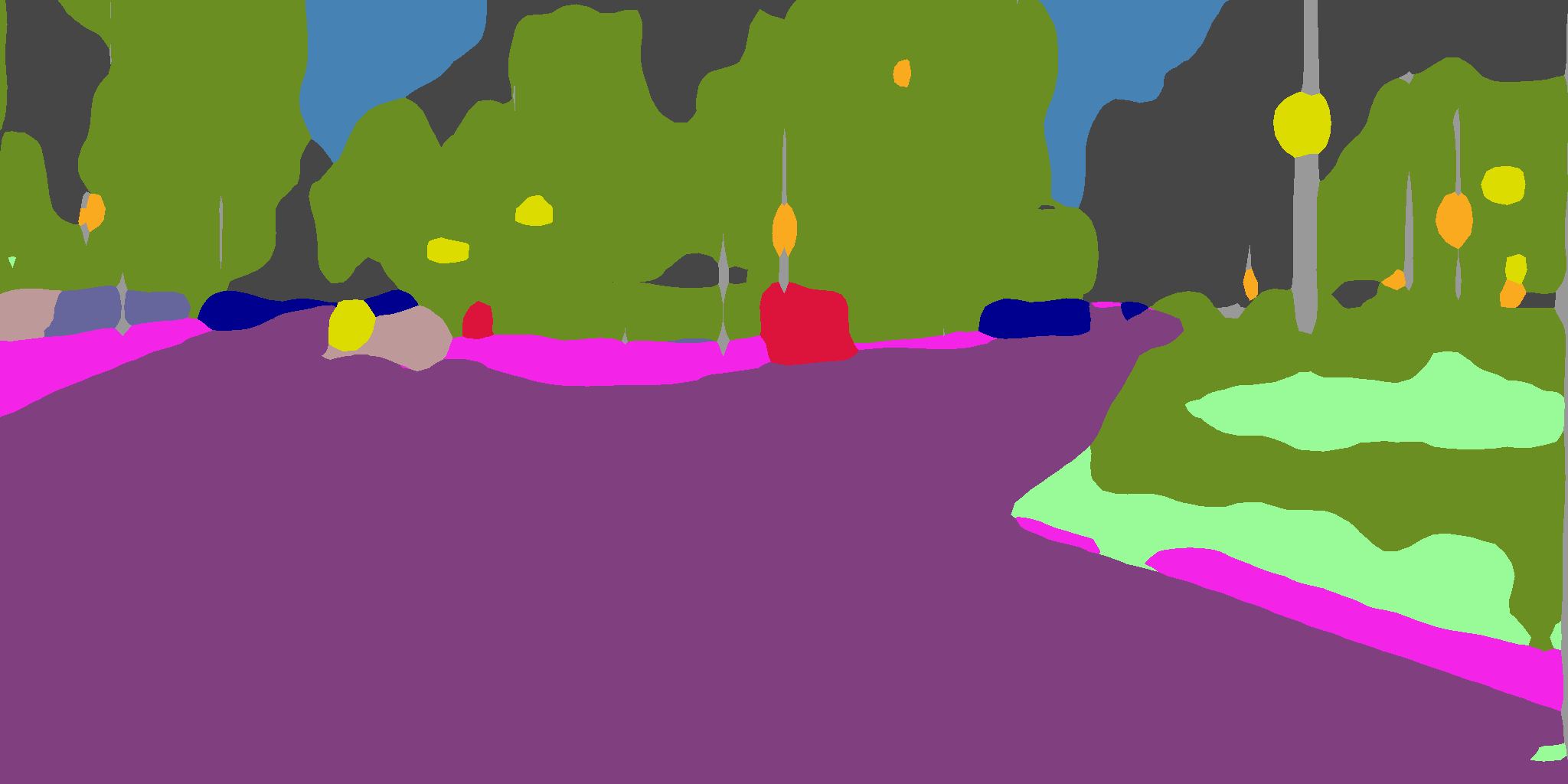}
  }
  \subfigure{%
    \includegraphics[width=.18\columnwidth]{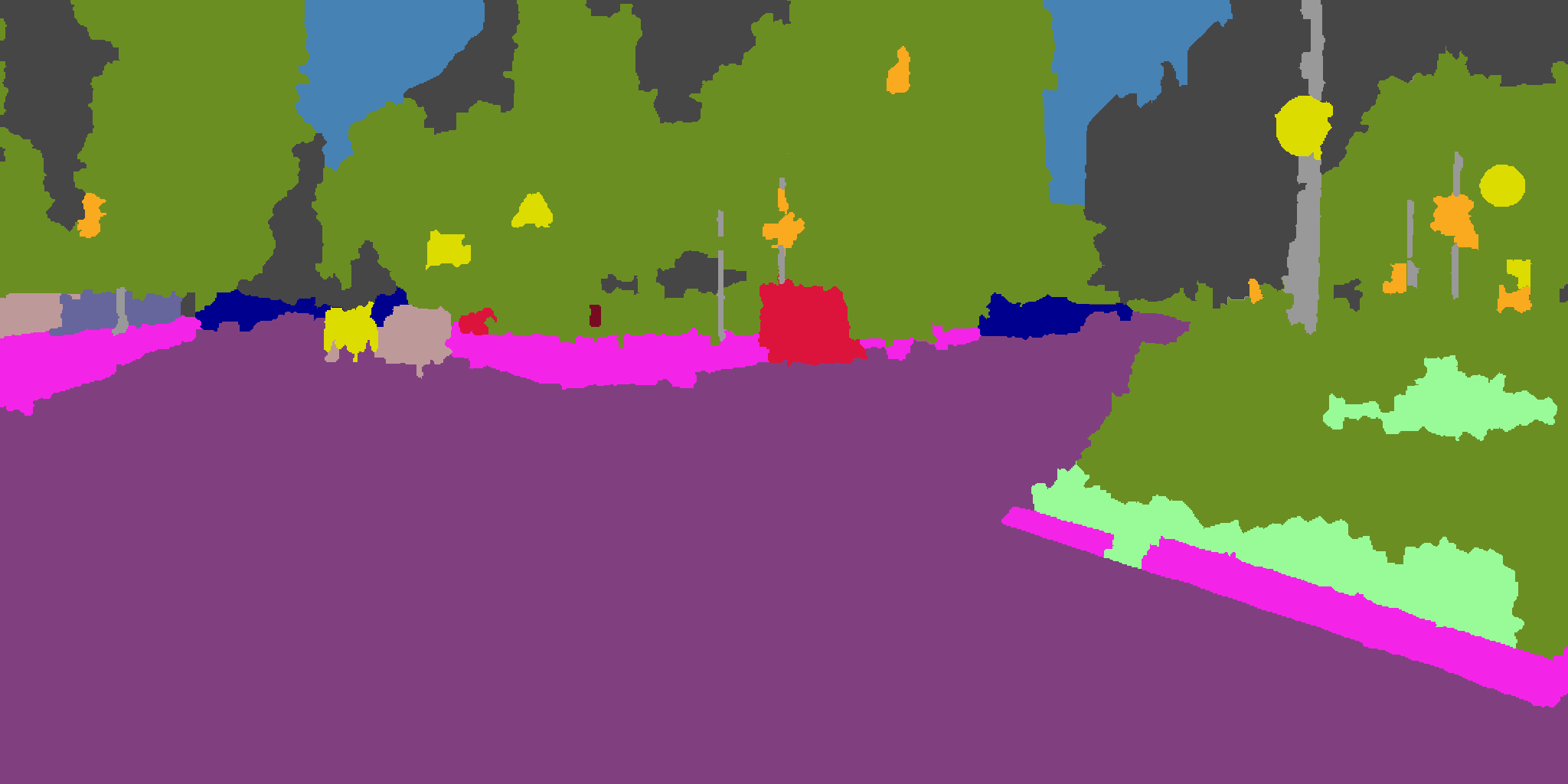}
  }\\[-2ex]

  \subfigure{%
    \includegraphics[width=.18\columnwidth]{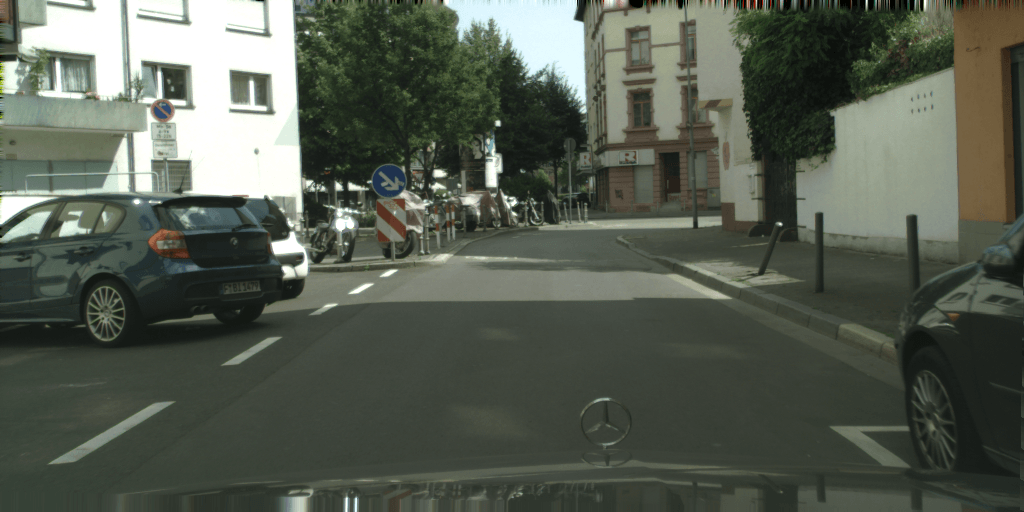}
  }
  \subfigure{%
    \includegraphics[width=.18\columnwidth]{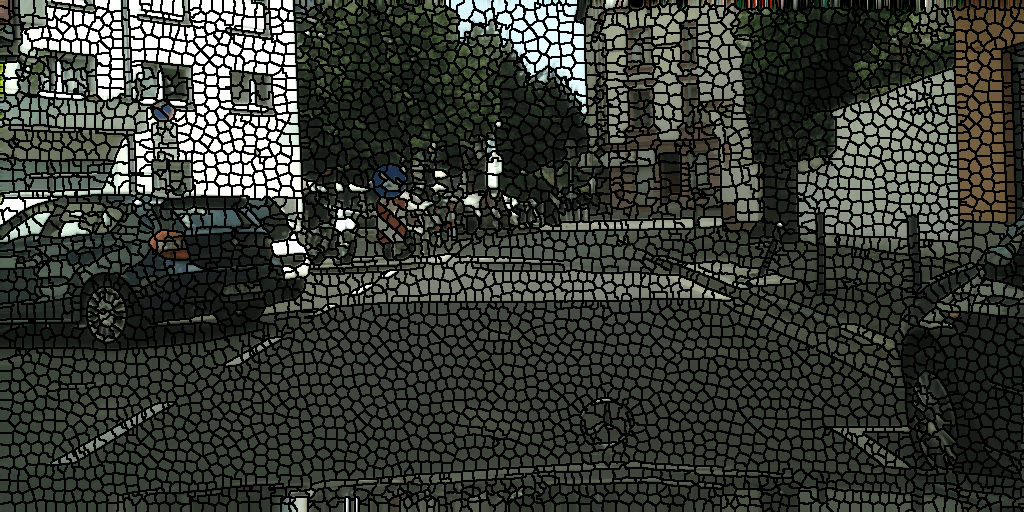}
  }
  \subfigure{%
    \includegraphics[width=.18\columnwidth]{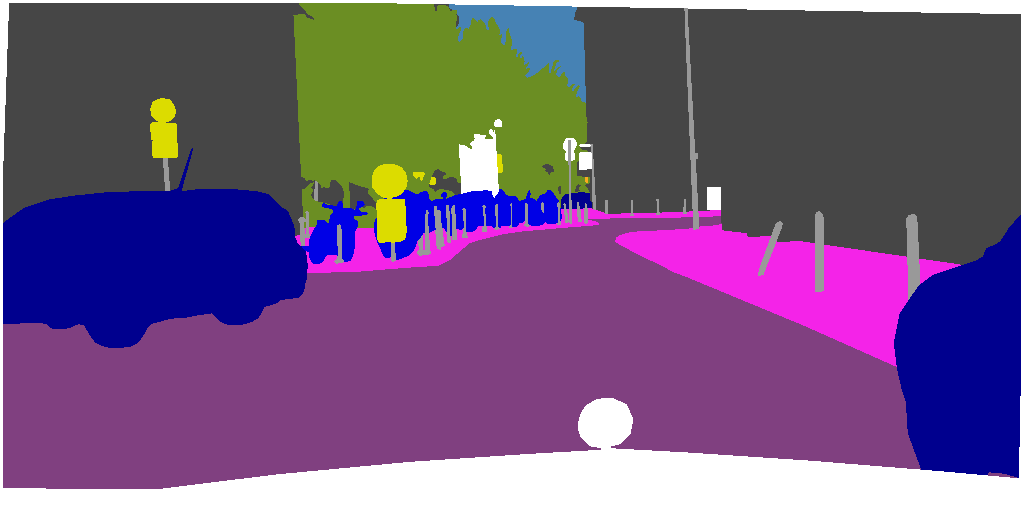}
  }
  \subfigure{%
    \includegraphics[width=.18\columnwidth]{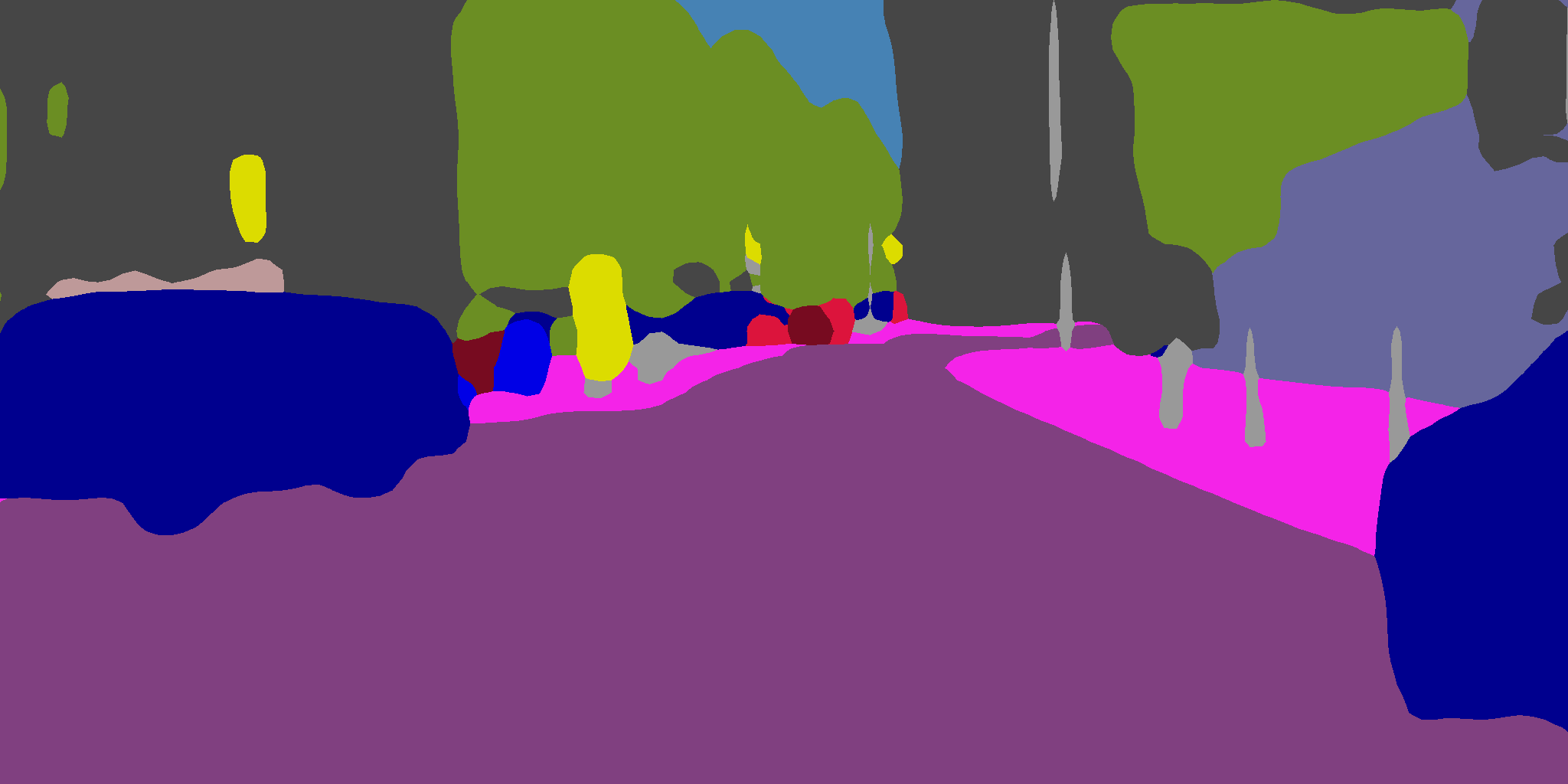}
  }
  \subfigure{%
    \includegraphics[width=.18\columnwidth]{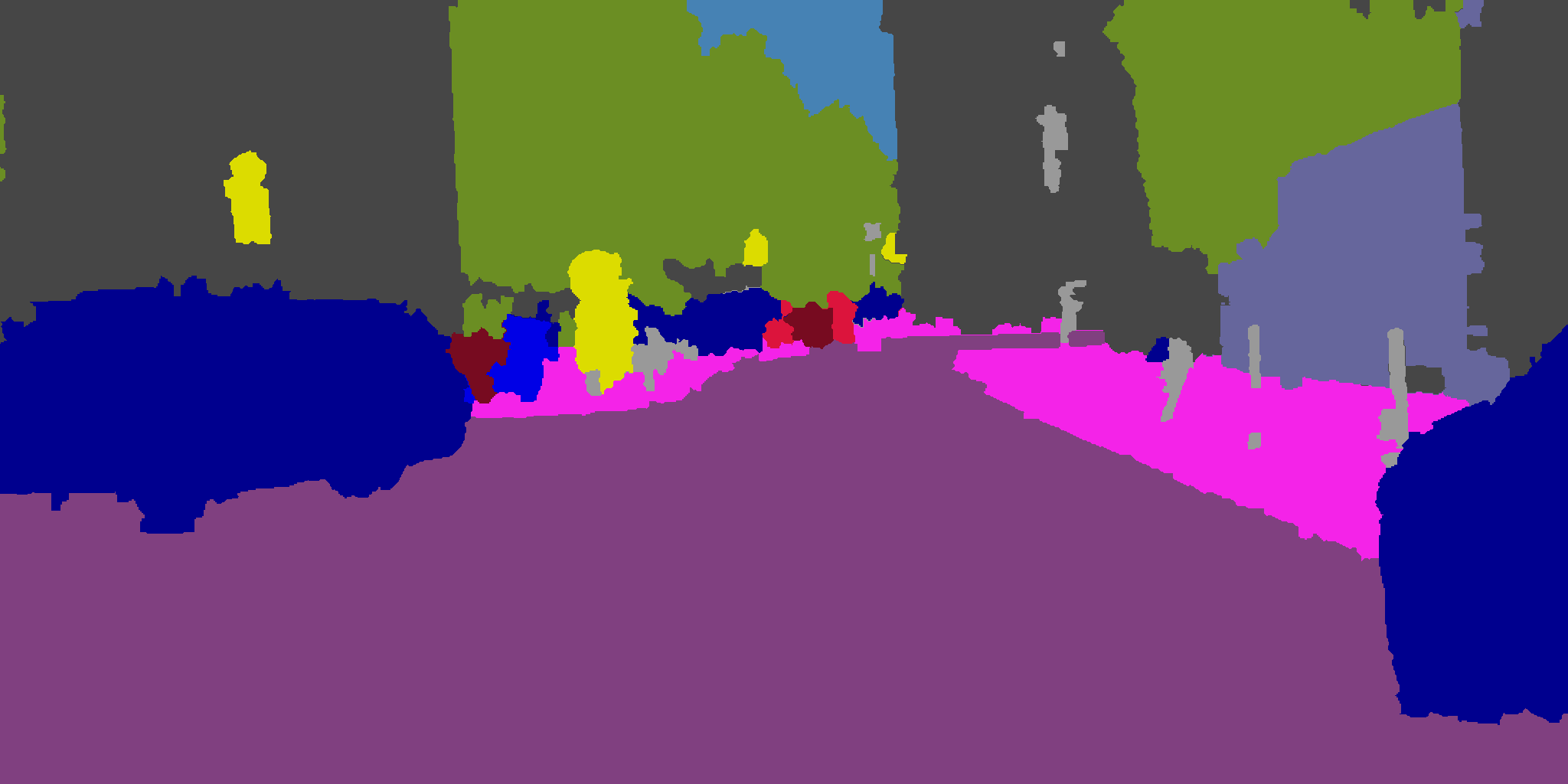}
  }\\[-2ex]

  \subfigure{%
    \includegraphics[width=.18\columnwidth]{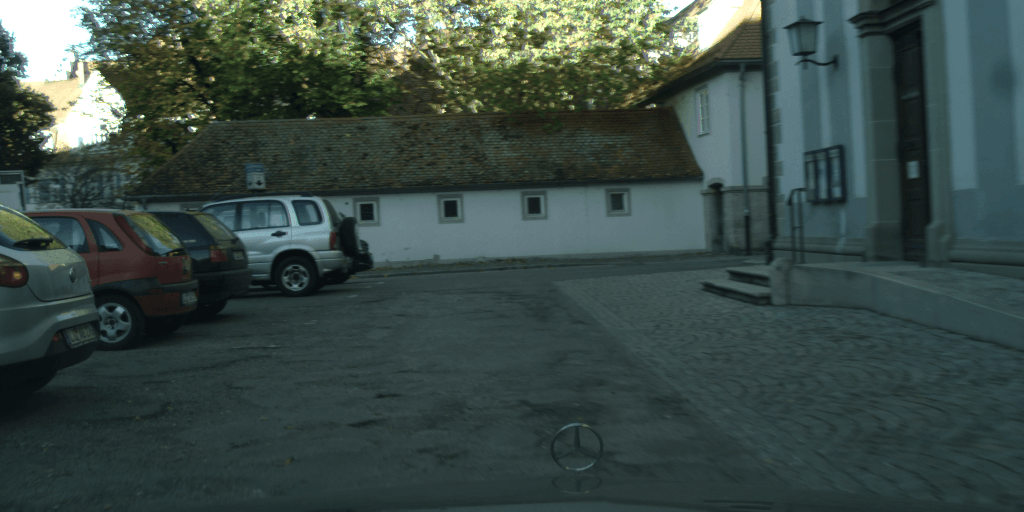}
  }
  \subfigure{%
    \includegraphics[width=.18\columnwidth]{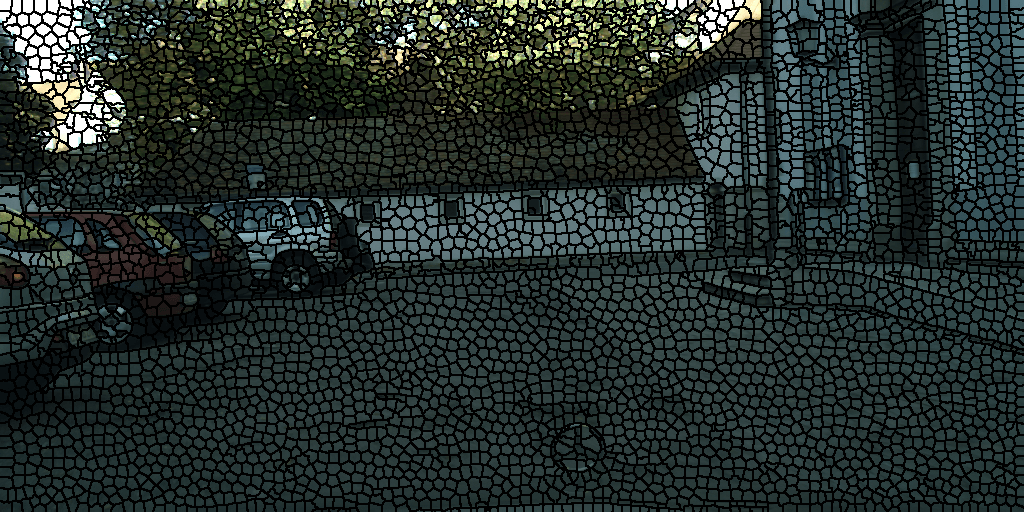}
  }
  \subfigure{%
    \includegraphics[width=.18\columnwidth]{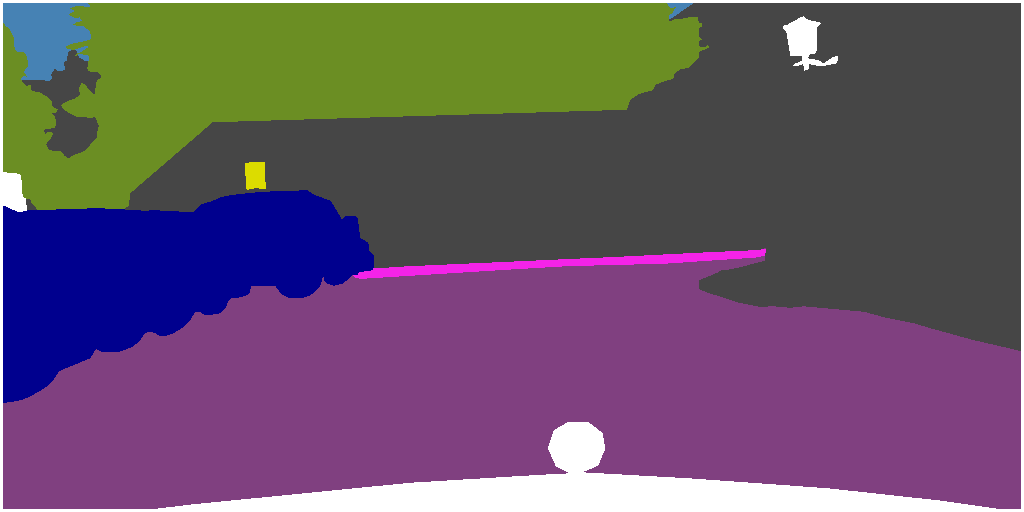}
  }
  \subfigure{%
    \includegraphics[width=.18\columnwidth]{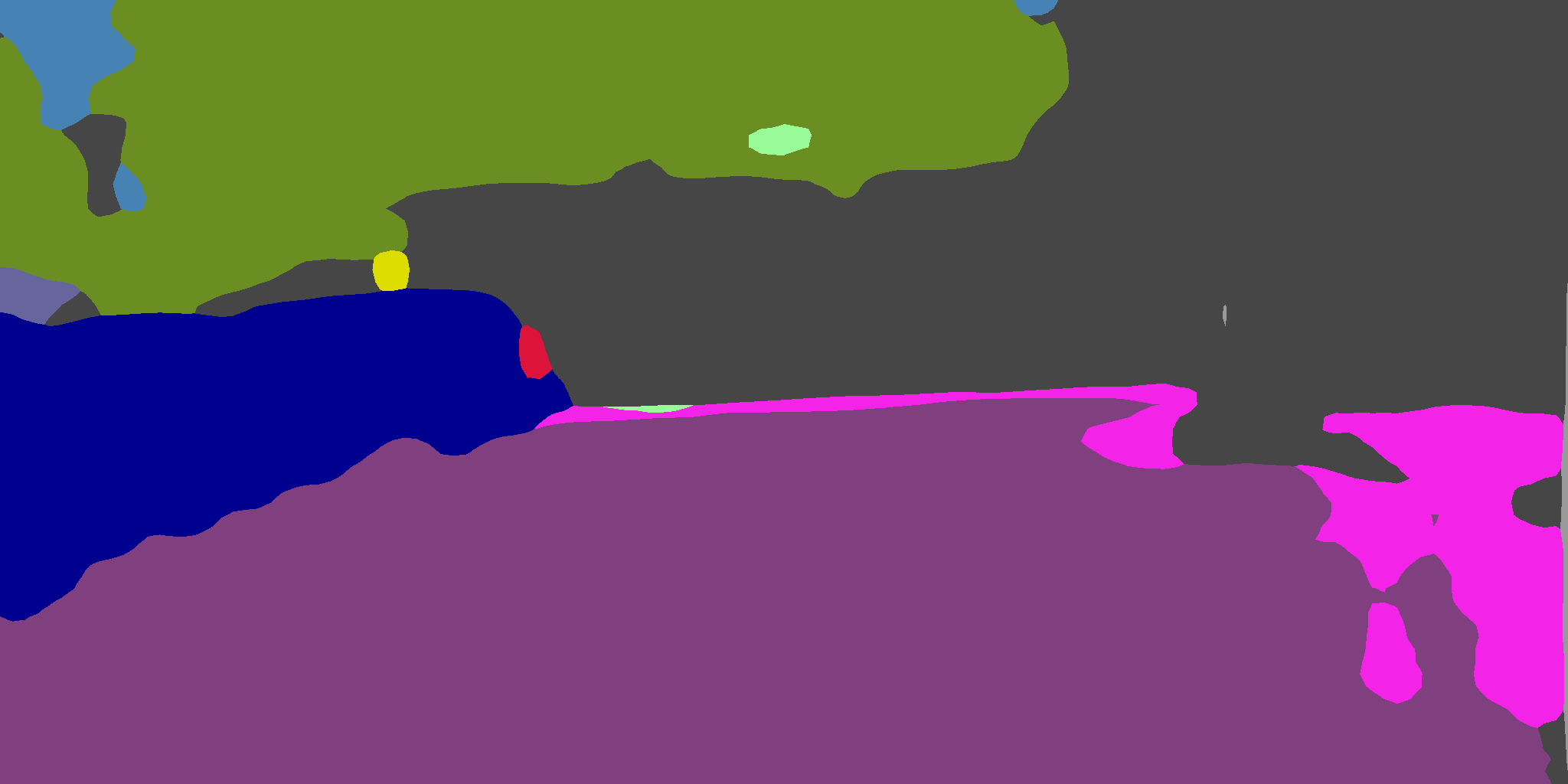}
  }
  \subfigure{%
    \includegraphics[width=.18\columnwidth]{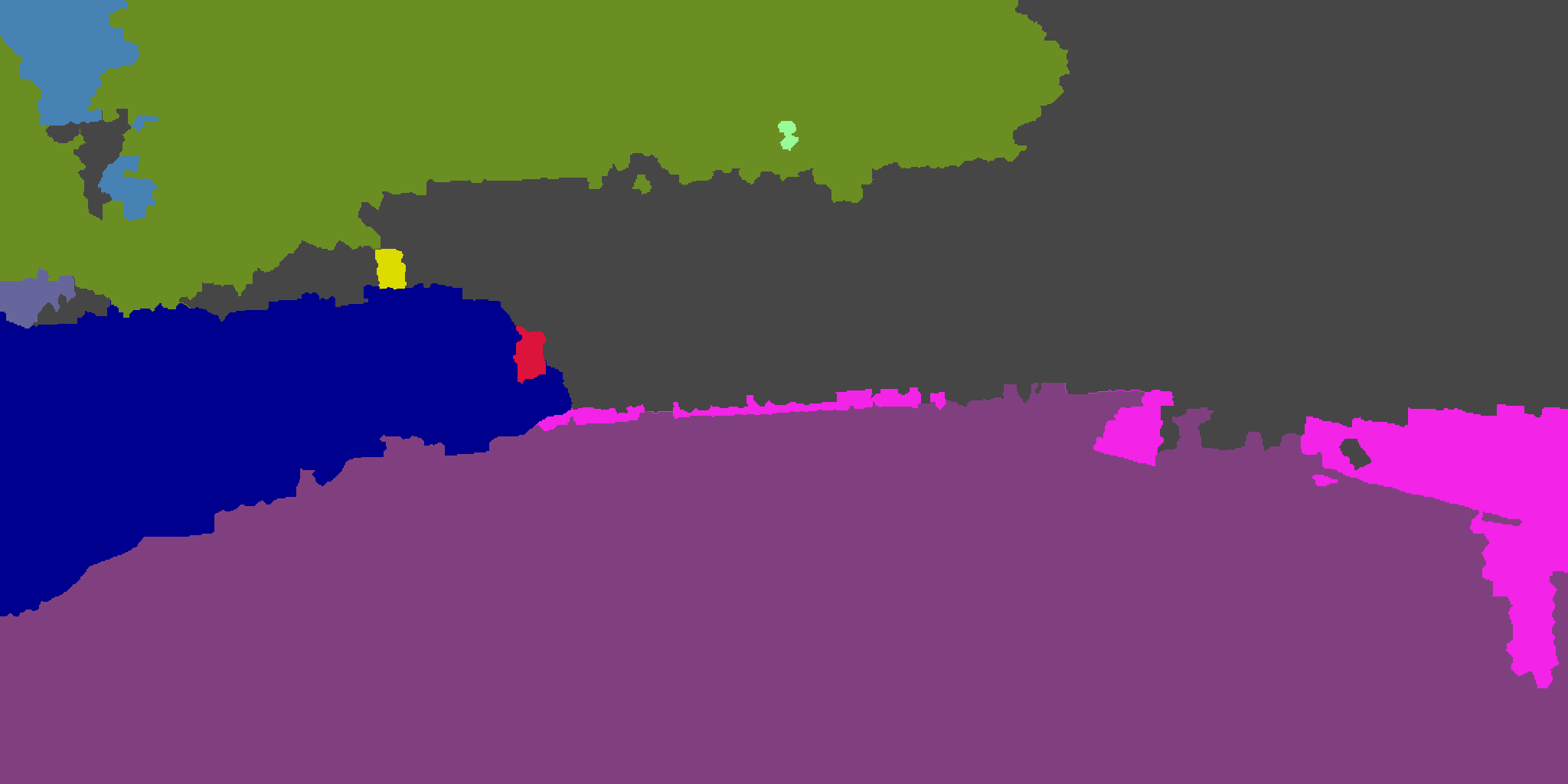}
  }\\[-2ex]

  \subfigure{%
    \includegraphics[width=.18\columnwidth]{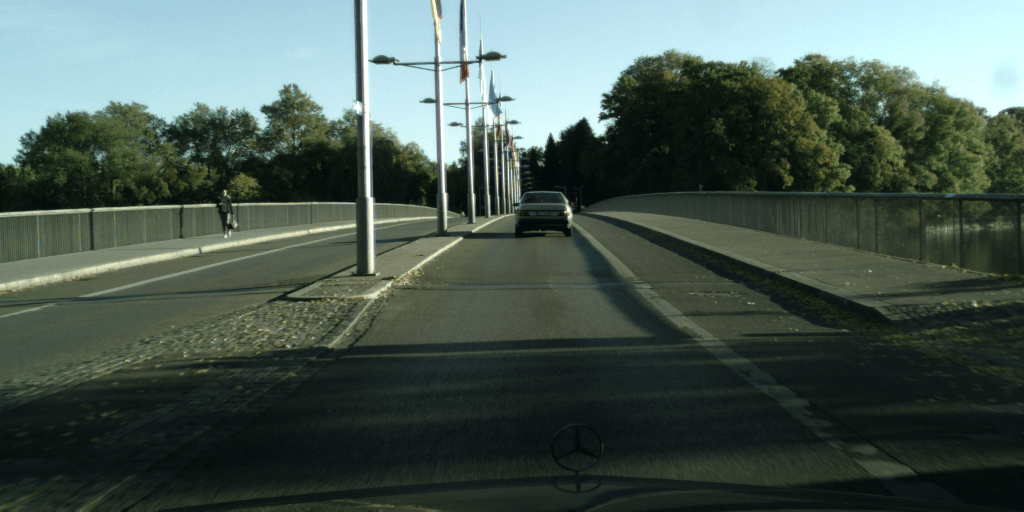}
  }
  \subfigure{%
    \includegraphics[width=.18\columnwidth]{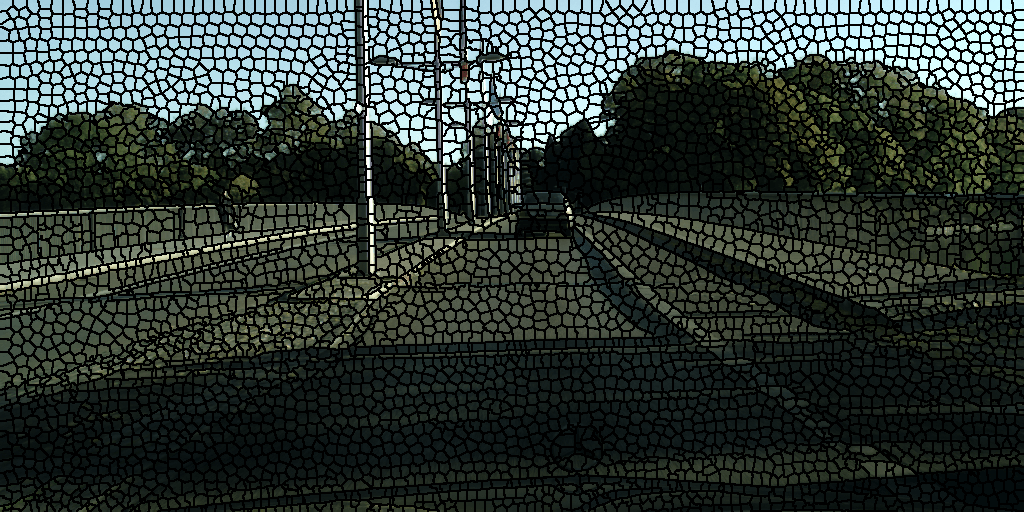}
  }
  \subfigure{%
    \includegraphics[width=.18\columnwidth]{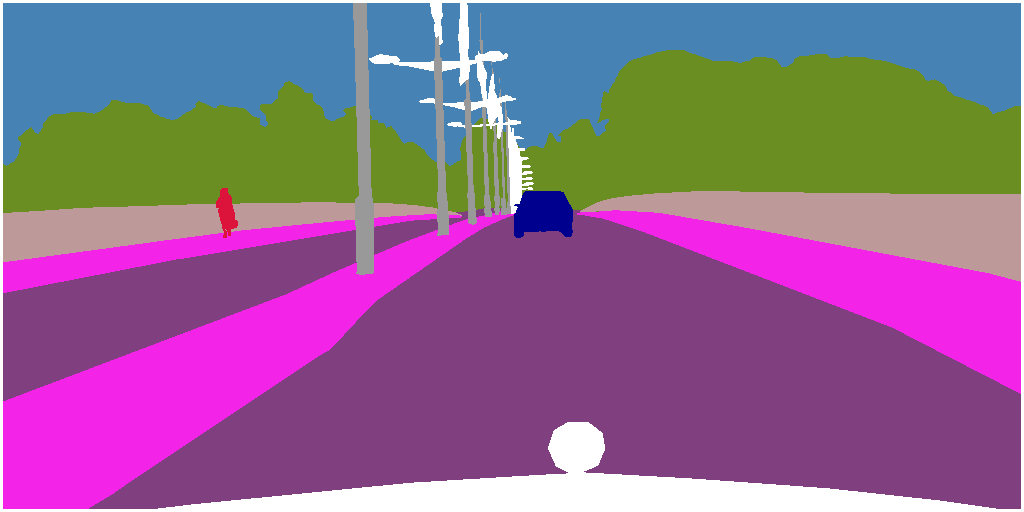}
  }
  \subfigure{%
    \includegraphics[width=.18\columnwidth]{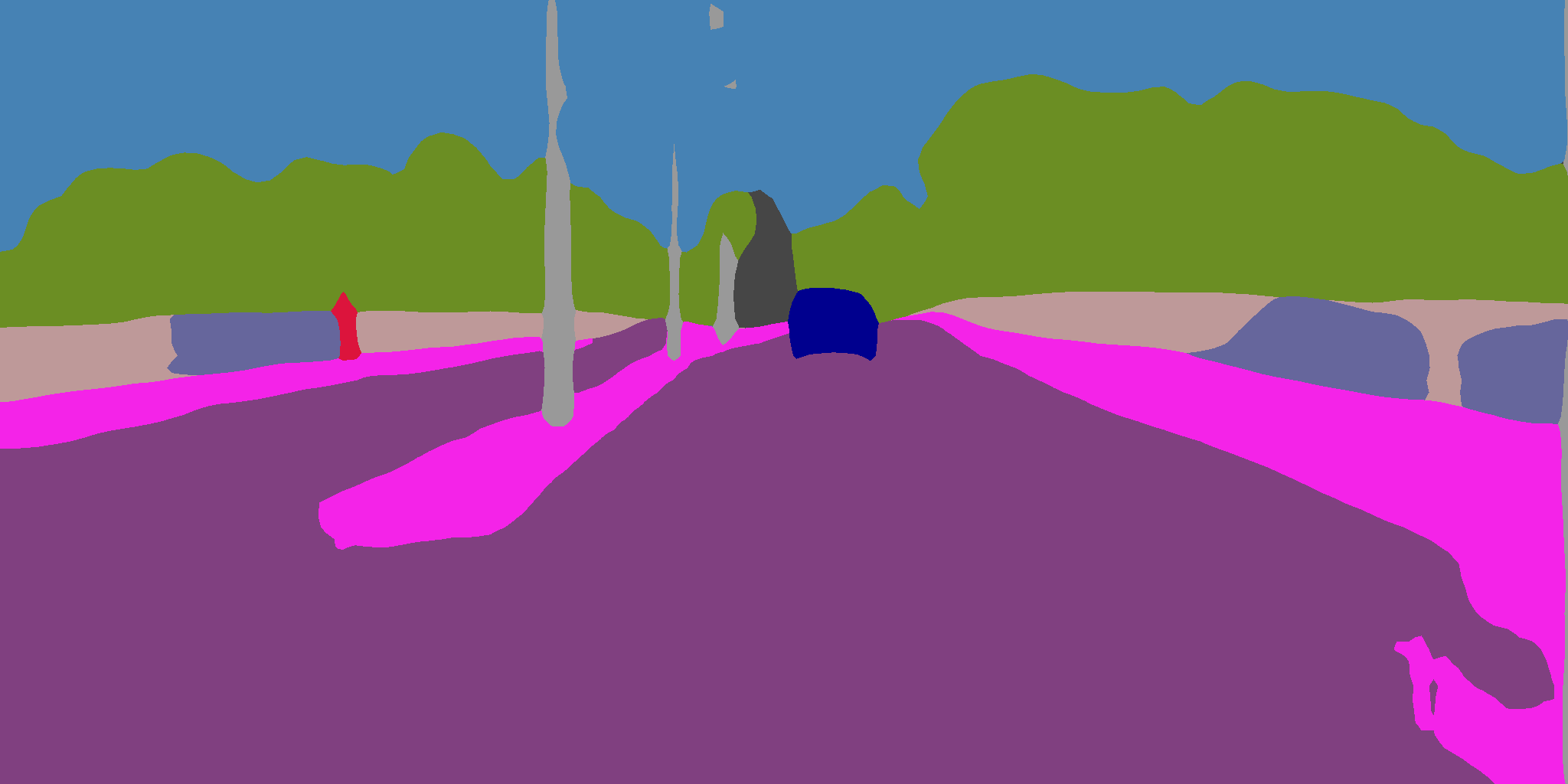}
  }
  \subfigure{%
    \includegraphics[width=.18\columnwidth]{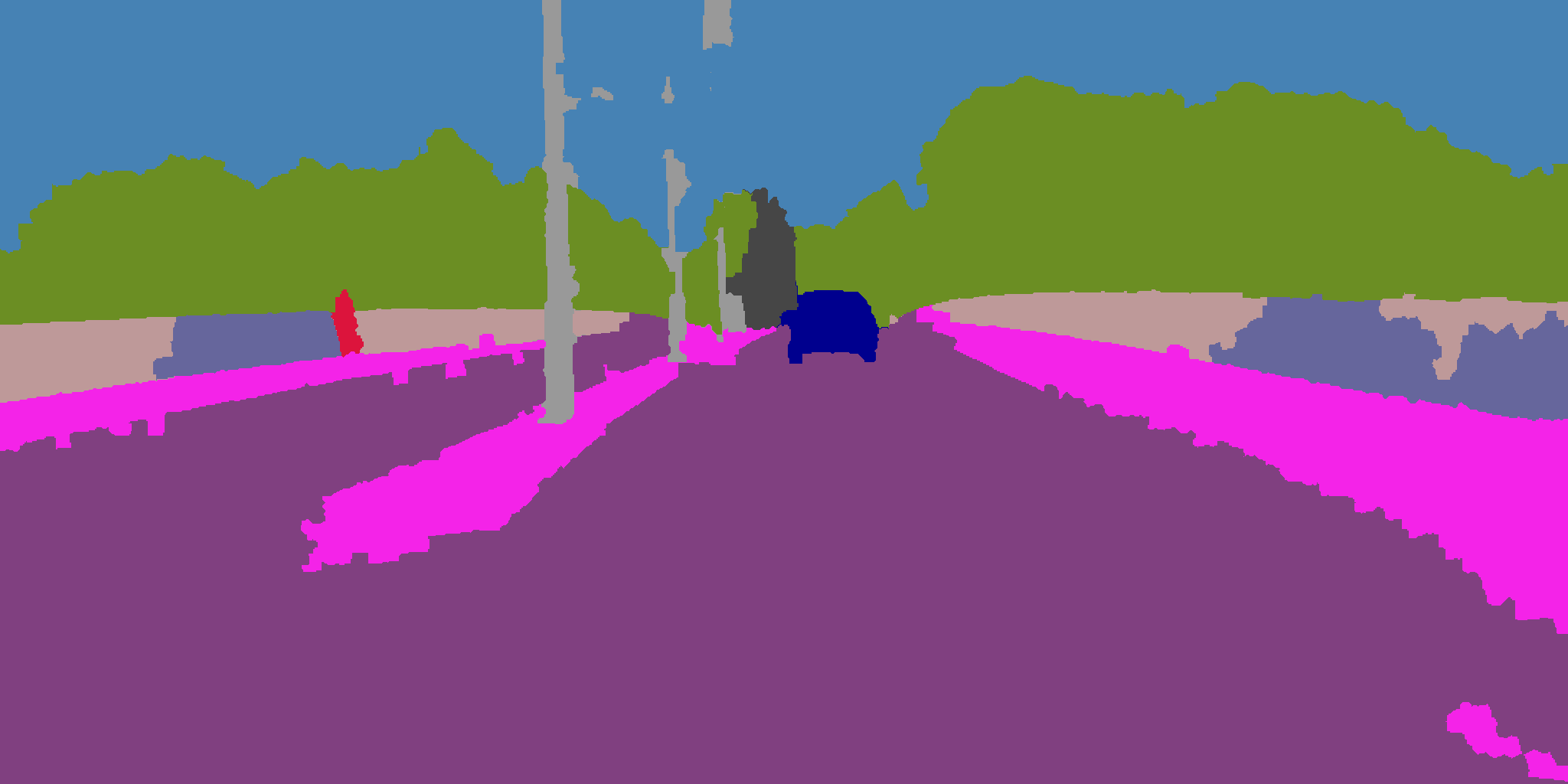}
  }\\[-2ex]

  \subfigure{%
    \includegraphics[width=.18\columnwidth]{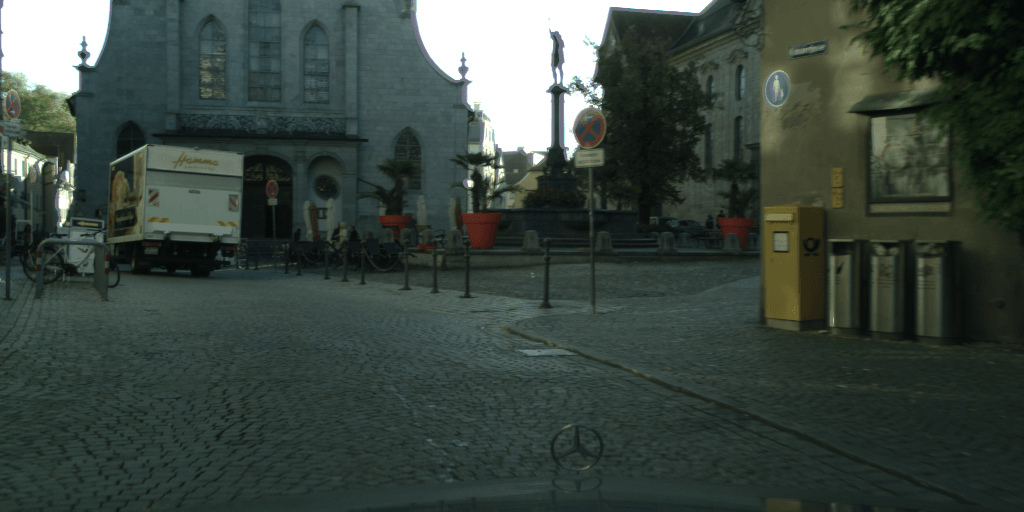}
  }
  \subfigure{%
    \includegraphics[width=.18\columnwidth]{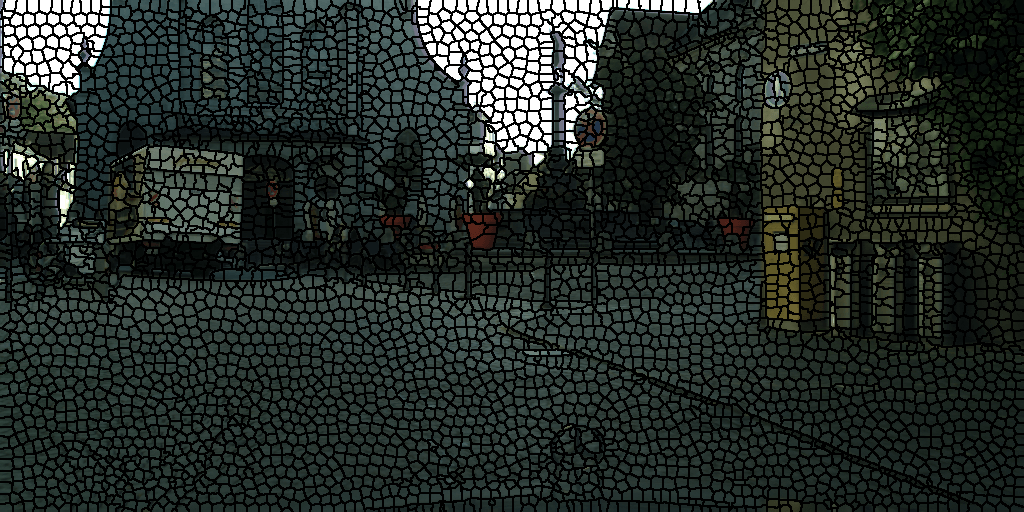}
  }
  \subfigure{%
    \includegraphics[width=.18\columnwidth]{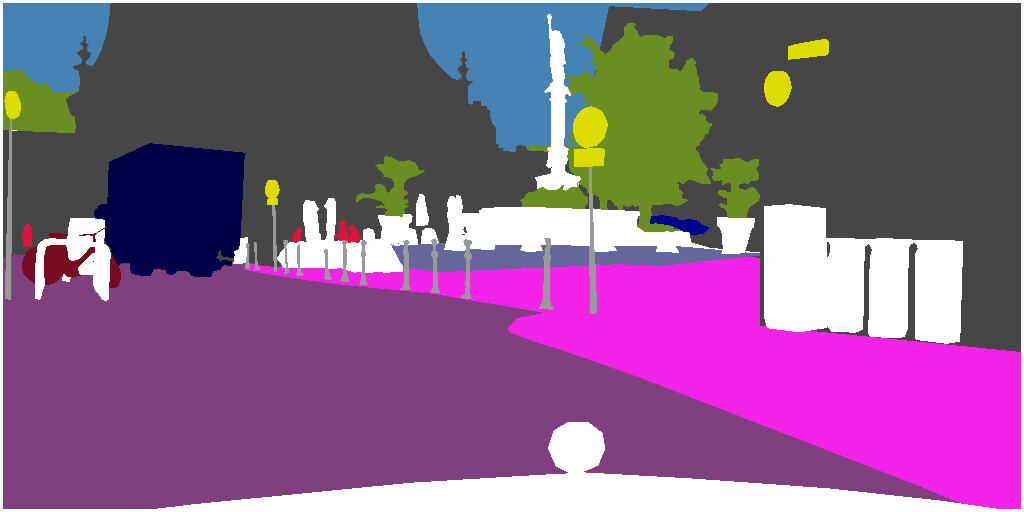}
  }
  \subfigure{%
    \includegraphics[width=.18\columnwidth]{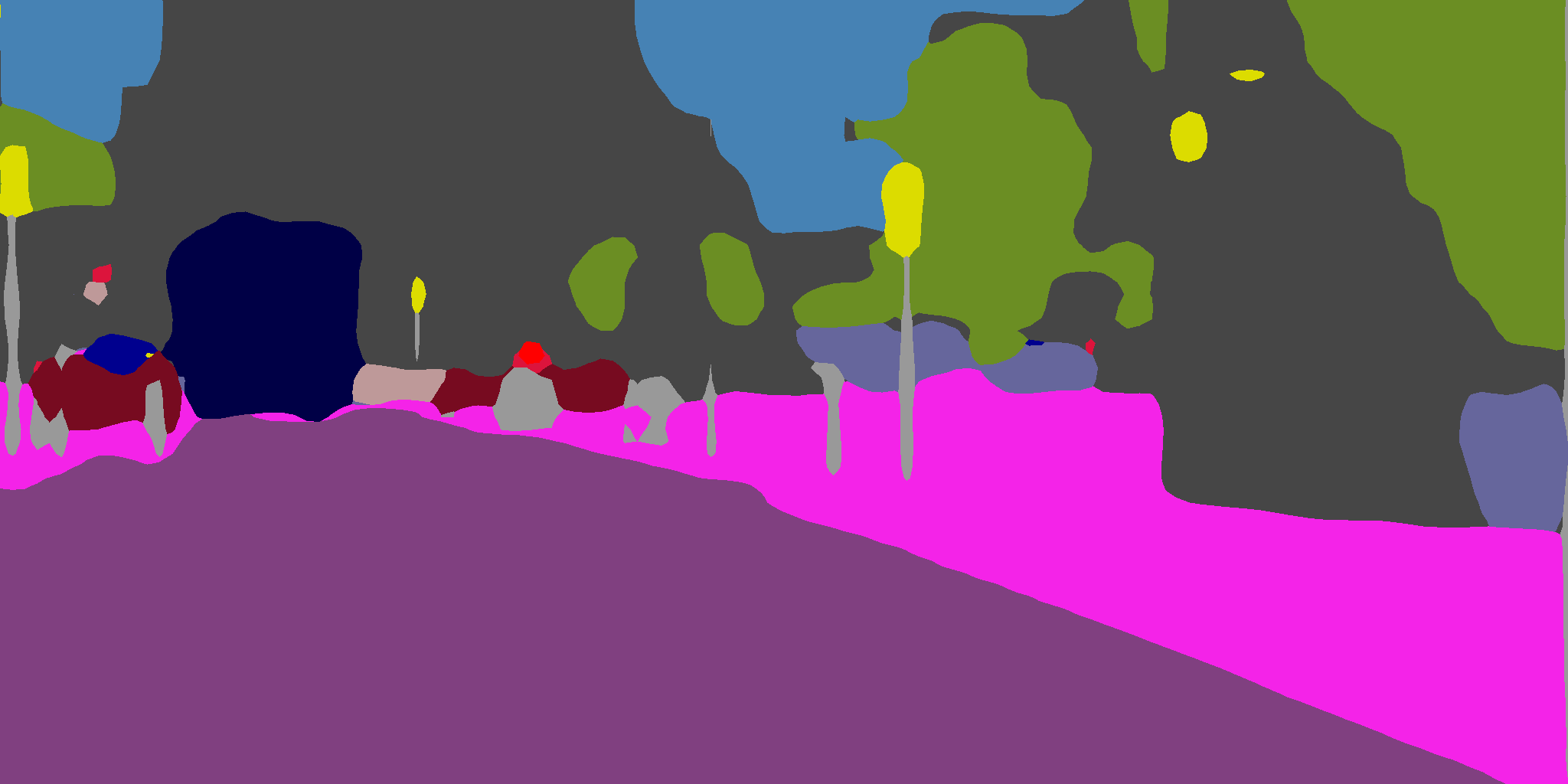}
  }
  \subfigure{%
    \includegraphics[width=.18\columnwidth]{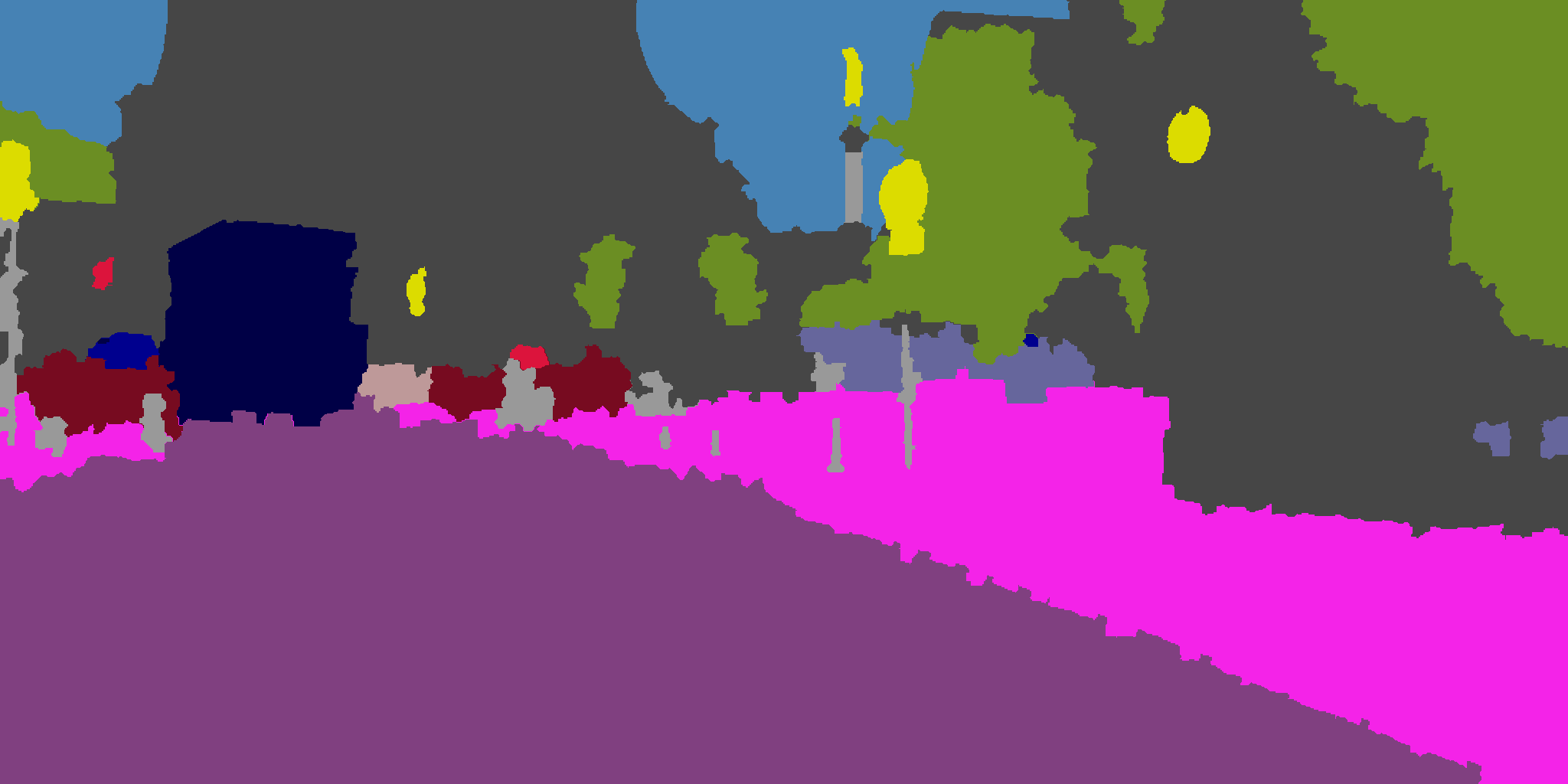}
  }\\[-2ex]

  \setcounter{subfigure}{0}
  \subfigure[\scriptsize Input]{%
    \includegraphics[width=.18\columnwidth]{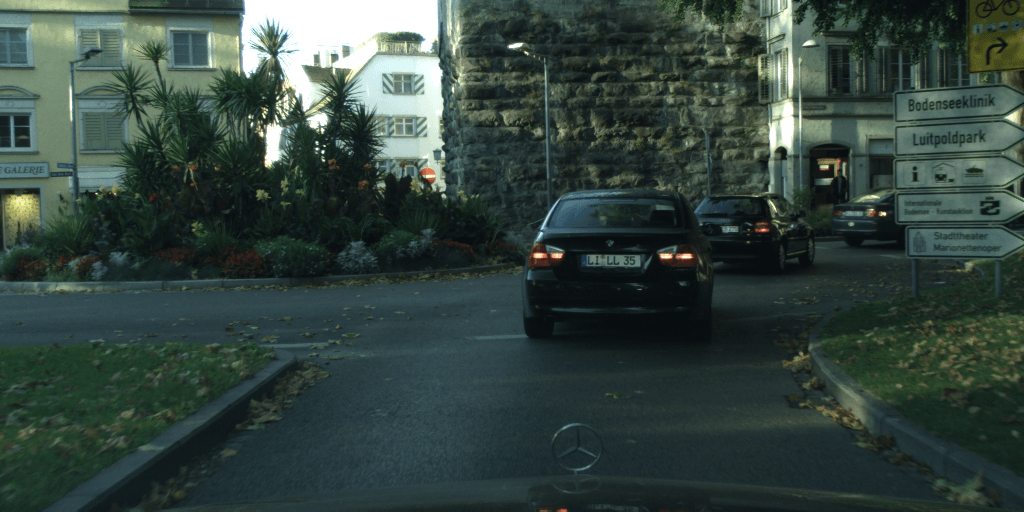}
  }
  \subfigure[\scriptsize Superpixels]{%
    \includegraphics[width=.18\columnwidth]{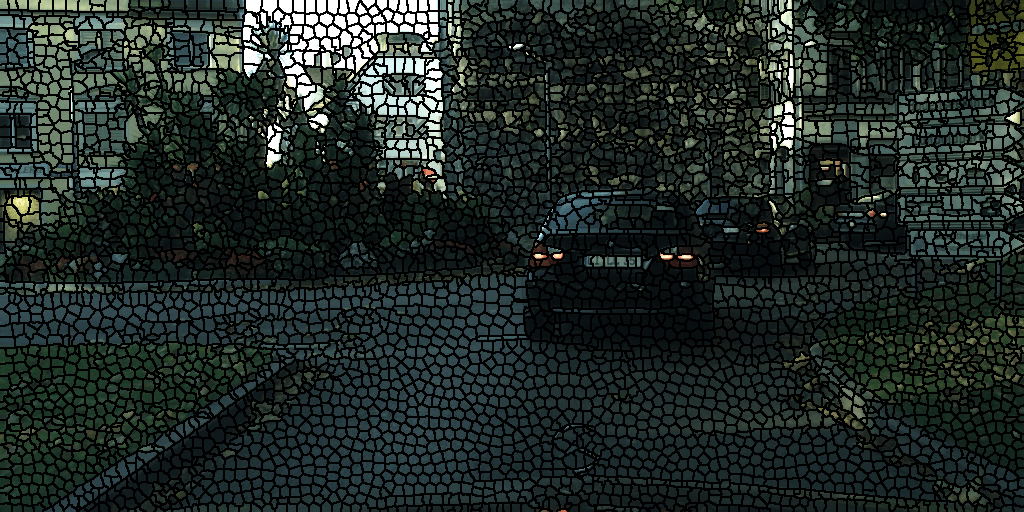}
  }
  \subfigure[\scriptsize GT]{%
    \includegraphics[width=.18\columnwidth]{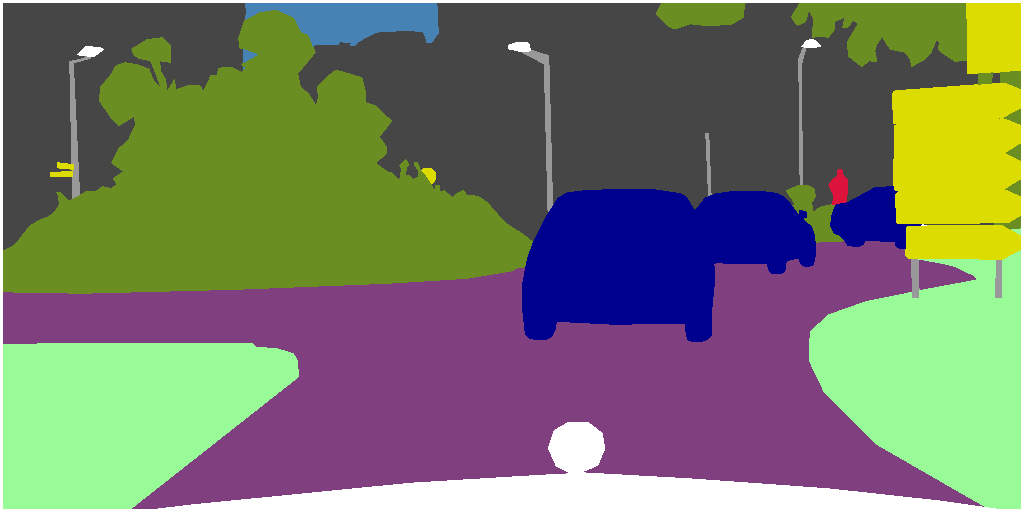}
  }
  \subfigure[\scriptsize Deeplab]{%
    \includegraphics[width=.18\columnwidth]{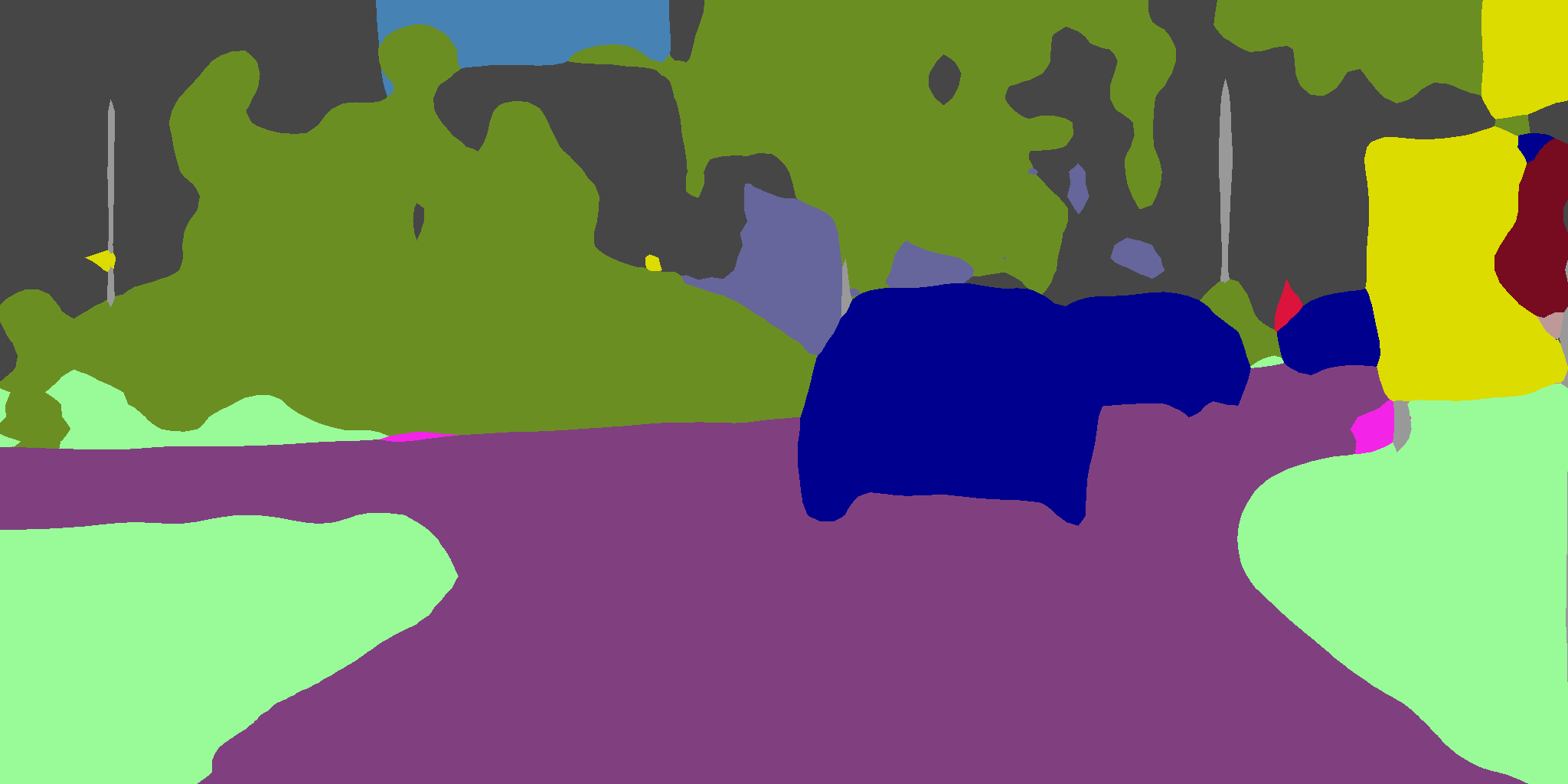}
  }
  \subfigure[\scriptsize Using BI]{%
    \includegraphics[width=.18\columnwidth]{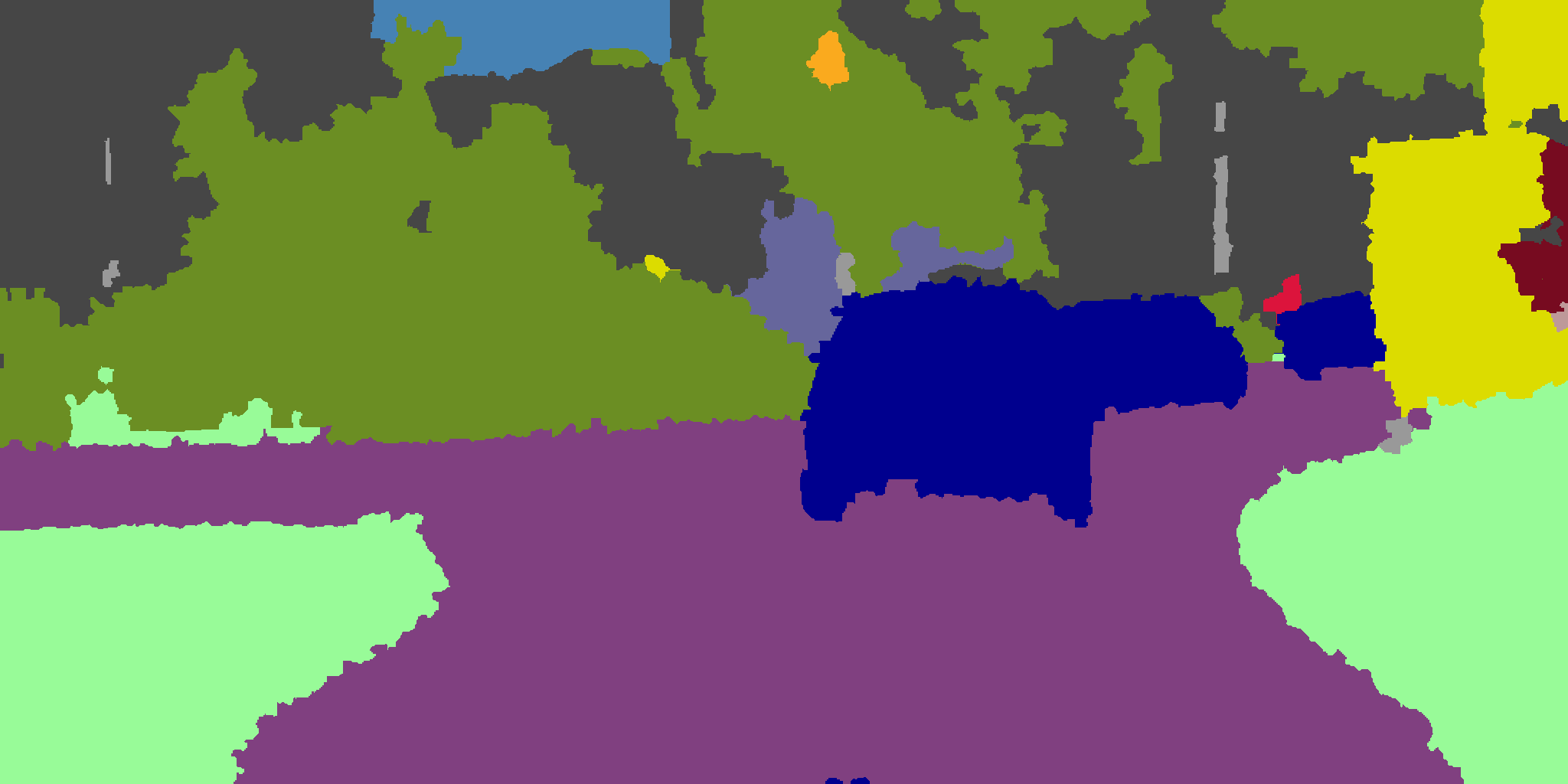}
  }

  \mycaption{Street Scene Segmentation}{Example results of street scene segmentation.
  (d)~depicts the DeepLab results, (e)~result obtained by adding bilateral inception (BI) modules (\bi{6}{2}+\bi{7}{6}) between \fc~layers.}
\label{fig:street_visuals-app}
\end{figure*}



\setlength{\bibsep}{0pt plus 0.3ex}
\bibliographystyle{abbrv}
\bibliography{refs.bib}
\end{document}